\documentclass{article}

\usepackage{microtype}
\usepackage{graphicx}
\graphicspath{{logs/train/runs/}{fig/}{logs/bnn_additional_experiments/}}
\usepackage{subcaption}
\usepackage{booktabs}
\usepackage{csquotes}
\usepackage{enumitem}
\usepackage{multirow}
\usepackage{multicol}
\usepackage[export]{adjustbox}
\usepackage{longtable}
\usepackage{outlines}
\usepackage{rotating}
\usepackage{array}
\newcolumntype{C}[1]{>{\centering\arraybackslash}p{#1}}
\usepackage{mathcomp}
\usepackage{caption}
\usepackage[most]{tcolorbox}%
\usepackage{subfloat}

\usepackage{hyperref}

\renewcommand{\arraystretch}{0.2}

\usepackage[accepted]{icml2024}

\usepackage{amsmath}
\usepackage{amssymb}
\usepackage{mathtools}
\usepackage{amsthm}
\usepackage{physics}
\usepackage{pifont}
\usepackage{makecell}

\usepackage[capitalize,noabbrev]{cleveref}

\theoremstyle{plain}

\theoremstyle{definition}

\theoremstyle{remark}

\usepackage[textsize=tiny]{todonotes}

\icmltitlerunning{The Importance of Architecture Choice in Deep Learning for Climate Applications}

\newcommand{\Fone}{\(\mathcal{F}_1\)}
\newcommand{\Ftwo}{\(\mathcal{F}_2\)}
\newcommand{\Fthree}{\(\mathcal{F}_3\)}
\newcommand{\Ffour}{\(\mathcal{F}_4\)}
\newcommand{\Ffive}{\(\mathcal{F}_5\)}
\newcommand{\Fsix}{\(\mathcal{F}_6\)}

\begin{document}

\twocolumn[
	\icmltitle{The Importance of Architecture Choice in Deep Learning for Climate Applications}



	\icmlsetsymbol{equal}{*}

	\begin{icmlauthorlist}
		\icmlauthor{Simon Dr{\"a}ger}{ucd}
		\icmlauthor{Maike Sonnewald}{ucd}
	\end{icmlauthorlist}

	\icmlaffiliation{ucd}{Department of Computer Science, University of California, Davis, USA}

	\icmlcorrespondingauthor{Simon Dr{\"a}ger}{\texttt{sdraeger@ucdavis.edu}}

	\icmlkeywords{Machine Learning, Deep Learning, Climate Science, Physical Oceanography, Dynamical Systems}

	\vskip 0.3in
]

\printAffiliationsAndNotice{}

\begin{abstract}%
	Machine Learning has become a pervasive tool in climate science applications.
	However, current models fail to address nonstationarity induced by anthropogenic
	alterations in greenhouse emissions and do not routinely quantify the uncertainty
	of proposed projections. In this paper, we model the Atlantic Meridional Overturning
	Circulation (AMOC) which is of major importance to climate in Europe and the US East
	Coast by transporting warm water to these regions, and has the potential for abrupt collapse. We can generate arbitrarily
	extreme climate scenarios through arbitrary time scales which we then predict using
	neural networks. Our analysis shows that the AMOC is predictable using neural networks
	under a diverse set of climate scenarios. Further experiments reveal that MLPs and Deep
	Ensembles can learn the physics of the AMOC instead of imitating its progression through
	autocorrelation. With quantified uncertainty, an intriguing pattern of \enquote{spikes}
	before critical points of collapse in the AMOC casts doubt on previous analyses that
	predicted an AMOC collapse within this century. Our results show that Bayesian Neural
	Networks perform poorly compared to more dense architectures and care should be taken
	when applying neural networks to nonstationary scenarios such as climate projections.
	Further, our results highlight that big NN models might have difficulty in modeling
	global Earth System dynamics accurately and be successfully applied in nonstationary
	climate scenarios due to the physics being challenging for neural networks to capture.
\end{abstract}

\begin{figure}[htb]
	\centering
	\includegraphics[scale=0.6]{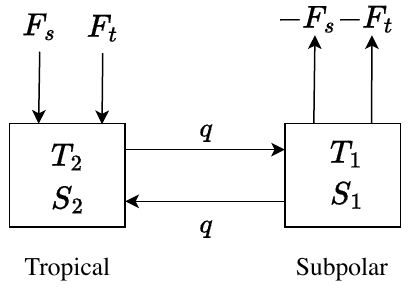}
	\caption{Stommel box model with a fresh water forcing and a temperature forcing.
	Note that this is the \enquote{extended variant} where we obtain the standard variant
	by setting \(F_t \equiv 0\).}%
	\label{fig:stommel_model_Fs_Ft}
\end{figure}

\section{Introduction}

Climate change has a significant impact on our planet and society, with many
unknown variables. One of these variables is the Atlantic Meridional Overturning
Circulation (AMOC) which transports warm water from the tropics to the US East Coast
and Europe. Without the AMOC, these places would be uninhabitable due to ice age
conditions, however, recent climate developments induce significant nonstationarity
into the AMOC system with largely uncertain outcomes. An interesting part of the AMOC
is its \emph{tipping point behavior}~\cite{tziperman2022global,lenton2008tipping}.
A tipping point~\cite{lenton2008tipping,van2016you} is defined as a point in a system
which upon a small perturbation leads to a dramatic change within the system. This
tipping point behavior is especially interesting due to anthropogenic climate change
driving the AMOC in the aforementioned \emph{nonstationary} way, ultimately leading
toward a tipping point~\cite{tziperman2022global}. By nonstationary, we mean for example
a non-constant amplitude of sinusoidal green house emissions (stationary, on the other
hand, would mean the emissions have the same mean and standard deviatino over time).
To deal with the challenge of increasingly nonstationary AMOC (and climate, by extension)
behavior, we wish to take advantage of the predictive abilities that Machine Learning (ML)
methods provide. Here, we present new insight into the extent that different neural network
architectures are able to learn the AMOC's physics, how they work in nonstationary climate
scenarios and how uncertain their predictions are.

\begin{figure}[htb]
	\centering
	\bgroup%
		\newcommand{\htrim}{1.5cm}%
		\newcommand{\bottrim}{2cm}%
		\newcommand{\toptrim}{3.2cm}%
		\includegraphics[scale=0.6,trim={{\htrim} {\bottrim} {\htrim} {\toptrim}},clip]{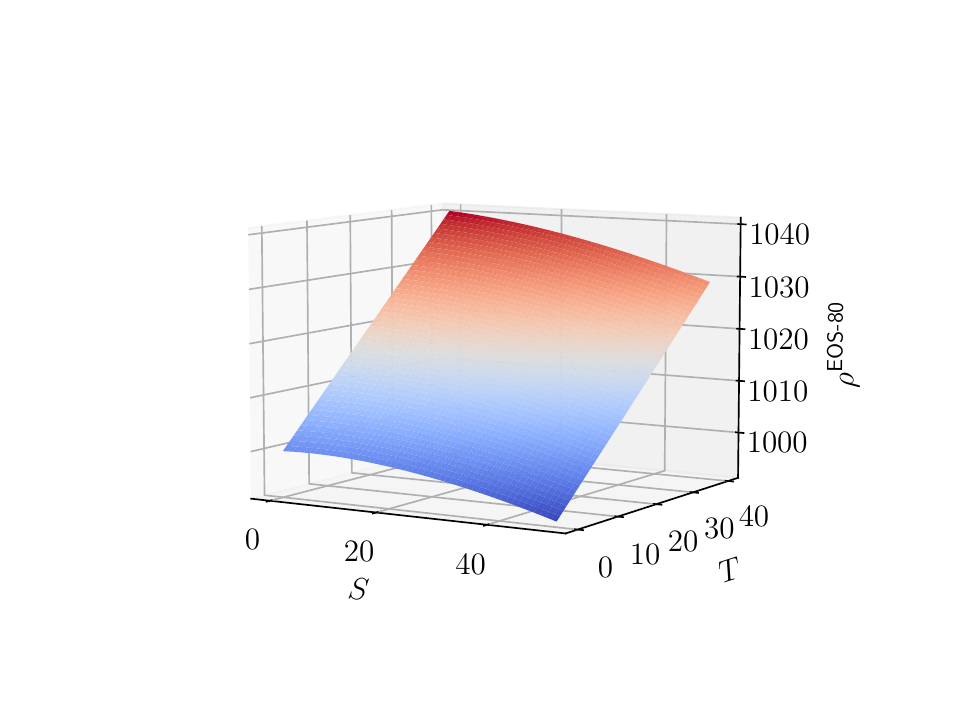}%
	\egroup%
	\caption{\(\rho^{\text{EOS-80}}(T, S)\) plotted on \(T \times S \subset {[0, 40]}^2\).
	\(\rho^{\text{EOS-80}}(T, S)\) is concave, and thus nonlinear, in \(S\).}%
	\label{fig:rho-eos80}
\end{figure}

With the inception of ML as a pervasive tool for predictive modeling, different questions
of societal importance, including climate science and oceanography (see for example~\cite{sonnewald2021bridging,jones2017machine,irrgang2021towards}),
have been addressed. We identify three key open questions that remain unaddressed: \emph{Firstly},
it remains unknown how well ML methods such as neural networks are able to capture the underlying
system behavior of the climate system. In order to use neural networks confidently we need to know
what determines a trained network's ability to estimate out-of-sample climate data reliably and
whether its skill corresponds to the underlying physics. If a net performs well only due to spurious
correlations in the input data, using the net for inference could lead to devastating societal
impact from wrong inference.\ \emph{Secondly}, it is unclear how confidently a neural net can
be used with the nonstationary forcing imposed by anthropogenically induced greenhouse gases.
As such, we are projecting systems, such as the AMOC, with known instabilities into an unknown
future by asking neural nets to be skillful out-of-sample, like predicting a \enquote{tipping point}
they previously have never seen.\ \emph{Thirdly}, there is no doubt that any prediction,
physics- or ML-based, will have some nonzero uncertainty associated with it. Thus, it is imperative
to be able to quantify the uncertainty of a neural net's prediction for it to be socially useful.

The question of whether a neural net has \enquote{learned} physics and if it can be applied out-of-sample
with low or quantified uncertainty are fascinating problems withing the field of ML\@. Here we
illustrate how skillful different architectures are at predicting the AMOC, how well neural networks
are able to capture the underlying physics using Explainable AI (XAI, described further below) and how confidently
they perform under nonstationary conditions using an ensemble approach. In particular, we are interested
in understanding the intrinsic differnces between neural nets trained on physics-informed features
vs.\ autoregressively, which established the duality \emph{understanding} vs.\ \emph{imitating} the
AMOC system.

Due to its importance, the AMOC and its potential for a shutdown or rapid change is studied using
conventional tools within climate science, such as climate models~\cite{menary2013mechanisms,swingedouw2013initialisation,cheng2013atlantic,wen2016active,jackson2023challenges},
but also using simpler statistical tools, such as stochastic process modeling using a Stochastic Differential
Equation (SDE)~\cite{ditlevsen2023warning}. The projections of these statistical tools recently caused
alarm, suggesting an AMOC shutdown could happen as soon as 2025~\cite{ditlevsen2023warning}. However,
such methods rely on making very strong assumptions about the AMOC system. With neural networks being
universal function approximators, we are less reliant on assumptions and we see much promise for the
use of Machine Learning.

The use of Machine Learning is already prevalent in weather forecasts~\cite{pathak2022fourcastnet,singh2019weather,salman2015weather},
as well as to predict and explain the impact of global heating on the North Atlantic Ocean~\cite{sonnewald2021revealing}.
More recently, research on the ocean~\cite{bire2023ocean} started employing the
Fourier~Neural~Operator (FNO;~\cite{li2020fourier}) in a novel ocean emulator system.
Previous efforts~\cite{hazeleger2013predicting,matei2012multiyear,mahajan2011predicting}
at predicting the AMOC are limited to short time scales within 10 years into the future
and are unable to simulate and quantify concrete climate change variables such as
greenhouse gas emissions.

\begin{table*}[htb]
	\centering
	\bgroup%
	\renewcommand{\arraystretch}{0.8}
		\begin{tabular}{cllc}
			\toprule
			Forcing Setup Identifier & \(F_s\) & \(F_t\) & \(\rho\) \\
			\midrule
			{\large {\Fone}} & Linear & --- & \multirow{3}{*}{\(\rho^{\text{lin}}\)} \\
			{\large {\Ftwo}} & Stationary sinusoidal & --- & \\
			{\large {\Fthree}} & Nonstationary sinusoidal & --- & \\
			\midrule
			{\large {\Ffour}} & Linear & Linear & \multirow{3}{*}{\(\rho^{\text{EOS-80}}\)} \\
			{\large {\Ffive}} & Stationary sinusoidal & Stationary sinusoidal & \\
			{\large {\Fsix}} & Nonstationary sinusoidal & Nonstationary sinusoidal & \\
			\bottomrule
		\end{tabular}%
	\egroup%
	\caption{The set of forcing setups that we consider in our experiments.}%
	\label{tab:forcing-setups}%
\end{table*}

Measured AMOC data is not abundant and only about 20 years of contiguous AMOC data
have been collected so far~\cite{mccarthy2015measuring}. This significant drawback
of using real world data to predict the AMOC is limiting accurate future projections
and usually does not incorporate the driving variables of the AMOC\@. A different
downside is that none of the previously mentioned works on ocean modeling assess the
question of whether the neural network has captured the underlying physical system,
or if the net is fit for deployment in nonstationary applications. In our work, we
model the North Atlantic Ocean using a physical system and subsequently attempt to
predict its behavior using neural networks. In addition, we not only predict but
\emph{explain} why different network architectures predicted the system in the way
they did, yielding an approach that is flexible and can use any time scale (months,
years, 1000s of years).

We consider a model of the Atlantic Ocean consisting of two boxes; a northern and a southern box.
This model is also called the \enquote{Stommel box model}~\cite{stommel1961thermohaline} and is
canonically used to model ocean circulation like the AMOC\@. In the Stommel box model, each box
is assigned a variable temperature \(T_i\) and a salinity \(S_i\) where \(i \in \{1, 2\}\). The
AMOC is represented as a variable \(q\) and \(F_s\) is an external fresh water forcing originating
from anthropogenic climate change. In \cref{fig:stommel_model_Fs_Ft}, we depict a schematic of the
Stommel model with a fresh water and a temperature forcing, i.e., external water that flows into the
southern box as well as greenhouse emissions altering its temperature.
We list all parameters for the box model along with their values for this paper in \cref{sec:box-model-parameters},
\cref{tab:parameters-box-model}.

\begin{figure*}[htb]
	\centering
	\bgroup%
	\newcommand{\scale}{0.2}
	\begin{tabular}{ccc}
		\toprule
		BNN & MLP & DE \\
		\midrule

		\includegraphics[scale=\scale,valign=m]{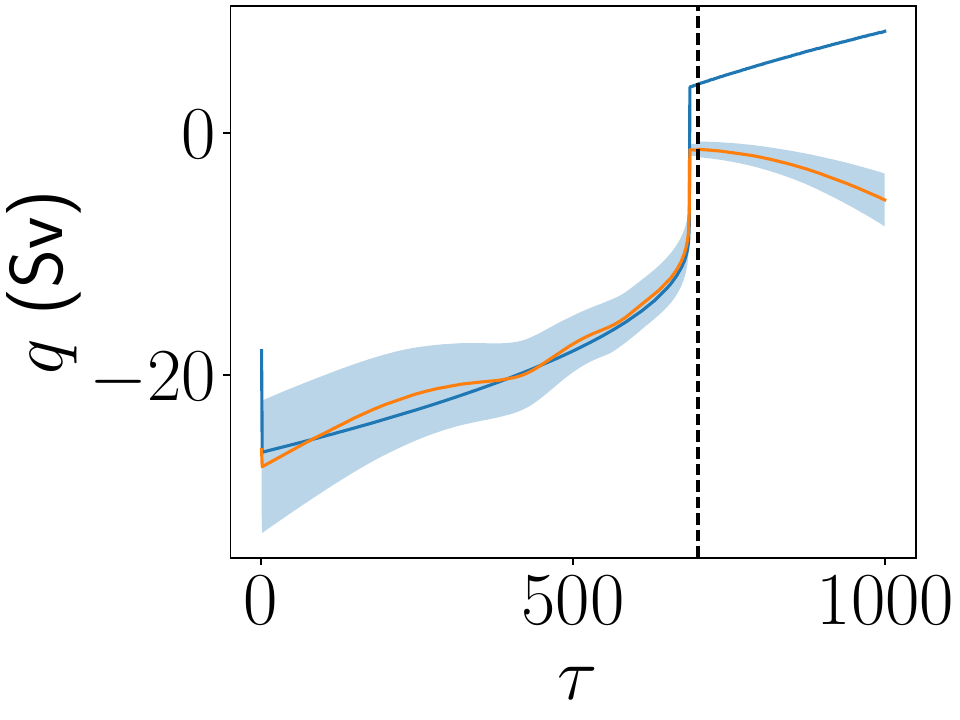}
		& \includegraphics[scale=\scale,valign=m]{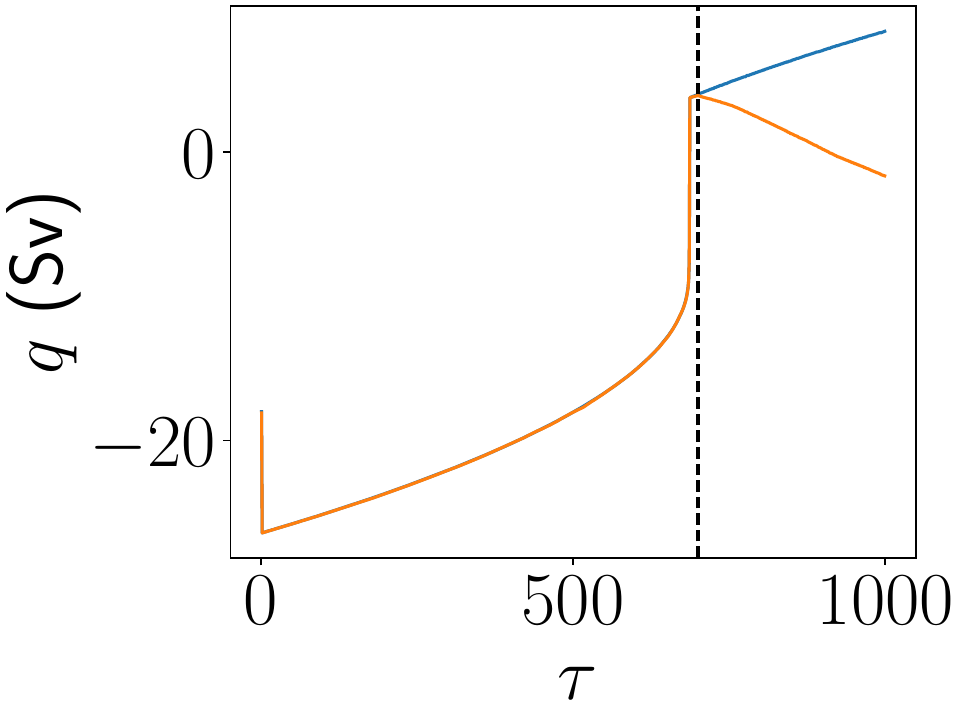}
		& \includegraphics[scale=\scale,valign=m]{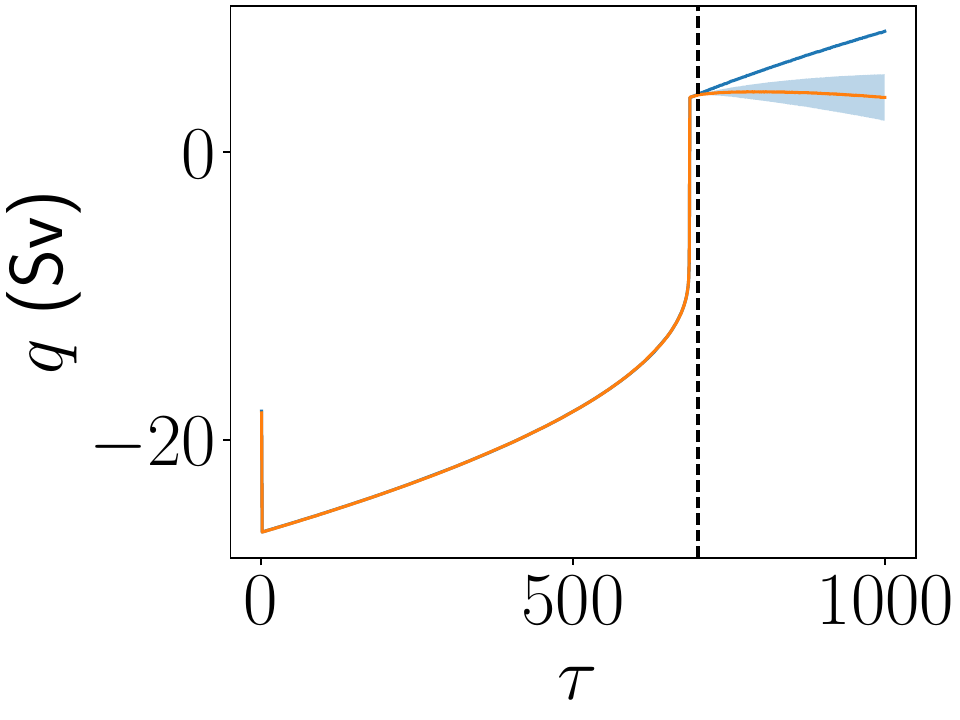} \\
		\includegraphics[scale=\scale,valign=m]{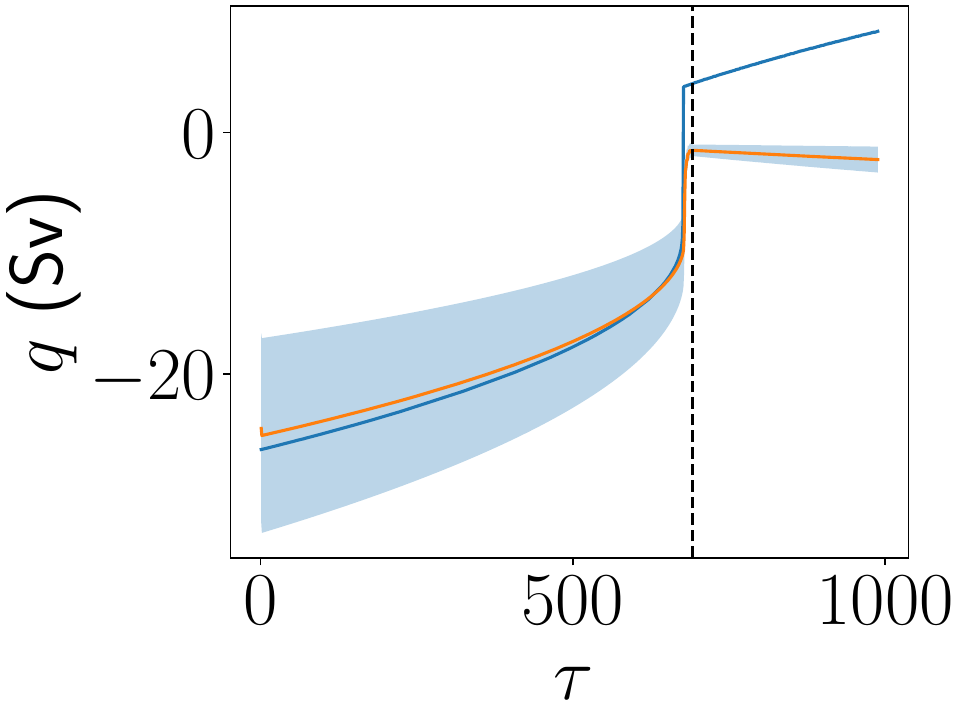}
		& \includegraphics[scale=\scale,valign=m]{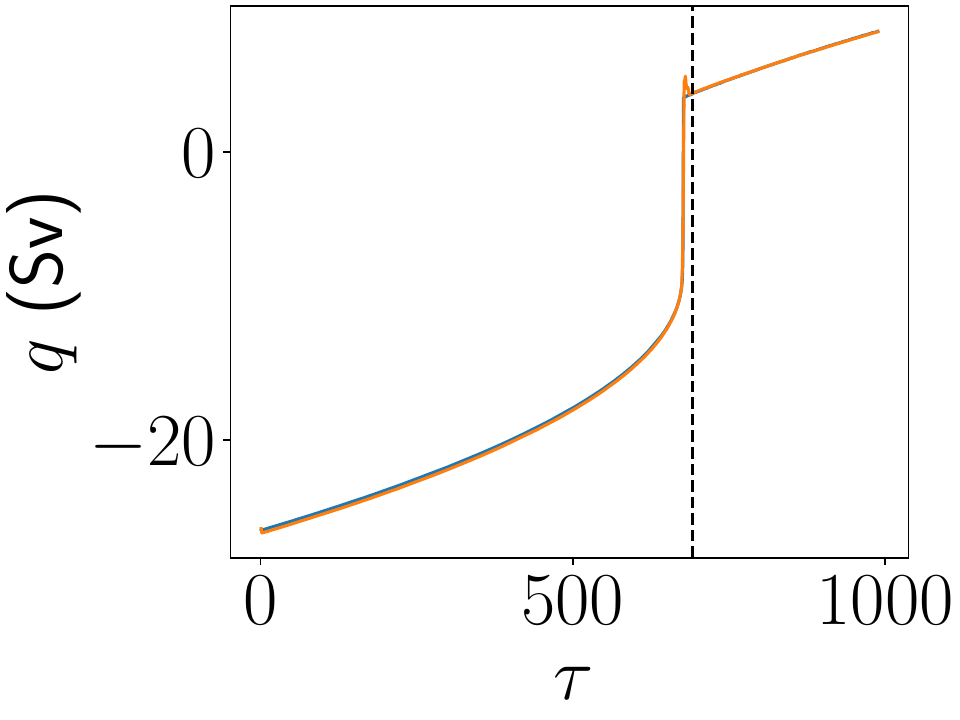}
		& \includegraphics[scale=\scale,valign=m]{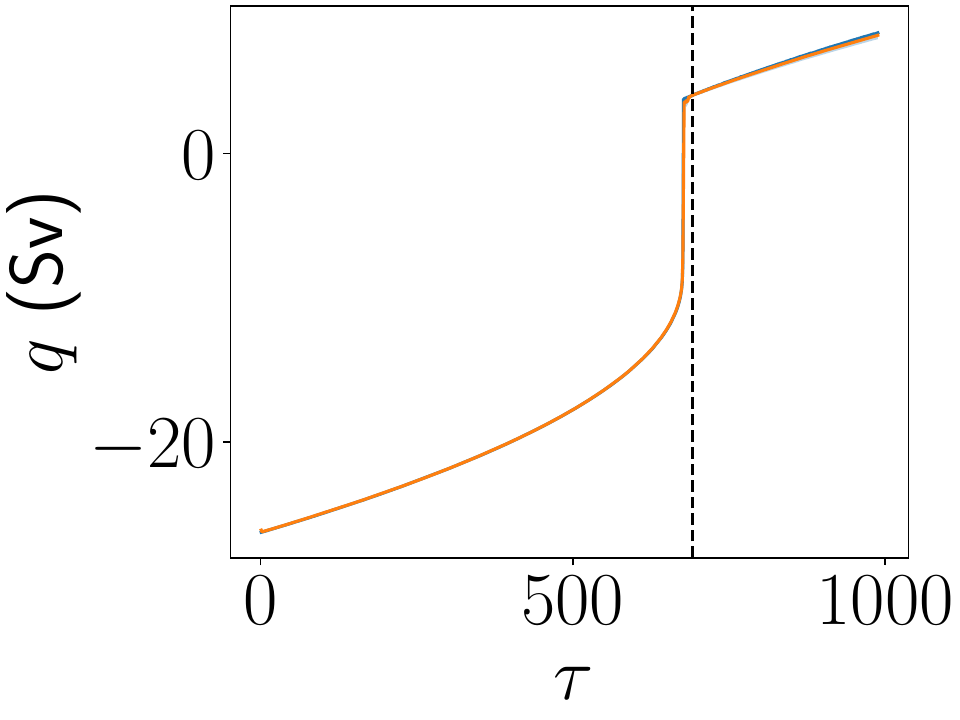} \\

		\multicolumn{3}{c}{\includegraphics[scale=0.6,valign=m]{legend_pred_gt.pdf}} \\

		\bottomrule
	\end{tabular}
	\egroup%
	\caption{Predictive performance for the considered architectures using physics-informed (PI\@; first row) and
	autoregressive (AR\@; second row) features under {\Fone}.}%
	\label{tab:F1-performance}%
\end{figure*}

\begin{figure}[htb]
	\centering
	\begin{adjustbox}{width=\columnwidth}
	\begin{tabular}{ccc}
		\toprule
		\resizebox{0.3\linewidth}{!}{BNN} & \resizebox{0.3\linewidth}{!}{MLP} & \resizebox{0.2\linewidth}{!}{DE} \\
		\midrule

		\includegraphics[width=\linewidth,valign=m]{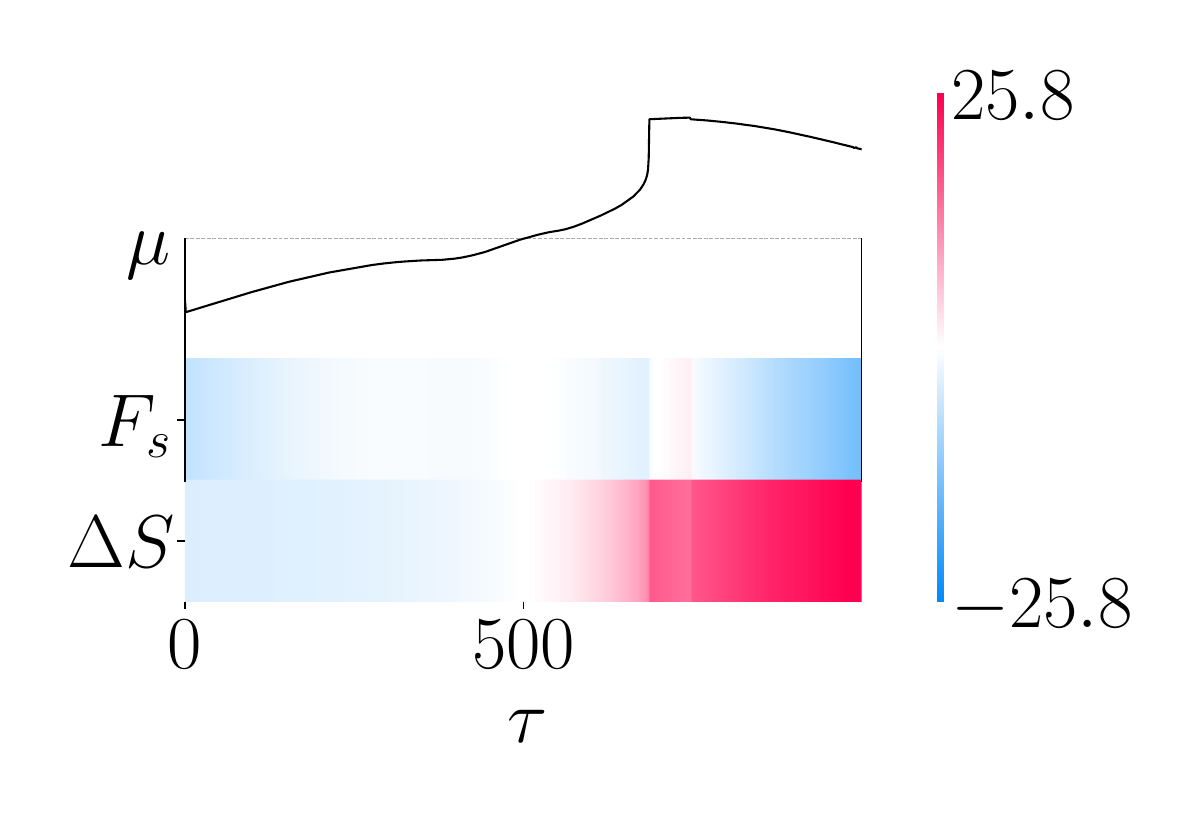}
		& \includegraphics[width=\linewidth,valign=m]{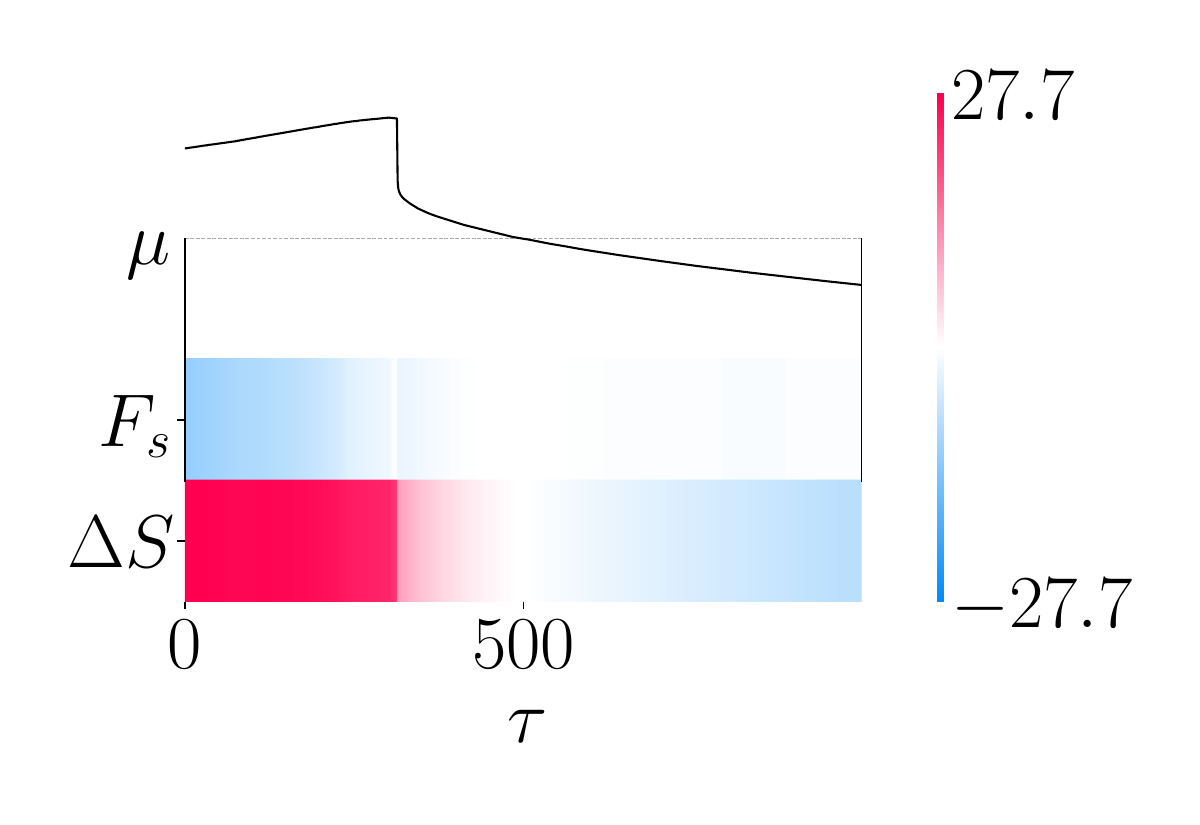}
		& \includegraphics[width=\linewidth,valign=m]{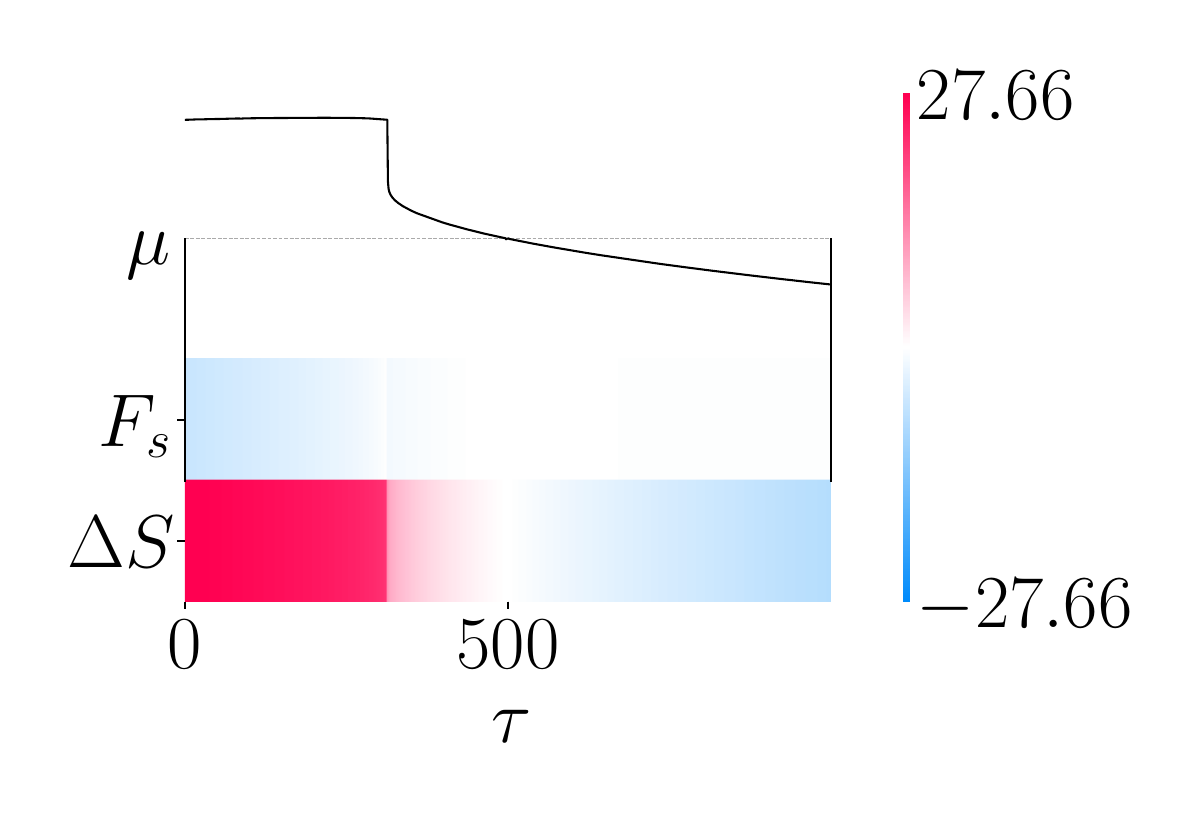} \\

		\includegraphics[width=\linewidth,trim={0 4cm 0 6cm},clip,valign=m]{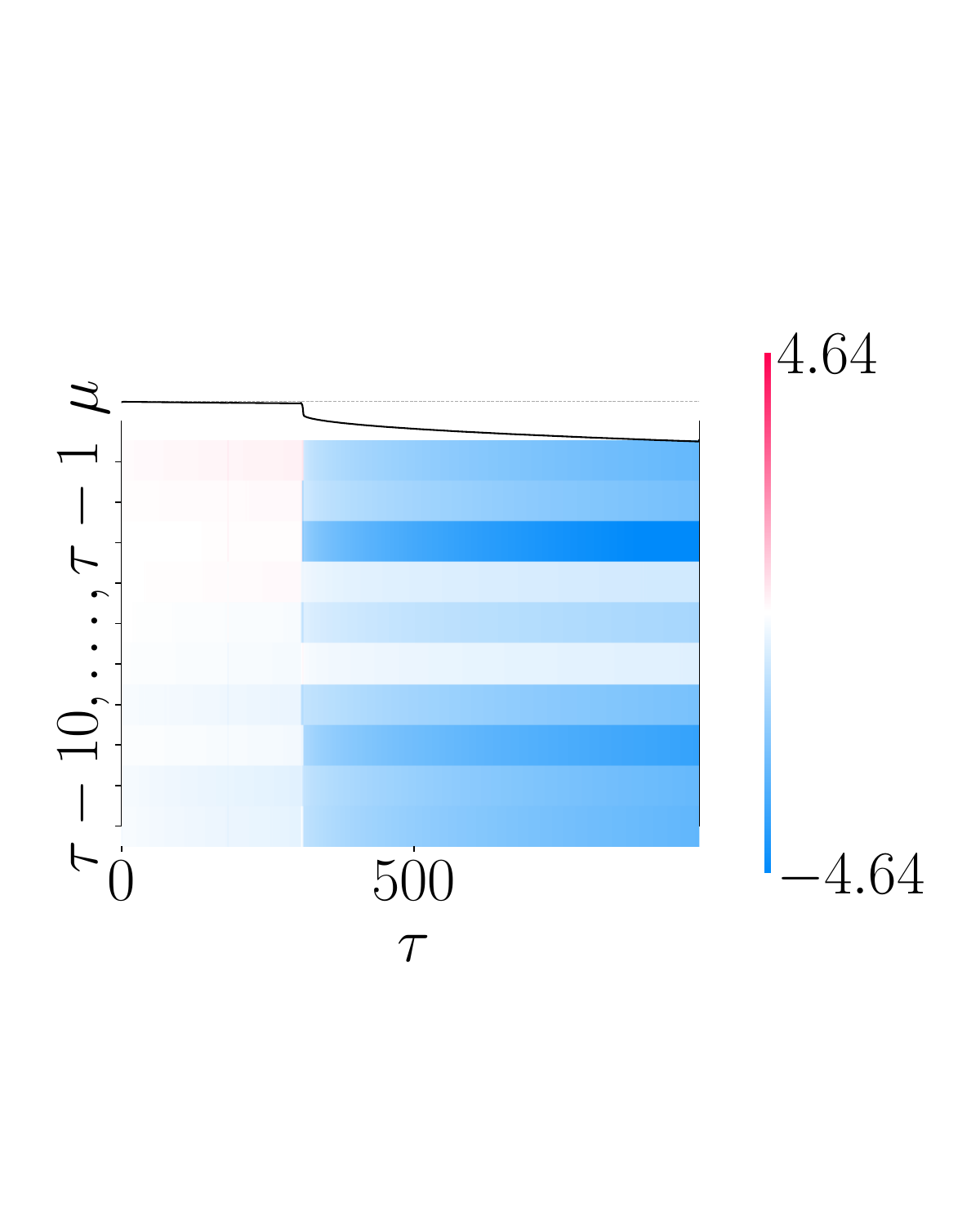}
		& \includegraphics[width=\linewidth,trim={0 4cm 0 6cm},clip,valign=m]{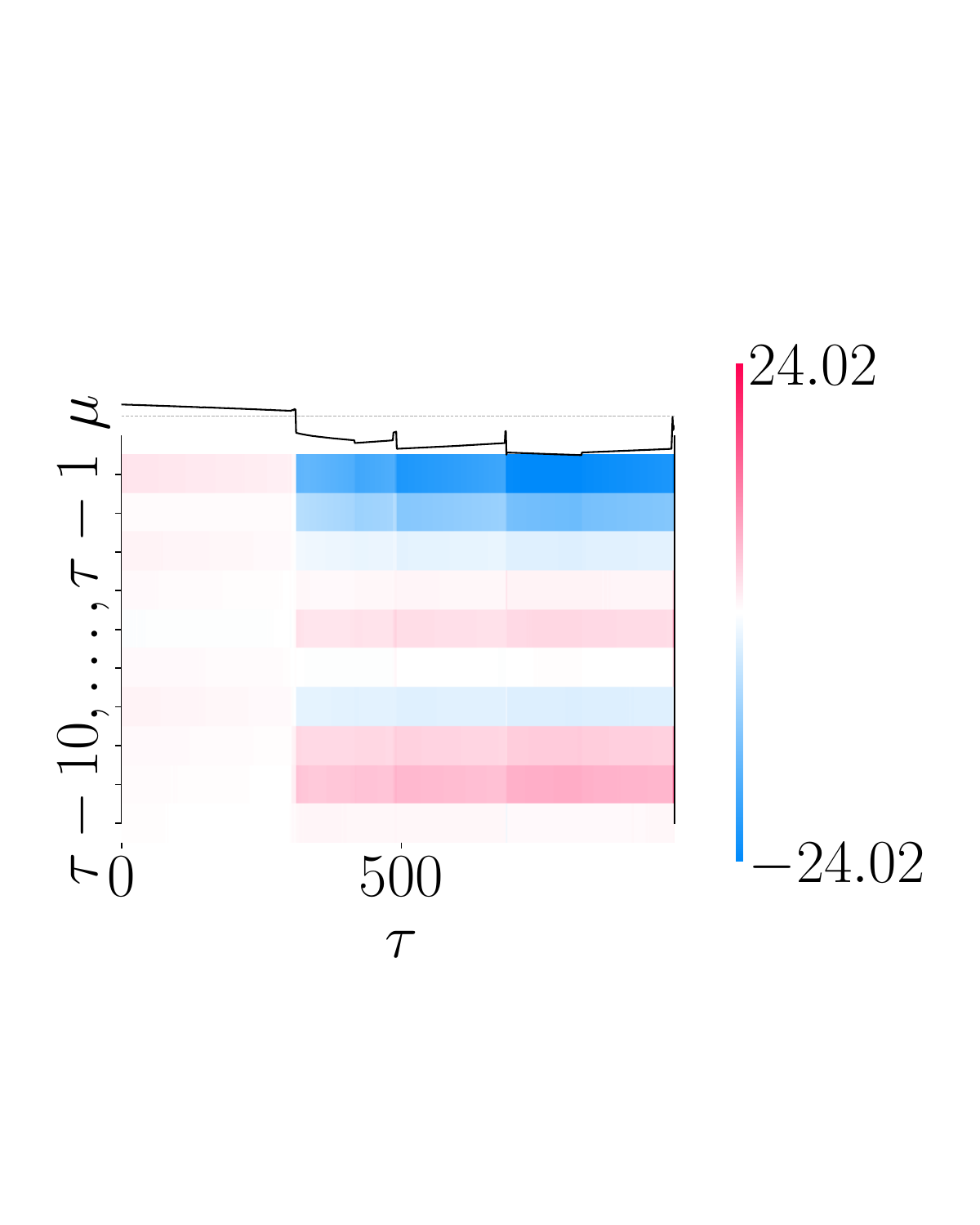}
		& \includegraphics[width=\linewidth,trim={0 4cm 0 6cm},clip,valign=m]{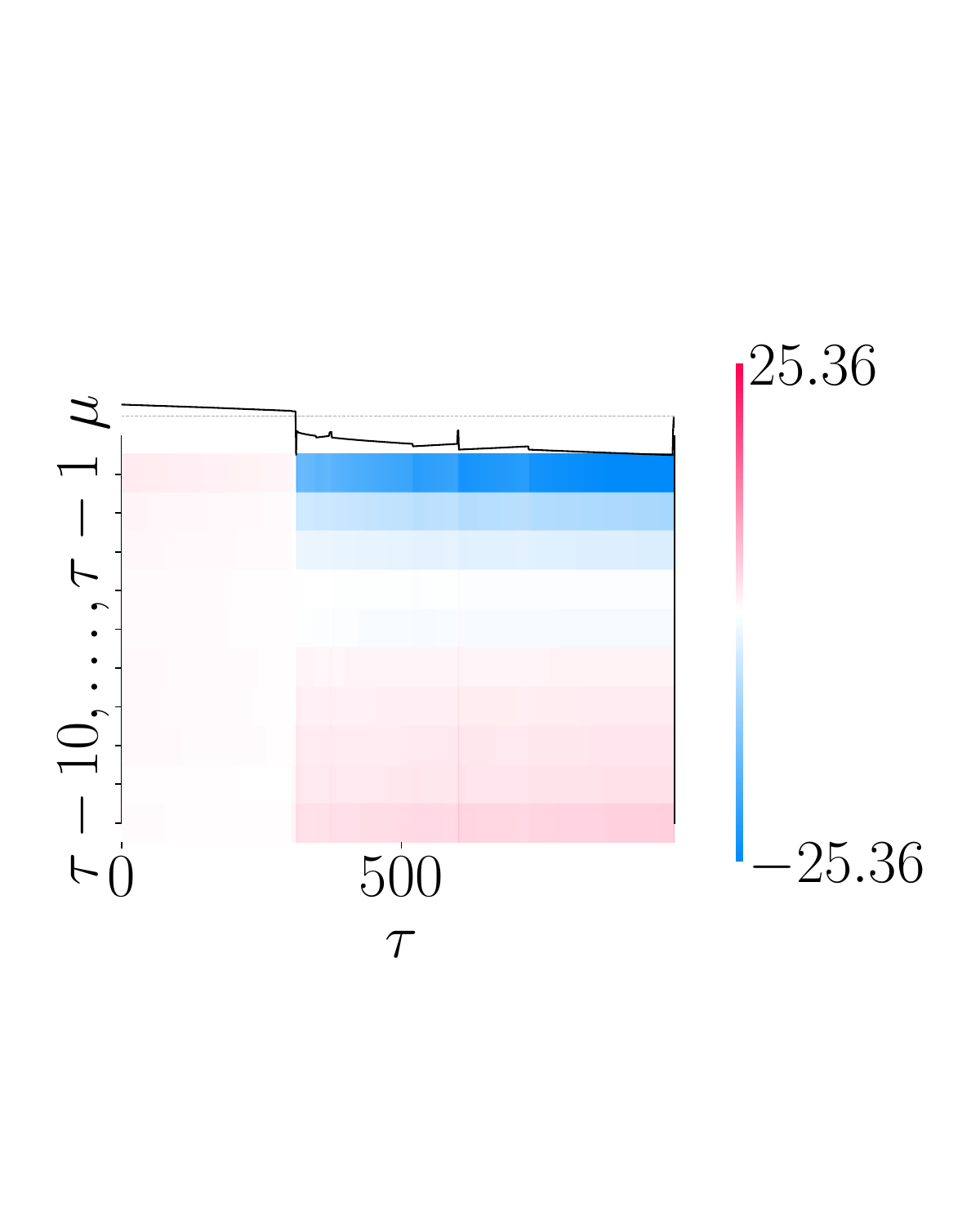} \\
		\bottomrule
	\end{tabular}
	\end{adjustbox}%
	\caption{DeepLIFT attribution maps for the considered architectures using physics-informed (PI\@; top row) and
	autoregressive (AR\@; bottom row) features under {\Fone}.}%
	\label{tab:F1-xai}%
\end{figure}

At the heart of the Stommel box model lie the concepts \emph{breakdown} and
\emph{recovery}.\ \emph{Breakdown} refers to the exceeding of a point in
the model that leads to a sudden collapse of the AMOC~\cite{liu2017overlooked,tziperman2022global}.
Similarly, the inverse concept \emph{recovery} denotes the return of the AMOC
to a previous state in which it had not collapsed~\cite{thomas2019mechanisms}.
The Stommel box model thus gives us three stable states a neural net should be
able to model and a relatively simple attractor basin that emerges through the
combination of equations describing the box model. This attractor basin and its
underlying equations are what the neural net should \enquote{learn}. From a neural
network perspective, ensemble neural nets offer an intuitive avenue in terms of
exploring the attractor basin since it is known that there exist three equally
important stable states in the Stommel box model~\cite{tziperman2022global}.

\section{Modeling the AMOC using Physical Equations and Neural Networks}

\subsection{Importance of Density and its Approximation}

The \emph{density} \(\rho_i\) in each box of the Stommel model depends on \(T_i\)
and \(S_i\) and fundamentally affects \(q\) via the difference in \(\rho_i\) between
both boxes~\cite{tziperman2022global}. There exist two ways to model the water density
in box 1 and 2. The first is a linear approximation
\begin{equation}
	\rho(T, S) = \rho_0 - \alpha (T - T_0) + \beta (S - S_0)
\end{equation}
where \(\rho_0\) is a baseline density value and \(T_0\), \(S_0\) are baseline
values for temperature and salinity, respectively.\ \(\alpha\) and \(\beta\) are
the thermal expansion and haline contraction coefficients. The second is the Equation
of State (EOS-80;~\cite{fofonoff1985physical}) for seawater, which for a reference
pressure (in this paper, atmospheric pressure) nonlinearly approximates seawater
density. Due to its size we do not outline it here.

\subsection{Modeling \(q\)}%
\label{sec:modeling-q}

\paragraph{Nonlinear case.} We begin with \(\rho\) being nonlinear. First, let \(\tau\)
be the time step which we use to model the evolution of \(S_i\) and \(T_i\) through
time~\cite{tziperman2022global}:
\begin{align}
	\pdv{S_1}{\tau} &= \frac{\abs{q} (S_2 - S_1) - F_s(\tau)}{V_1} \label{eq:s1-dt} \\
	\pdv{S_2}{\tau} &= \frac{\abs{q} (S_1 - S_2) + F_s(\tau)}{V_2} \label{eq:s2-dt} \\
	\pdv{T_1}{\tau} &= \frac{\abs{q} (T_2 - T_1) - F_t(\tau)}{V_1} \label{eq:t1-dt} \\
	\pdv{T_2}{\tau} &= \frac{\abs{q} (T_1 - T_2) + F_t(\tau)}{V_2} \label{eq:t2-dt}
\end{align}
with \(V_i\) being the volume of the \(i\)-th box. \(F_s(\tau)\), a fresh water forcing,
and \(F_t(\tau)\), a temperature forcing, represent ice melting in the polar region or
increased heat due to elevated greenhouse gas levels. These allow us to effectively
simulate different climate scenarios. Note that \(F_t(\tau)\) is optional (fallback
to the standard box model variant) and leaving it out omits \cref{eq:t1-dt,eq:t2-dt}.
\Cref{fig:rho-eos80} shows a plot of \(\rho^{\text{EOS-80}}(T, S)\), demonstrating
that it is concave in \(S\). For the experiments using \(\rho^{\text{EOS-80}}\),
we select the initial \(S_i\) such that they lie in the interval \([5, 15]\) in
order to introduce as much nonlinearity as possible into the box model, thereby
establishing a stark contrast between experiments with a linear and nonlinear density.

\paragraph{Linear case.} Following~\cite{tziperman2022global}, we define
\begin{align}
	\Delta S &= S_1 - S_2 \\
	\Delta T &= T_1 - T_2
\end{align}
which due to the linearity of differentiation leads to the formulation of the
partial differential equations that model the evolution of \(\Delta S\) and
\(\Delta T\) under forcings \(F_s\) and \(F_t\) over time, respectively~\cite{tziperman2022global}:
\begin{align}
	\pdv{\Delta S}{\tau} &= -2\frac{\abs{q} \Delta S + F_s(\tau)}{V} \label{eq:delta-s} \\
    \pdv{\Delta T}{\tau} &= -2\frac{\abs{q} \Delta T + F_t(\tau)}{V} \label{eq:delta-t}
\end{align}
where \(V = V_1 = V_2\) is the box volume.

Finally, \(q\) is modeled as a proportional density difference between box 1 and 2:
\begin{equation}
	q = k (\rho_1 - \rho_2) =
	\begin{cases}
		k (\beta \Delta S - \alpha \Delta T), & \rho = \rho^{\text{lin}} \\
		k (\rho_1^{\text{EOS-80}} - \rho_2^{\text{EOS-80}}), & \rho = \rho^{\text{EOS-80}}
	\end{cases}
\end{equation}

\subsection{Obtaining Simulated Climate Data with the Stommel Box Model}

By solving \cref{eq:delta-s,eq:delta-t} or \cref{eq:s1-dt,eq:s2-dt,eq:t1-dt,eq:t2-dt},
respectively, we receive sets of values ordered by time:
\begin{align}
	{\left\{ \Delta S_\tau \right\}}_{\tau = 0}^{\tau_{\text{max}}} && {\left\{ \Delta T_\tau \right\}}_{\tau = 0}^{\tau_{\text{max}}}
\end{align}
and
\begin{align}
	{\left\{ S_{1, \tau} \right\}}_{\tau = 0}^{\tau_{\text{max}}} && {\left\{ S_{2, \tau} \right\}}_{\tau = 0}^{\tau_{\text{max}}} \\
	{\left\{ T_{1, \tau} \right\}}_{\tau = 0}^{\tau_{\text{max}}} && {\left\{ T_{2, \tau} \right\}}_{\tau = 0}^{\tau_{\text{max}}}
\end{align}
with \({\{F_s(\tau)\}}_{\tau = 0}^{\tau_{\text{max}}}\) and \({\{F_t(\tau)\}}_{\tau = 0}^{\tau_{\text{max}}}\)
as additional sets of values over time. Note that for the standard box model, we do not include \(\Delta T\)
as an input feature due to the fact that for \(F_t \equiv 0\), \({\{\Delta T_\tau\}}_{\tau = 0}^{\tau_\text{max}} \approx {\{0\}}_{\tau = 0}^{\tau_\text{max}}\).

Now let \(f_\theta\) be a neural network with parameters \(\theta\). We consider two
fundamental approaches;
\begin{enumerate}[wide=0pt,widest=99,leftmargin={\parindent},label= \textbf{Approach (\Roman*):}]
	\item Physics-informed (PI), in the sense that the inputs to \(f_\theta\) are
	physical variables:
	\begin{equation}
		f_\theta(\Delta S_\tau, \Delta T_\tau, F_s(\tau), F_t(\tau)) = q_\tau
	\end{equation}
	and
	\begin{equation}
		f_\theta(S_{1, \tau}, S_{2, \tau}, T_{1, \tau}, T_{2, \tau}, F_s(\tau), F_t(\tau)) = q_\tau.
	\end{equation}
	\item Autoregressive (AR), where we discard the physical variables and
	predict \(q\) from the last \(\omega\) values of \(q\) in time:
	\begin{equation}
		f_\theta(q_{\tau - \omega}, \dots, q_{\tau - 1}) = q_\tau.
	\end{equation}
\end{enumerate}

\begin{figure*}
	\centering
	\bgroup%
	\newcommand{\scale}{0.15}
		\begin{tabular}{cccc}
			\toprule
			Forcing Setup & Forcing~~~\includegraphics[scale=0.5,valign=m]{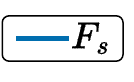} & Variables~~~\includegraphics[scale=0.5,valign=m]{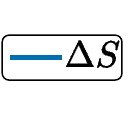} & \(q\) \\
			\midrule

			{\Fone}
			& \includegraphics[scale=\scale,valign=m]{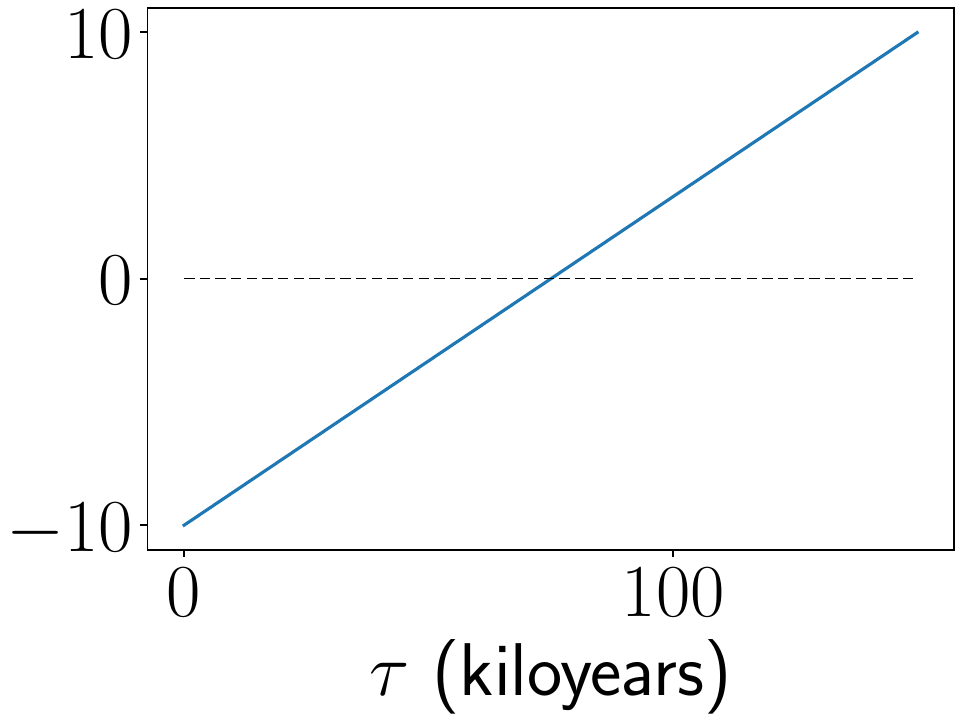}
			& \includegraphics[scale=\scale,valign=m]{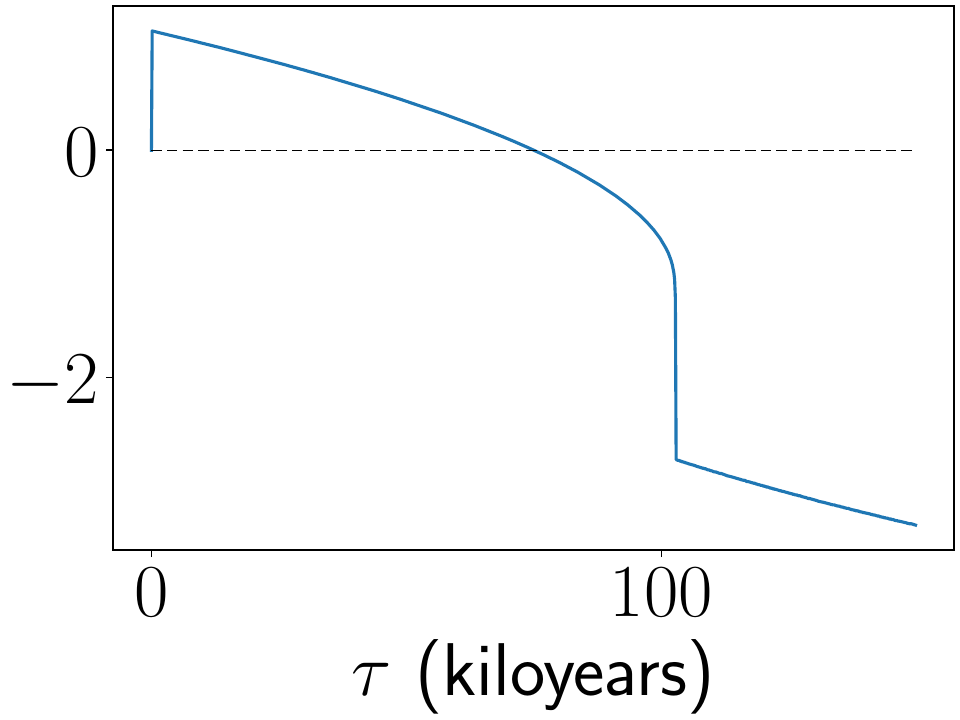}
			& \includegraphics[scale=\scale,valign=m]{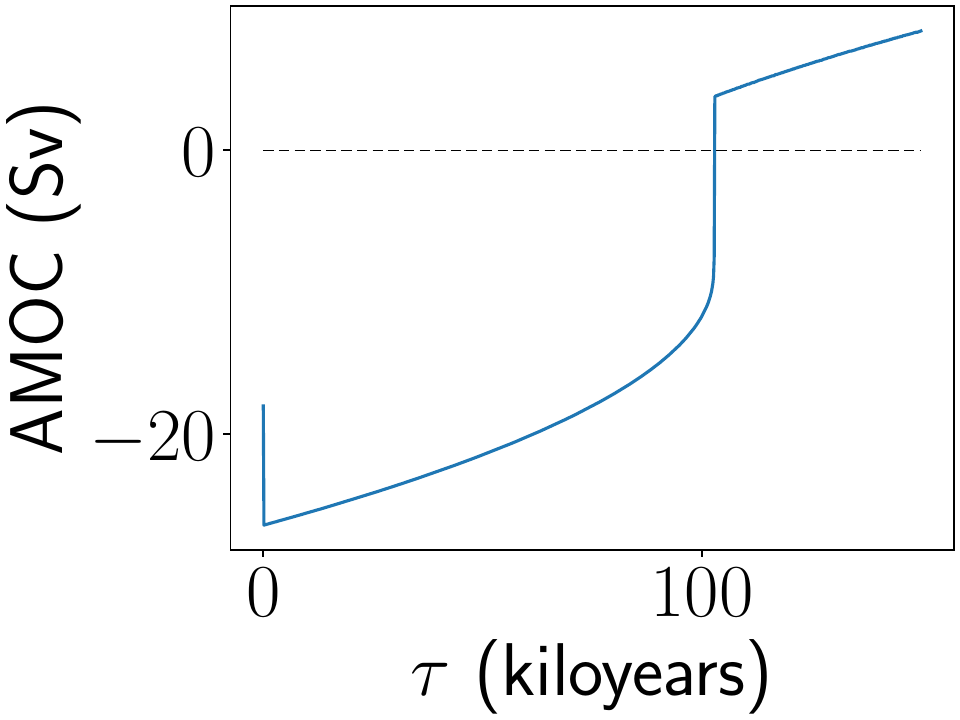} \\
			\midrule

			{\Ftwo}
			& \includegraphics[scale=\scale,valign=m]{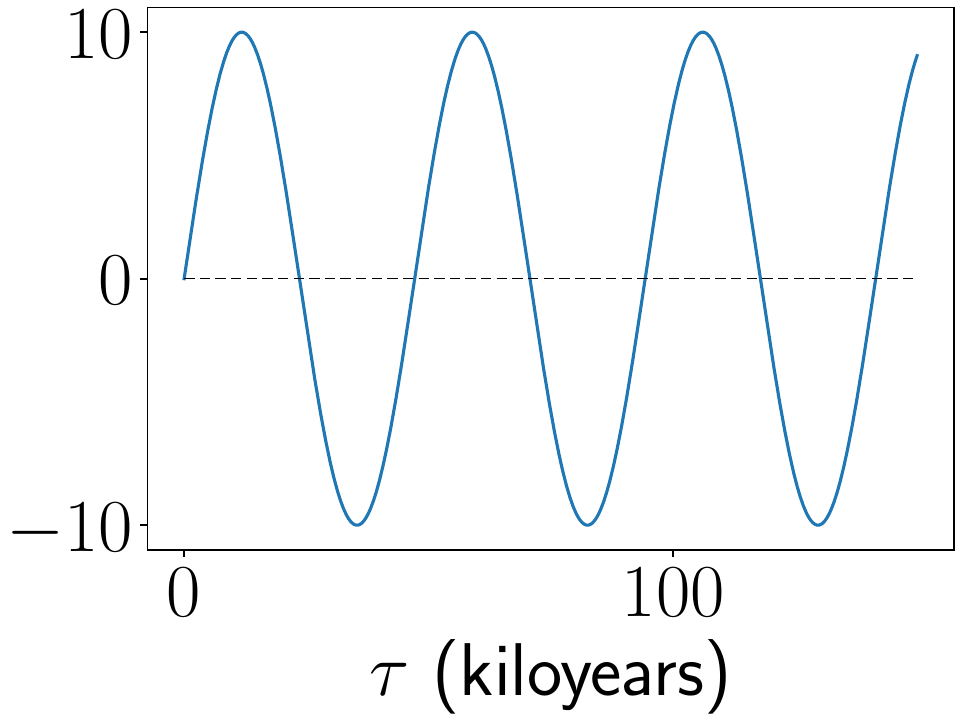}
			& \includegraphics[scale=\scale,valign=m]{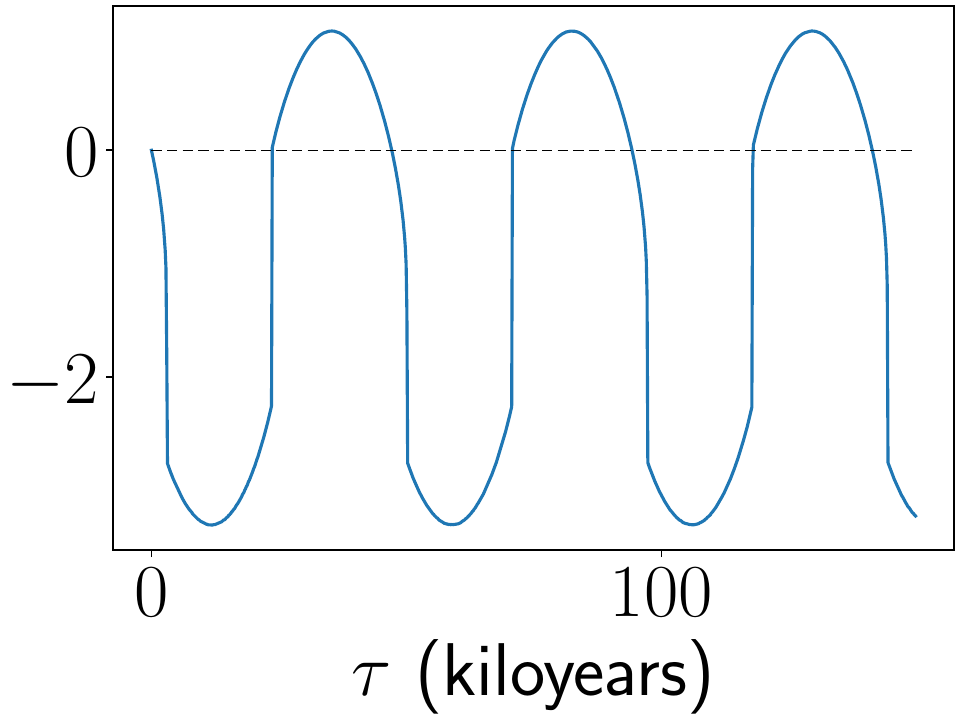}
			& \includegraphics[scale=\scale,valign=m]{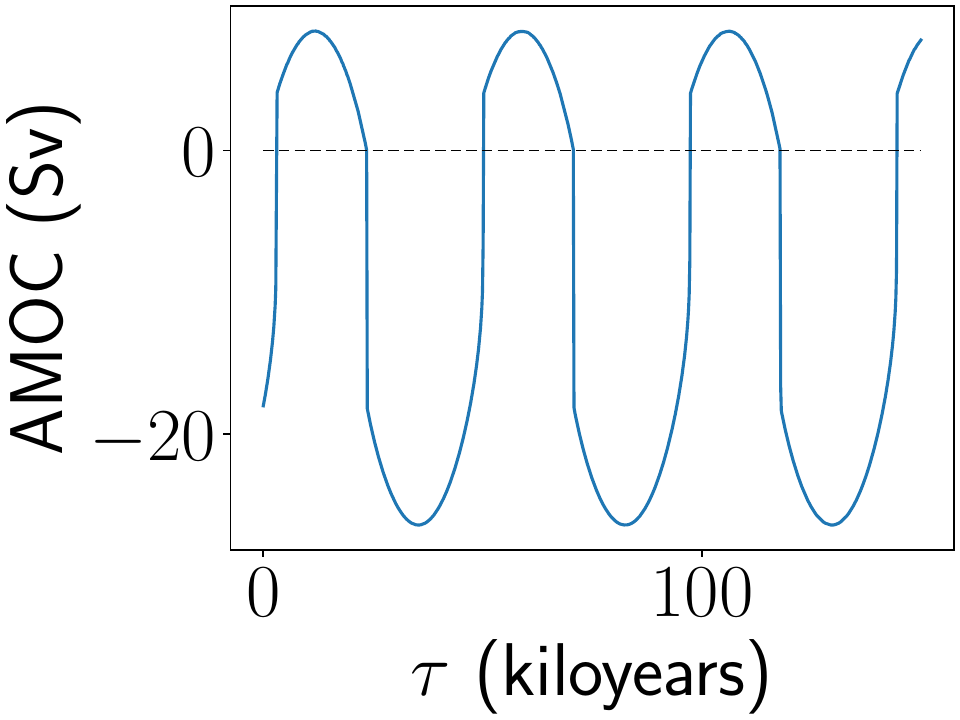} \\
			\midrule

			{\Fthree}
			& \includegraphics[scale=\scale,valign=m]{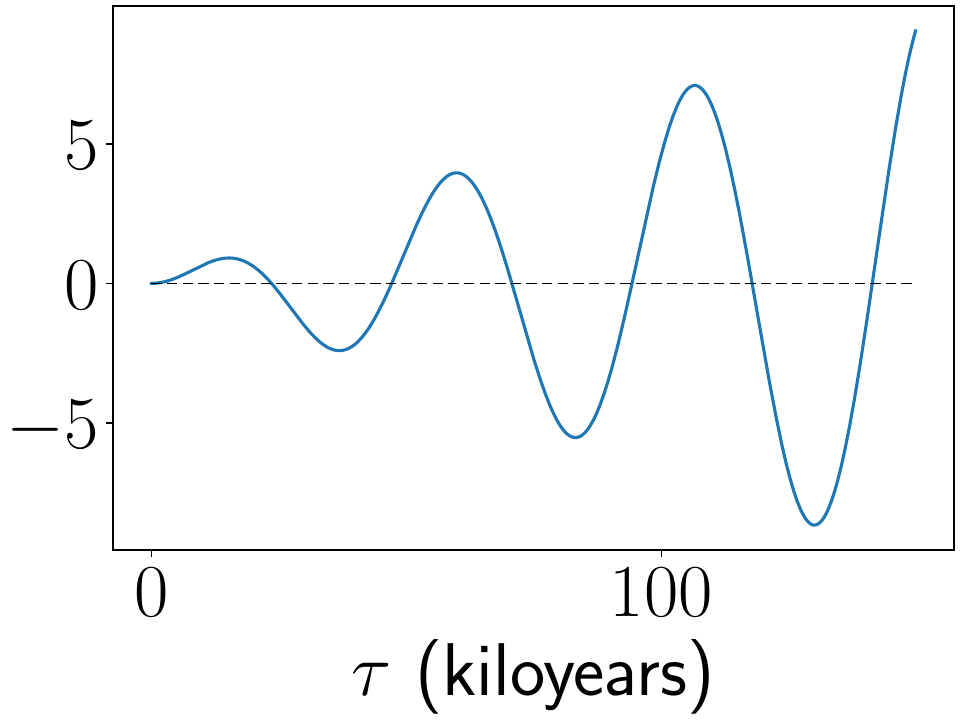}
			& \includegraphics[scale=\scale,valign=m]{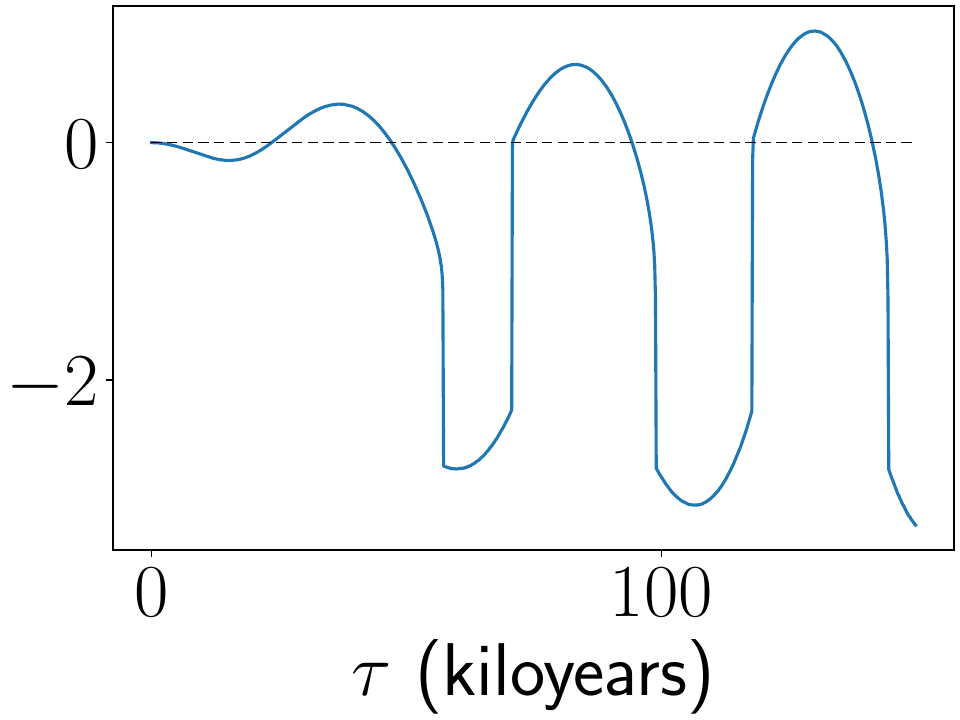}
			& \includegraphics[scale=\scale,valign=m]{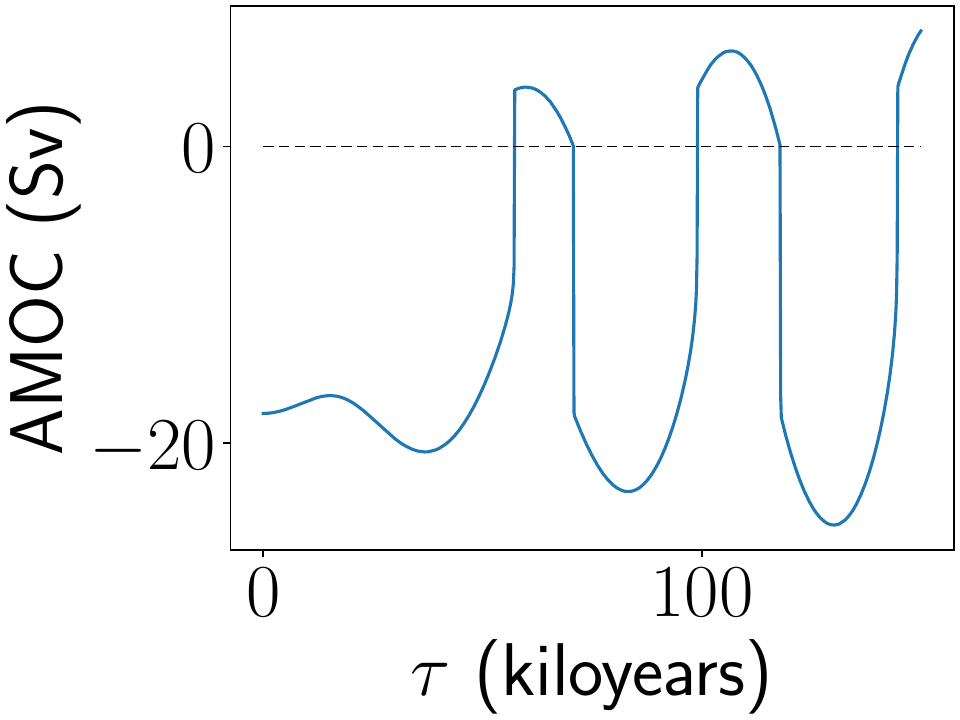} \\

			\bottomrule
		\end{tabular}
	\egroup%
	\caption{Forcing trajectory, physical variables and AMOC for {\Fone}, {\Ftwo} and {\Fthree}.}%
	\label{tab:F1-F2-F3}%
\end{figure*}

\subsection{Neural Network Architectures}

We consider three neural network architectures that we use to predict the AMOC\@:
Multi-Layer Perceptron (MLP), Deep Ensemble (DE) and Bayesian Neural Network (BNN).
Only two of the aforementioned architectures are able to quantify uncertainty; the Deep Ensemble
by averaging and the BNN by Monte Carlo sampling and successive computation of the
mean and standard deviation~\cite{jospin2022hands}. BNNs have previously been used
for ocean applications~\cite{clare2022bayesian,clare2022explainable,yik2023southern,rasouli2020forecast,bittig2018alternative,juan2023uncertainties,clare2022bayesian},
as well as different tasks using them for uncertainty quantification~\cite{thodberg1996review,springenberg2016bayesian,zhang2021bayesian,bao2020epistemic}.
Deep Ensembles consist of training multiple independent deep neural networks whose
predictions are averaged during inference. They have been used extensively in climate
science applications to generate more accurate predictions~\cite{sonnewald2021bridging,cho2022novel,jahanbakht2021sea,qi2023advancing}.

\subsection{Explainability}

Current works~\cite{lobelle2020detectability,ditlevsen2023warning} on predicting the AMOC focus
on predicting the AMOC's progression over time with an emphasis on a potential tipping point.
They do not, however, attempt to explain why their predictive models output the resulting
predictions. This is a major restriction with respect to determining \emph{why} the AMOC is
projected to evolve in a certain way and impedes climate actionability. We overcome this
restriction by generating \emph{feature attributions} per \(\tau\), visualizing what input
feature had a strong contribution for \(q_\tau\).

We apply DeepLIFT~\cite{shrikumar2017learning} and SHAP~\cite{lundberg2017unified}
to each trained \(f_\theta\) in order to determine whether \(f_\theta\) learned the underlying system
dynamics or simply imitated the time series \({\{q_\tau\}}_{\tau = 0}^{\tau_{\text{max}}}\).
For example, in forcing scenarios with the only forcing being \(F_s\), \(f_\theta\) should
not make a prediction \(\hat{q}_\tau = f_\theta(\dots)\) such that the corresponding attribution
\(\abs{\mathcal{A}(\hat{q}_\tau, \Delta S)}\) for \(\Delta S\) is small. This is because
salinity differences, induced by the fresh water forcing, drive the AMOC\@. We chose DeepLIFT
in combination with SHAP due to their relative recency among comparable methods~\cite{selvaraju2016grad,ribeiro2016should}
and continued impactfulness~\cite{li2021deep,sixt2020explanations,zhou2023interpretable,cakiroglu2024data,dewi2023xai}.
Besides this, we aim to use different attribution perspectives; sampling feature subsets as
well as propagating internal contribution scores can lead to different attribution maps and
tell us something about the internal and external attribution mechanics of a trained predictor,
respectively. This can also be seen as examining sensitivity to data vs.\ the internal network
parameters. Prior work~\cite{clare2022explainable} has examined this, linking different explainability
algorithms to aleatoric vs.\ epistemic uncertainty. Aleatoric uncertainty refers to uncertainty
that is inherent in the data whereas epistemic uncertainty denotes uncertainty inherent in the model
that is being used~\cite{clare2022explainable}. For the autoregressive cases (using SHAP) due to
an input length of 10, we do not use all feature subsets but rather sample them. We repeat this
sampling process 20 times and take the mean for each \(\tau\) in each attribution map.

\begin{figure}[htb]
	\centering
	\begin{adjustbox}{width=\columnwidth}
		\begin{tabular}{ccc}
			\toprule
			{\Huge BNN} & {\Huge MLP} & {\Huge DE} \\
			\midrule

			\includegraphics[width=1.5\linewidth,valign=m]{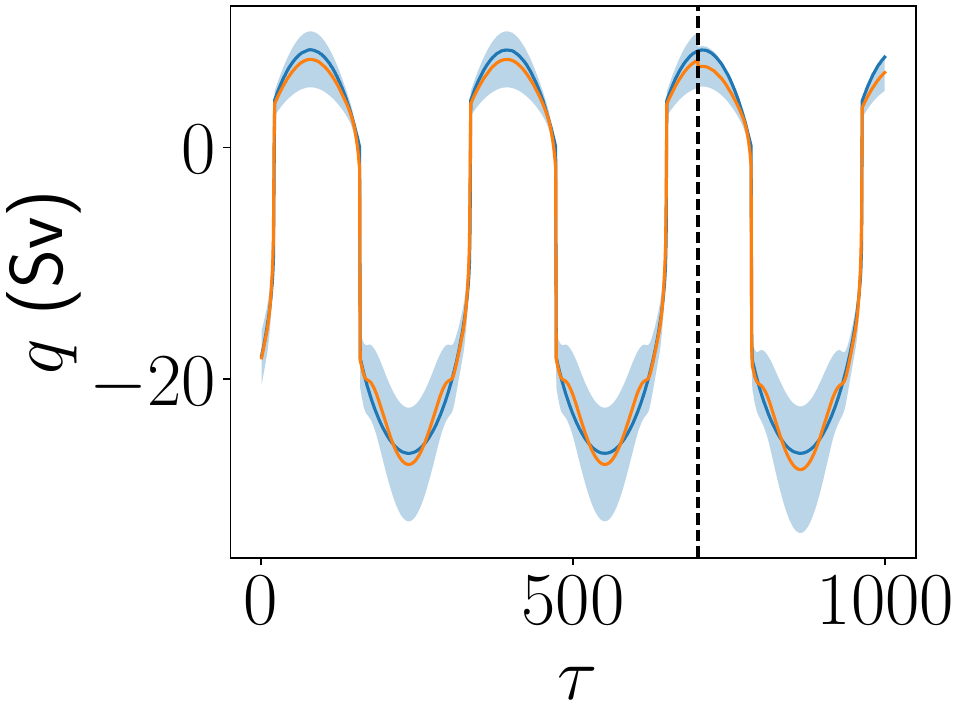}
			& \includegraphics[width=1.5\linewidth,valign=m]{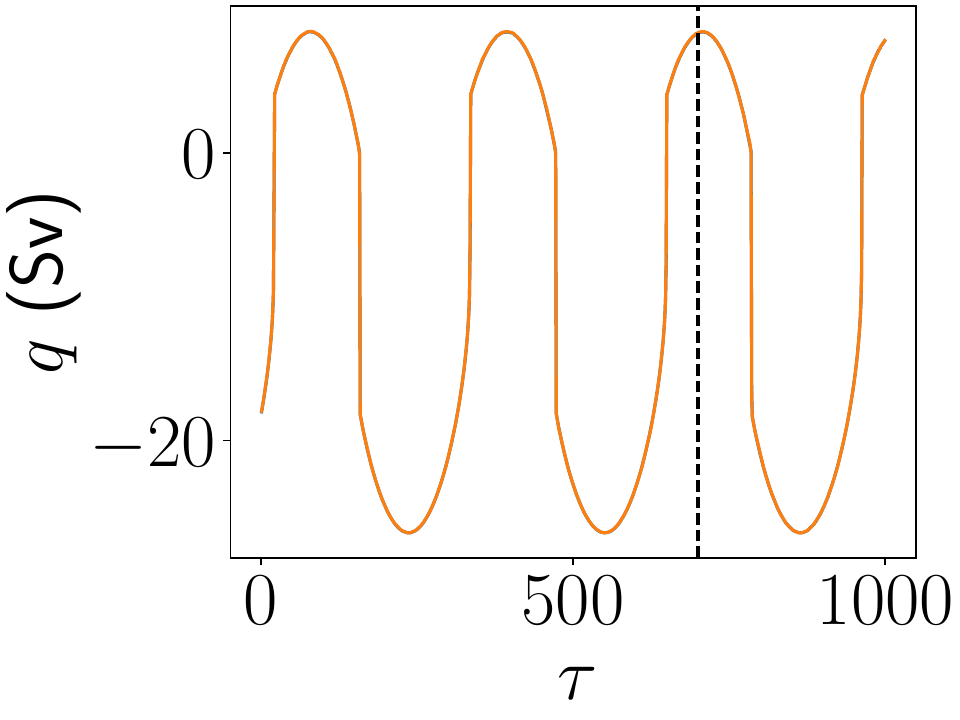}
			& \includegraphics[width=1.5\linewidth,valign=m]{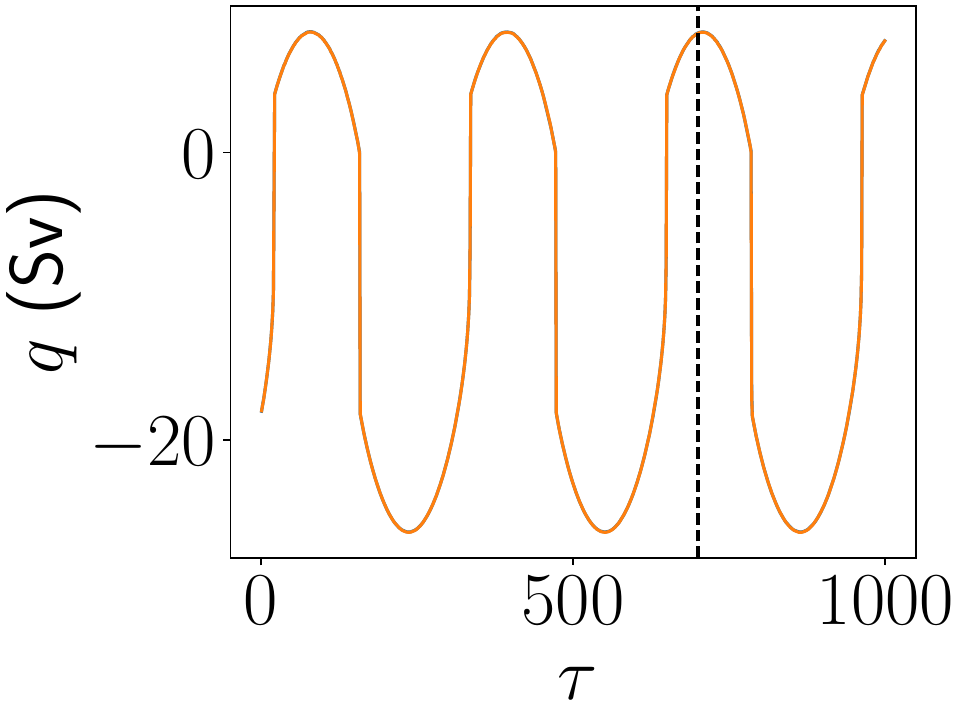} \\
			\includegraphics[width=1.5\linewidth,valign=m]{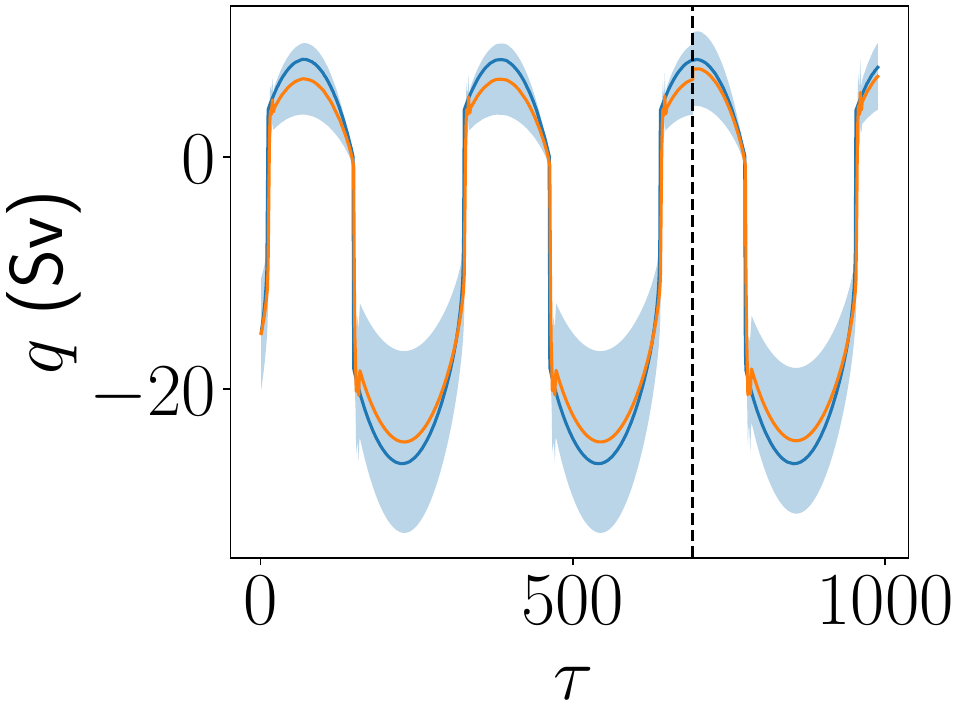}
			& \includegraphics[width=1.5\linewidth,valign=m]{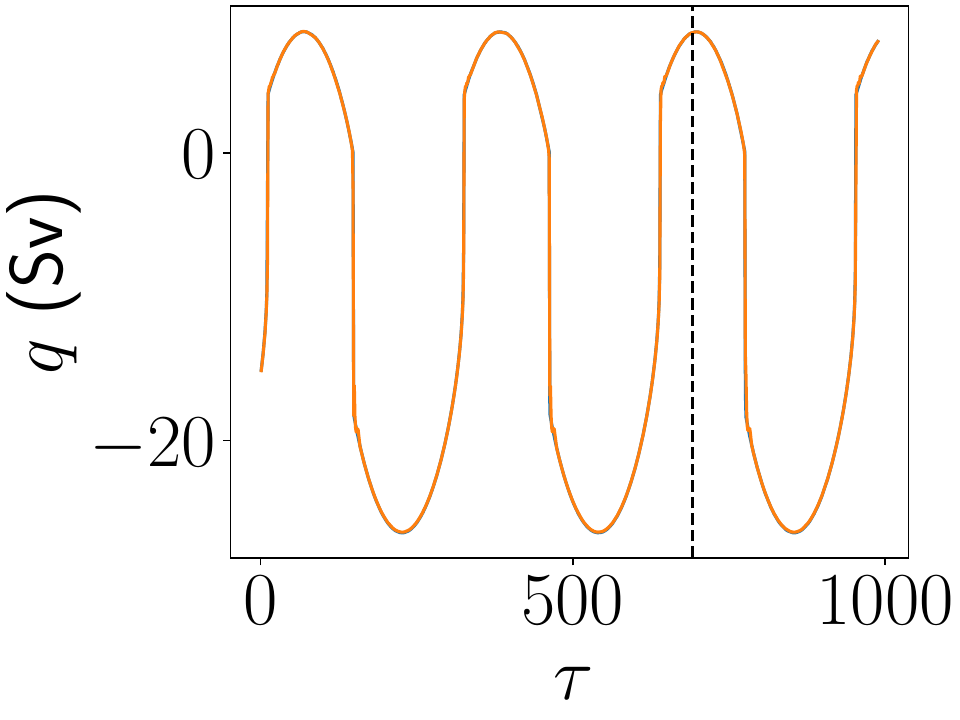}
			& \includegraphics[width=1.5\linewidth,valign=m]{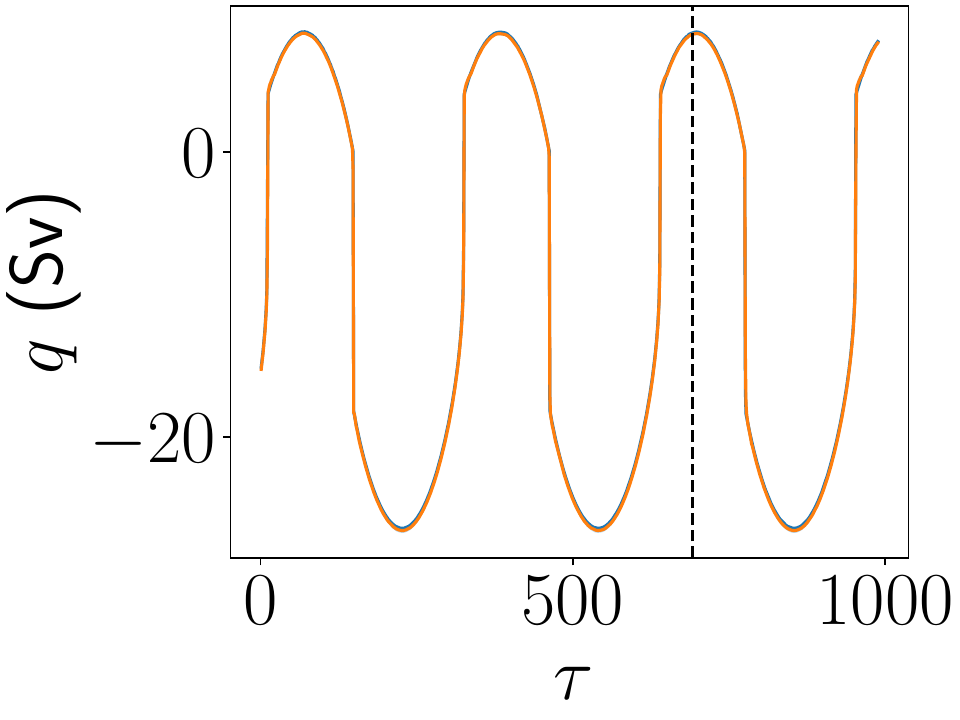} \\

			\multicolumn{3}{c}{\includegraphics[scale=2,valign=m]{legend_pred_gt.pdf}} \\

			\bottomrule
		\end{tabular}
	\end{adjustbox}%
	\caption{Predictive performance for the considered architectures using physics-informed (PI\@; first row) and
	autoregressive (AR\@; second row) features under {\Ftwo}.}%
	\label{tab:F2-performance}%
\end{figure}

\subsection{Detecting Network Understanding vs.\ Imitation}

One question we address is whether \(f_\theta\) learned or imitated the physics of \(q\).
Although there exists no formal method of identifying \(f_\theta\)'s understanding of physics,
we can infer some physical understanding and imitation by identifying cases in which a plausible
physical understanding was learned vs.\ when spurious correlations were learned. Mainly, if \(\Delta S\) or \(S_i\)
receive high absolute attribution scores, we can infer learned physics as well as an attribution
pattern assigning \(S_1\) a score with polarity \(a \in \{-1, 1\}\) and \(S_2\) a score with
polarity \(-a\). This would concur with natural AMOC dynamics; differences in salinity cause
differences in density which then drive \(q\).

\begin{figure}[htb]
	\centering
	\begin{adjustbox}{width=\columnwidth}
		\bgroup%
		\renewcommand{\arraystretch}{0.0}
		\begin{tabular}{cc}
			\toprule
			{\huge MLP} & {\huge DE} \\
			\midrule

			\includegraphics[trim={0 1cm 0 0.8cm},clip,width=\linewidth,valign=m]{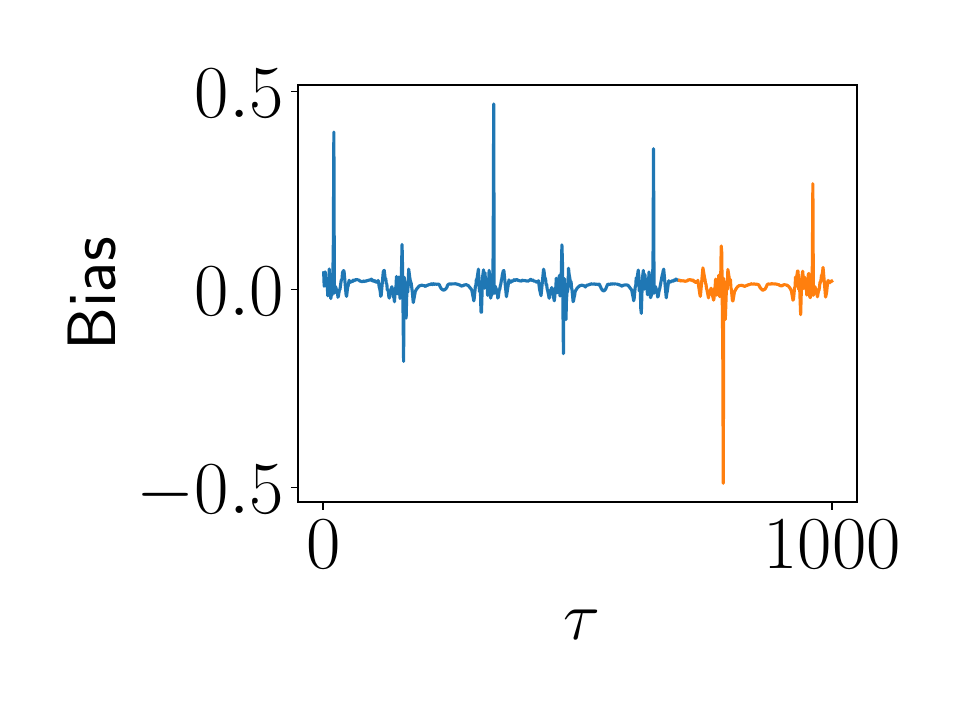}
			& \includegraphics[trim={0 1cm 0 0.8cm},clip,width=\linewidth,valign=m]{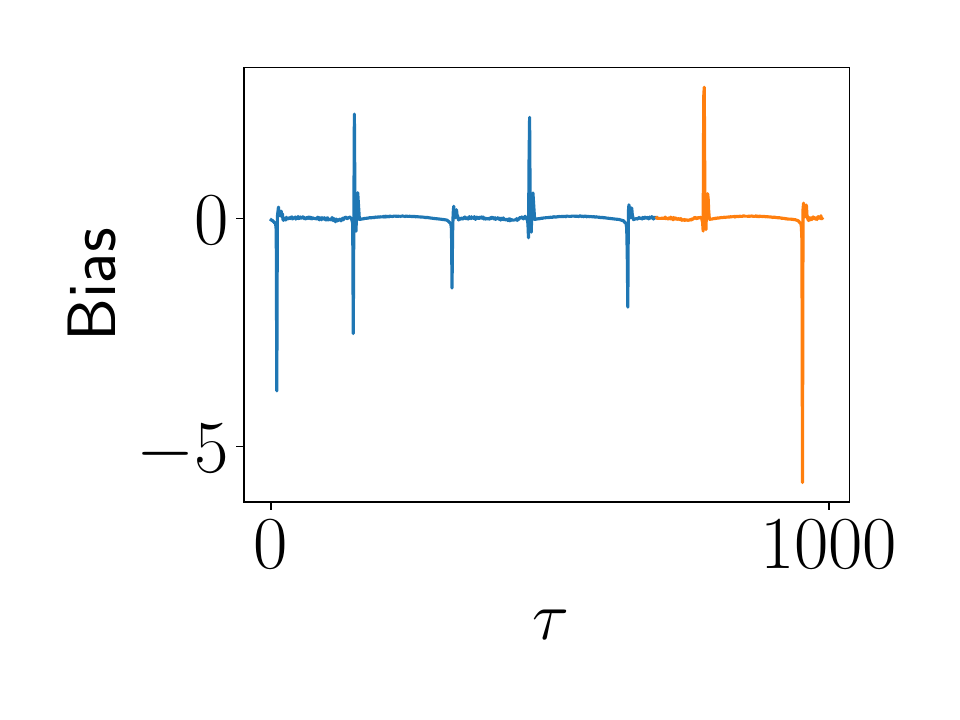} \\

			\includegraphics[trim={0 1cm 0 0.8cm},clip,width=\linewidth,valign=m]{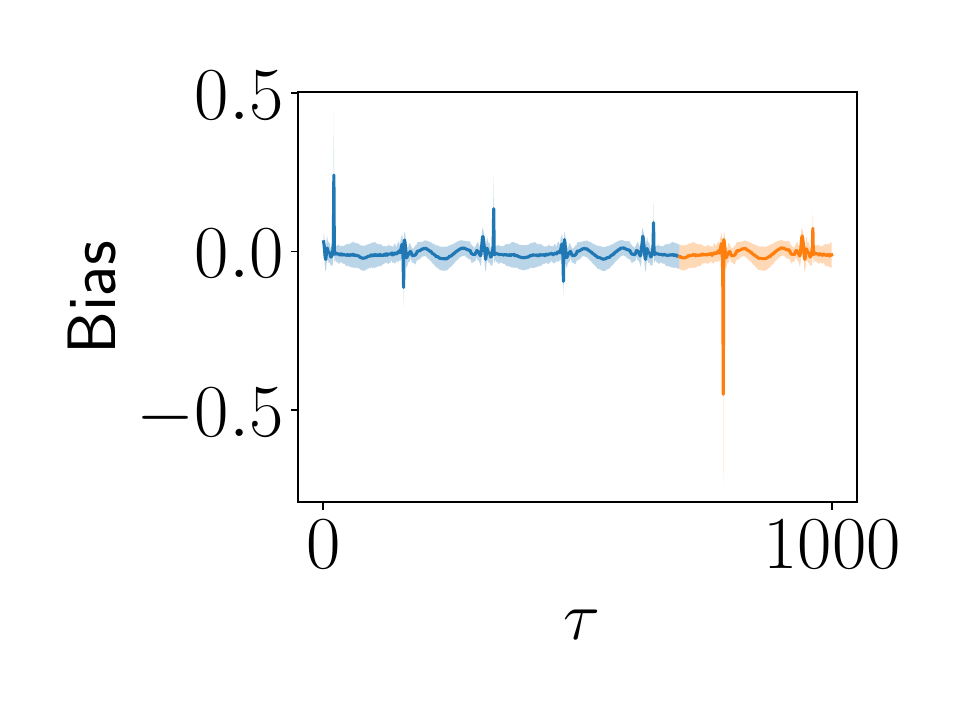}
			& \includegraphics[trim={0 1cm 0 0.8cm},clip,width=\linewidth,valign=m]{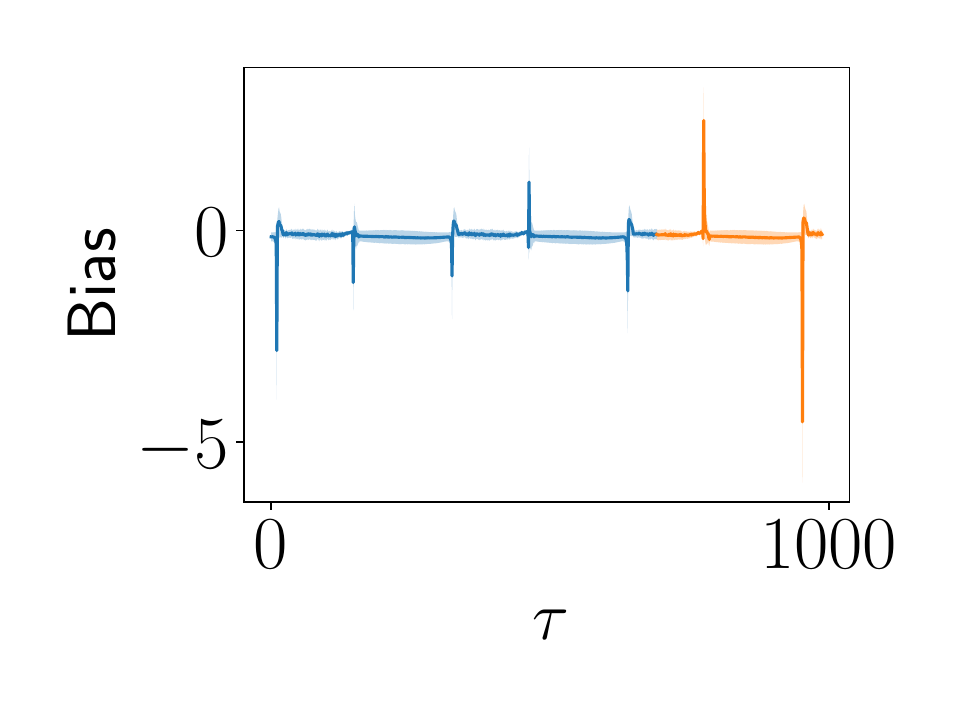} \\

			\multicolumn{2}{c}{\includegraphics[valign=m]{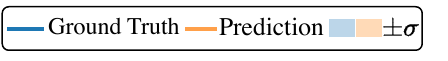}} \\

			\bottomrule
		\end{tabular}%
		\egroup%
	\end{adjustbox}%
	\caption{Bias (\(\hat{q}_\tau - q_\tau\)) for the MLP and DE architectures under {\Ftwo}.}%
	\label{tab:F2-bias}%
\end{figure}

\begin{figure*}[htb]
	\centering
	\bgroup%
	\newcommand{\scale}{0.15}
	\begin{tabular}{ccc}
		\toprule
		BNN & MLP & DE \\
		\midrule

		\includegraphics[scale=\scale,valign=m]{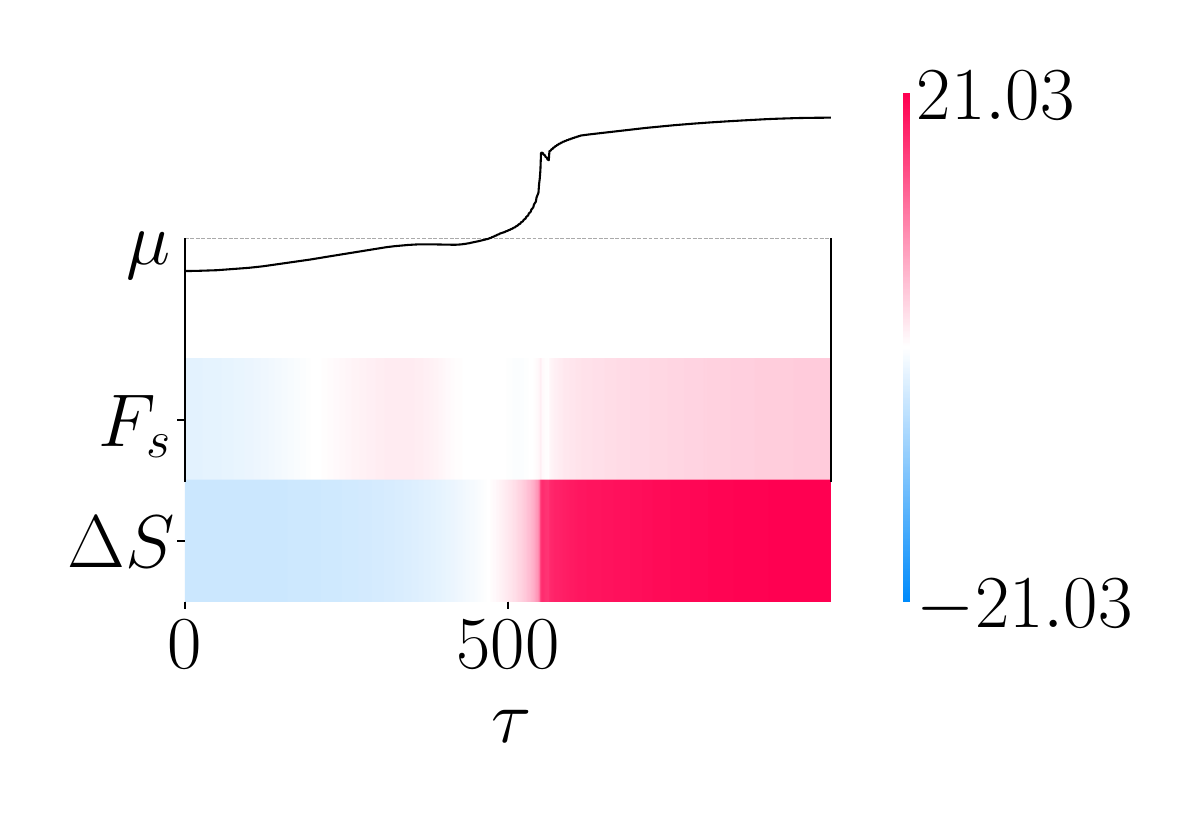}
		& \includegraphics[scale=\scale,valign=m]{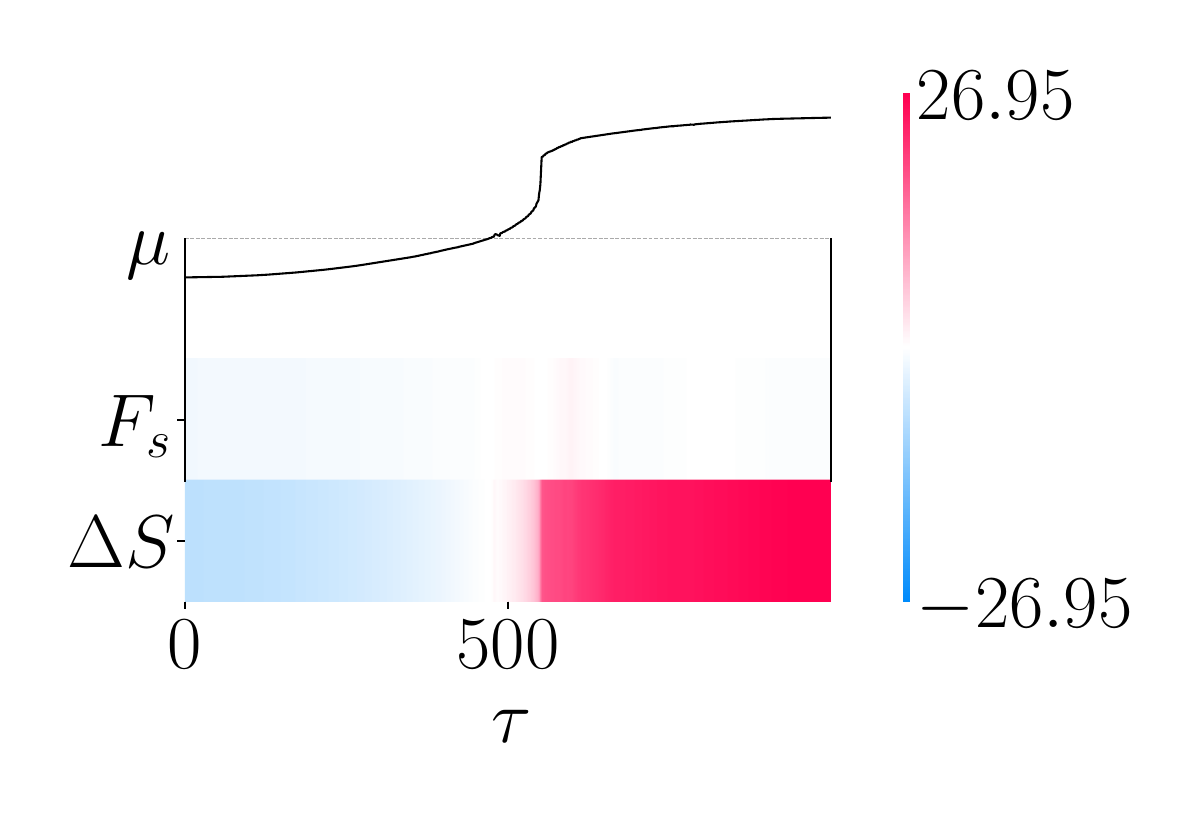}
		& \includegraphics[scale=\scale,valign=m]{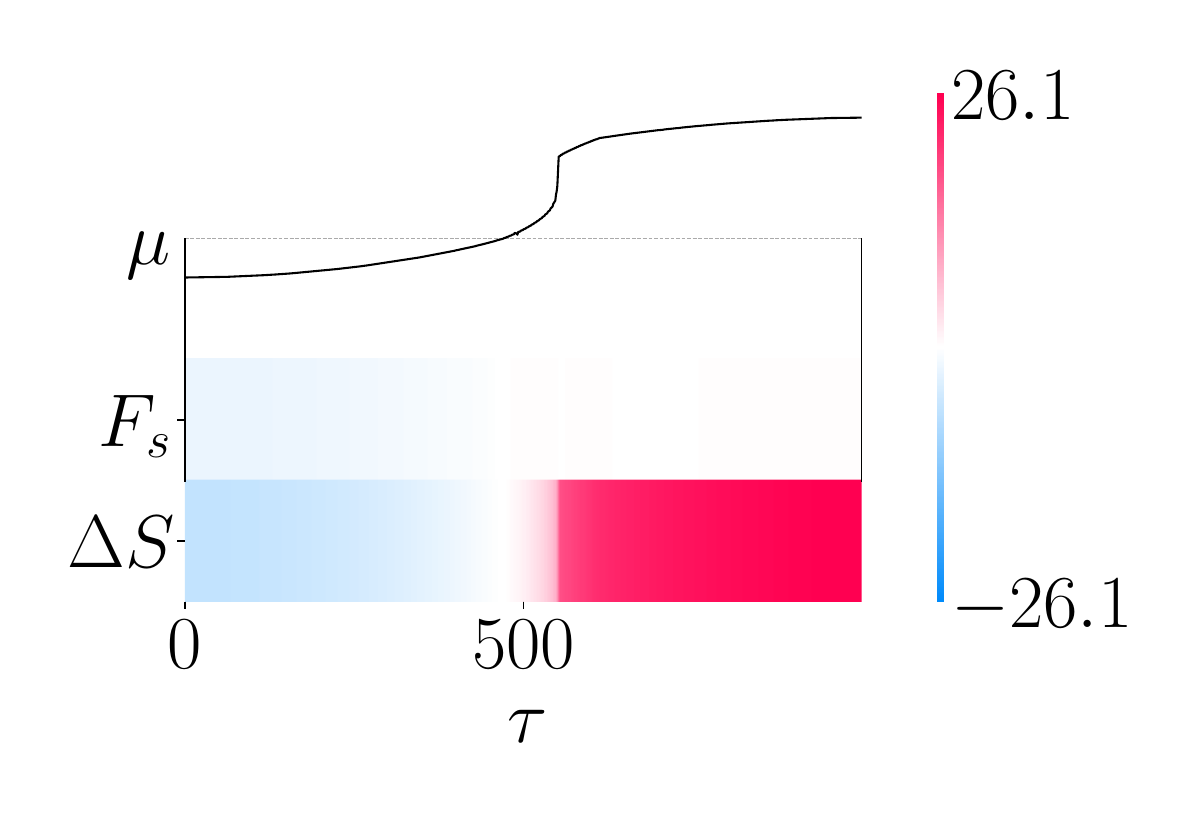} \\
		\includegraphics[trim={0 4cm 0 7cm},clip,scale=\scale,valign=m]{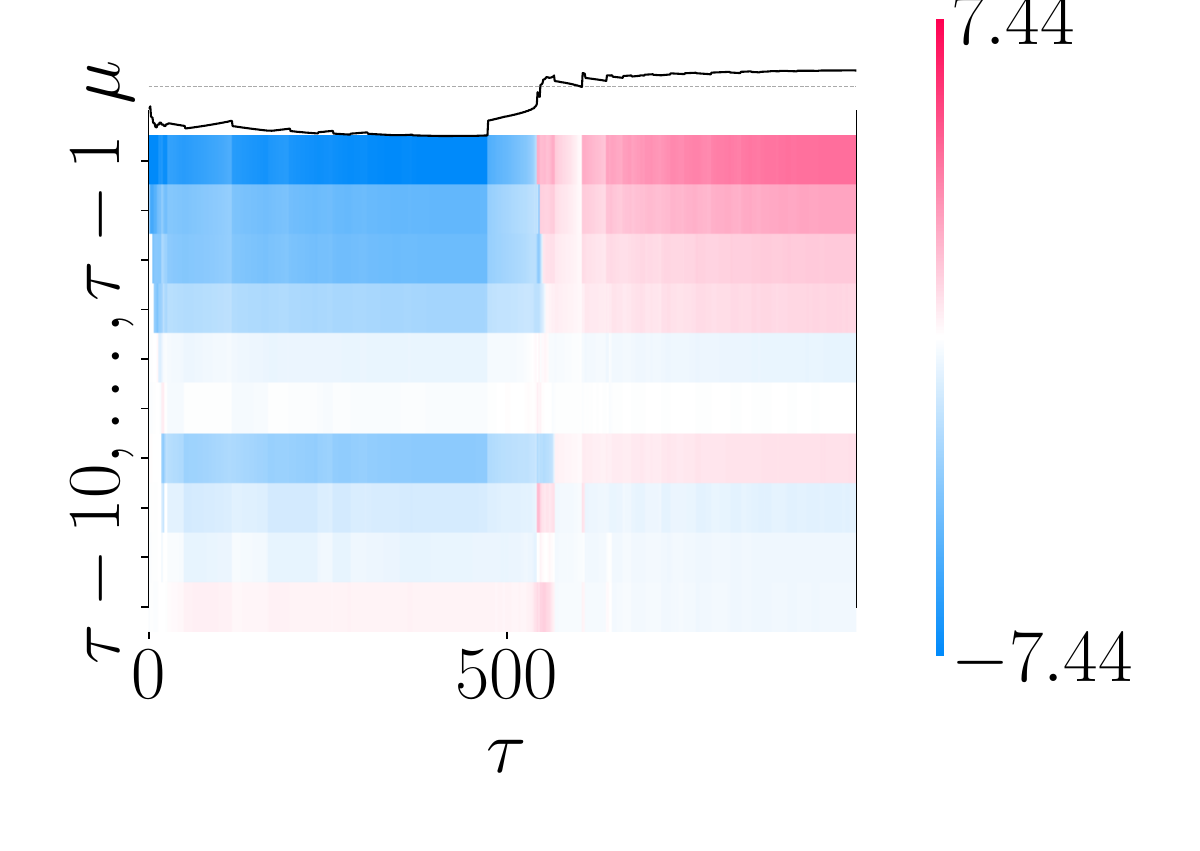}
		& \includegraphics[trim={0 4cm 0 7cm},clip,scale=\scale,valign=m]{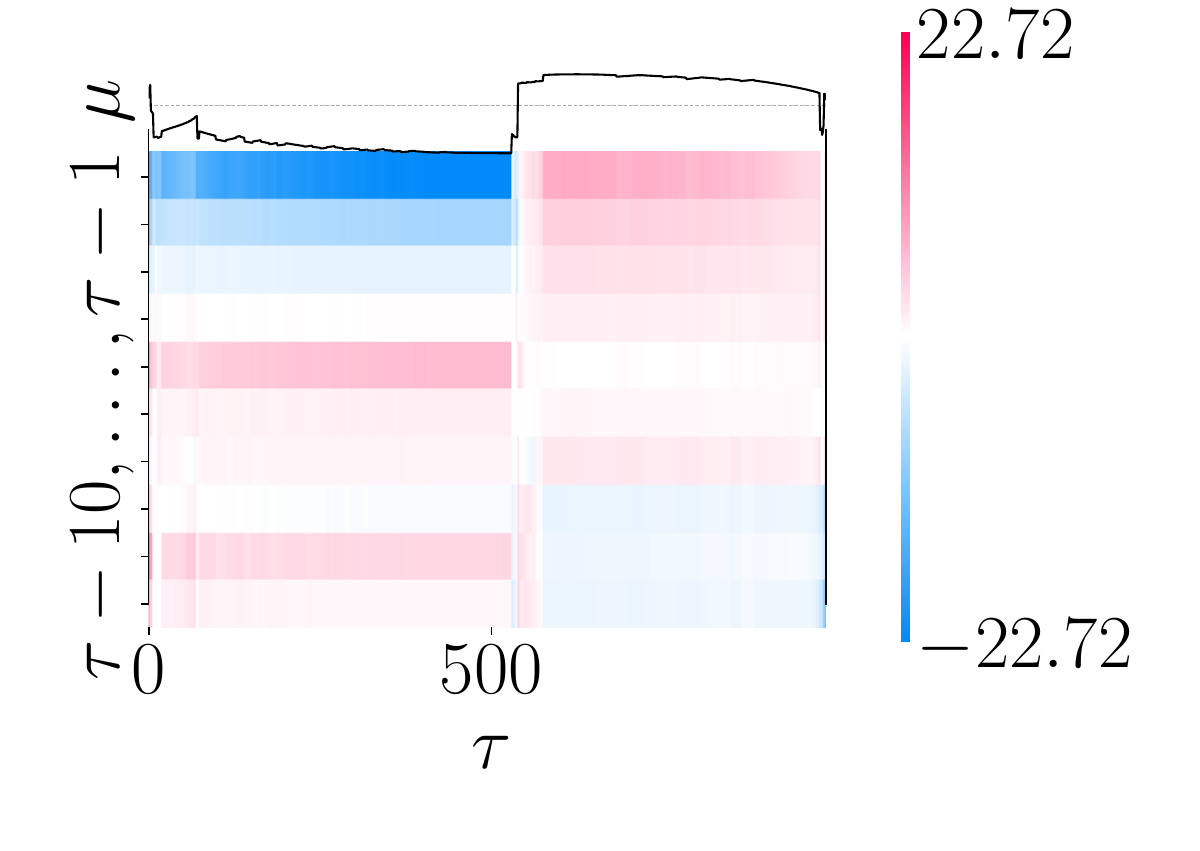}
		& \includegraphics[trim={0 4cm 0 7cm},clip,scale=\scale,valign=m]{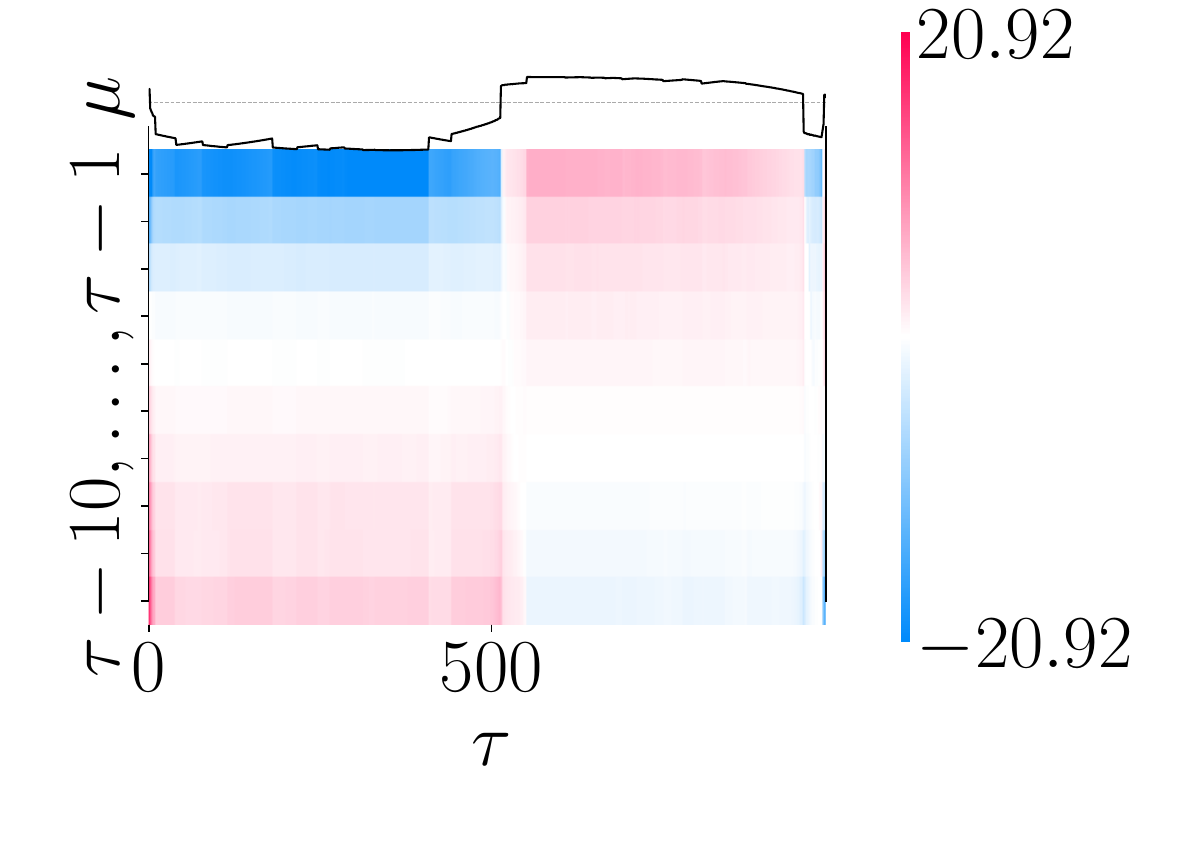} \\

		\bottomrule
	\end{tabular}
	\egroup%
	\caption{DeepLIFT attribution maps for the considered architectures using physics-informed (PI\@; top row) and
	autoregressive (AR\@; bottom row) features under {\Ftwo}.}%
	\label{tab:F2-xai}%
\end{figure*}

\section{Experiments}

\subsection{Experimental Design}

We aim to understand two perspectives: how does the AMOC evolve according to
the physical ruleset proposed in \cref{sec:modeling-q} vs.\ autoregressively
where we only use measured data that is sampled from an externally observable
physical process? On the macro level, we have the difference between a linear
box model (\(\rho^{\text{lin}}\)) and a nonlinear one (\(\rho^{\text{EOS-80}}\))
which yields a more accurate but difficult-to-predict AMOC representation.

\begin{figure*}[htb]
	\centering
	\bgroup%
	\newcommand{\scale}{0.15}
		\begin{tabular}{cccc}
			\toprule
			Forcing Setup & Forcing & Variables & \(q\) \\
			\midrule

			& \includegraphics[scale=0.5,valign=m]{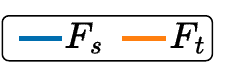}
			& \includegraphics[scale=0.3,valign=m]{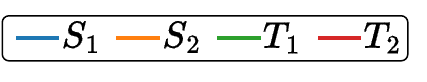}
			& \\

			{\Ffour}
			& \includegraphics[scale=\scale,valign=m]{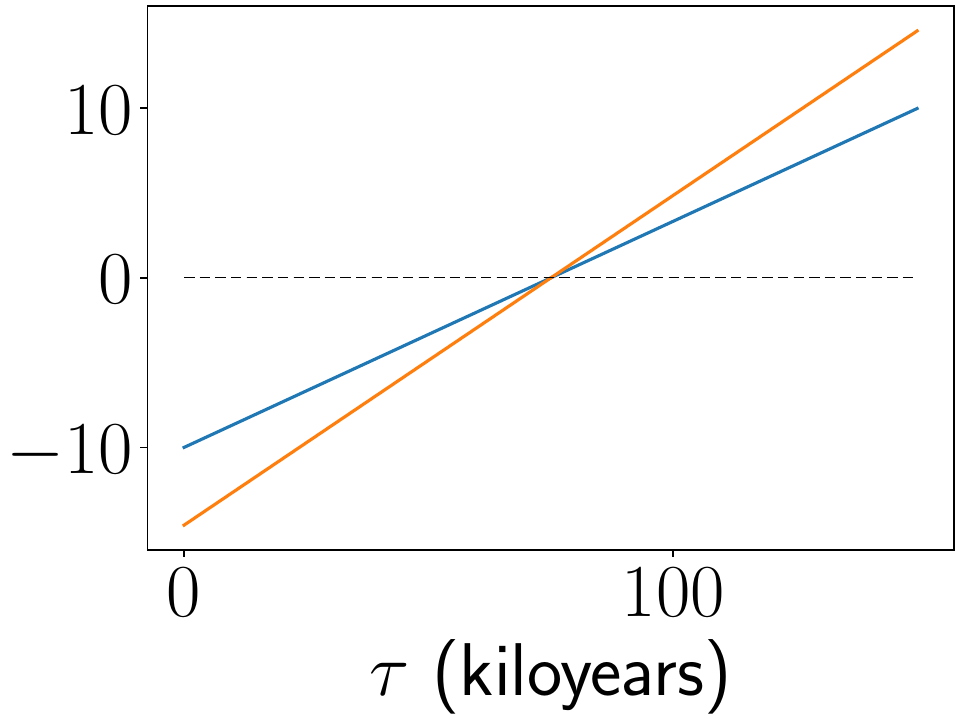}
			& \includegraphics[scale=\scale,valign=m]{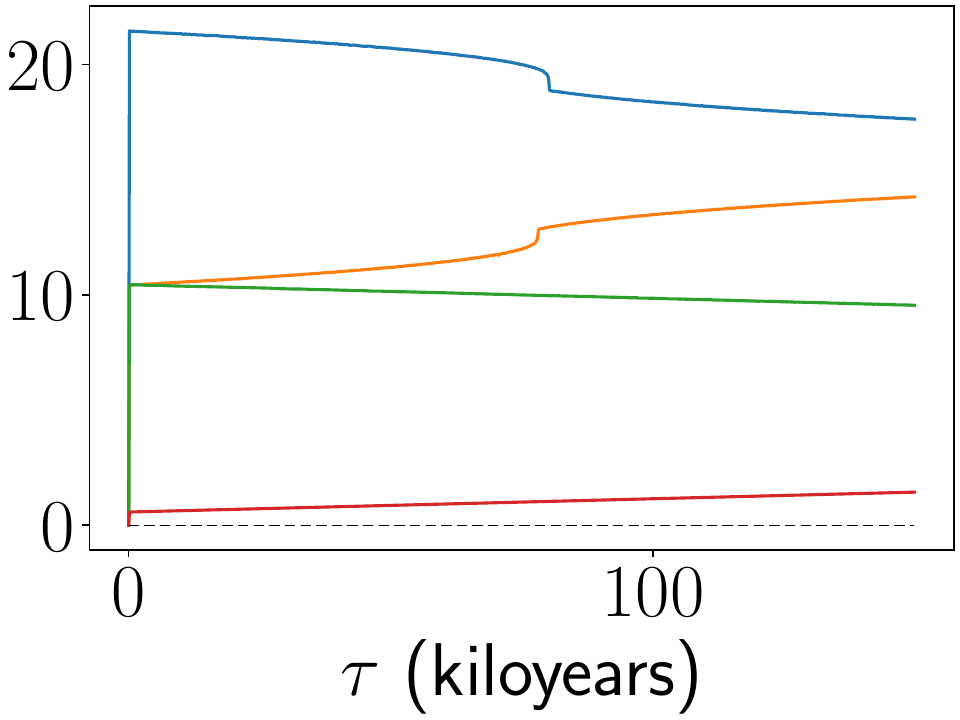}
			& \includegraphics[scale=\scale,valign=m]{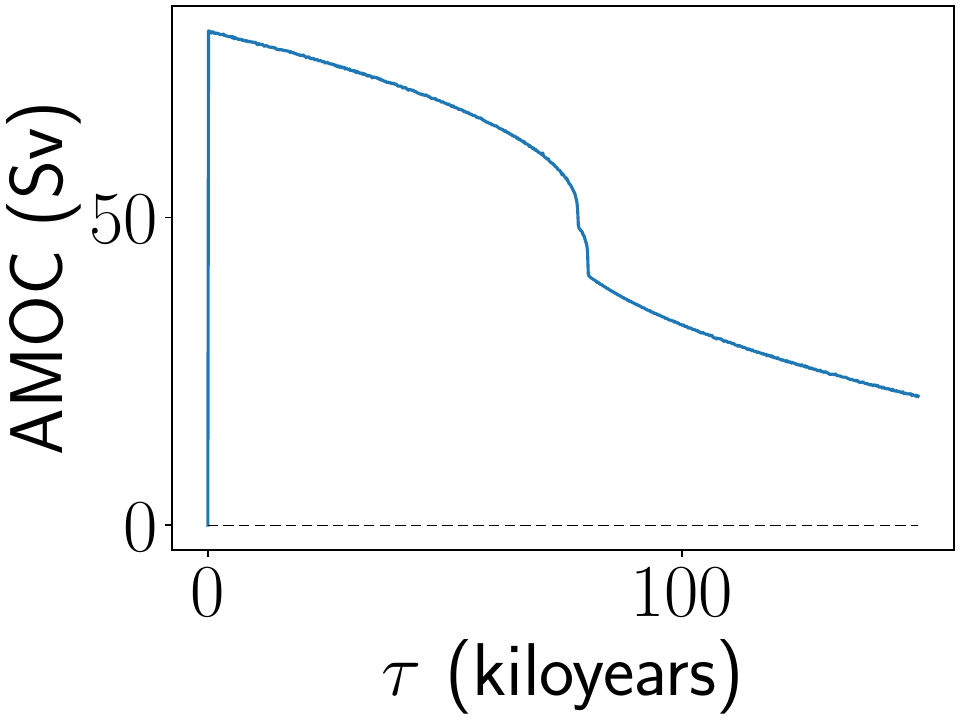} \\
			{\Ffive}
			& \includegraphics[scale=\scale,valign=m]{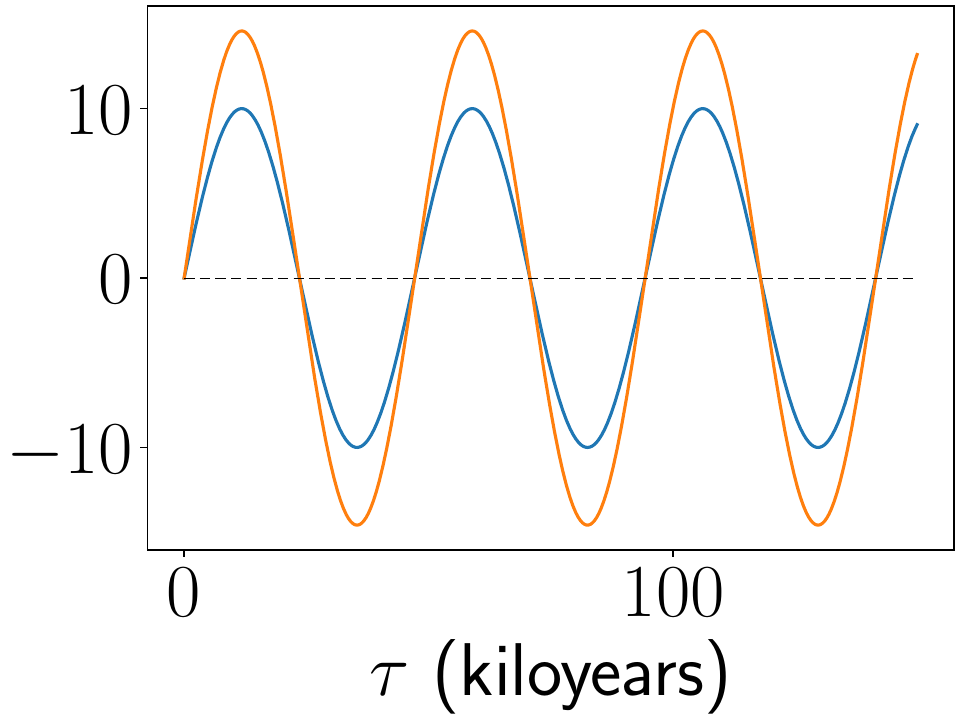}
			& \includegraphics[scale=\scale,valign=m]{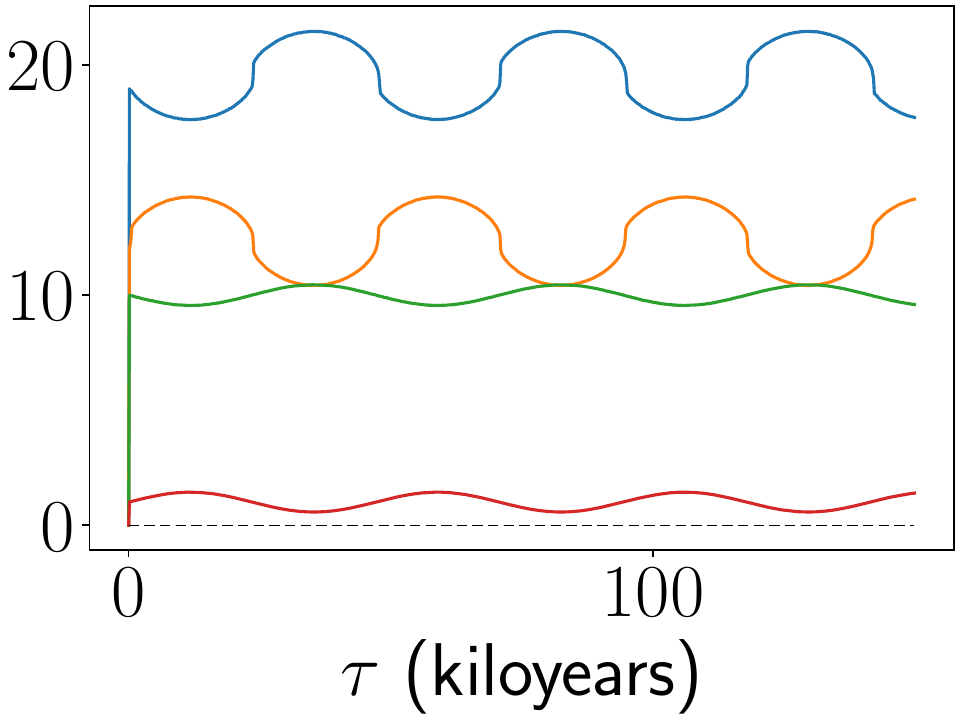}
			& \includegraphics[scale=\scale,valign=m]{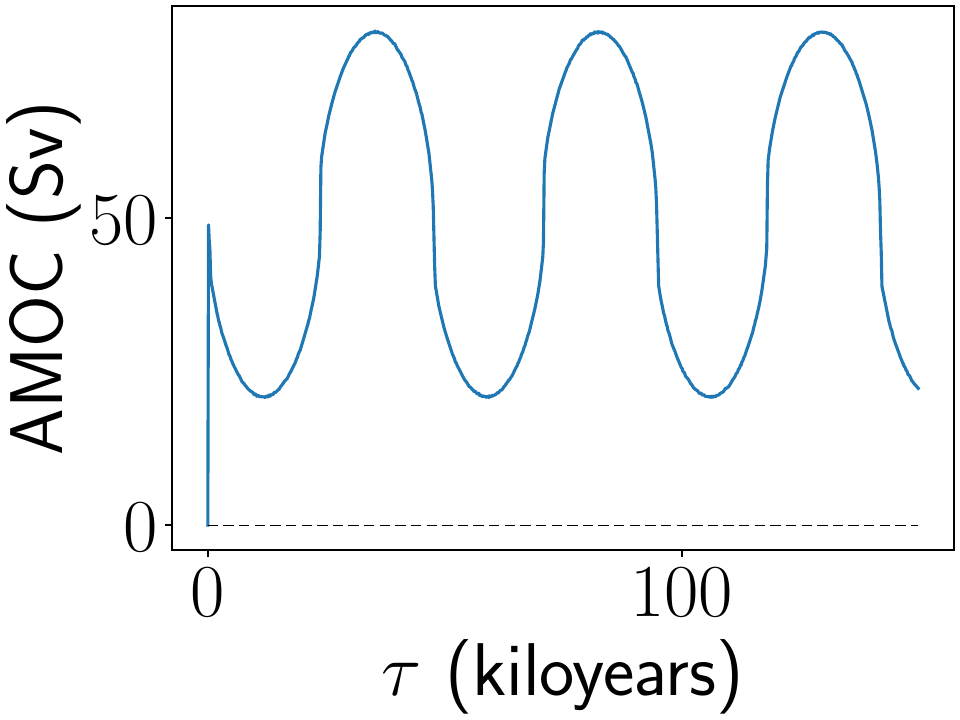} \\
			{\Fsix}
			& \includegraphics[scale=\scale,valign=m]{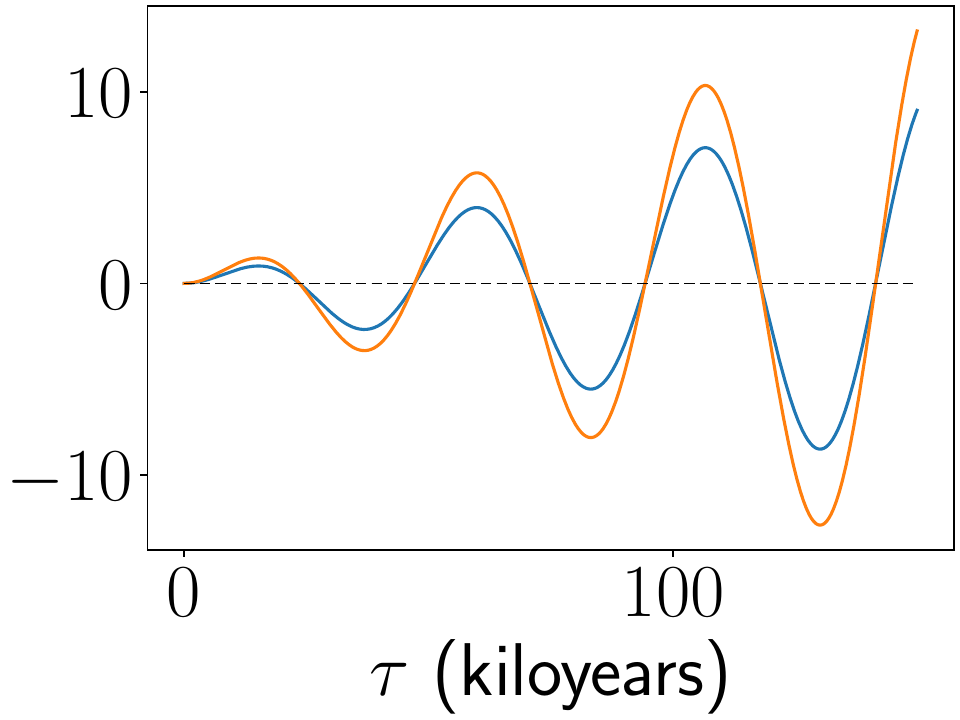}
			& \includegraphics[scale=\scale,valign=m]{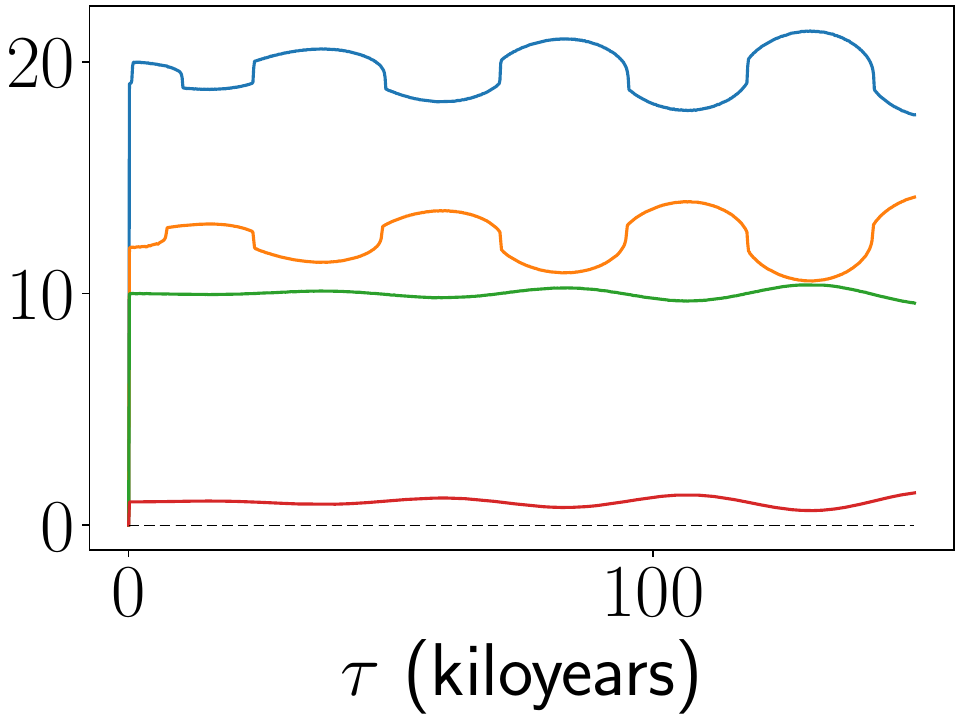}
			& \includegraphics[scale=\scale,valign=m]{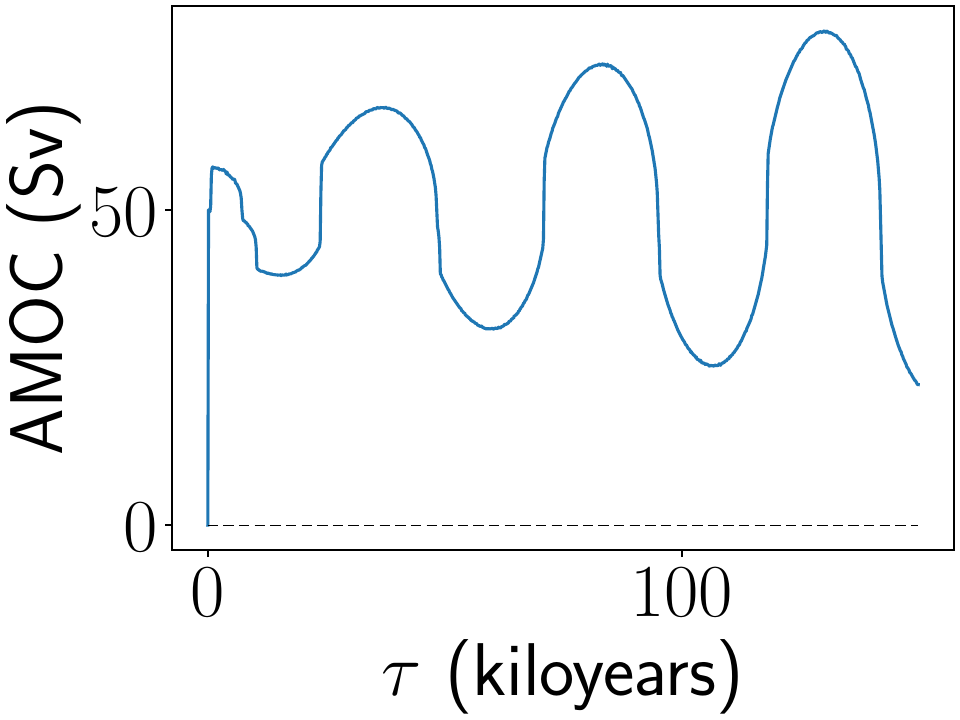} \\

			\bottomrule
		\end{tabular}
	\egroup%
	\caption{Forcing trajectory, physical variables and AMOC for {\Ffour}, {\Ffive} and {\Fsix}.}%
	\label{tab:F4-F5-F6}%
\end{figure*}

In \cref{tab:forcing-setups}, we list the combinations of forcings we use in
our experiments. For the standard box model we set \(\rho = \rho^{\text{lin}}\)
and include a linear and two sinusoidal forcings with one nonstationary forcing.
For the extended box model, we consider the same forcings in conjunction with
\(F_t\) and \(\rho = \rho^{\text{EOS-80}}\). All forcings with their corresponding
\({\Delta \{S, T\}}\) and \(\{S_i, T_i\}\) including the resulting AMOC are plotted
in \cref{tab:F1-F2-F3,tab:F4-F5-F6}, respectively. We train all models for 120
epochs using the Adam optimizer~\cite{kingma2014adam}.

\subsection{Results}

\begin{figure}[htb]
	\centering
	\begin{adjustbox}{width=\columnwidth}
		\begin{tabular}{ccc}
			\toprule
			{\Huge BNN} & {\Huge MLP} & {\Huge DE} \\
			\midrule

			\includegraphics[width=1.5\linewidth,valign=m]{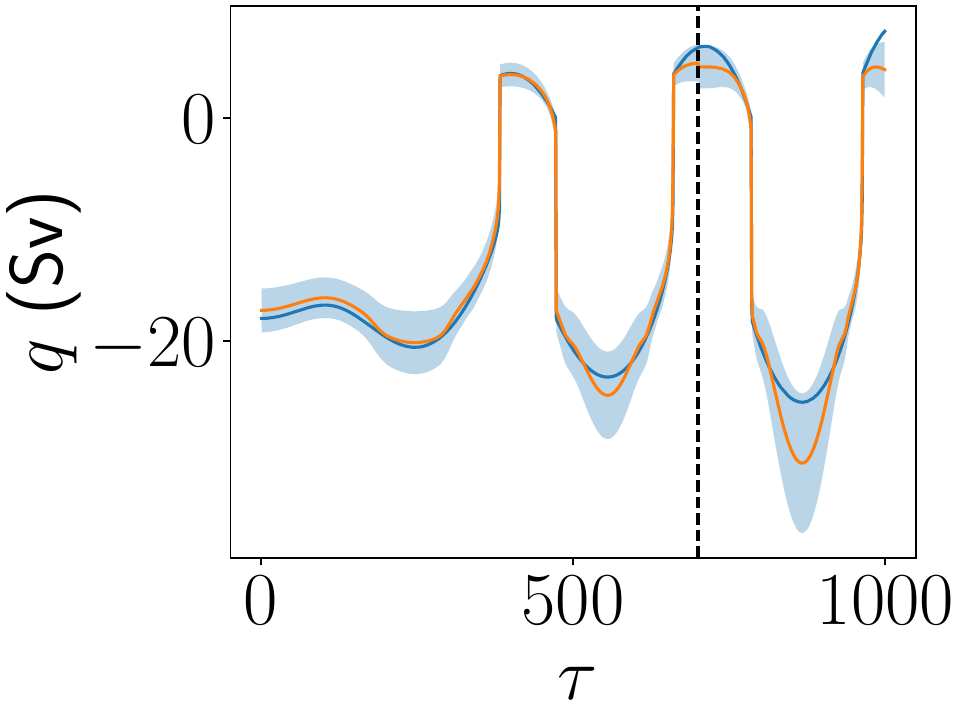}
			& \includegraphics[width=1.5\linewidth,valign=m]{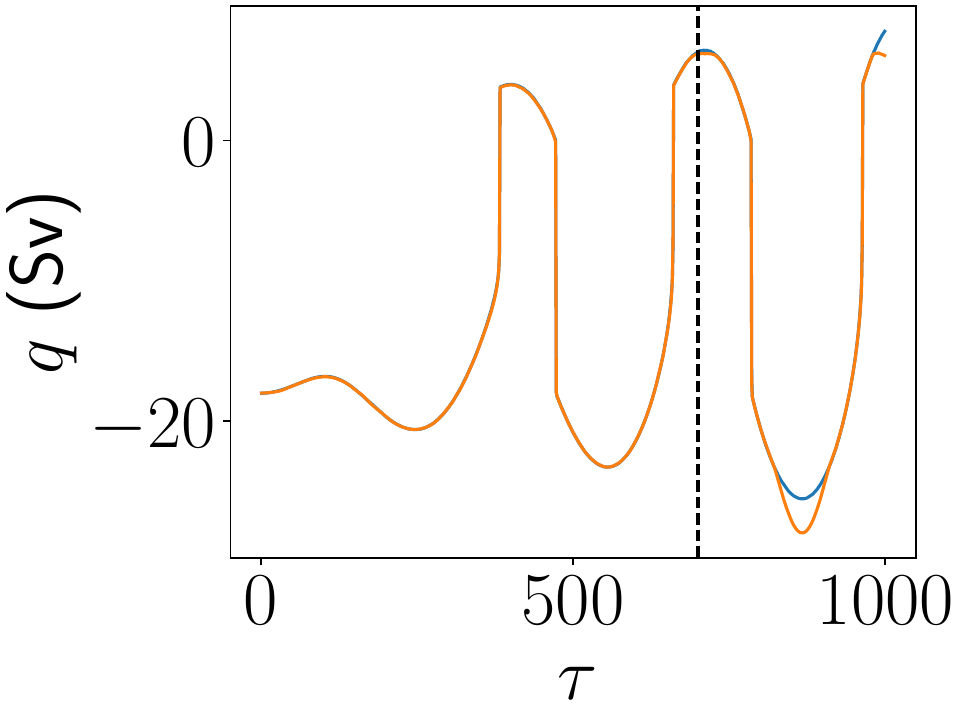}
			& \includegraphics[width=1.5\linewidth,valign=m]{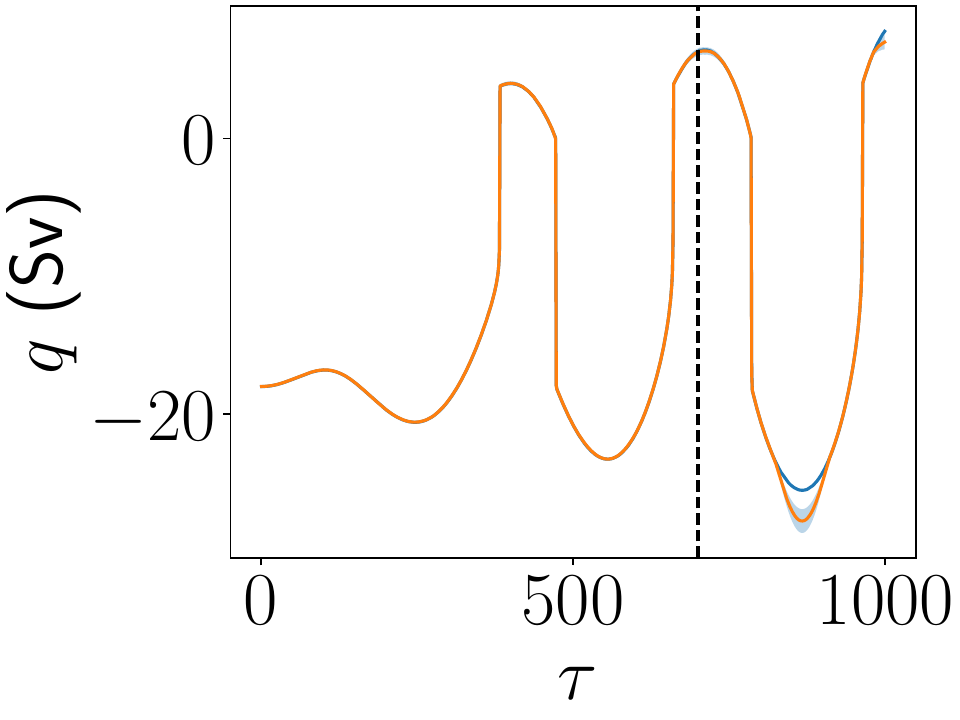} \\
			\includegraphics[width=1.5\linewidth,valign=m]{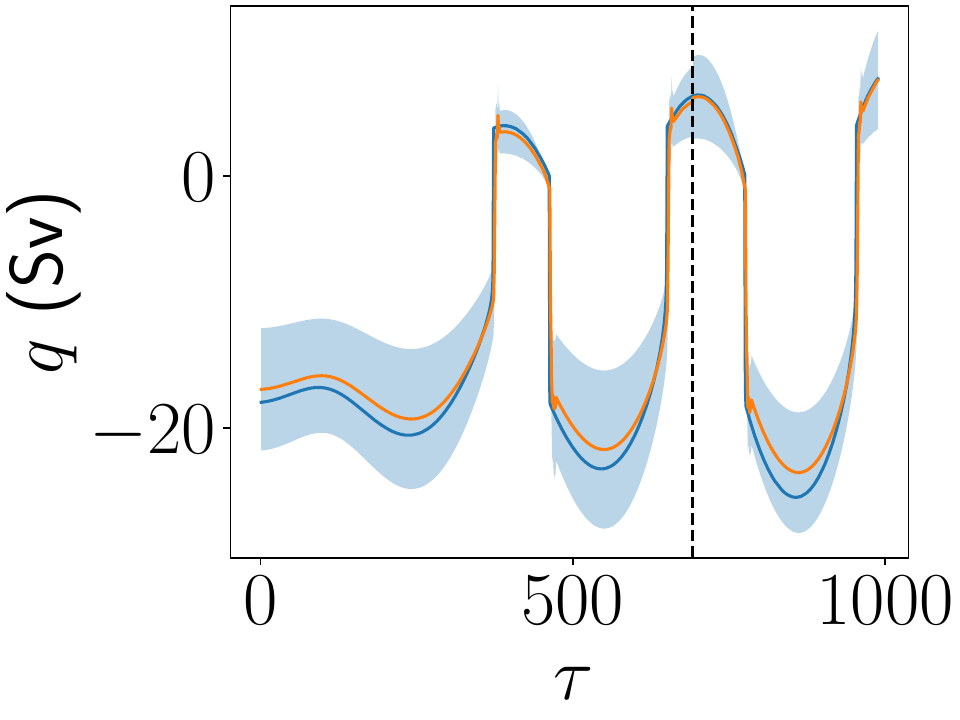}
			& \includegraphics[width=1.5\linewidth,valign=m]{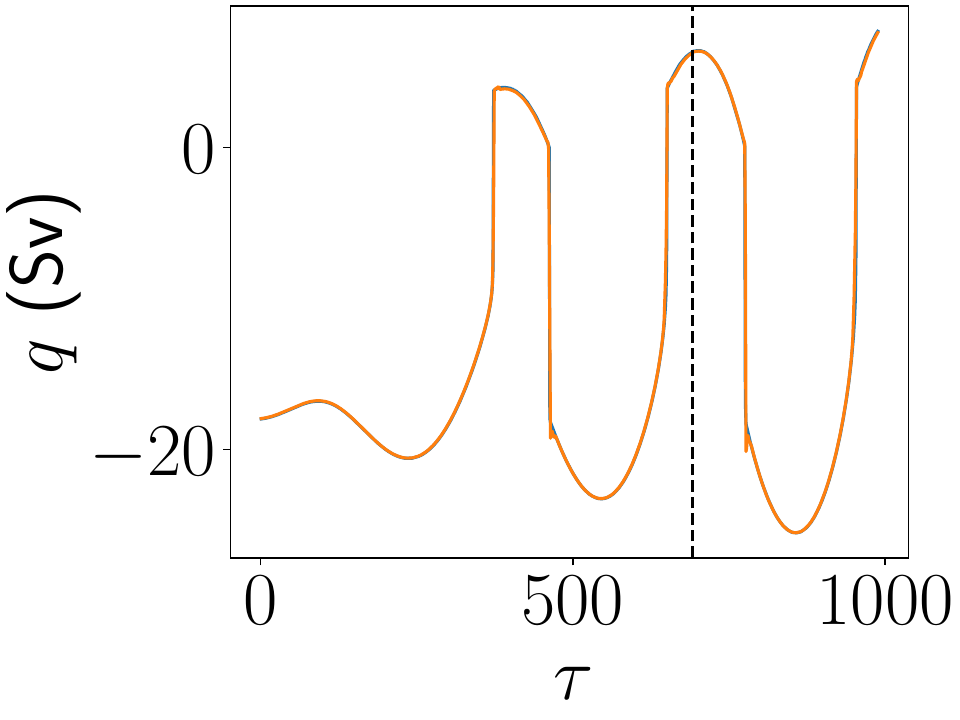}
			& \includegraphics[width=1.5\linewidth,valign=m]{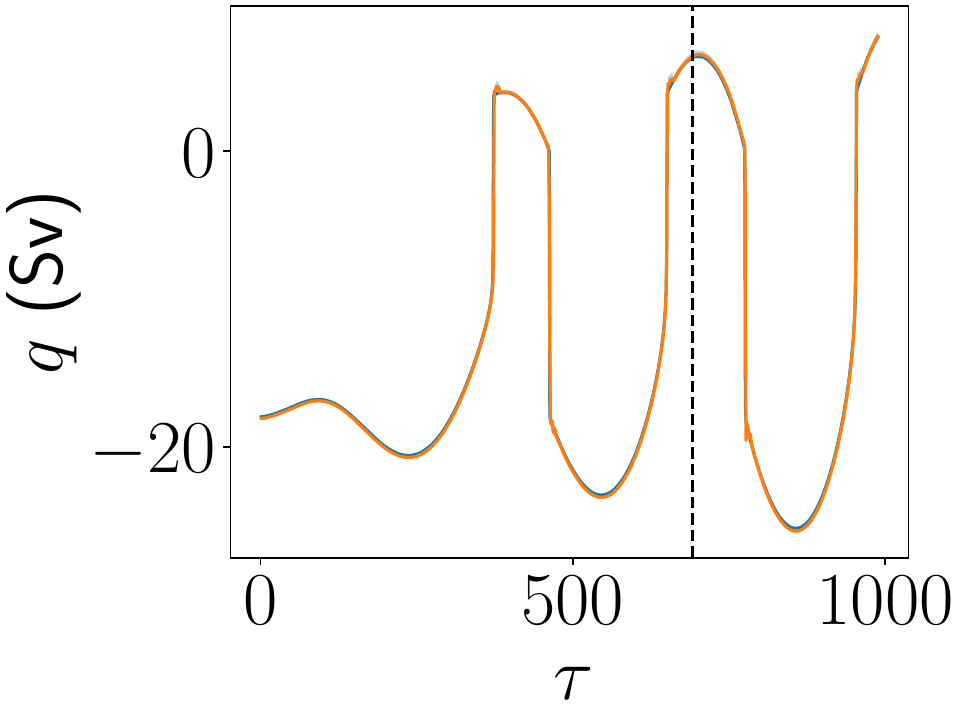} \\

			\multicolumn{3}{c}{\includegraphics[scale=2,valign=m]{legend_pred_gt.pdf}} \\

			\bottomrule
		\end{tabular}
	\end{adjustbox}%
	\caption{Predictive performance for the considered architectures using physics-informed (PI\@; first row) and
	autoregressive (AR\@; second row) features under {\Fthree}.}%
	\label{tab:F3-performance}%
\end{figure}

\begin{figure}[htb]
	\centering
	\begin{adjustbox}{width=\columnwidth}
	\begin{tabular}{cc}
		\toprule
		{\huge MLP} & {\huge DE} \\
		\midrule

		\includegraphics[trim={0 1cm 0 0.8cm},clip,width=\linewidth,valign=m]{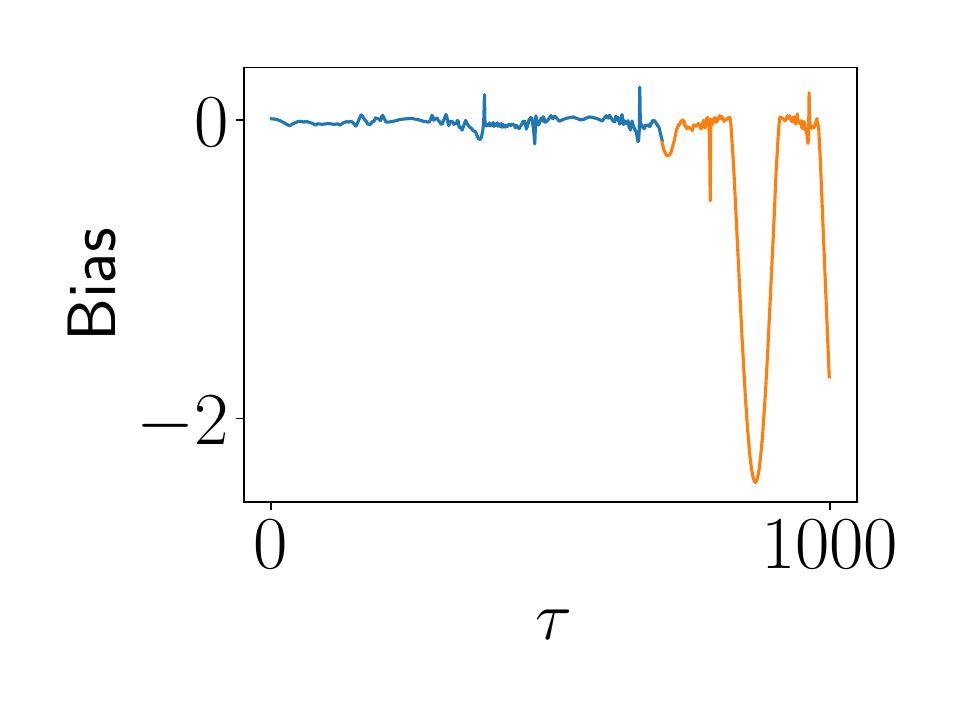}
		& \includegraphics[trim={0 1cm 0 0.8cm},clip,width=\linewidth,valign=m]{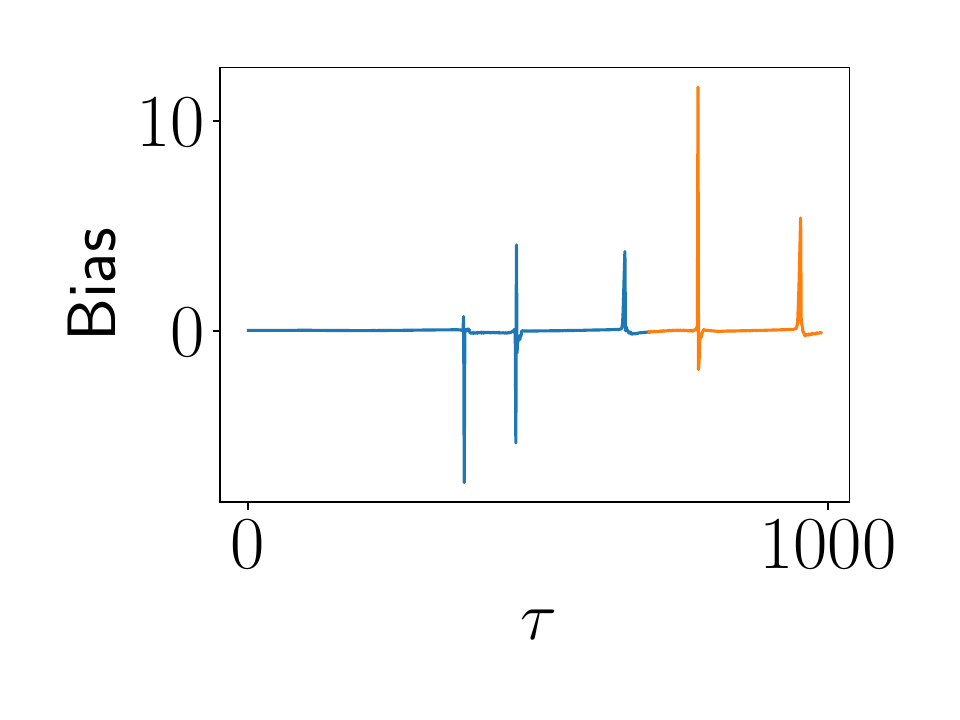} \\

		\includegraphics[trim={0 1cm 0 0.8cm},clip,width=\linewidth,valign=m]{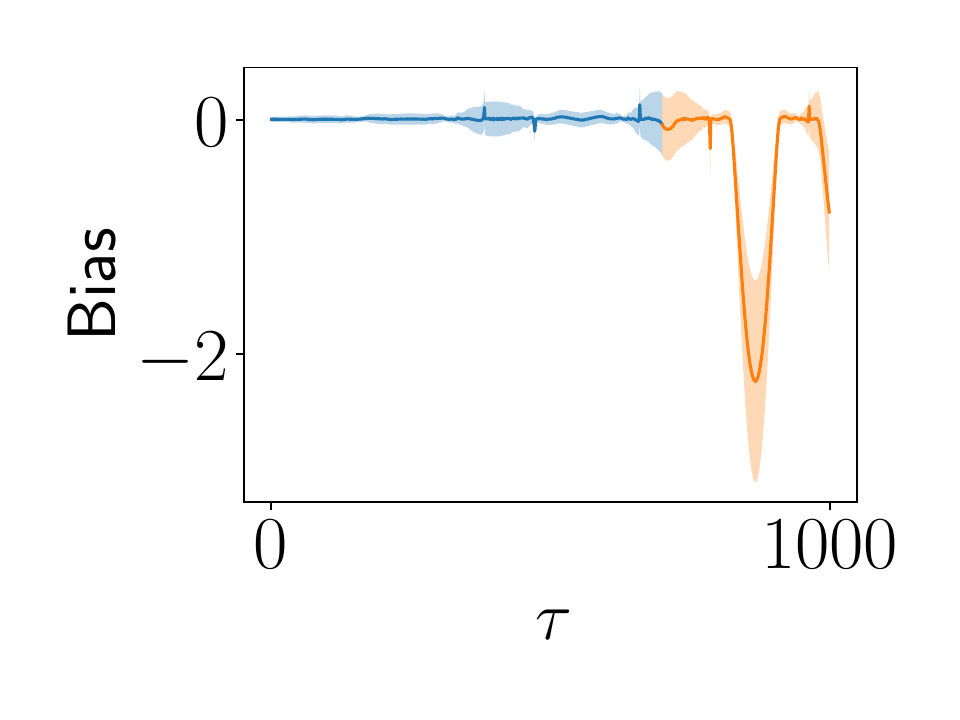}
		& \includegraphics[trim={0 1cm 0 0.8cm},clip,width=\linewidth,valign=m]{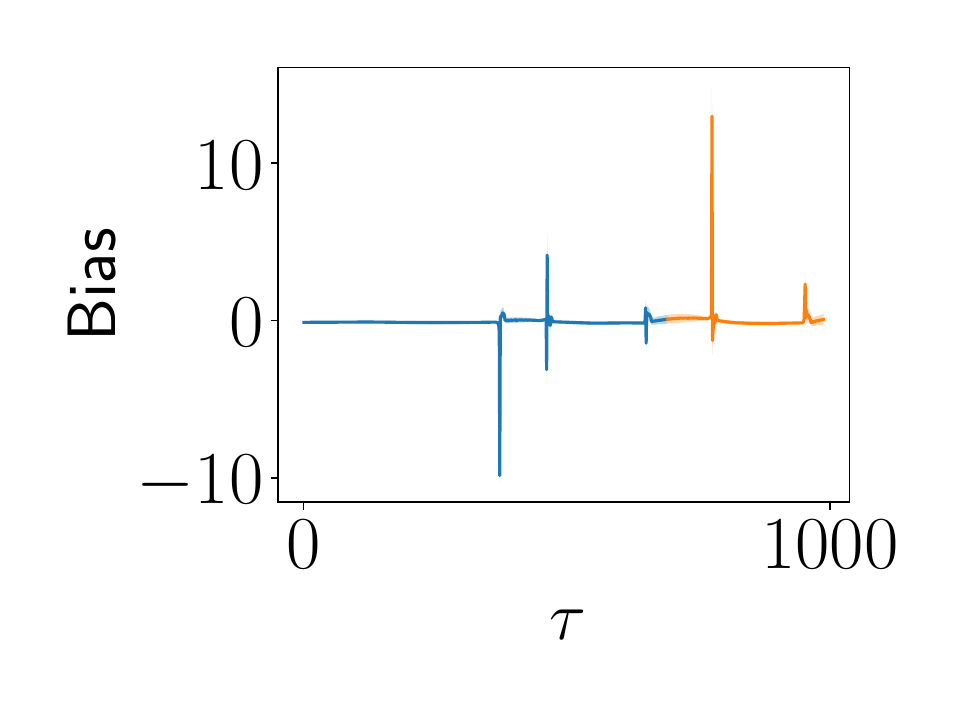} \\

		\multicolumn{2}{c}{\includegraphics[valign=m]{legend_bias.pdf}} \\

		\bottomrule
	\end{tabular}%
	\end{adjustbox}%
	\caption{Bias (\(\hat{q}_\tau - q_\tau\)) for the MLP and DE architectures under {\Fthree}.}%
	\label{tab:F3-bias}%
\end{figure}

\begin{figure*}[htb]
	\centering
	\bgroup%
	\newcommand{\scale}{0.2}
	\begin{tabular}{ccc}
		\toprule
		BNN & MLP & DE \\
		\midrule

		\includegraphics[scale=\scale,valign=m]{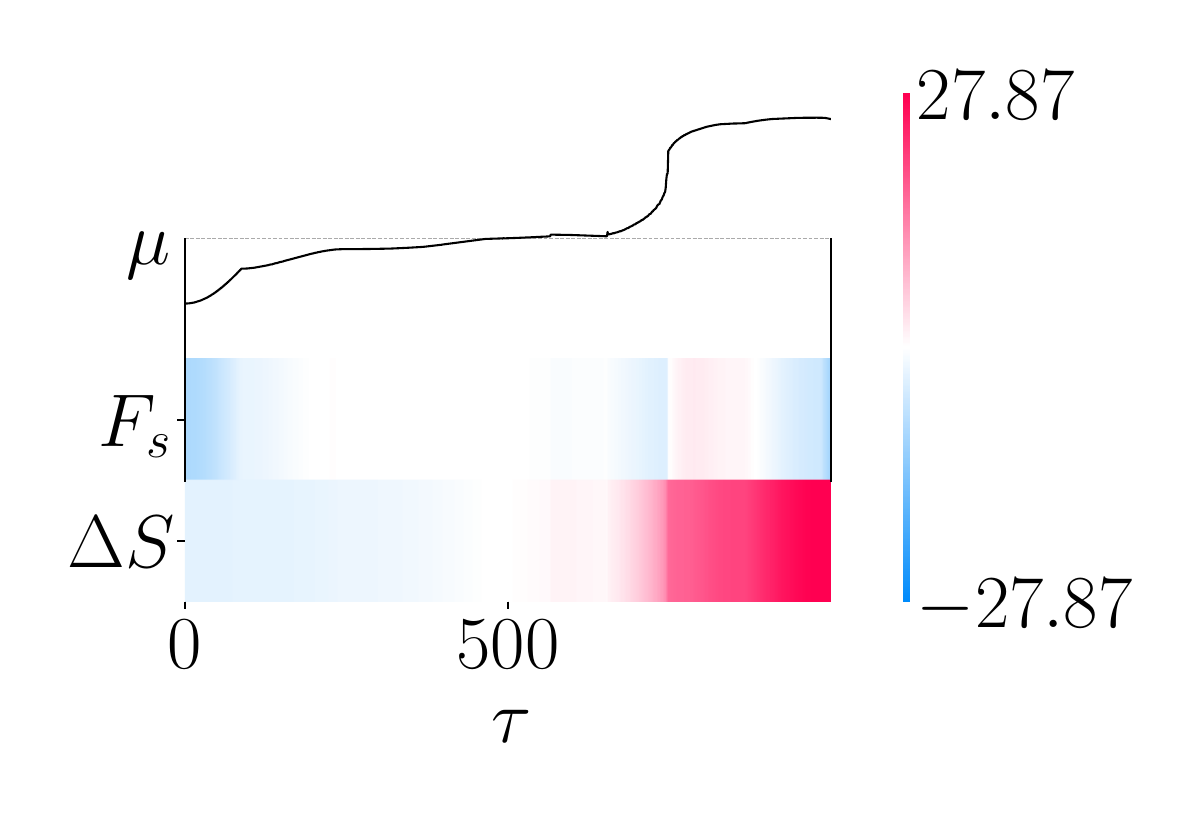}
		& \includegraphics[scale=\scale,valign=m]{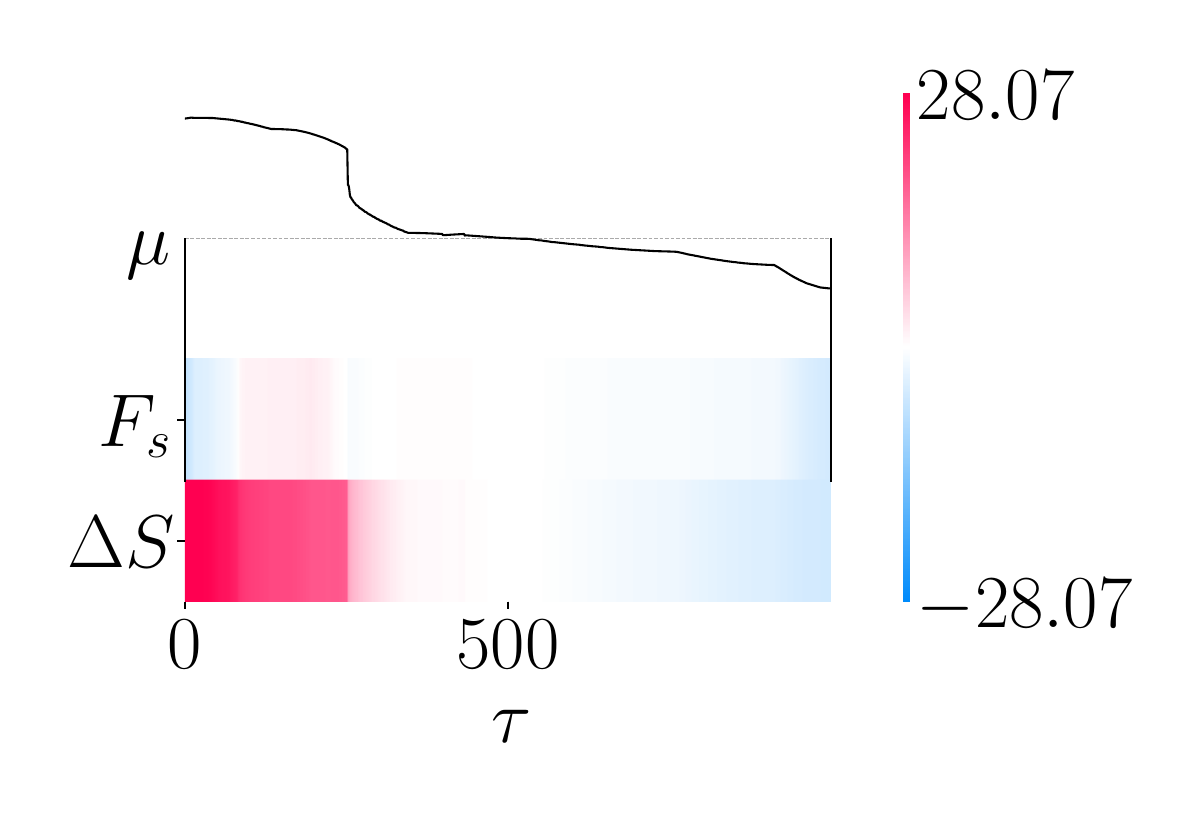}
		& \includegraphics[scale=\scale,valign=m]{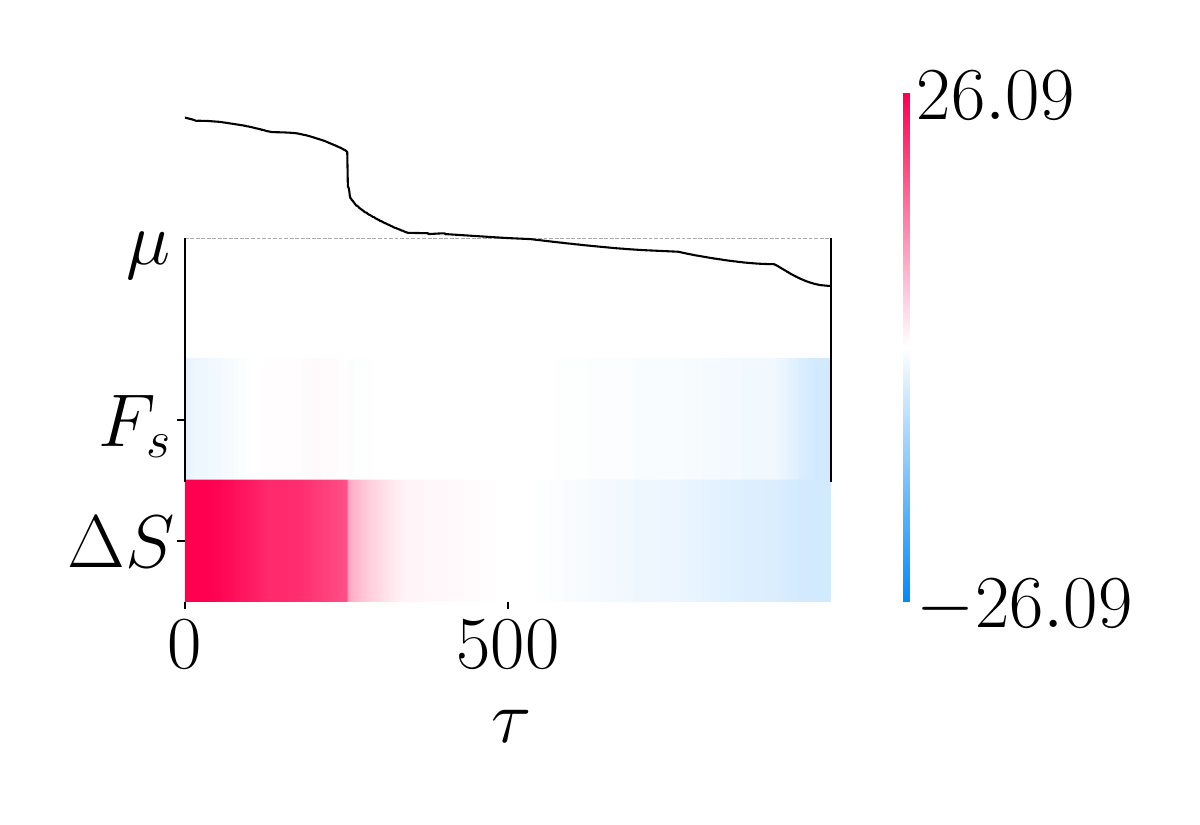} \\

		\includegraphics[trim={0 4cm 0 6cm},clip,scale=\scale,valign=m]{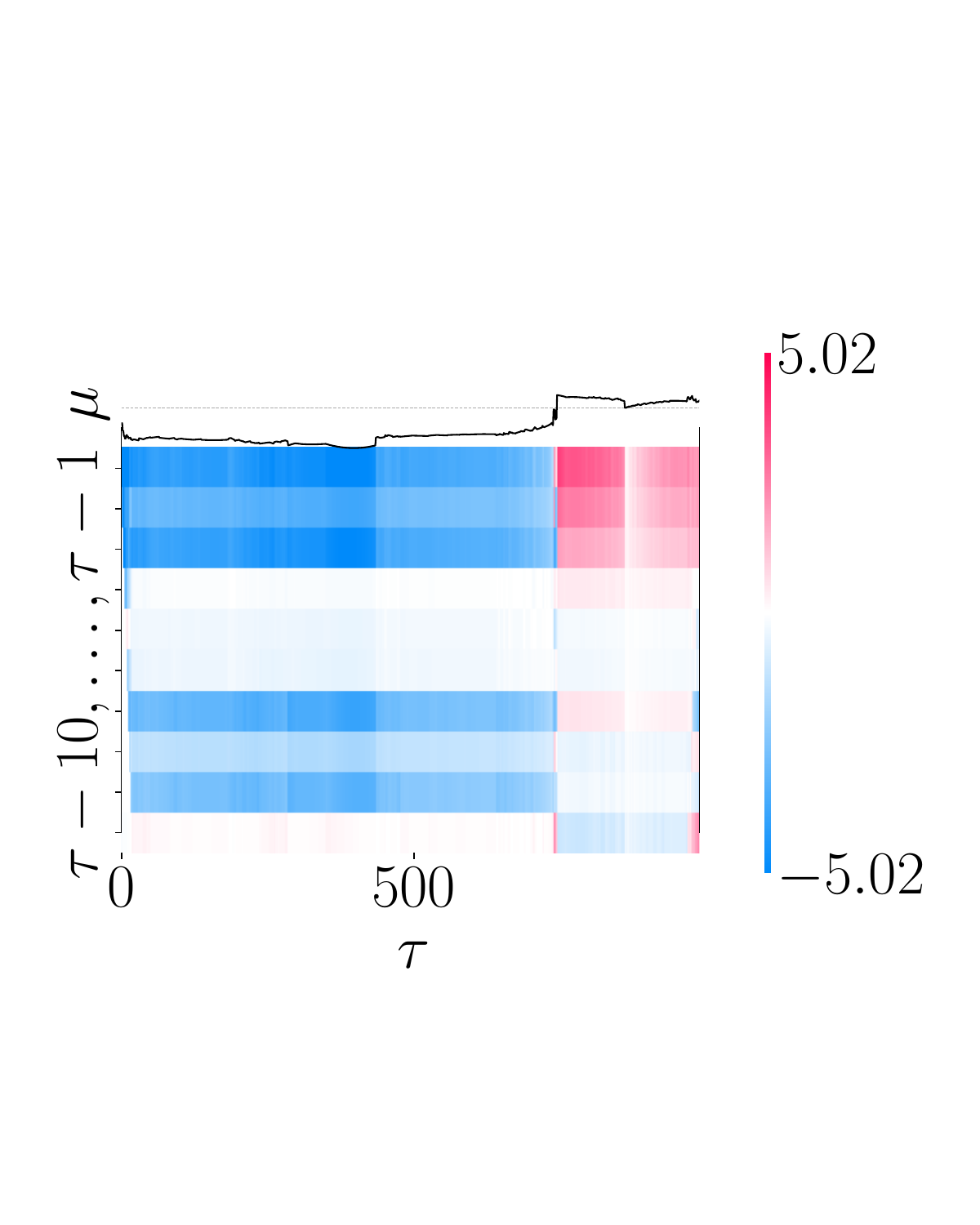}
		& \includegraphics[trim={0 4cm 0 6cm},clip,scale=\scale,valign=m]{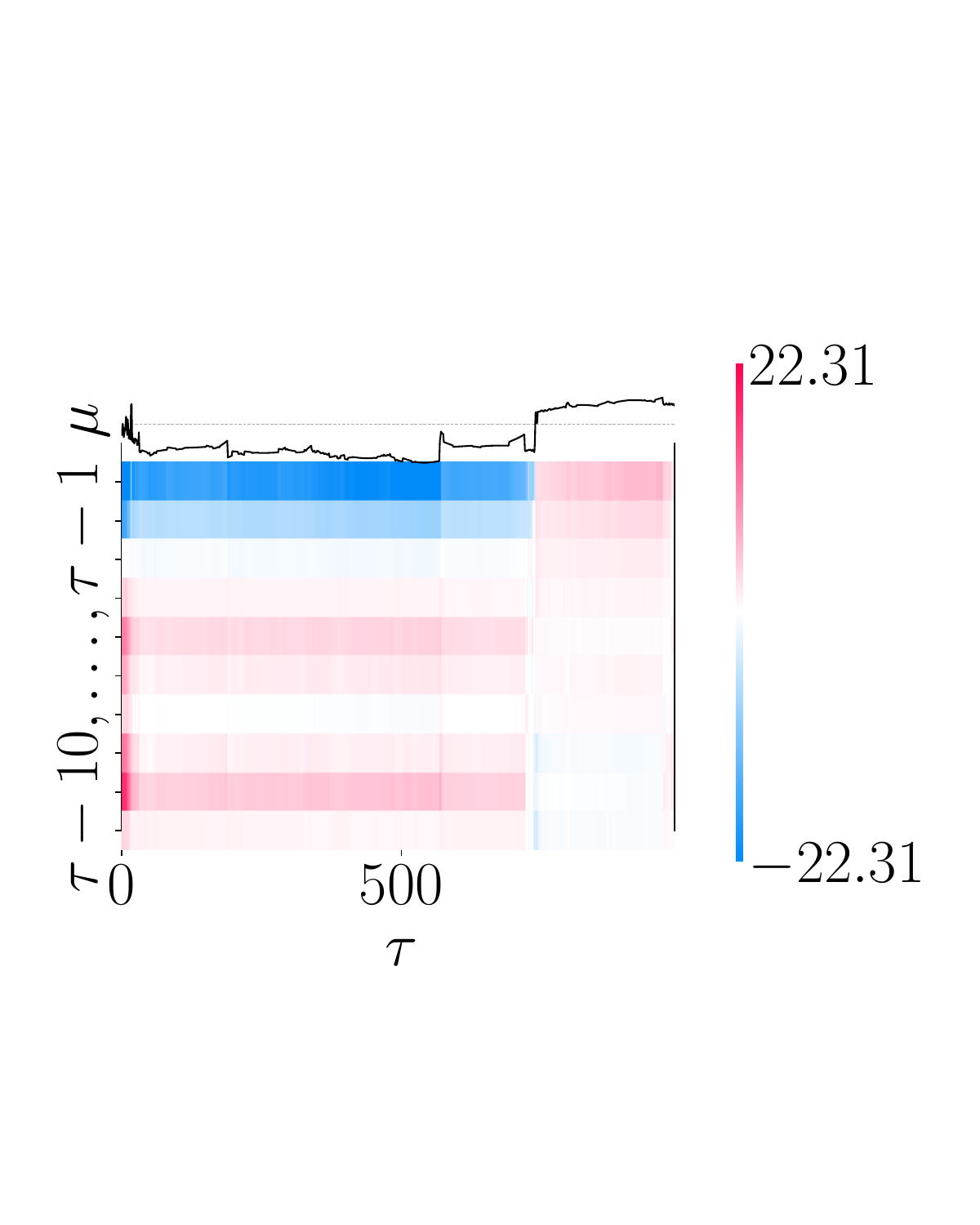}
		& \includegraphics[trim={0 4cm 0 6cm},clip,scale=\scale,valign=m]{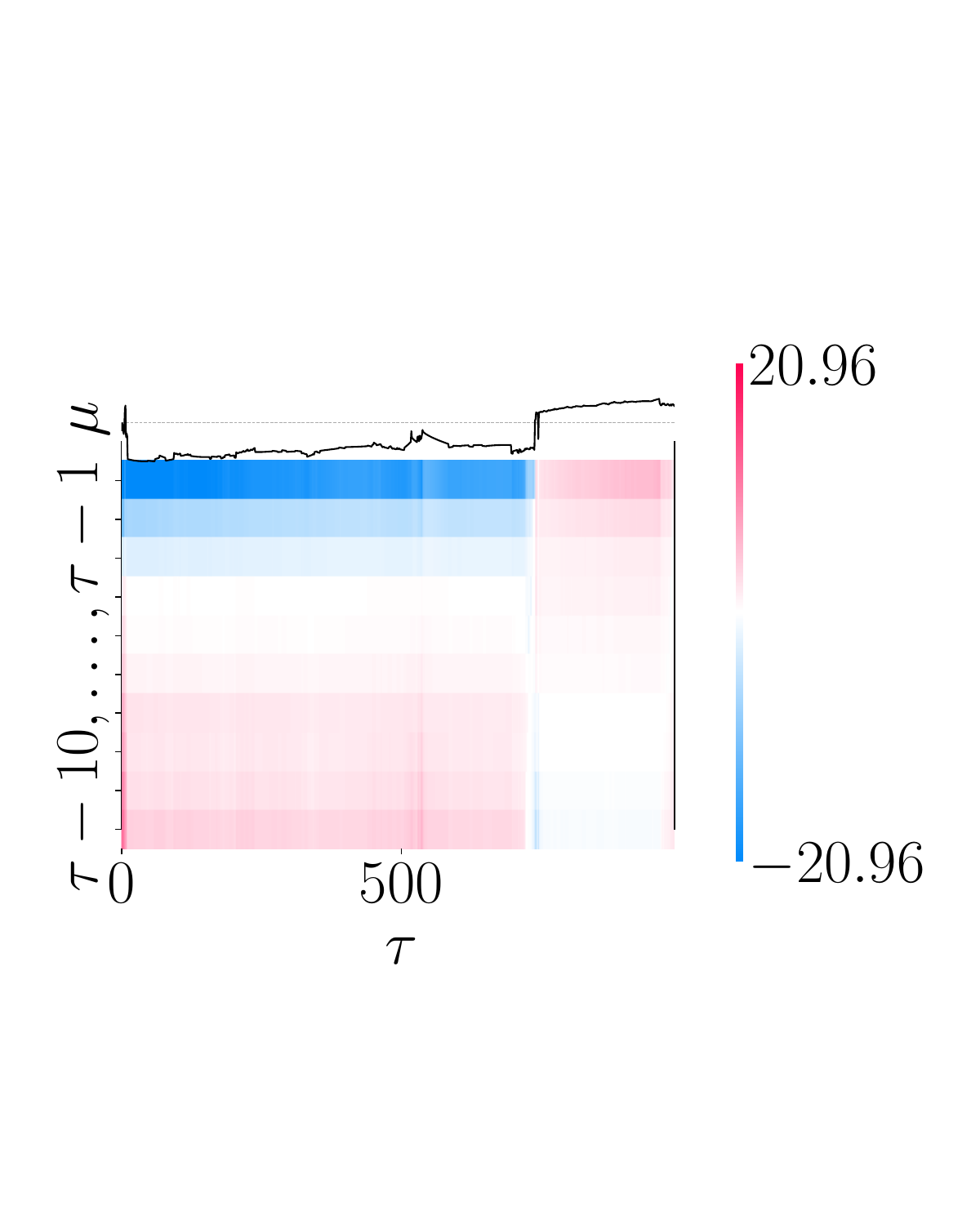} \\

		\bottomrule
	\end{tabular}
	\egroup%
	\caption{DeepLIFT attribution maps for the considered architectures using physics-informed (PI\@; top row) and
	autoregressive (AR\@; bottom row) features under {\Fthree}.}%
	\label{tab:F3-xai}%
\end{figure*}

\begin{figure}[htb]
	\centering
	\begin{adjustbox}{width=\columnwidth}
		\begin{tabular}{ccc}
			\toprule
			{\Huge BNN} & {\Huge MLP} & {\Huge DE} \\
			\midrule

			\includegraphics[width=1.5\linewidth,valign=m]{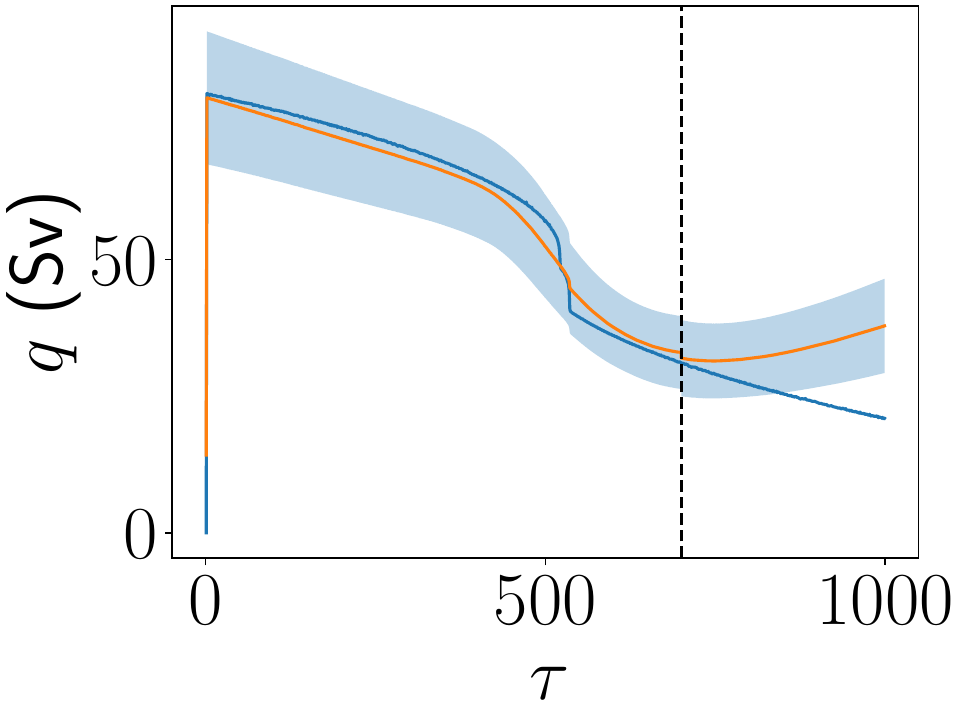}
			& \includegraphics[width=1.5\linewidth,valign=m]{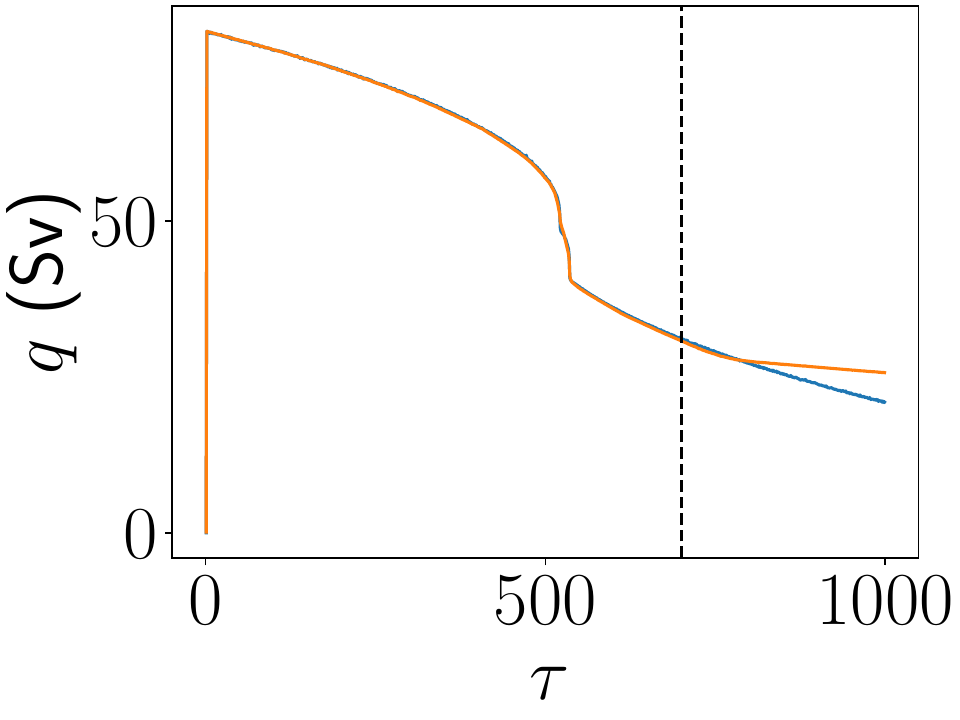}
			& \includegraphics[width=1.5\linewidth,valign=m]{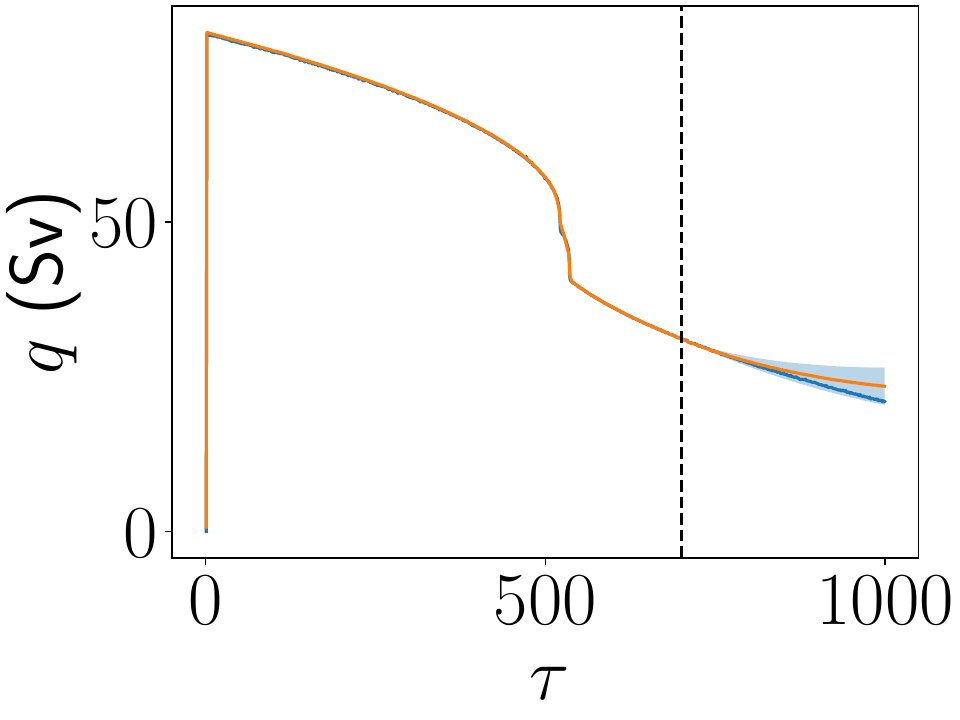} \\
			\includegraphics[width=1.5\linewidth,valign=m]{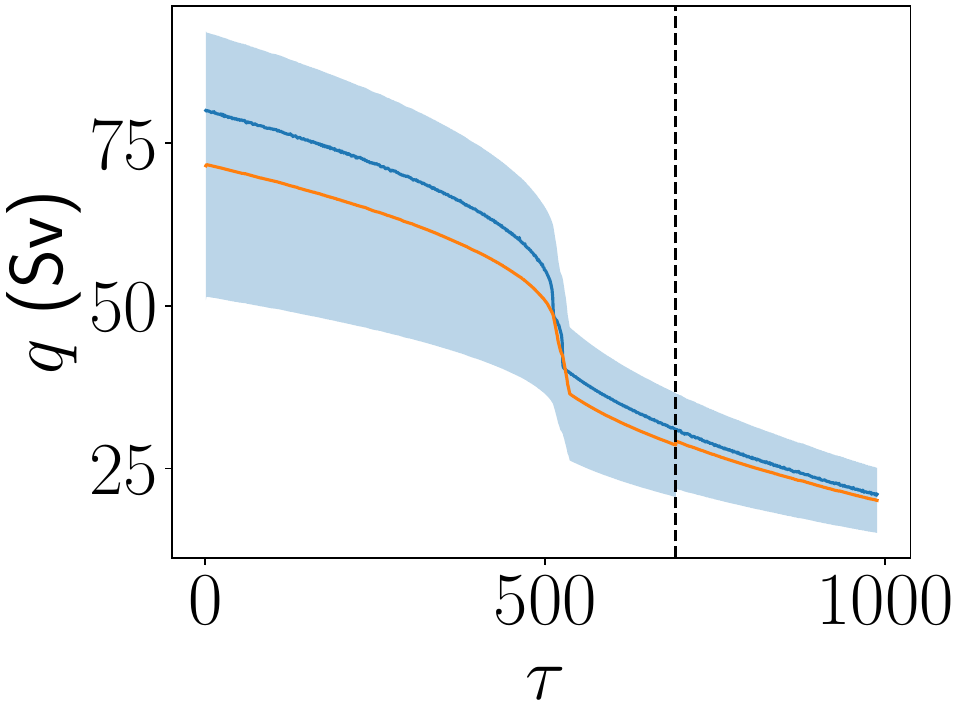}
			& \includegraphics[width=1.5\linewidth,valign=m]{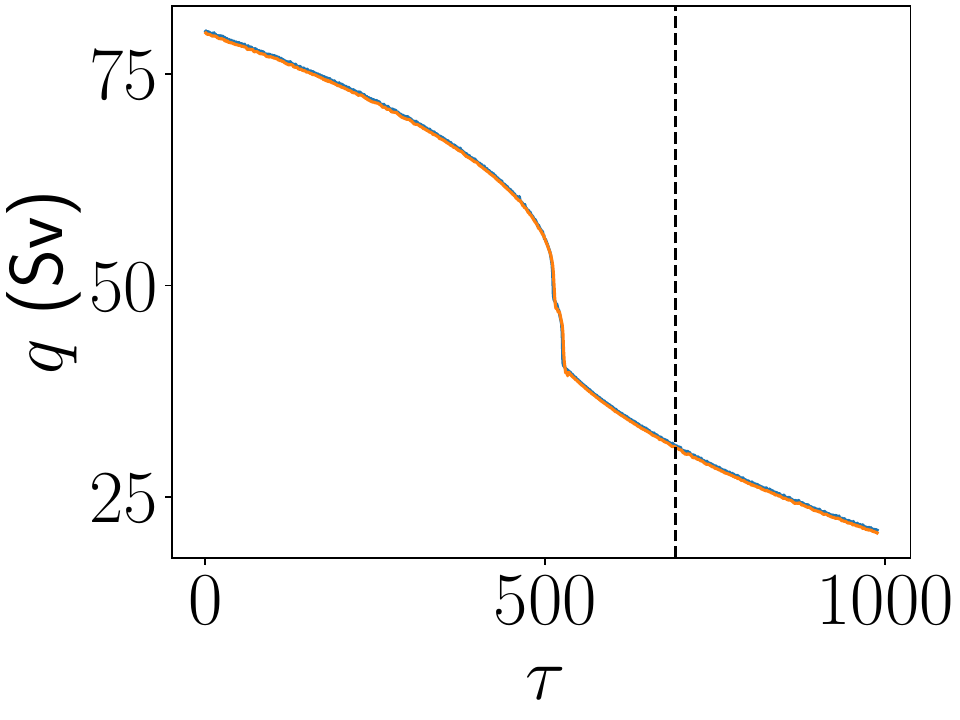}
			& \includegraphics[width=1.5\linewidth,valign=m]{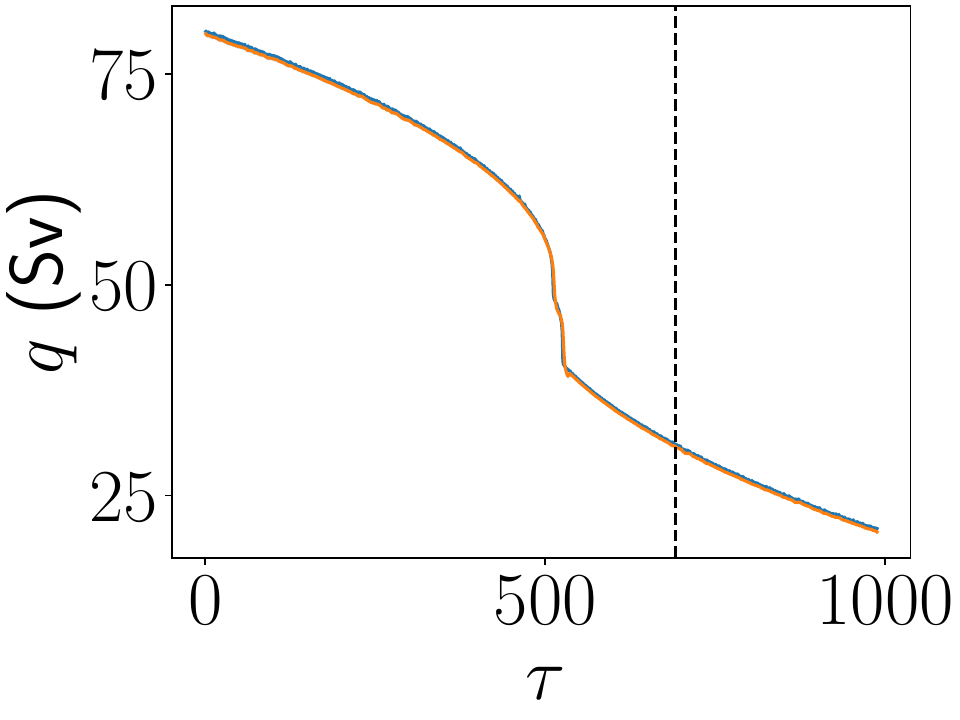} \\

			\multicolumn{3}{c}{\includegraphics[scale=2,valign=m]{legend_pred_gt.pdf}} \\

			\bottomrule
		\end{tabular}
	\end{adjustbox}%
	\caption{Predictive performance for the considered architectures using physics-informed (PI\@; first row) and
	autoregressive (AR\@; second row) features under {\Ffour}.}%
	\label{tab:F4-performance}%
\end{figure}

\begin{figure}[htb]
	\centering
	\begin{adjustbox}{width=\columnwidth}
		\begin{tabular}{ccc}
			\toprule
			\resizebox{0.3\linewidth}{!}{BNN} & \resizebox{0.3\linewidth}{!}{MLP} & \resizebox{0.2\linewidth}{!}{DE} \\
			\midrule

			\includegraphics[width=1.6\linewidth,valign=m]{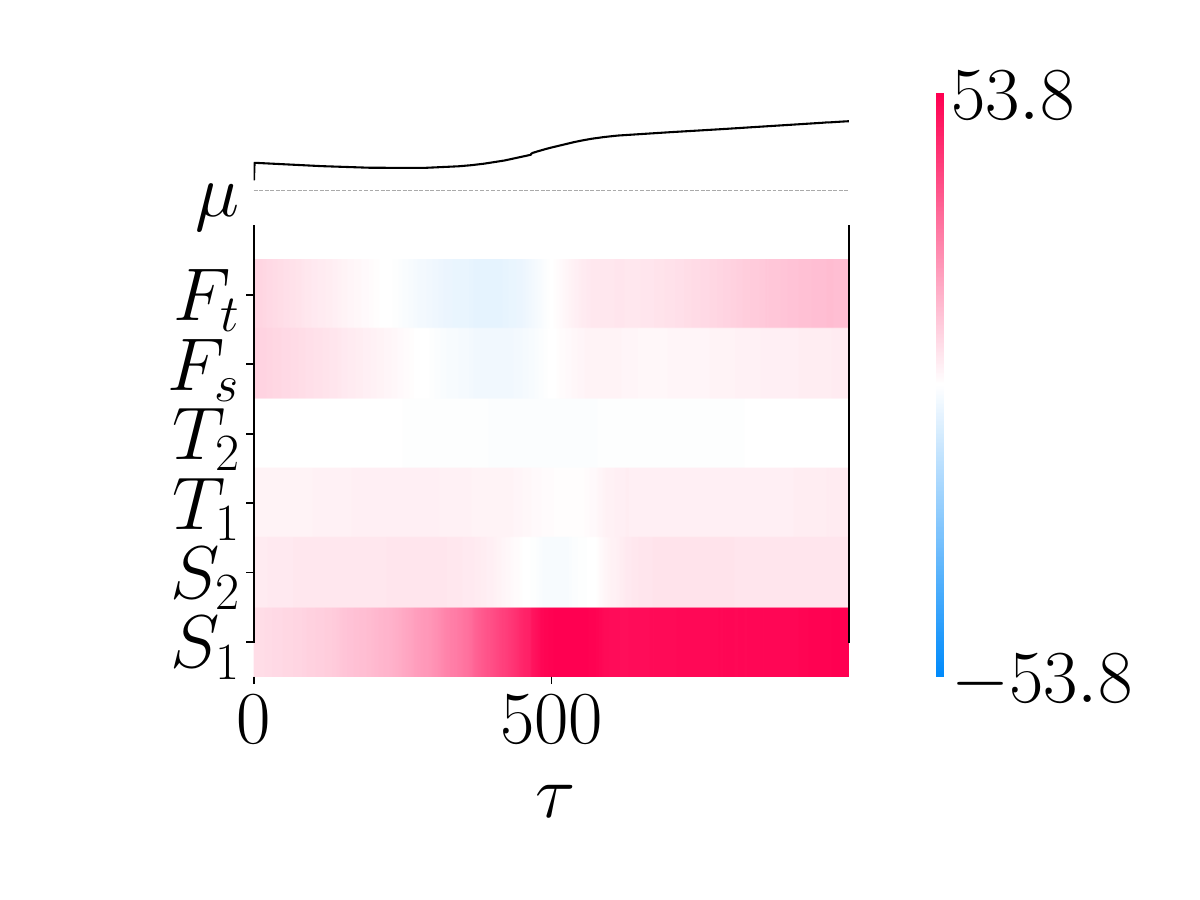}
			& \includegraphics[width=1.6\linewidth,valign=m]{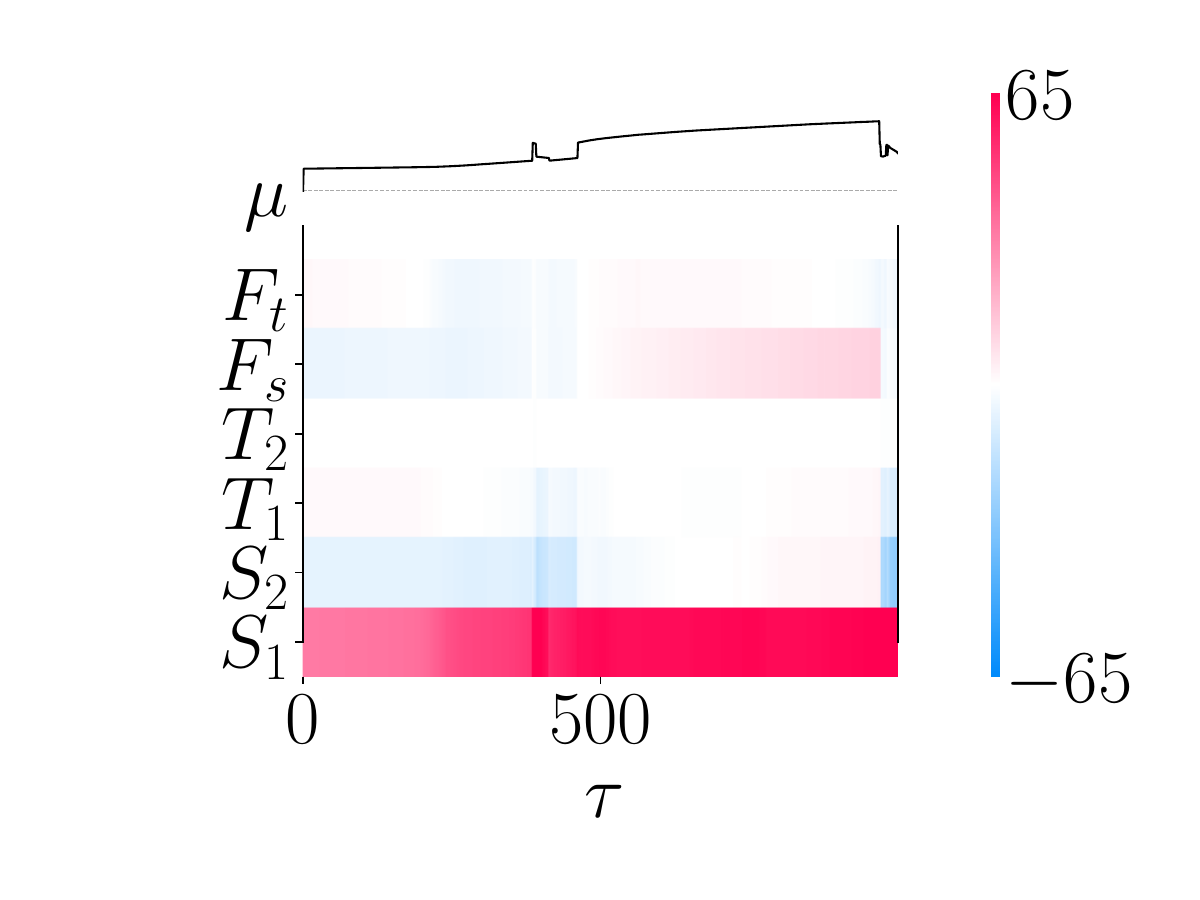}
			& \includegraphics[width=1.6\linewidth,valign=m]{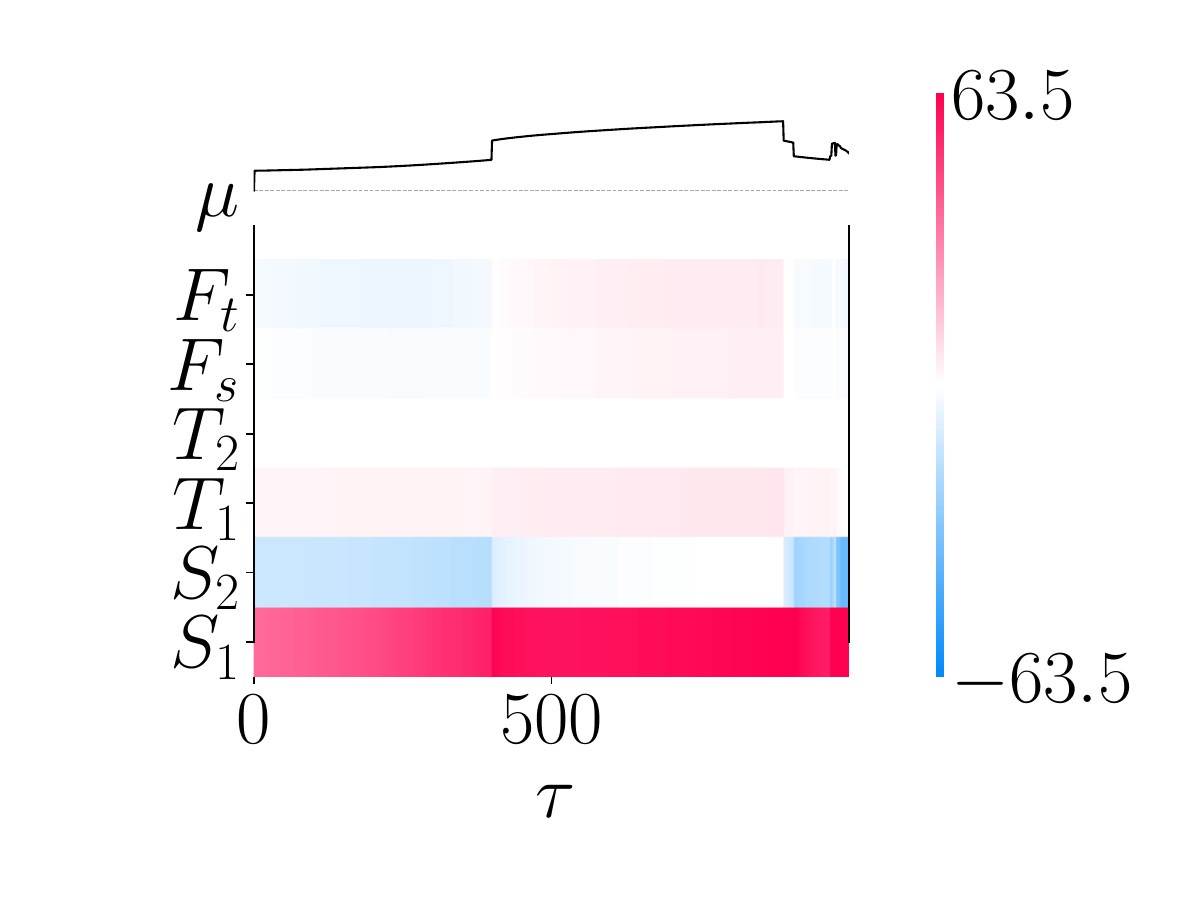} \\

			\includegraphics[trim={0 4cm 0 6cm},clip,width=1.6\linewidth,valign=m]{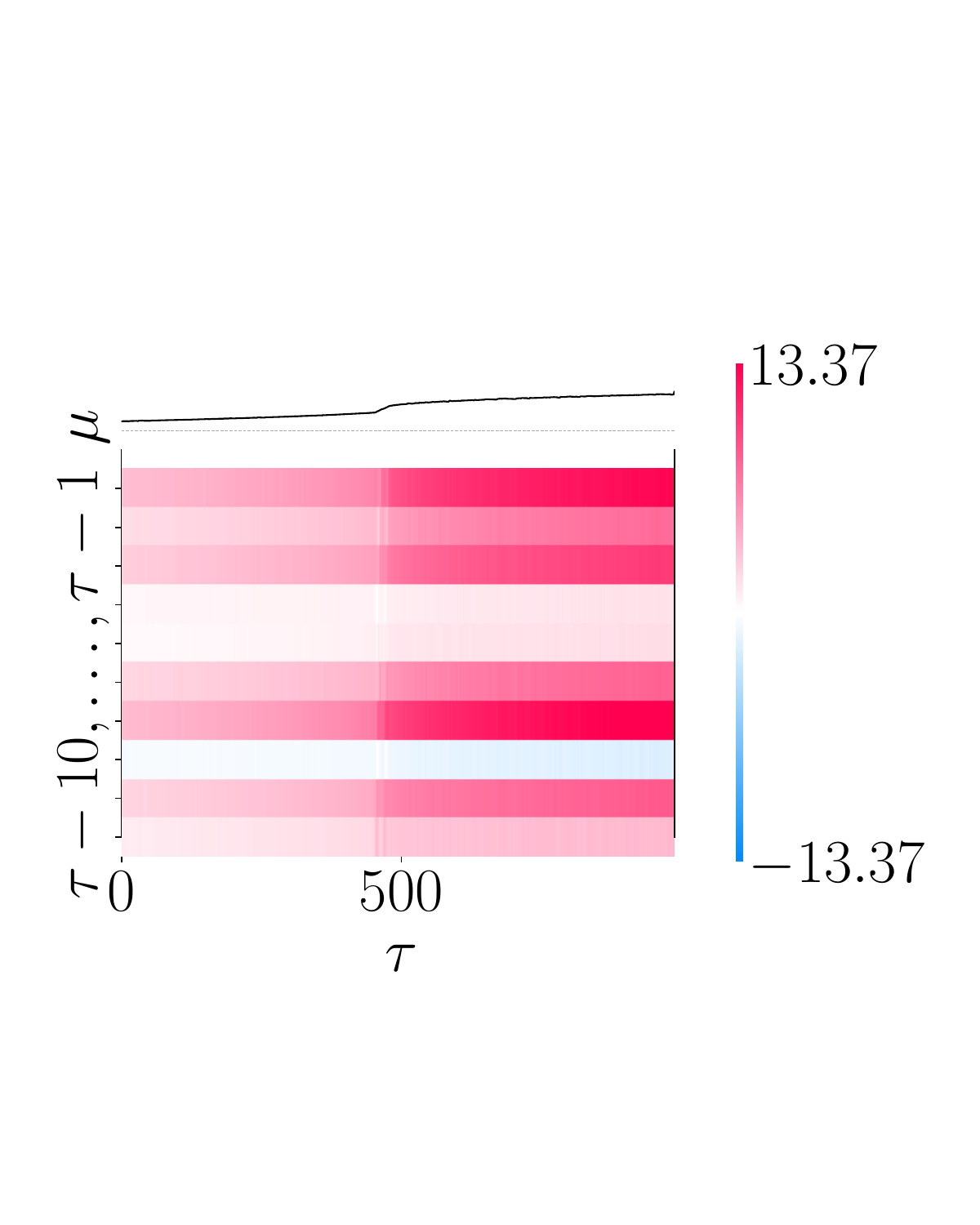}
			& \includegraphics[trim={0 4cm 0 6cm},clip,width=1.6\linewidth,valign=m]{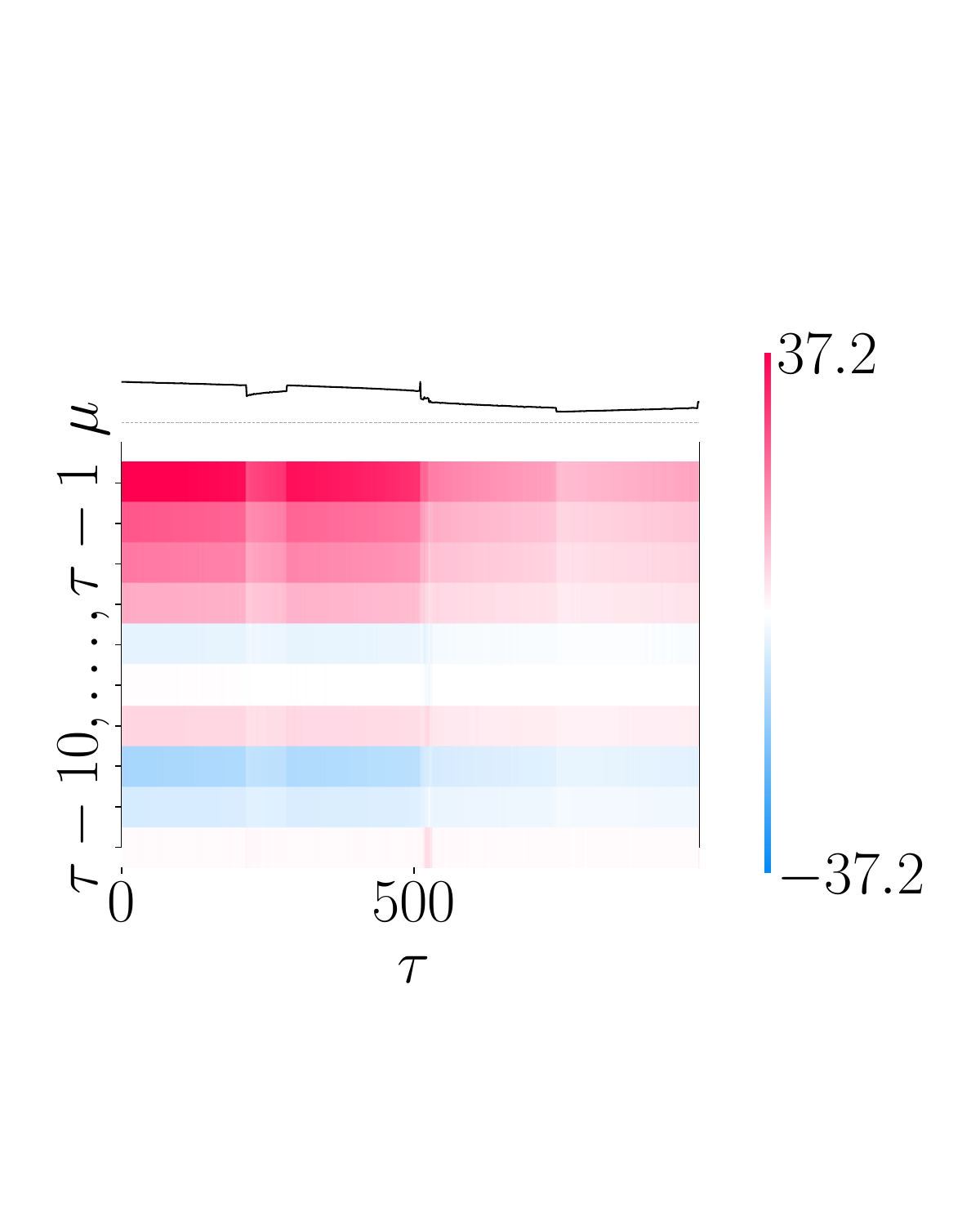}
			& \includegraphics[trim={0 4cm 0 6cm},clip,width=1.6\linewidth,valign=m]{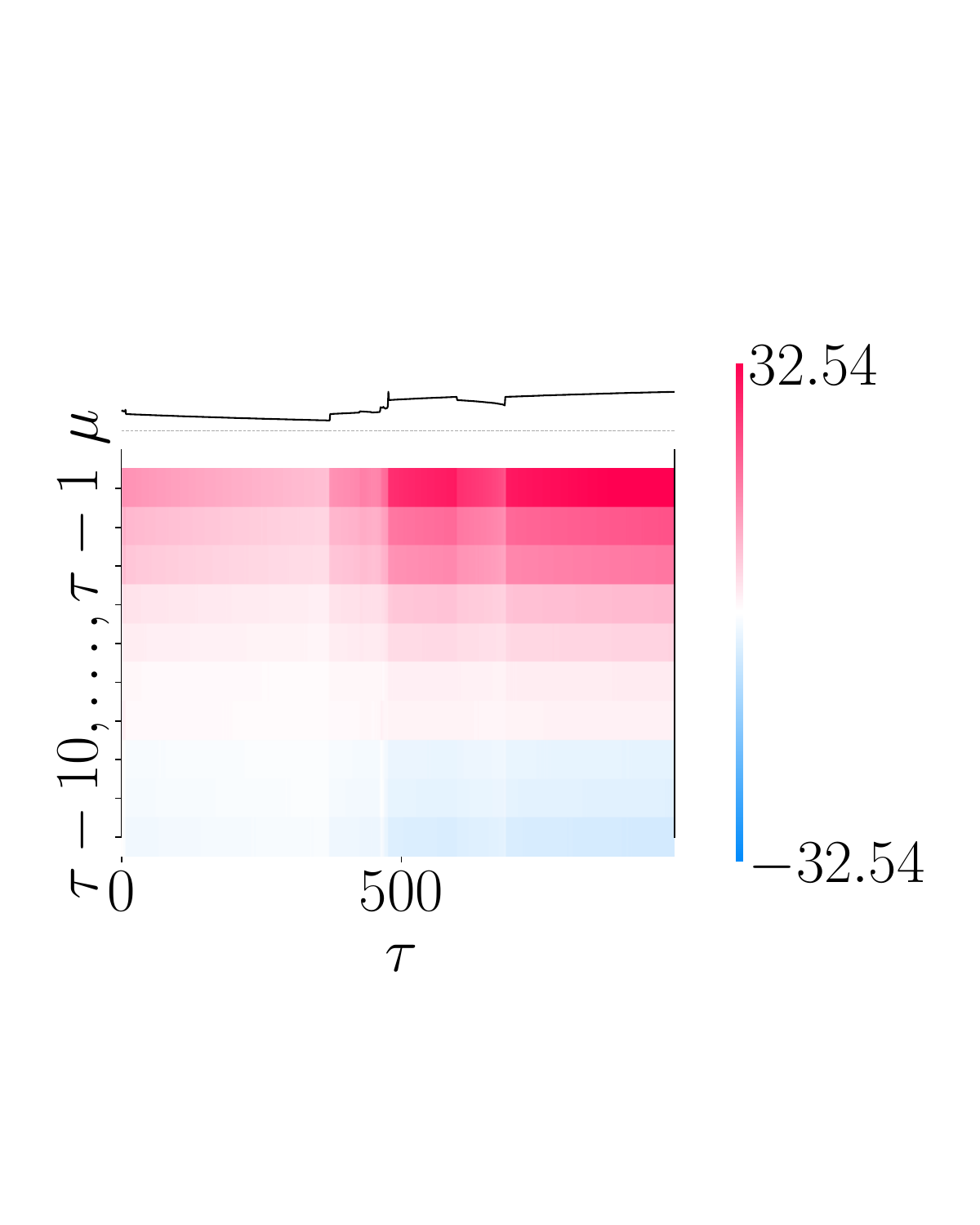} \\

			\bottomrule
		\end{tabular}
	\end{adjustbox}%
	\caption{DeepLIFT attribution maps for the considered architectures using physics-informed (PI\@; top row) and
	autoregressive (AR\@; bottom row) features under {\Ffour}.}%
	\label{tab:F4-xai}%
\end{figure}

\begin{figure}[htb]
	\centering
	\begin{adjustbox}{width=\columnwidth}
		\begin{tabular}{ccc}
			\toprule
			{\Huge BNN} & {\Huge MLP} & {\Huge DE} \\
			\midrule

			\includegraphics[width=1.5\linewidth,valign=m]{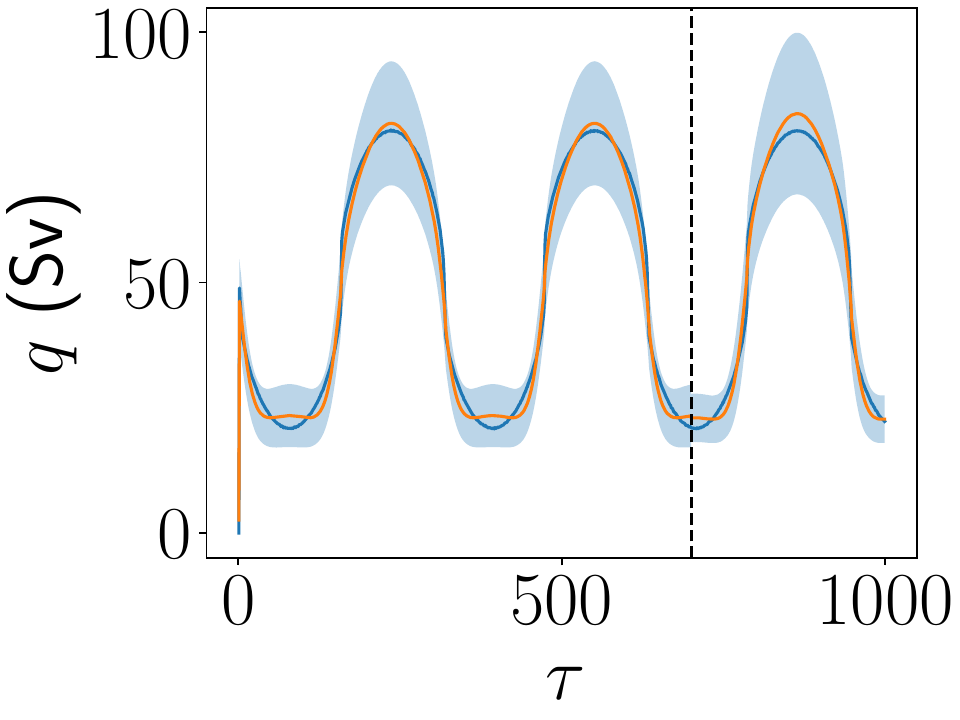}
			& \includegraphics[width=1.5\linewidth,valign=m]{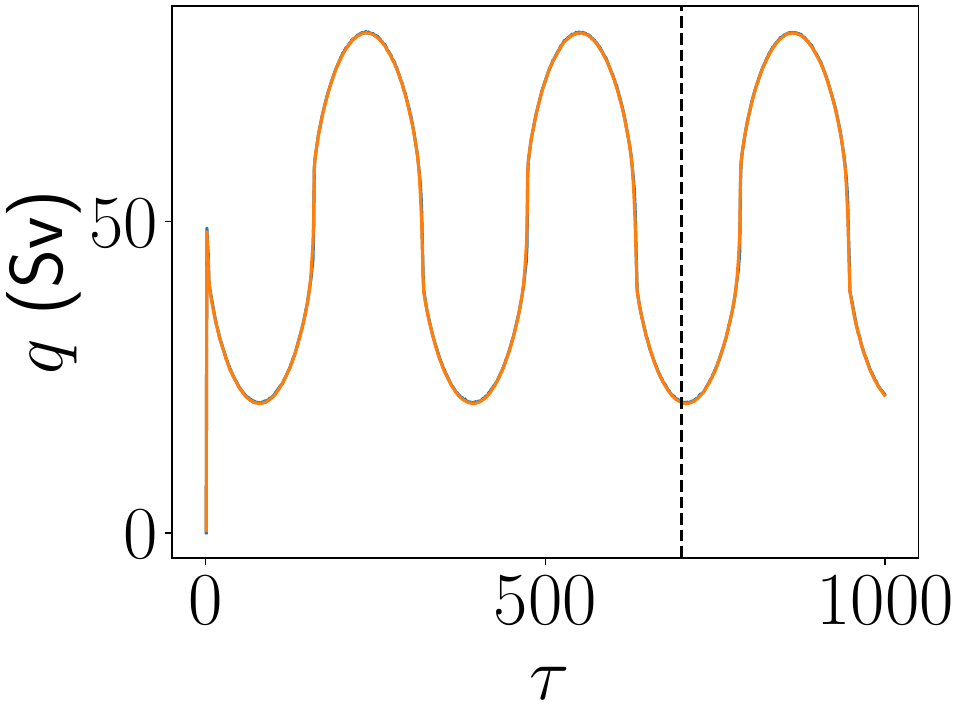}
			& \includegraphics[width=1.5\linewidth,valign=m]{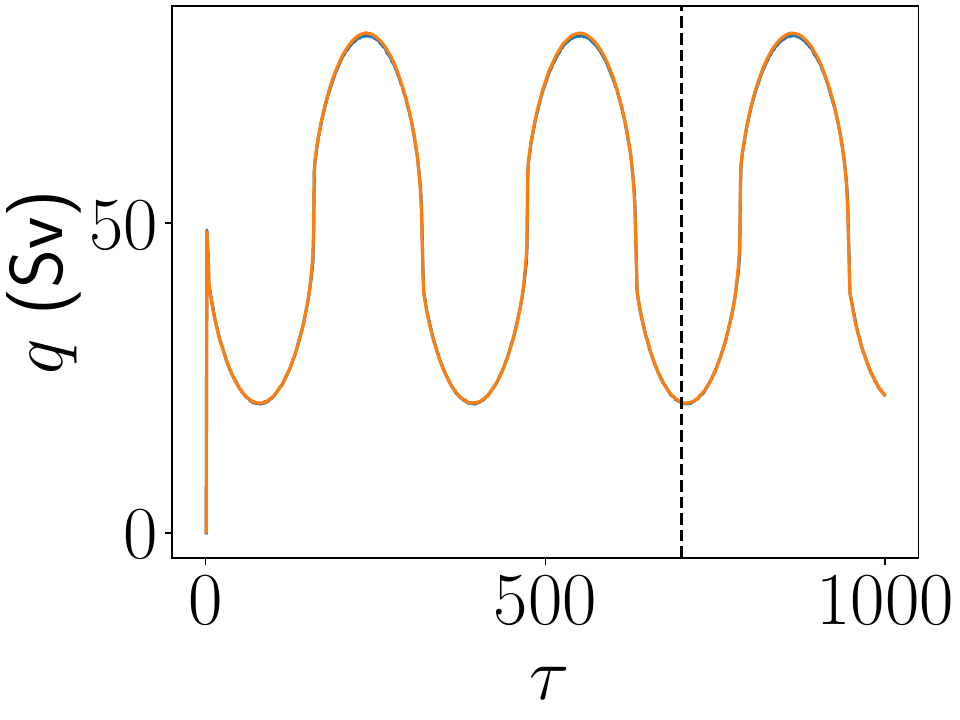} \\
			\includegraphics[width=1.5\linewidth,valign=m]{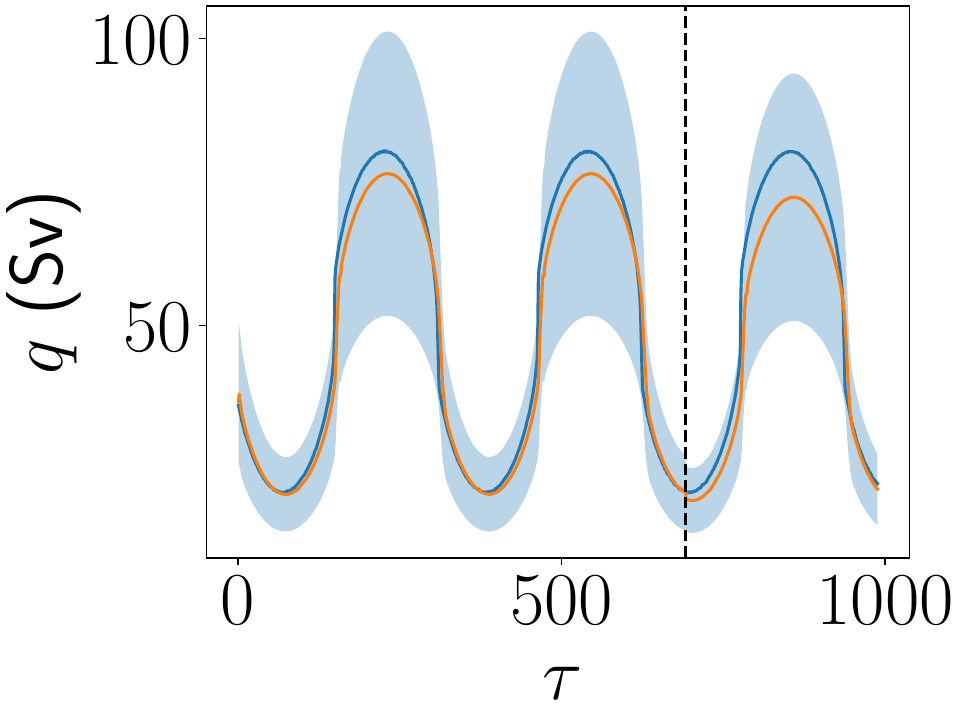}
			& \includegraphics[width=1.5\linewidth,valign=m]{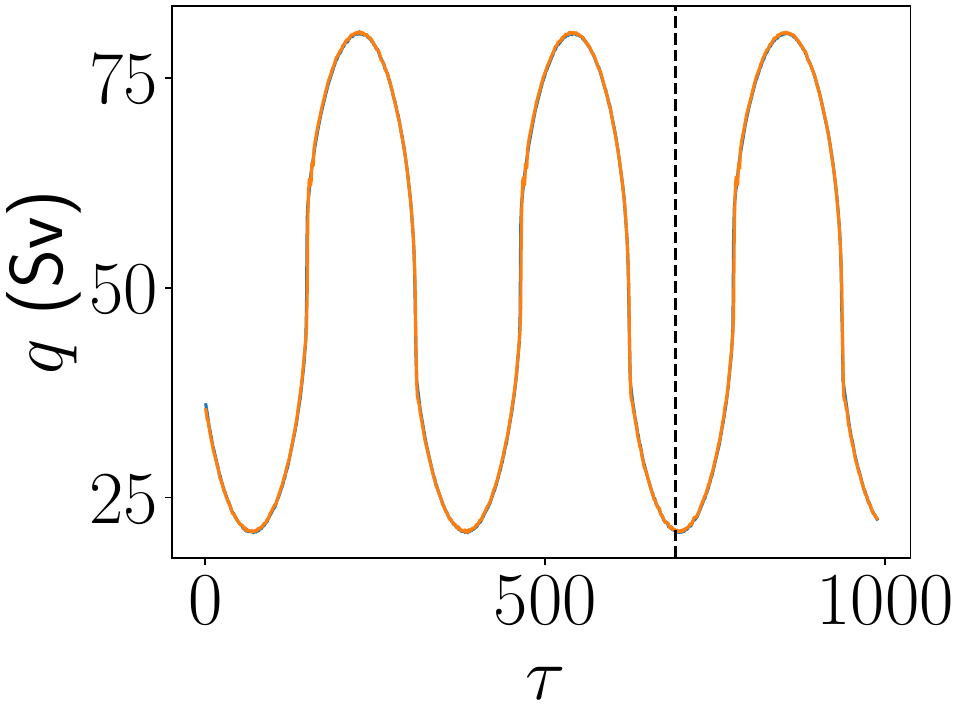}
			& \includegraphics[width=1.5\linewidth,valign=m]{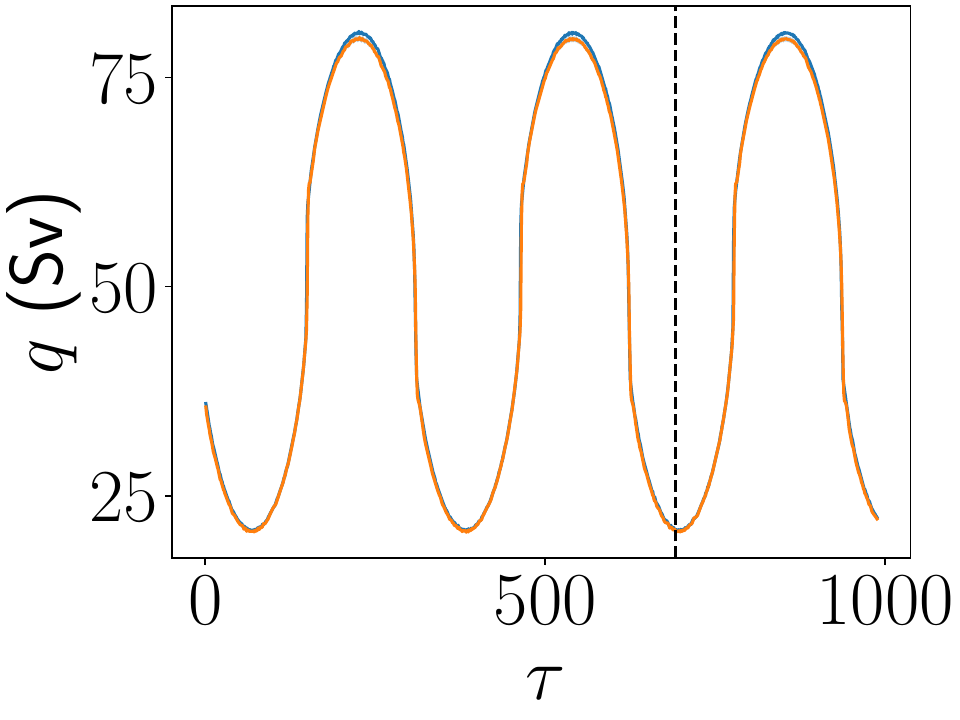} \\

			\multicolumn{3}{c}{\includegraphics[scale=2,valign=m]{legend_pred_gt.pdf}} \\

			\bottomrule
		\end{tabular}
	\end{adjustbox}%
	\caption{Predictive performance for the considered architectures using physics-informed (PI\@; first row) and
	autoregressive (AR\@; second row) features under {\Ffive}.}%
	\label{tab:F5-performance}%
\end{figure}

\begin{figure}[H]
	\centering
	\begin{adjustbox}{width=\columnwidth}
		\begin{tabular}{cc}
			\toprule
			{\Large MLP} & {\Large DE} \\
			\midrule

			\includegraphics[trim={0 1cm 0 0.8cm},clip,width=\linewidth,valign=m]{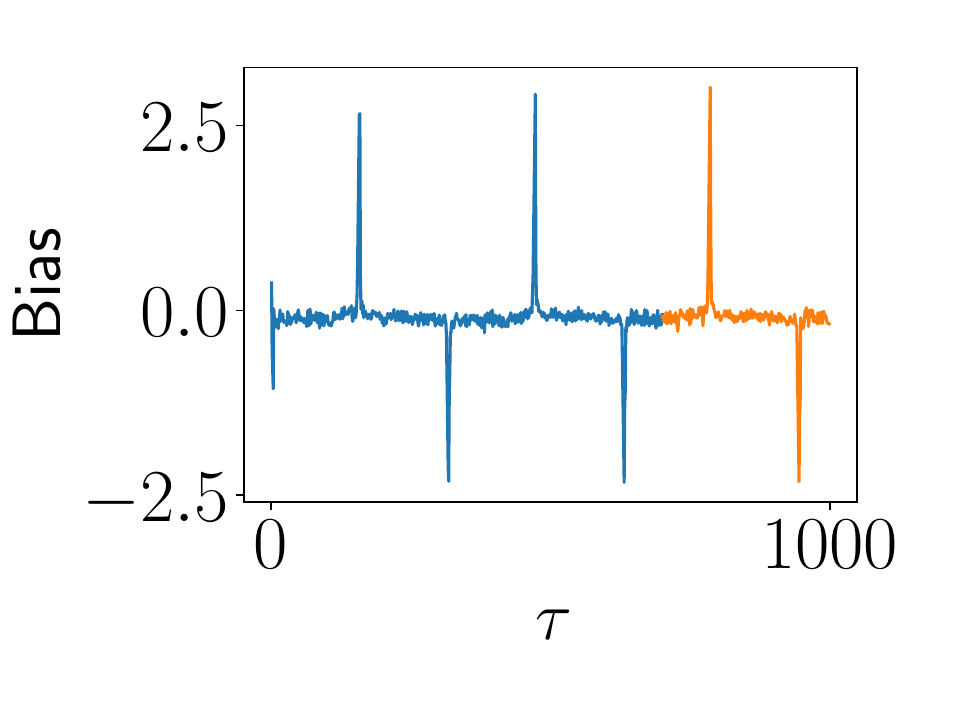}
			& \includegraphics[trim={0 1cm 0 0.8cm},clip,width=\linewidth,valign=m]{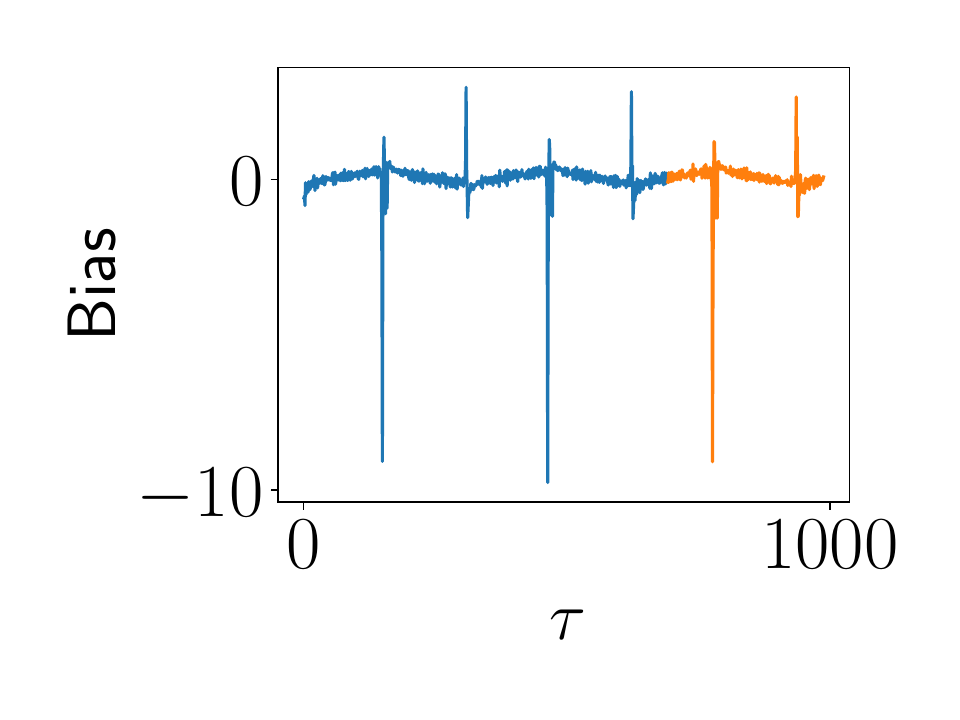} \\

			\includegraphics[trim={0 1cm 0 0.8cm},clip,width=\linewidth,valign=m]{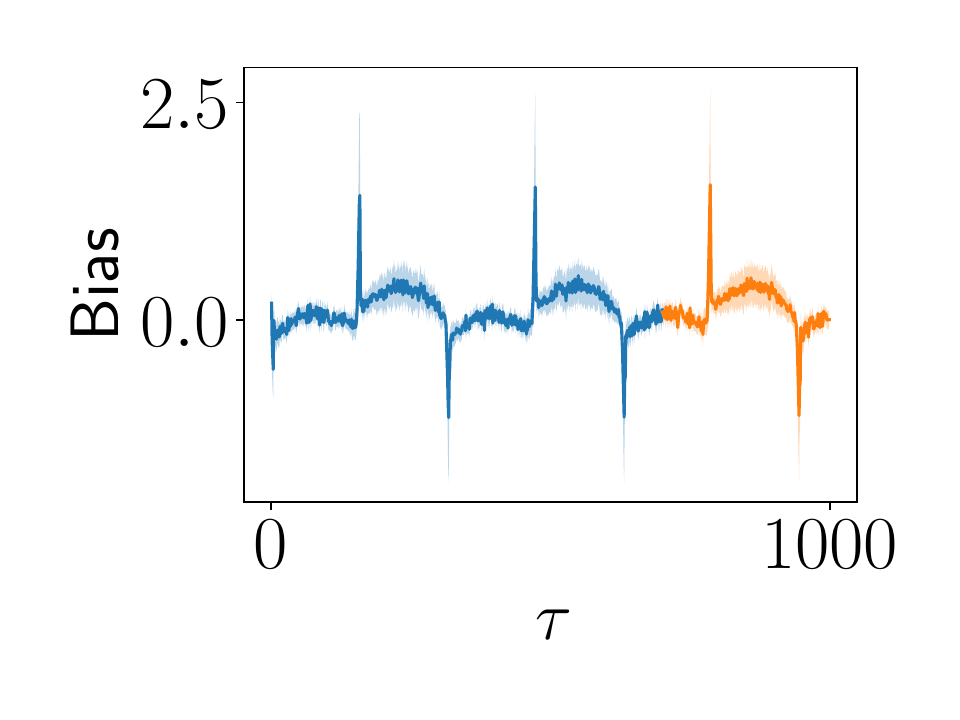}
			& \includegraphics[trim={0 1cm 0 0.8cm},clip,width=\linewidth,valign=m]{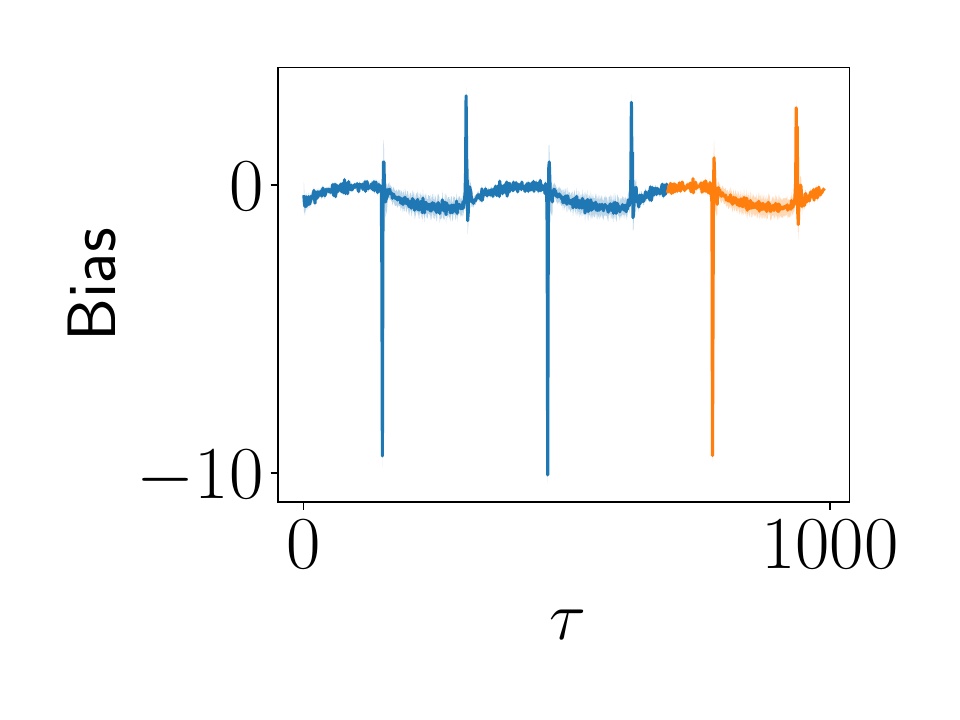} \\

			\multicolumn{2}{c}{\includegraphics[valign=m]{legend_bias.pdf}} \\

			\bottomrule
		\end{tabular}%
	\end{adjustbox}%
	\caption{Bias (\(\hat{q}_\tau - q_\tau\)) for the MLP and DE architectures under {\Ffive}.}%
	\label{tab:F5-bias}%
\end{figure}

\begin{figure}[H]
	\centering
	\begin{adjustbox}{width=\columnwidth}
		\begin{tabular}{ccc}
			\toprule
			{\Huge BNN} & {\Huge MLP} & {\Huge DE} \\
			\midrule

			\includegraphics[width=\linewidth,valign=m]{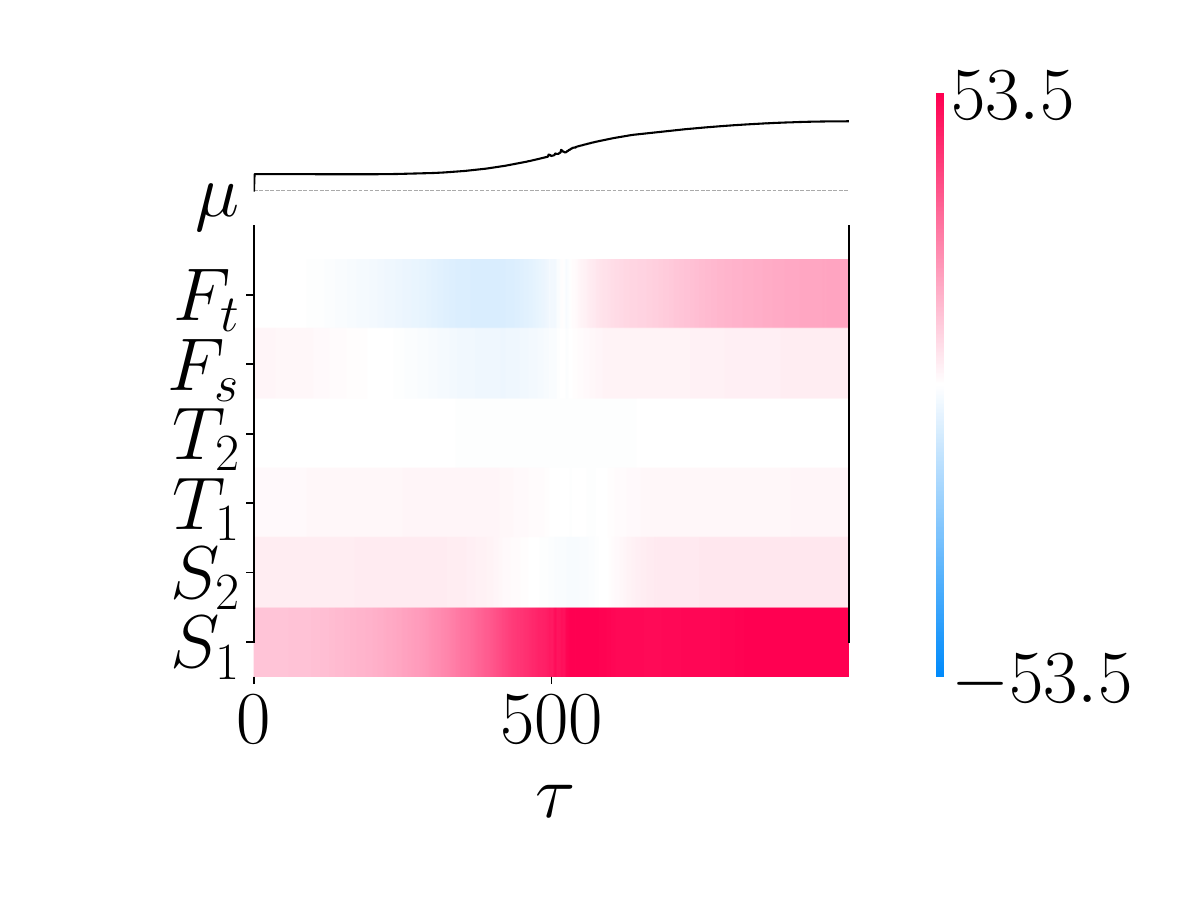}
			& \includegraphics[width=\linewidth,valign=m]{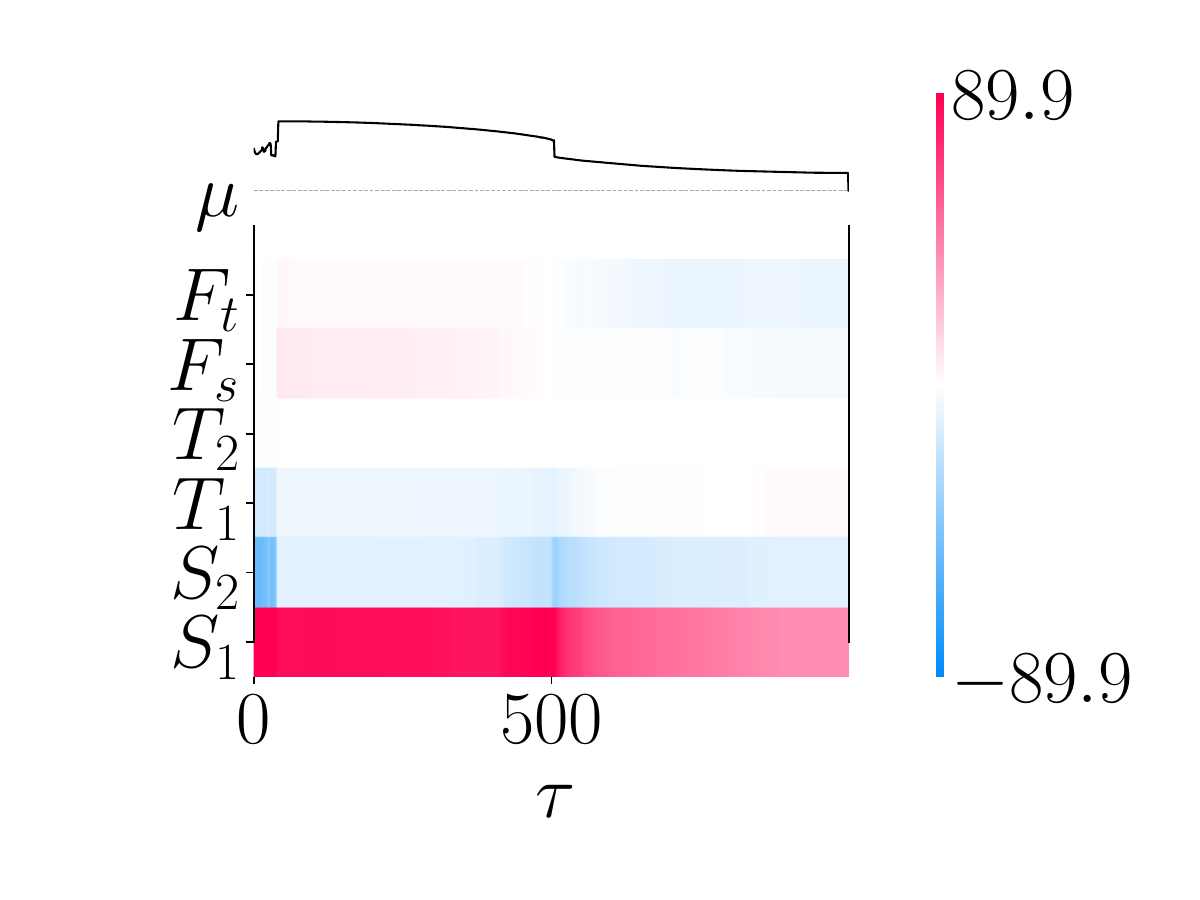}
			& \includegraphics[width=\linewidth,valign=m]{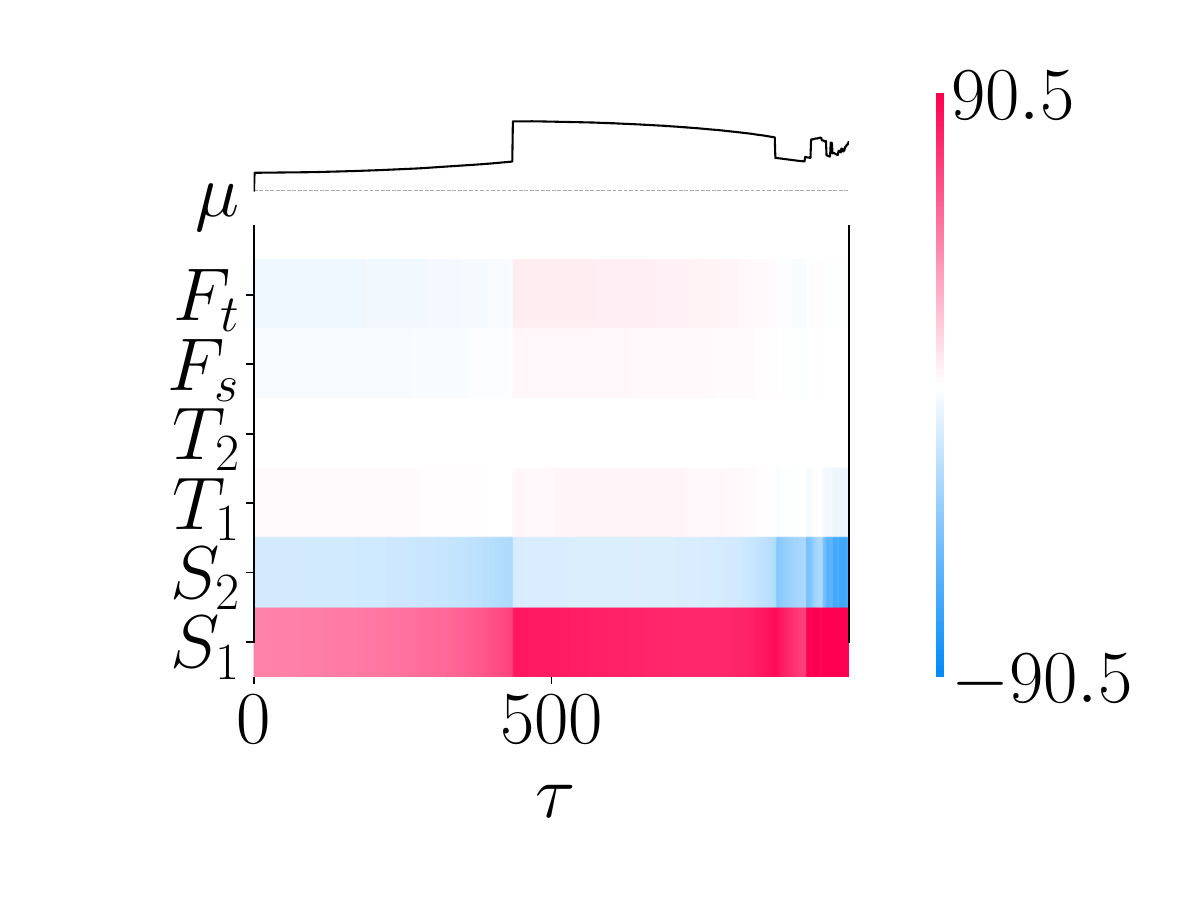} \\

			\includegraphics[trim={0 3cm 0 5cm},clip,width=\linewidth,valign=m]{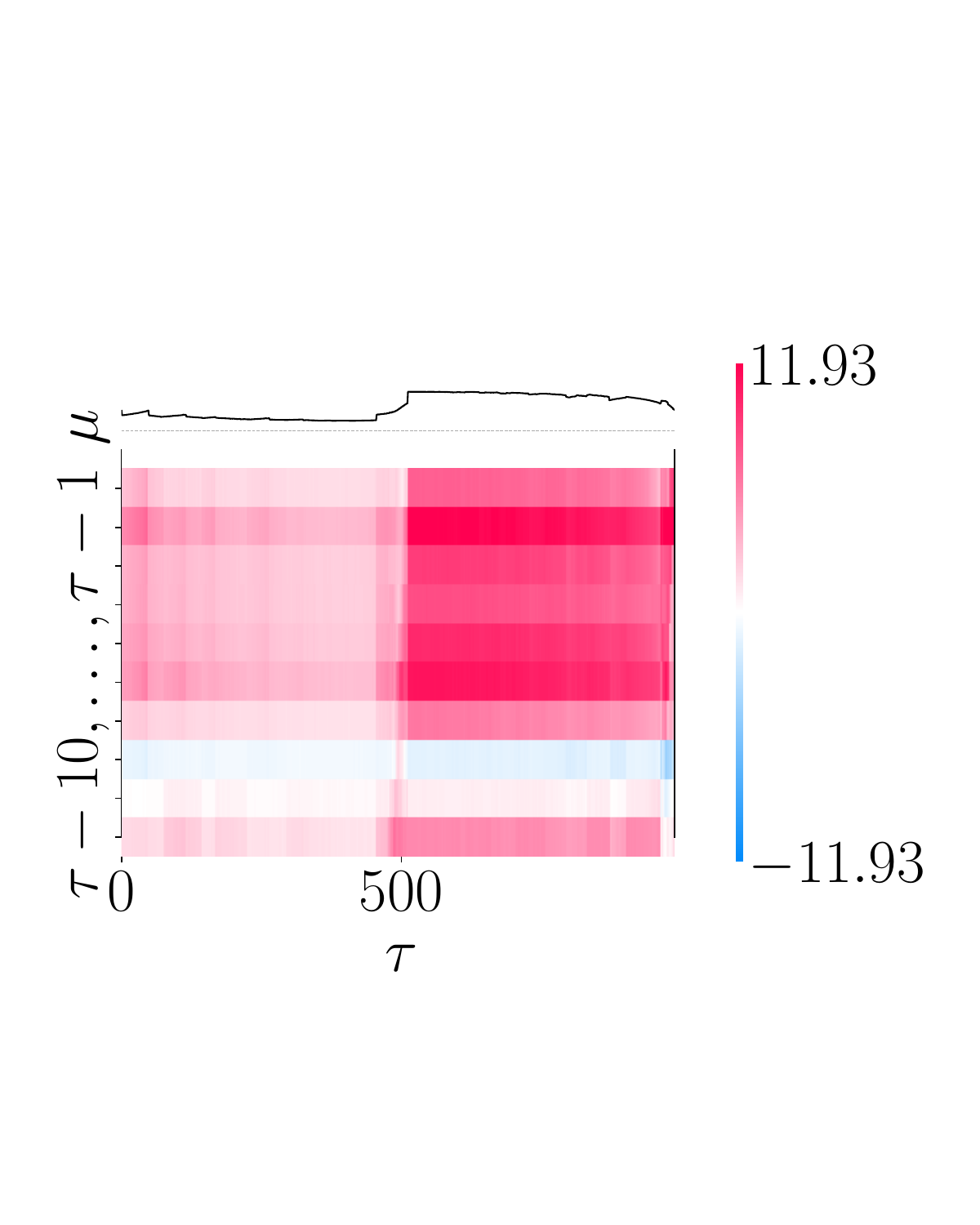}
			& \includegraphics[trim={0 3cm 0 5cm},clip,width=\linewidth,valign=m]{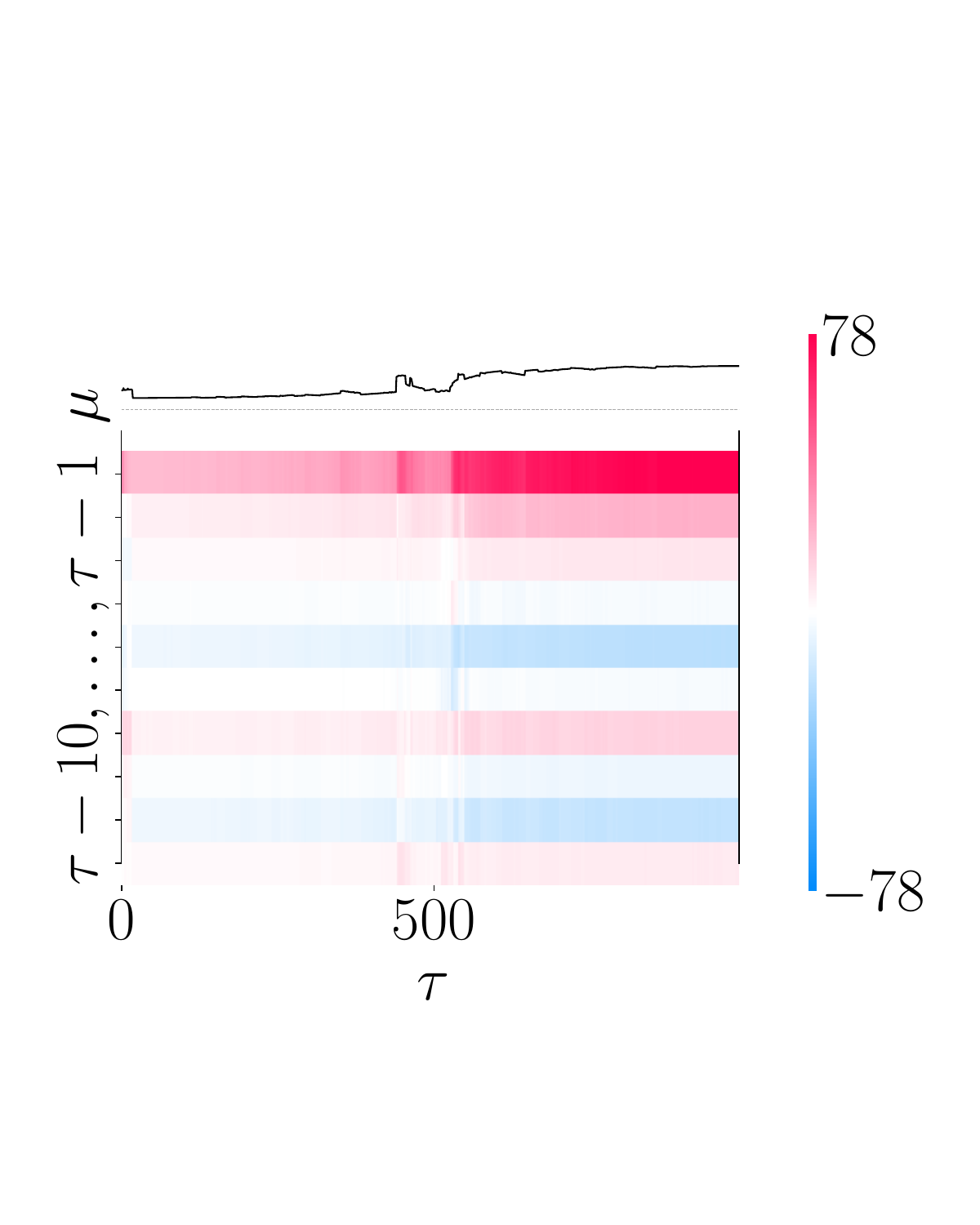}
			& \includegraphics[trim={0 3cm 0 5cm},clip,width=\linewidth,valign=m]{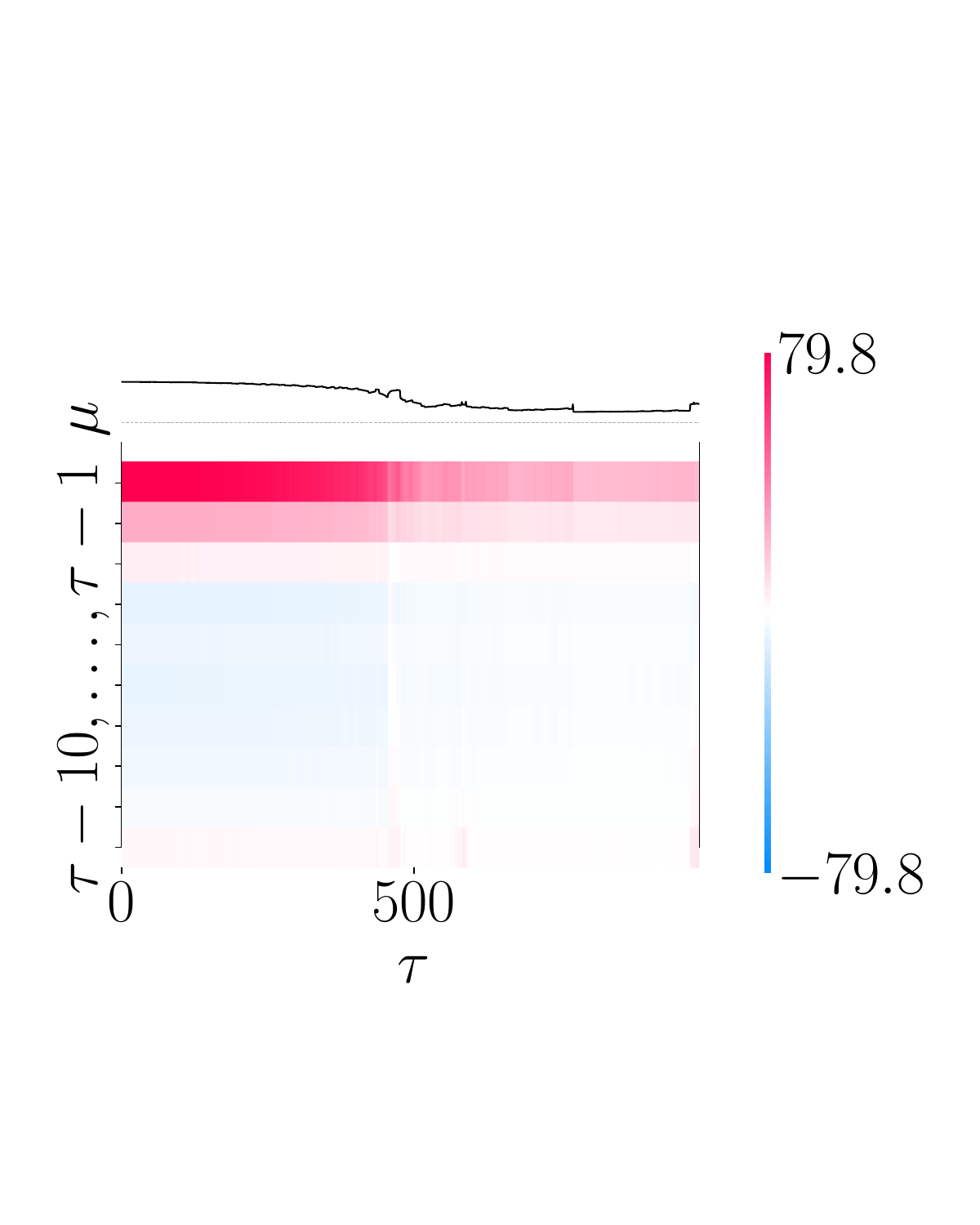} \\

			\bottomrule
		\end{tabular}
	\end{adjustbox}%
	\caption{DeepLIFT attribution maps for the considered architectures using physics-informed (PI\@; top row) and
	autoregressive (AR\@; bottom row) features under {\Ffive}.}%
	\label{tab:F5-xai}%
\end{figure}

\begin{figure}[htb]
	\centering
	\begin{adjustbox}{width=\columnwidth}
		\begin{tabular}{ccc}
			\toprule
			{\Huge BNN} & {\Huge MLP} & {\Huge DE} \\
			\midrule

			\includegraphics[width=1.5\linewidth,valign=m]{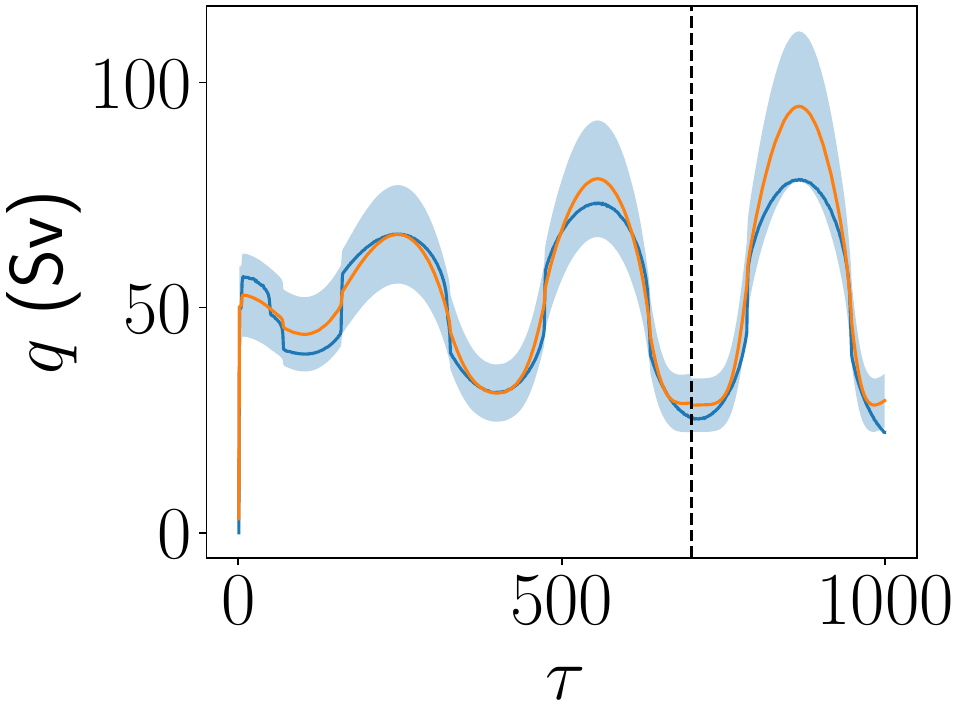}
			& \includegraphics[width=1.5\linewidth,valign=m]{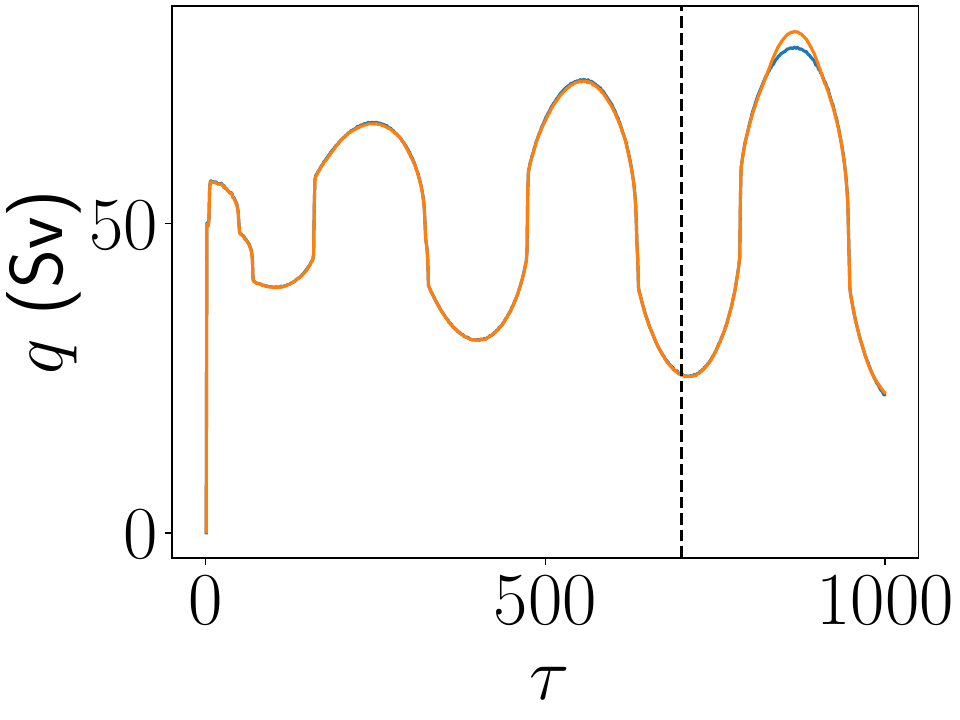}
			& \includegraphics[width=1.5\linewidth,valign=m]{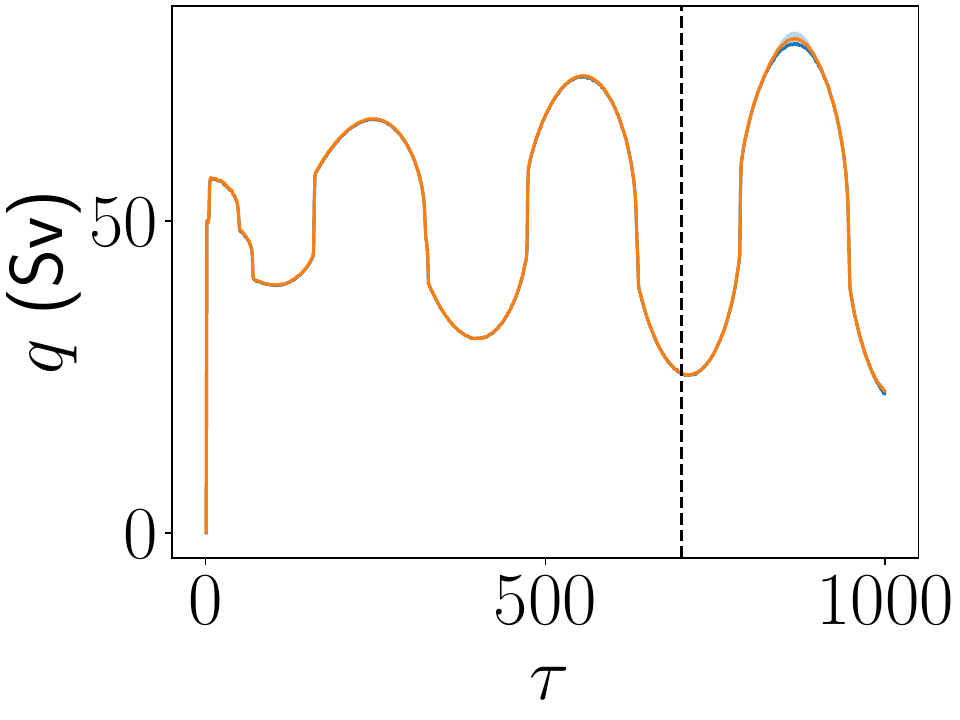} \\
			\includegraphics[width=1.5\linewidth,valign=m]{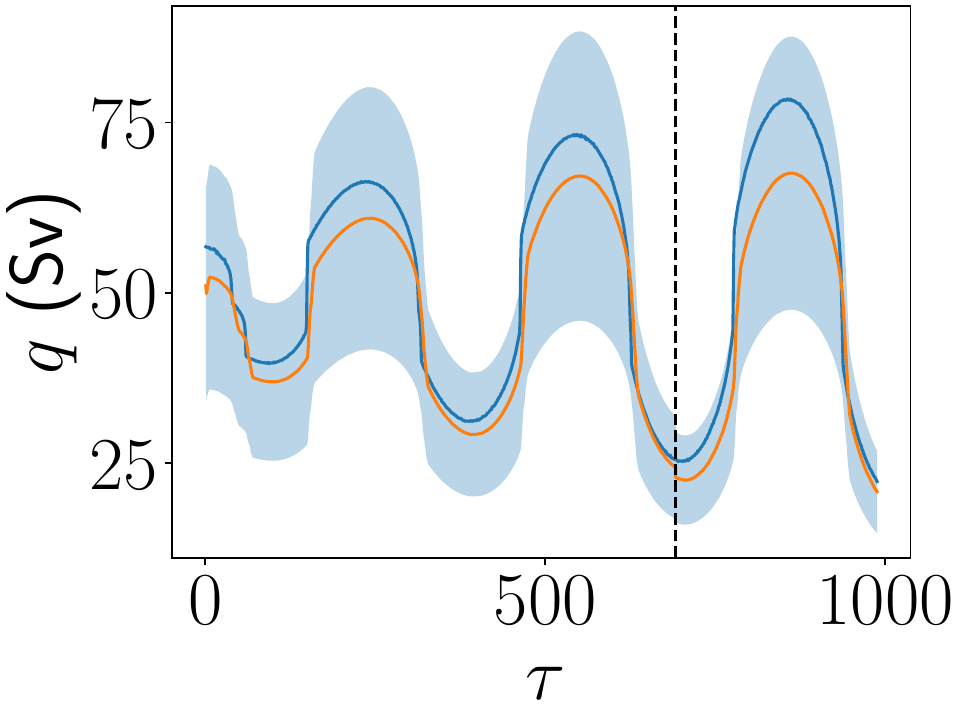}
			& \includegraphics[width=1.5\linewidth,valign=m]{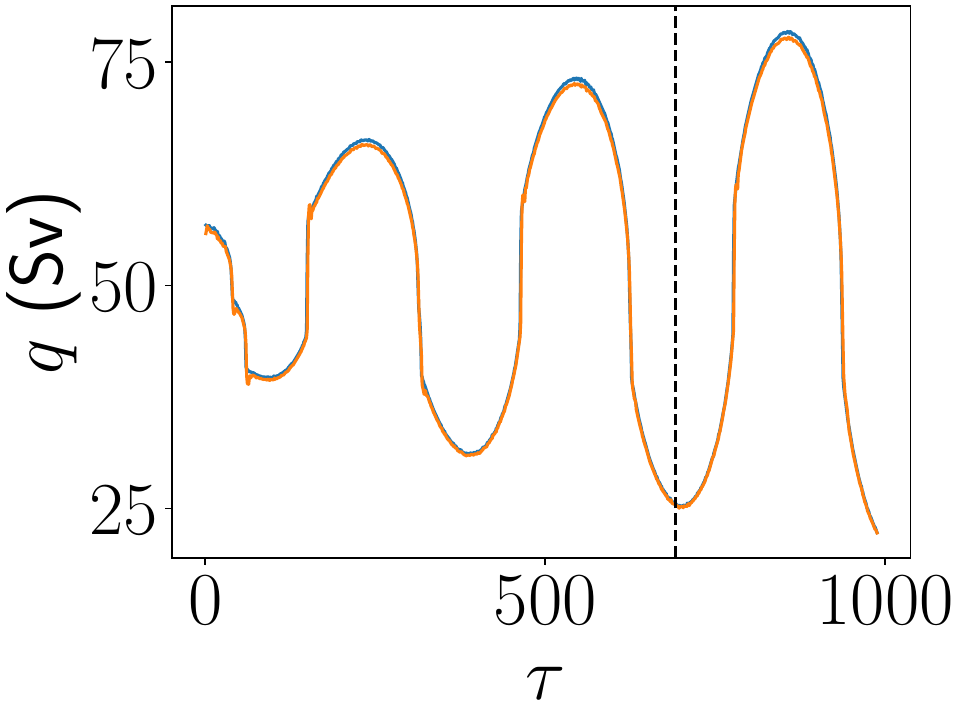}
			& \includegraphics[width=1.5\linewidth,valign=m]{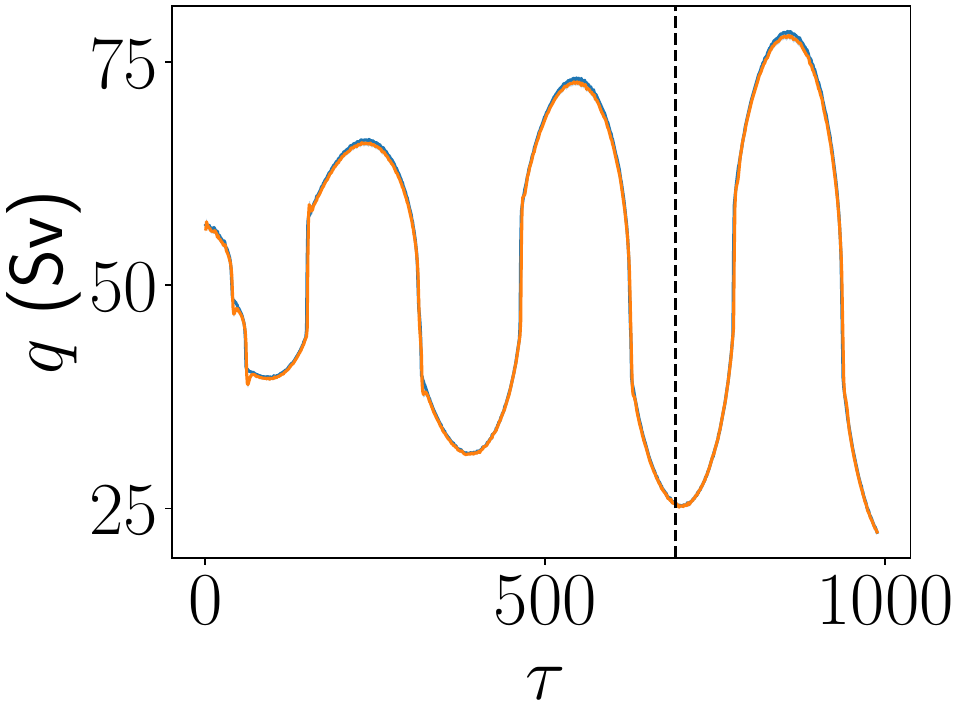} \\

			\multicolumn{3}{c}{\includegraphics[scale=2,valign=m]{legend_pred_gt.pdf}} \\

			\bottomrule
		\end{tabular}
	\end{adjustbox}%
	\caption{Predictive performance for the considered architectures using physics-informed (PI\@; first row) and
	autoregressive (AR\@; second row) features under {\Fsix}.}%
	\label{tab:F6-performance}%
\end{figure}

\begin{figure}[htb]
	\centering
	\begin{adjustbox}{width=\columnwidth}
		\begin{tabular}{cc}
			\toprule
			{\huge MLP} & {\huge DE} \\
			\midrule

			\includegraphics[trim={0 1cm 0 0.8cm},clip,width=\linewidth,valign=m]{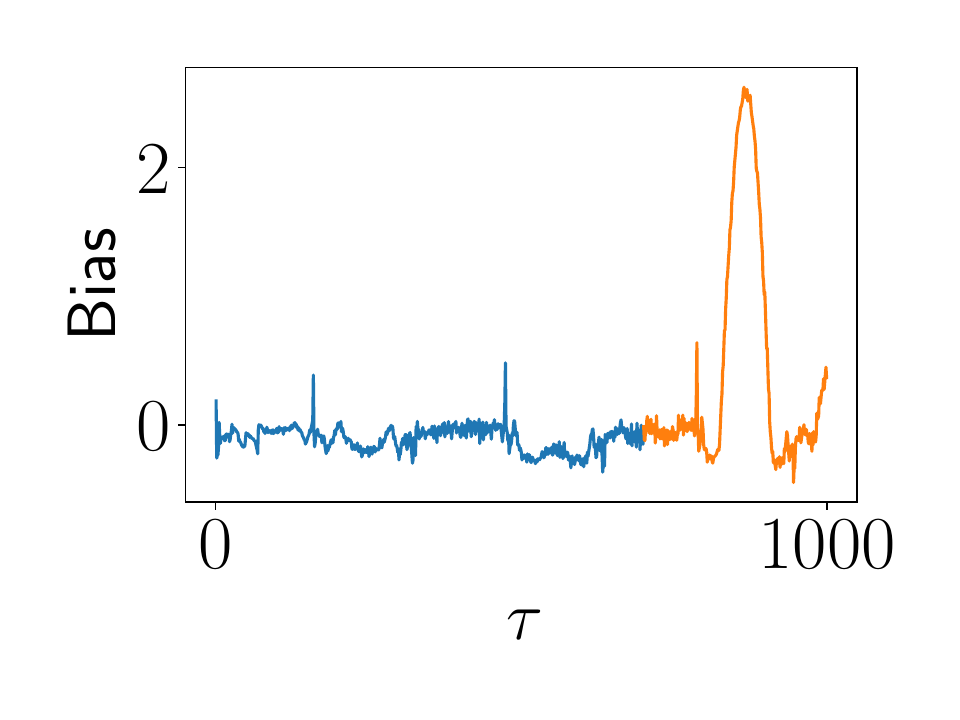}
			& \includegraphics[trim={0 1cm 0 0.8cm},clip,width=\linewidth,valign=m]{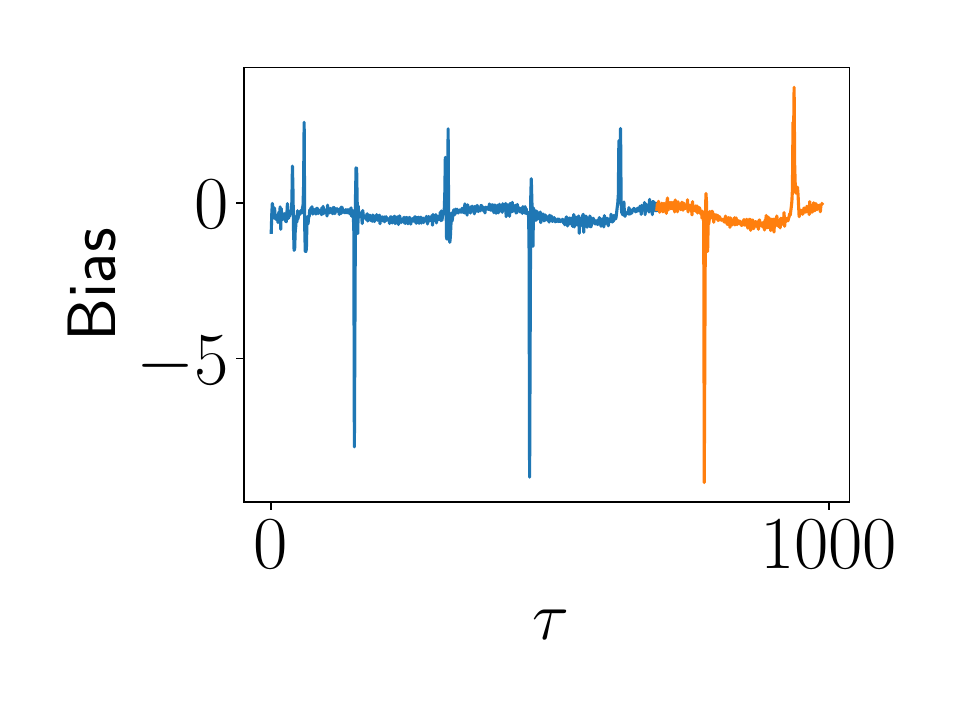} \\

			\includegraphics[trim={0 1cm 0 0.8cm},clip,width=\linewidth,valign=m]{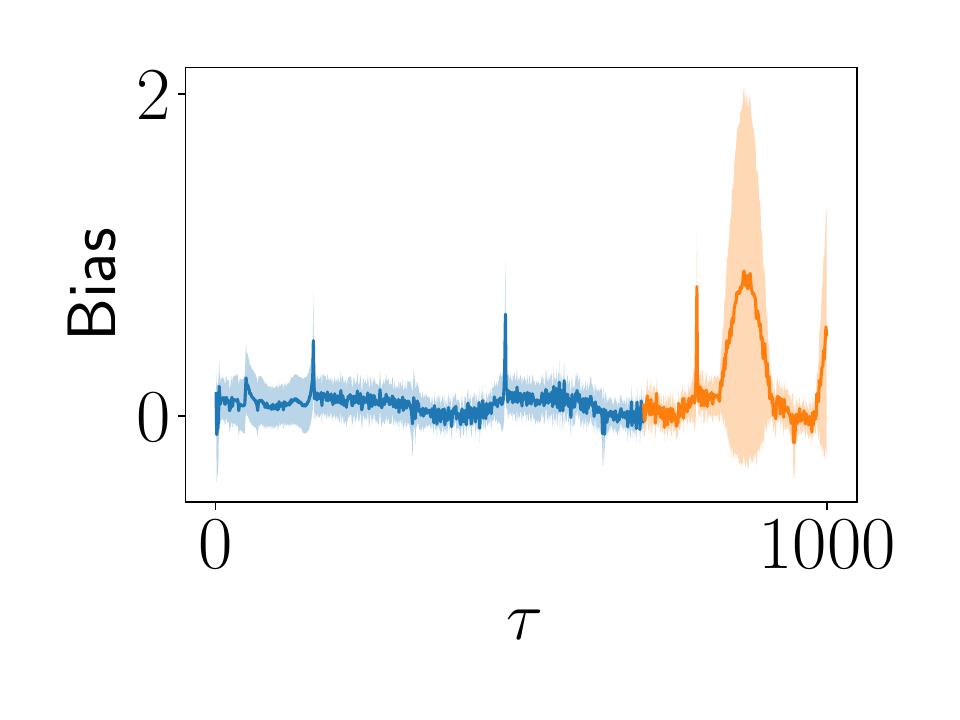}
			& \includegraphics[trim={0 1cm 0 0.8cm},clip,width=\linewidth,valign=m]{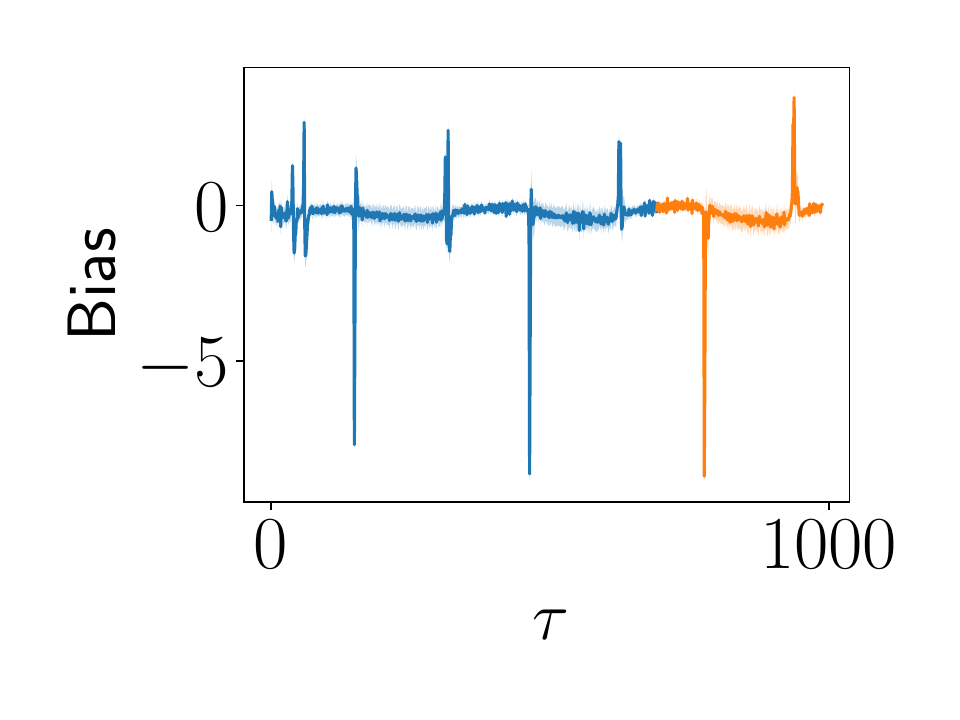} \\

			\multicolumn{2}{c}{\includegraphics[valign=m]{legend_bias.pdf}} \\

			\bottomrule
		\end{tabular}%
	\end{adjustbox}%
	\caption{Bias (\(\hat{q}_\tau - q_\tau\)) for the MLP and DE architectures under {\Fsix}.}%
	\label{tab:F6-bias}%
\end{figure}

\begin{figure}[htb]
	\centering
	\begin{adjustbox}{width=\columnwidth}
		\begin{tabular}{ccc}
			\toprule
			{\Huge BNN} & {\Huge MLP} & {\Huge DE} \\
			\midrule

			\includegraphics[width=\linewidth,valign=m]{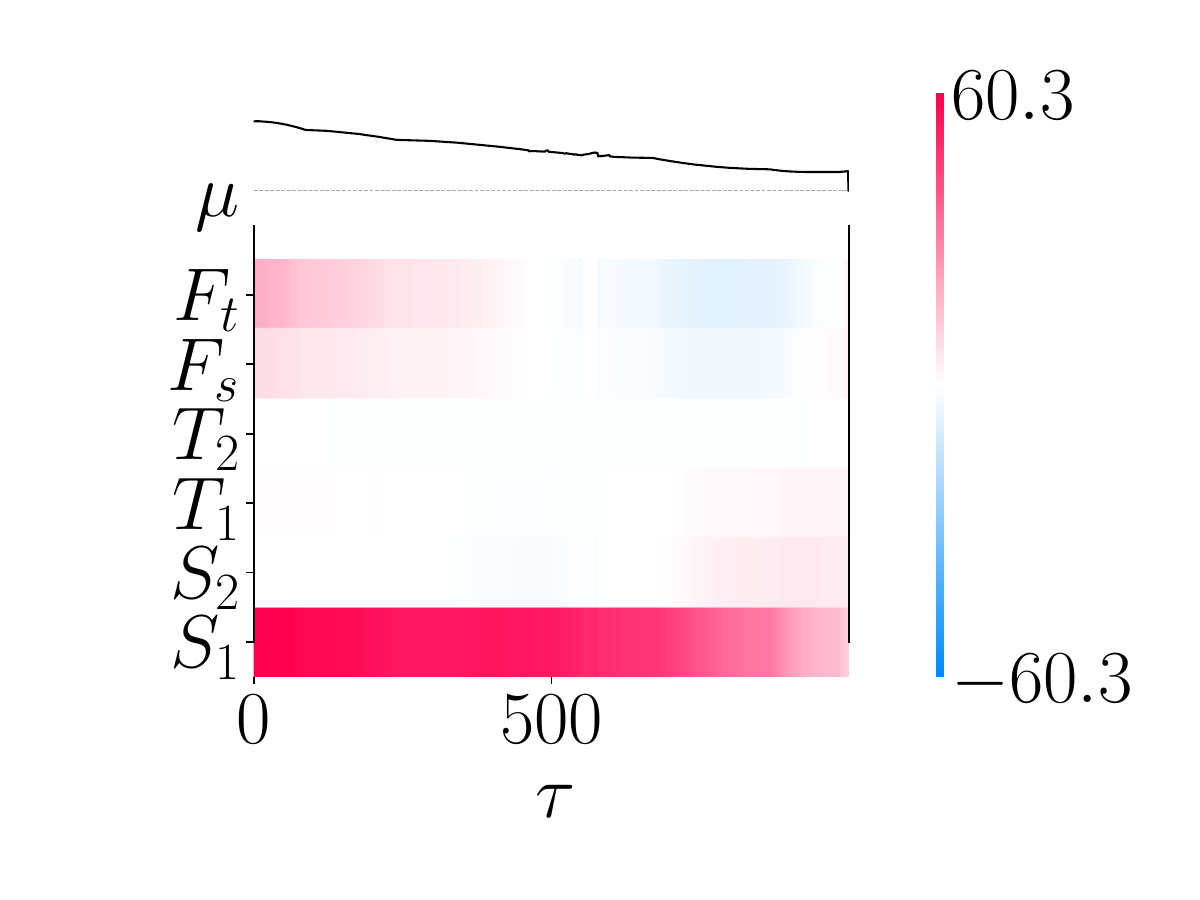}
			& \includegraphics[width=\linewidth,valign=m]{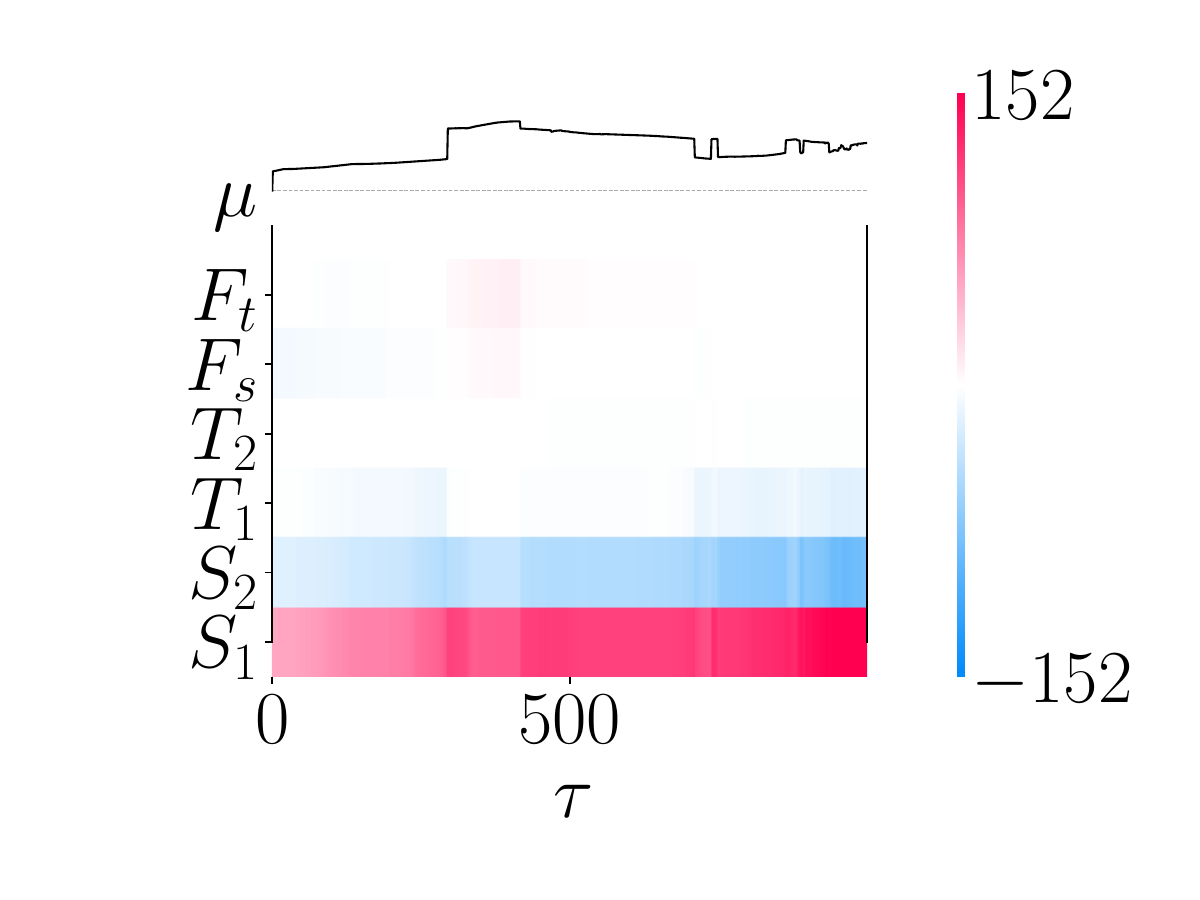}
			& \includegraphics[width=\linewidth,valign=m]{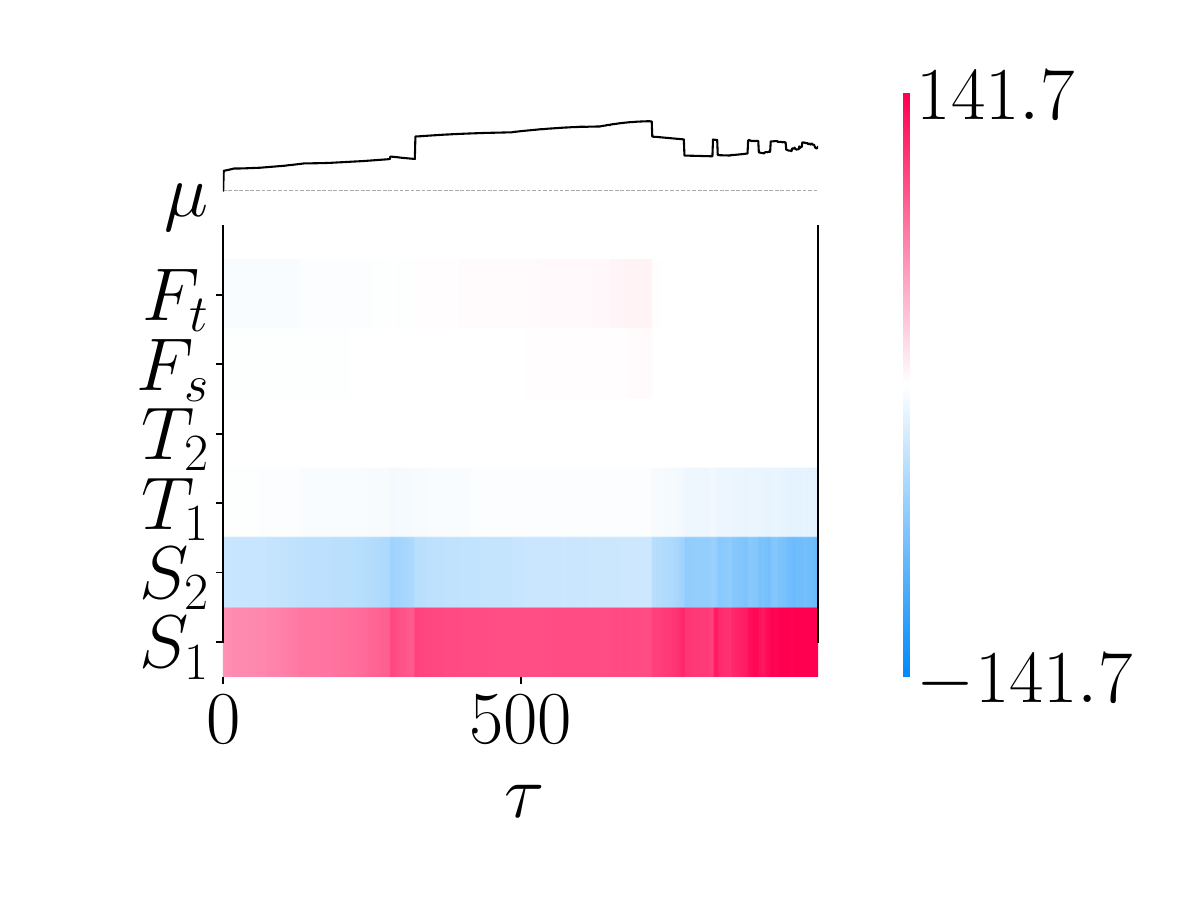} \\

			\includegraphics[trim={0 4cm 0 7cm},clip,width=\linewidth,valign=m]{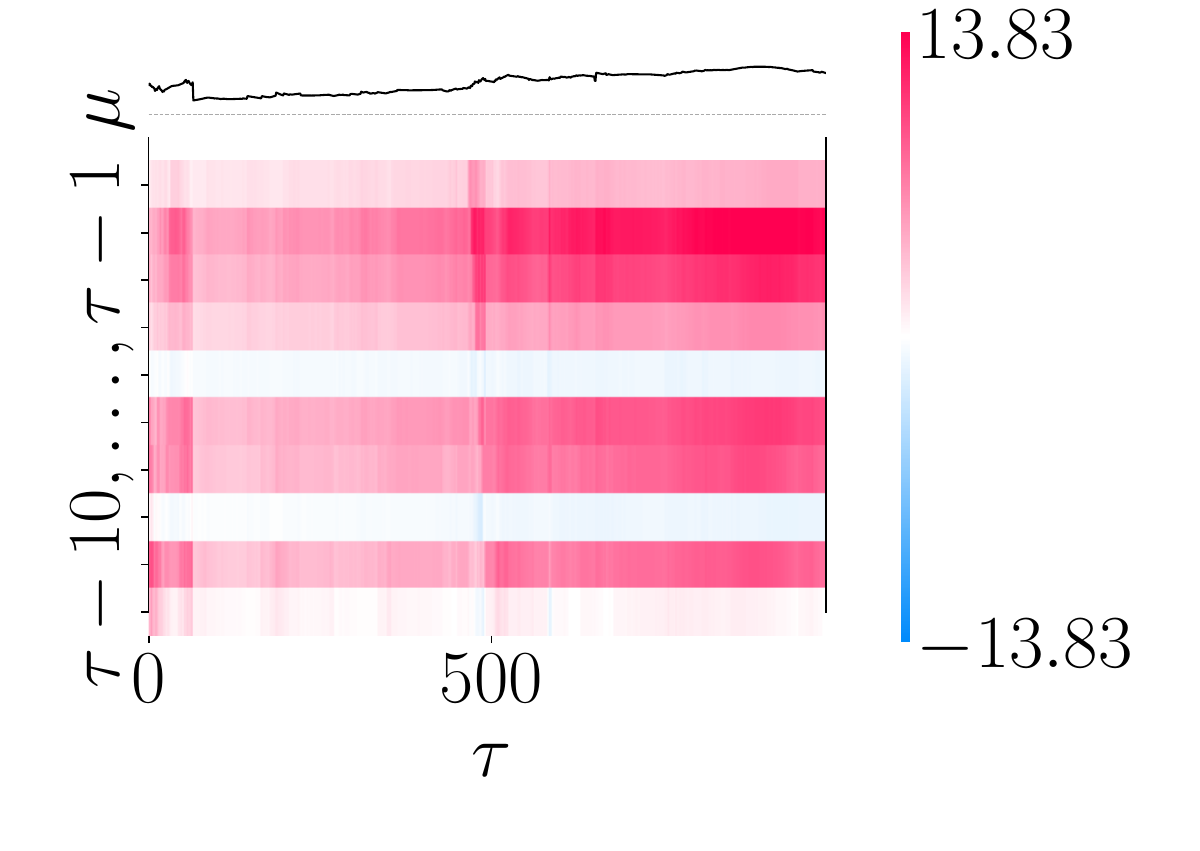}
			& \includegraphics[trim={0 4cm 0 7cm},clip,width=\linewidth,valign=m]{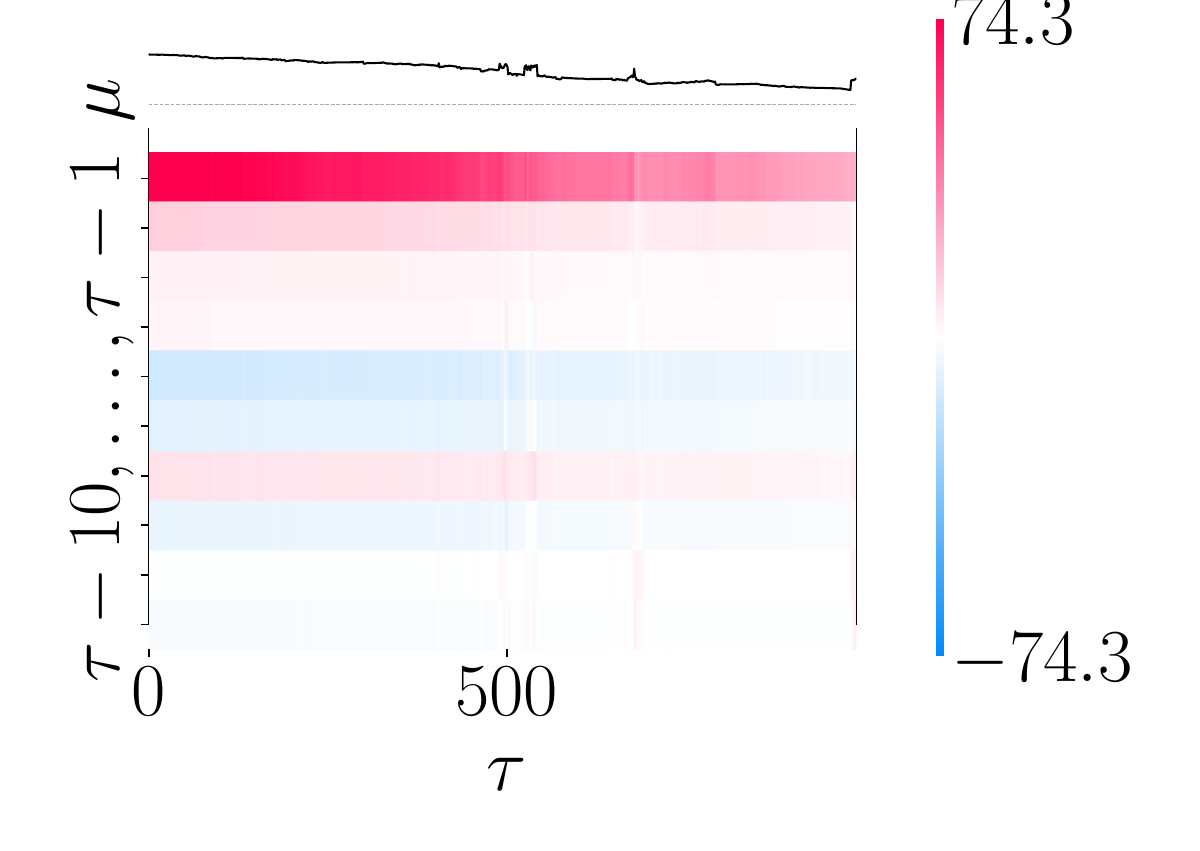}
			& \includegraphics[trim={0 4cm 0 7cm},clip,width=\linewidth,valign=m]{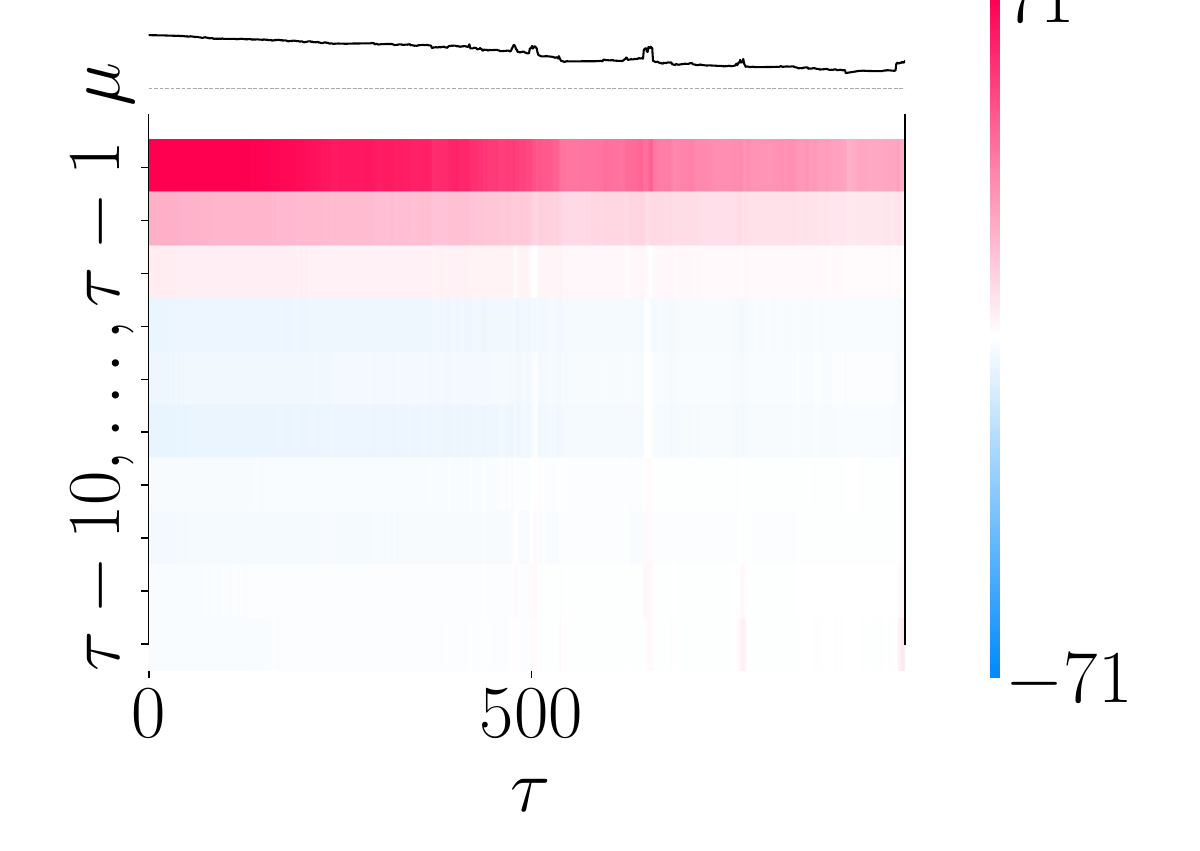} \\

			\bottomrule
		\end{tabular}
	\end{adjustbox}%
	\caption{DeepLIFT attribution maps for the considered architectures using physics-informed (PI\@; top row) and
	autoregressive (AR\@; bottom row) features under {\Fsix}.}%
	\label{tab:F6-xai}%
\end{figure}

Here, we only include the DeepLIFT attribution plots while the SHAP plots are located
in \cref{sec:shap} due to space constraints and their similarity to the DeepLIFT attributions.
The resulting ground-truth-prediction plots after 120 epochs of training are plotted in
\cref{tab:F1-performance,tab:F2-performance,tab:F3-performance,tab:F4-performance,tab:F5-performance,tab:F6-performance}
with DeepLIFT attribution heatmaps being plotted in \cref{tab:F1-xai,tab:F2-xai,tab:F3-xai,tab:F4-xai,tab:F5-xai,tab:F6-xai}.
A vertical dashed line indicates the train/test split (70{\%} / 30{\%}). We supply plots
of the bias \(\hat{q}_\tau - q_\tau\) for \(\{\mathcal{F}_2, \mathcal{F}_3, \mathcal{F}_5, \mathcal{F}_6\}\)
in \cref{tab:F2-bias,tab:F3-bias,tab:F5-bias,tab:F6-bias}.
We conduct additional tuning experiments on the BNN in \cref{sec:bnn-prior}.

In \cref{tab:F1-performance,tab:F1-xai}, we plot the predictions and ground truth with
uncertainty bands under {\Fone}. Immediately, we notice that the BNN architecture results
in highly uncertain predictions with both physics-informed (PI) and autoregressive (AR)
features. This is not the case for the MLP and DE, the latter of which recovers
the best approximation for the validation data. In the physics-informed case, we notice
slight uncertainty where the MLP is producing divergent predictions. Autoregressively,
we notice a small spike in the predicted AMOC (compare bottom row of~\cref{tab:F1-performance})
that is dampened using a DE\@. This is likely indicative of a failed tipping point
capture since these spikes occur just around the recovery of the only tipping point present
in the data. The attribution maps under physics-informed features are fairly similar up to
a sign flip in the attributions, which comes from different random initializations of the
models. Where there are physically plausible attributions in the physics-informed case, the
autoregressive differ significantly between the BNN and the dense architectures; while the
dense architectures primarily use the feature subset \(\{\tau - 3, \dots, \tau - 1\}\), the
BNN appears to pick up spurious correlations even in a simple linear forcing scenario,
indicating that it likely did not develop a plausible physical understanding.

Concerning predictive performance under {\Ftwo} (see~\cref{tab:F2-performance}), we have the
BNN failing to accurately articulate all of the tipping points under stationarity (physics-informed
and autoregressively). Instead, it smoothly approximates breakdown and recovery phases with a
large amount of uncertainty. The MLP and DE perform remarkably well using both PI and
AR data and have seemingly equal performance. However, from~\cref{tab:F2-bias} it becomes clear
that the DE performs better due to a smaller and smoother bias which also reflects in
the attributions (\cref{tab:F2-xai}); smoother transitions in \(\Delta S\) as well as \(\{\tau - 3, \dots, \tau - 1\}\).


Under {\Fthree}, we notice an interesting failure to capture an out-of-sample tipping point
breakdown and its recovery (compare top row of~\cref{tab:F3-performance}) with PI data.
Interestingly, this is consistent among all three architectures and thus raises the question
of whether a system under nonstationary forcing is too complex for neural networks to represent.
However, the DE allows for a slight amount of uncertainty in this particular tipping point. In
the AR case, the BNN generates profusely uncertain outputs during the recovery of a tipping point
and displays a consistent positive shift in the AMOC\@.
Performance between the MLP and Deep
Ensemble appears equal up to a bias (see~\cref{tab:F3-bias}) which is smoothed with PI data
but tends to be larger than the MLP's with AR data. Curiously, the BNN leaves out ranges of
AR features (compare~\cref{tab:F3-xai}, bottom row) in its prediction which could indicate
that it simply learned to approximate a rough estimate of \(q\) while leaving out certain
features. This hypothesis arises due to the dense architectures having most of their attribution
scores distributed among the feature range \(\{\tau - 2, \tau - 1\}\), while attaining good
generalization abilities.

In the more complicated {\Ffour}, which requires significantly more input variables, the BNN
architecture tends to produce a version of the AMOC that either diverges (see~\cref{tab:F4-performance},
top row) or is shifted (see~\cref{tab:F4-performance}, bottom row). This is linked to
a large amount of spuriously picked up features (see~\cref{tab:F4-xai}, first row) with PI and
AR data. In particular, the MLP and DE approximate \(q\) well under {\Ffour} with a
significantly smaller number of attributed features (see~\cref{tab:F4-xai}), which is a likely
indication that these architectures learned the physics of the AMOC\@.

Under {\Ffive}, the BNN architecture significantly simplifies the AMOC and tends to approximate
it by a nearly constant function during recovery periods. This is paired with high uncertainties
during the tipping points for AR data (compare~\cref{tab:F5-performance}). It is likely that this
stems from the BNN picking up additional spurious correlations (see~\cref{tab:F5-xai}, left column)
which the dense architectures do not pick up. Interestingly, it appears that salinity is the main
driver of the AMOC in highly nonlinear and extreme climate scenarios (compare~\cref{tab:F5-xai},
row 2 and 3). In contrast to the BNN architecture, the MLP and DE architectures are good
at handling both PI and AR data, with the ensemble's bias tending towards zero (see~\cref{tab:F5-bias}).
With respect to {\Fsix}, a slightly misaligned tipping point in the validation set is observed
(see~\cref{tab:F6-performance}, top row), although it is dampened by the Ensemble architecture.
With AR data, the BNN appears to have learned a vertically negatively shifted version of the AMOC
(see~\cref{tab:F6-performance}, bottom row) that is not learned by the other architectures. Curiously,
the misaligned tipping point is not revealed by the attribution maps (\cref{tab:F6-xai}). It is,
however, shown that the BNN obtained correlations from a number of features which does not seem
to be necessary to learn an accurate physical understanding of the AMOC\@. A further observation
is that the bias for both MLP and ensemble is the same in this forcing scenario.


We see that adjusting the prior standard deviation (\cref{sec:bnn-prior})
of the BNN leads to a diverse set of outcomes with respect to predictive performance
and explainability. For example, in {\Fone} (see~\cref{fig:pgt-1}) with PI data, none
of the values of \(\sigma\) can produce very accurate predictions. Pathological cases
in which the prediction is near-constant show almost vanishing attributions (\cref{fig:xai-1}~(e,~f,~o,~p,~q,~r)).
Interestingly, the amount of failure cases with near-constant predictions when varying
\(\sigma\) is equal across equal density approximations; in {\Fone}, {\Ftwo} and {\Fthree},
there are exactly three such cases (compare~\cref{fig:pgt-1,fig:pgt-2,fig:pgt-3}) while
in {\Ffour}, {\Ffive} and {\Fsix} there are exactly two (compare~\cref{fig:pgt-4,fig:pgt-5,fig:pgt-6}).
These occur for very low values of \(\sigma\) in which the BNN is unable to sample appropriately.

\section{Conclusion {\&} Future Work}

In this paper, we have addressed the question of whether neural networks are able to
learn the underlying physical system of the AMOC or entirely derive predictive skill
from \enquote{unphysical} or spurious correlations. In both the physics-informed and
autoregressive approach, the BNN architecture is unable to approximate the AMOC sufficiently
well~(compare~\cref{tab:F1-performance,tab:F2-performance,tab:F3-performance,tab:F4-performance,tab:F5-performance,tab:F6-performance}).
It is unlikely that these results make an absolute statement about the capabilities of BNNs
but rather tell us that they are more sensitive to careful hyperparameter tuning than other
architectures. As shown in \cref{sec:bnn-prior}, the BNN is able to capture the AMOC but only
after tuning another parameter. Especially its imposed latent space regularization could be a crucial factor
with respect to its predictive performance. Using the BNN architecture autoregressively leads
to it picking up spurious correlations which is unwanted and indicates failure to learn the
AMOC physics. Such spurious correlations are not seen in the dense architectures, leading us
to the conclusion that these learned a sensible representation of the AMOC's physics. Apart
from this, the BNN appears to produce a consistently negative shift in the \(y\)-axis for
autoregressive data which is again not observed in the MLP and DE\@.
In general, both dense architectures (MLP {\&} DE) generalize better, possibly
due to the better attractor space optimization which is seen especially in the DE\@.
Both architectures' ability to perform well out-of-sample is, other than in the BNNs, another
plausible indicator that they retained AMOC physics during training. Importantly, due to the
spiking behavior near imminent breakdown and recovery phases in AR data, we raise concern
about the catastrophic predictions made using the methodologies of \citet{ditlevsen2023warning}
and emphasize that our approach makes fewer assumptions that would bias an effective AMOC forecast.
In the future, we see the need for experiments on more forcing scenarios, especially those
that represent current greenhouse gas emissions and fresh water forcing in the North Atlantic
Ocean. Besides this, different neural network architectures like the FNO or Graph Neural
Networks (GNNs;~\cite{scarselli2008graph}) could be applied to predict \(q\), however, without
the ready ability to apply XAI\@.

\section*{Acknowledgements}

We thank Mariana Clare and Redouane Lguensat for their helpful comments.

\clearpage

\bibliography{references}
\bibliographystyle{icml2024}

\newpage
\appendix
\onecolumn


\section{Box Model Parameters}%
\label{sec:box-model-parameters}

We show the parameter configuration for the Stommel box model in this paper.

\begin{table}[htb]
	\centering
	\bgroup%
	\renewcommand{\arraystretch}{0.8}
	\begin{tabular}{rr}
		\toprule
		Parameter & Value \\
		\midrule
		\(S_0\) & 35.0 ppt \\
		\(T_0\) & 24.0 \(\tccentigrade\) \\
		\(S_1\) & 12.0 ppt \\
		\(S_2\) & 20.0 ppt \\
		\(T_1\) & 1.0 \(\tccentigrade\) \\
		\(T_2\) & 10.0 \(\tccentigrade\) \\
		\(A\) (area) & \({(5 \times 10^7)} \ \text{m}^2\) \\
		\(k\) & \(1 \times 10^{10}\) \\
		Depth & 4000 m \\
		\(\alpha\) & 0.2 \\
		\(\beta\) & 0.8 \\
		Years of prediction & 150,000 \\
		\bottomrule
	\end{tabular}%
	\egroup%
	\caption{Parameters for the box model used in our analysis.}%
	\label{tab:parameters-box-model}%
\end{table}

\section{SHAP Attributions}%
\label{sec:shap}

We present additional attribution plots given by the SHAP algorithm.
These concur narrowly with the DeepLIFT attribution plots.

\begin{figure}[H]
	\centering
	\begin{adjustbox}{width=\columnwidth}
		\begin{tabular}{ccc}
			\toprule
			{\Huge BNN} & {\Huge MLP} & {\Huge DE} \\
			\midrule

			\includegraphics[valign=m]{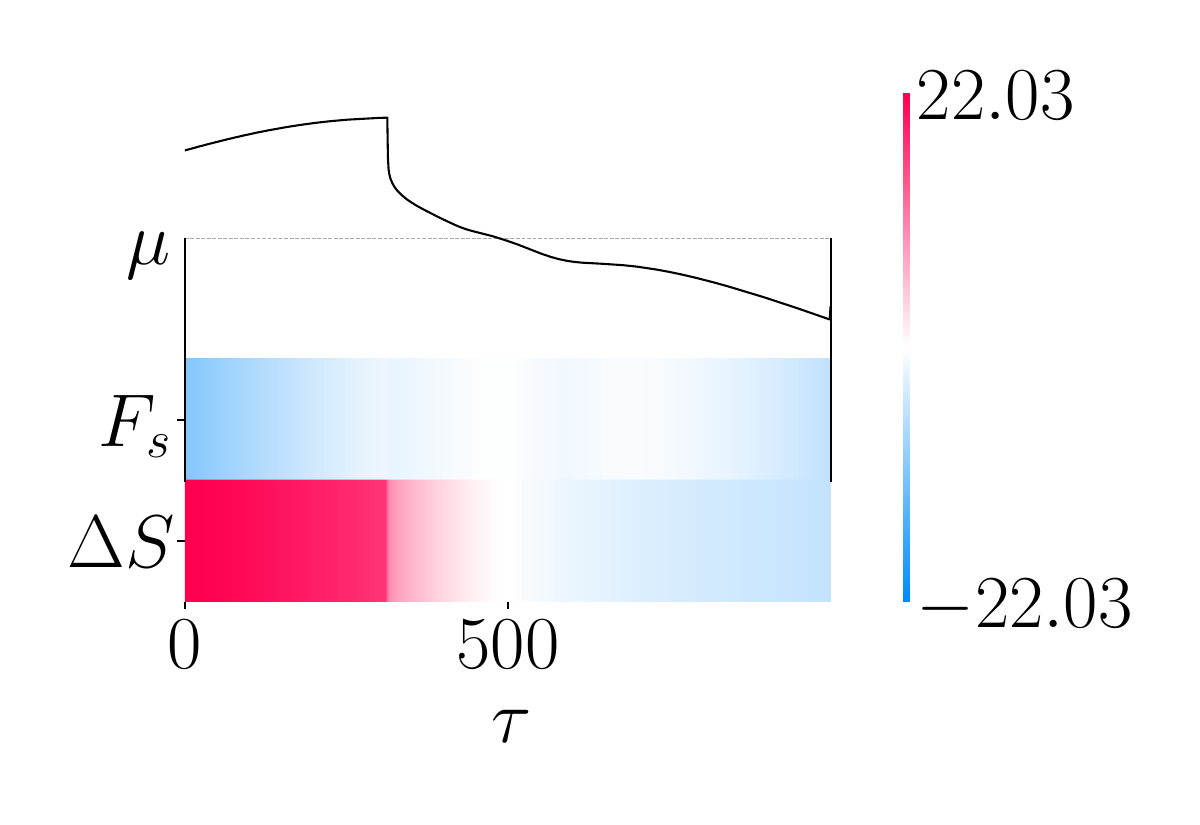}
			& \includegraphics[valign=m]{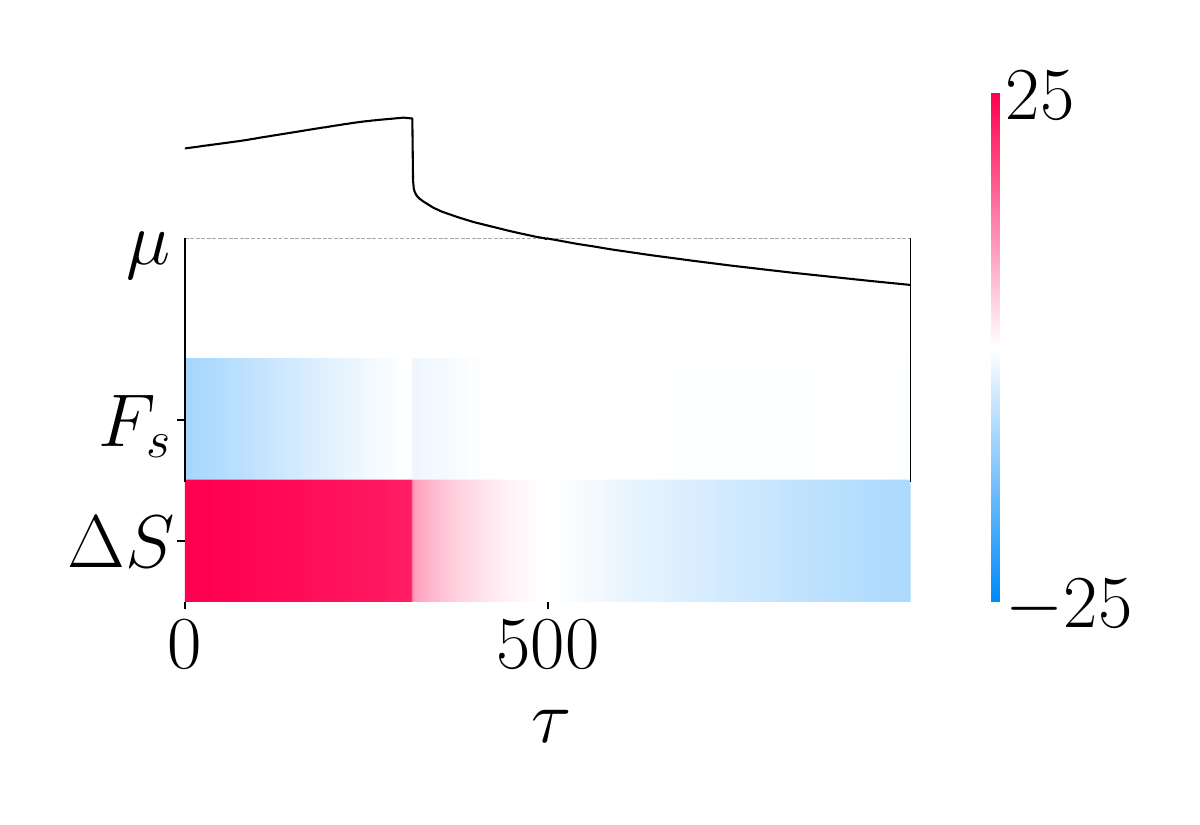}
			& \includegraphics[valign=m]{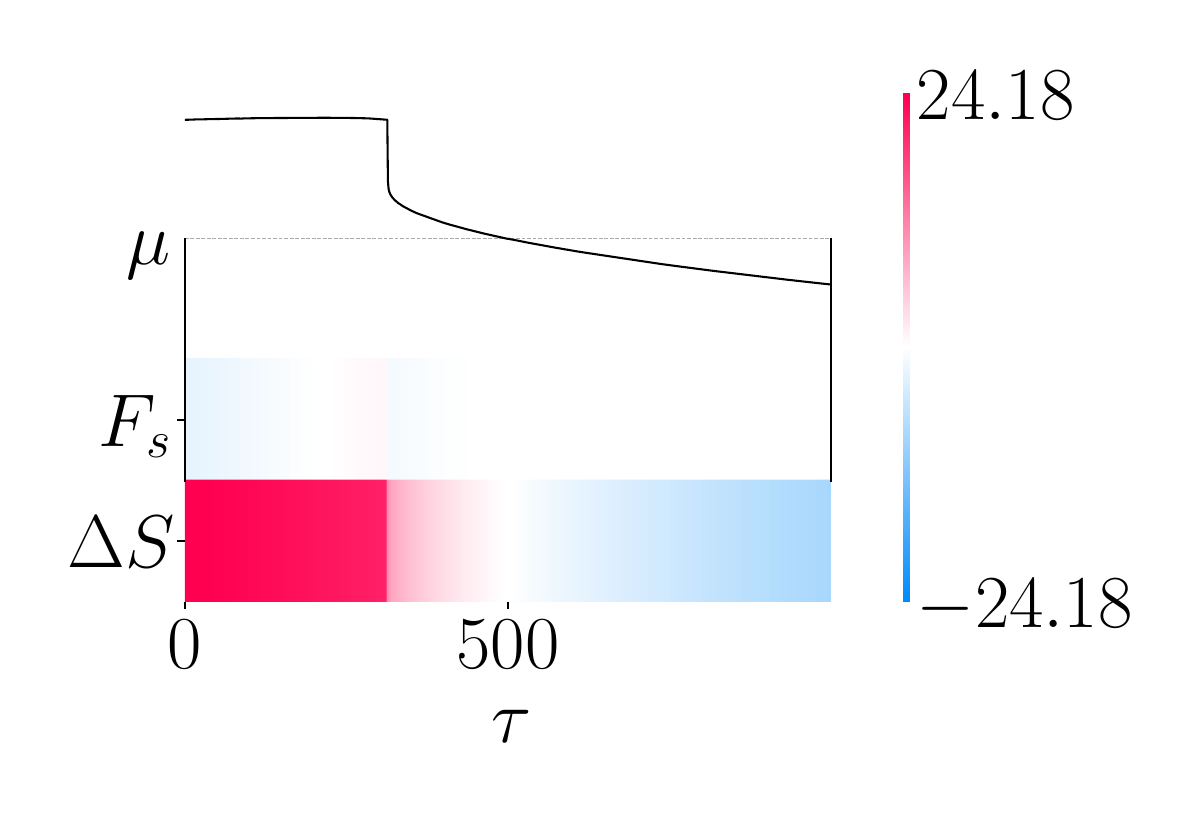} \\

			\includegraphics[trim={0 4cm 0 6cm},clip,valign=m]{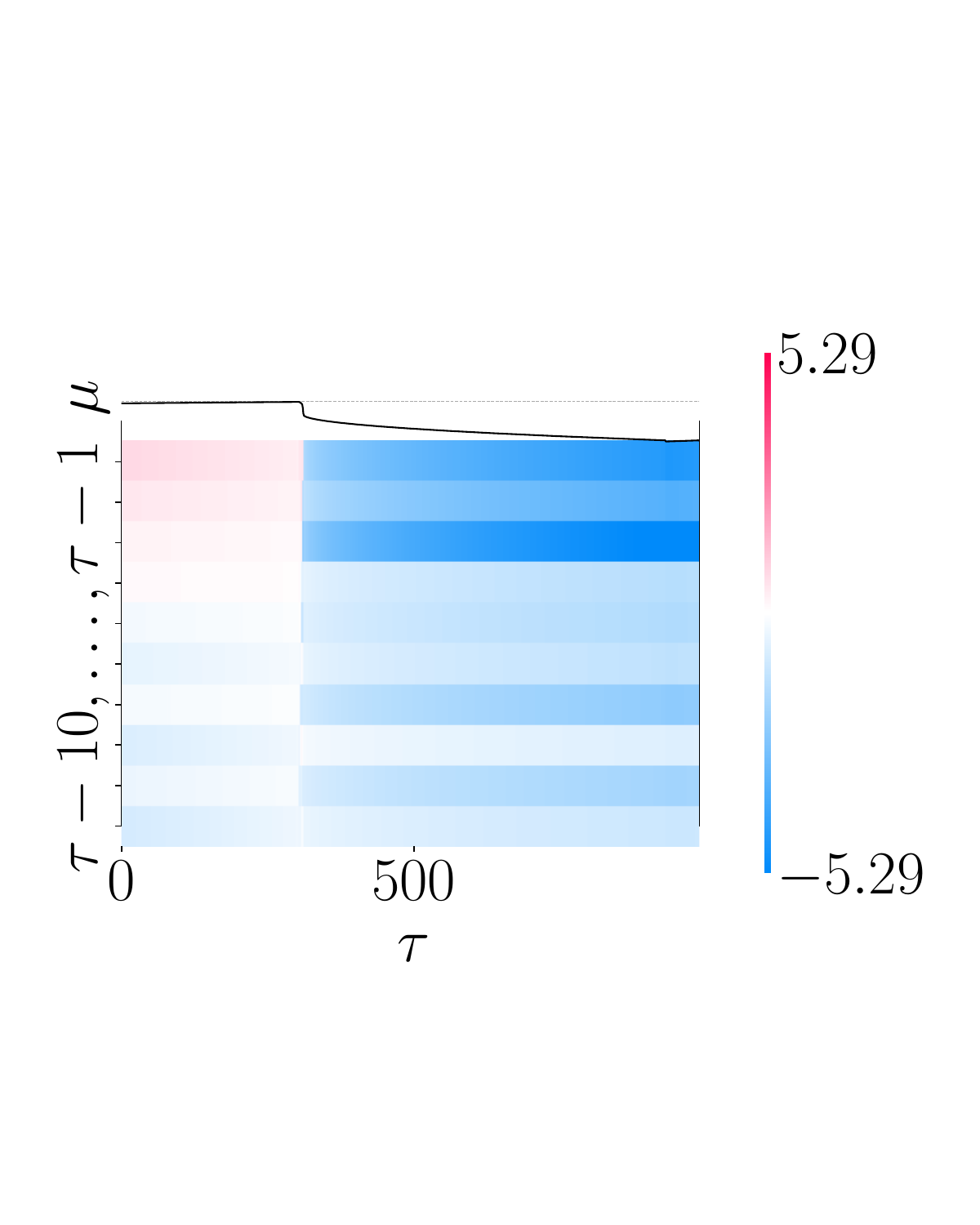}
			& \includegraphics[trim={0 4cm 0 6cm},clip,valign=m]{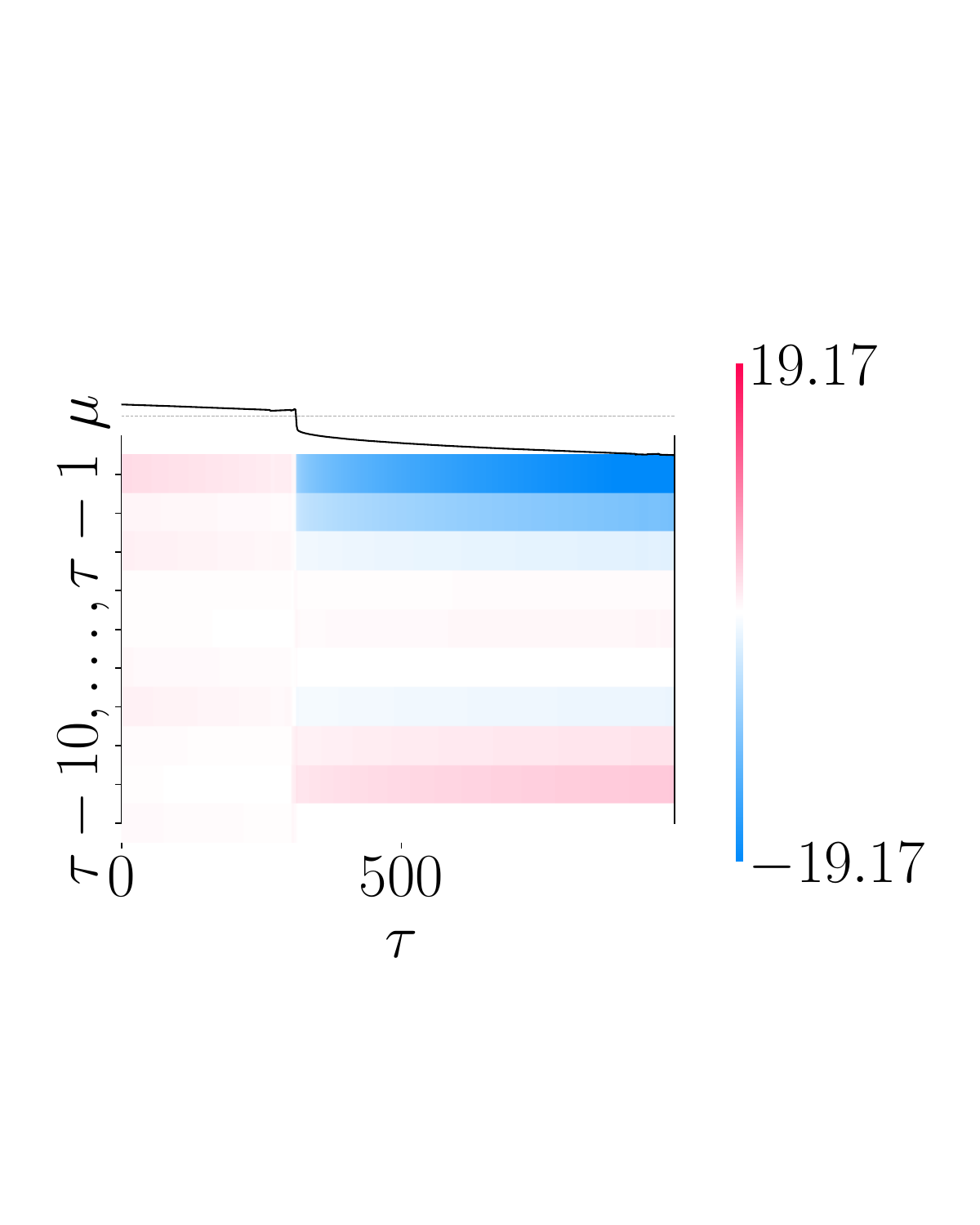}
			& \includegraphics[trim={0 4cm 0 6cm},clip,valign=m]{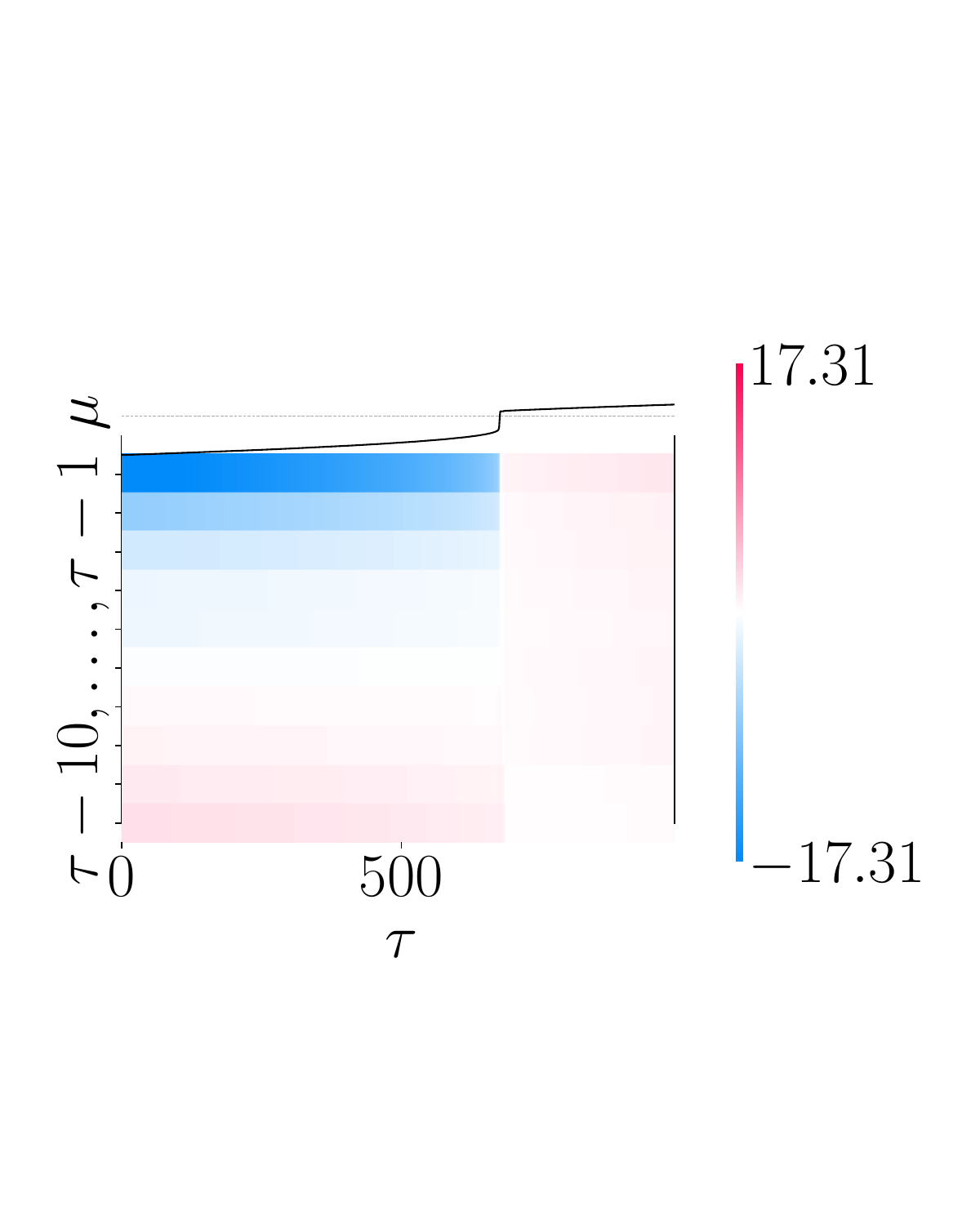} \\
			\bottomrule
		\end{tabular}
	\end{adjustbox}%
	\caption{SHAP attribution maps for the considered architectures using physics-informed (PI\@; top row) and
	autoregressive (AR\@; bottom row) features under {\Fone}.}%
	\label{tab:F1-shap}%
\end{figure}

\begin{figure}[H]
	\centering
	\begin{adjustbox}{width=\columnwidth}
		\begin{tabular}{ccc}
			\toprule
			{\Huge BNN} & {\Huge MLP} & {\Huge DE} \\
			\midrule

			\includegraphics[width=\linewidth,valign=m]{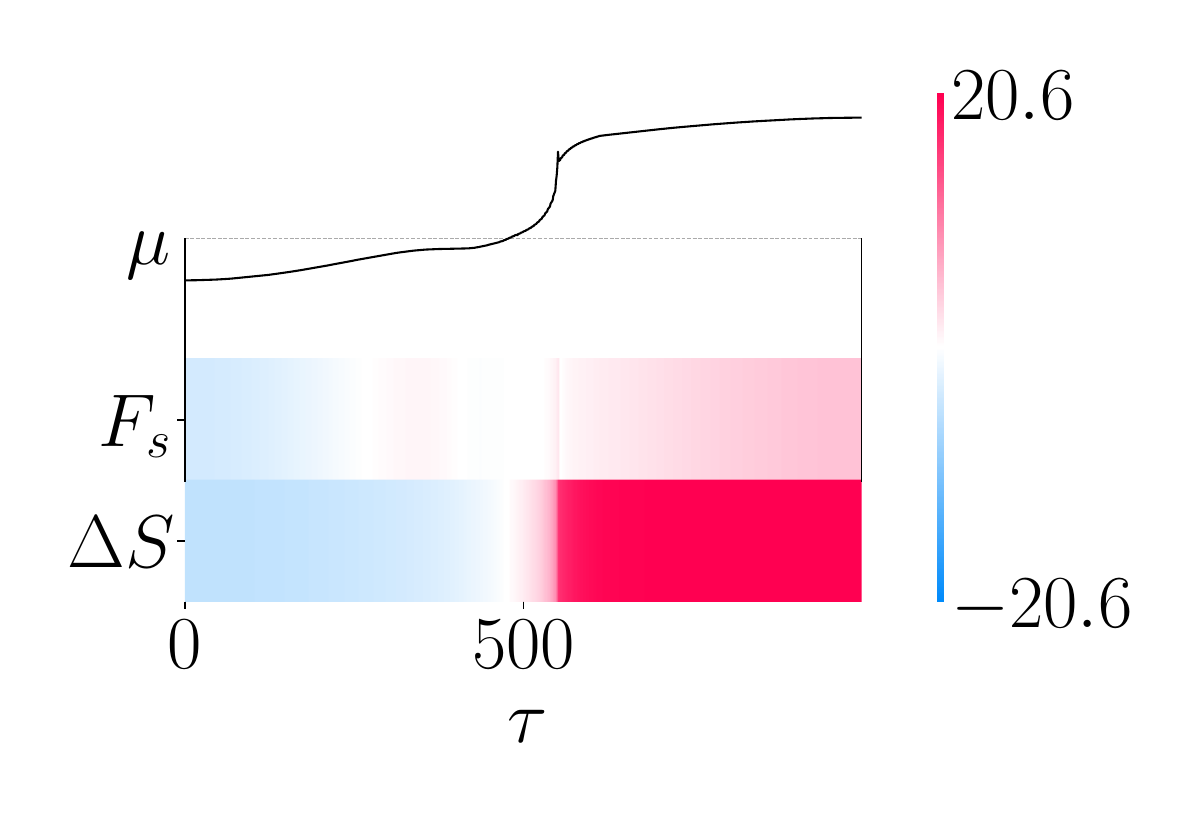}
			& \includegraphics[width=\linewidth,valign=m]{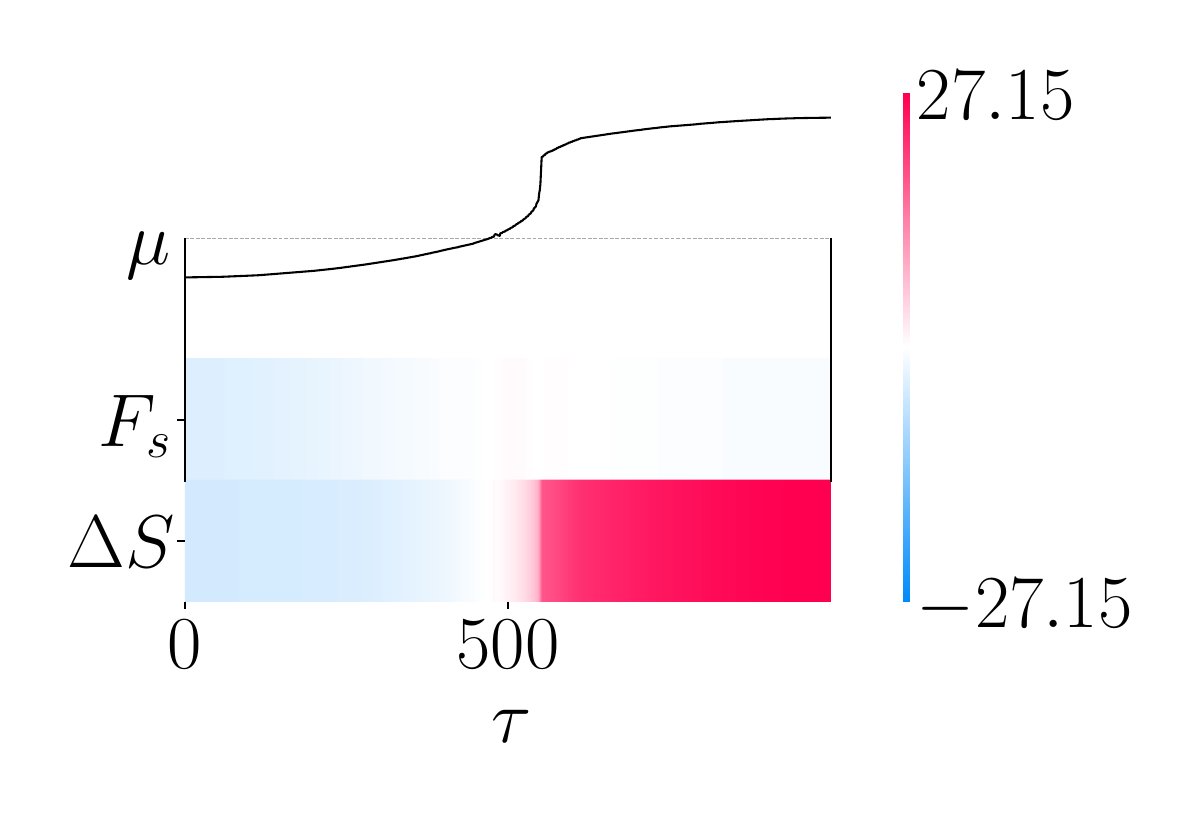}
			& \includegraphics[width=\linewidth,valign=m]{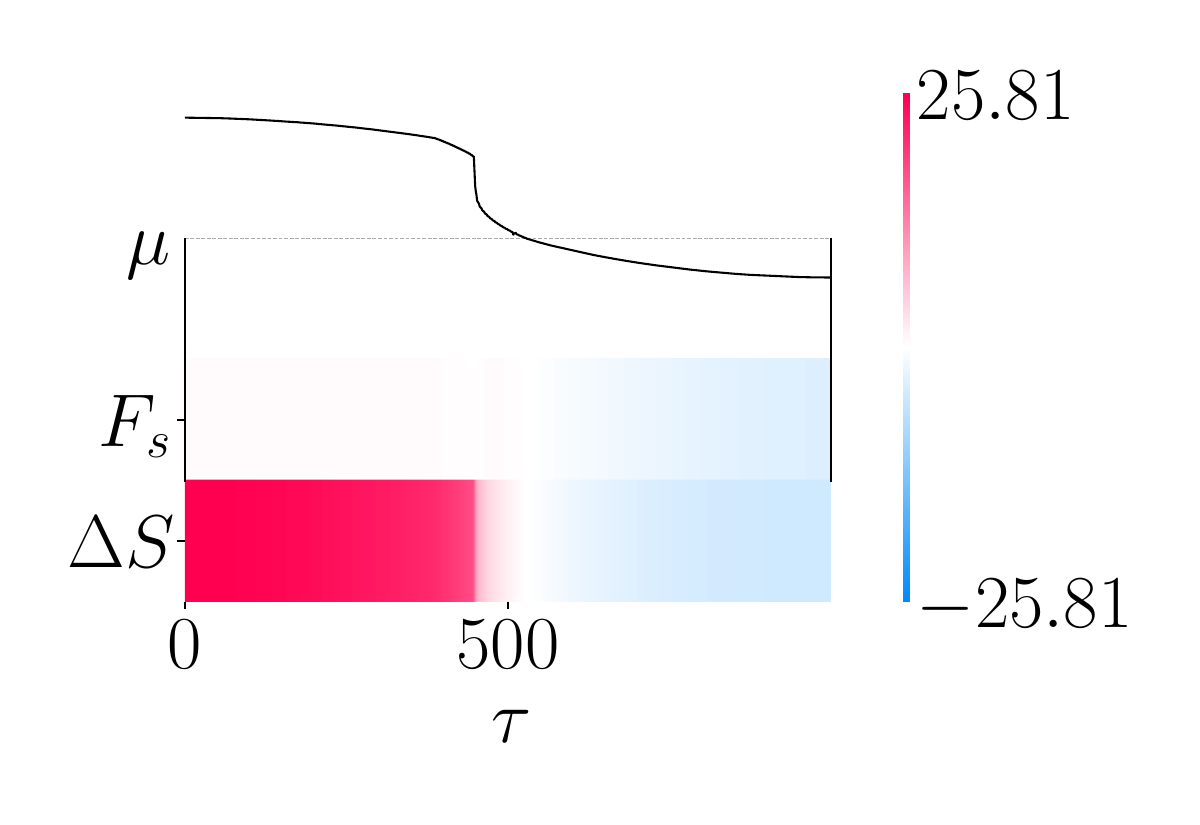} \\
			\includegraphics[trim={0 4cm 0 6cm},clip,width=\linewidth,valign=m]{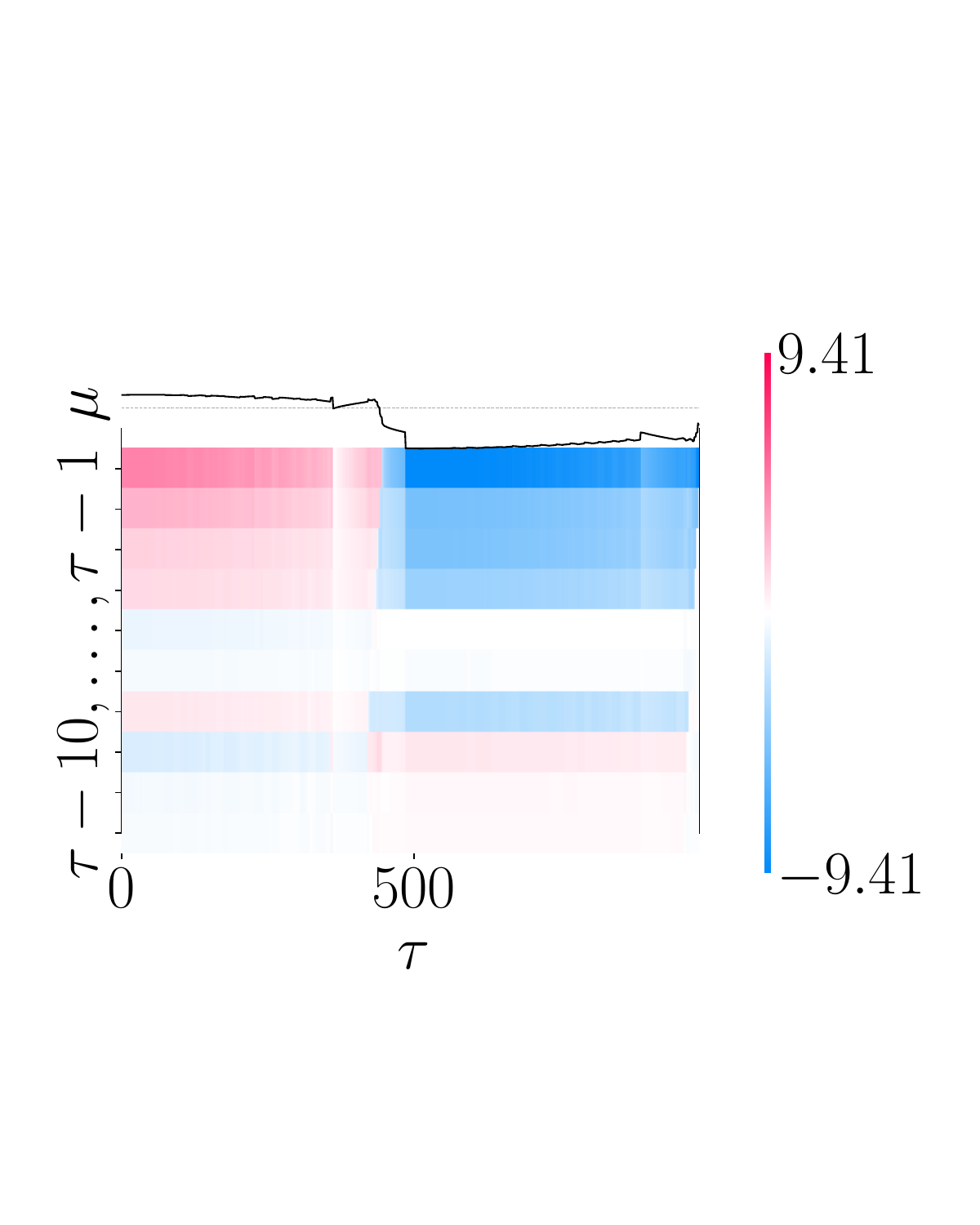}
			& \includegraphics[trim={0 4cm 0 6cm},clip,width=\linewidth,valign=m]{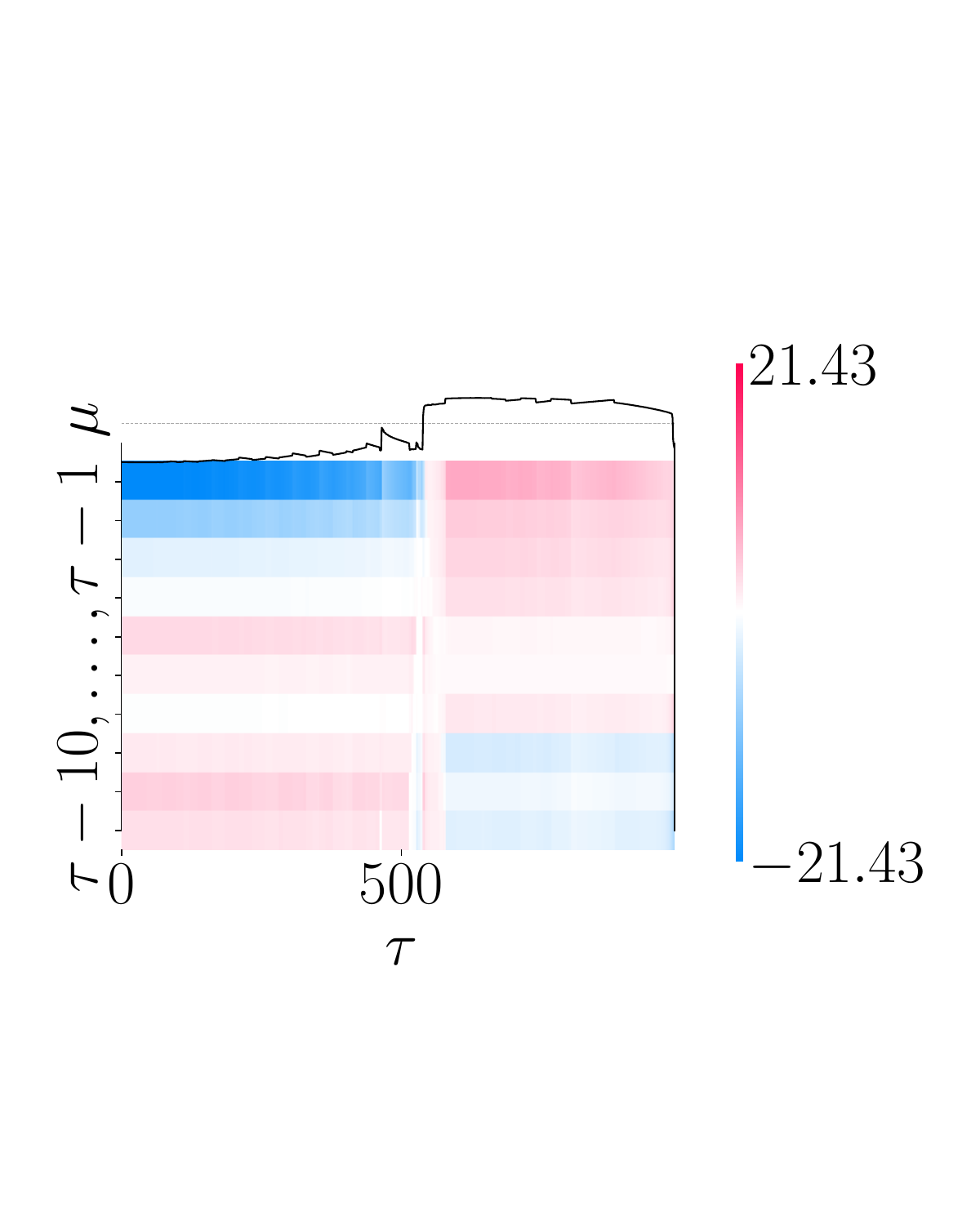}
			& \includegraphics[trim={0 4cm 0 6cm},clip,width=\linewidth,valign=m]{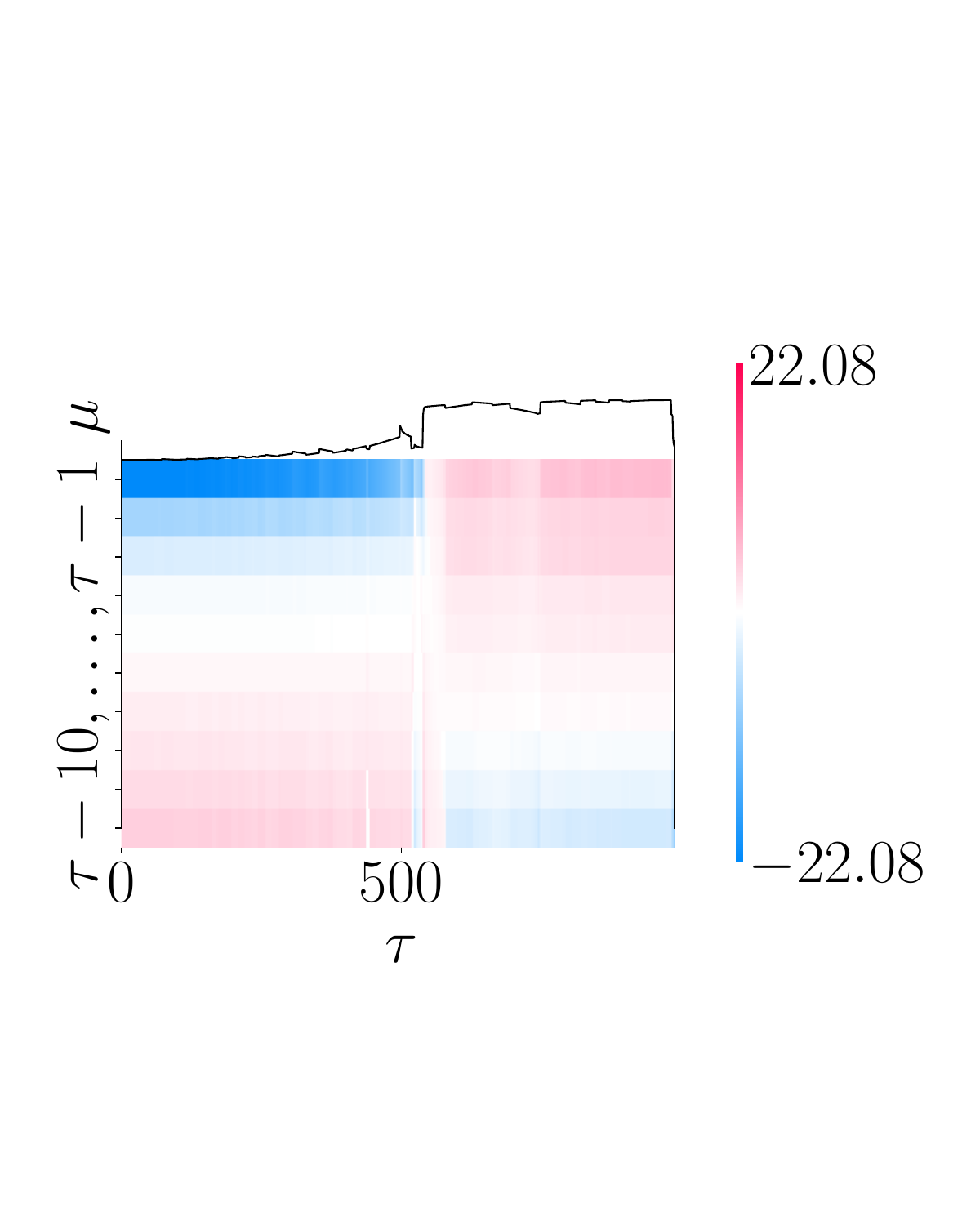} \\

			\bottomrule
		\end{tabular}
	\end{adjustbox}%
	\caption{SHAP attribution maps for the considered architectures using physics-informed (PI\@; top row) and
	autoregressive (AR\@; bottom row) features under {\Ftwo}.}%
	\label{tab:F2-shap}%
\end{figure}

\begin{figure}[H]
	\centering
	\begin{adjustbox}{width=\columnwidth}
		\begin{tabular}{ccc}
			\toprule
			{\Huge BNN} & {\Huge MLP} & {\Huge DE} \\
			\midrule

			\includegraphics[width=\linewidth,valign=m]{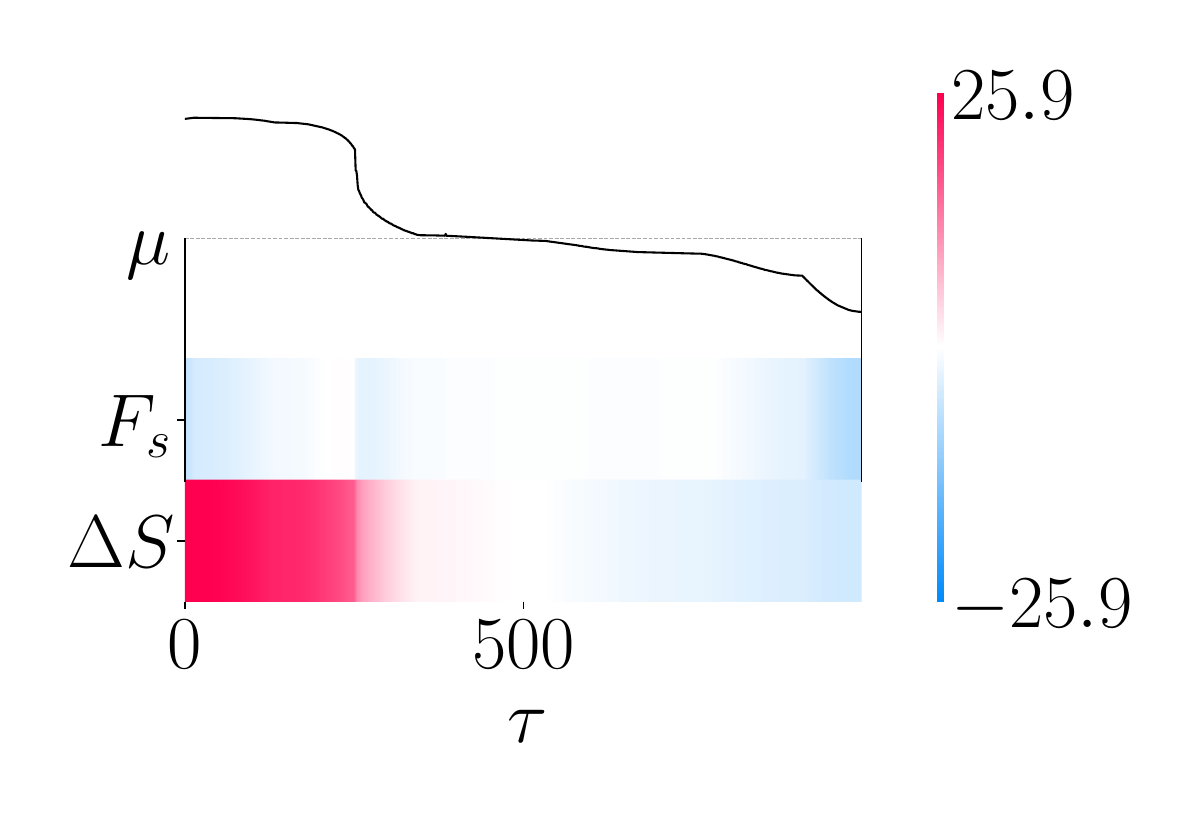}
			& \includegraphics[width=\linewidth,valign=m]{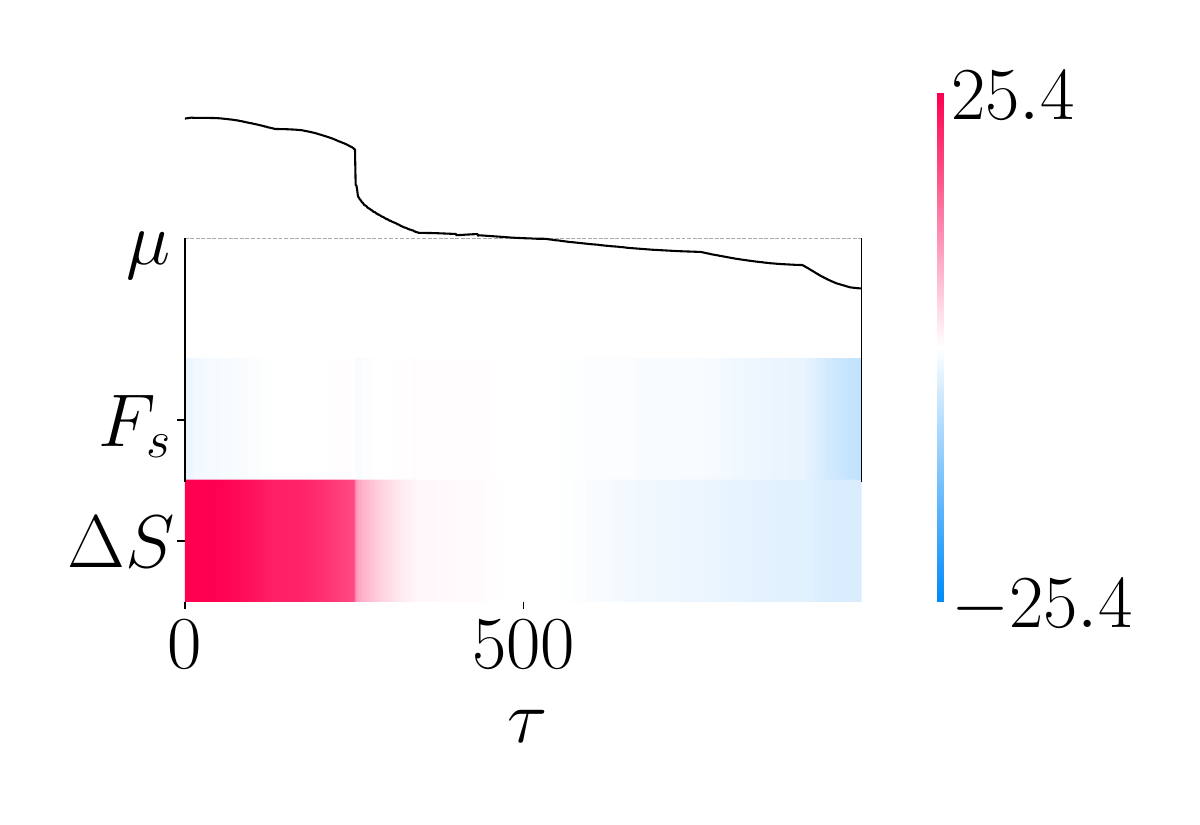}
			& \includegraphics[width=\linewidth,valign=m]{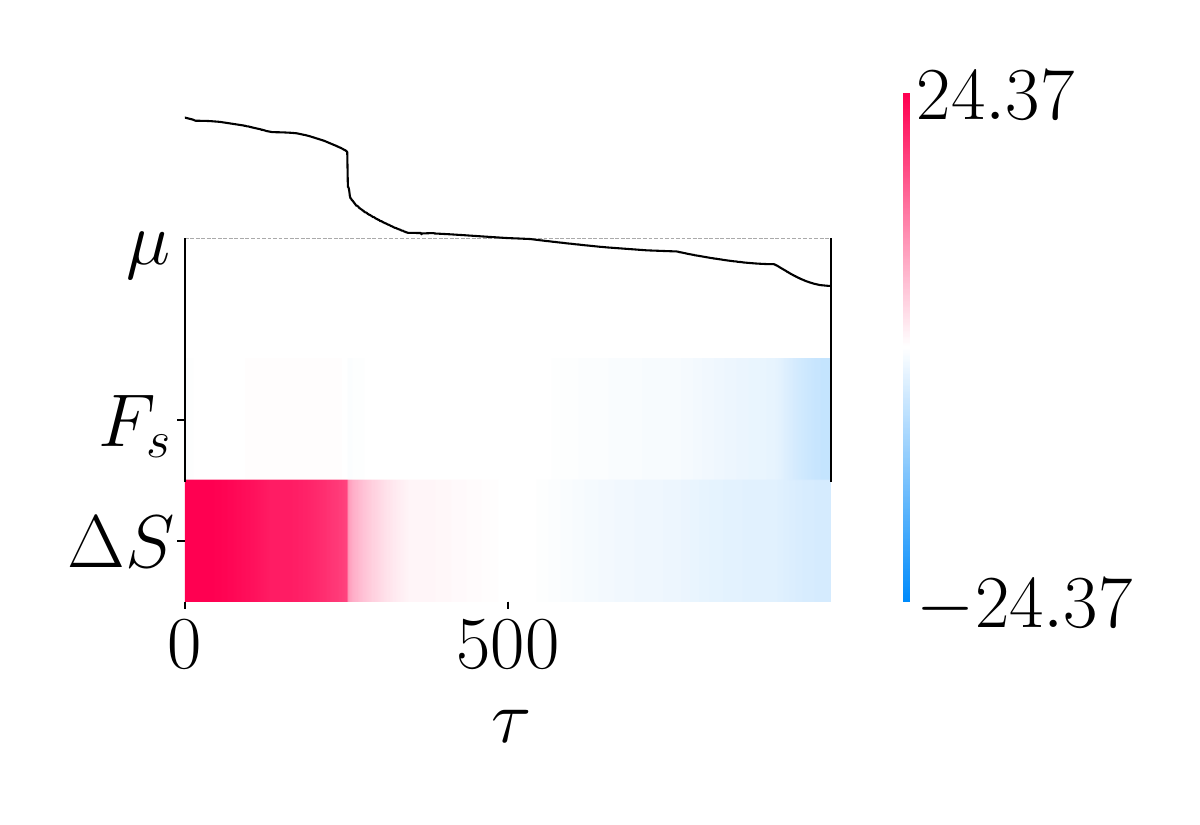} \\

			\includegraphics[trim={0 4cm 0 6cm},clip,width=\linewidth,valign=m]{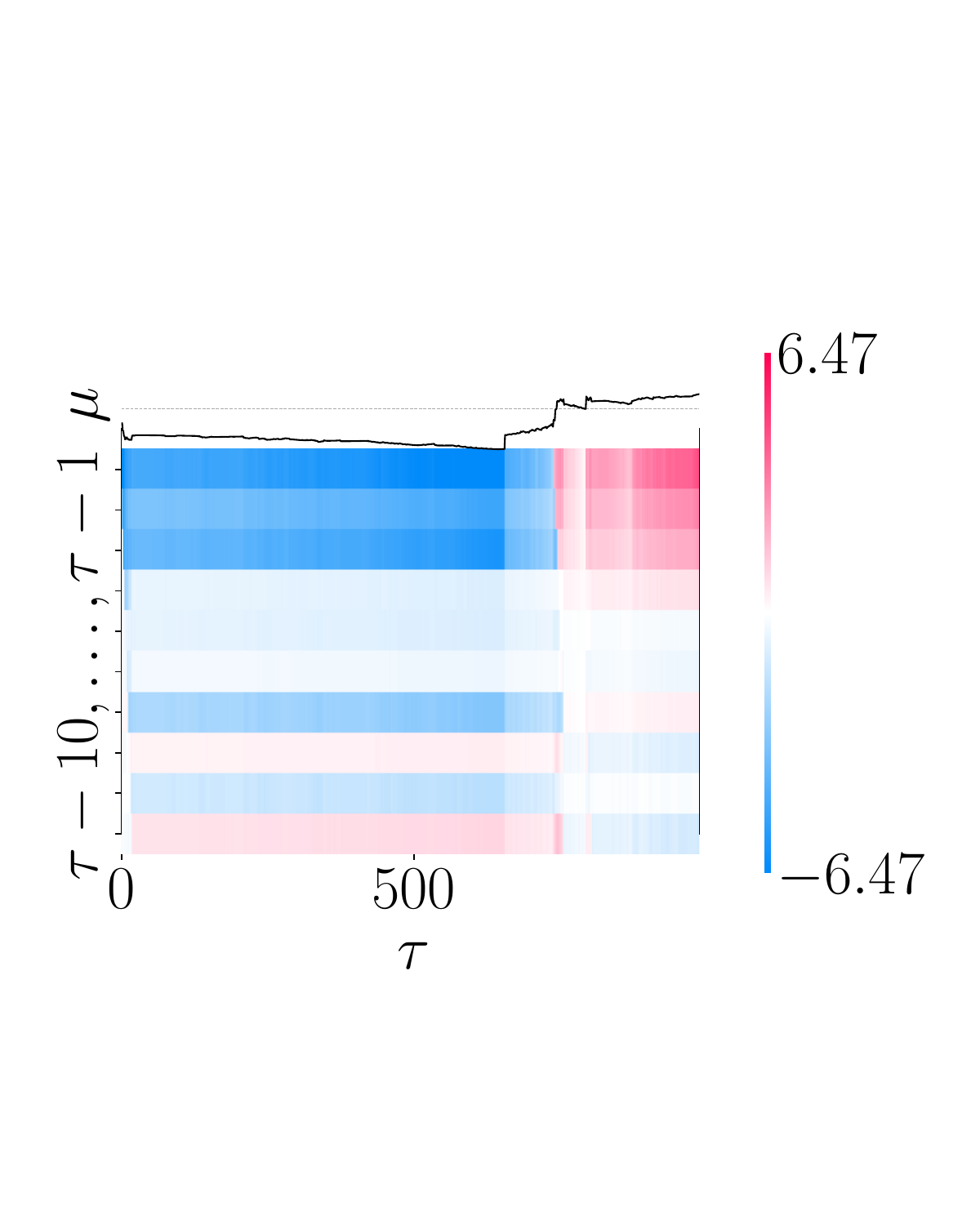}
			& \includegraphics[trim={0 4cm 0 6cm},clip,width=\linewidth,valign=m]{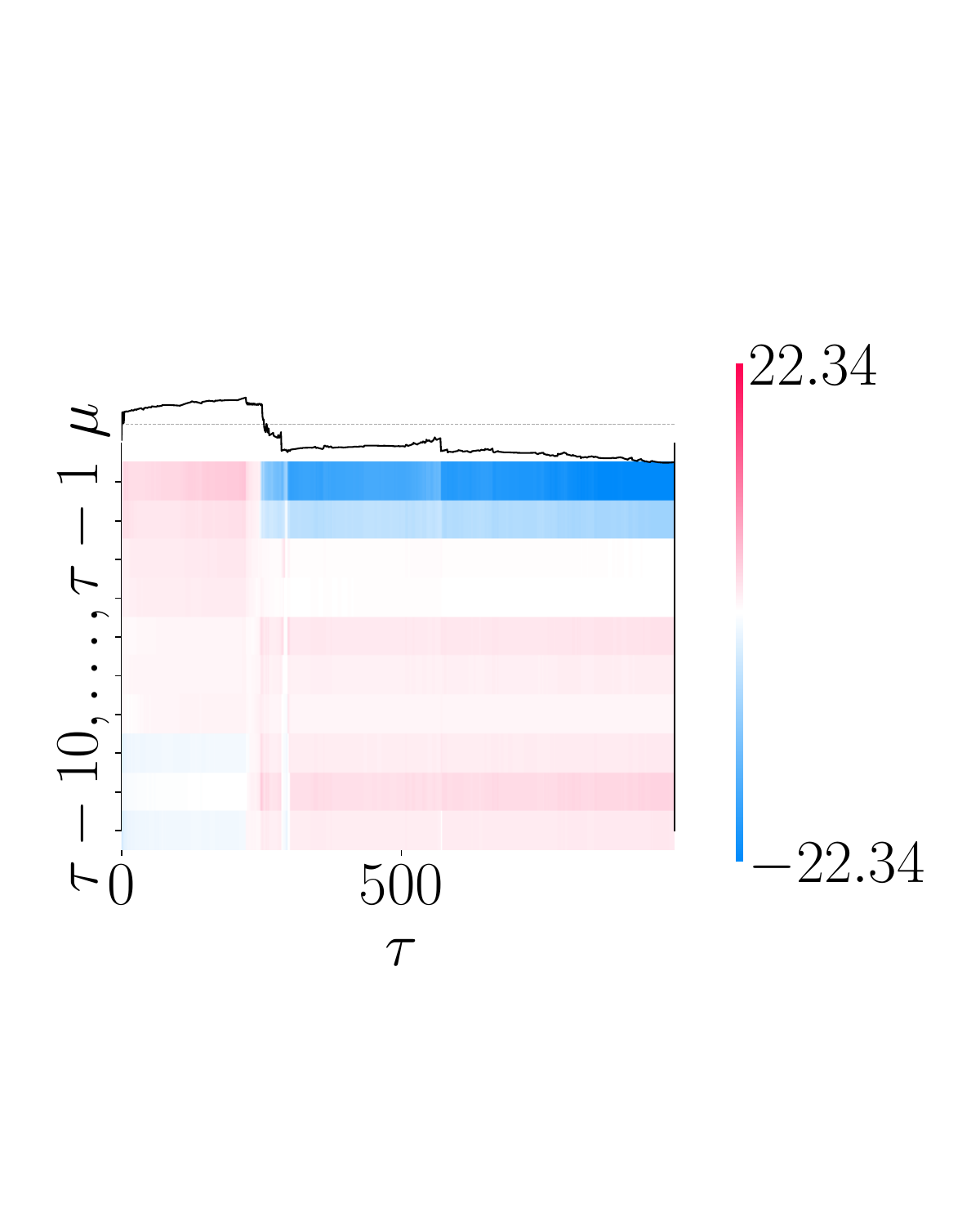}
			& \includegraphics[trim={0 4cm 0 6cm},clip,width=\linewidth,valign=m]{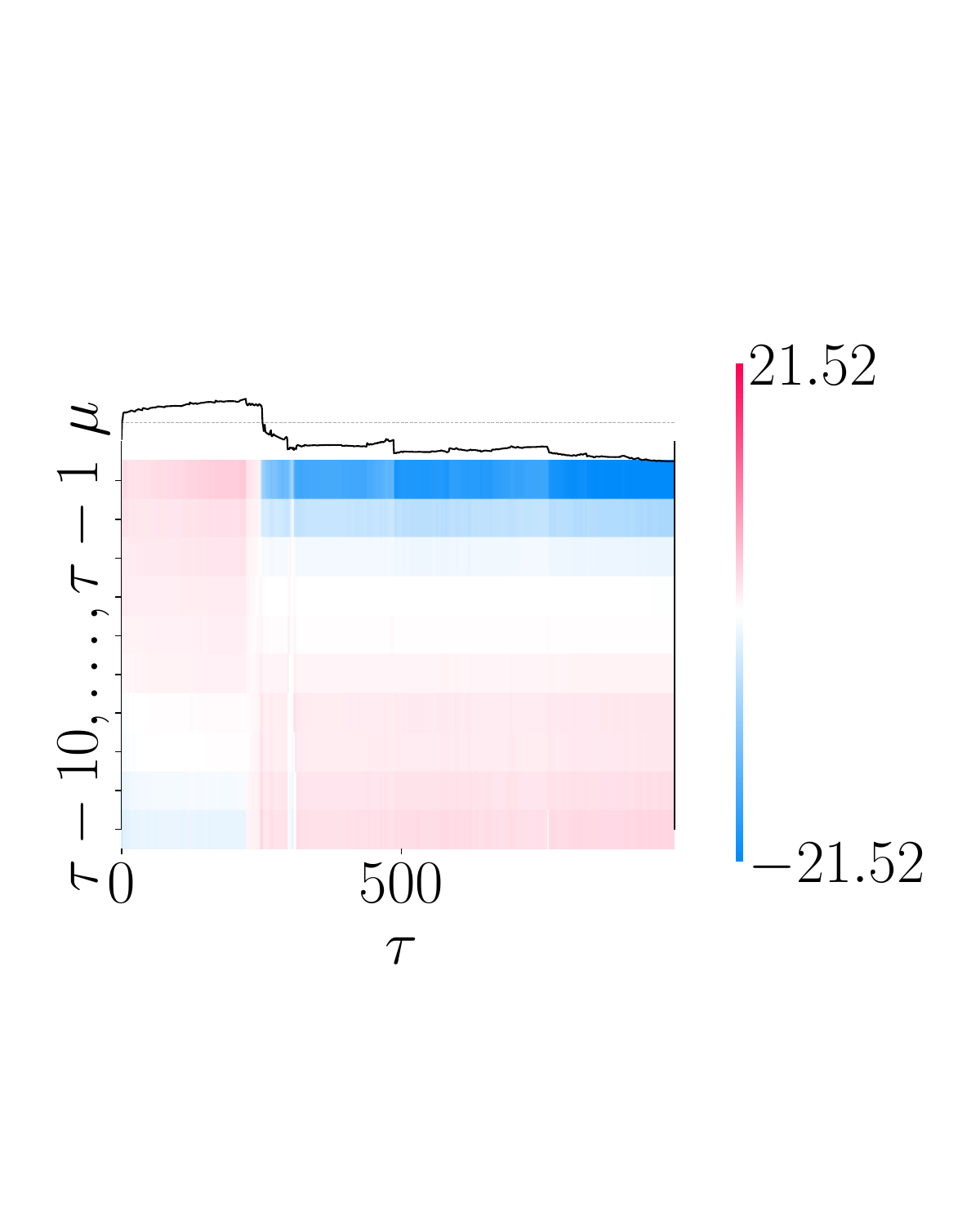} \\

			\bottomrule
		\end{tabular}
	\end{adjustbox}%
	\caption{SHAP attribution maps for the considered architectures using physics-informed (PI\@; top row) and
	autoregressive (AR\@; bottom row) features under {\Fthree}.}%
	\label{tab:F3-shap}%
\end{figure}

\begin{figure}[H]
	\centering
	\begin{adjustbox}{width=\columnwidth}
		\begin{tabular}{ccc}
			\toprule
			{\Huge BNN} & {\Huge MLP} & {\Huge DE} \\
			\midrule

			\includegraphics[width=1.6\linewidth,valign=m]{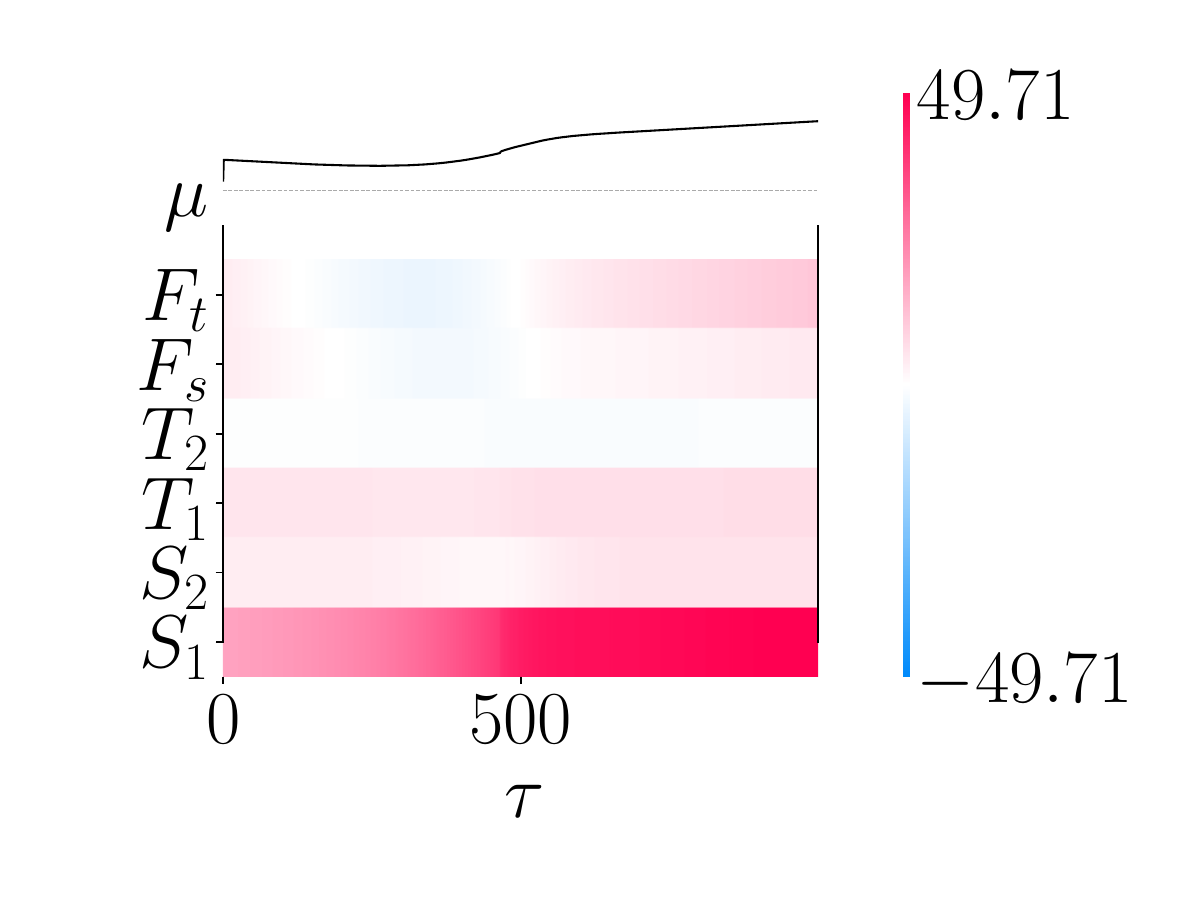}
			& \includegraphics[width=1.6\linewidth,valign=m]{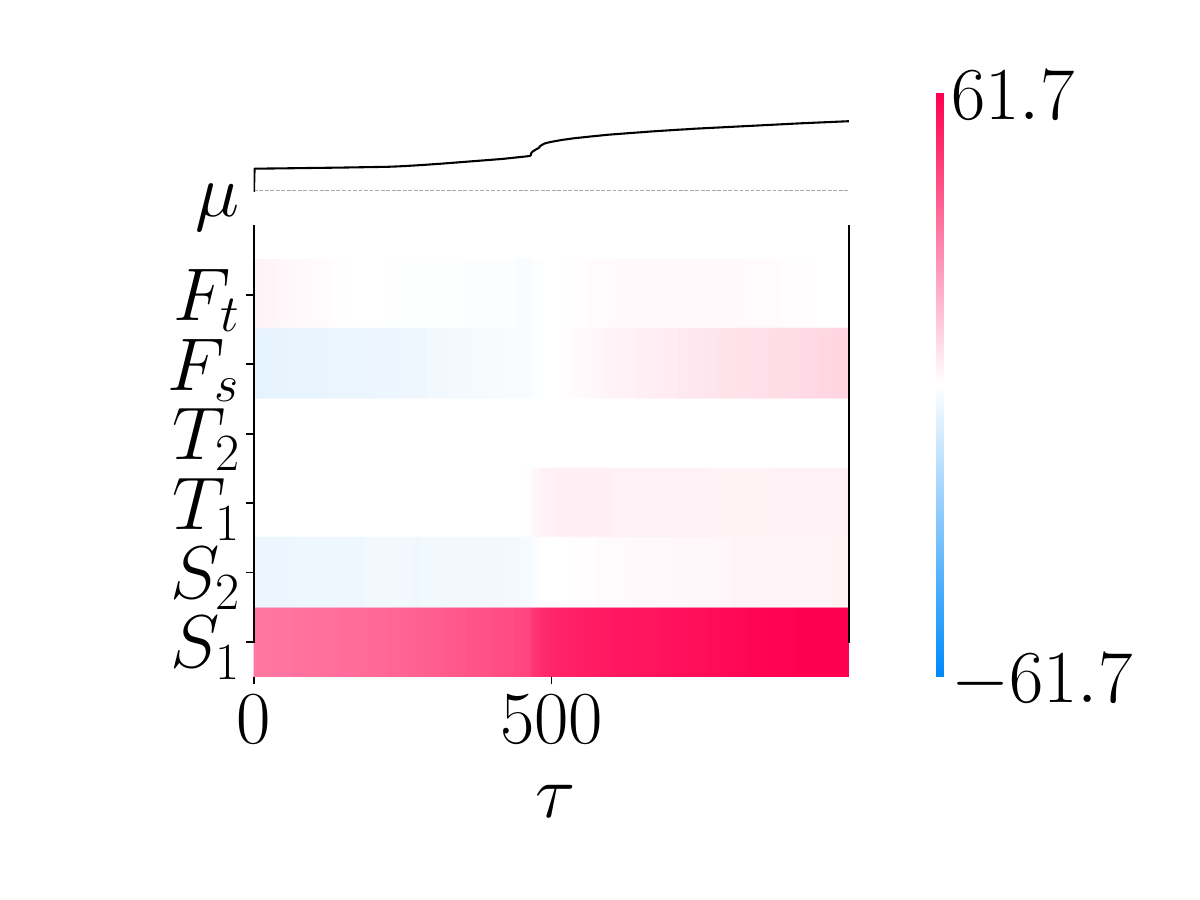}
			& \includegraphics[width=1.6\linewidth,valign=m]{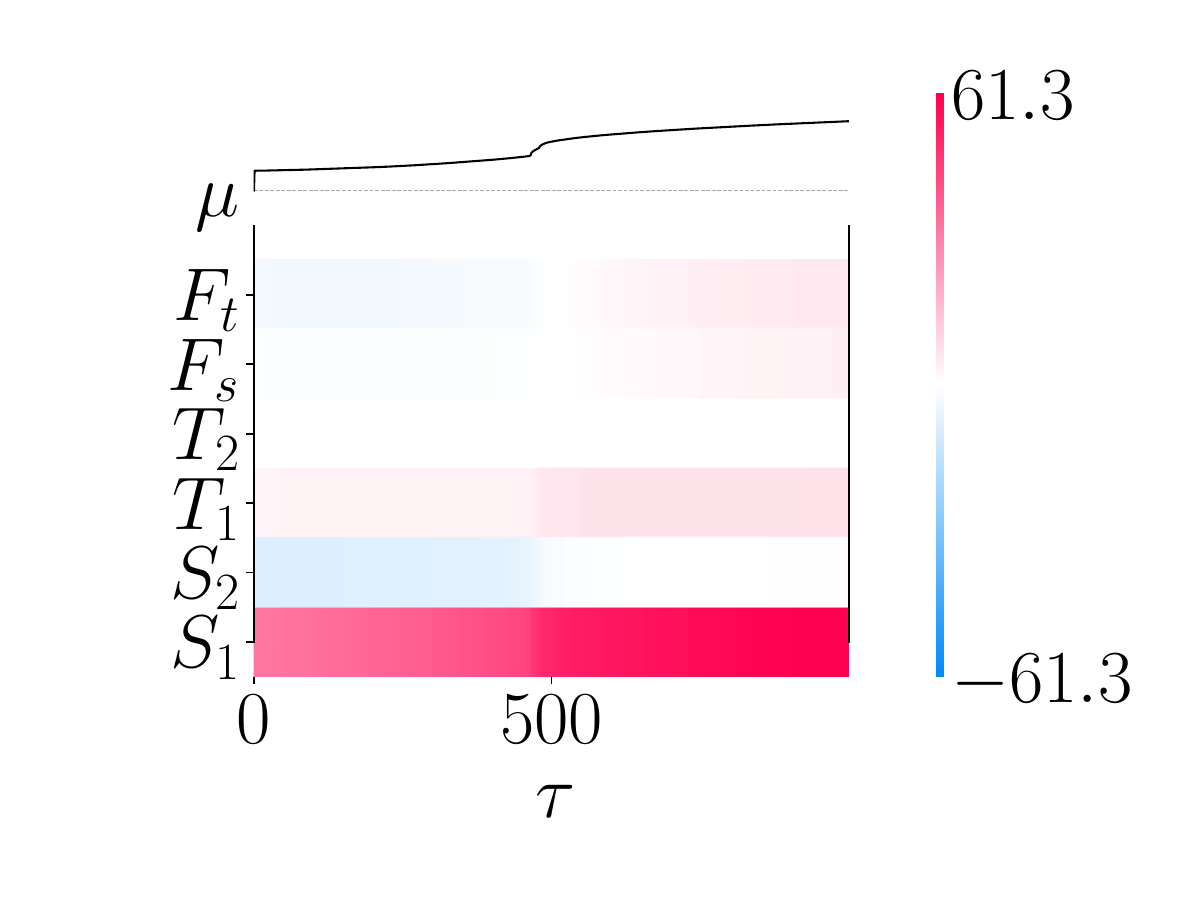} \\

			\includegraphics[trim={0 4cm 0 8cm},clip,width=1.6\linewidth,valign=m]{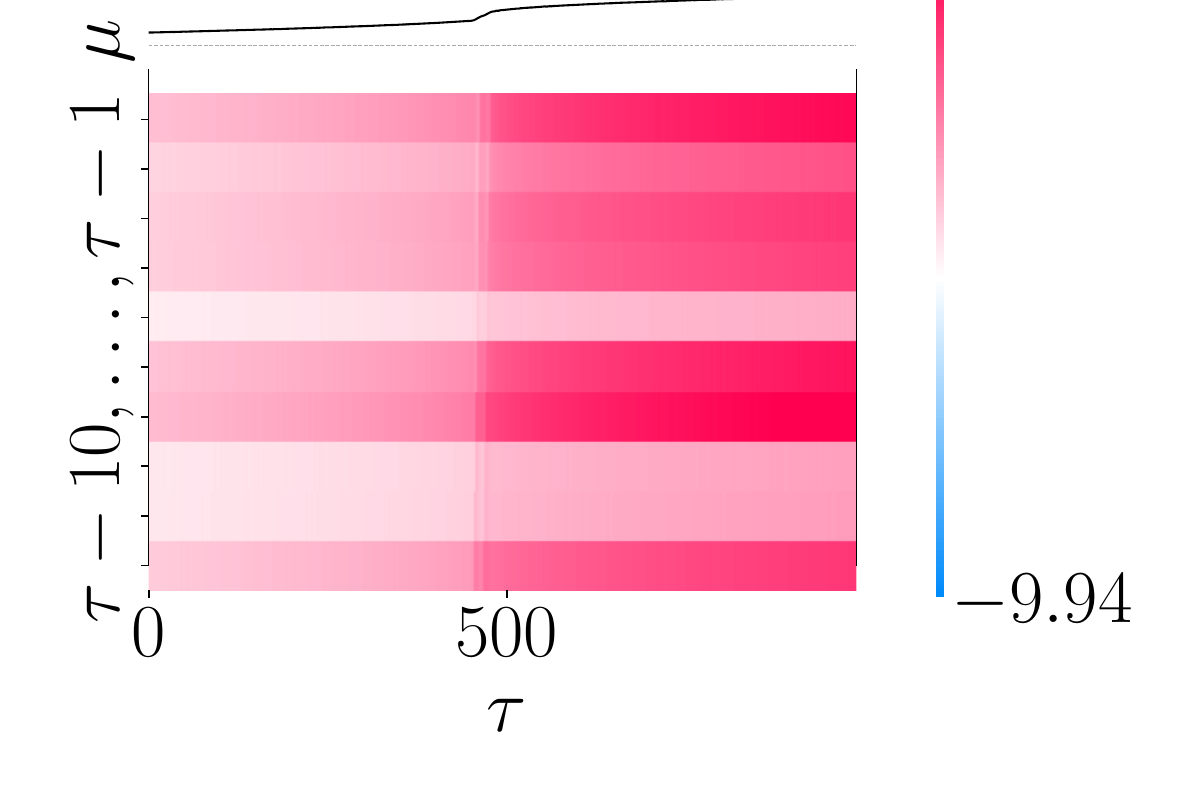}
			& \includegraphics[trim={0 4cm 0 8cm},clip,width=1.6\linewidth,valign=m]{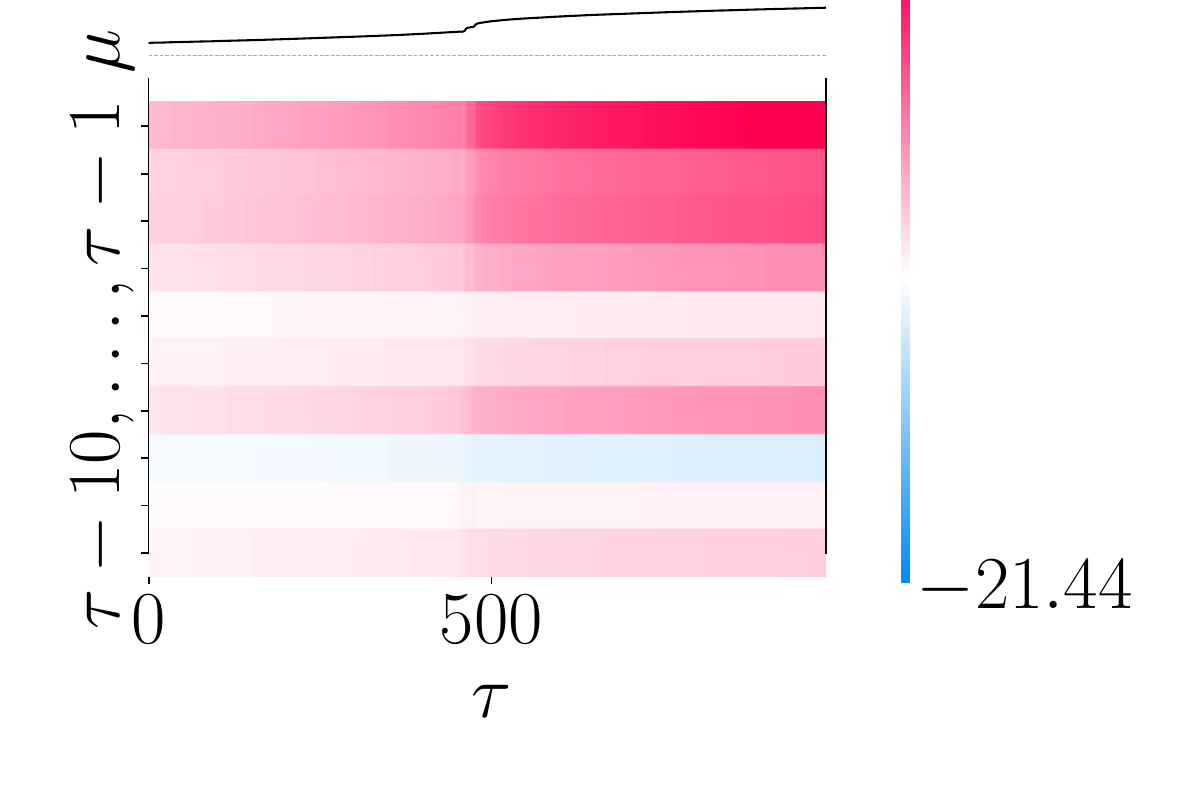}
			& \includegraphics[trim={0 4cm 0 8cm},clip,width=1.6\linewidth,valign=m]{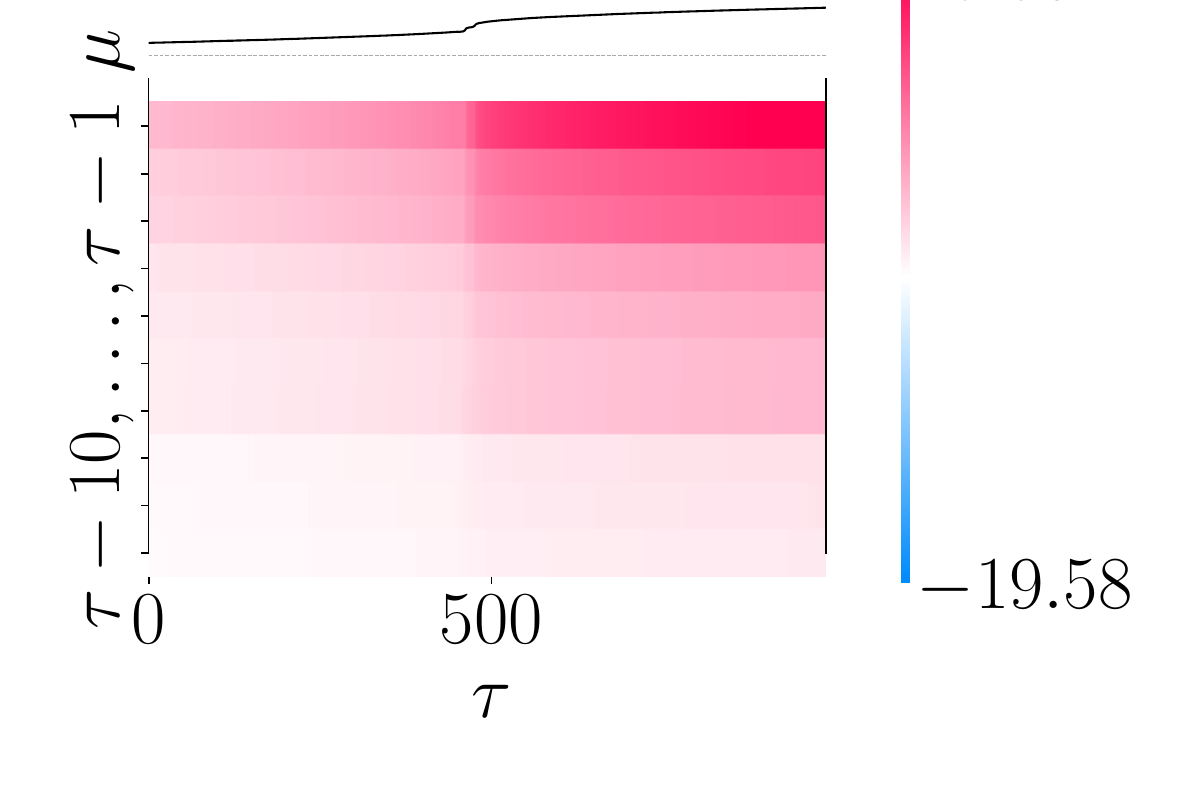} \\

			\bottomrule
		\end{tabular}
	\end{adjustbox}%
	\caption{SHAP attribution maps for the considered architectures using physics-informed (PI\@; top row) and
	autoregressive (AR\@; bottom row) features under {\Ffour}.}%
	\label{tab:F4-shap}%
\end{figure}

\begin{figure}[H]
	\centering
	\begin{adjustbox}{width=\columnwidth}
		\begin{tabular}{ccc}
			\toprule
			{\Huge BNN} & {\Huge MLP} & {\Huge DE} \\
			\midrule

			\includegraphics[width=\linewidth,valign=m]{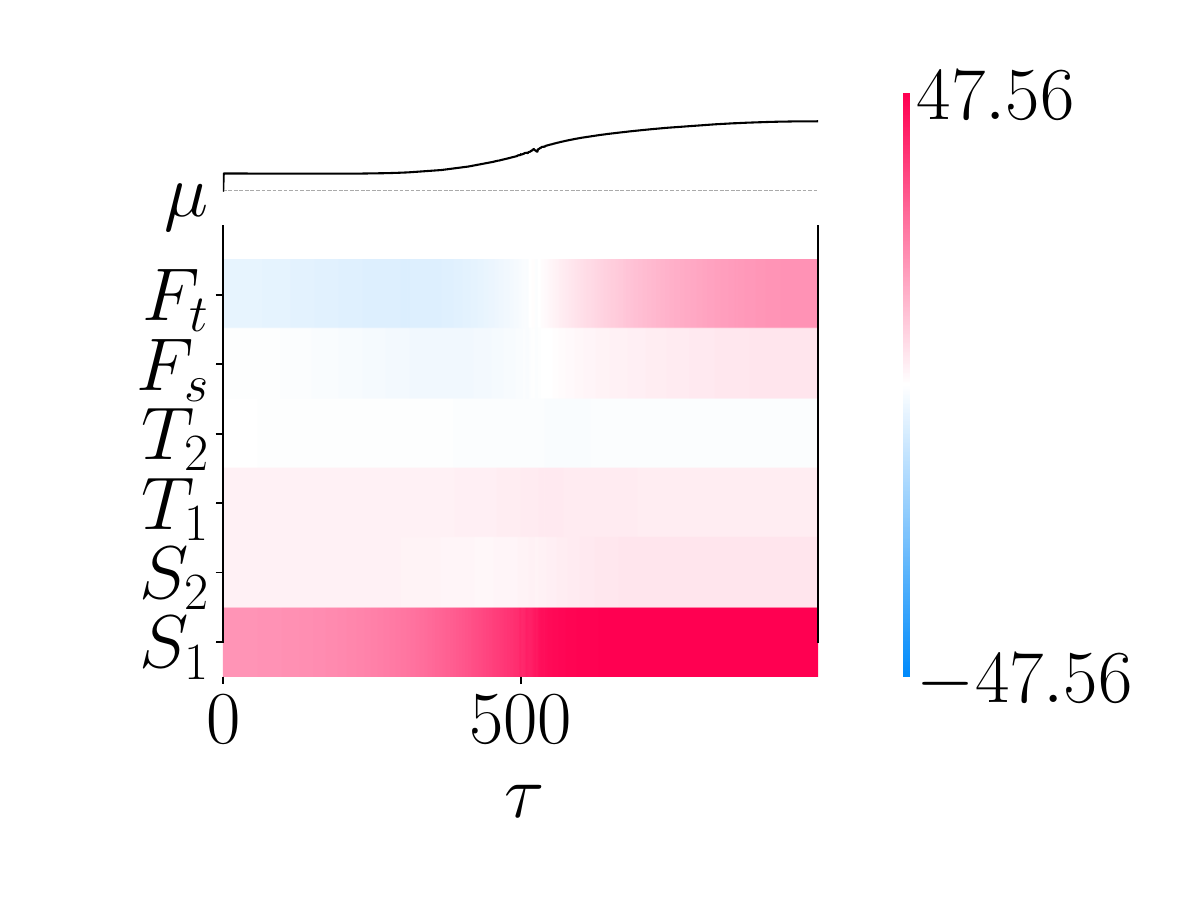}
			& \includegraphics[width=\linewidth,valign=m]{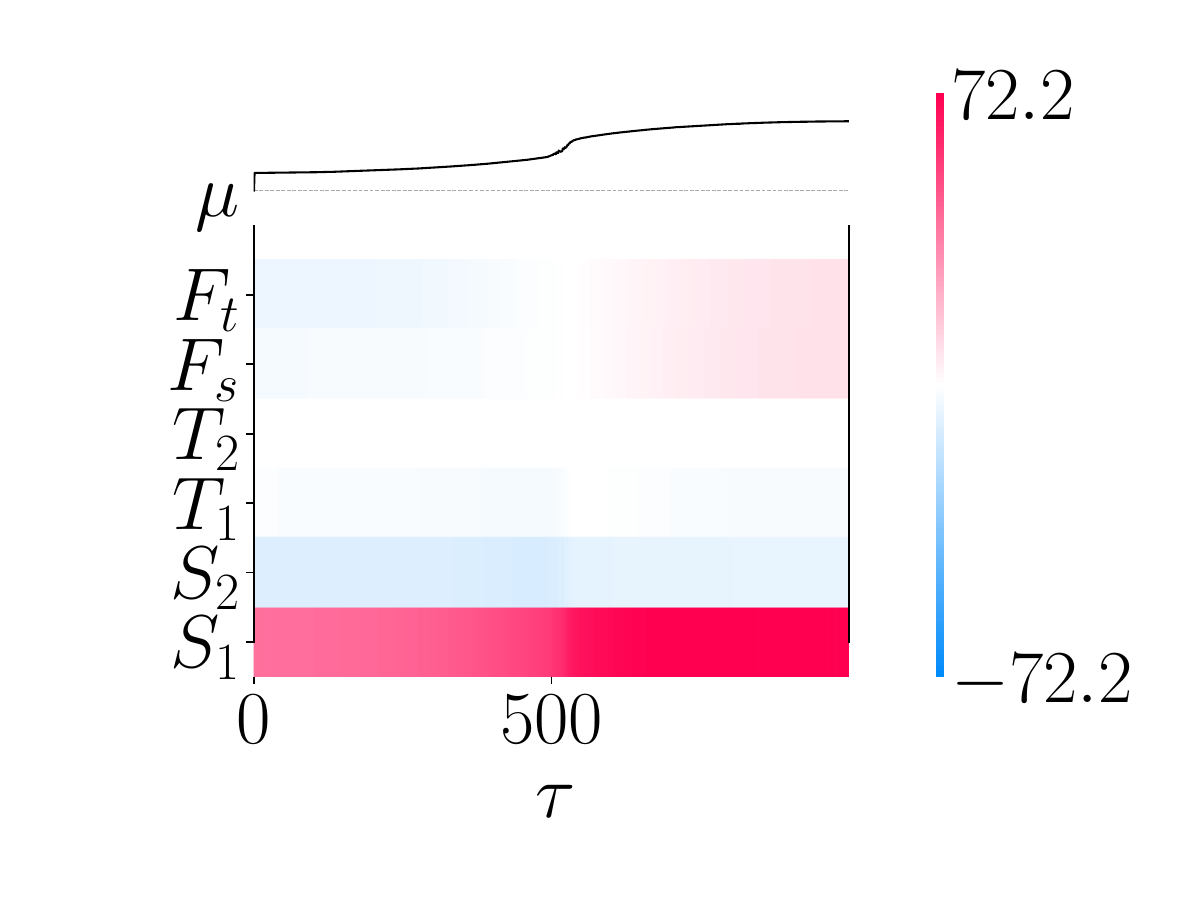}
			& \includegraphics[width=\linewidth,valign=m]{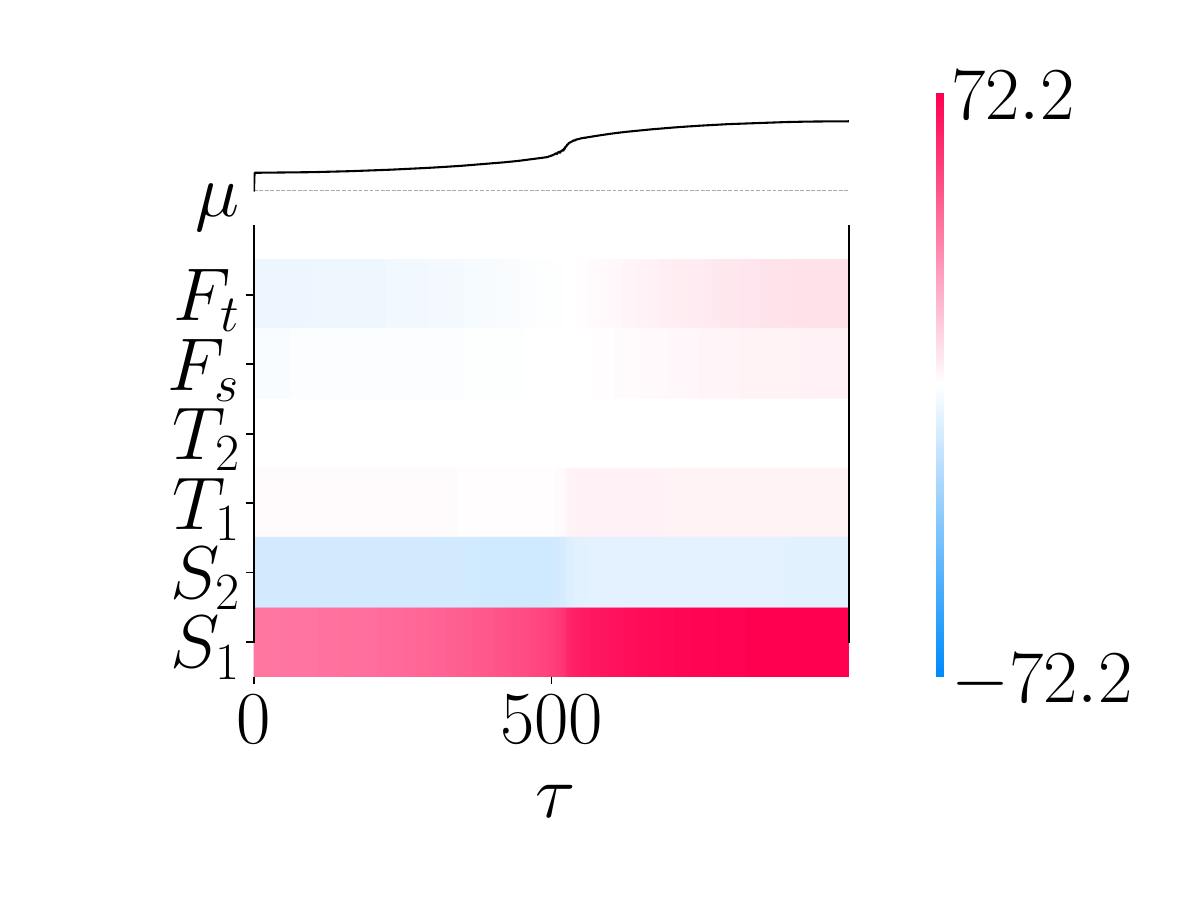} \\

			\includegraphics[trim={0 4cm 0 6cm},clip,width=\linewidth,valign=m]{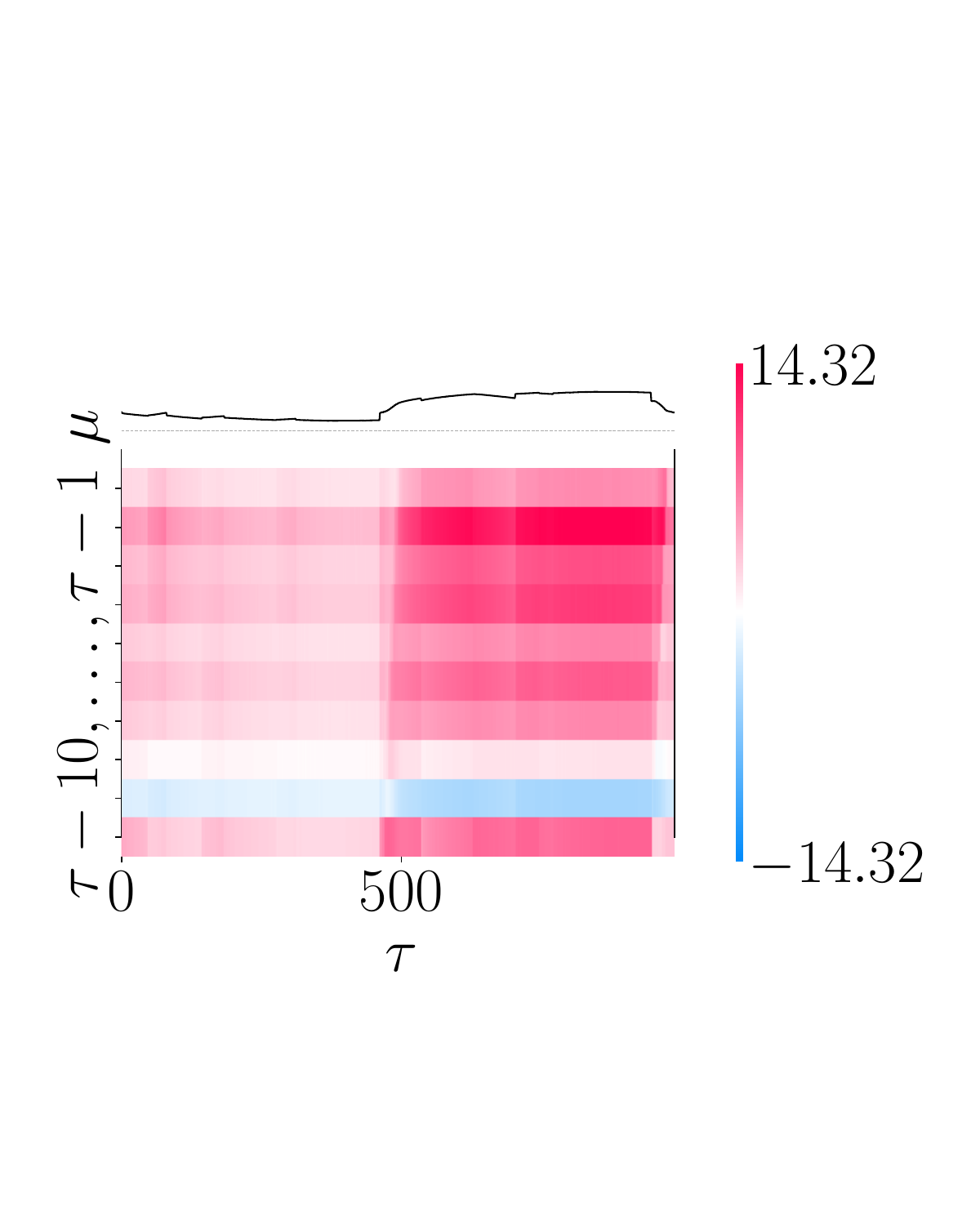}
			& \includegraphics[trim={0 4cm 0 6cm},clip,width=\linewidth,valign=m]{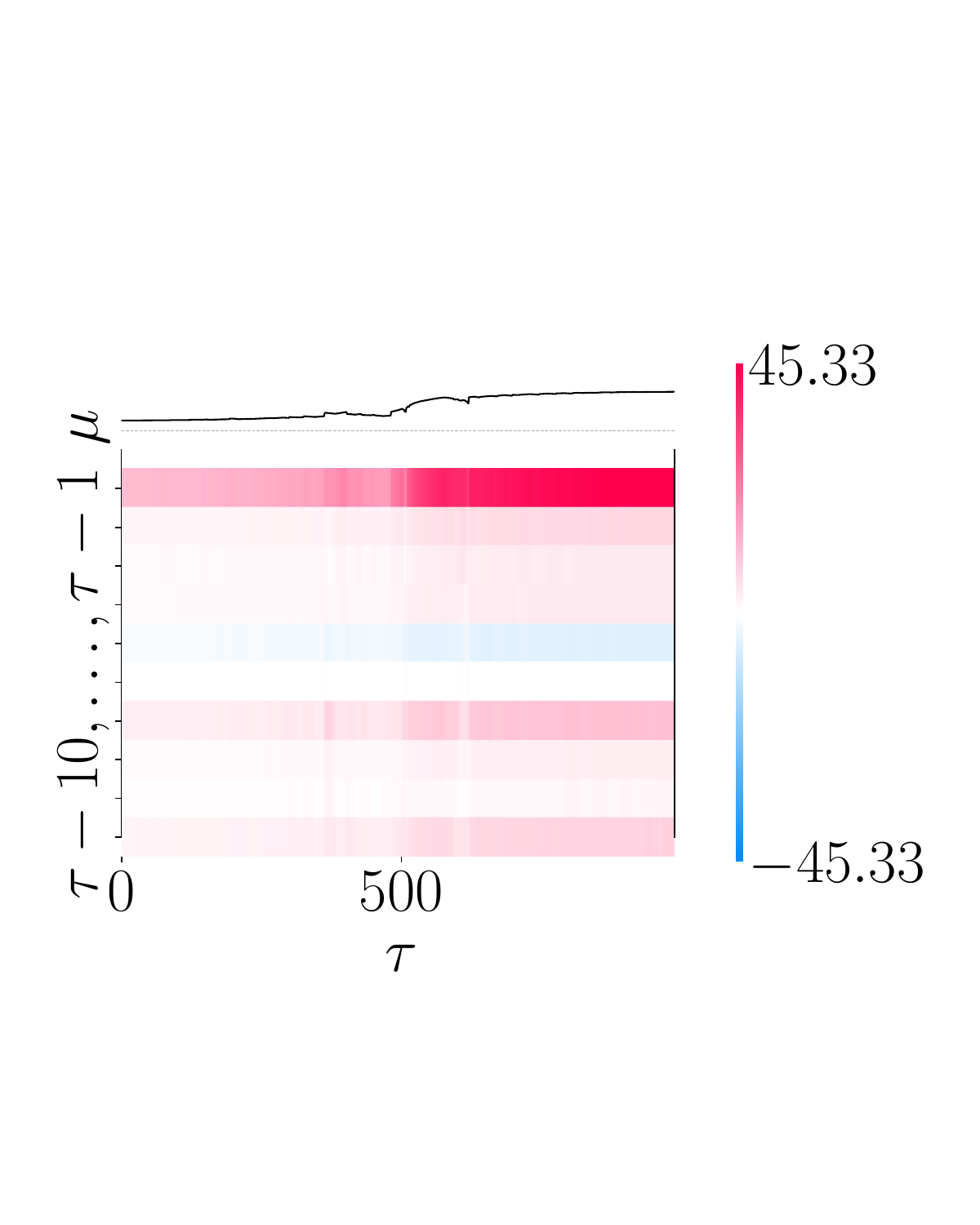}
			& \includegraphics[trim={0 4cm 0 6cm},clip,width=\linewidth,valign=m]{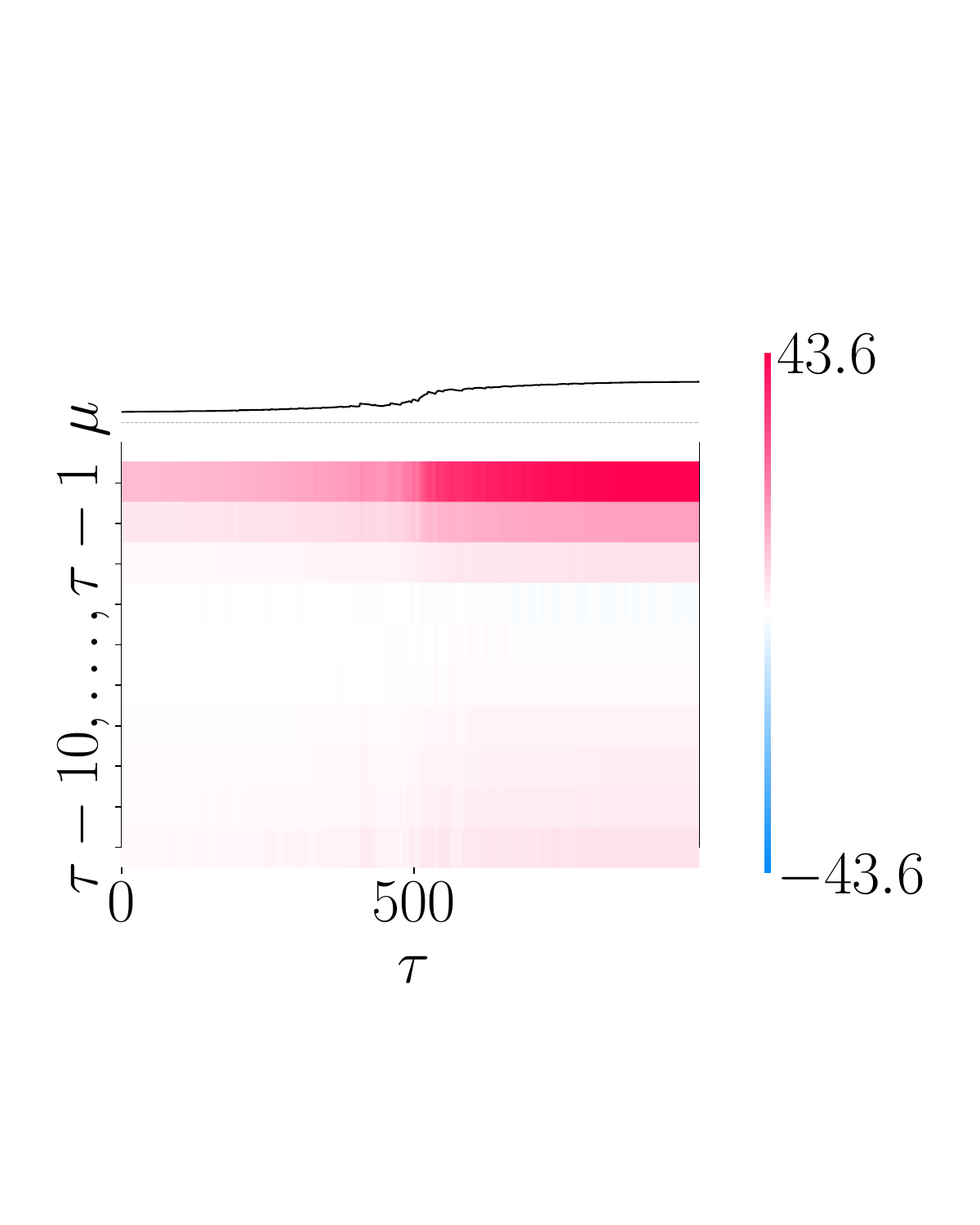} \\

			\bottomrule
		\end{tabular}
	\end{adjustbox}%
	\caption{SHAP attribution maps for the considered architectures using physics-informed (PI\@; top row) and
	autoregressive (AR\@; bottom row) features under {\Ffive}.}%
	\label{tab:F5-shap}%
\end{figure}

\begin{figure}[H]
	\centering
	\begin{adjustbox}{width=\columnwidth}
		\begin{tabular}{ccc}
			\toprule
			{\Huge BNN} & {\Huge MLP} & {\Huge DE} \\
			\midrule

			\includegraphics[width=\linewidth,valign=m]{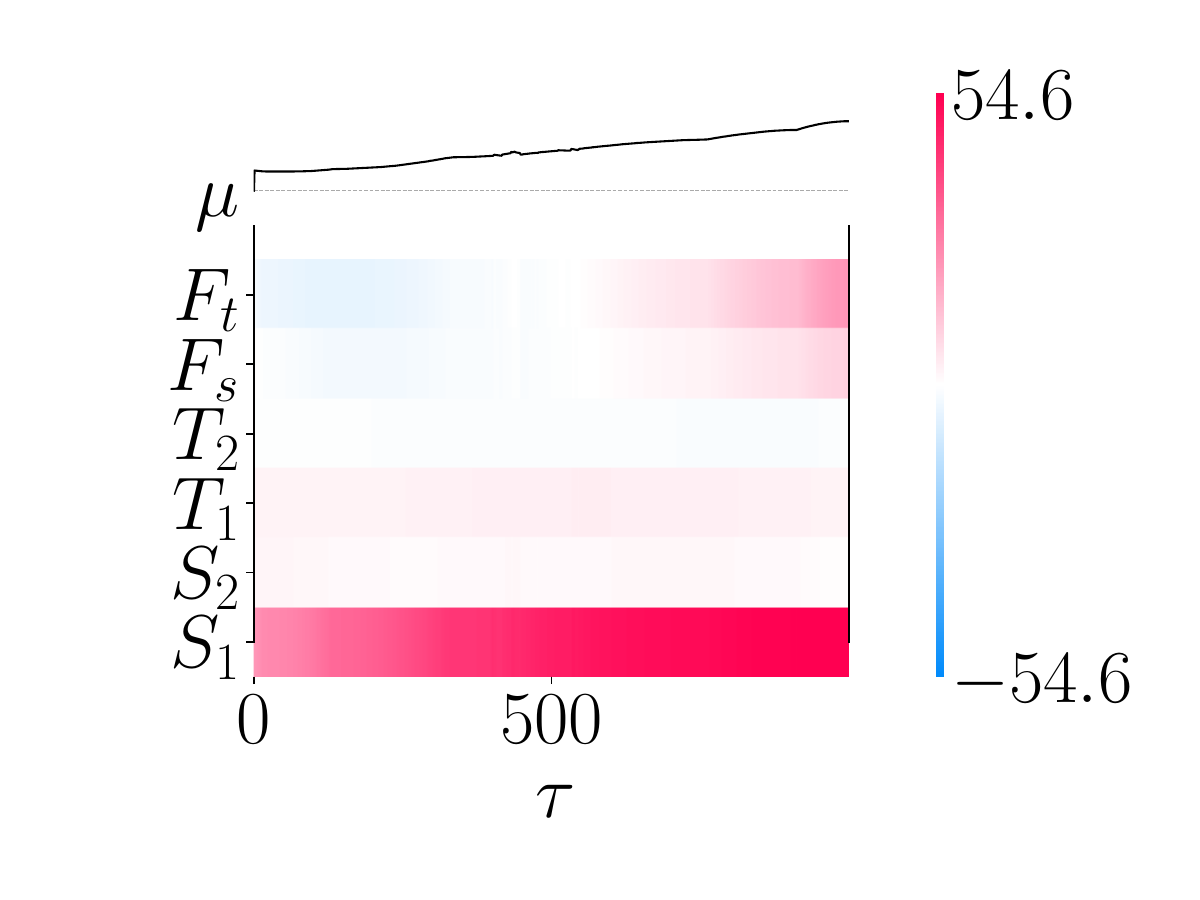}
			& \includegraphics[width=\linewidth,valign=m]{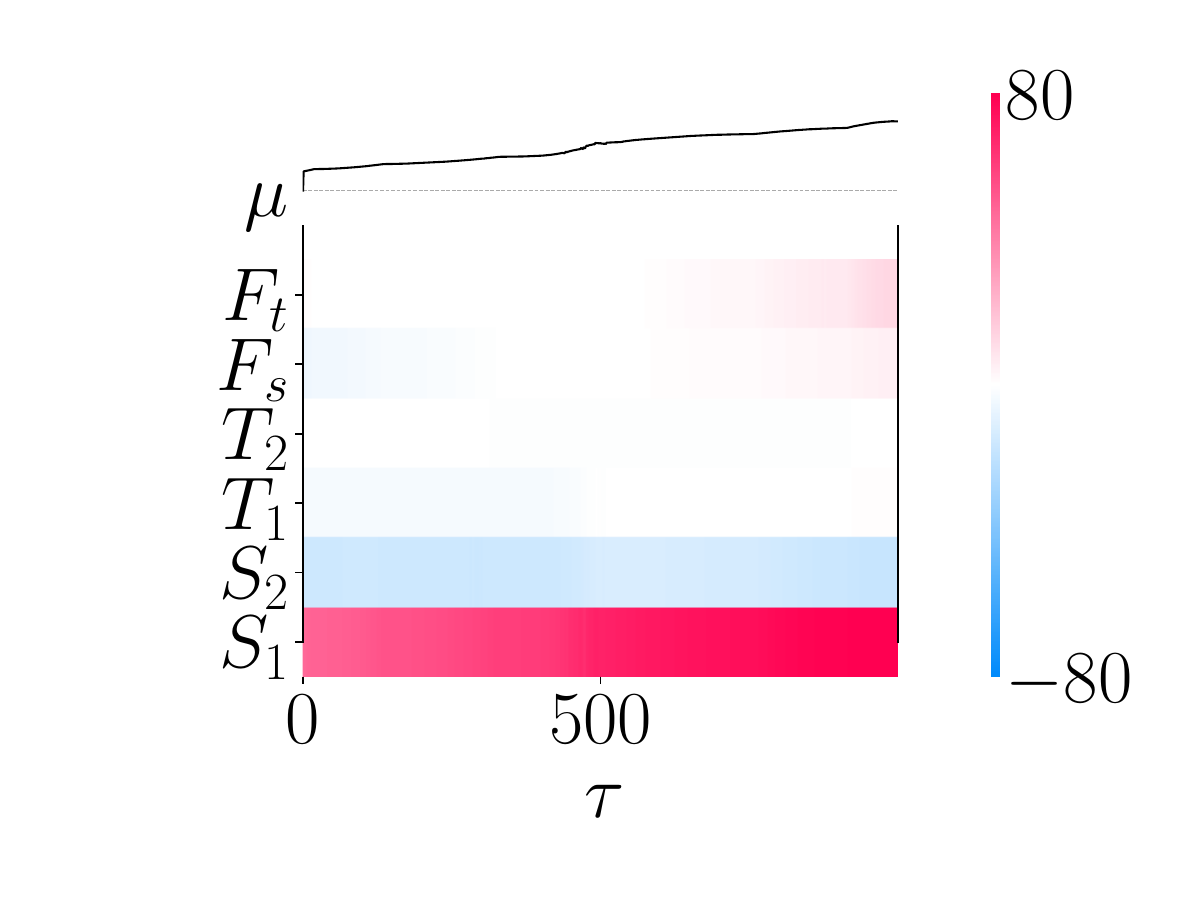}
			& \includegraphics[width=\linewidth,valign=m]{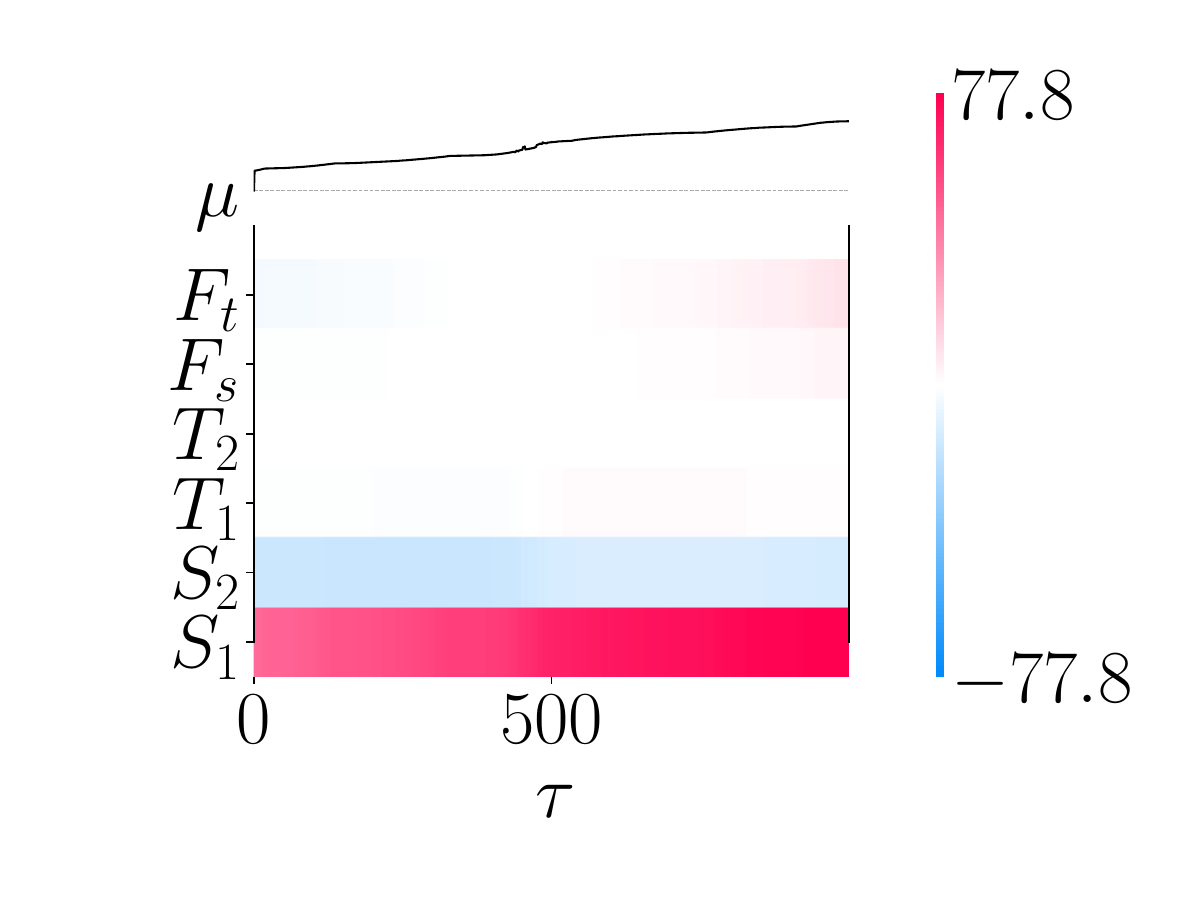} \\

			\includegraphics[trim={0 4cm 0 6cm},clip,width=\linewidth,valign=m]{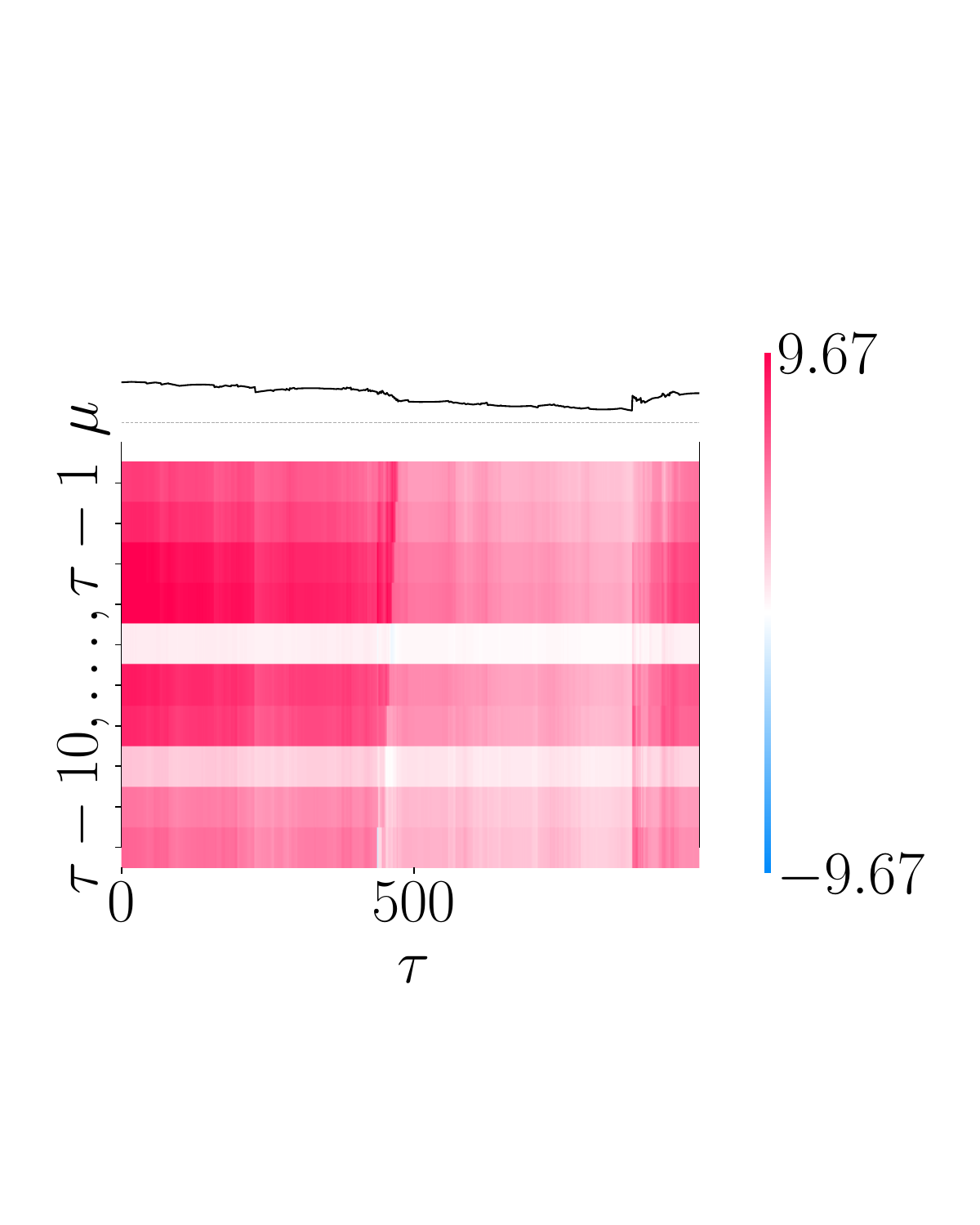}
			& \includegraphics[trim={0 4cm 0 6cm},clip,width=\linewidth,valign=m]{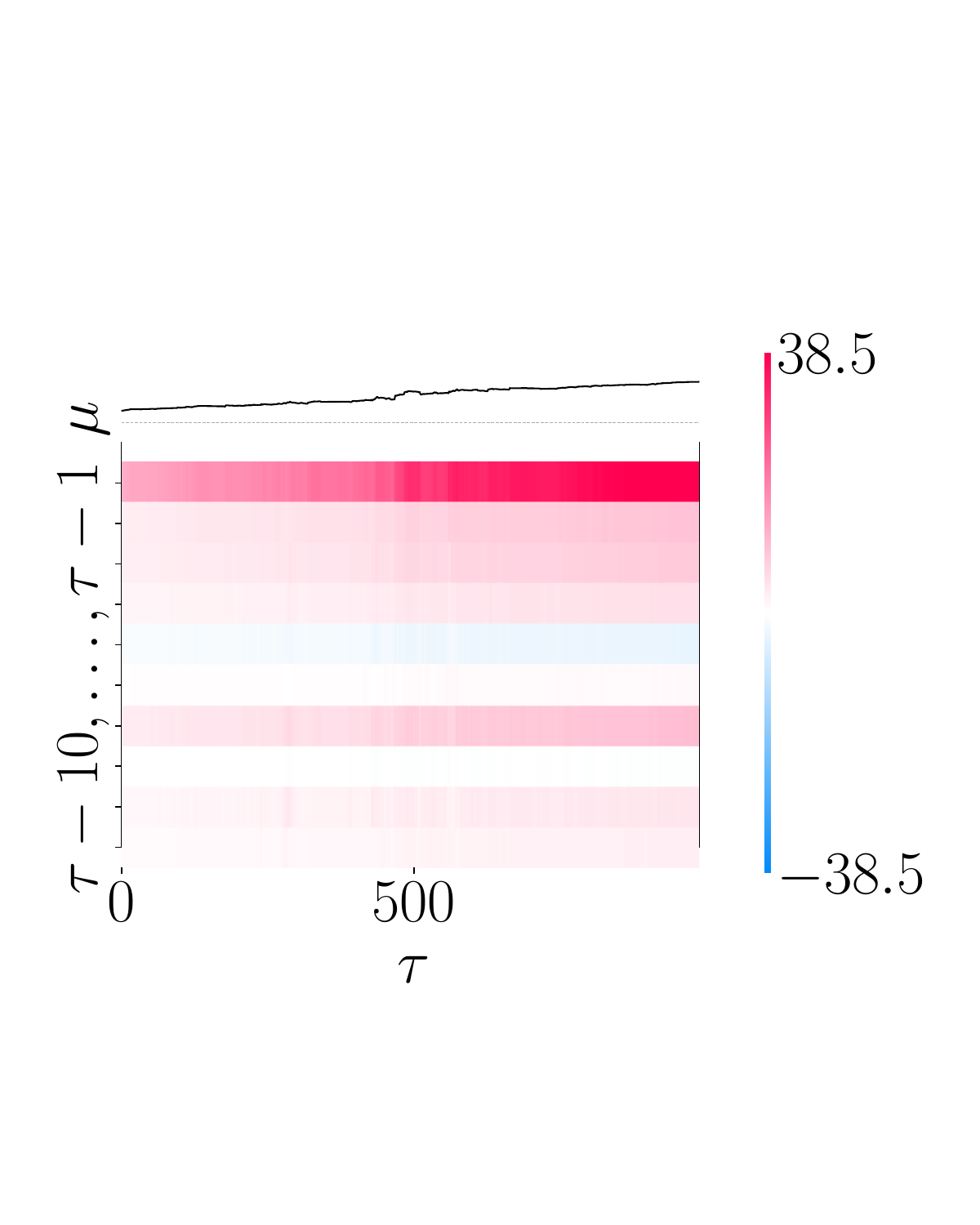}
			& \includegraphics[trim={0 4cm 0 6cm},clip,width=\linewidth,valign=m]{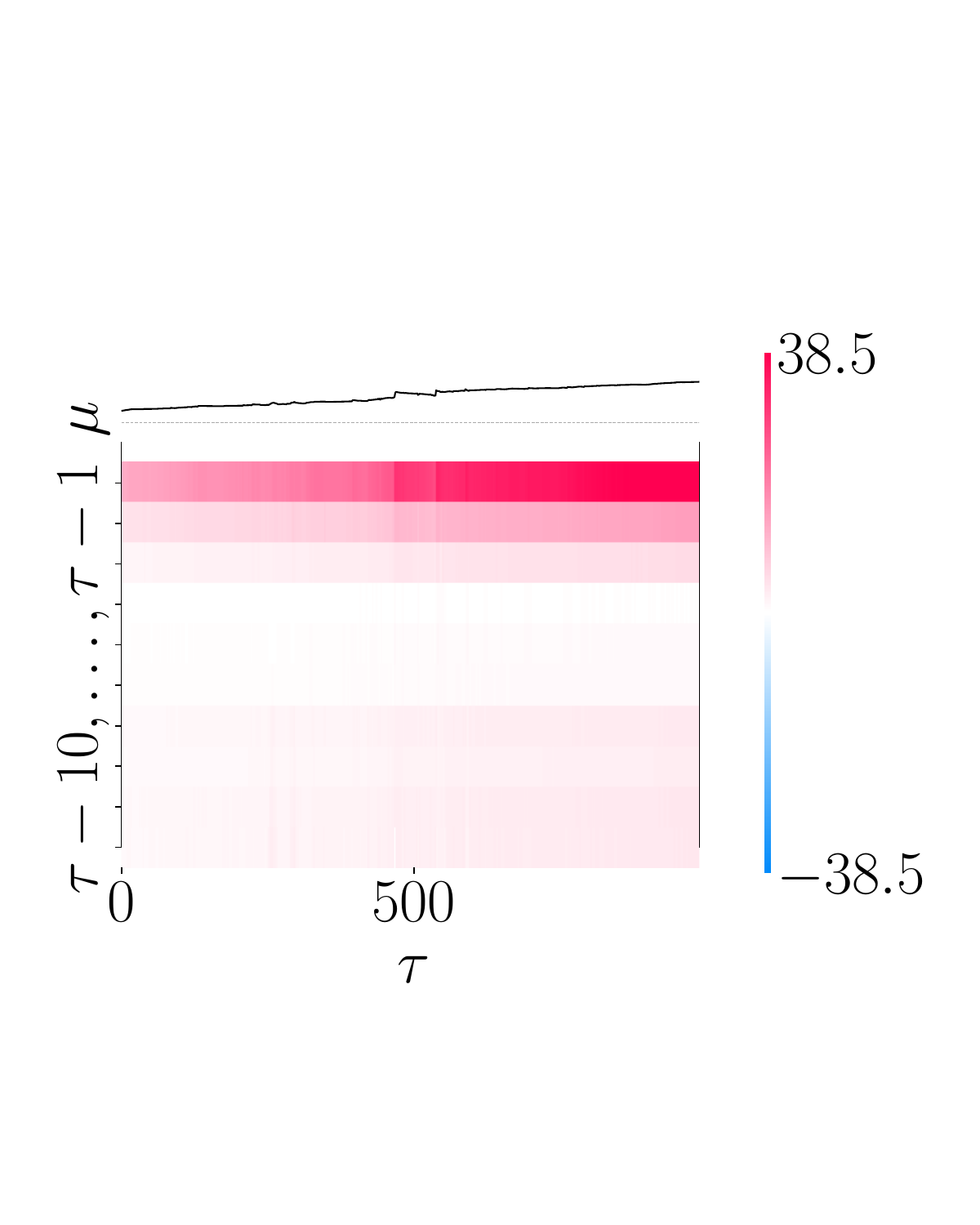} \\

			\bottomrule
		\end{tabular}
	\end{adjustbox}%
	\caption{SHAP attribution maps for the considered architectures using physics-informed (PI\@; top row) and
	autoregressive (AR\@; bottom row) features under {\Fsix}.}%
	\label{tab:F6-shap}%
\end{figure}

\section{\(\rho^{\text{lin}}\)}

In this Appendix, we include supplementary forcing scenarios using \(\rho^{\text{lin}}\) that
complement \(\{\mathcal{F}_1, \dots, \mathcal{F}_6\}\) from the main experiments. We include a
Recurrent Neural Network (RNN) as an additional architecture.

\subsection{Extended Box Model}

\subsubsection{Performance}

\begin{longtable}[c]{p{.1\columnwidth}p{.1\columnwidth}C{.25\columnwidth}C{.25\columnwidth}}
	\toprule
	Scenario & Architecture & Prediction: PI & Prediction: AR \\
	\midrule

	\multirow[c]{3}{*}{\rotatebox[origin=c]{90}{\makecell{\(F_s\): Lin. \\ \(F_t\): Lin.}}}
	& BNN
	& \includegraphics[width=\linewidth,valign=m]{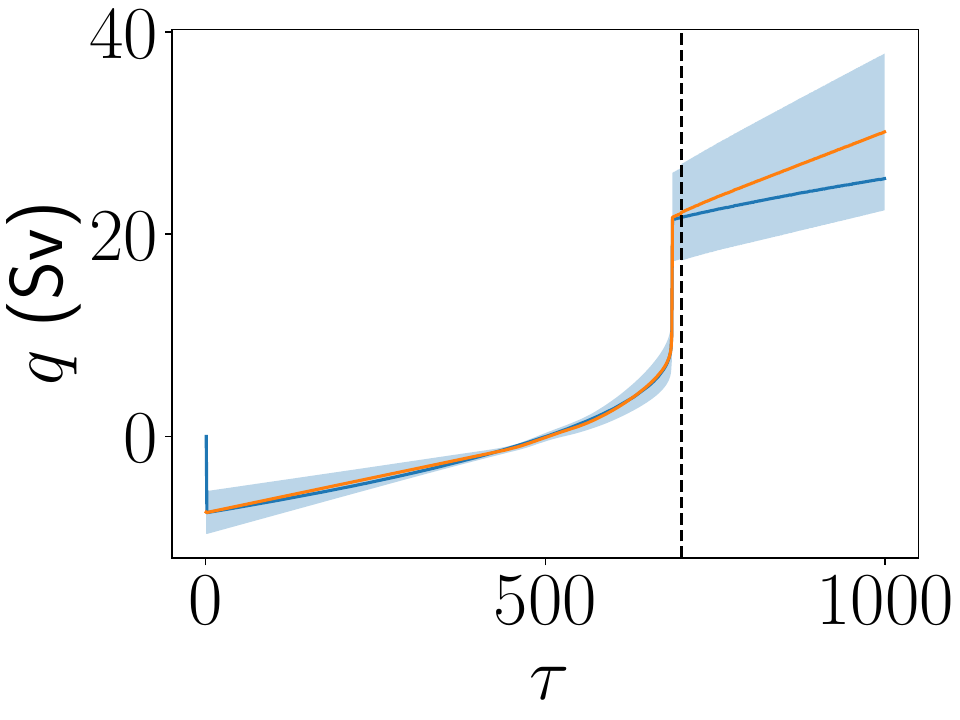}
	& \includegraphics[width=\linewidth,valign=m]{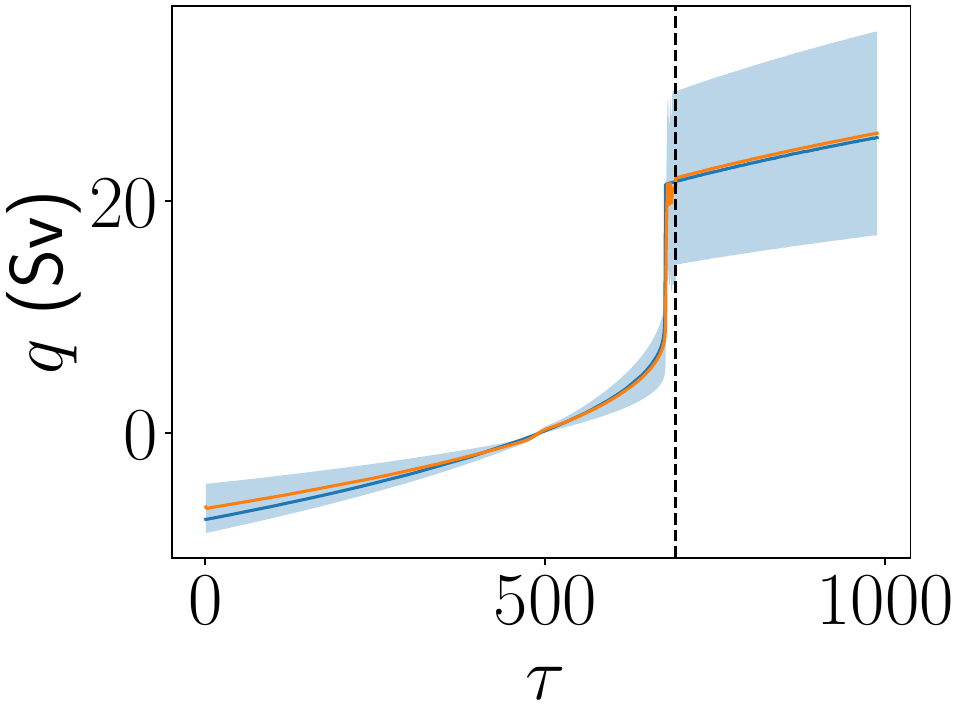} \\
	& MLP
	& \includegraphics[width=\linewidth,valign=m]{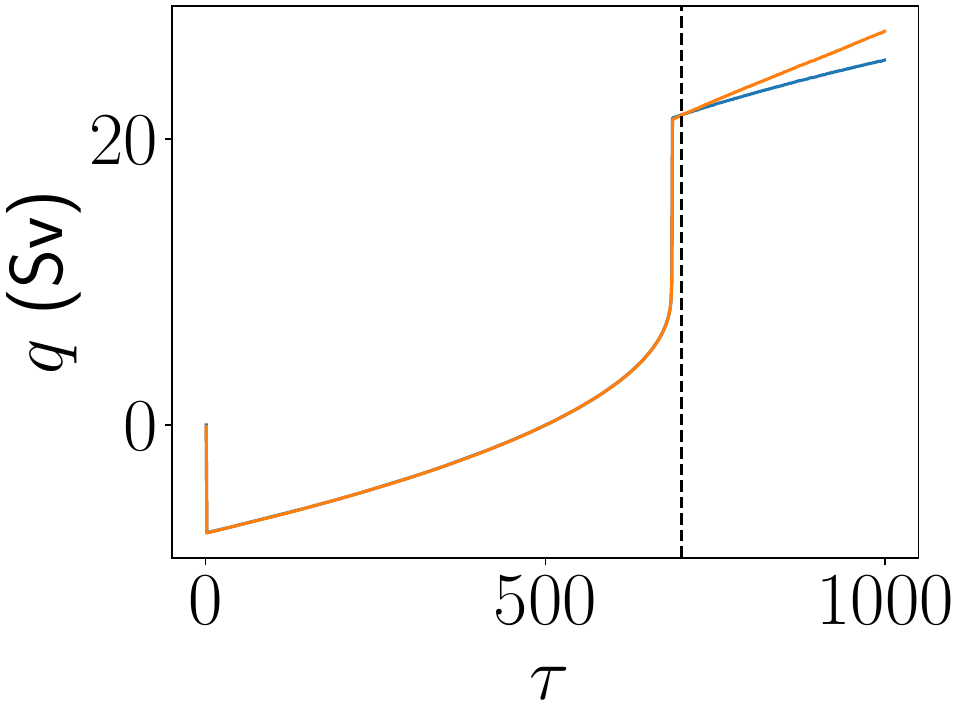}
	& \includegraphics[width=\linewidth,valign=m]{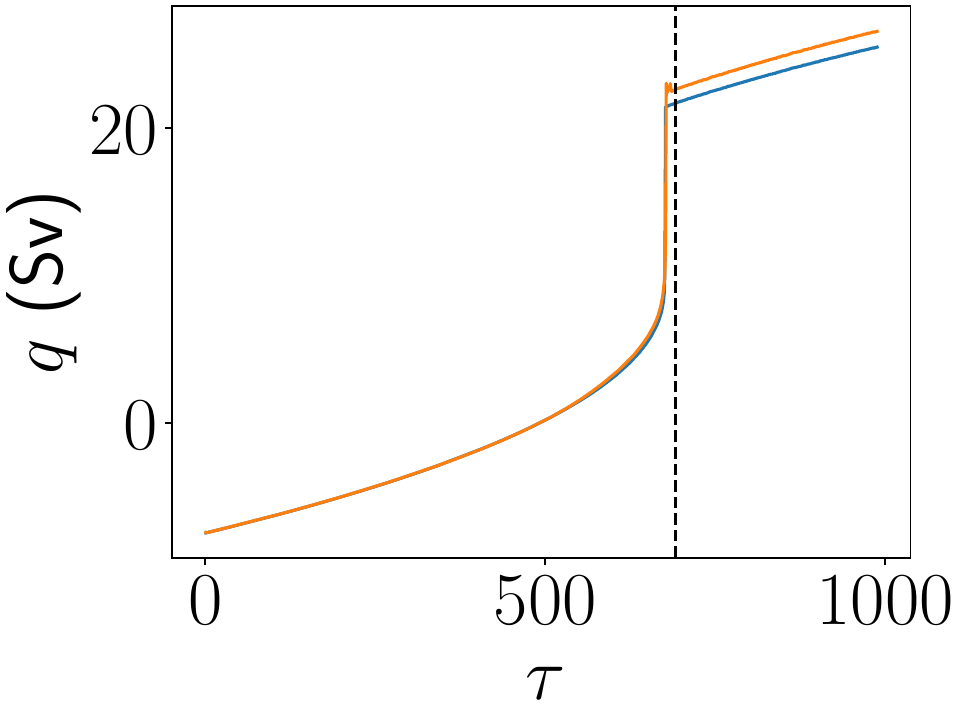} \\
	& Deep Ensemble
	& \includegraphics[width=\linewidth,valign=m]{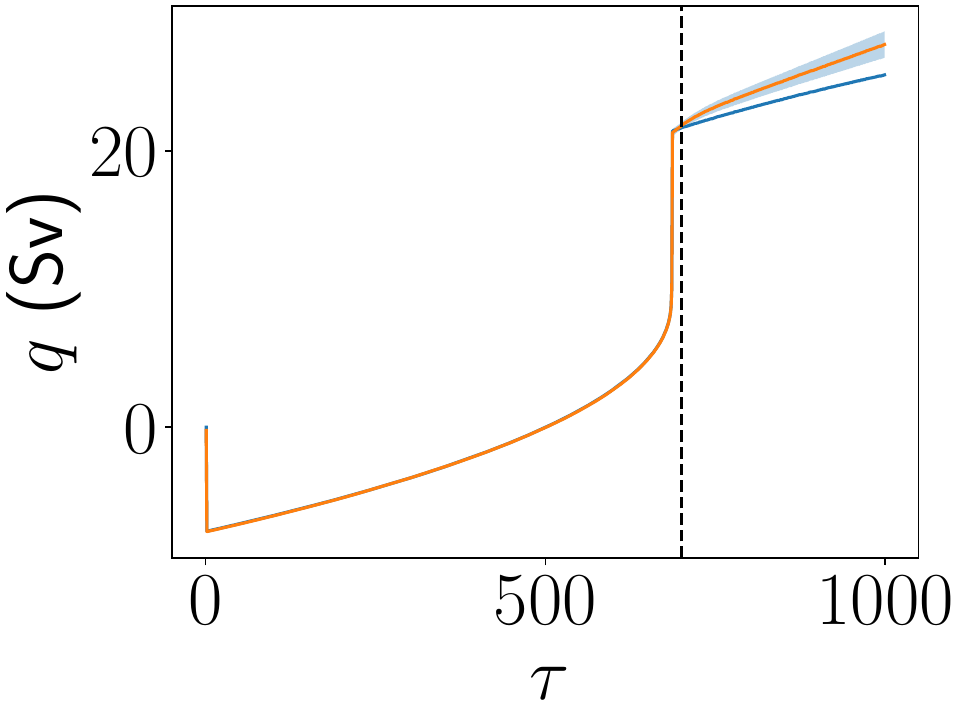}
	& \includegraphics[width=\linewidth,valign=m]{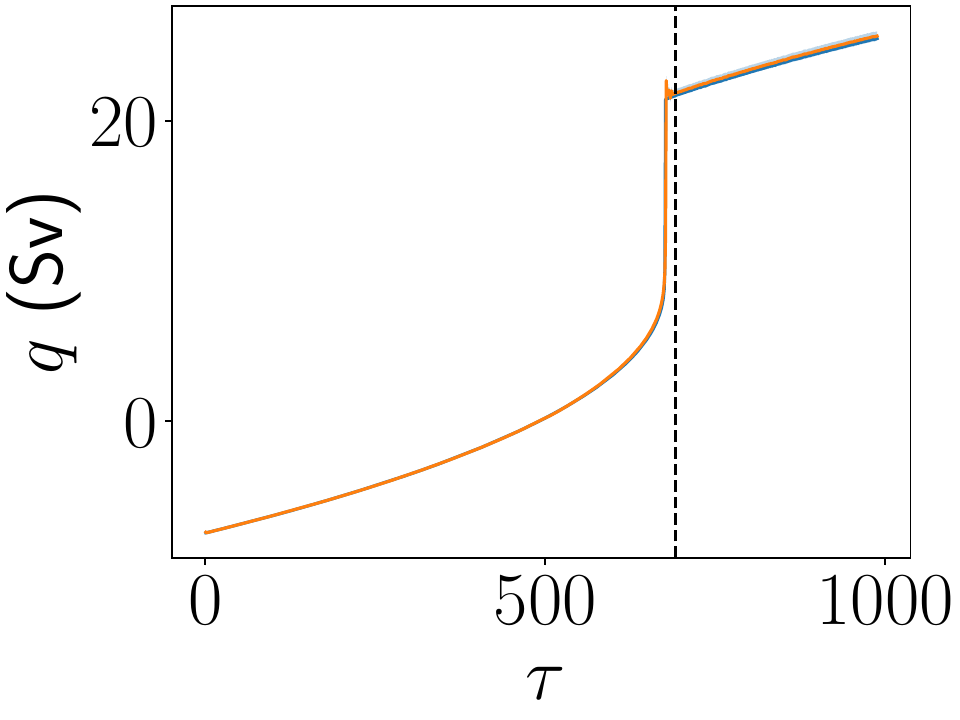} \\
	& RNN
	& ---
	& \includegraphics[width=\linewidth,valign=m]{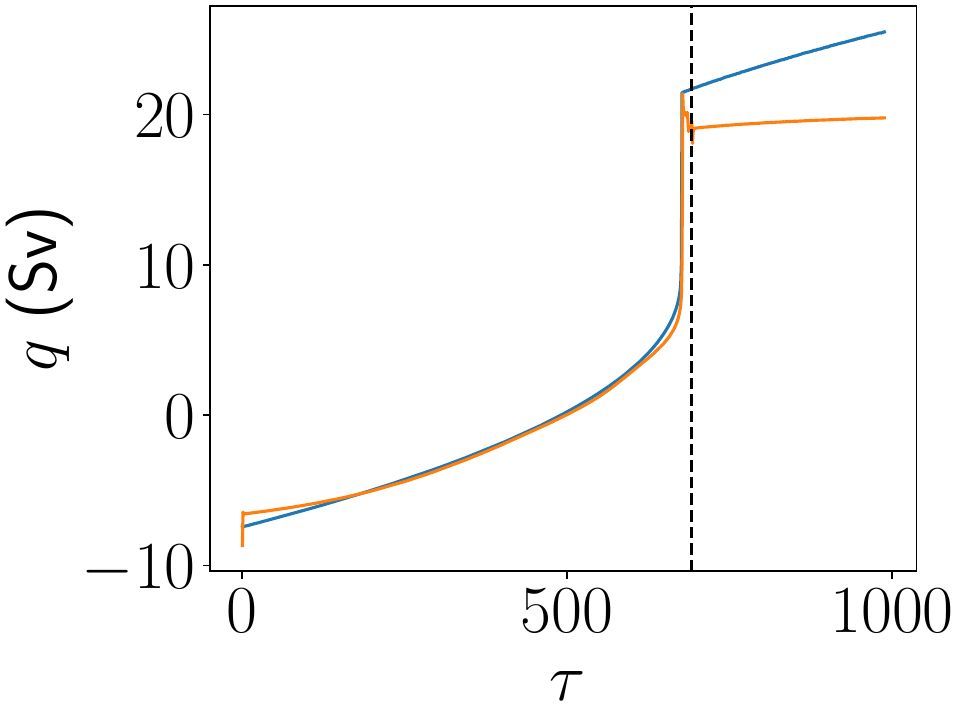} \\

	\midrule

	\multirow[c]{3}{*}{\rotatebox[origin=c]{90}{\makecell{\(F_s\): Sin.\ (stationary) \\ \(F_t\): Sin.\ (stationary)}}}
	& BNN
	& \includegraphics[width=\linewidth,valign=m]{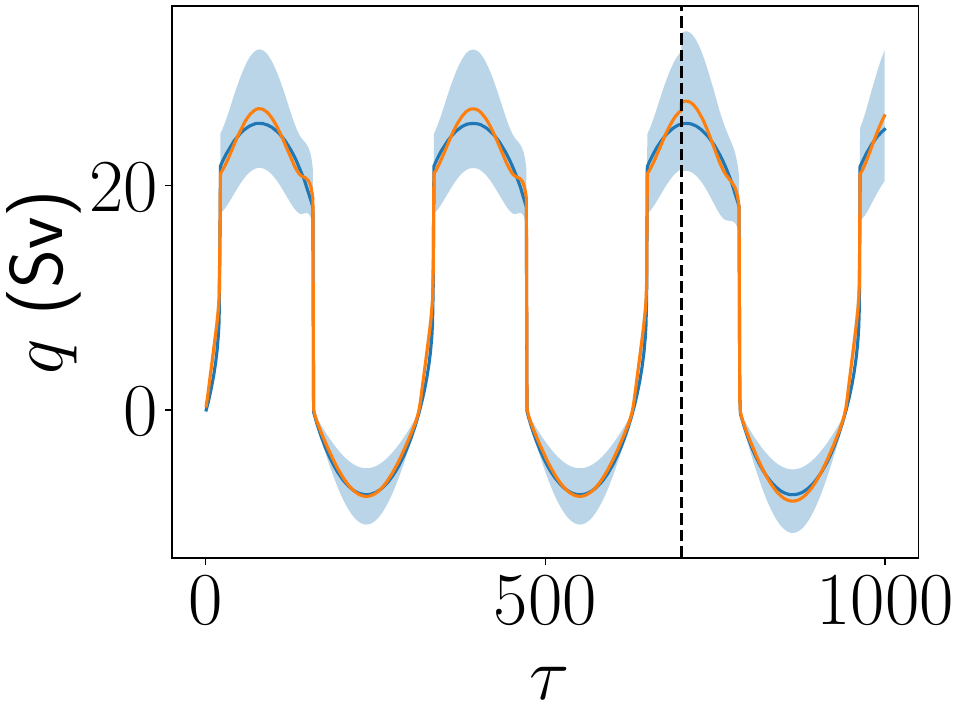}
	& \includegraphics[width=\linewidth,valign=m]{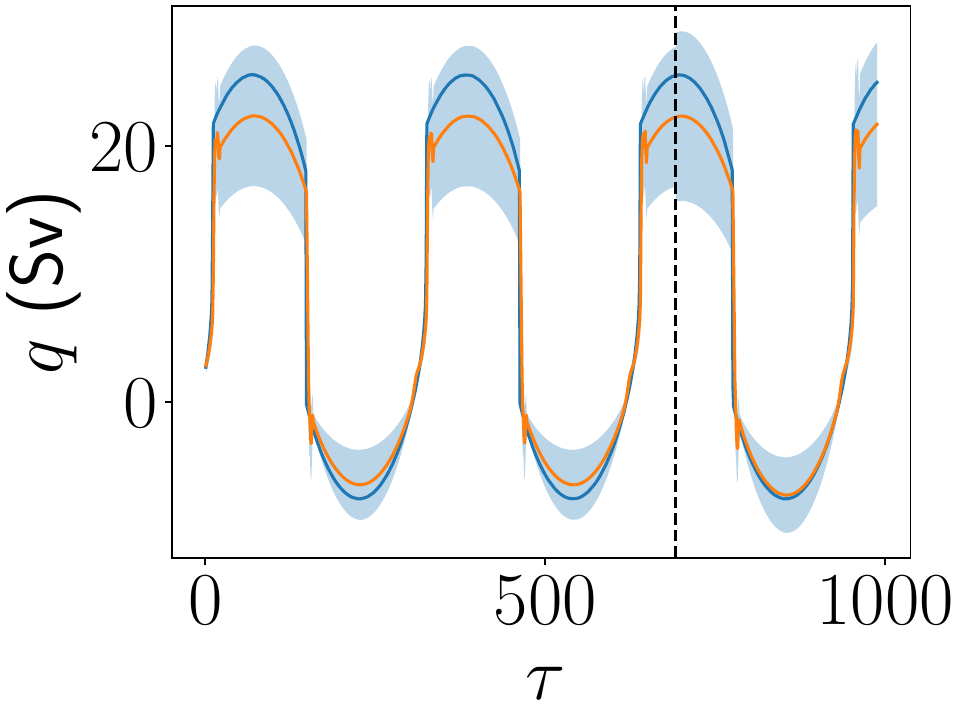} \\
	& MLP
	& \includegraphics[width=\linewidth,valign=m]{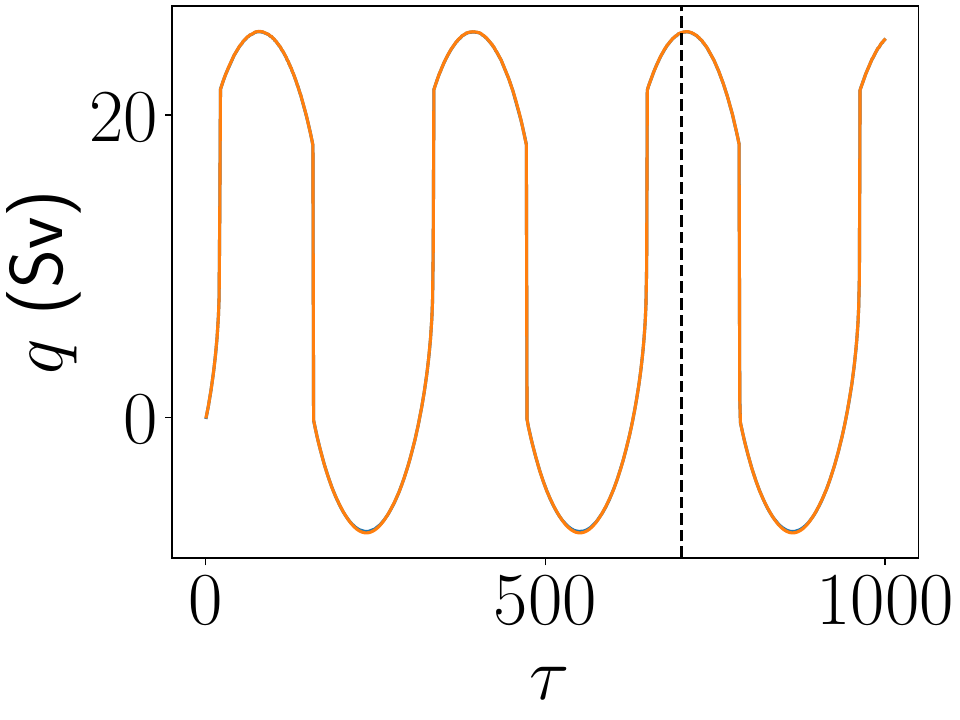}
	& \includegraphics[width=\linewidth,valign=m]{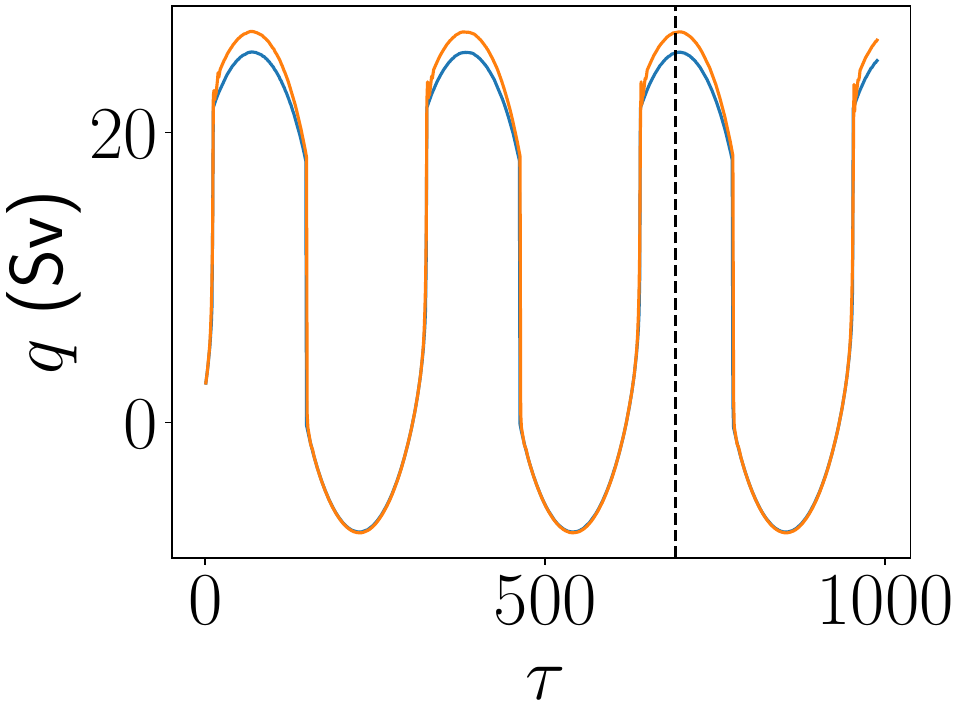} \\
	& Deep Ensemble
	& \includegraphics[width=\linewidth,valign=m]{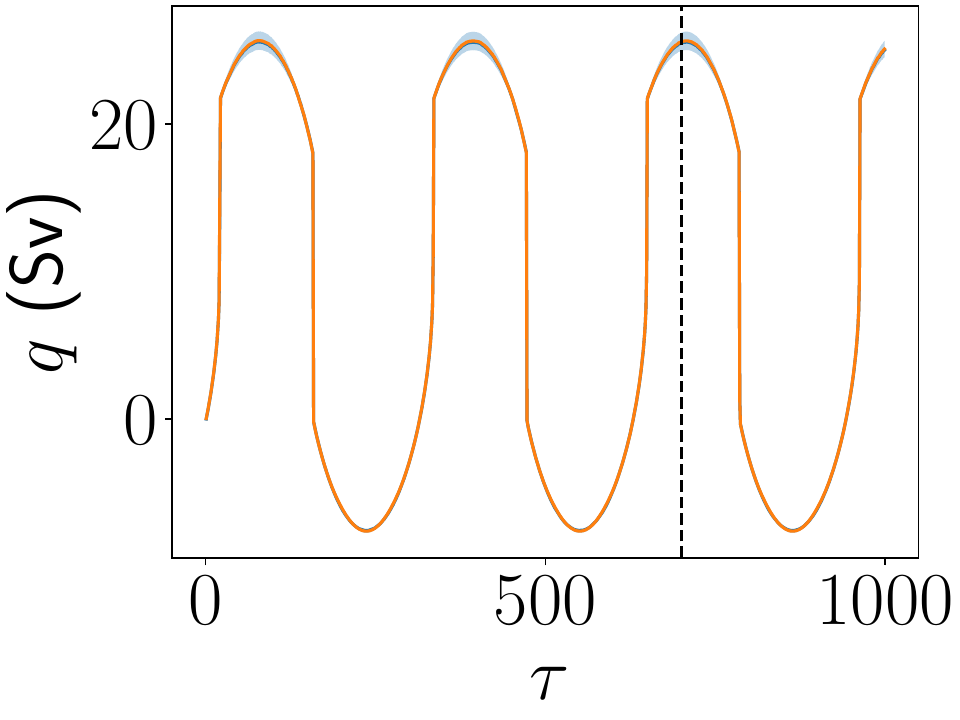}
	& \includegraphics[width=\linewidth,valign=m]{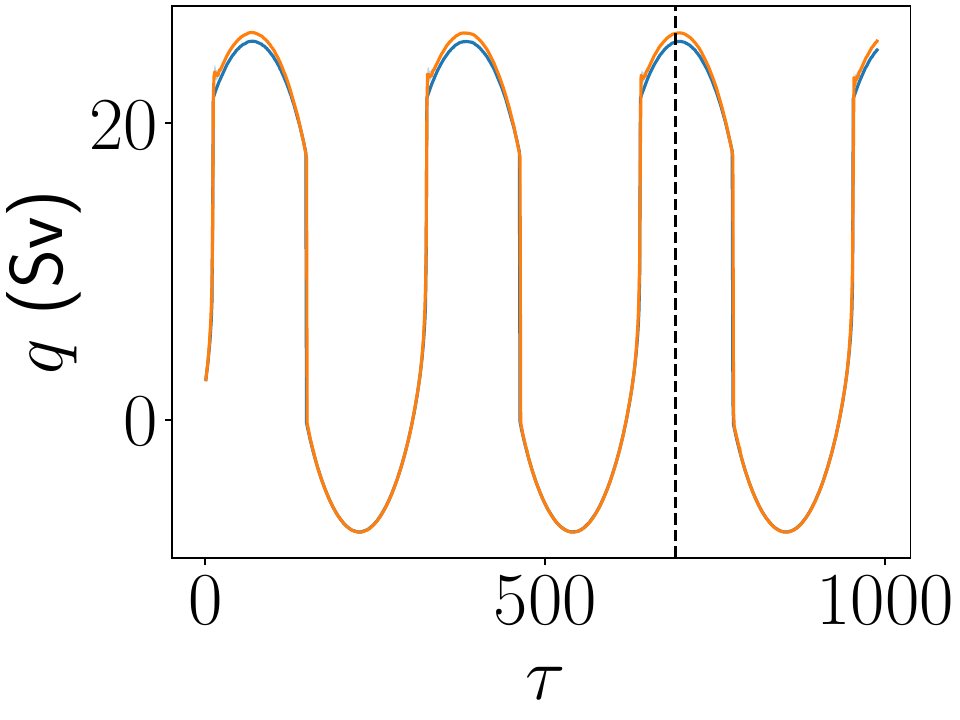} \\
	& RNN
	& ---
	& \includegraphics[width=\linewidth,valign=m]{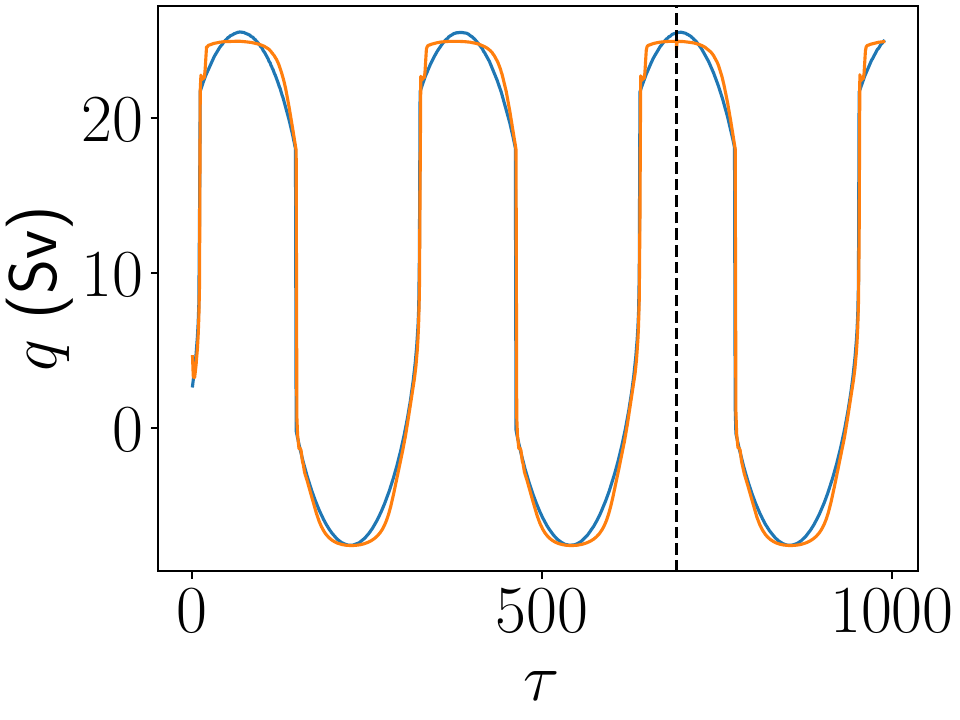} \\

	\midrule

	\multirow[c]{3}{*}{\rotatebox[origin=c]{90}{\makecell{\(F_s\): Sin.\ (nonstationary) \\ \(F_t\): Sin.\ (nonstationary)}}}
	& BNN
	& \includegraphics[width=\linewidth,valign=m]{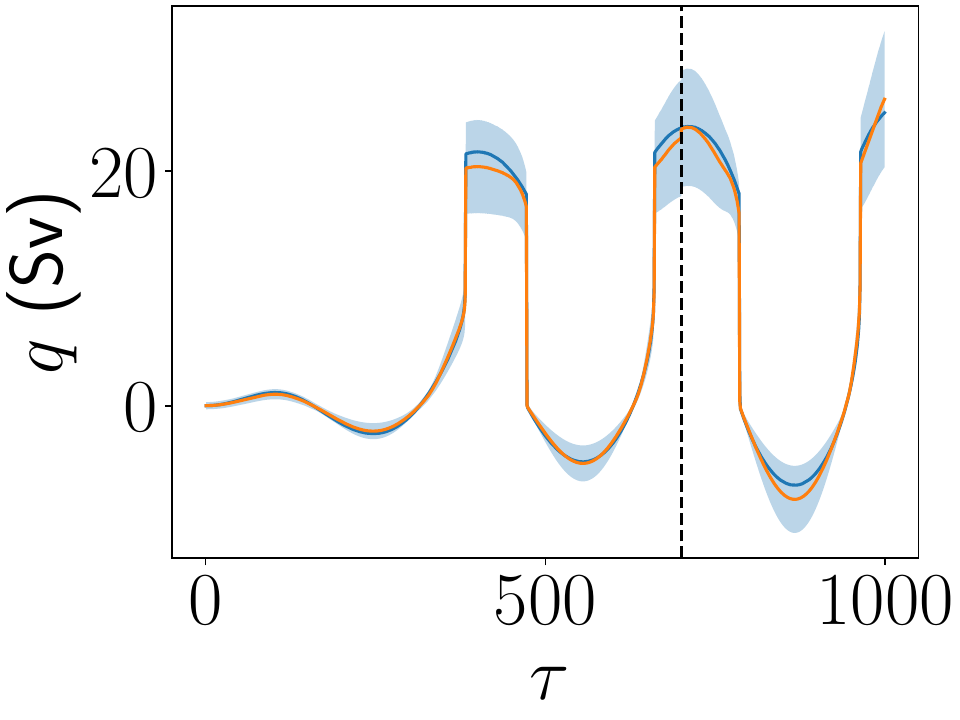}
	& \includegraphics[width=\linewidth,valign=m]{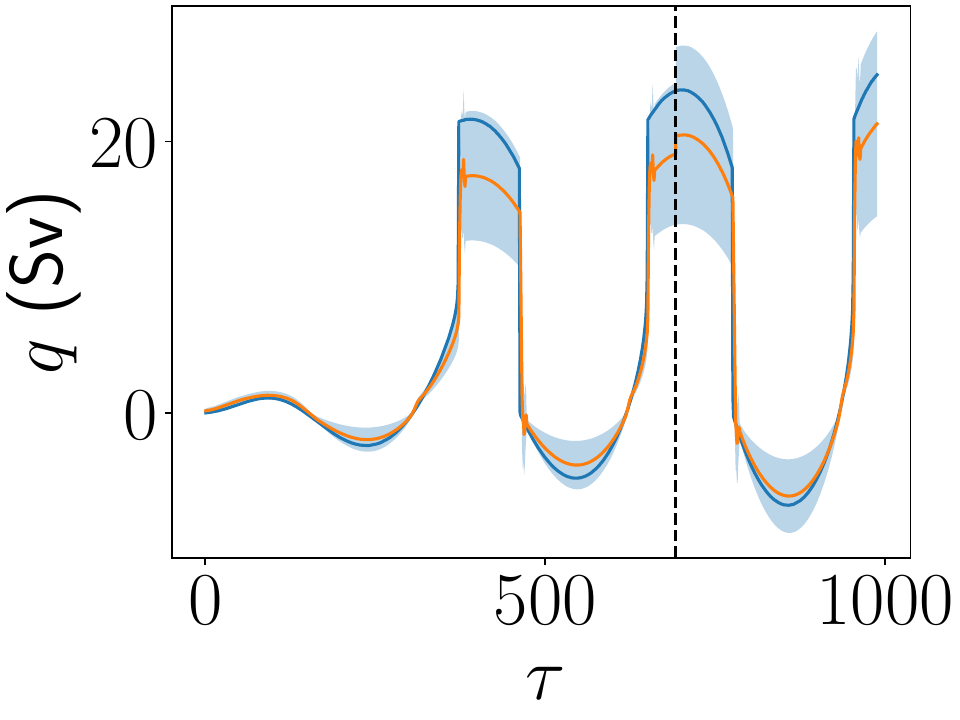} \\
	& MLP
	& \includegraphics[width=\linewidth,valign=m]{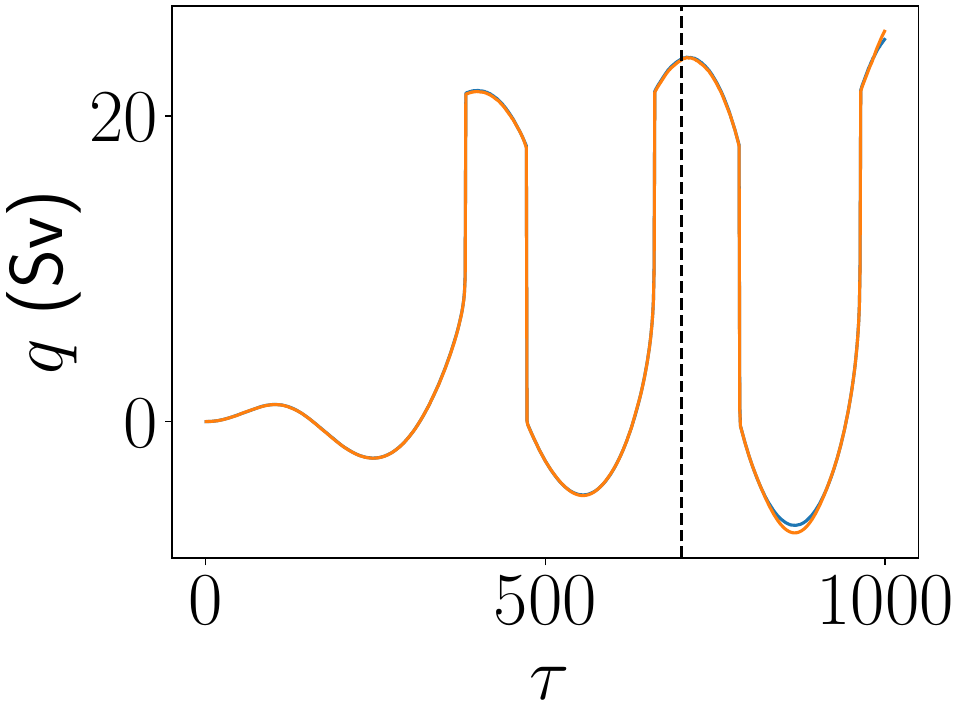}
	& \includegraphics[width=\linewidth,valign=m]{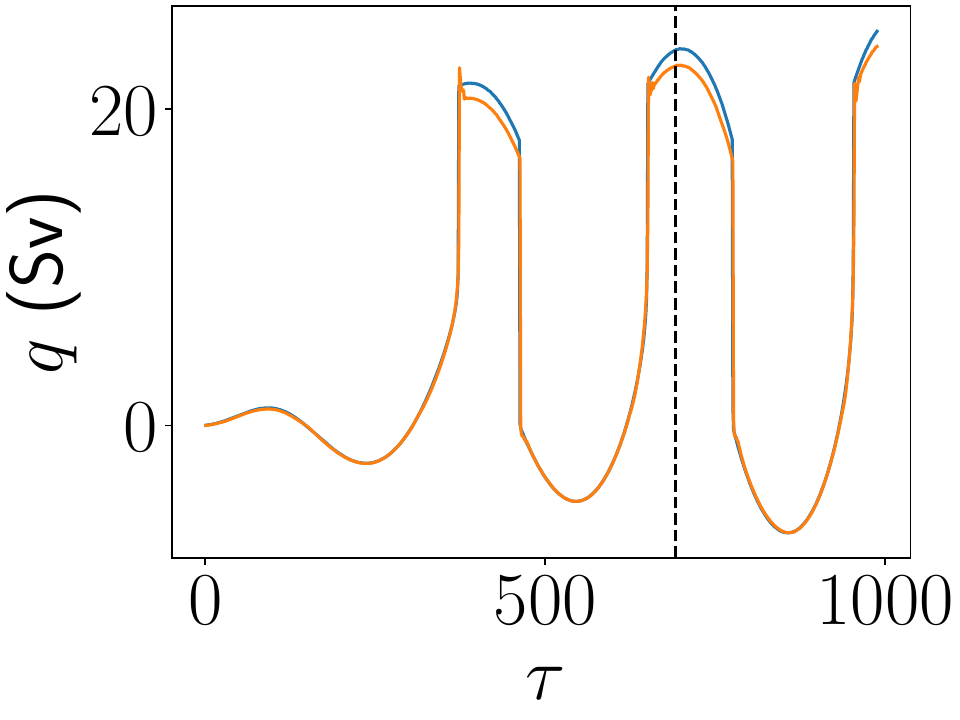} \\
	& Deep Ensemble
	& \includegraphics[width=\linewidth,valign=m]{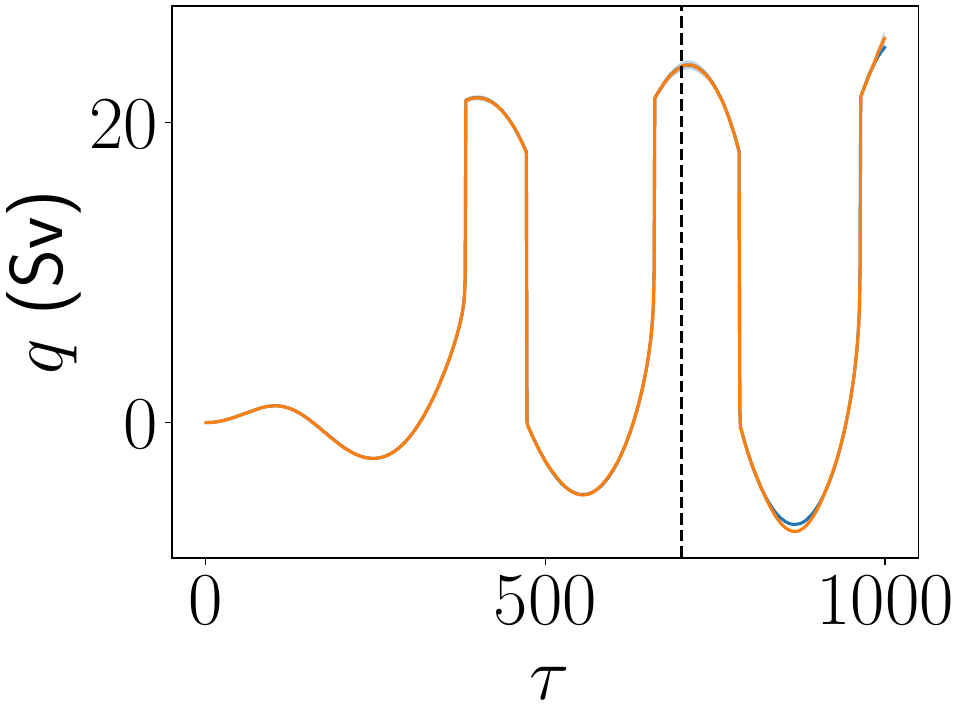}
	& \includegraphics[width=\linewidth,valign=m]{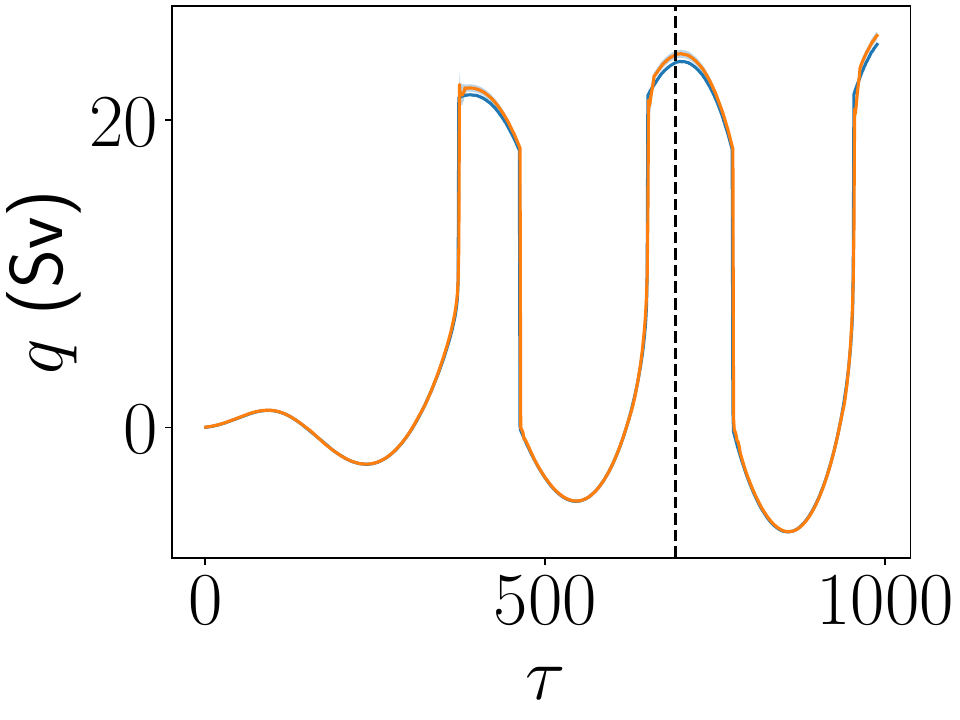} \\
	& RNN
	& ---
	& \includegraphics[width=\linewidth,valign=m]{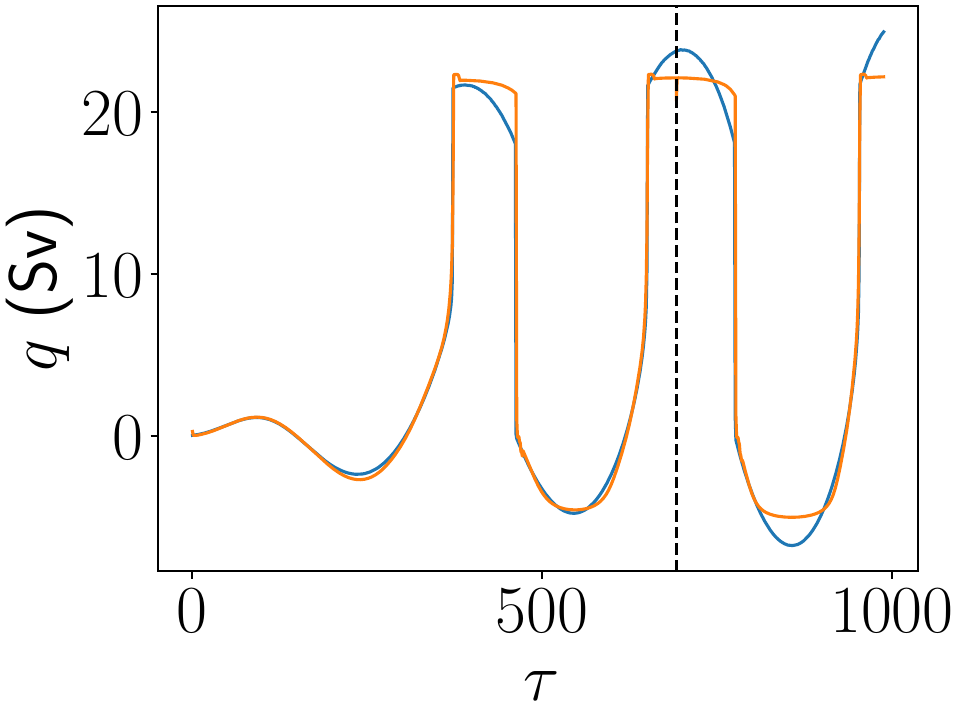} \\

	\bottomrule
	\caption{Predictive performance using \(F_s\), \(F_t\) and \(\rho^{\text{lin}}\).}%
	\label{tab:performance-extended-linear}%
\end{longtable}

\subsubsection{Explainability}

\begin{longtable}{p{.06\columnwidth}p{.08\columnwidth}C{.2\columnwidth}C{.2\columnwidth}C{.2\columnwidth}C{.2\columnwidth}}
	\toprule
	Scenario & Architecture & DeepLIFT\@: PI & SHAP\@: PI & DeepLIFT\@: AR & SHAP\@: AR \\
	\midrule

	\multirow[c]{3}{*}{\rotatebox[origin=c]{90}{\makecell{\(F_s\): Lin. \\ \(F_t\): Lin.}}}
	& BNN
	& \includegraphics[width=.7\linewidth,valign=m]{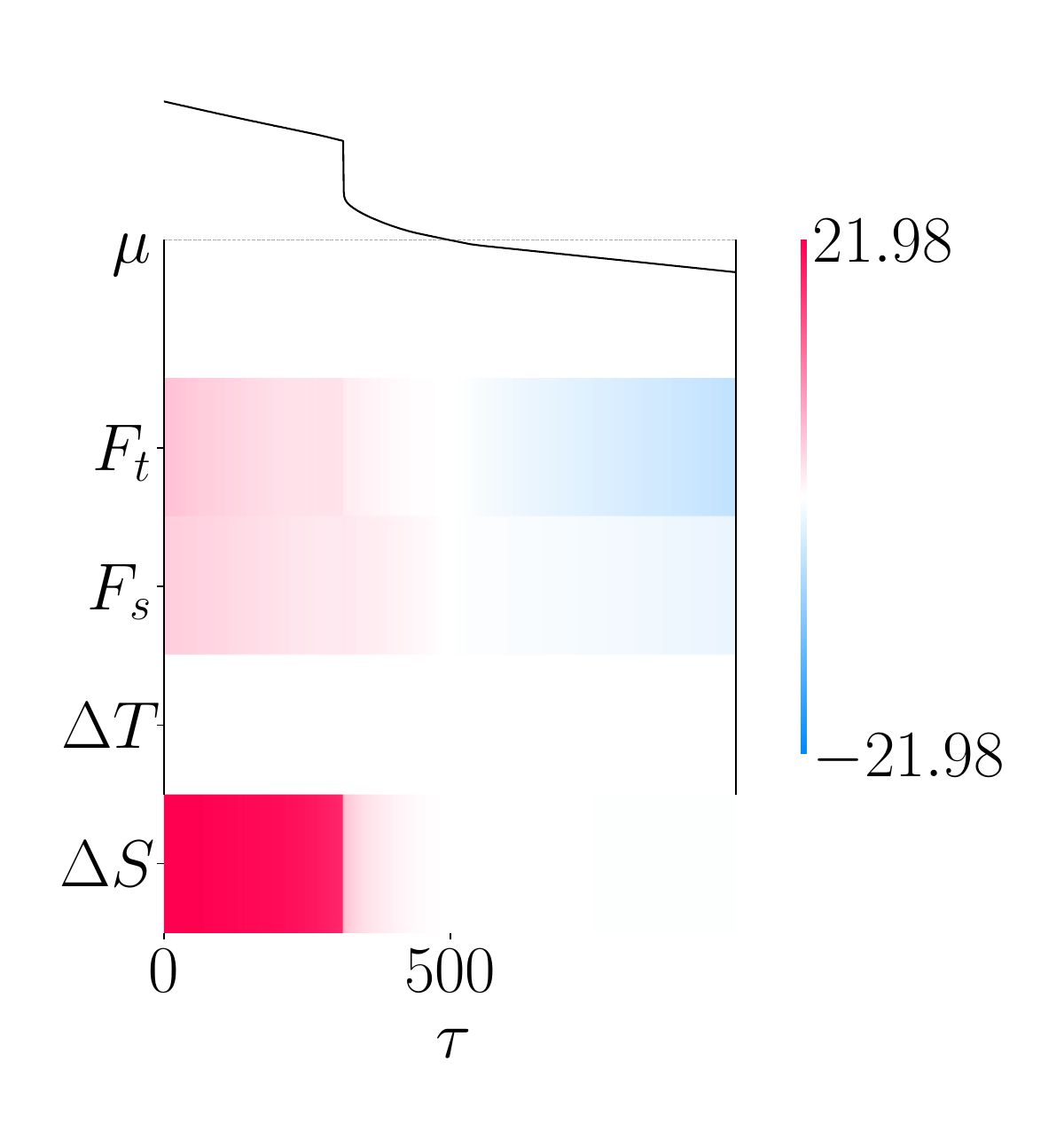}
	& \includegraphics[width=.7\linewidth,valign=m]{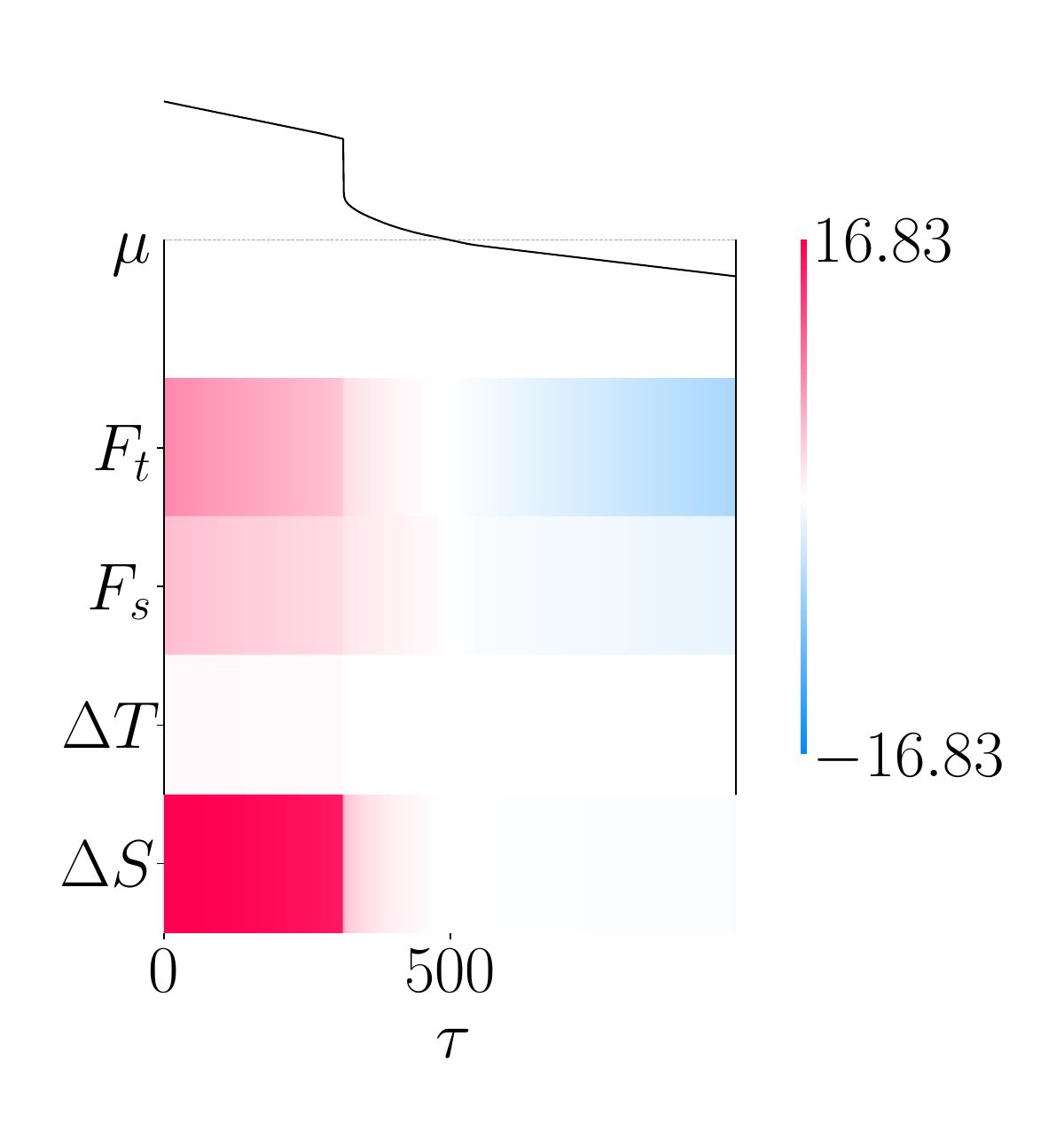}
	& \includegraphics[width=\linewidth,valign=m]{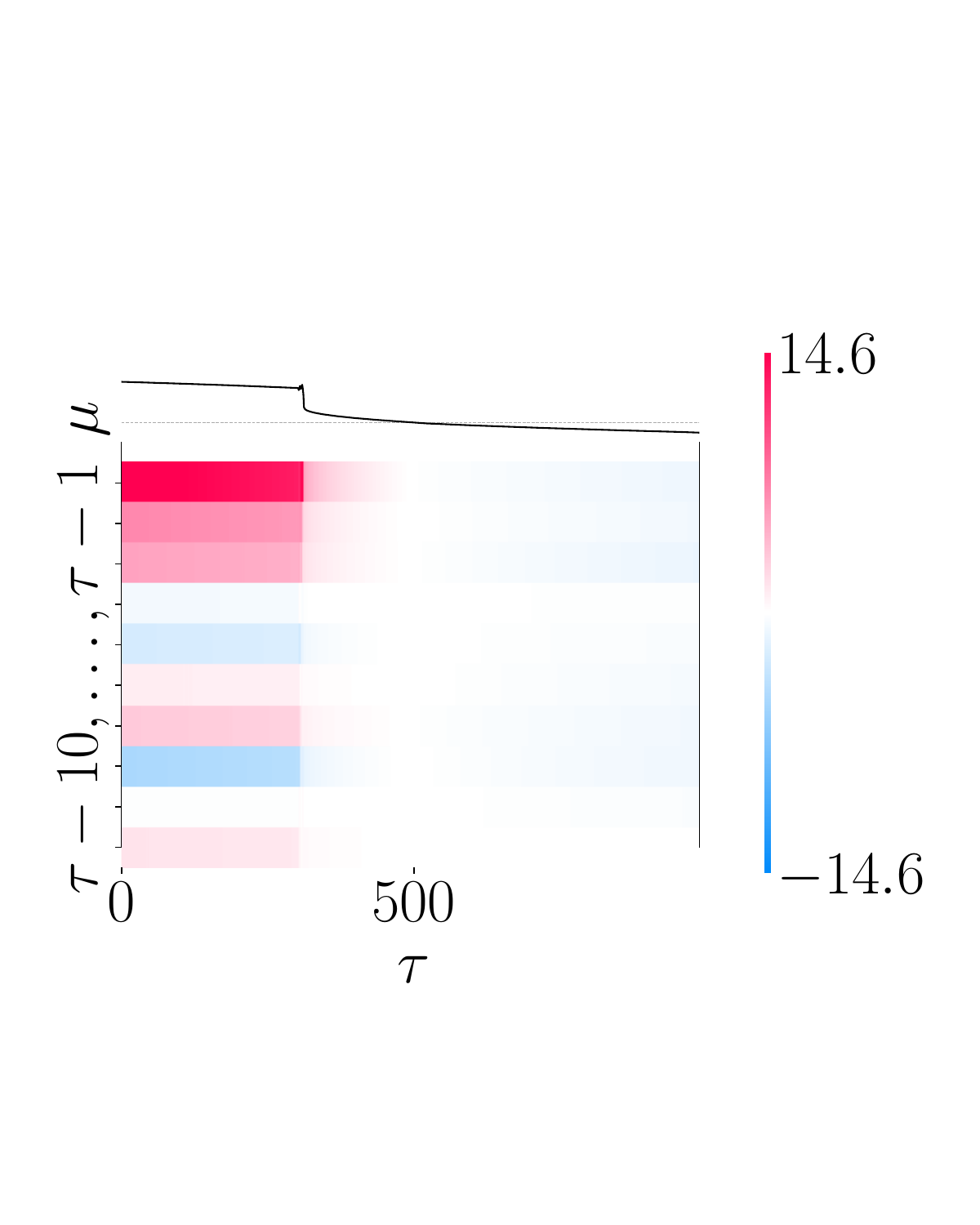}
	& \includegraphics[width=\linewidth,valign=m]{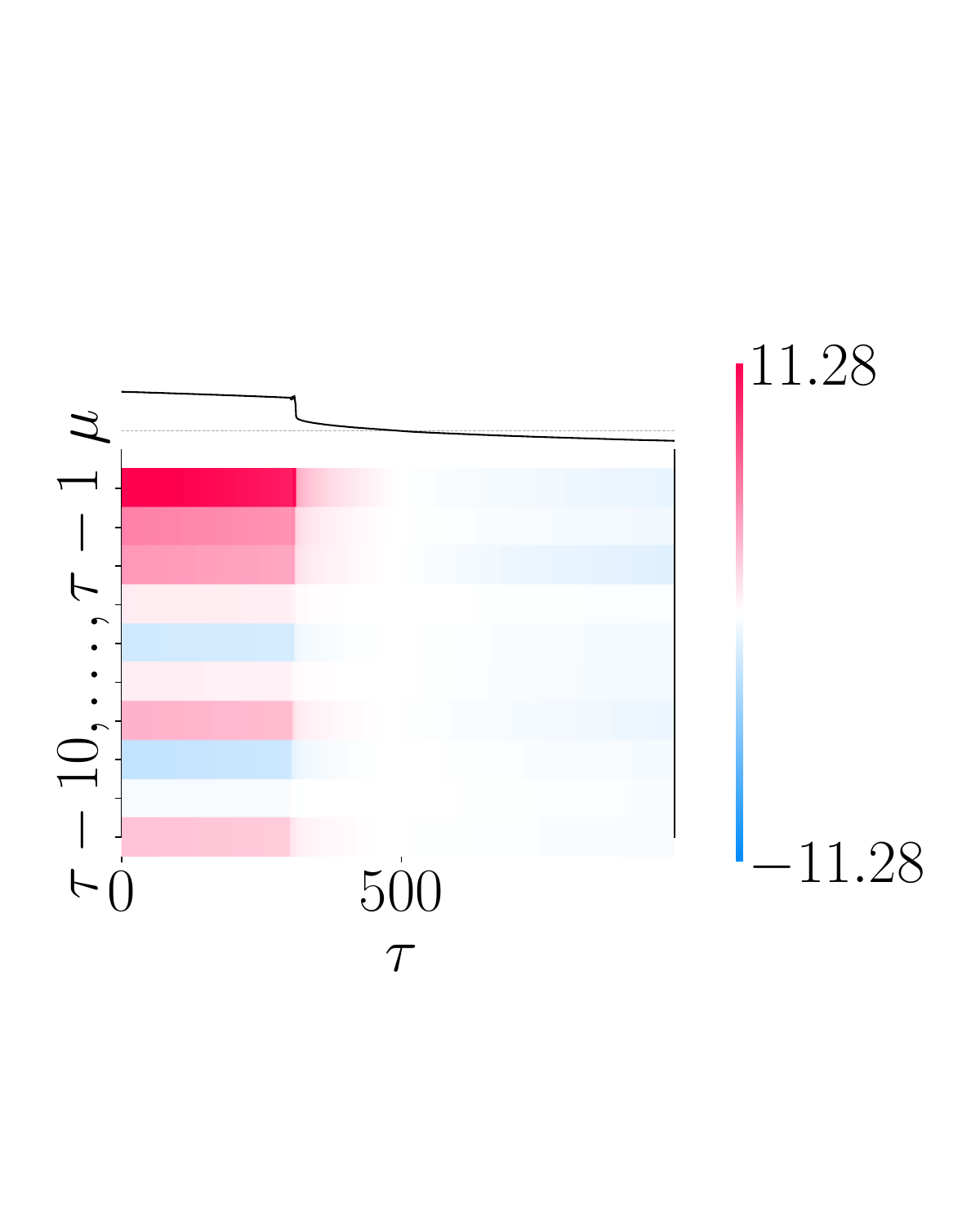} \\
	& MLP
	& \includegraphics[width=.7\linewidth,valign=m]{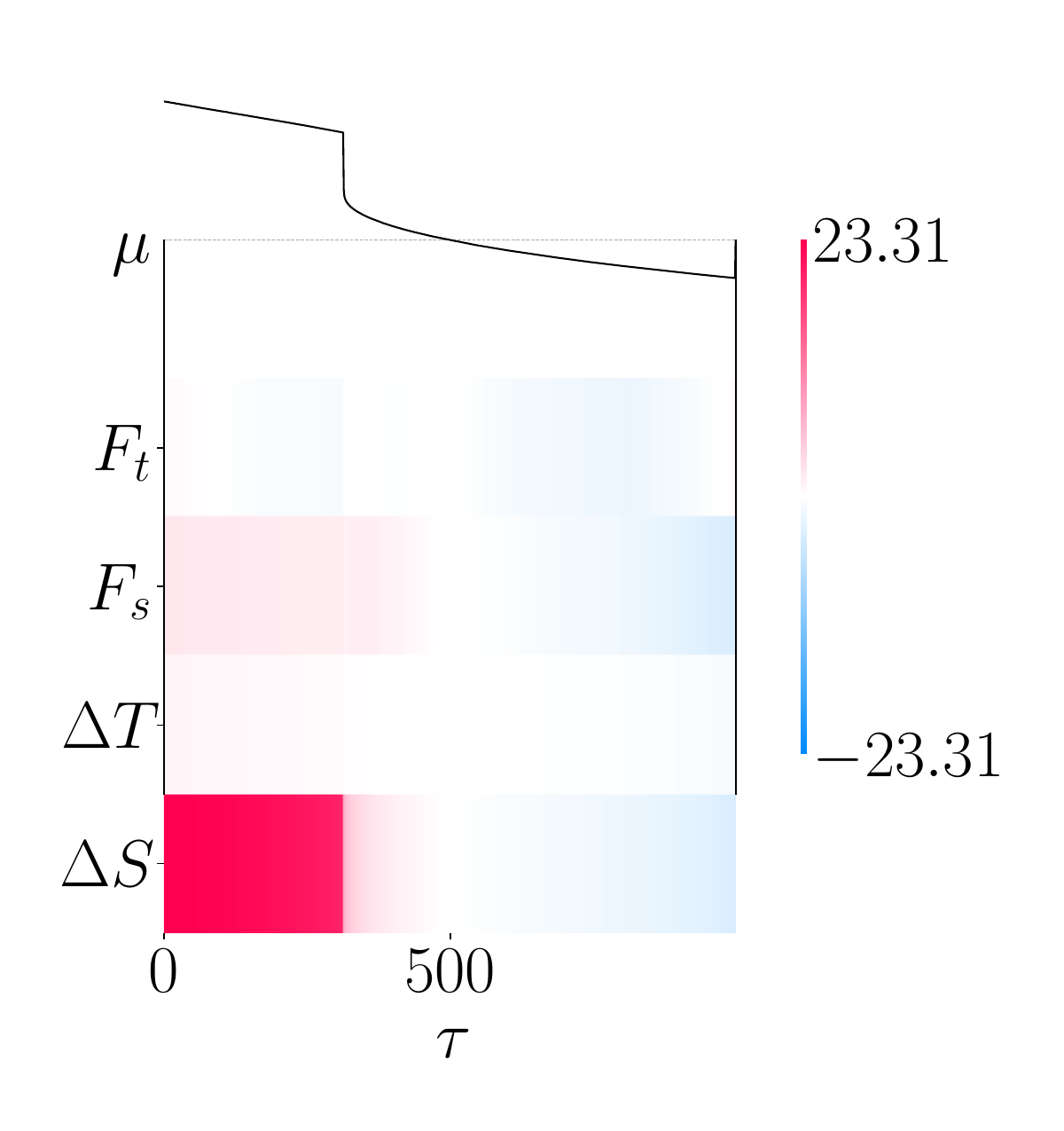}
	& \includegraphics[width=.7\linewidth,valign=m]{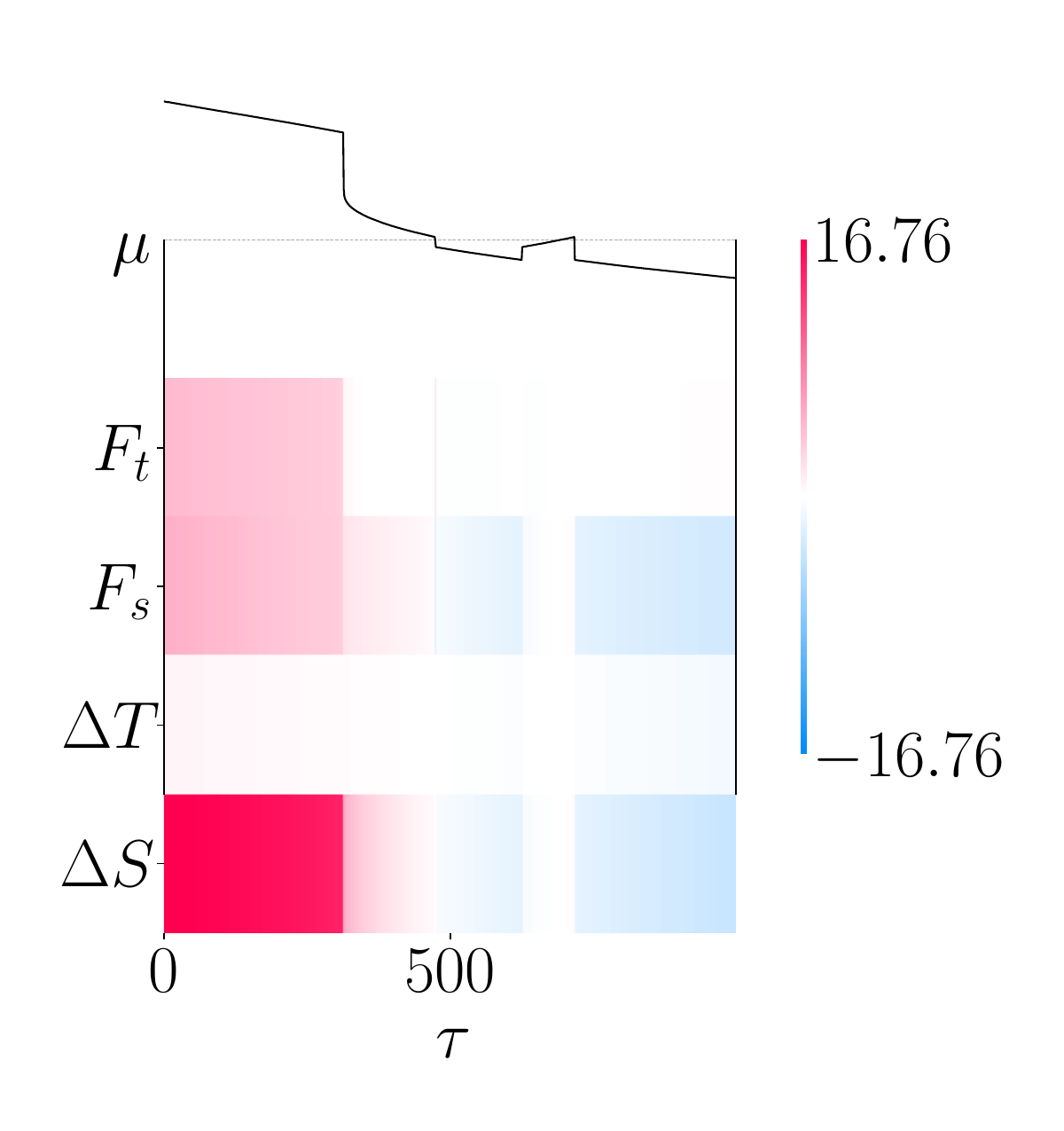}
	& \includegraphics[width=\linewidth,valign=m]{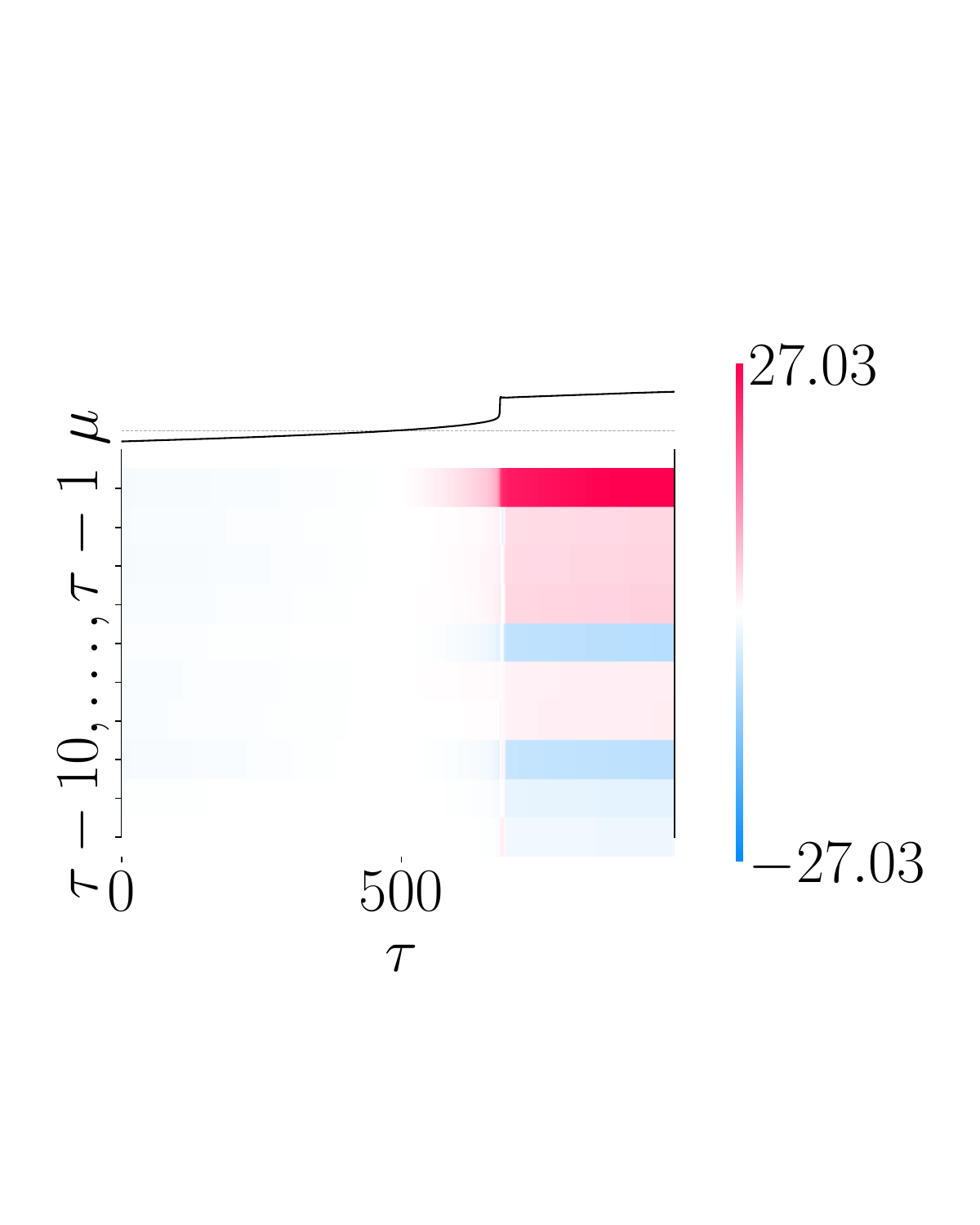}
	& \includegraphics[width=\linewidth,valign=m]{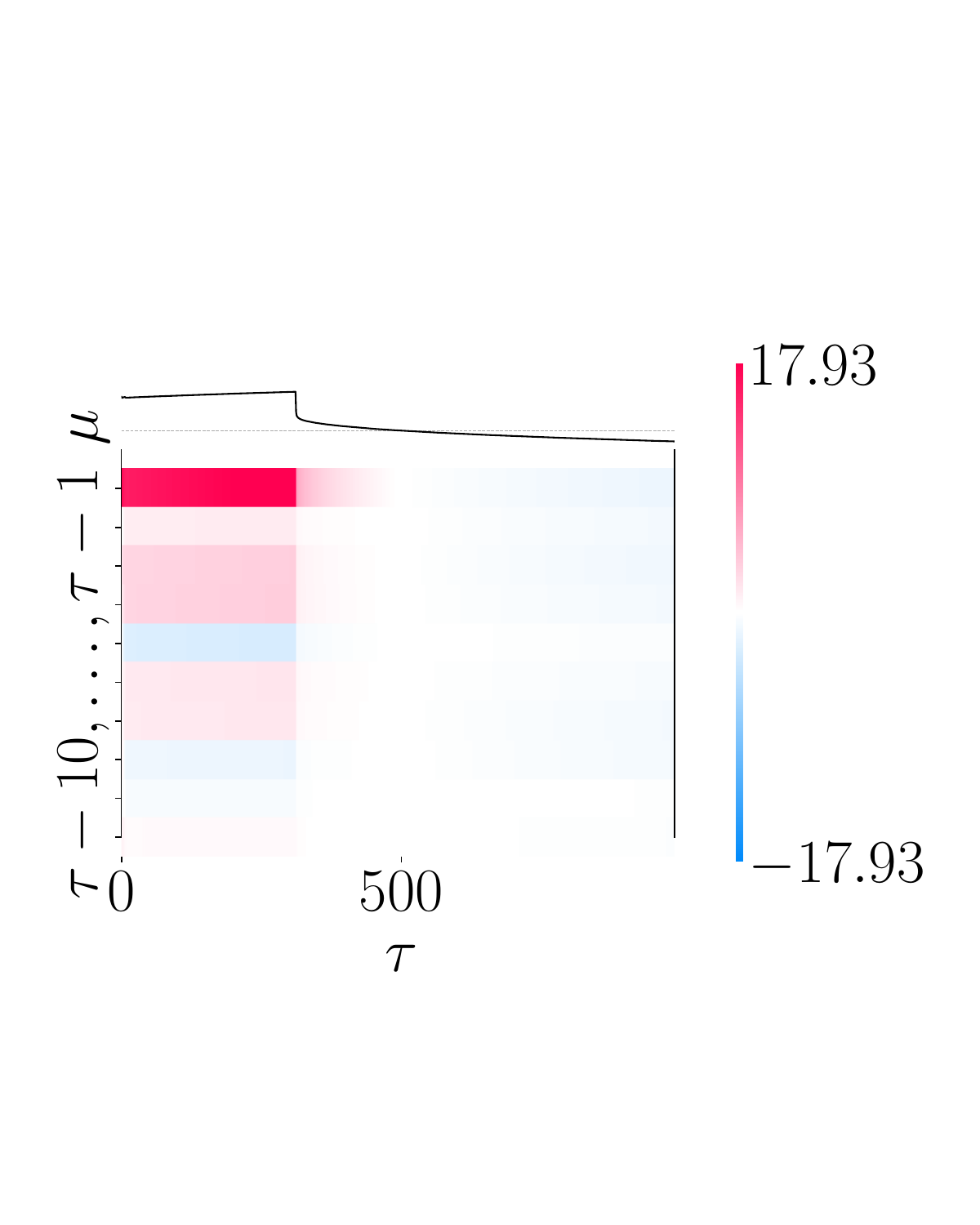} \\
	& Deep Ensemble
	& \includegraphics[width=.7\linewidth,valign=m]{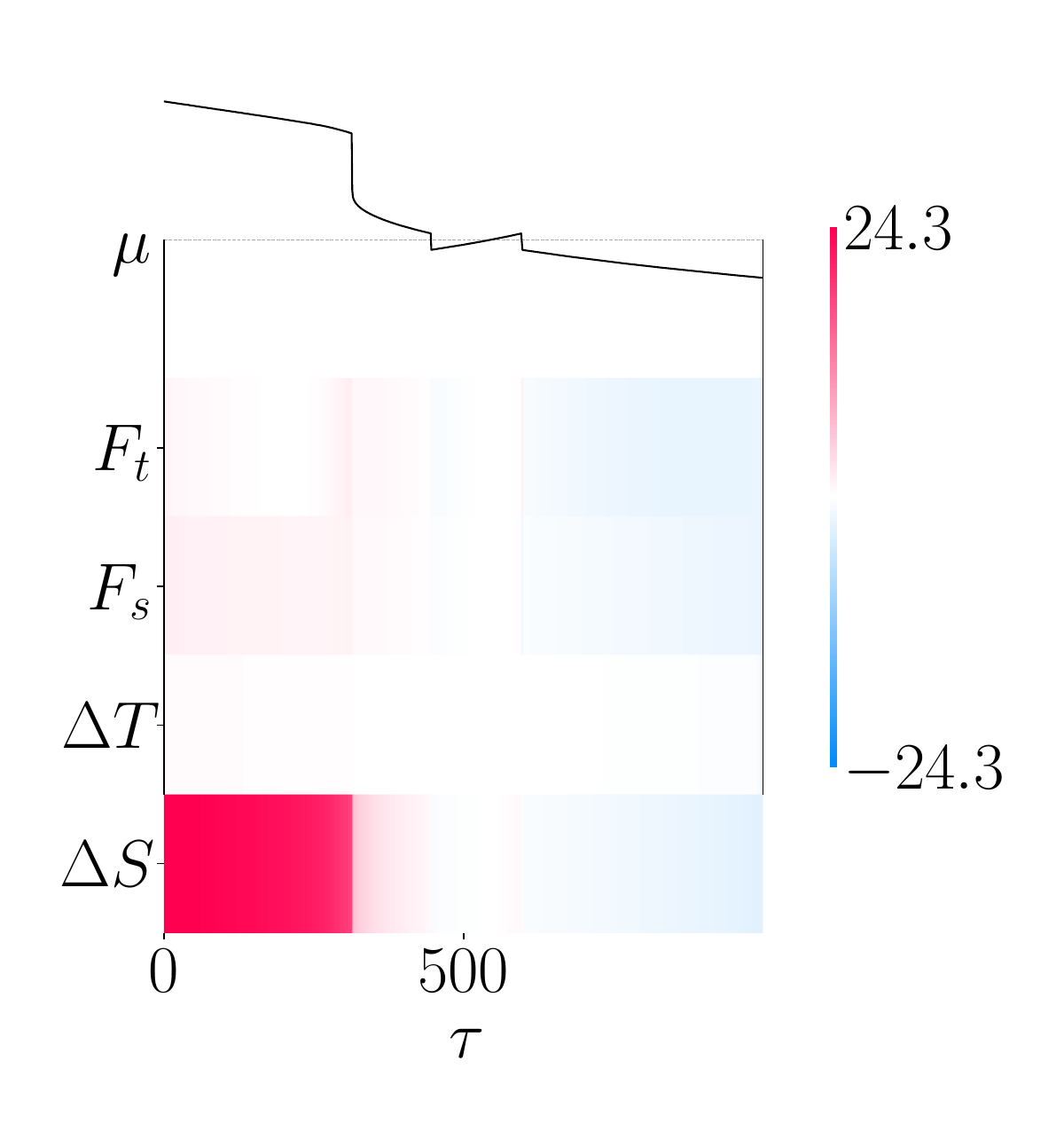}
	& \includegraphics[width=.7\linewidth,valign=m]{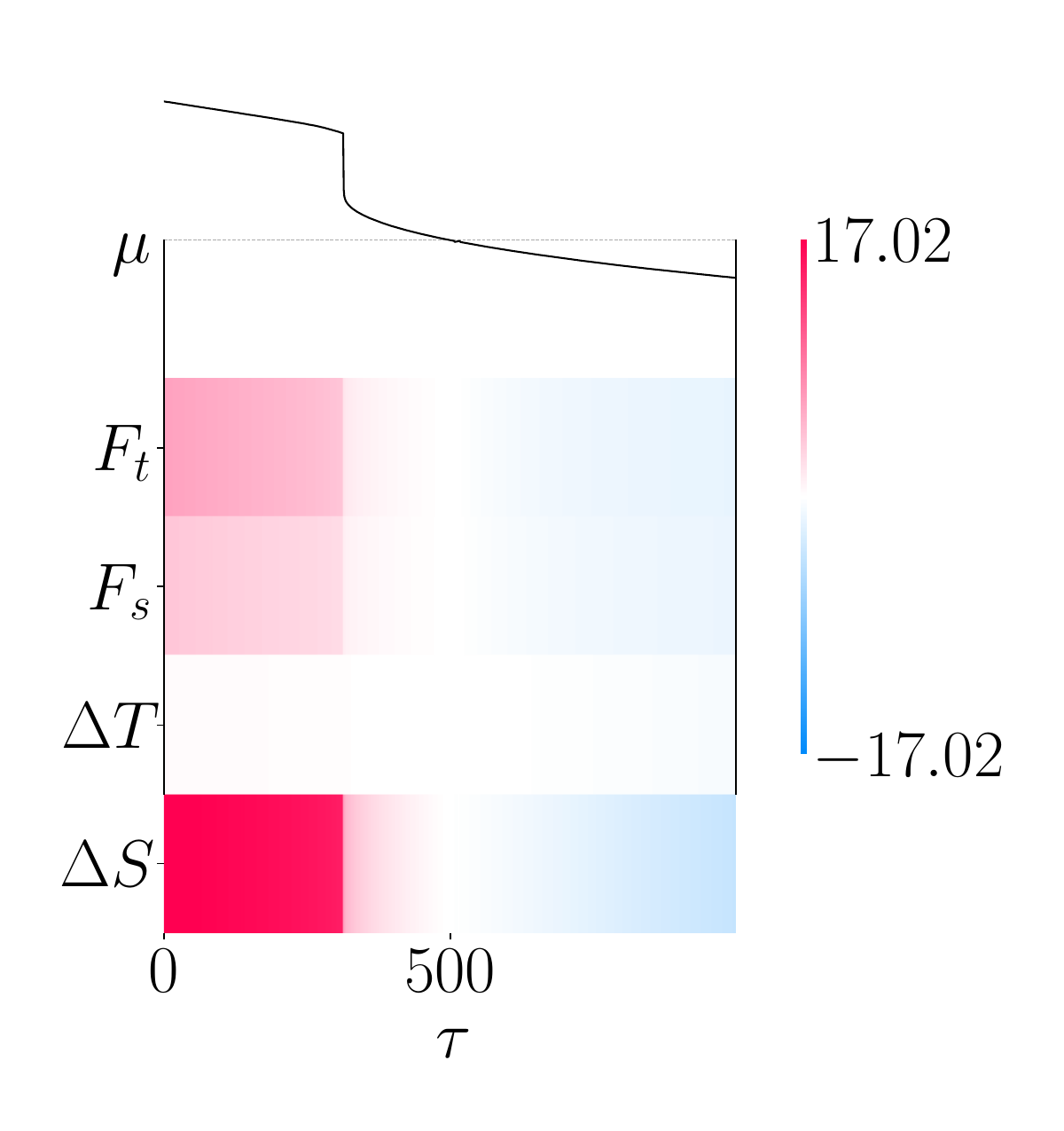}
	& \includegraphics[width=\linewidth,valign=m]{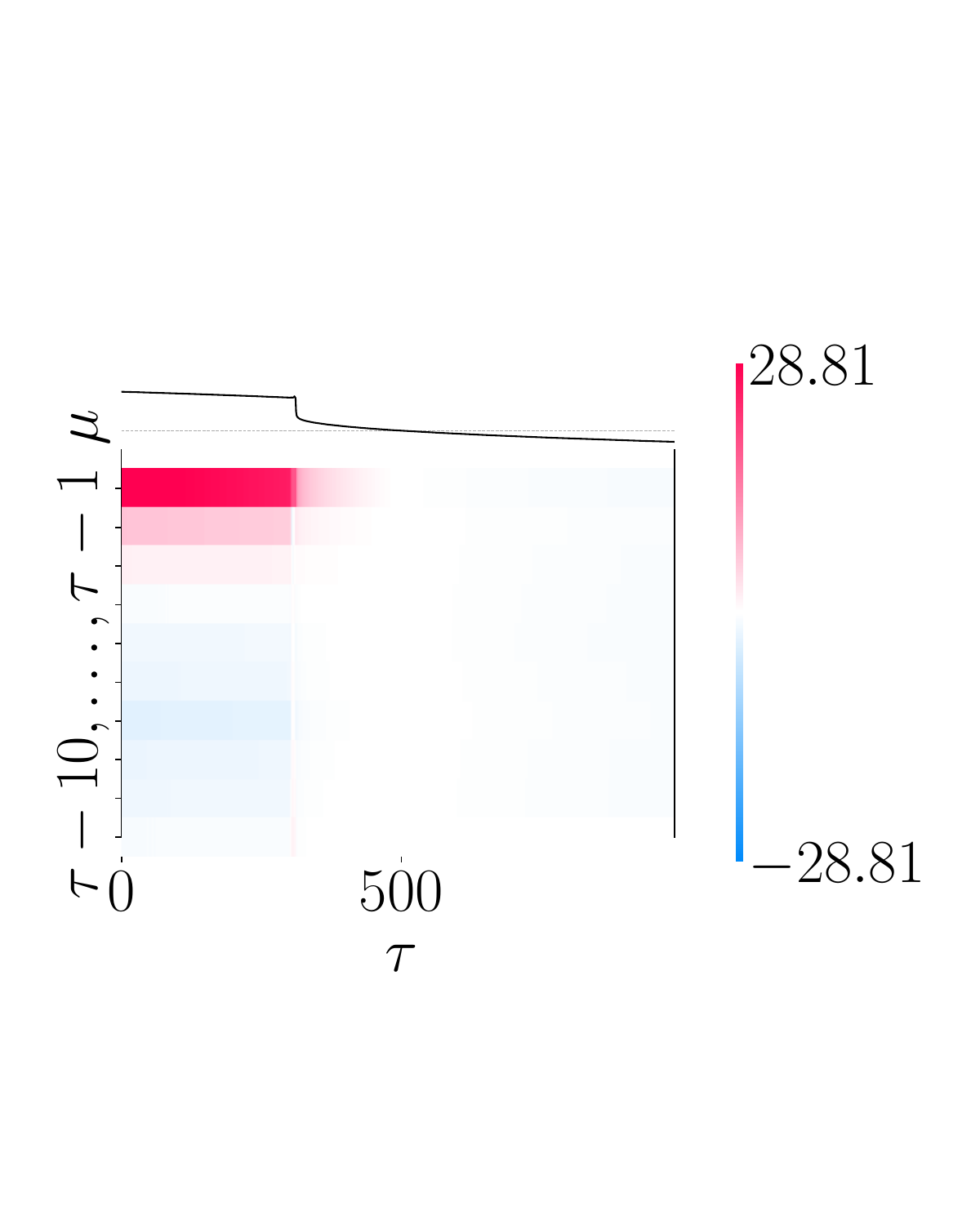}
	& \includegraphics[width=\linewidth,valign=m]{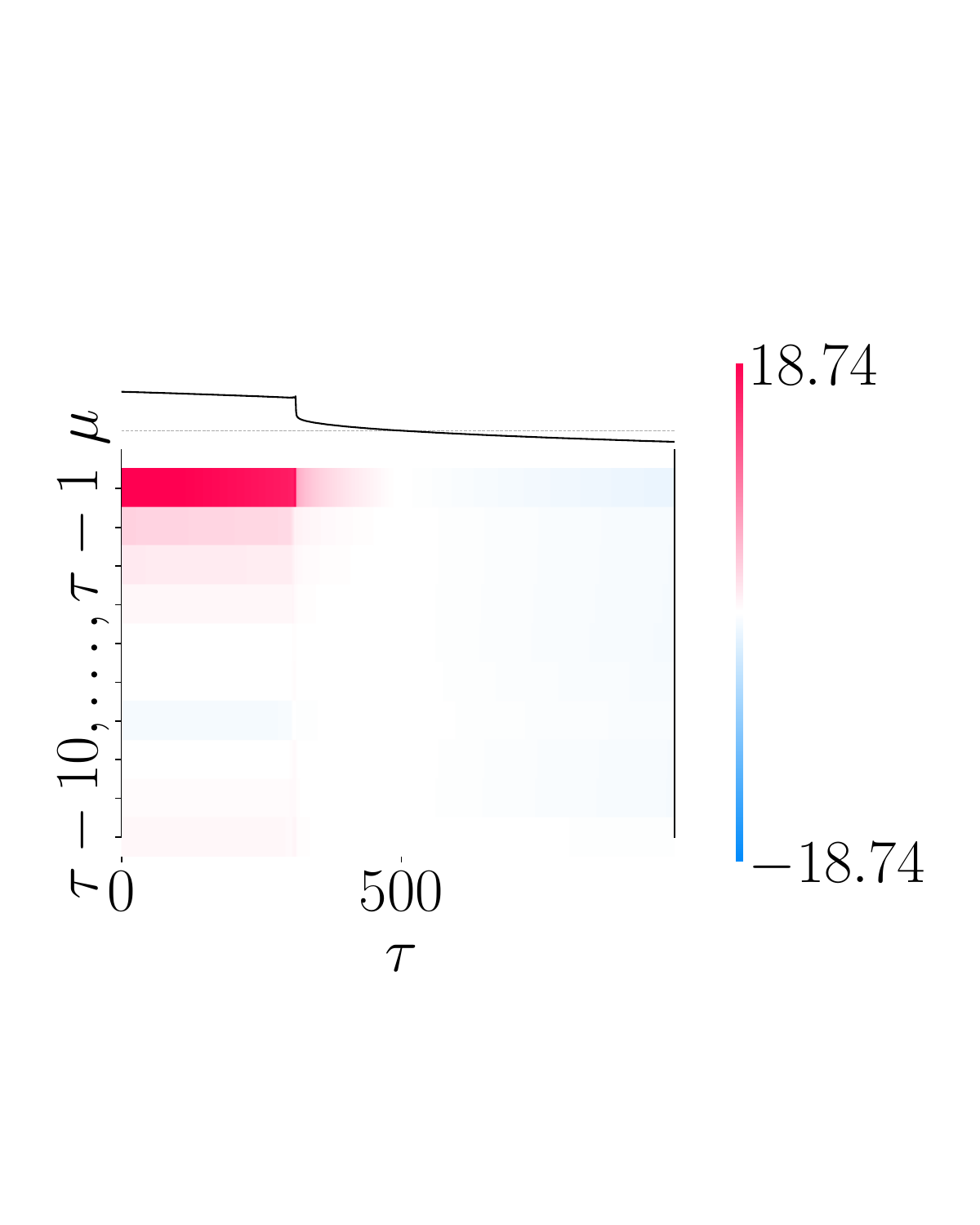} \\
	& RNN
	& ---
	& ---
	& \includegraphics[width=\linewidth,valign=m]{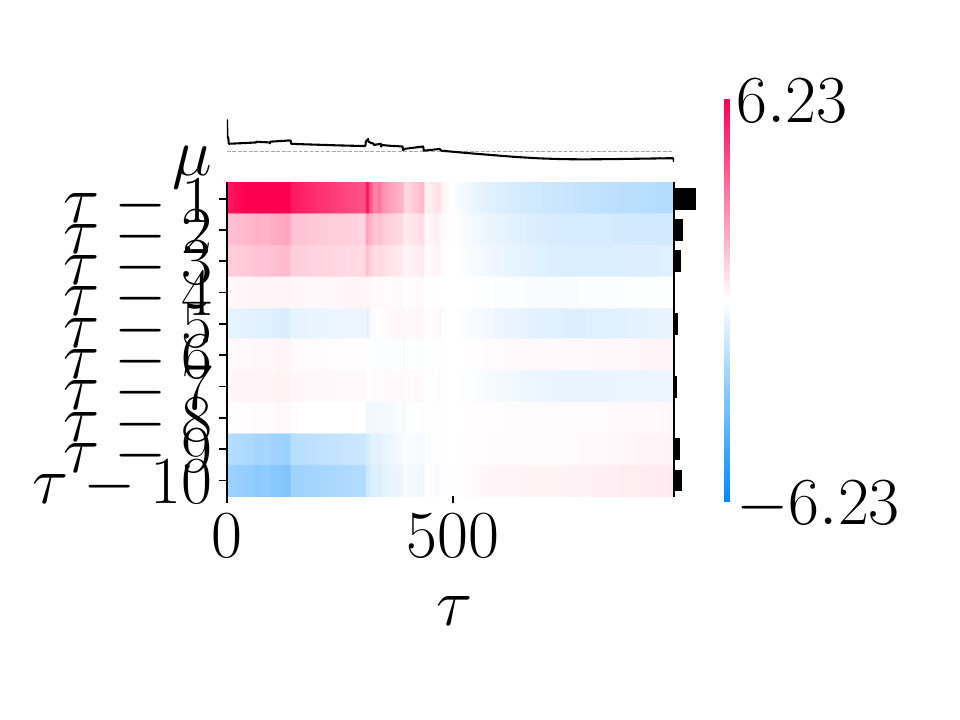}
	& \includegraphics[width=\linewidth,valign=m]{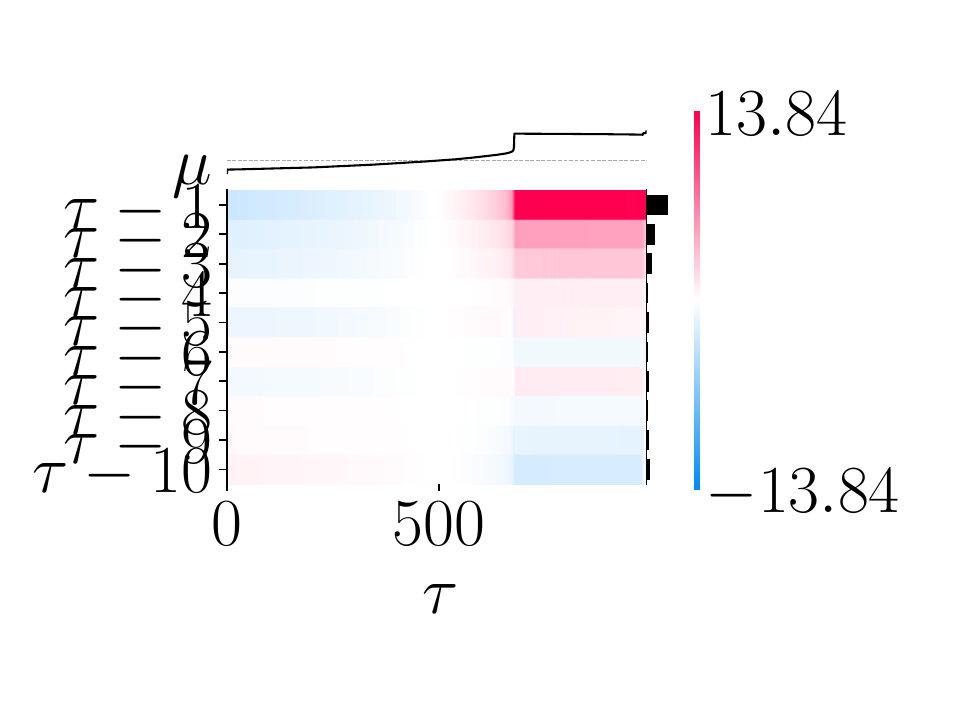} \\

	\midrule

	\multirow[c]{3}{*}{\rotatebox[origin=c]{90}{\makecell{\(F_s\): Sin.\ (stationary) \\ \(F_t\): Sin.\ (stationary)}}}
	& BNN
	& \includegraphics[width=.7\linewidth,valign=m]{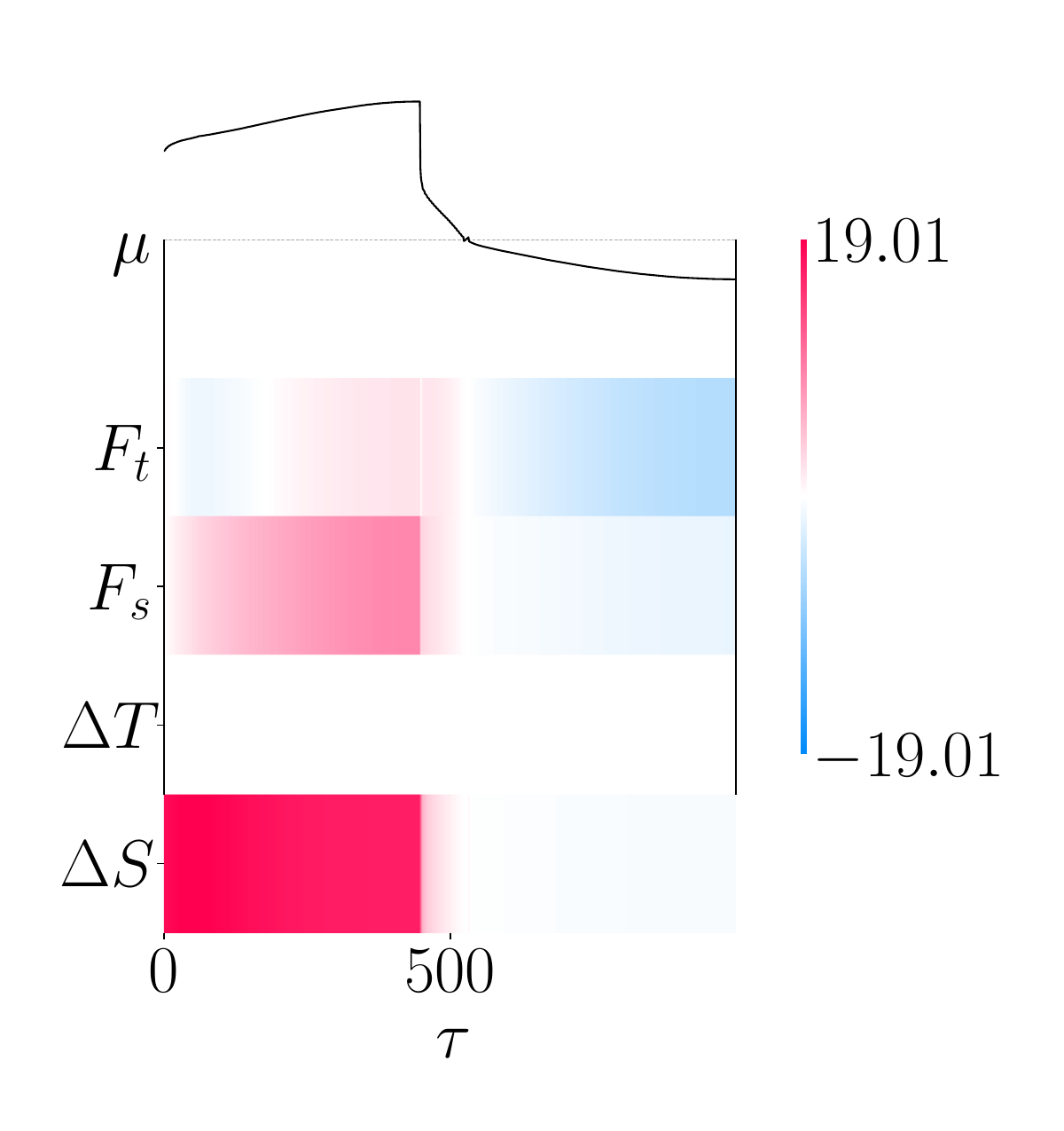}
	& \includegraphics[width=.7\linewidth,valign=m]{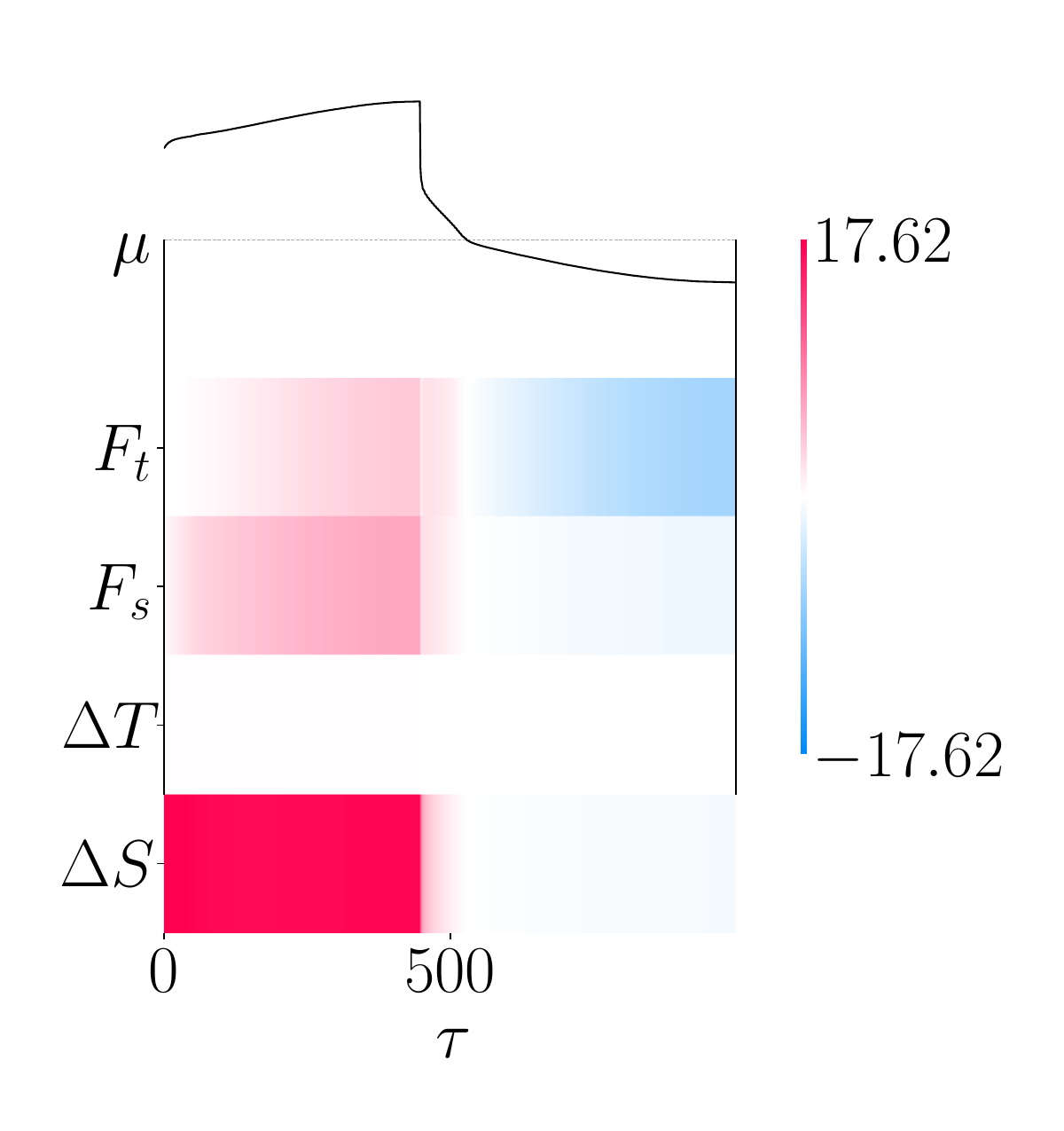}
	& \includegraphics[width=\linewidth,valign=m]{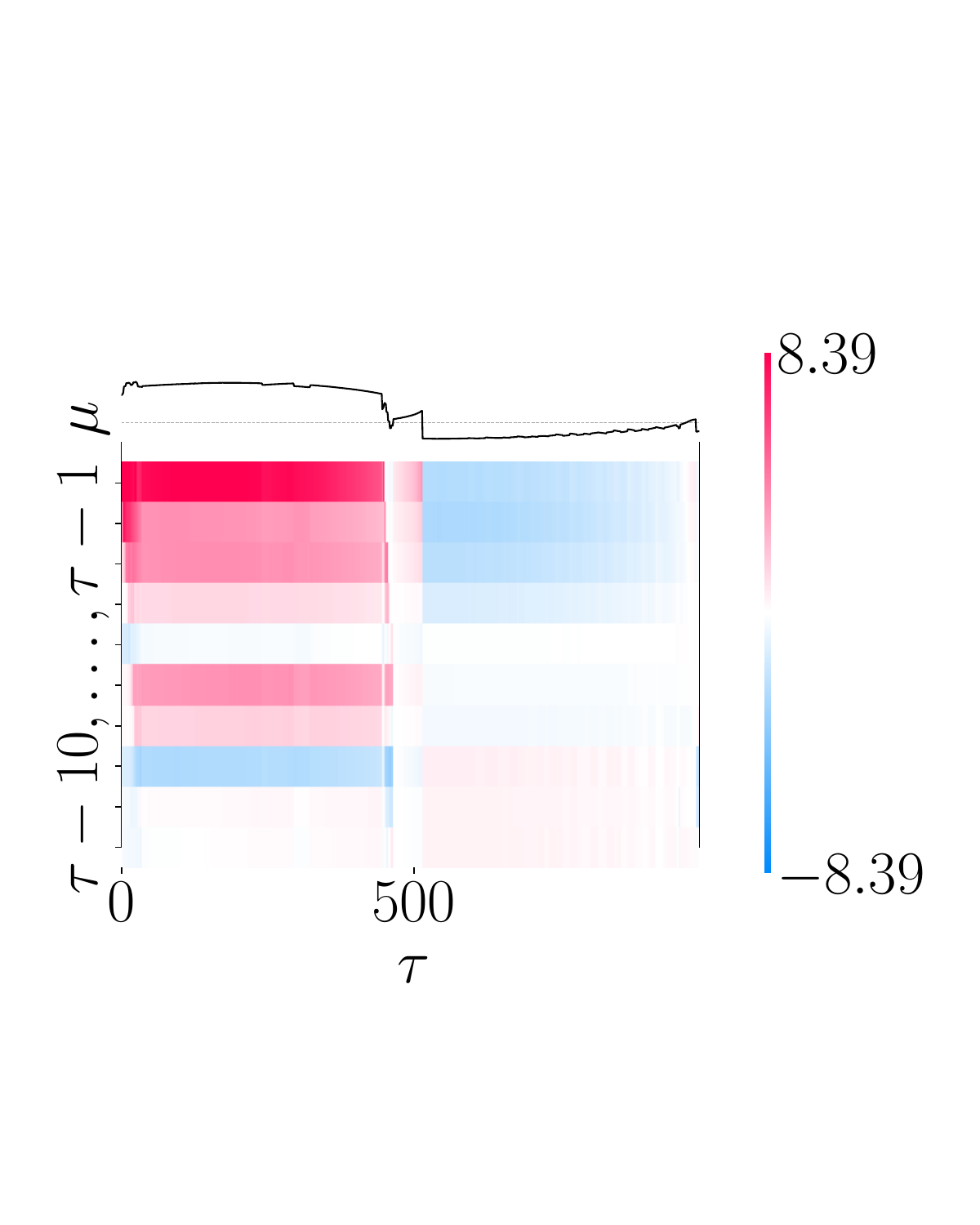}
	& \includegraphics[width=\linewidth,valign=m]{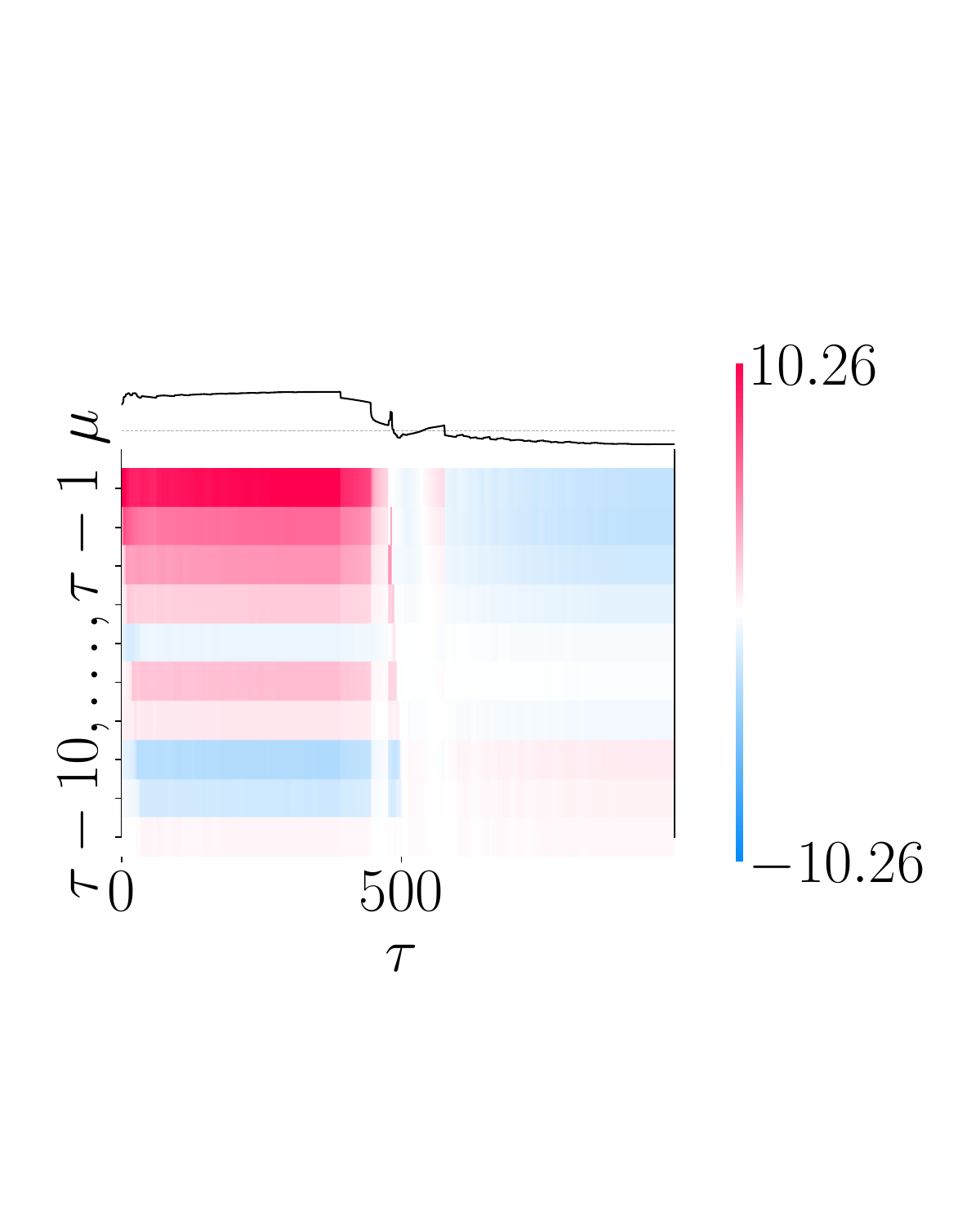} \\
	& MLP
	& \includegraphics[width=.7\linewidth,valign=m]{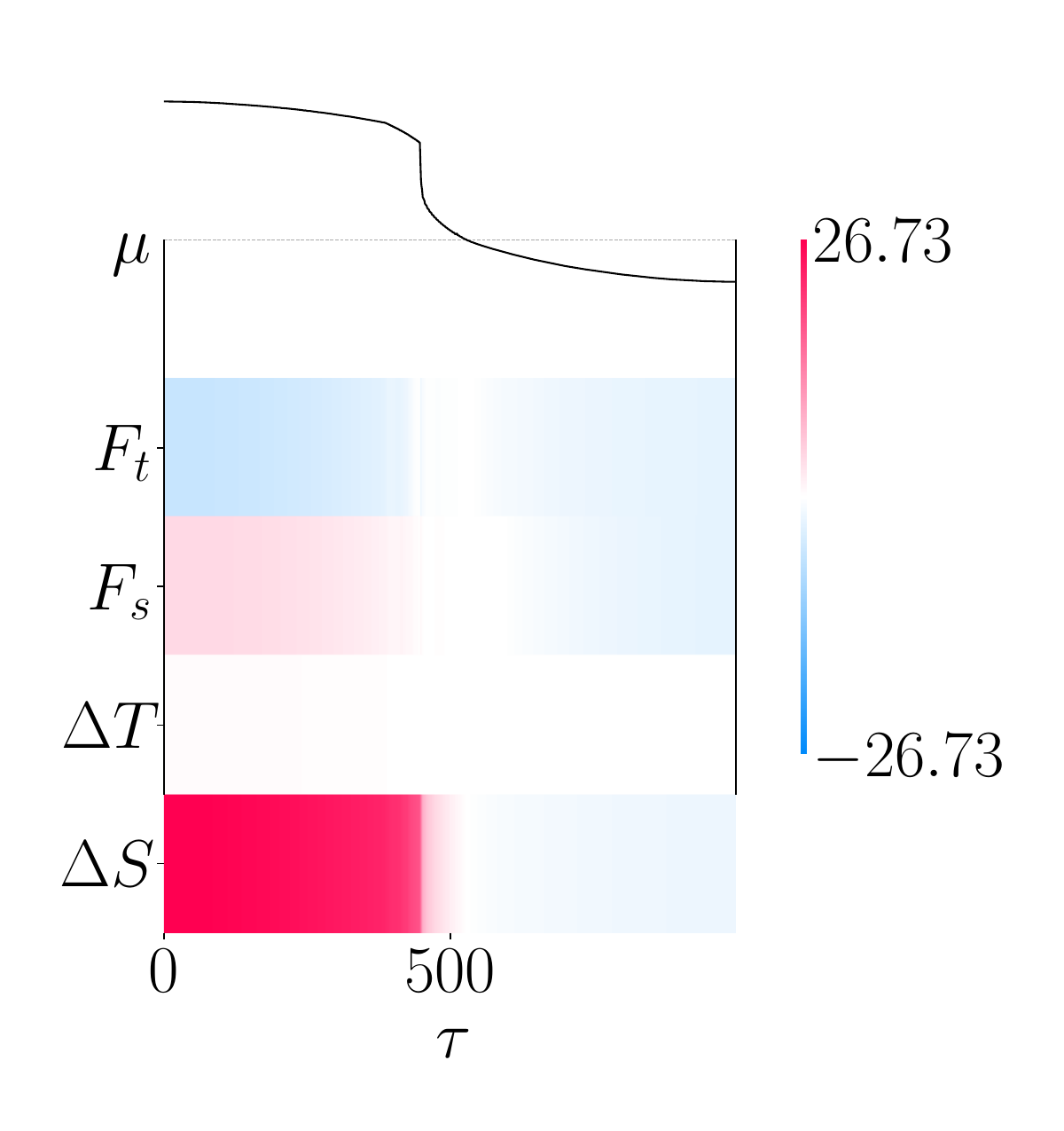}
	& \includegraphics[width=.7\linewidth,valign=m]{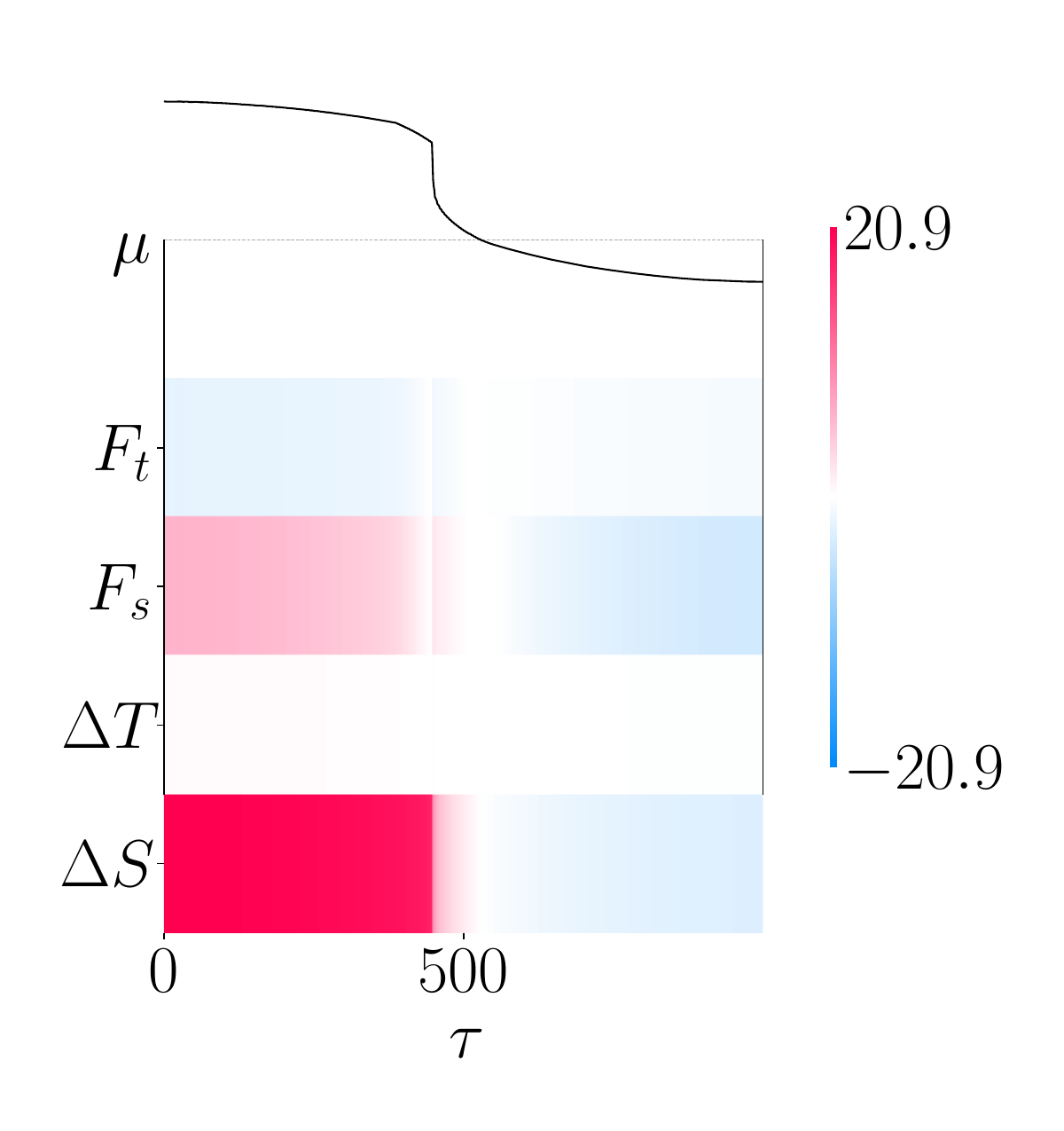}
	& \includegraphics[width=\linewidth,valign=m]{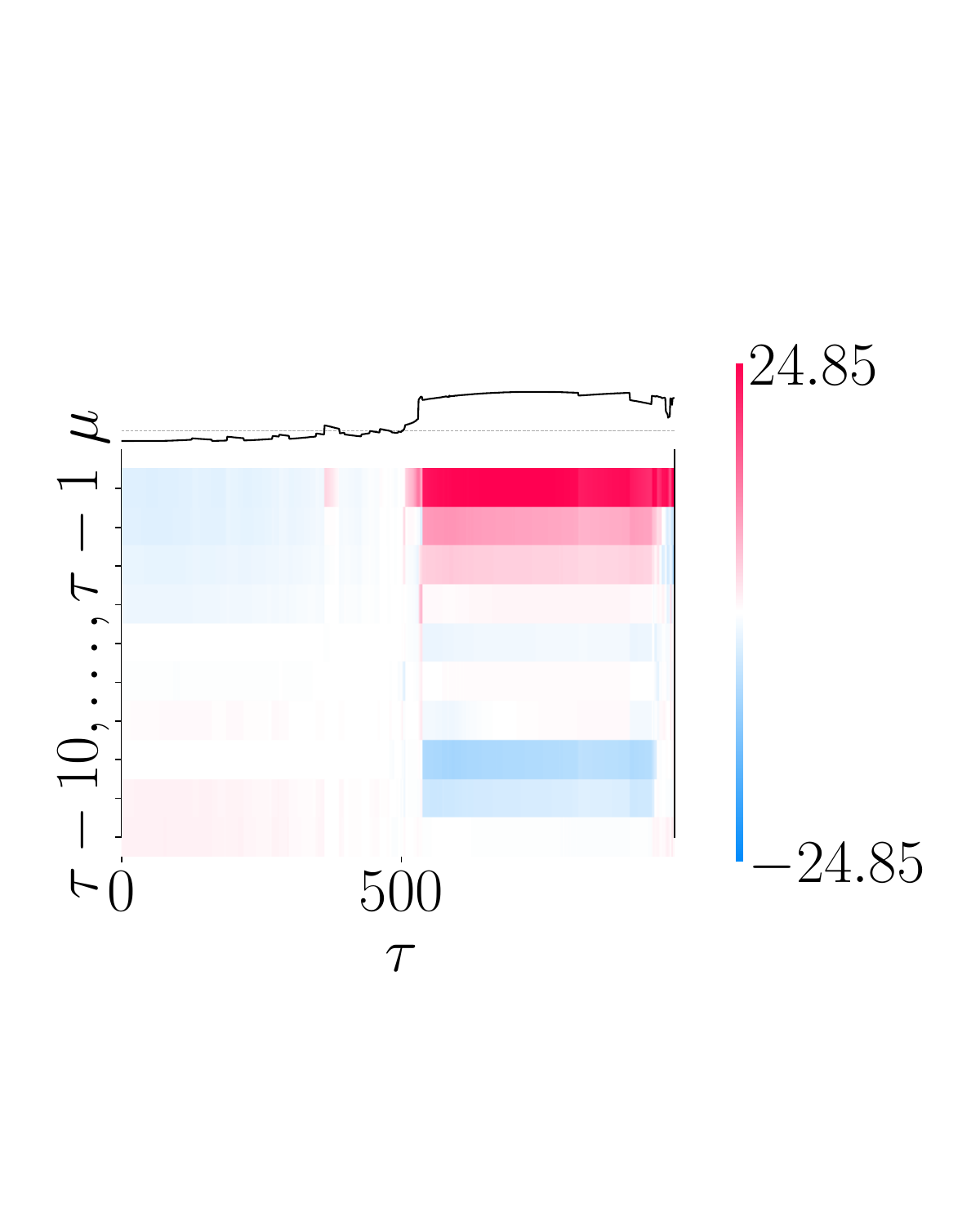}
	& \includegraphics[width=\linewidth,valign=m]{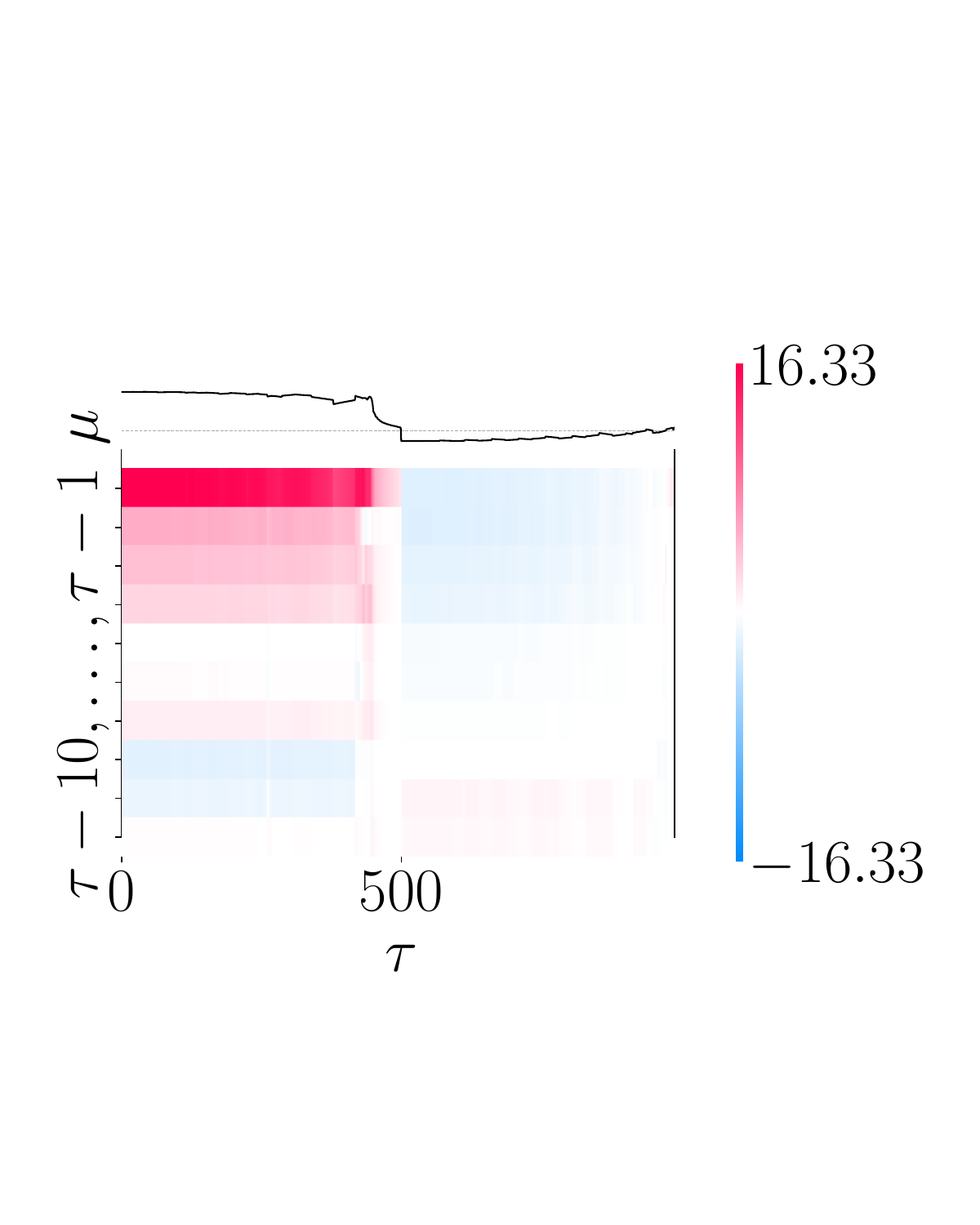} \\
	& Deep Ensemble
	& \includegraphics[width=.7\linewidth,valign=m]{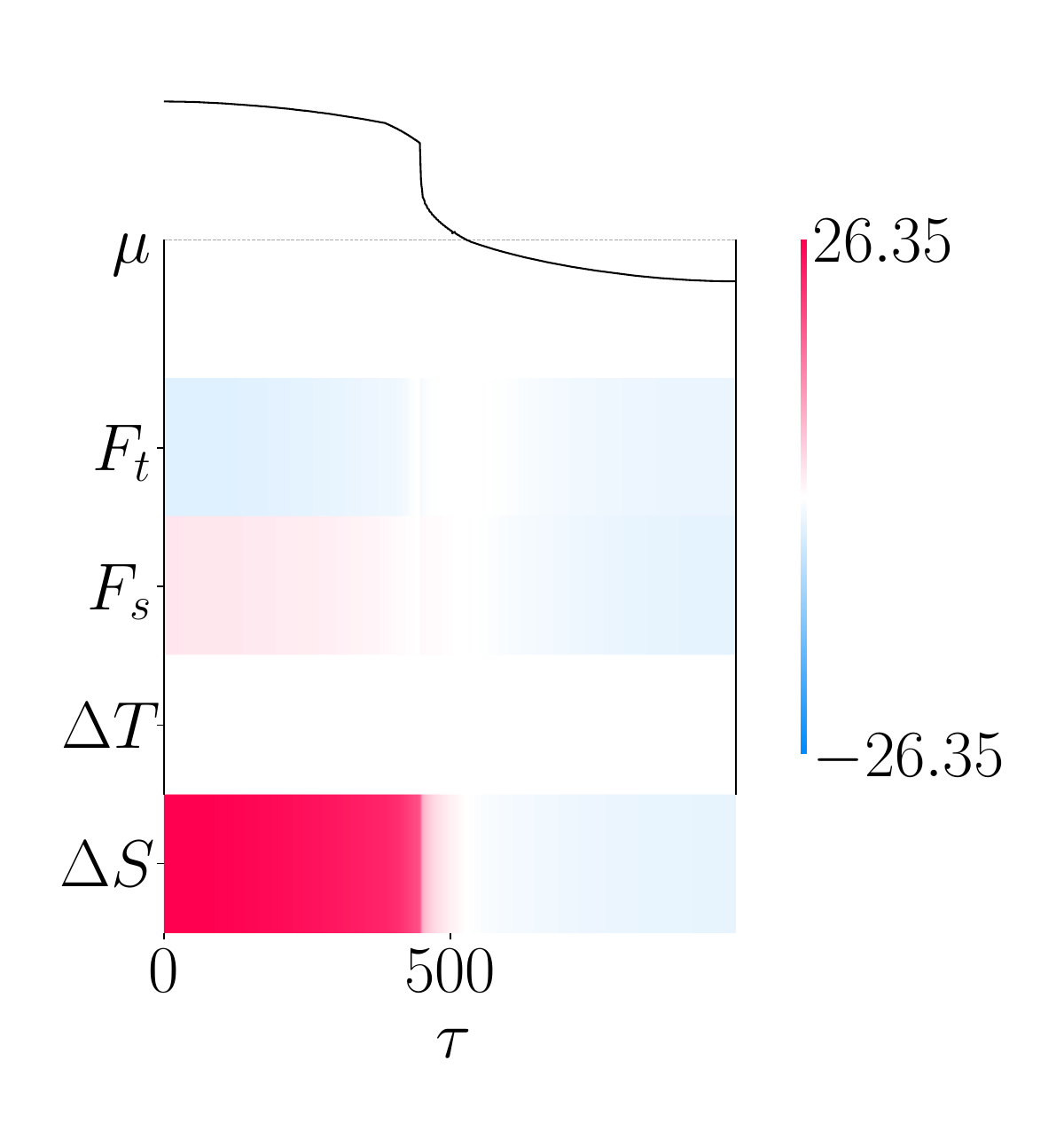}
	& \includegraphics[width=.7\linewidth,valign=m]{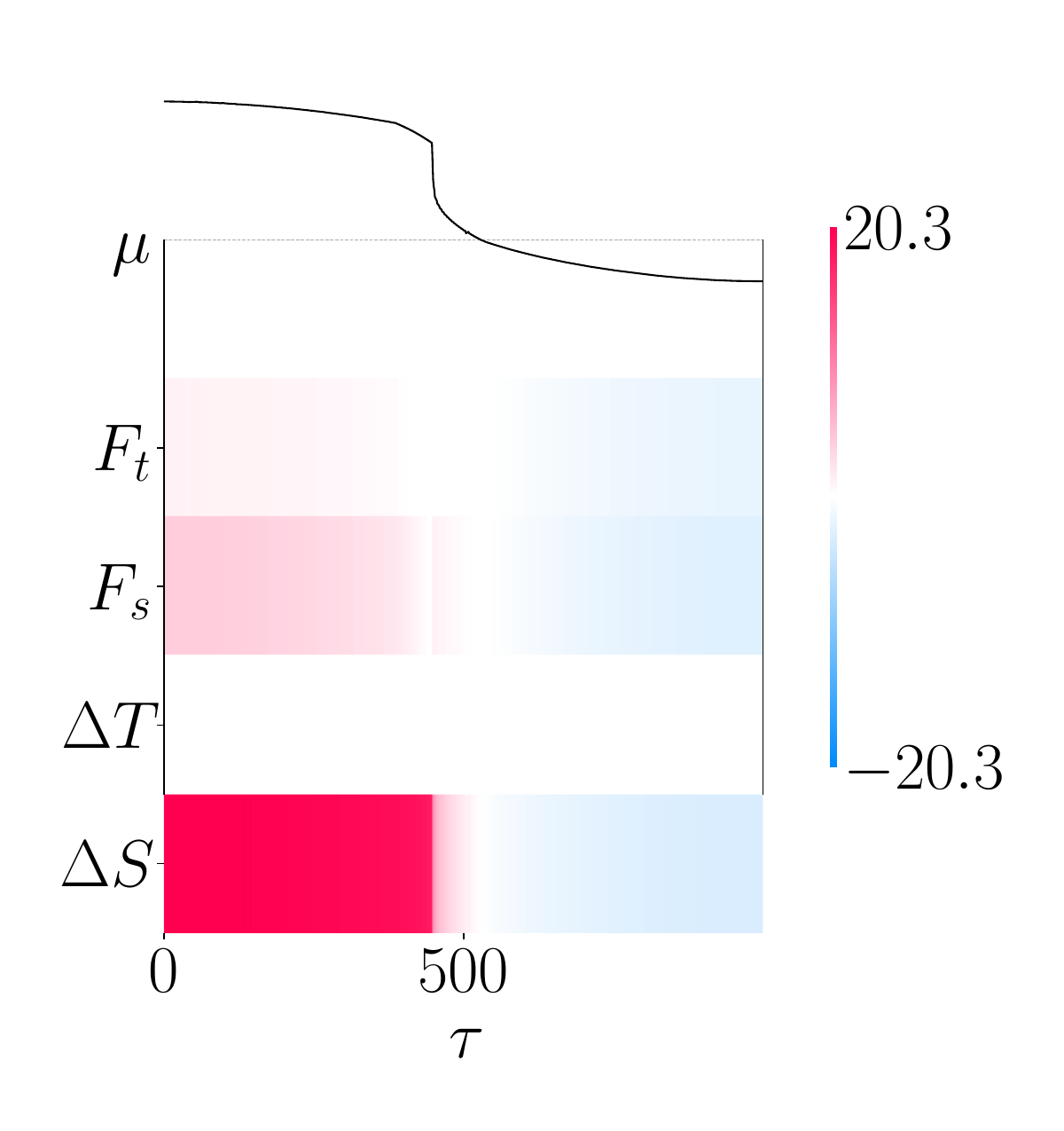}
	& \includegraphics[width=\linewidth,valign=m]{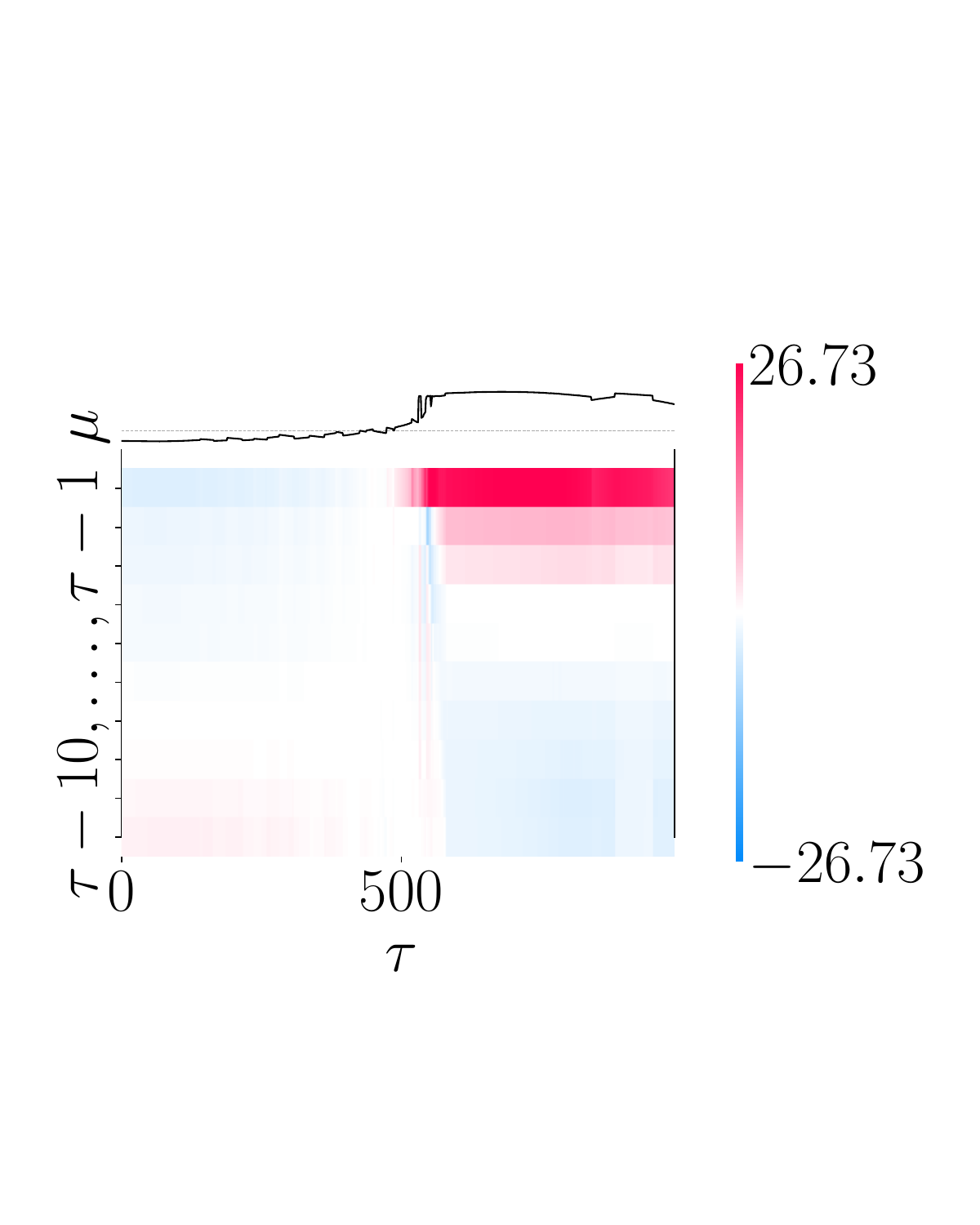}
	& \includegraphics[width=\linewidth,valign=m]{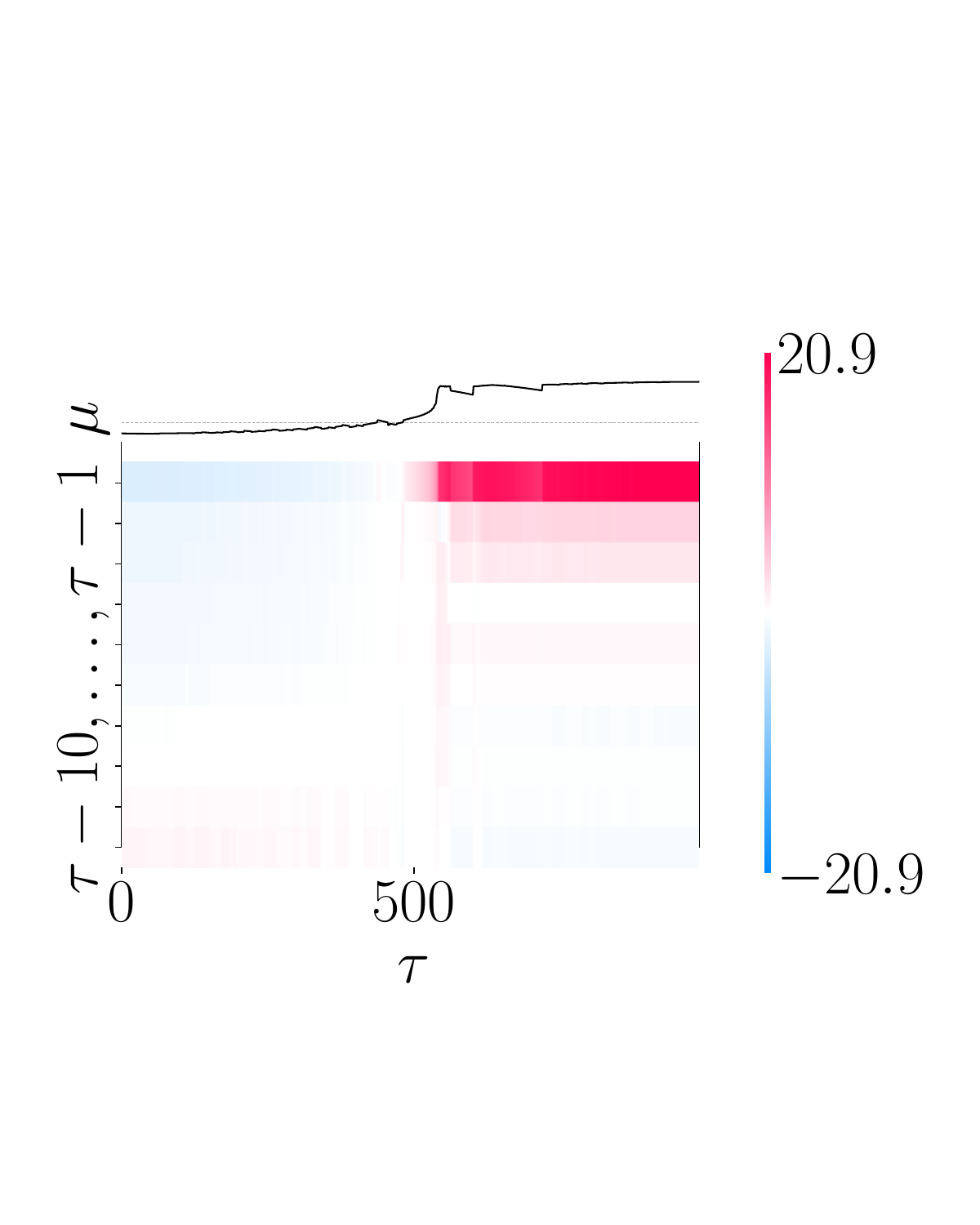} \\
	& RNN
	& ---
	& ---
	& \includegraphics[width=\linewidth,valign=m]{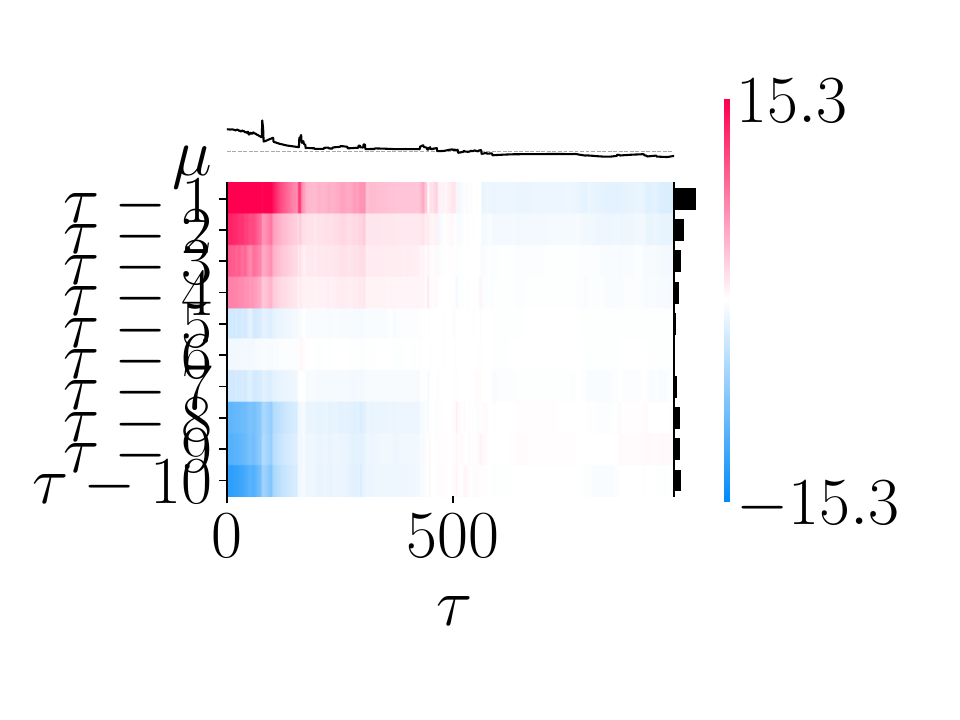}
	& \includegraphics[width=\linewidth,valign=m]{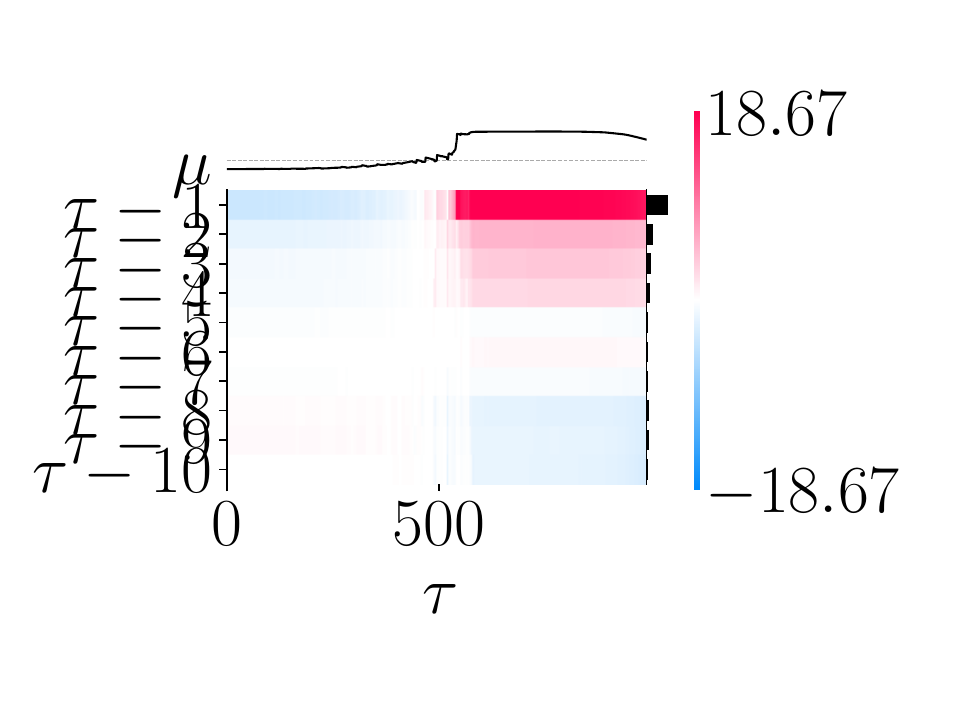} \\

	\midrule

	\multirow[c]{3}{*}{\rotatebox[origin=c]{90}{\makecell{\(F_s\): Sin.\ (nonstationary) \\ \(F_t\): Sin.\ (nonstationary)}}}
	& BNN
	& \includegraphics[width=.7\linewidth,valign=m]{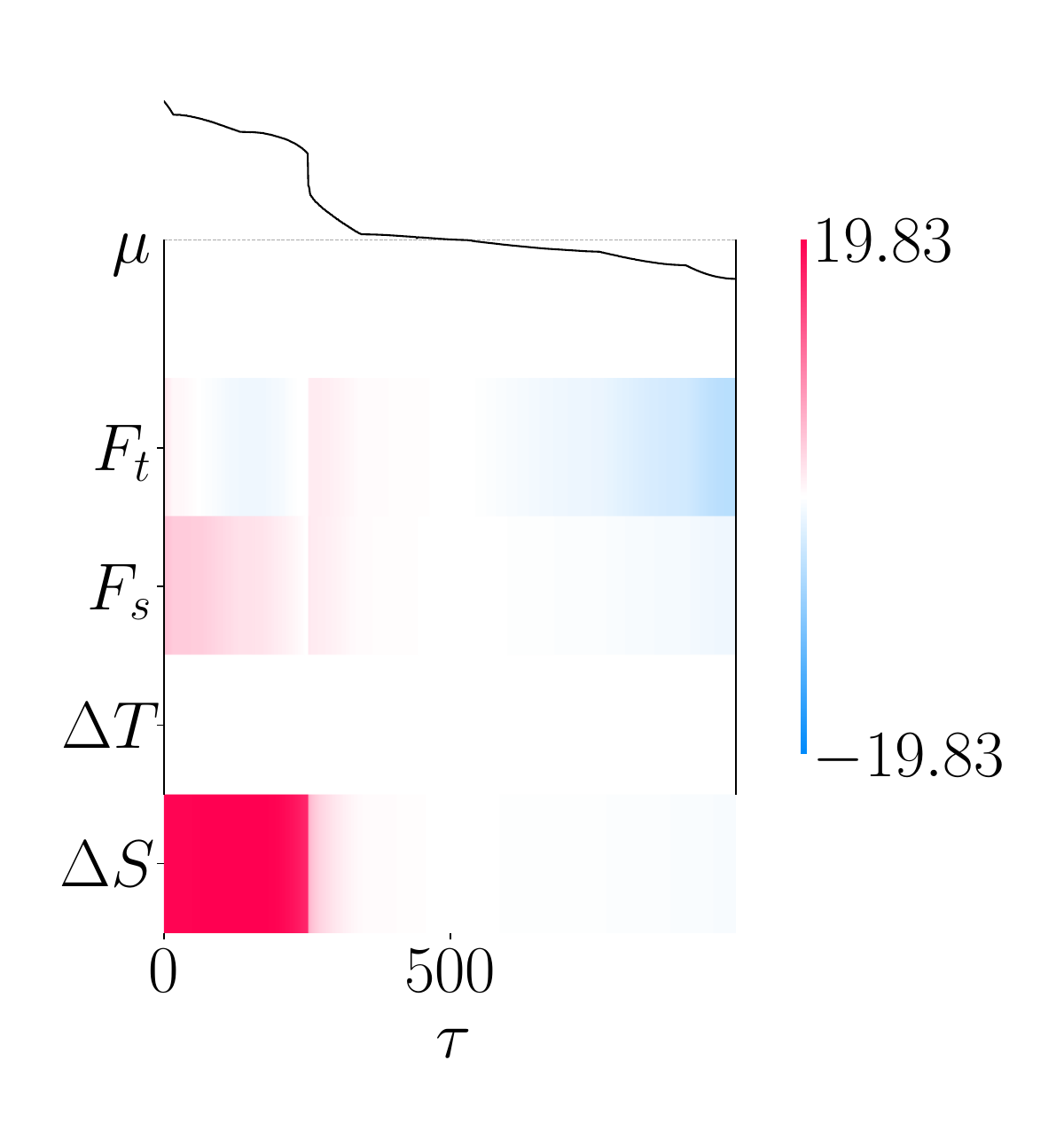}
	& \includegraphics[width=.7\linewidth,valign=m]{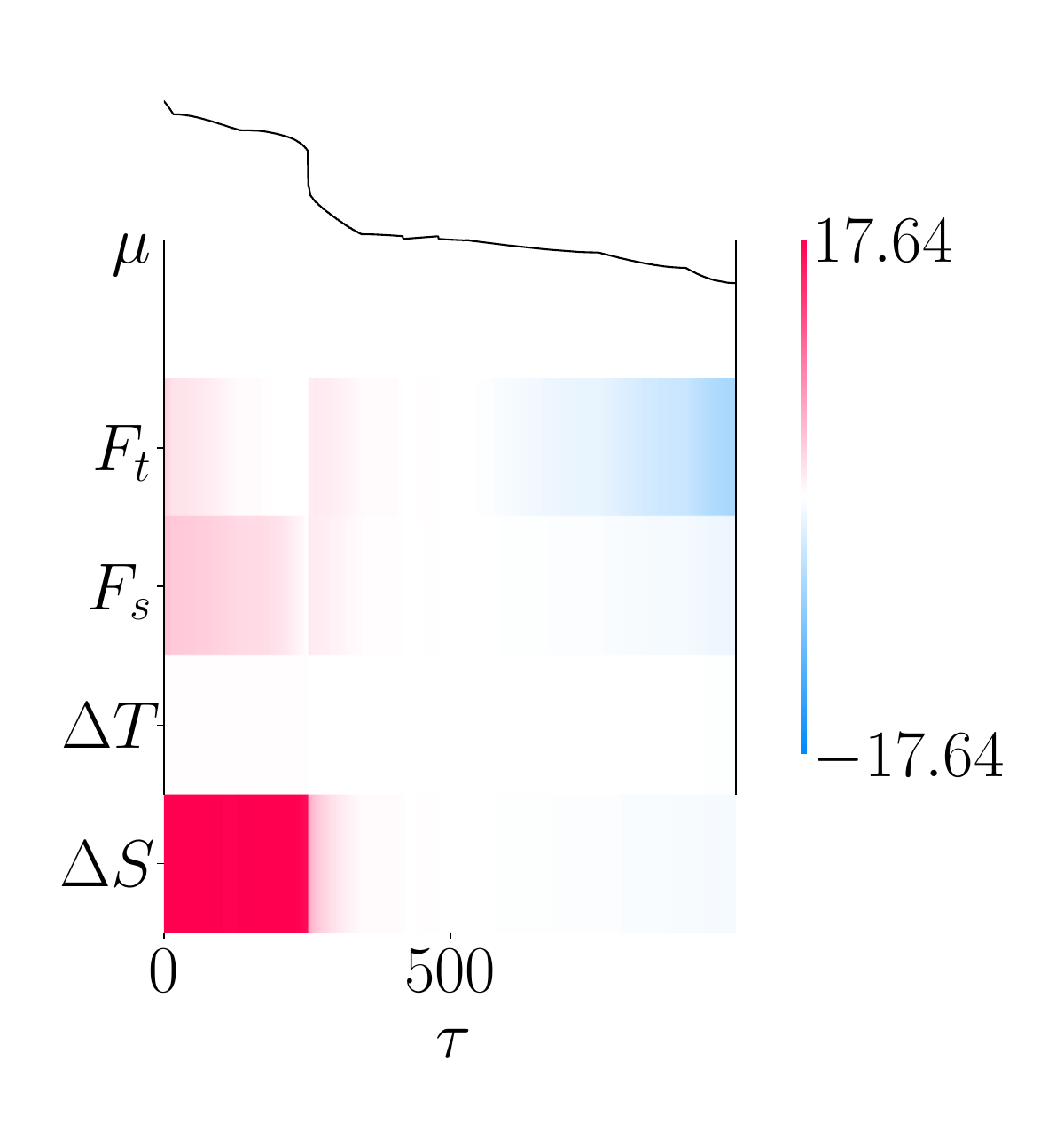}
	& \includegraphics[width=\linewidth,valign=m]{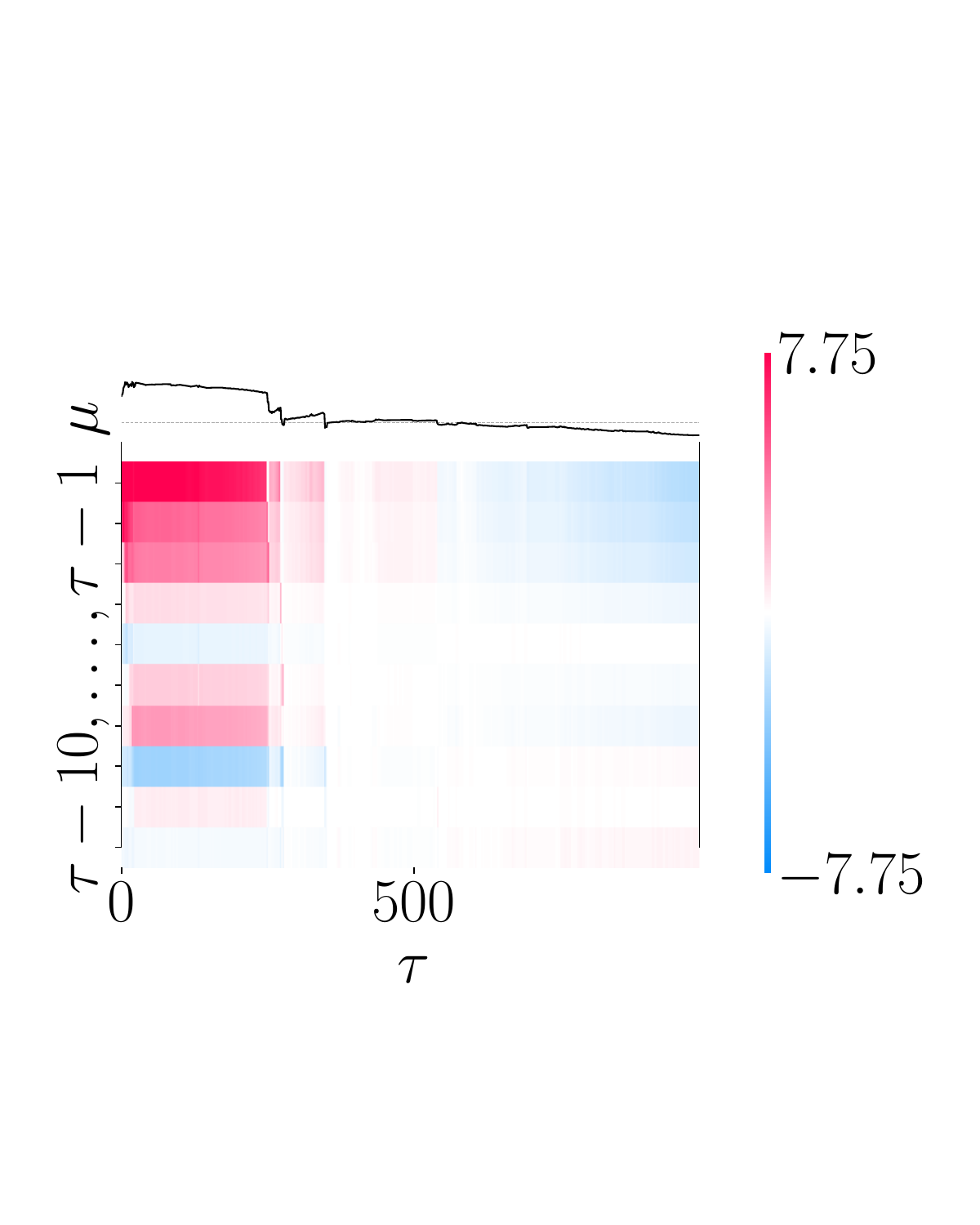}
	& \includegraphics[width=\linewidth,valign=m]{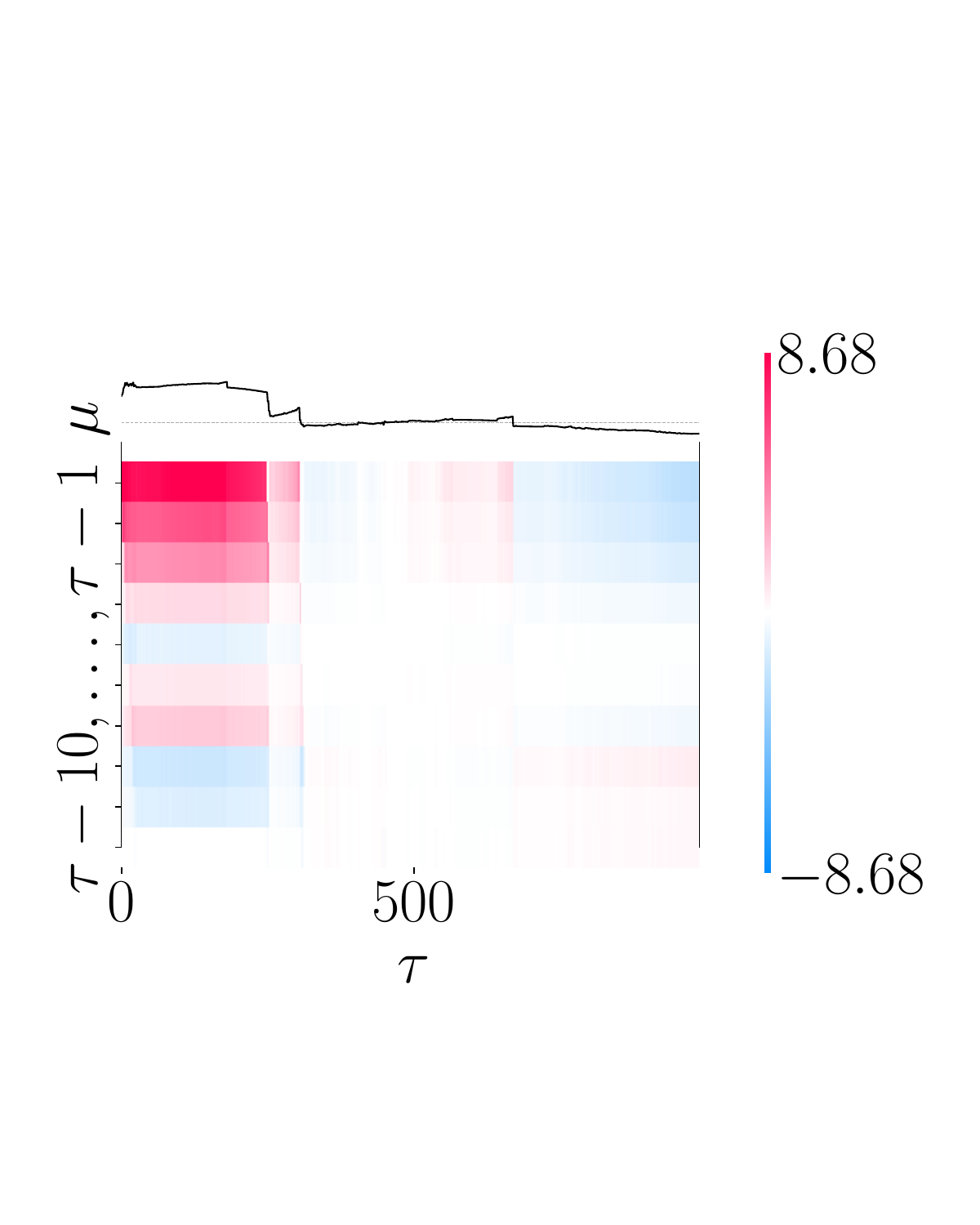} \\
	& MLP
	& \includegraphics[width=.7\linewidth,valign=m]{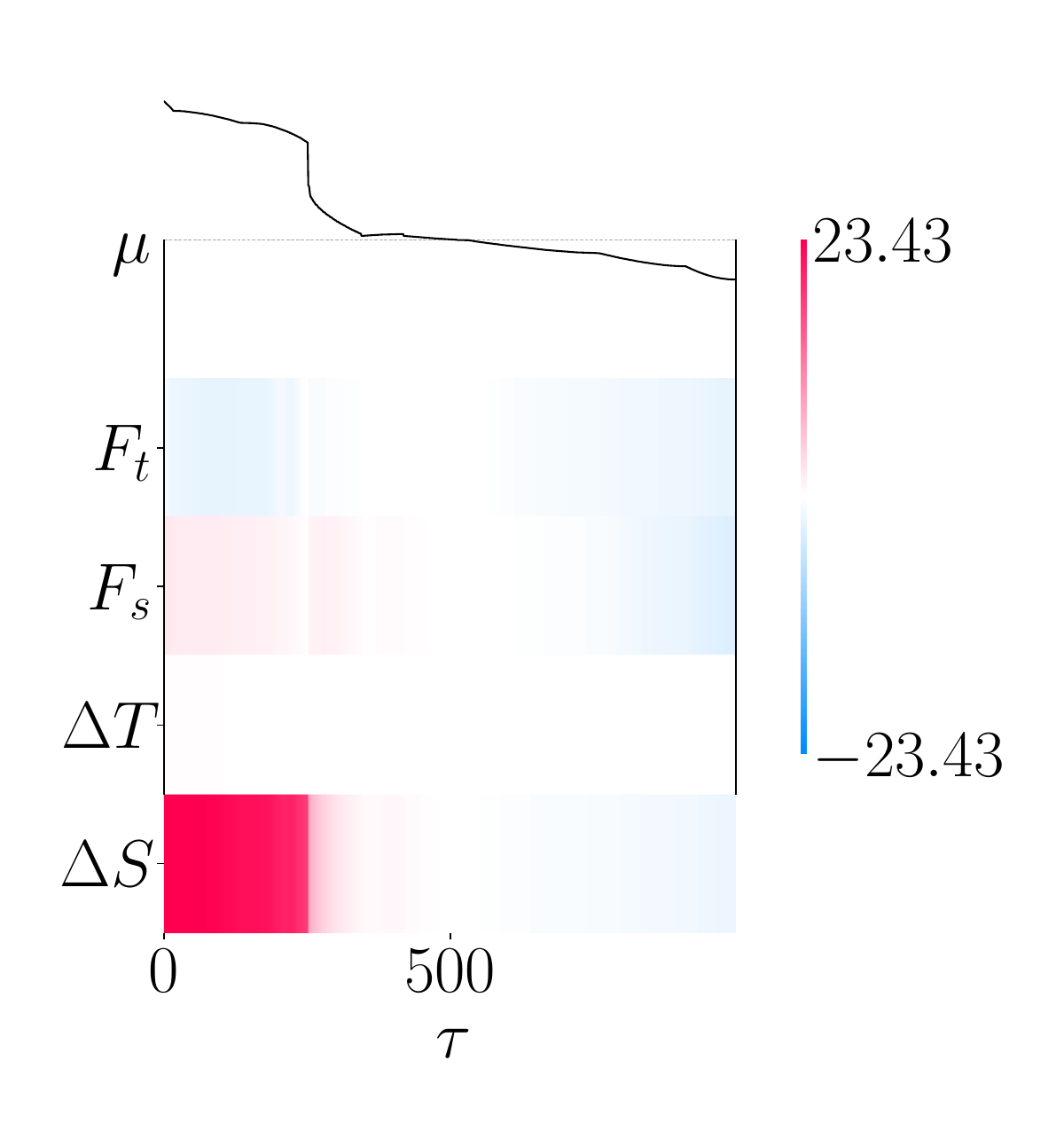}
	& \includegraphics[width=.7\linewidth,valign=m]{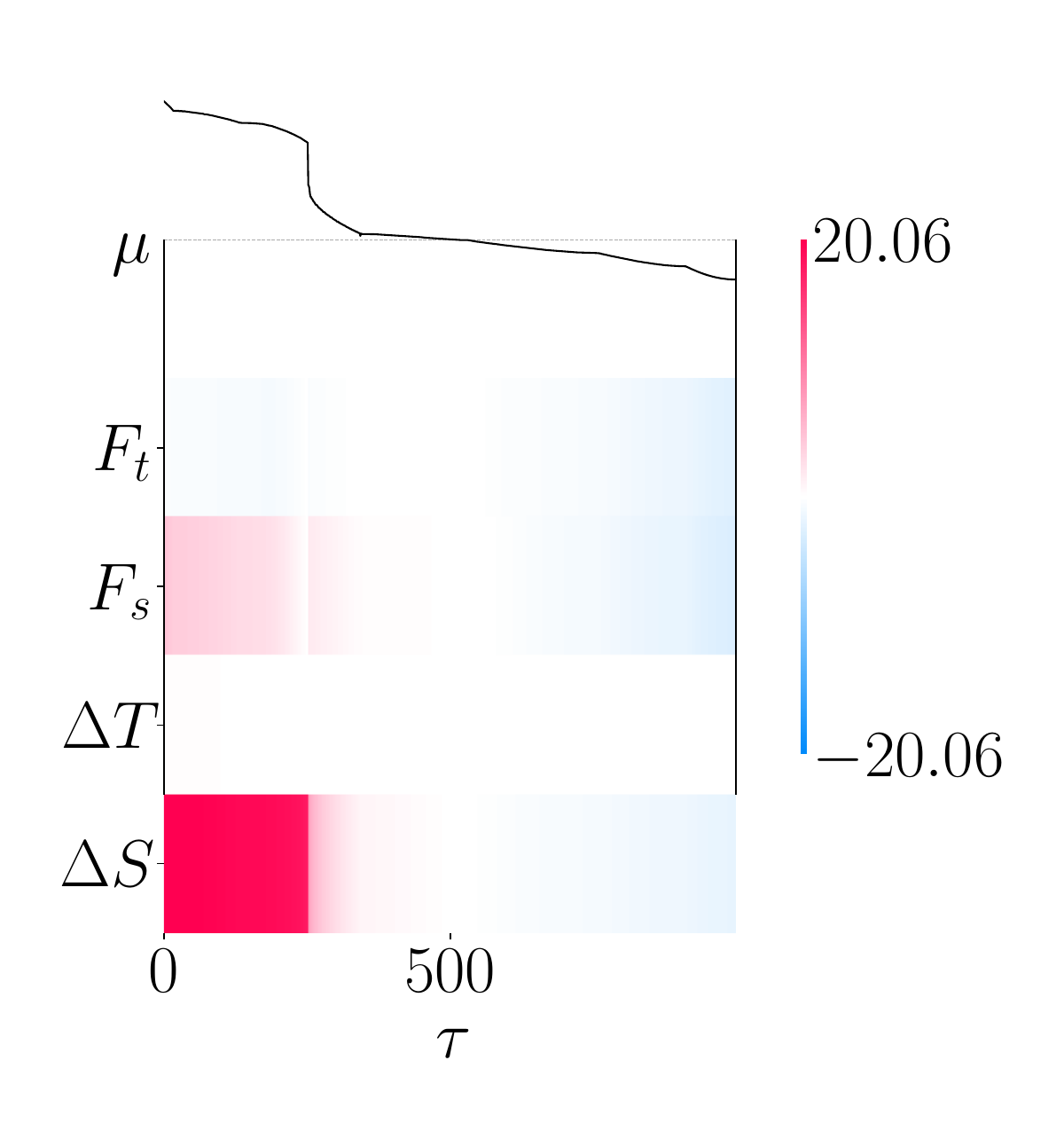}
	& \includegraphics[width=\linewidth,valign=m]{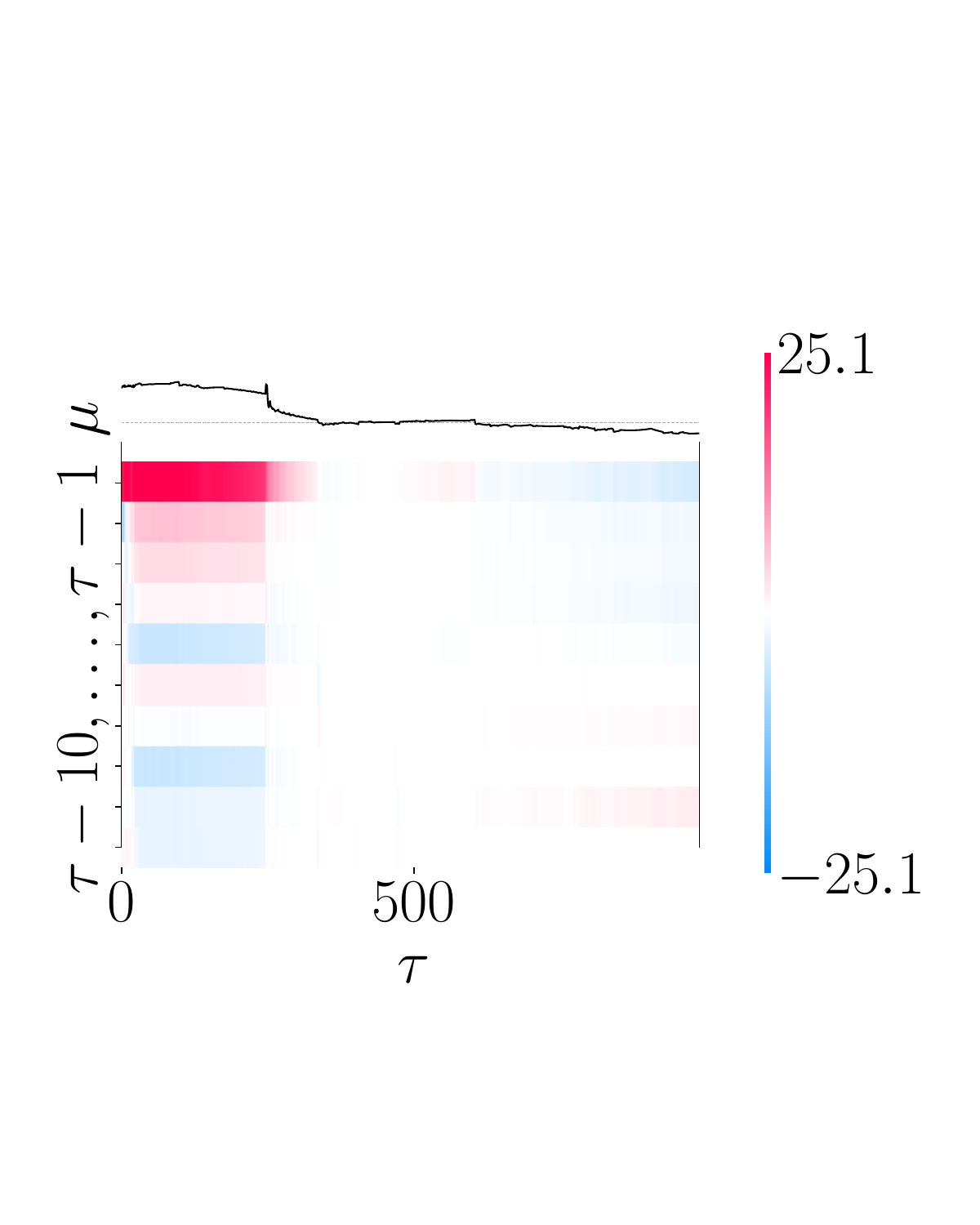}
	& \includegraphics[width=\linewidth,valign=m]{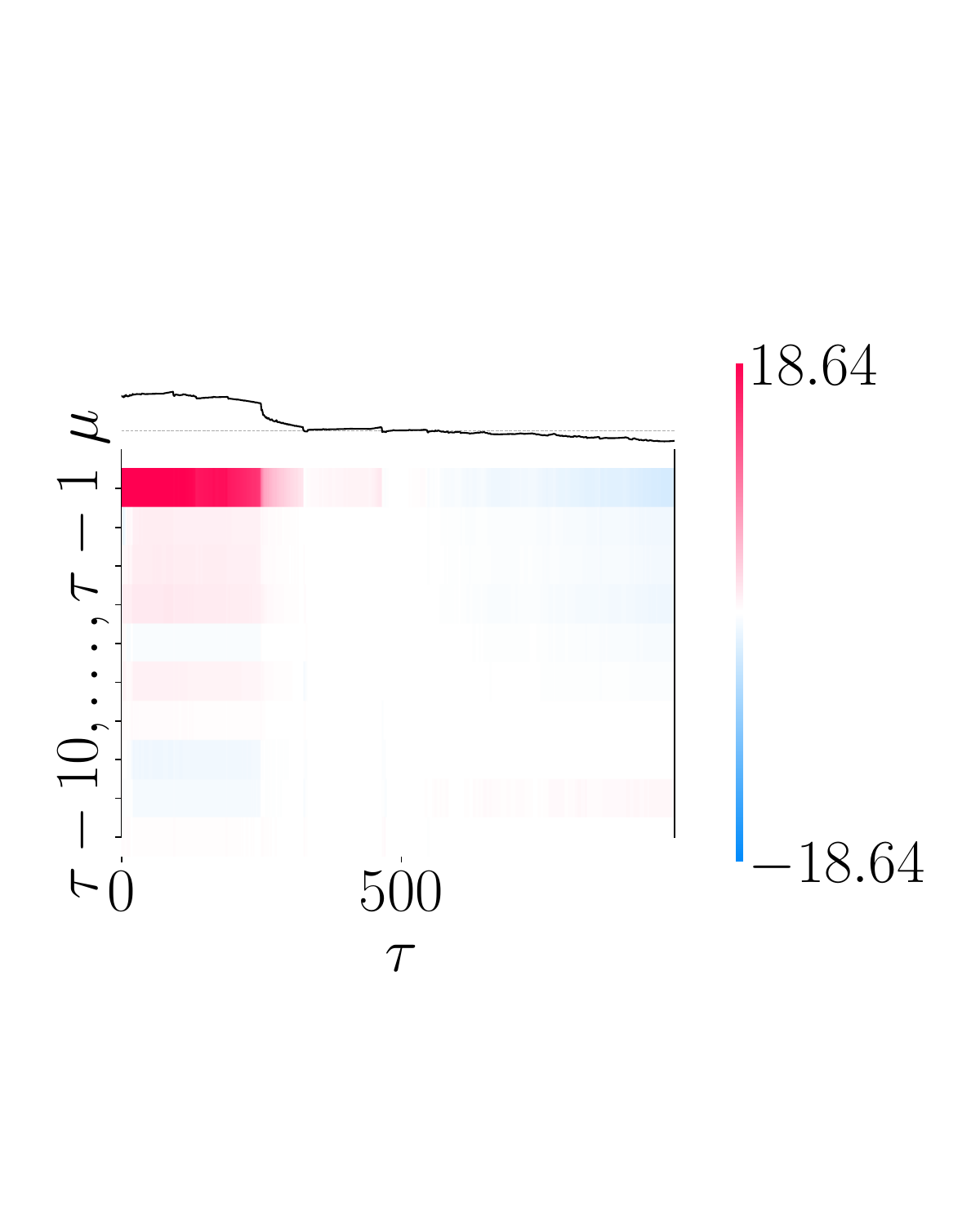} \\
	& Deep Ensemble
	& \includegraphics[width=.7\linewidth,valign=m]{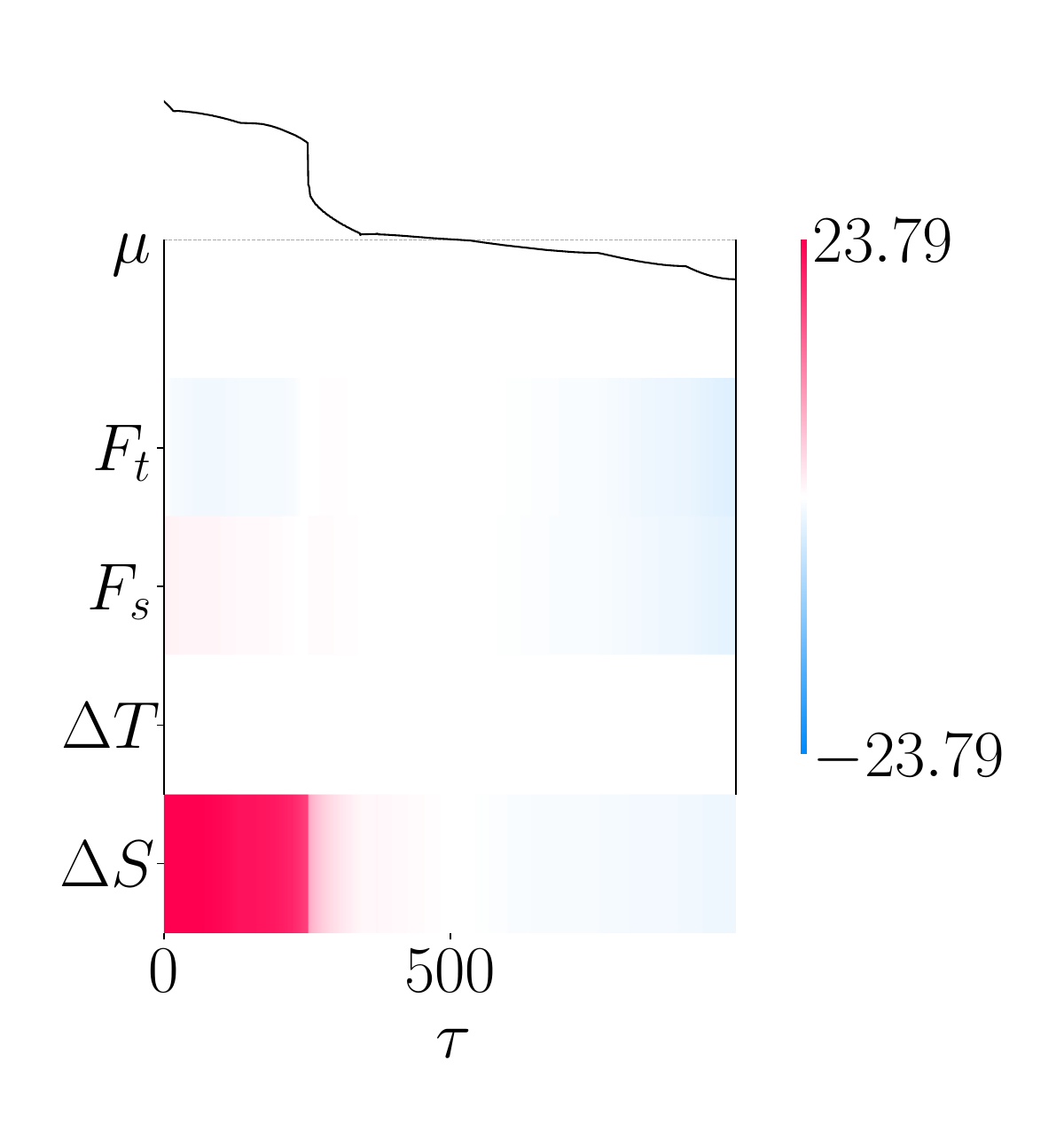}
	& \includegraphics[width=.7\linewidth,valign=m]{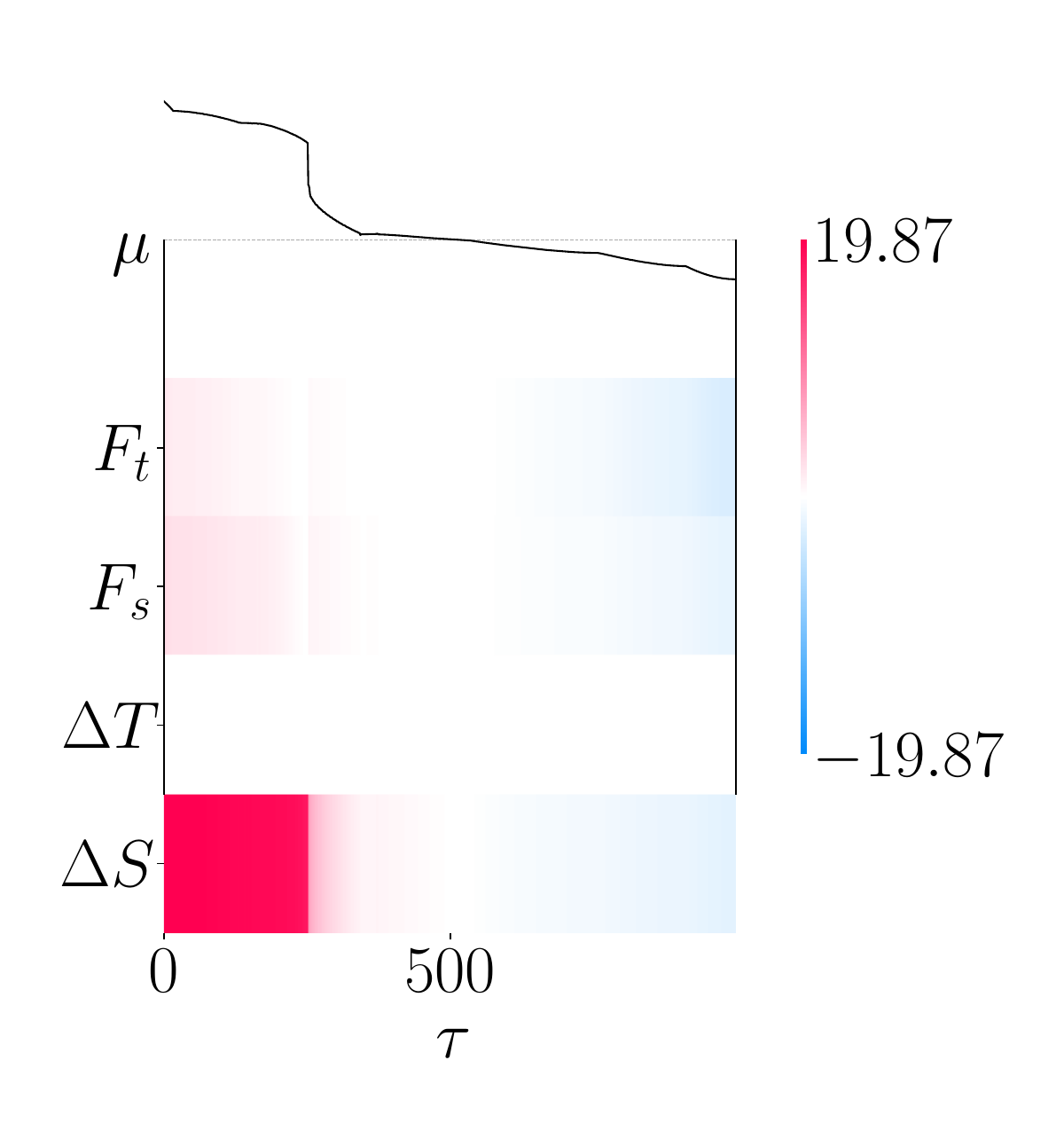}
	& \includegraphics[width=\linewidth,valign=m]{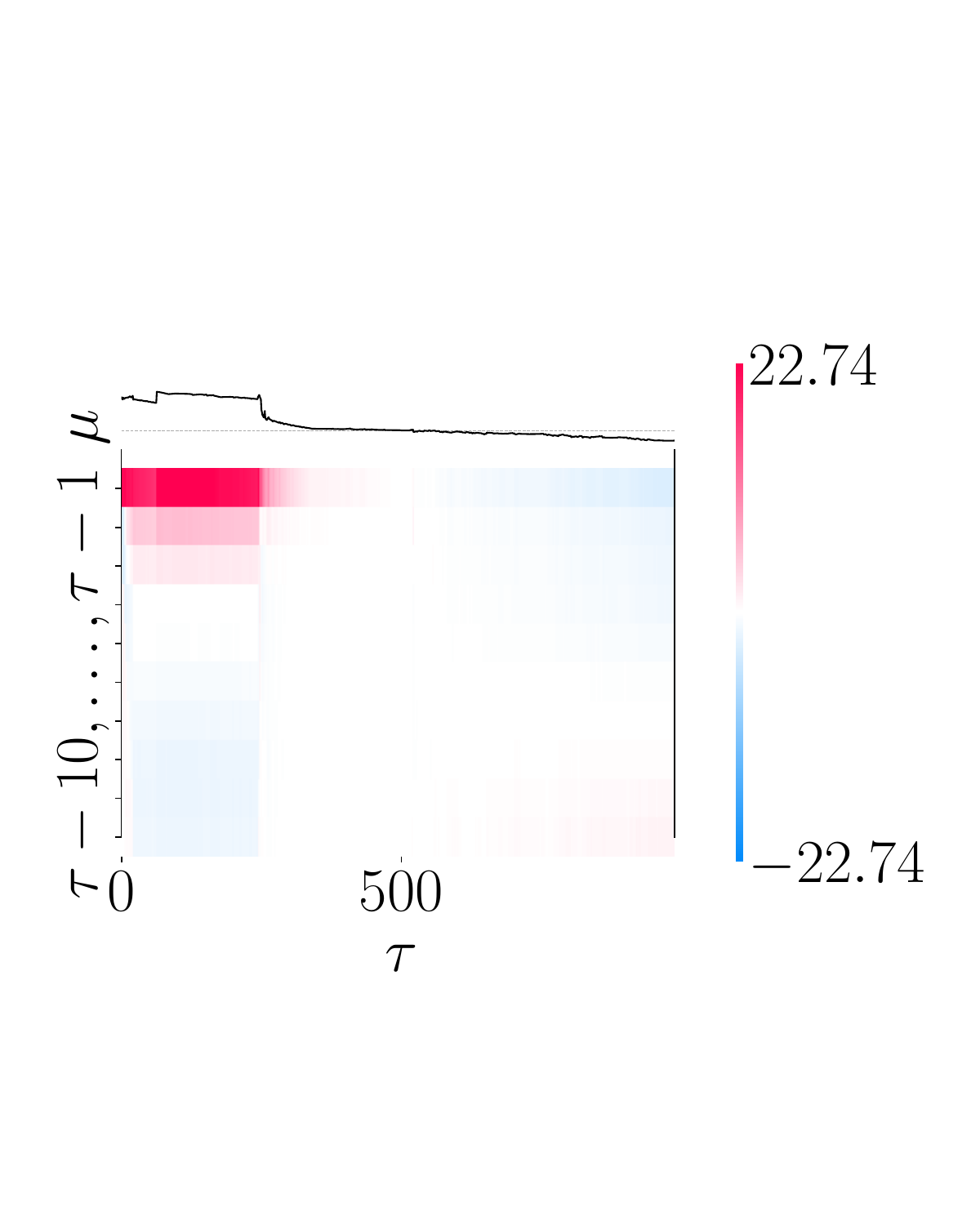}
	& \includegraphics[width=\linewidth,valign=m]{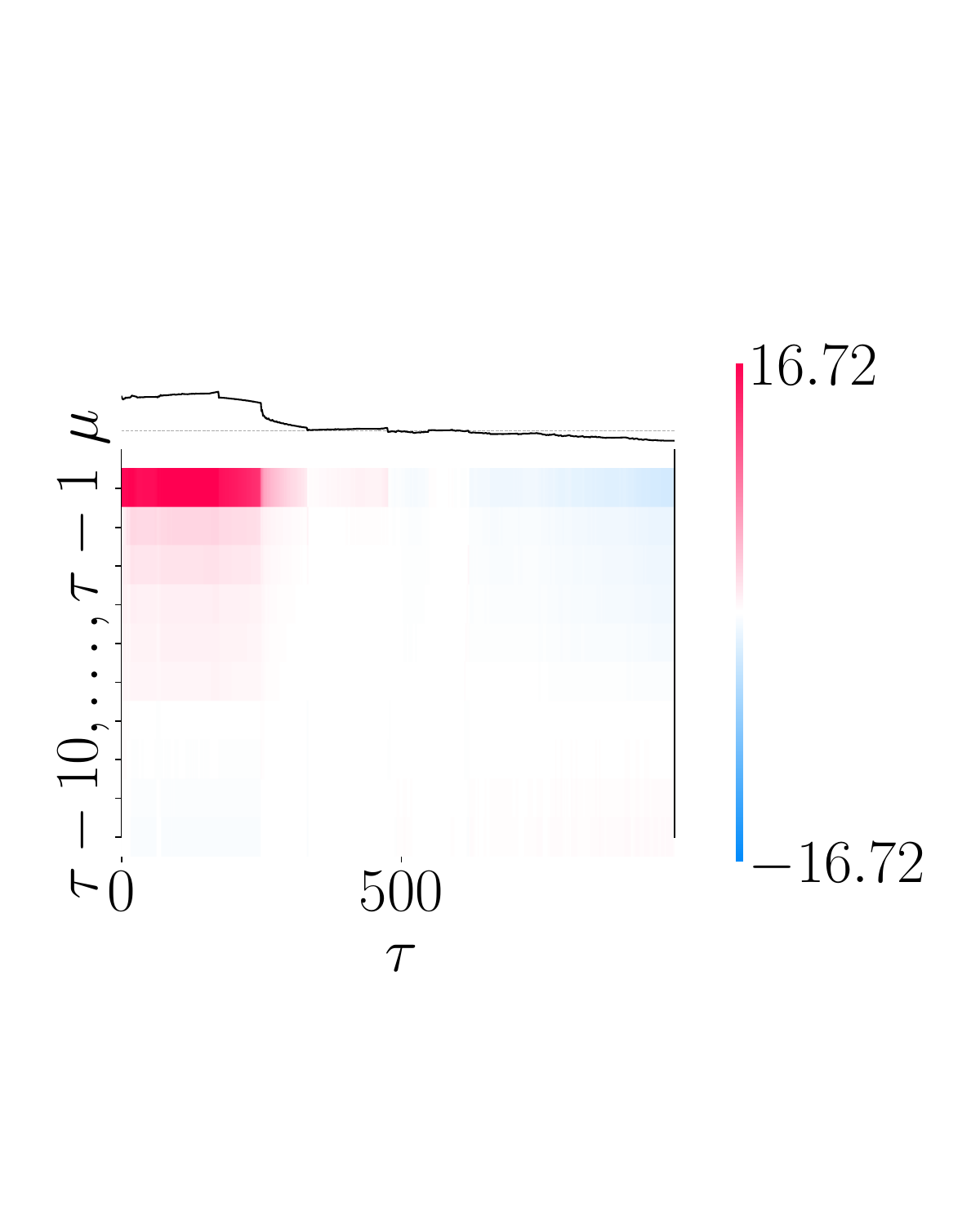} \\
	& RNN
	& ---
	& ---
	& \includegraphics[width=\linewidth,valign=m]{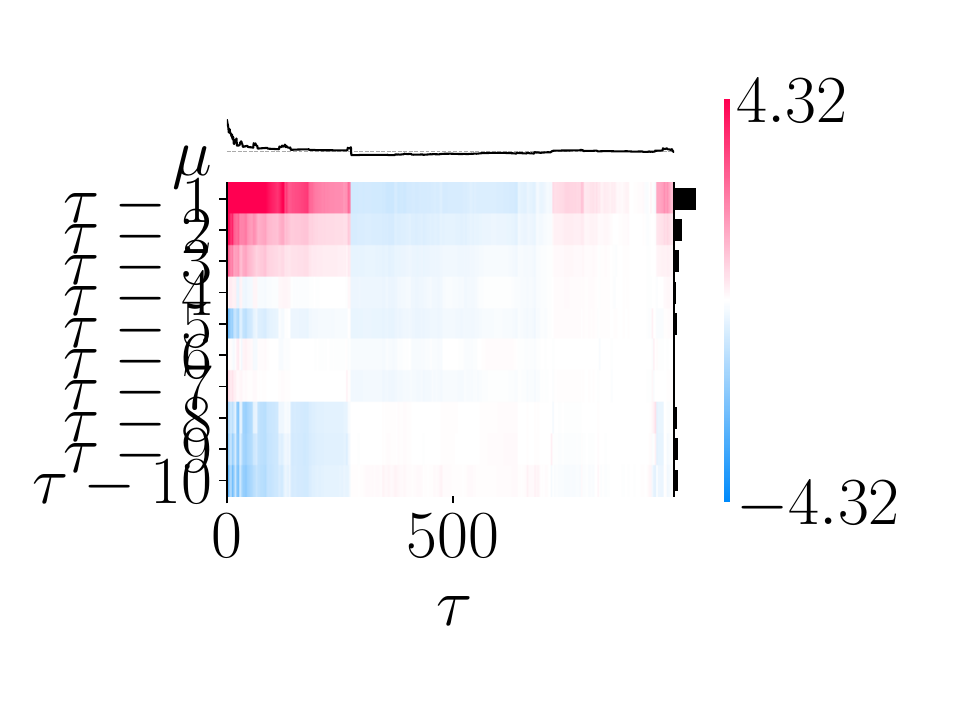}
	& \includegraphics[width=\linewidth,valign=m]{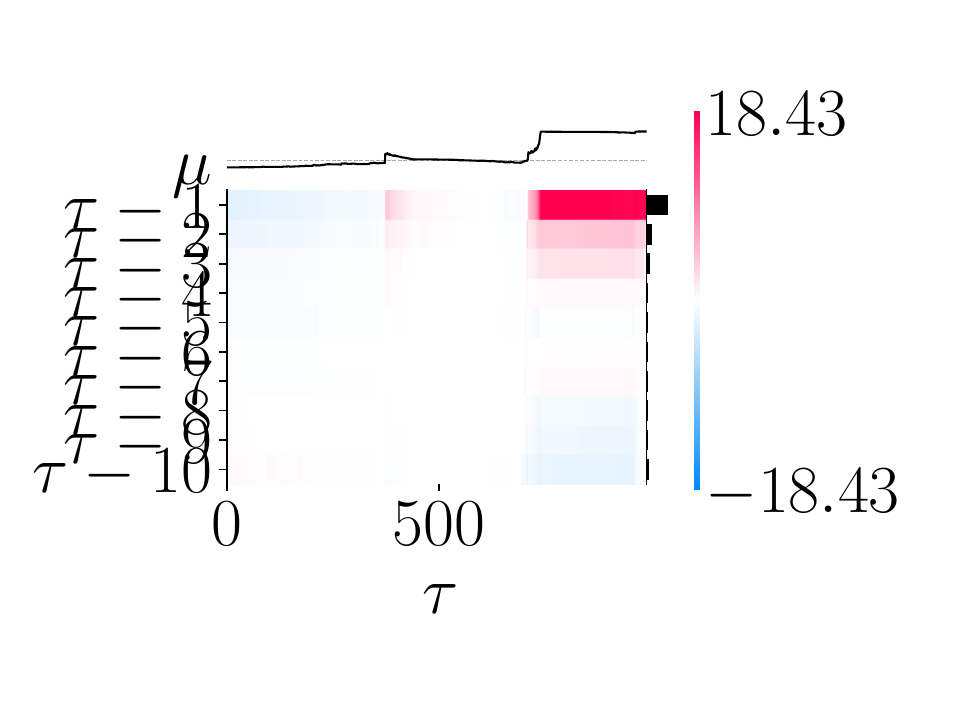} \\

	\bottomrule
	\caption{Attribution maps using \(F_s\), \(F_t\) and \(\rho^{\text{lin}}\).}%
	\label{tab:attributions-extended-linear}%
\end{longtable}

\section{\(\rho^{\text{EOS-80}}\)}

In this Appendix, we include supplementary forcing scenarios using \(\rho^{\text{EOS-80}}\) that
complement \(\{\mathcal{F}_1, \dots, \mathcal{F}_6\}\) from the main experiments. We include a
Recurrent Neural Network (RNN) as an additional architecture.

\subsection{Standard Box Model}

\subsubsection{Forcing Scenarios}

\begin{longtable}[c]{p{.25\columnwidth}C{.25\columnwidth}C{.25\columnwidth}C{.25\columnwidth}}
			\toprule
			Scenario & Forcing & Variables & \(q\) \\
			\midrule
			\(F_s\): Linear
			& \includegraphics[width=0.8\linewidth,valign=m]{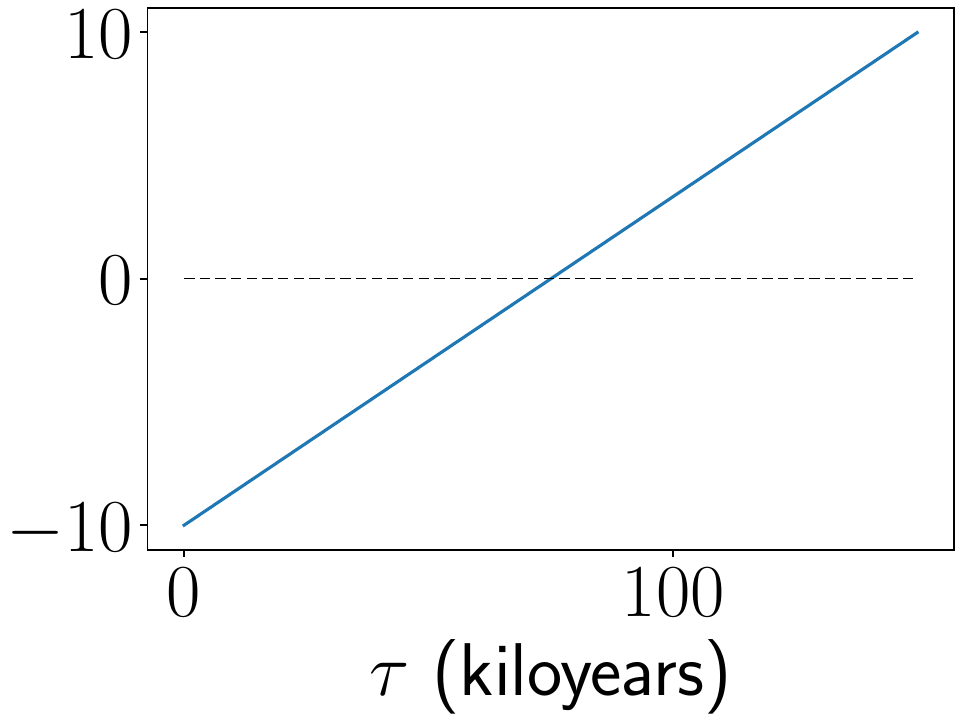}
			& \includegraphics[width=0.8\linewidth,valign=m]{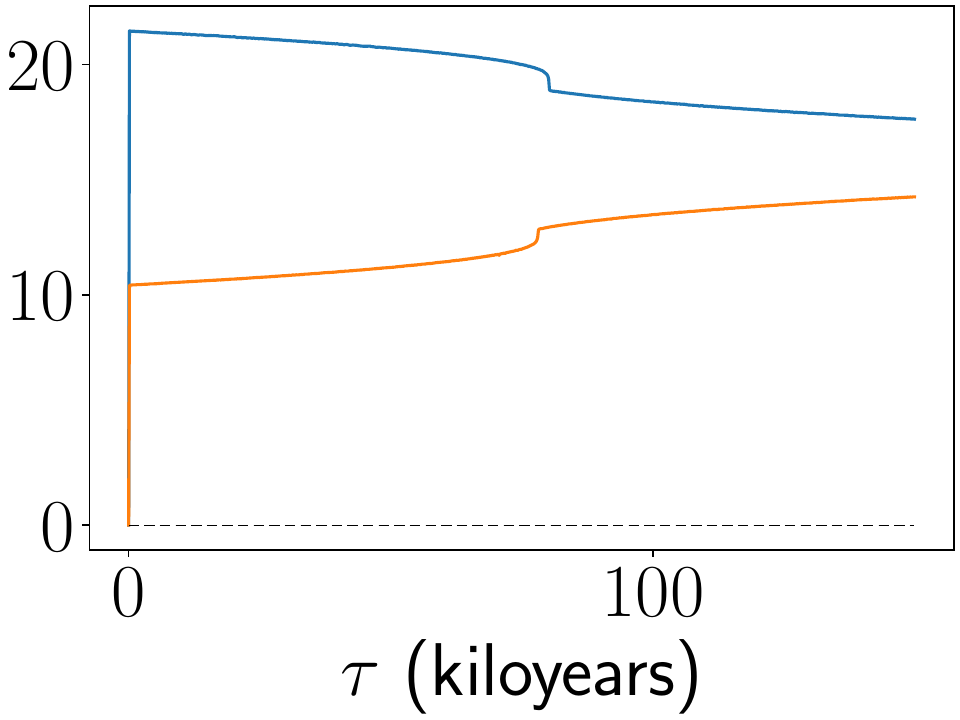}
			& \includegraphics[width=0.8\linewidth,valign=m]{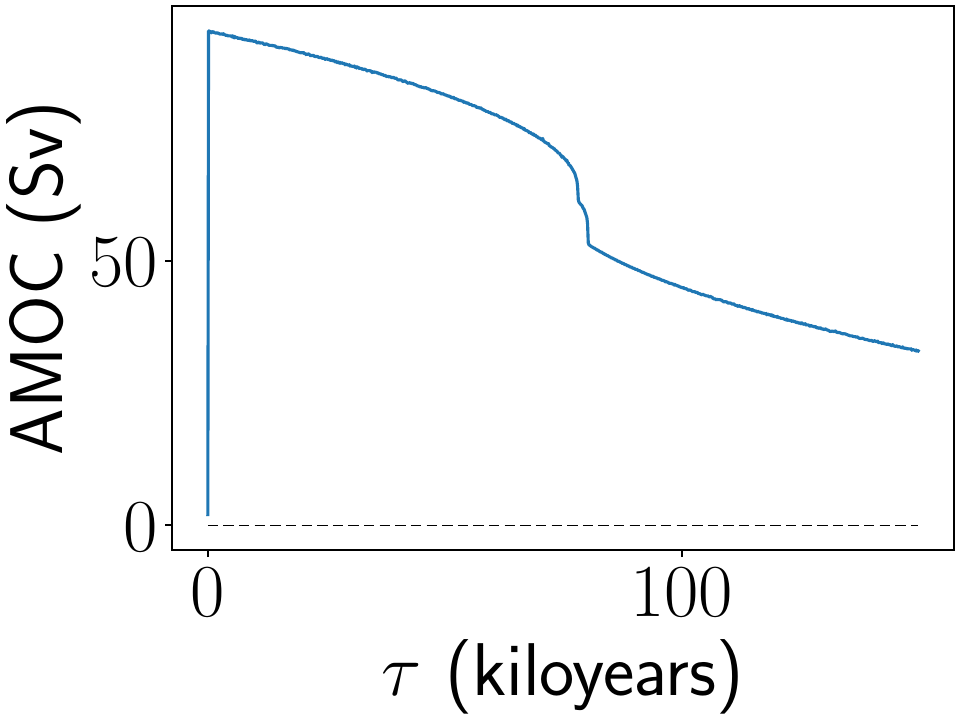} \\
			\(F_s\): Sinusoidal (stationary)
			& \includegraphics[width=0.8\linewidth,valign=m]{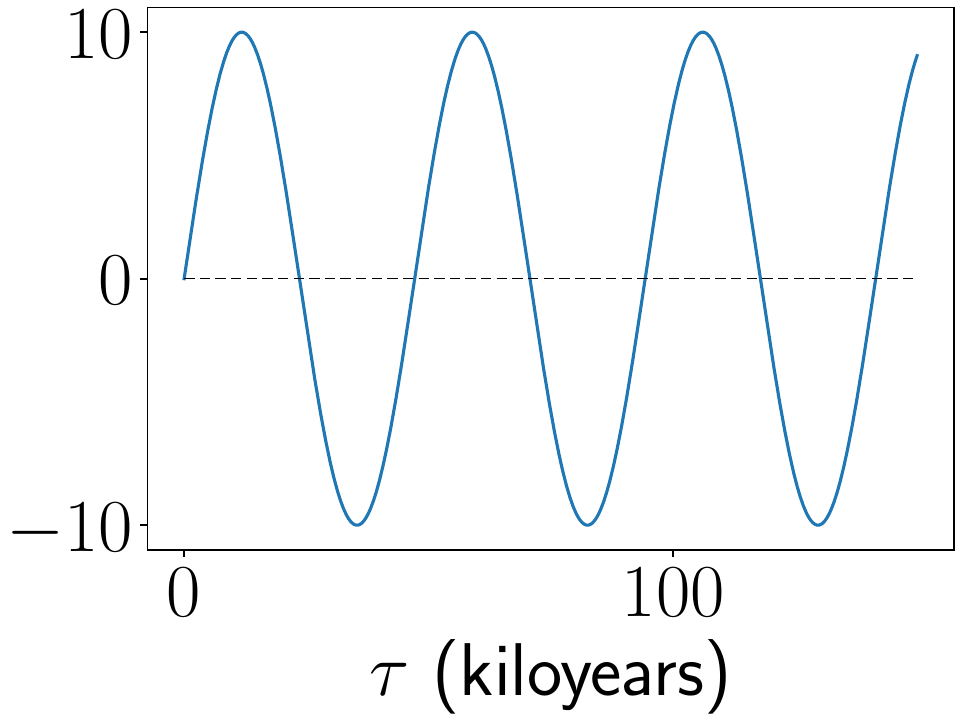}
			& \includegraphics[width=0.8\linewidth,valign=m]{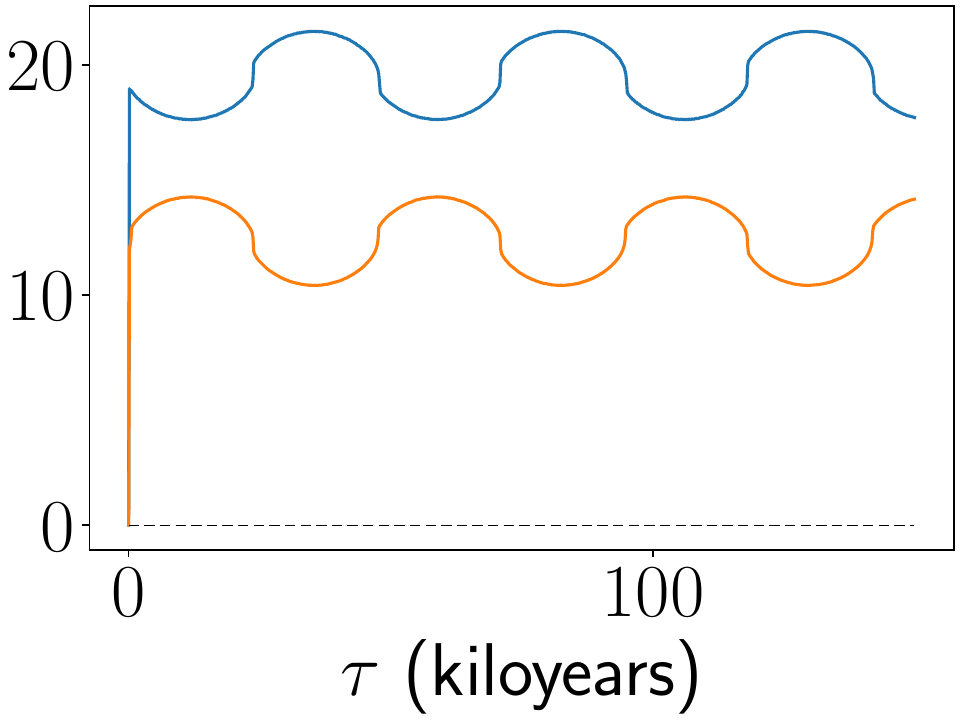}
			& \includegraphics[width=0.8\linewidth,valign=m]{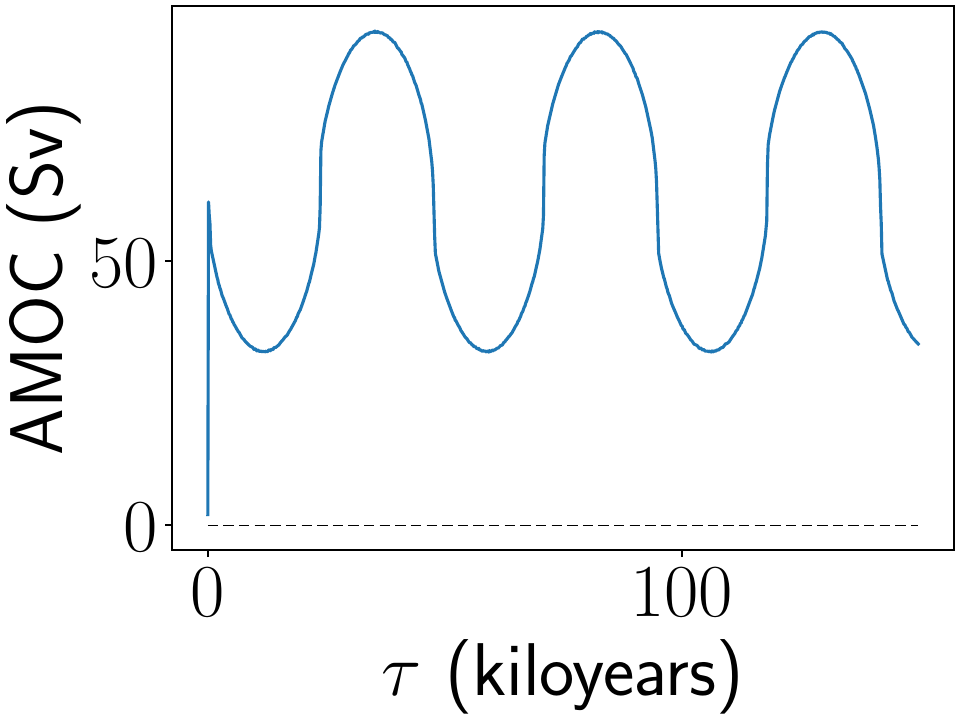} \\
			\(F_s\): Sinusoidal (nonstationary)
			& \includegraphics[width=0.8\linewidth,valign=m]{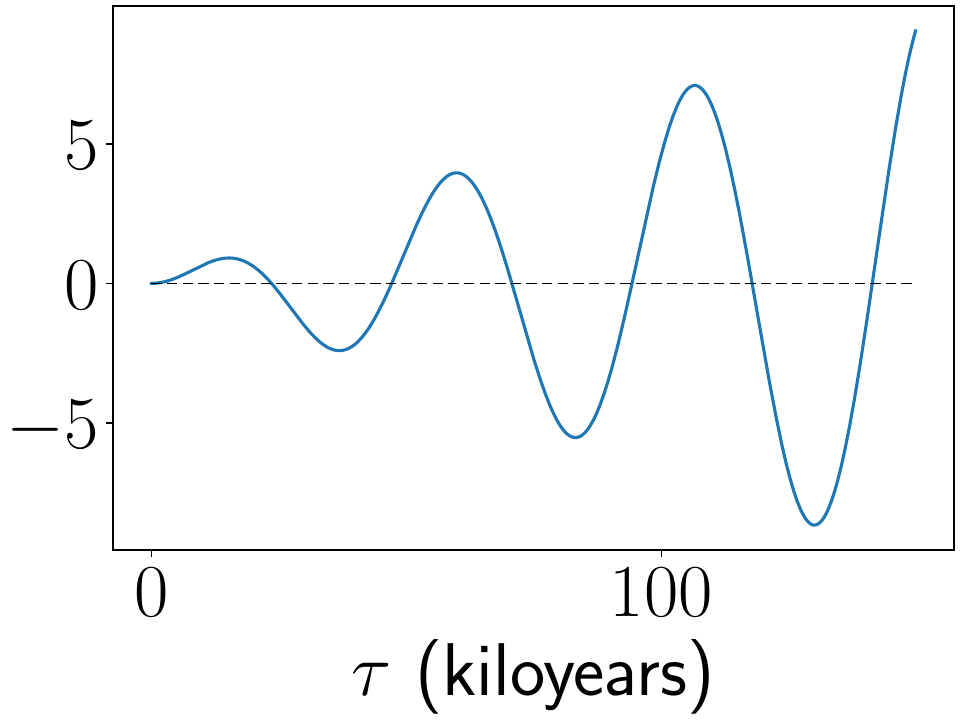}
			& \includegraphics[width=0.8\linewidth,valign=m]{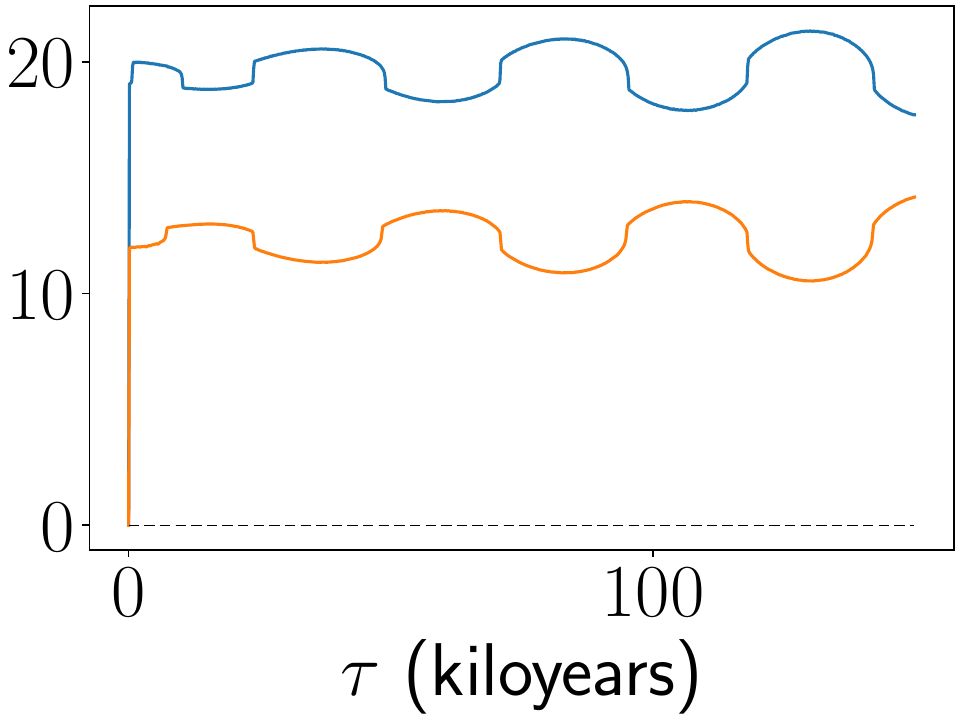}
			& \includegraphics[width=0.8\linewidth,valign=m]{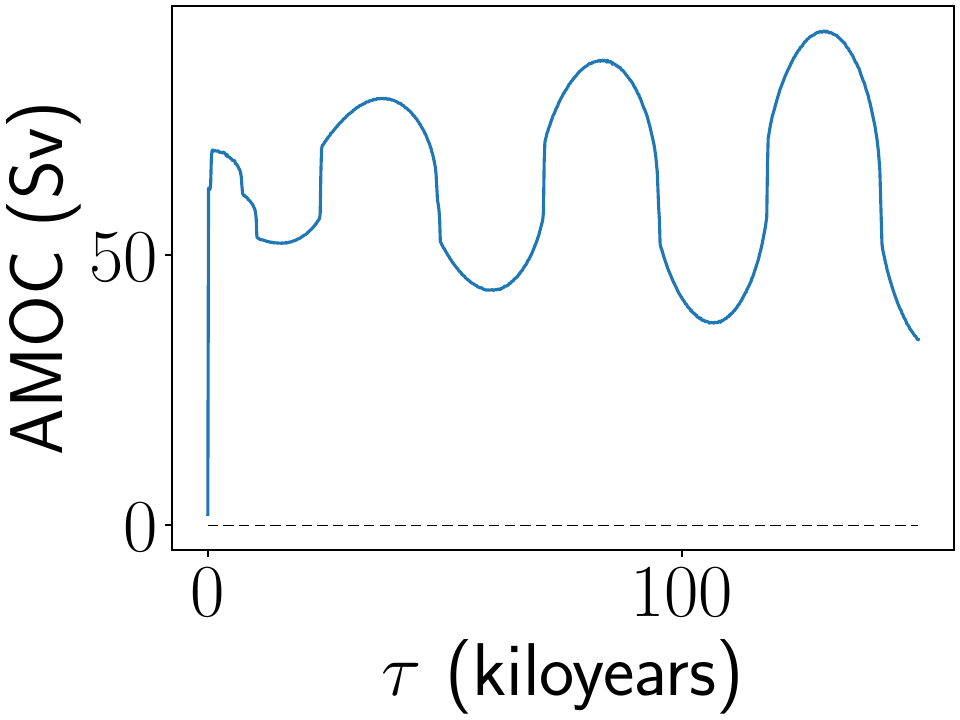} \\
			\bottomrule
	\caption{Overview of the scenarios and their variables using only fresh
	water forcing.}%
	\label{tab:scenarios-standard-nonlinear}%
\end{longtable}

\subsubsection{Performance}

\begin{longtable}[c]{p{.06\columnwidth}p{.1\columnwidth}C{.25\columnwidth}C{.25\columnwidth}}
	\toprule
	Scenario & Architecture & Prediction: PI & Prediction: AR \\
	\midrule

	\multirow[c]{3}{*}{\rotatebox[origin=c]{90}{\(F_s\): Linear}}
	& BNN
	& \includegraphics[width=\linewidth,valign=m]{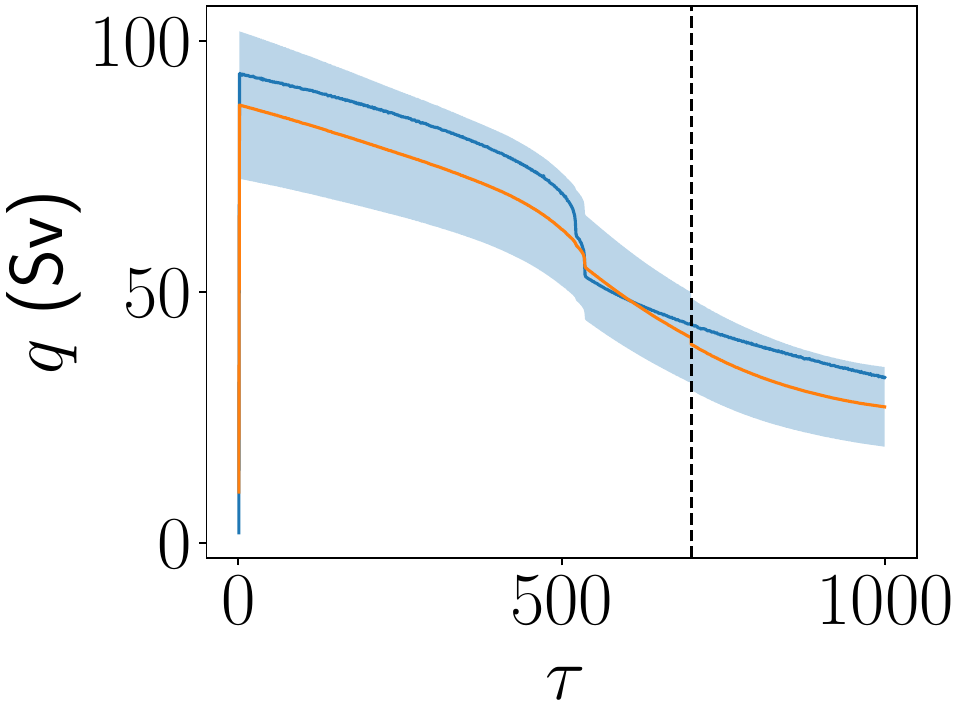}
	& \includegraphics[width=\linewidth,valign=m]{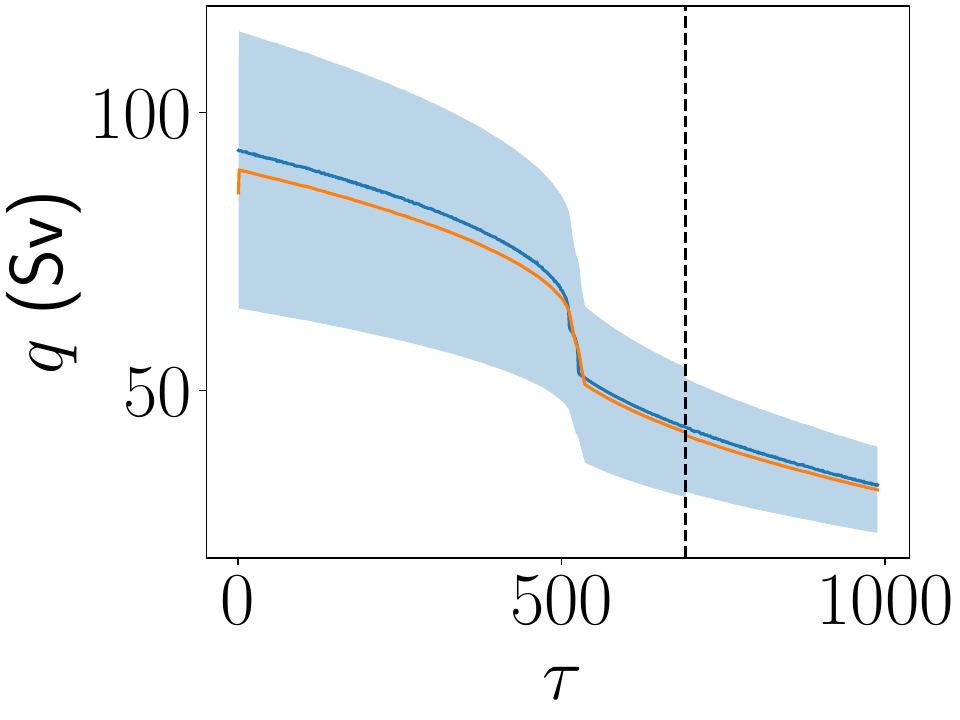} \\
	& MLP
	& \includegraphics[width=\linewidth,valign=m]{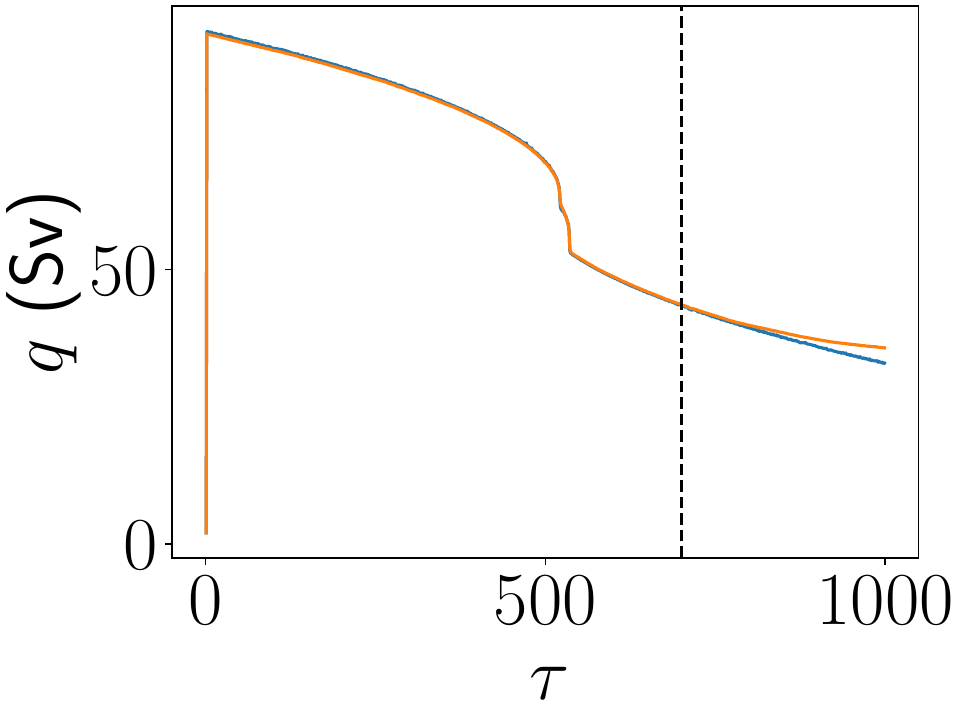}
	& \includegraphics[width=\linewidth,valign=m]{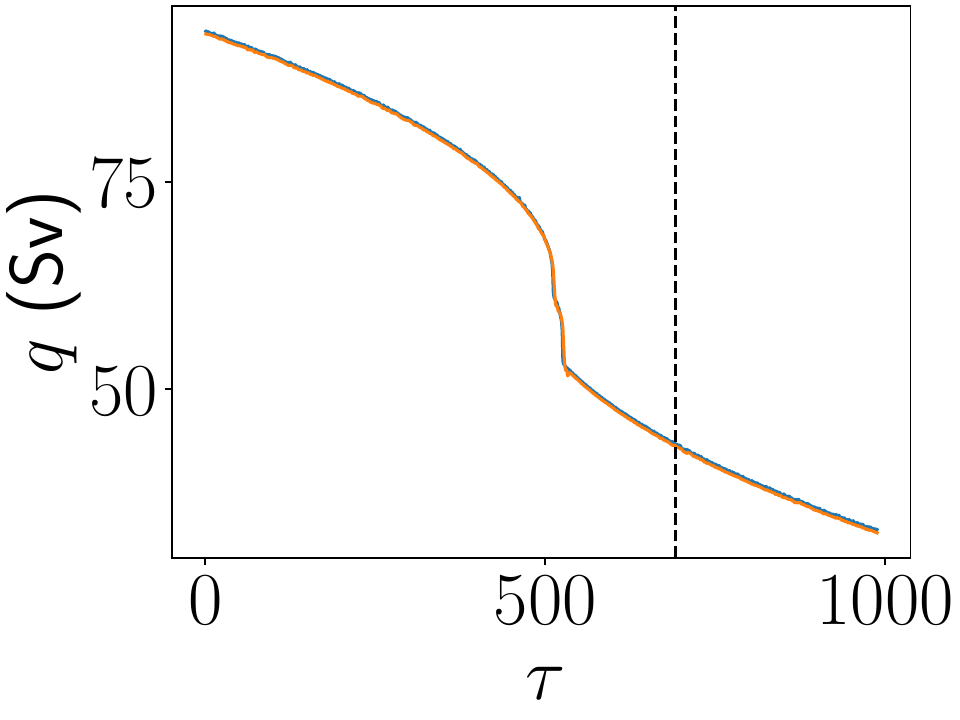} \\
	& Deep Ensemble
	& \includegraphics[width=\linewidth,valign=m]{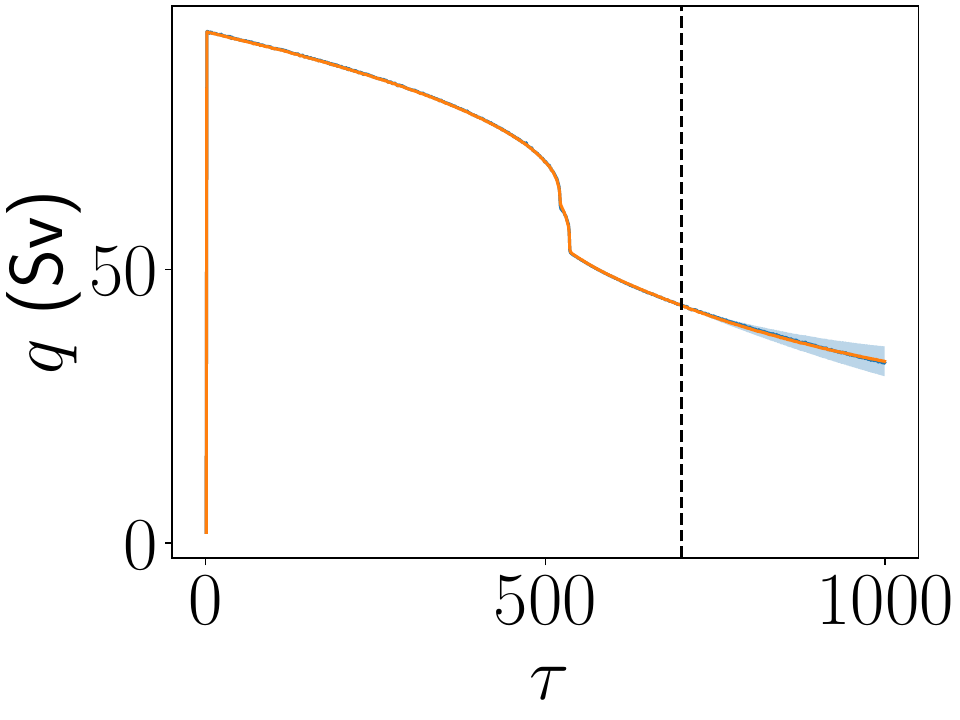}
	& \includegraphics[width=\linewidth,valign=m]{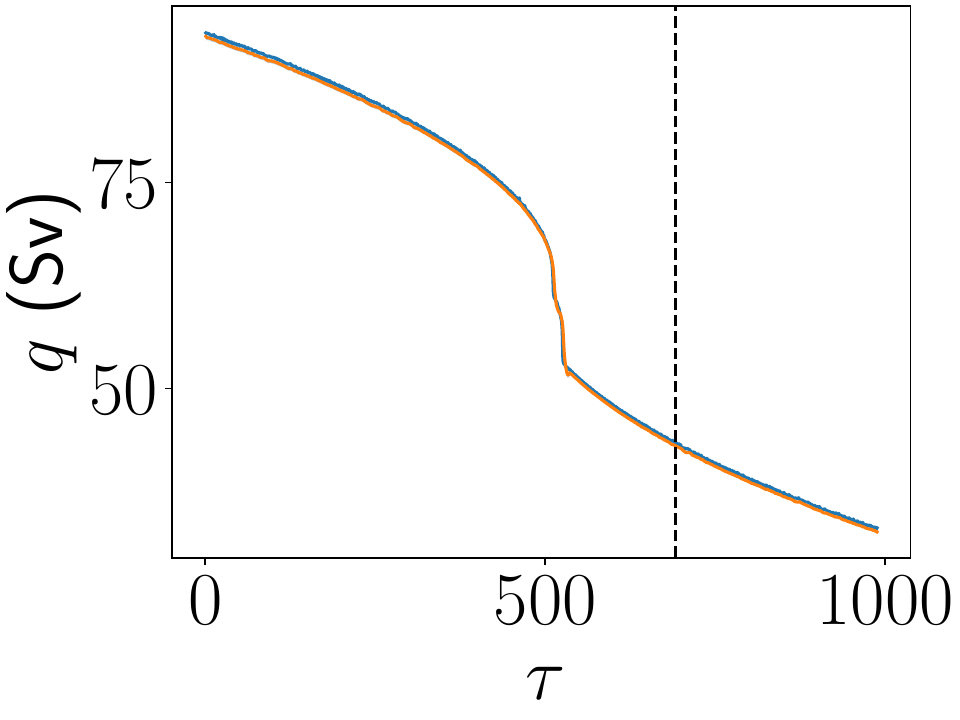} \\
	& RNN
	& ---
	& \includegraphics[width=\linewidth,valign=m]{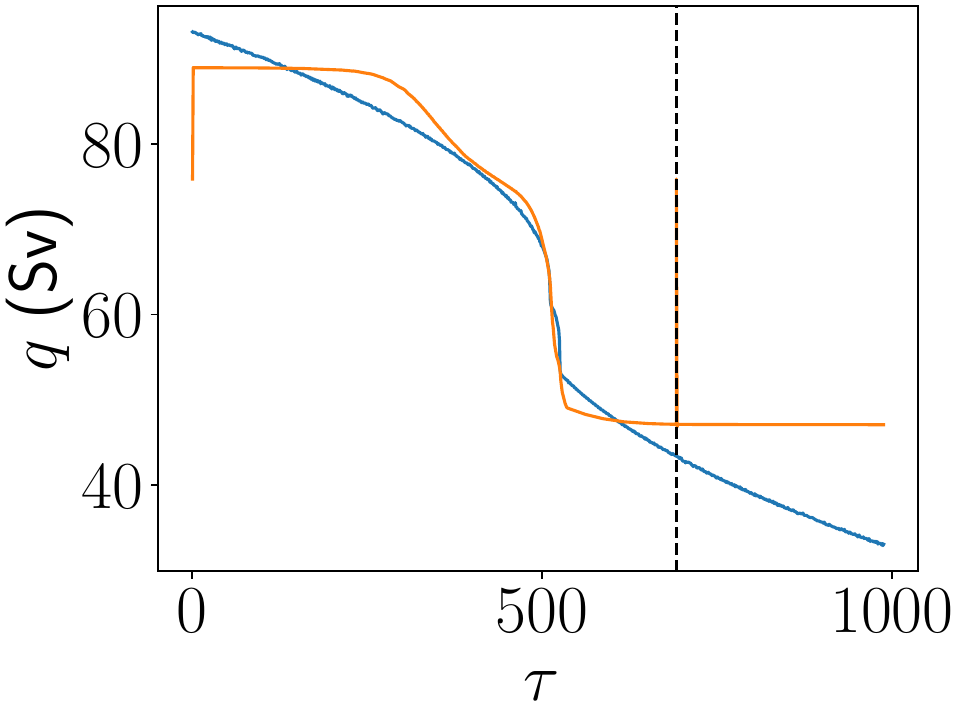} \\

	\midrule

	\multirow[c]{3}{*}{\rotatebox[origin=c]{90}{\(F_s\): Sinusoidal (stationary)}}
	& BNN
	& \includegraphics[width=\linewidth,valign=m]{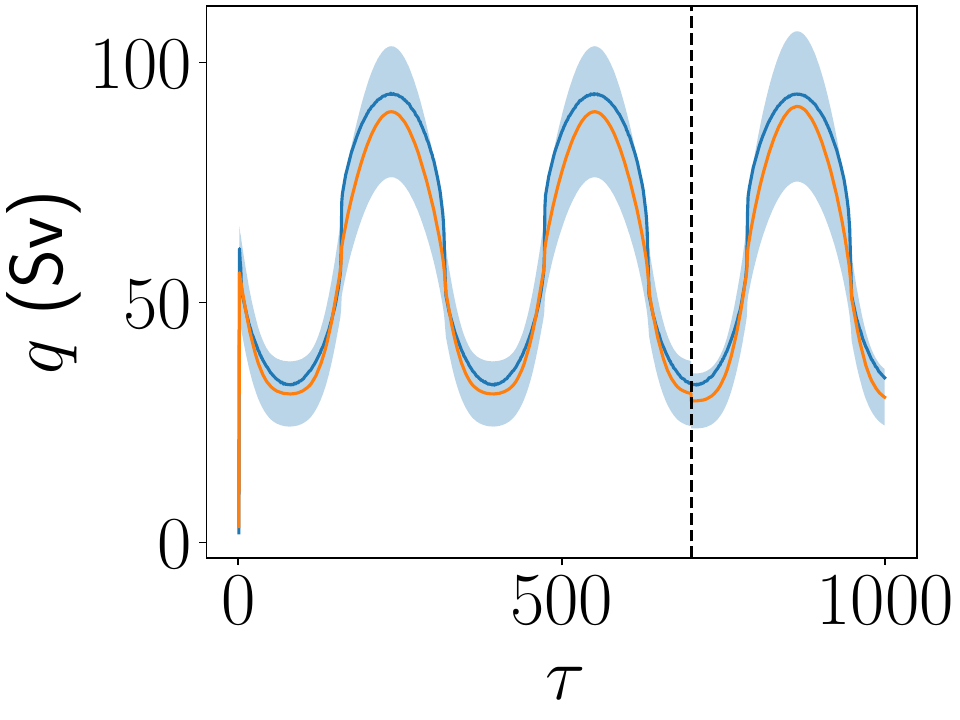}
	& \includegraphics[width=\linewidth,valign=m]{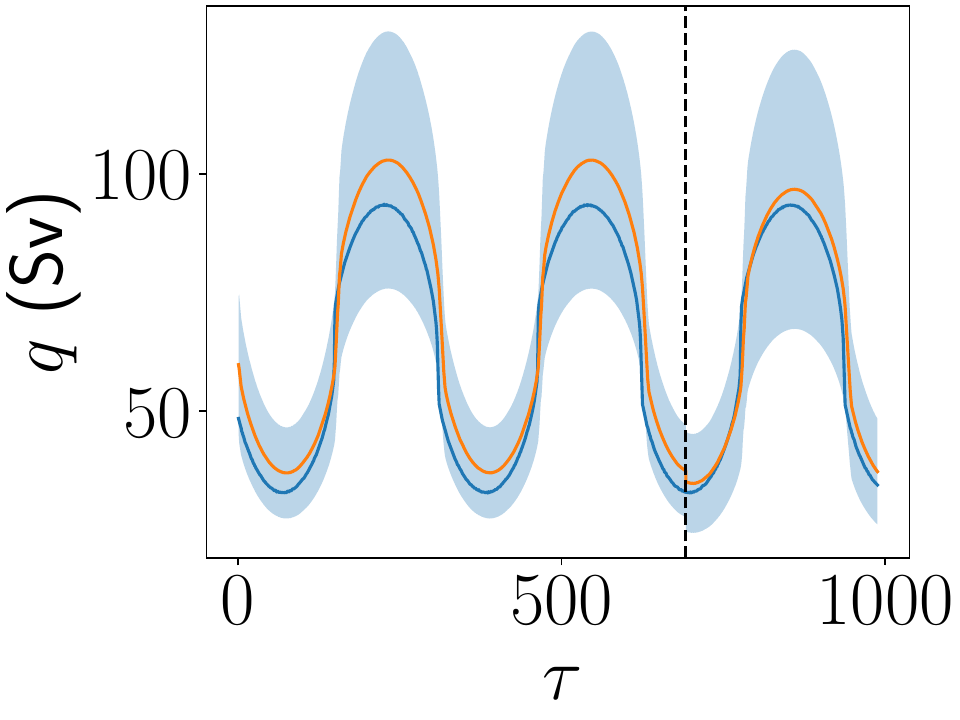} \\
	& MLP
	& \includegraphics[width=\linewidth,valign=m]{sinusoidal_stationary/linear_density/mlp/physics_informed/groundtruth-prediction.pdf}
	& \includegraphics[width=\linewidth,valign=m]{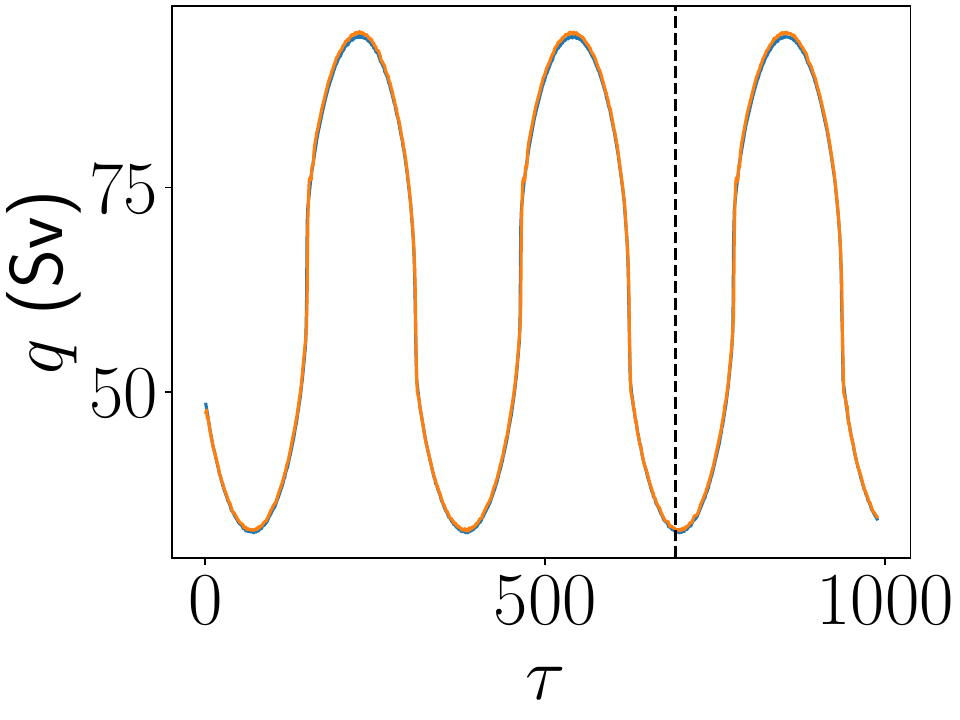} \\
	& Deep Ensemble
	& \includegraphics[width=\linewidth,valign=m]{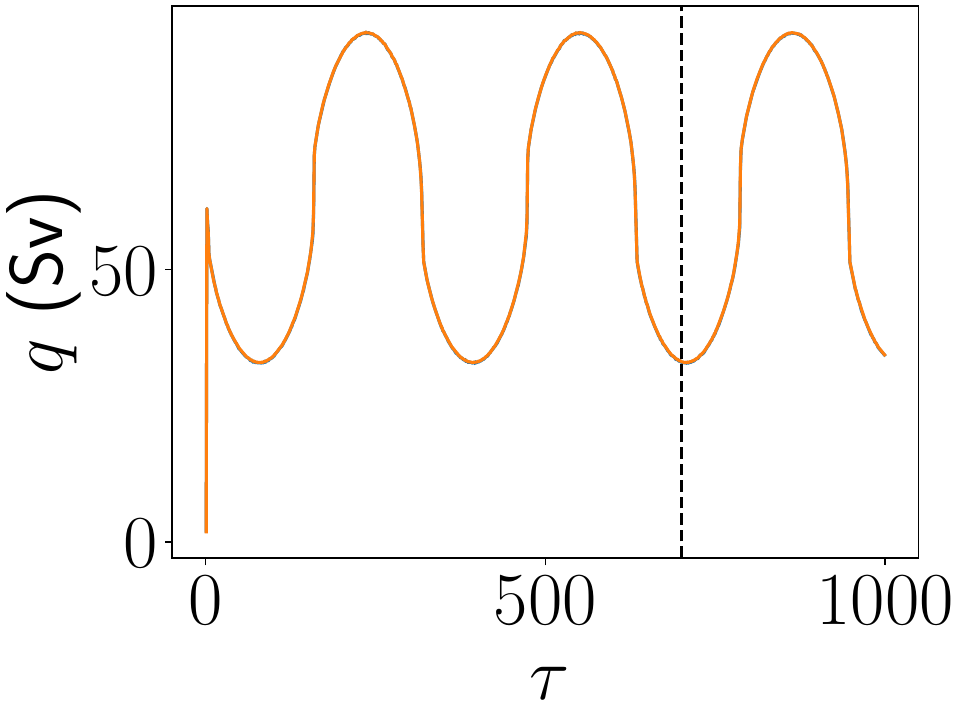}
	& \includegraphics[width=\linewidth,valign=m]{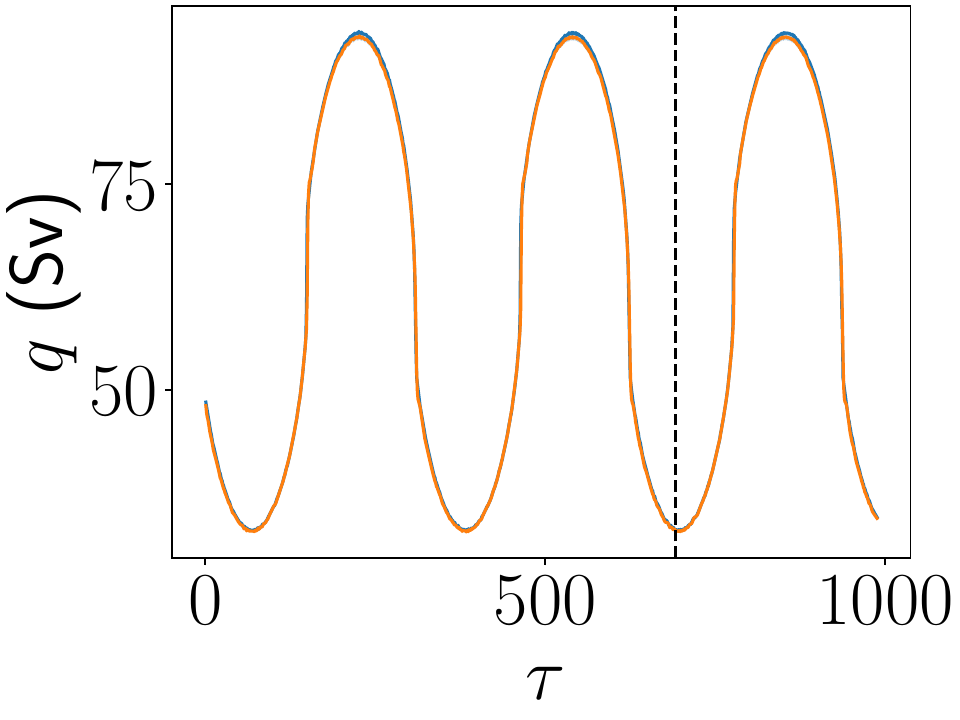} \\
	& RNN
	& ---
	& \includegraphics[width=\linewidth,valign=m]{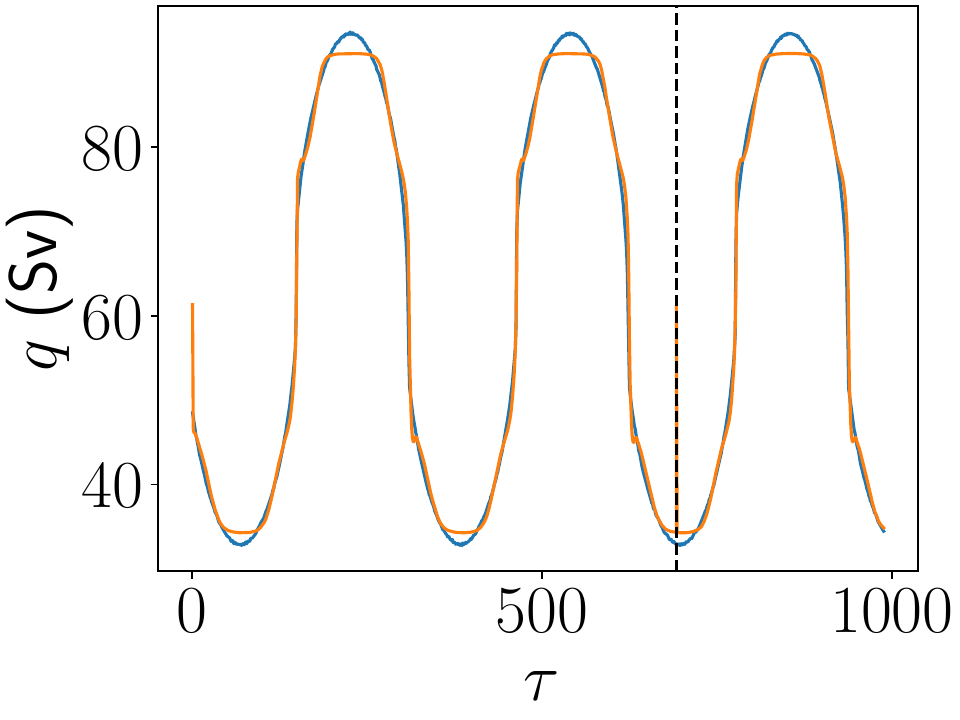} \\

	\midrule

	\multirow[c]{3}{*}{\rotatebox[origin=c]{90}{\(F_s\): Sinusoidal (nonstationary)}}
	& BNN
	& \includegraphics[width=\linewidth,valign=m]{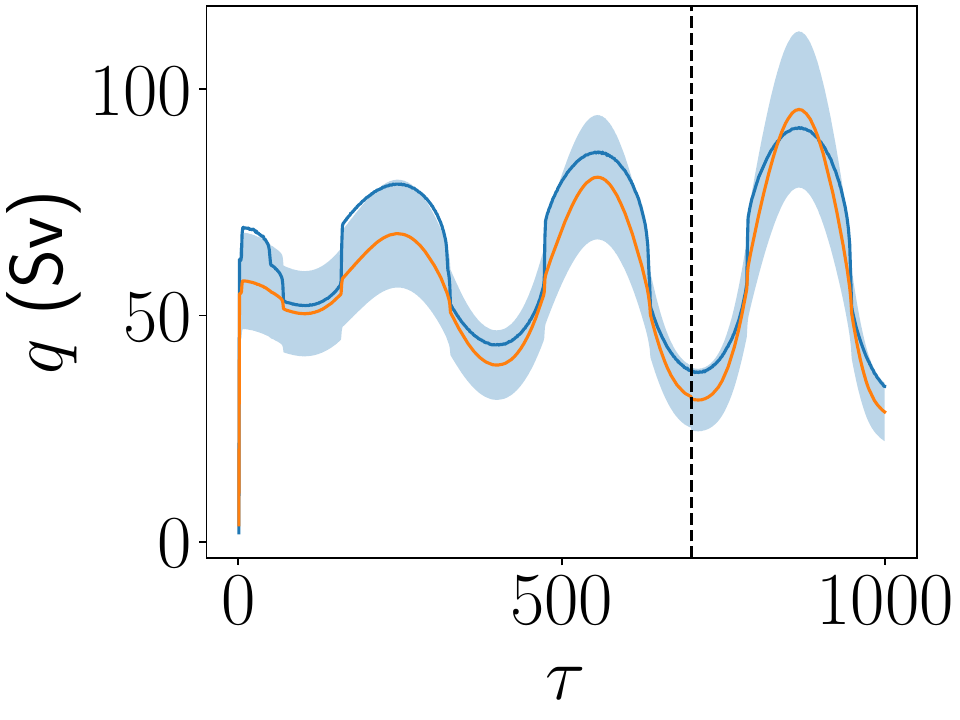}
	& \includegraphics[width=\linewidth,valign=m]{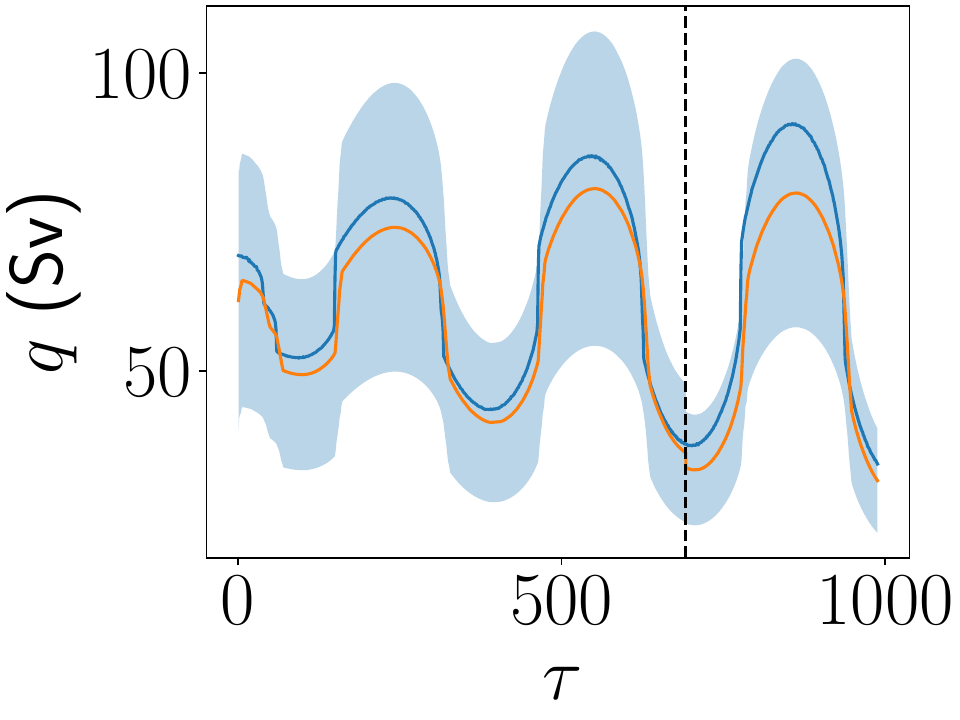} \\
	& MLP
	& \includegraphics[width=\linewidth,valign=m]{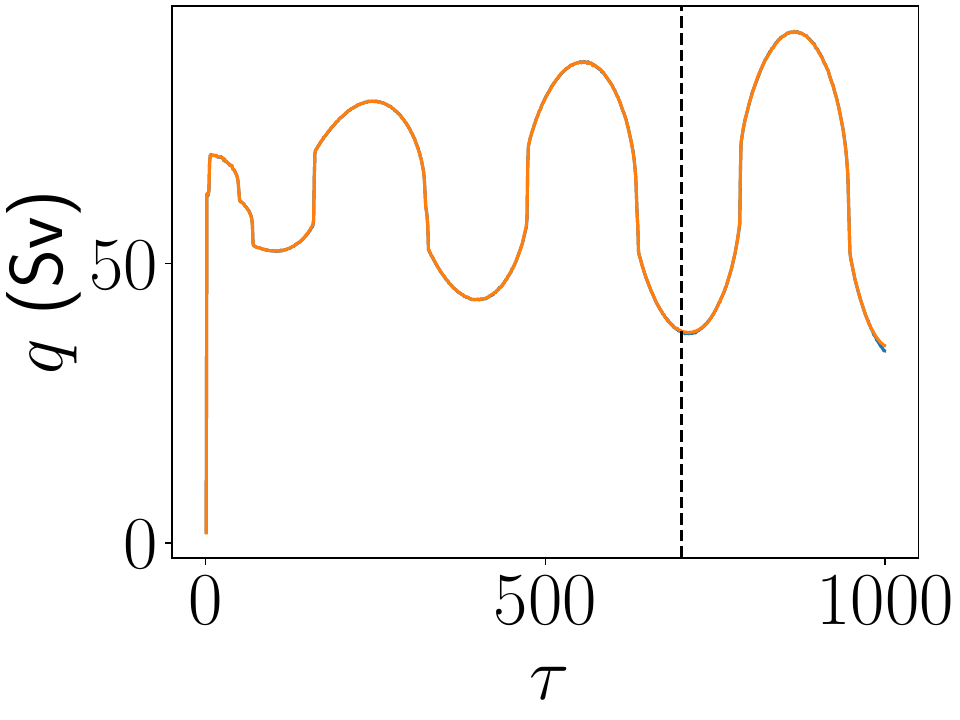}
	& \includegraphics[width=\linewidth,valign=m]{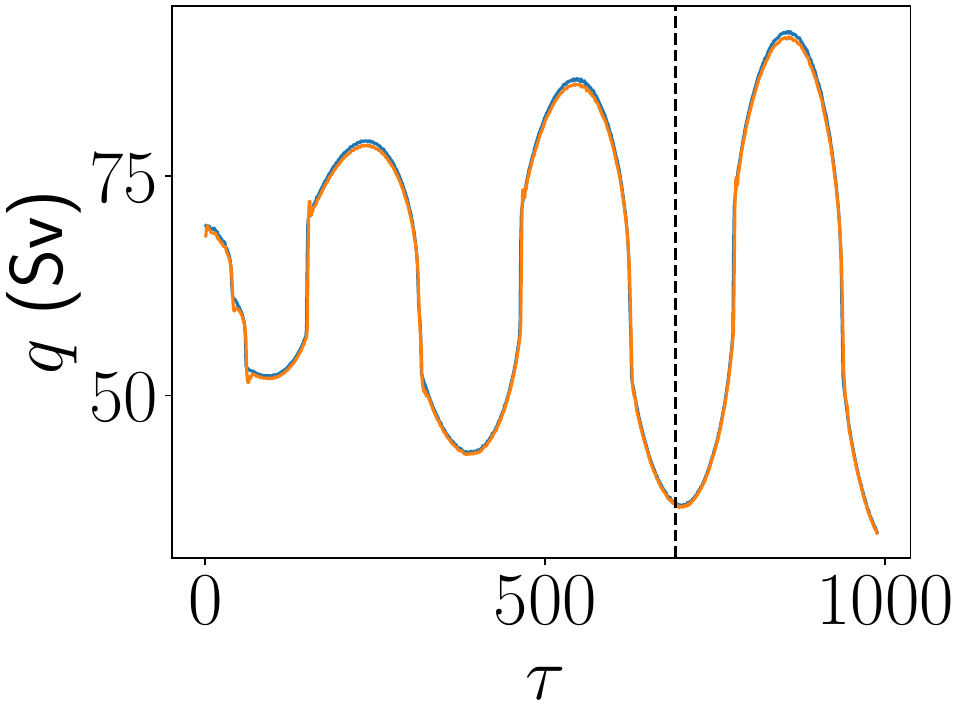} \\
	& Deep Ensemble
	& \includegraphics[width=\linewidth,valign=m]{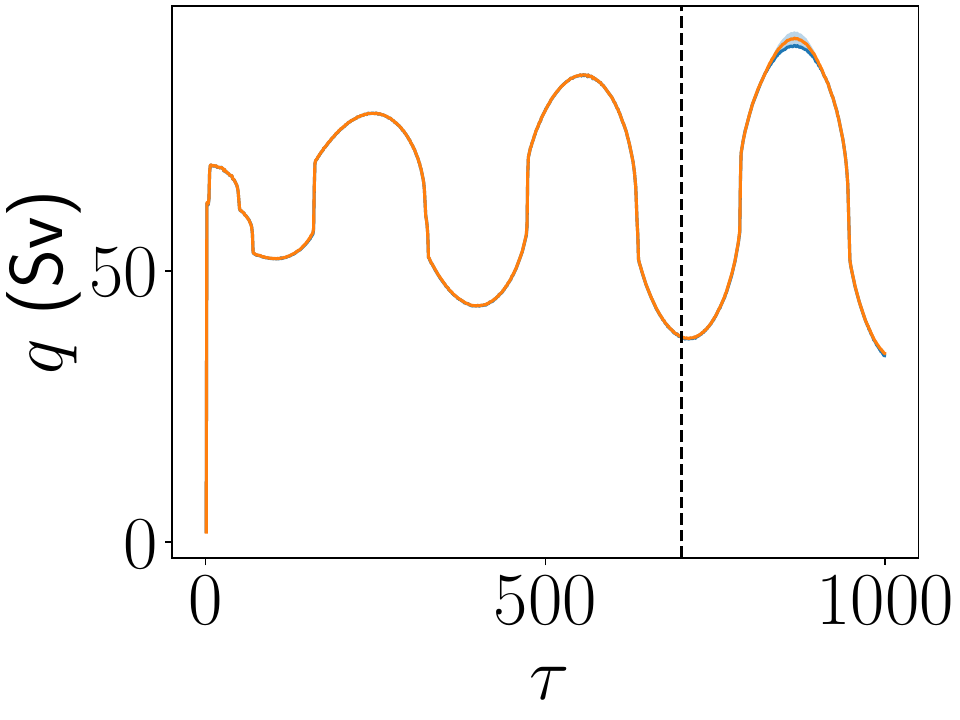}
	& \includegraphics[width=\linewidth,valign=m]{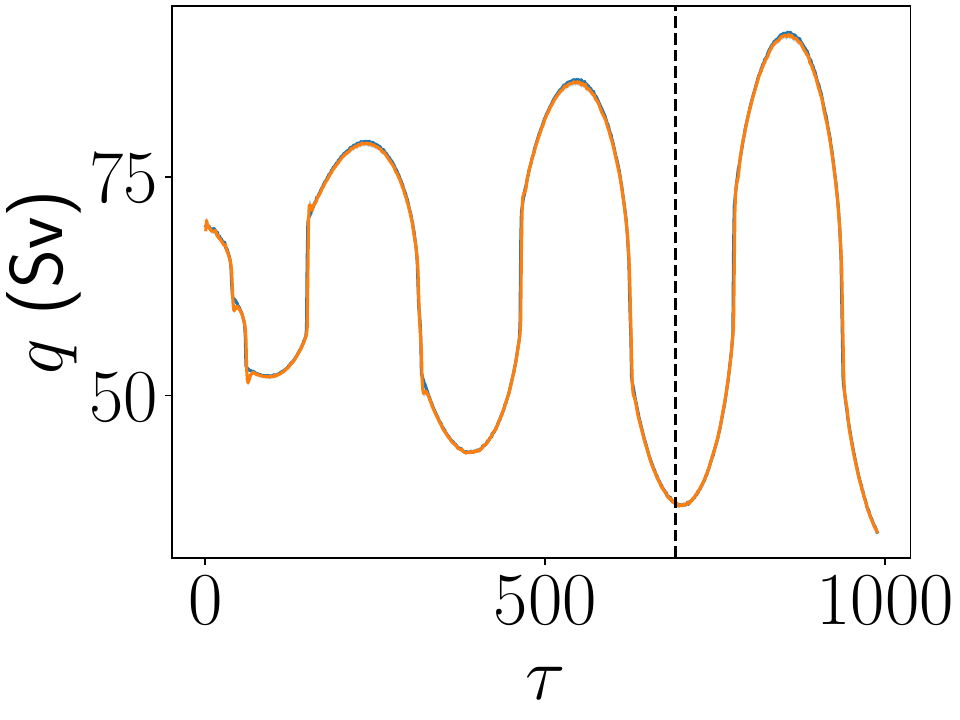} \\
	& RNN
	& ---
	& \includegraphics[width=\linewidth,valign=m]{sinusoidal_stationary/nonlinear_density/rnn/autoregressive/groundtruth-prediction.pdf} \\

	\bottomrule
	\caption{Predictive performance using \(F_s\) and \(\rho^{\text{EOS-80}}\).}%
	\label{fig:performance-standard-nonlinear}%
\end{longtable}

\subsubsection{Explainability}

\begin{longtable}{p{.06\columnwidth}p{.08\columnwidth}C{.2\columnwidth}C{.2\columnwidth}C{.2\columnwidth}C{.2\columnwidth}}
	\toprule
	Scenario & Architecture & DeepLIFT\@: PI & SHAP\@: PI & DeepLIFT\@: AR & SHAP\@: AR \\
	\midrule

	\multirow[c]{3}{*}{\rotatebox[origin=c]{90}{\(F_s\): Linear}}
	& BNN
	& \includegraphics[width=\linewidth,valign=m]{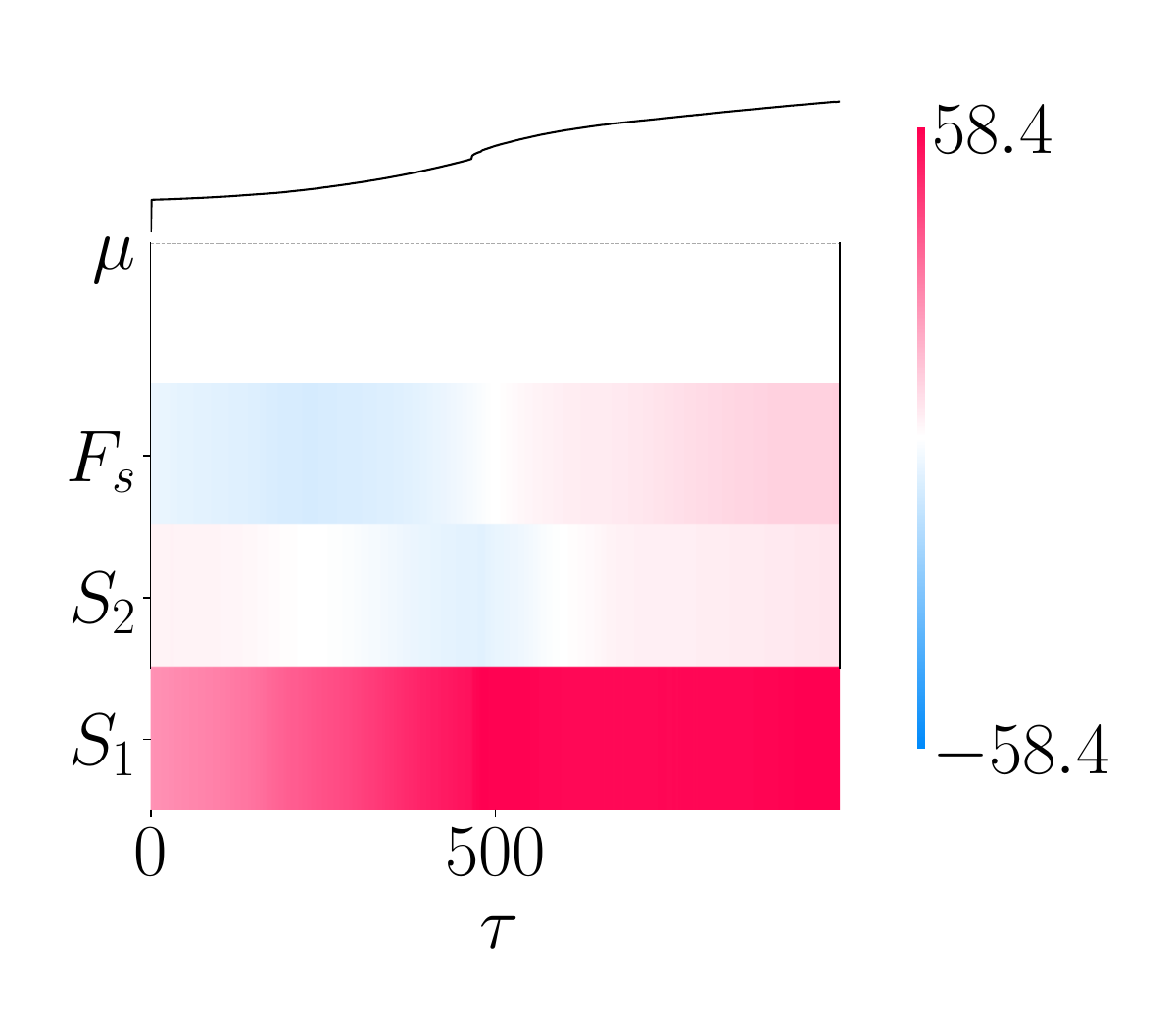}
	& \includegraphics[width=\linewidth,valign=m]{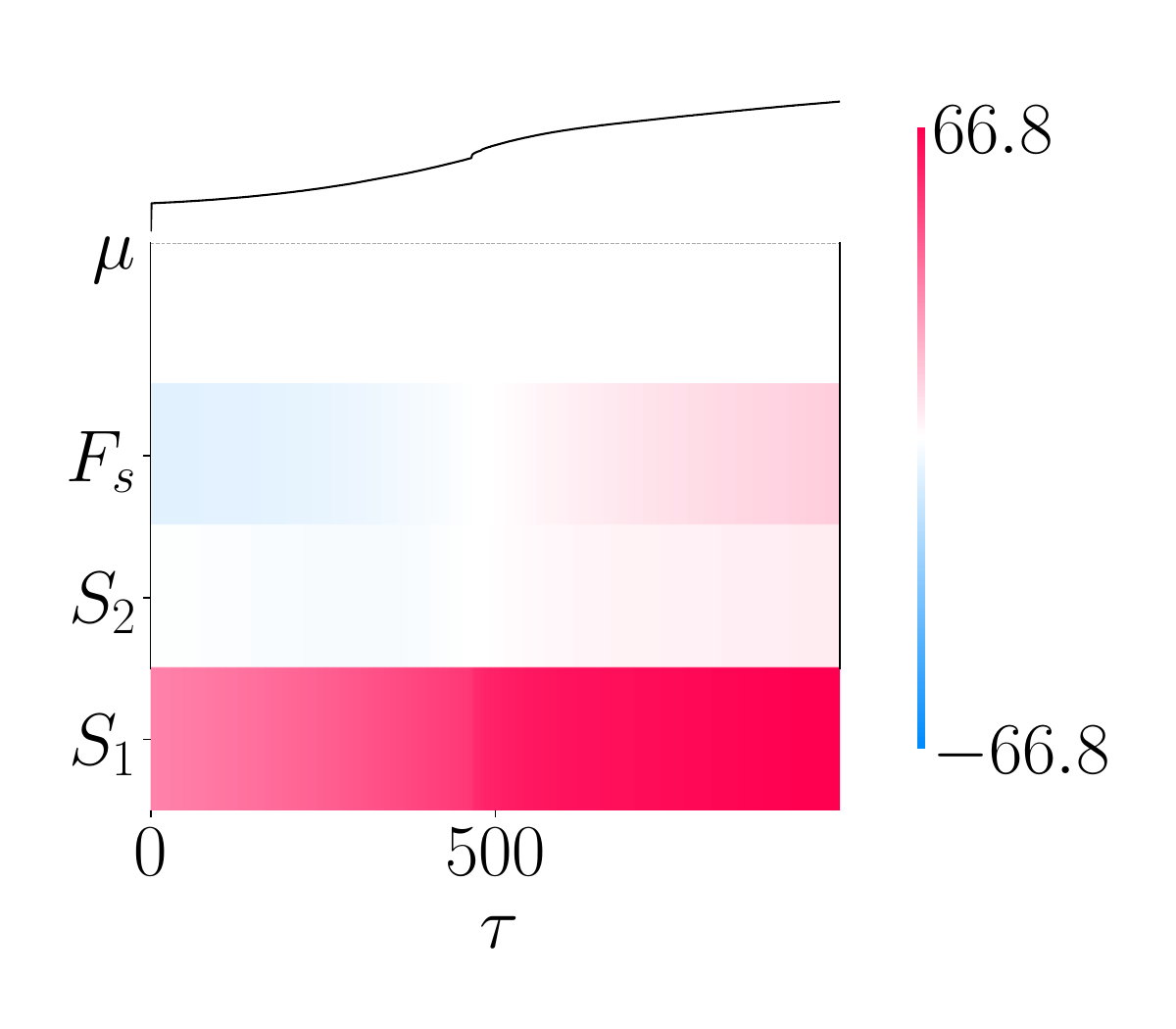}
	& \includegraphics[width=\linewidth,valign=m]{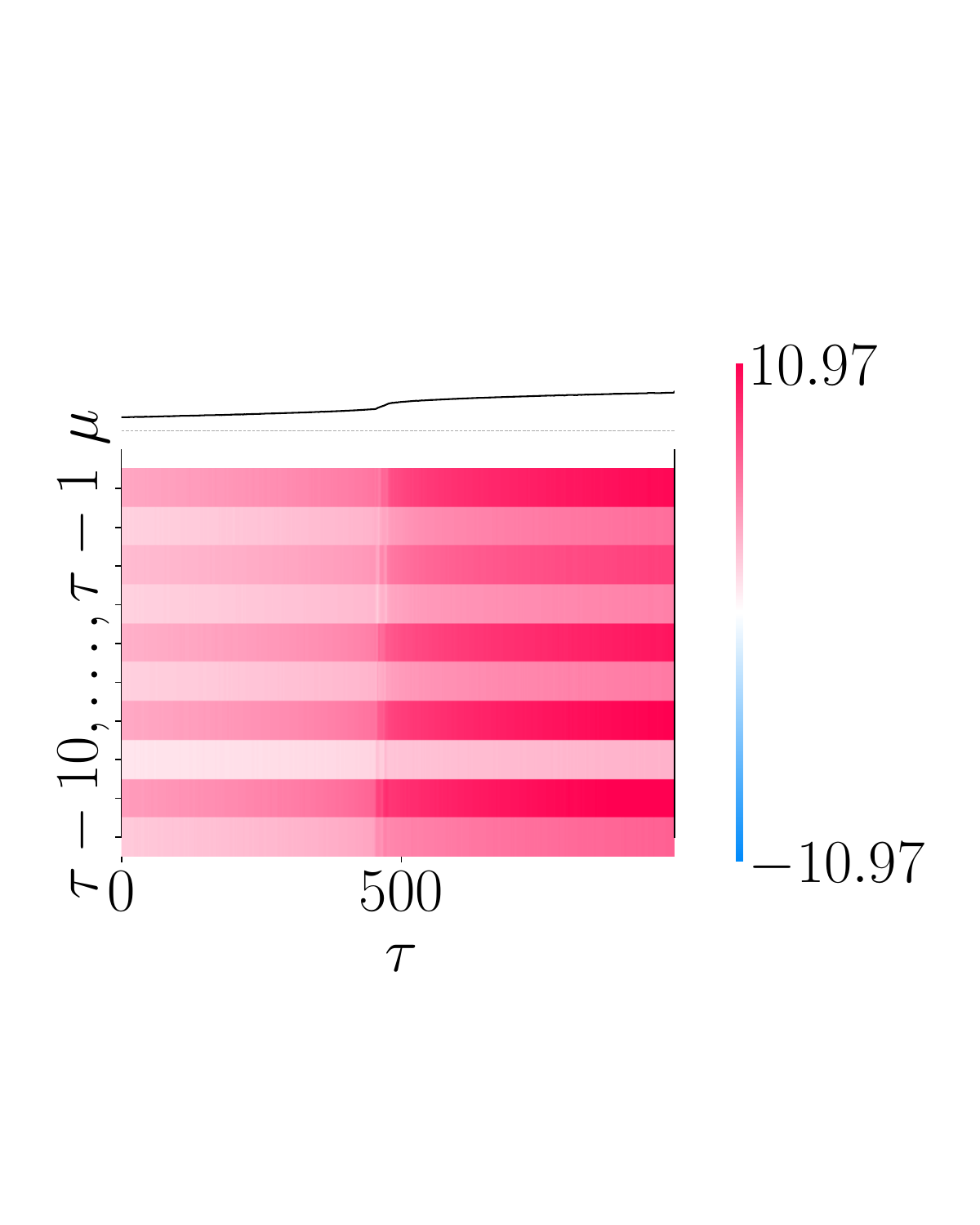}
	& \includegraphics[width=\linewidth,valign=m]{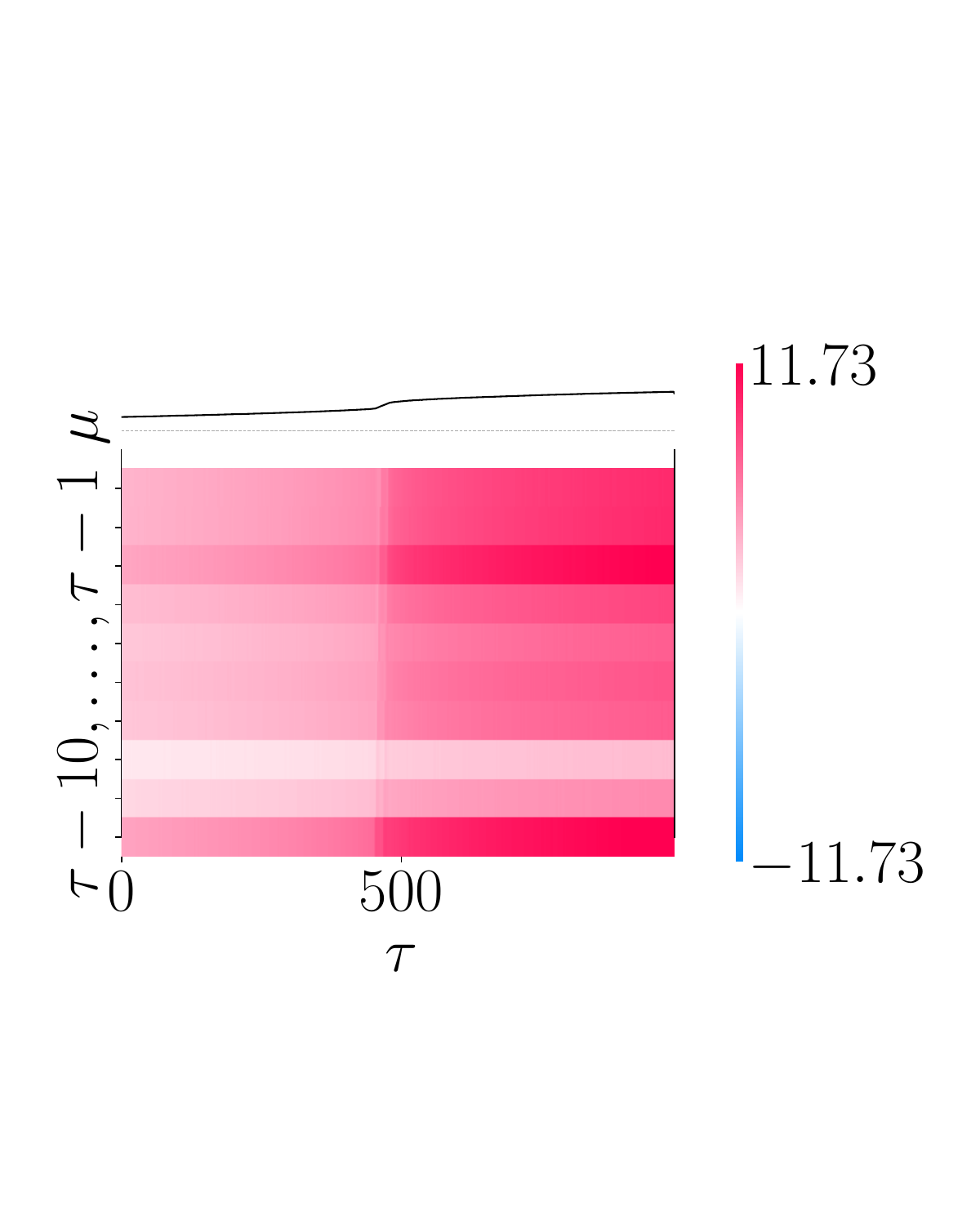} \\
	& MLP
	& \includegraphics[width=\linewidth,valign=m]{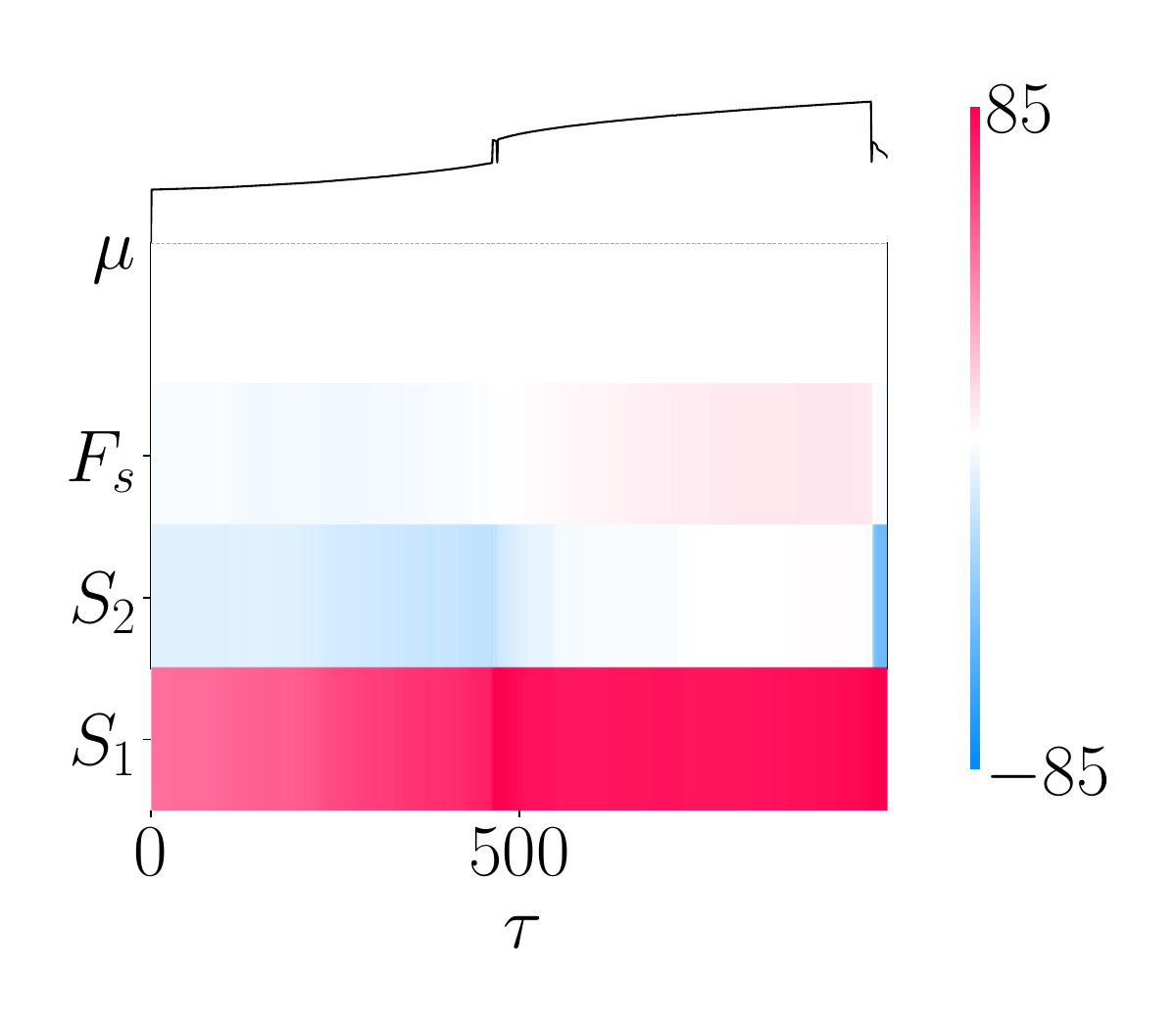}
	& \includegraphics[width=\linewidth,valign=m]{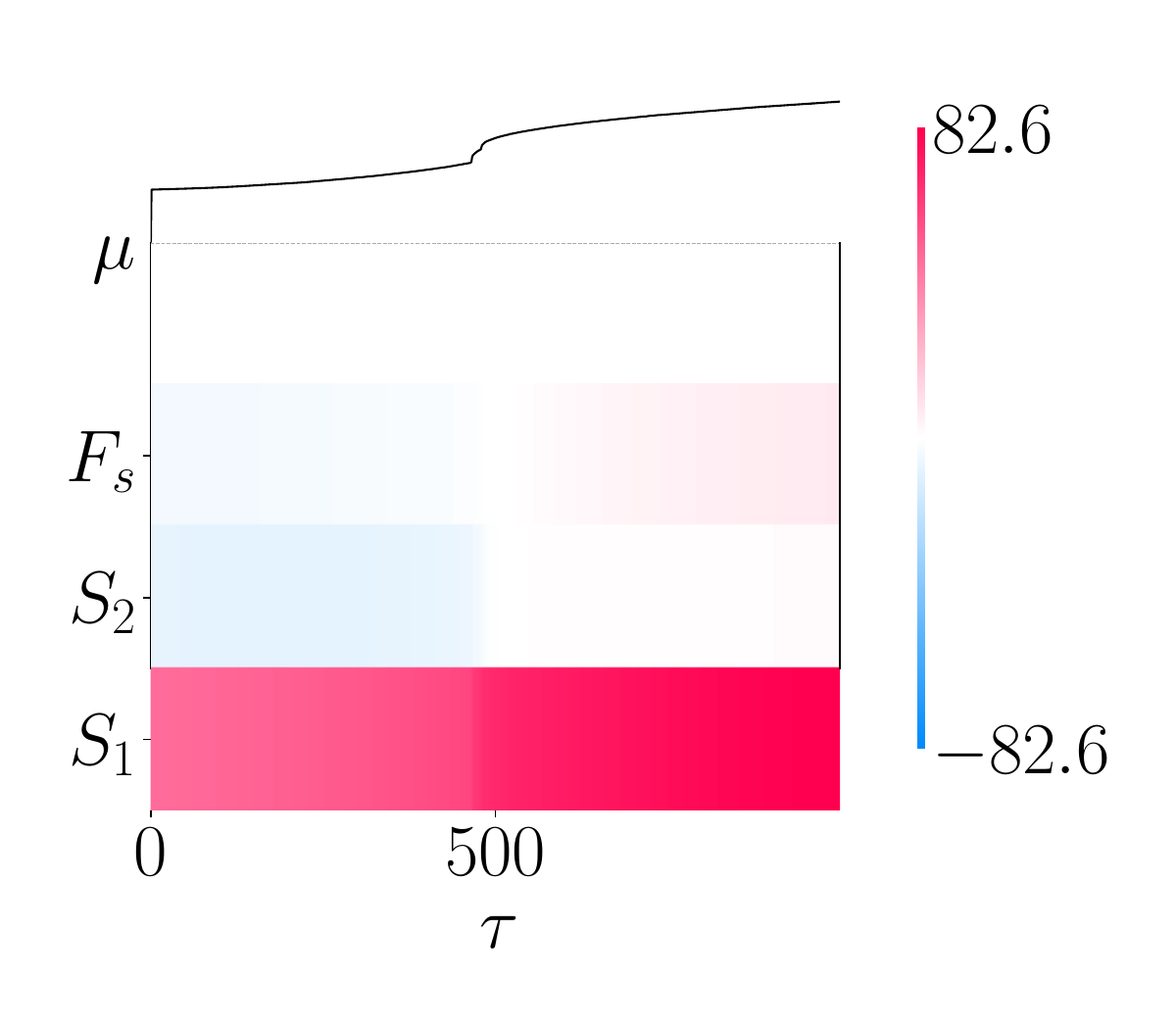}
	& \includegraphics[width=\linewidth,valign=m]{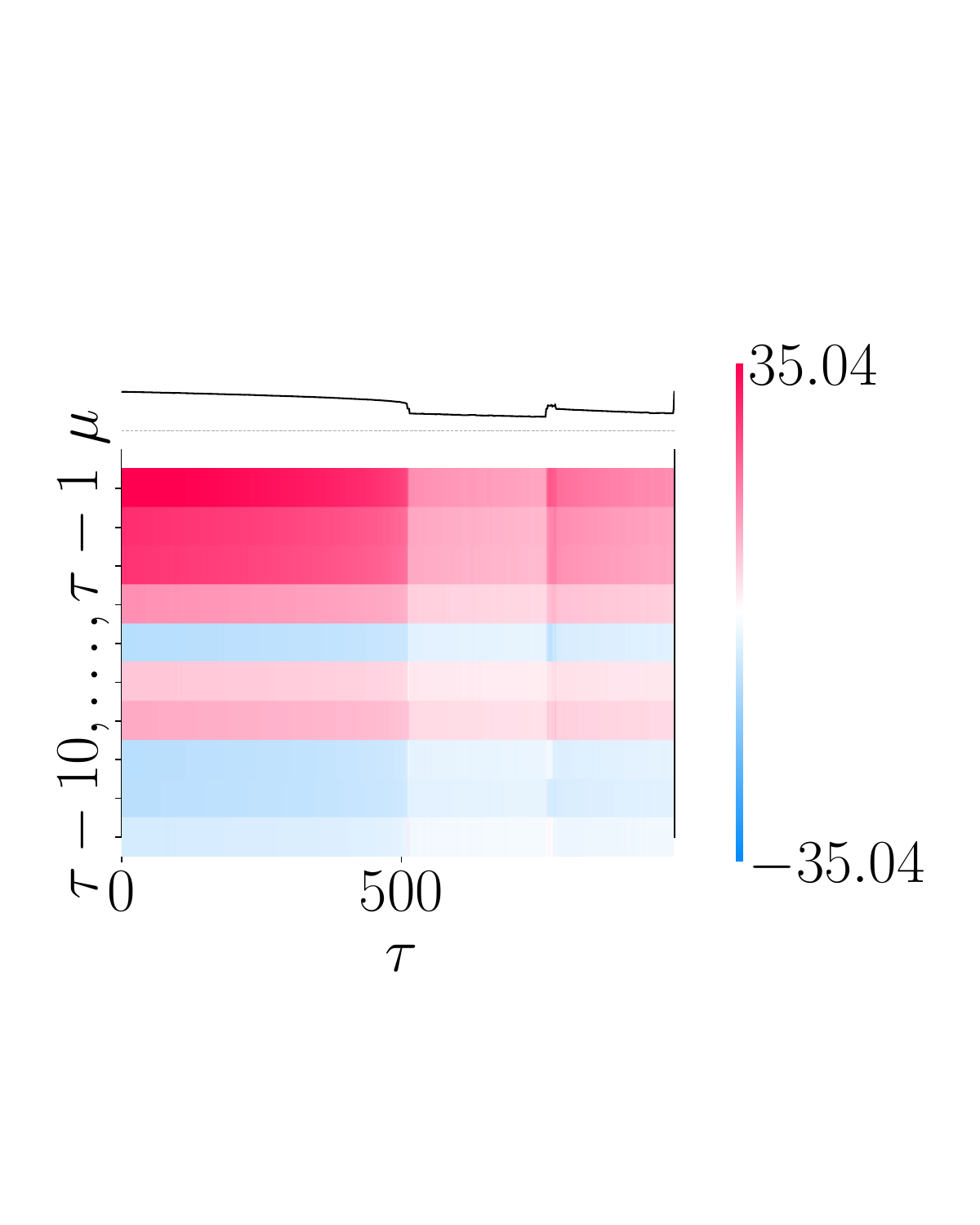}
	& \includegraphics[width=\linewidth,valign=m]{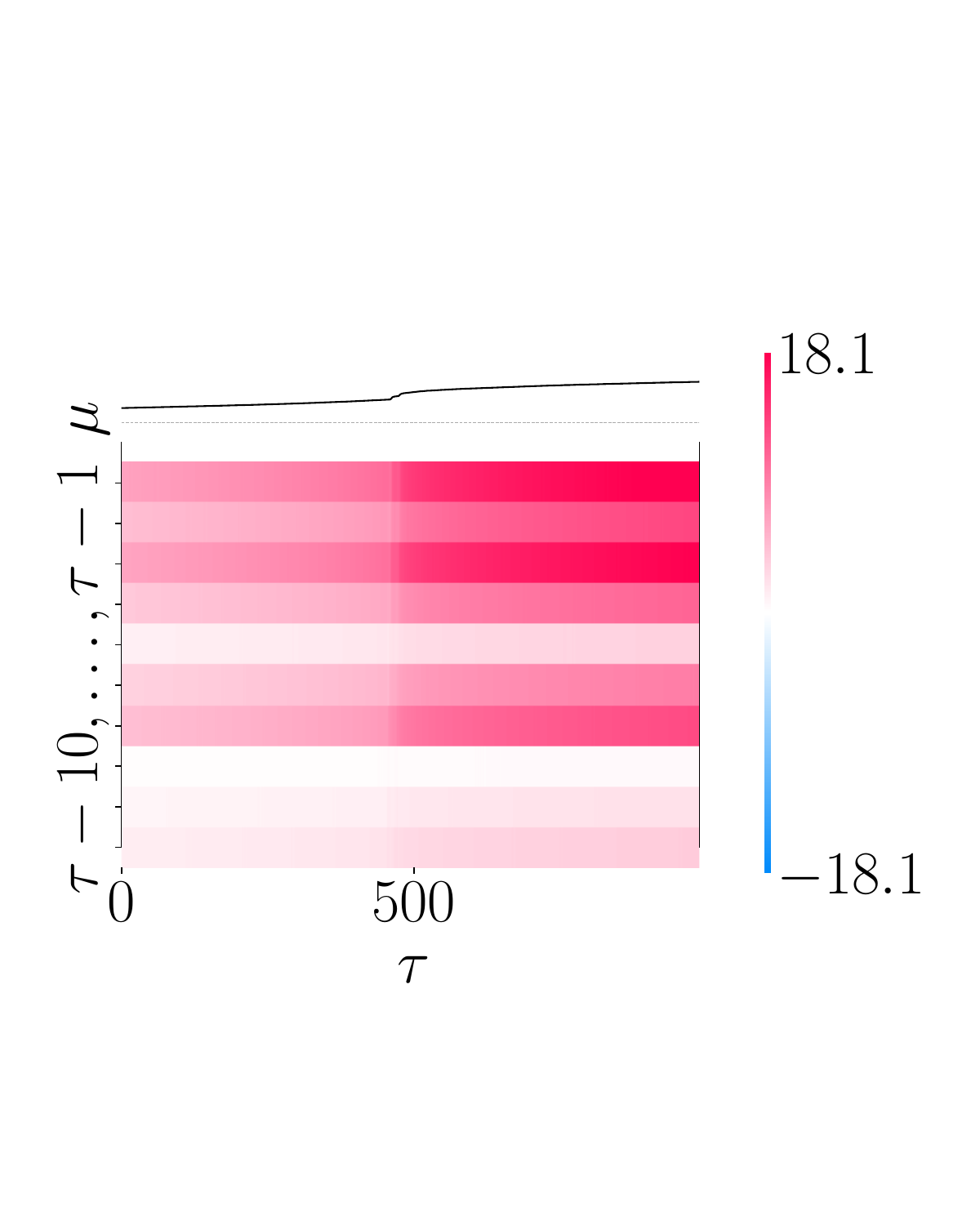} \\
	& Deep Ensemble
	& \includegraphics[width=\linewidth,valign=m]{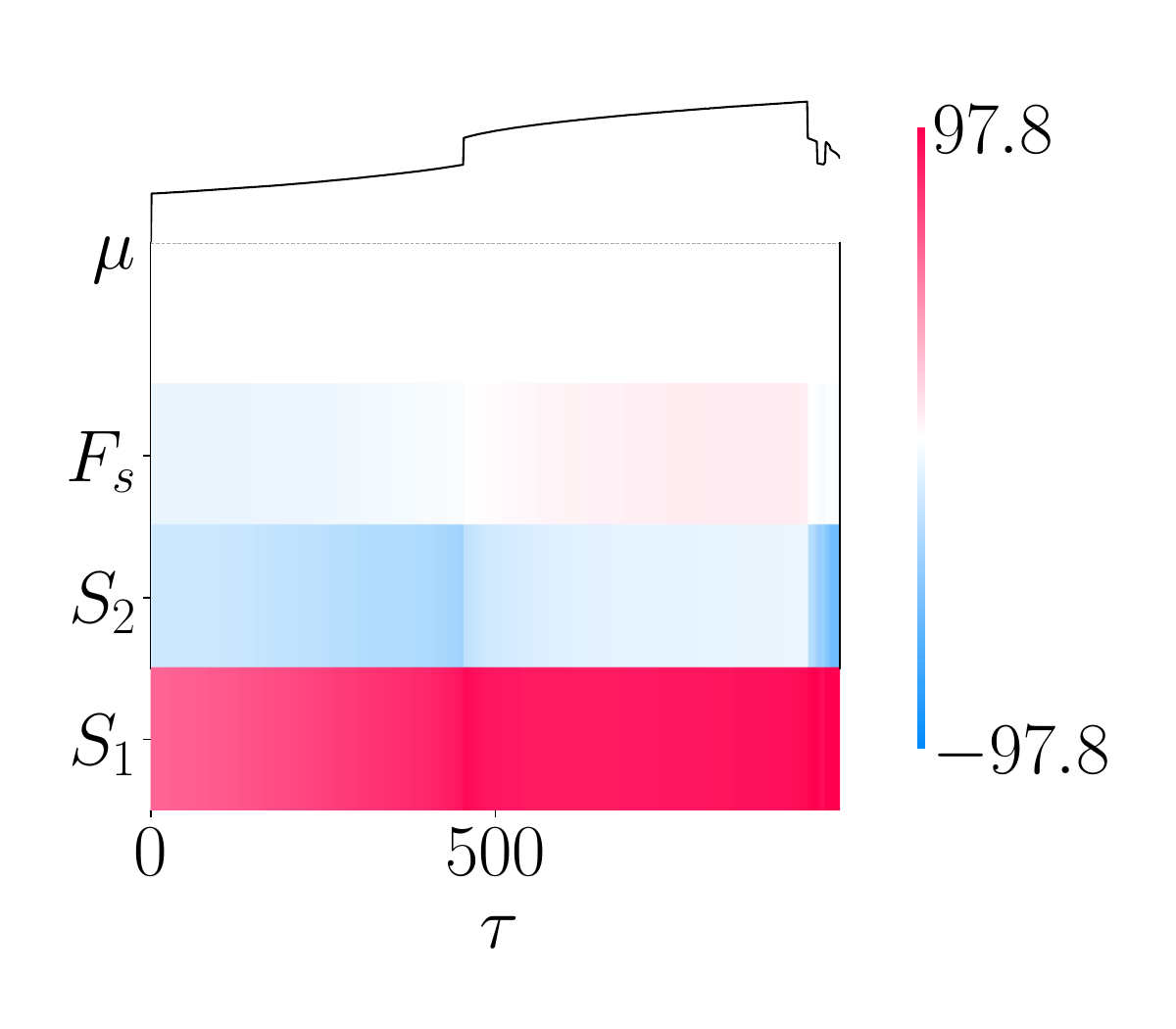}
	& \includegraphics[width=\linewidth,valign=m]{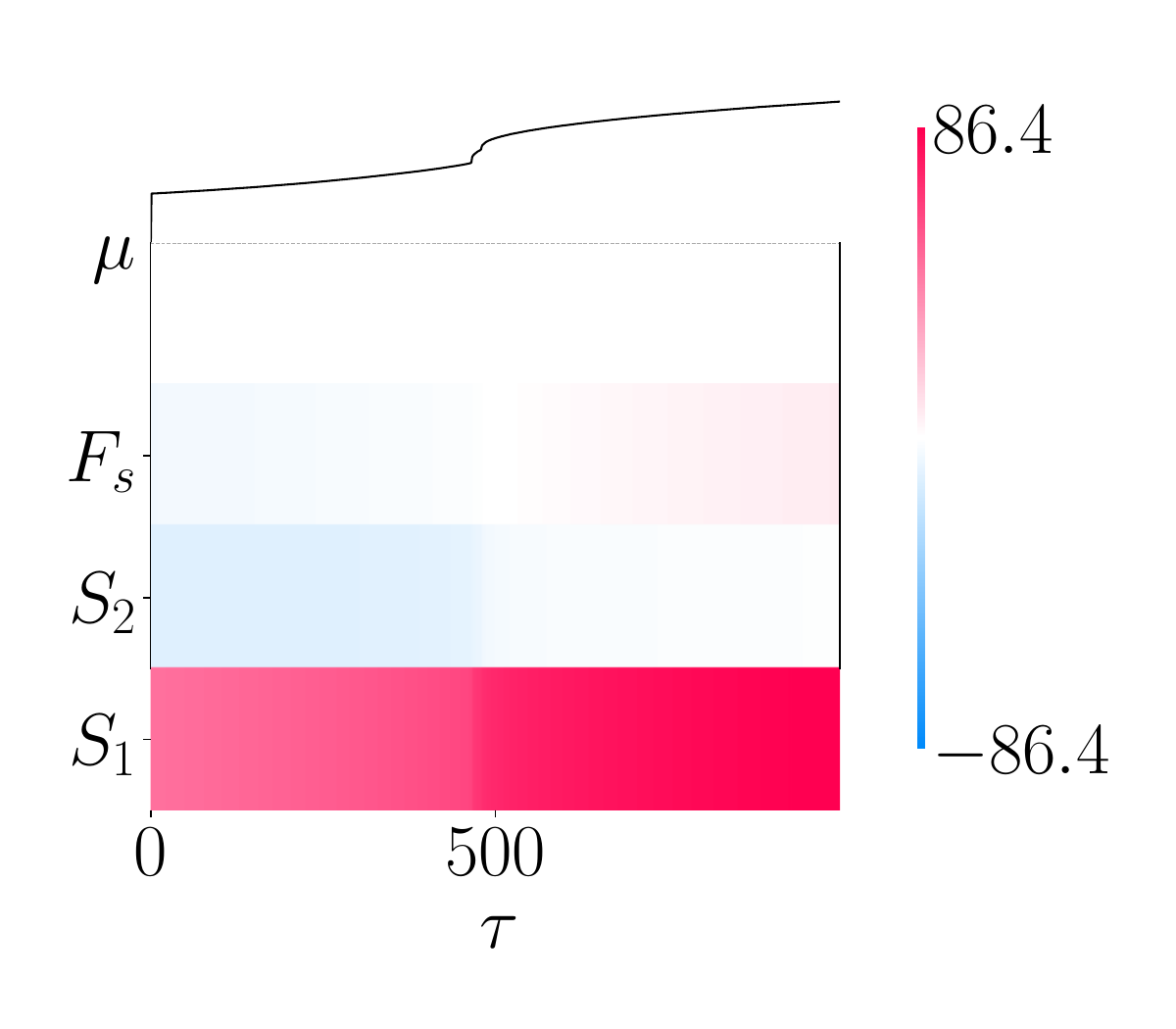}
	& \includegraphics[width=\linewidth,valign=m]{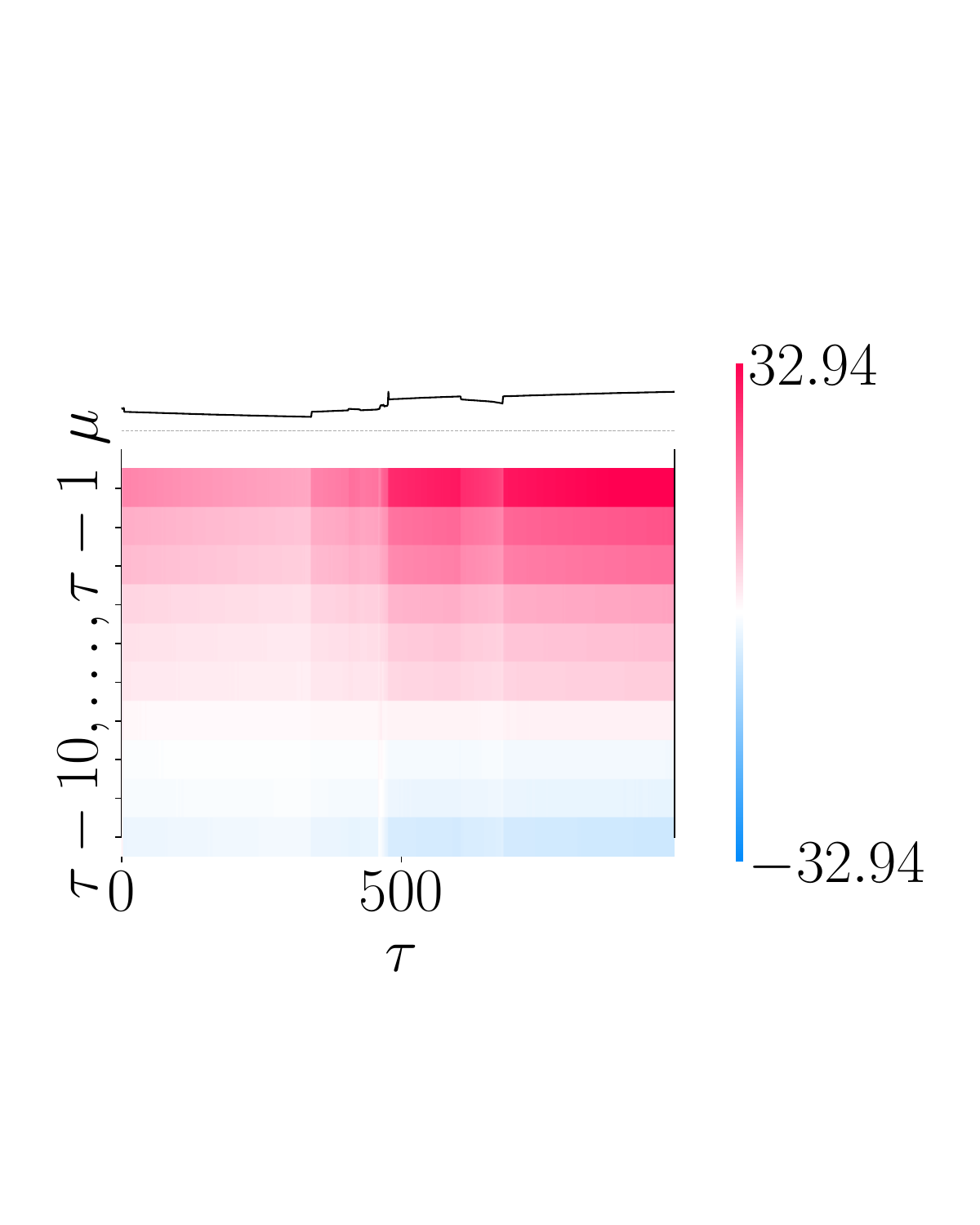}
	& \includegraphics[width=\linewidth,valign=m]{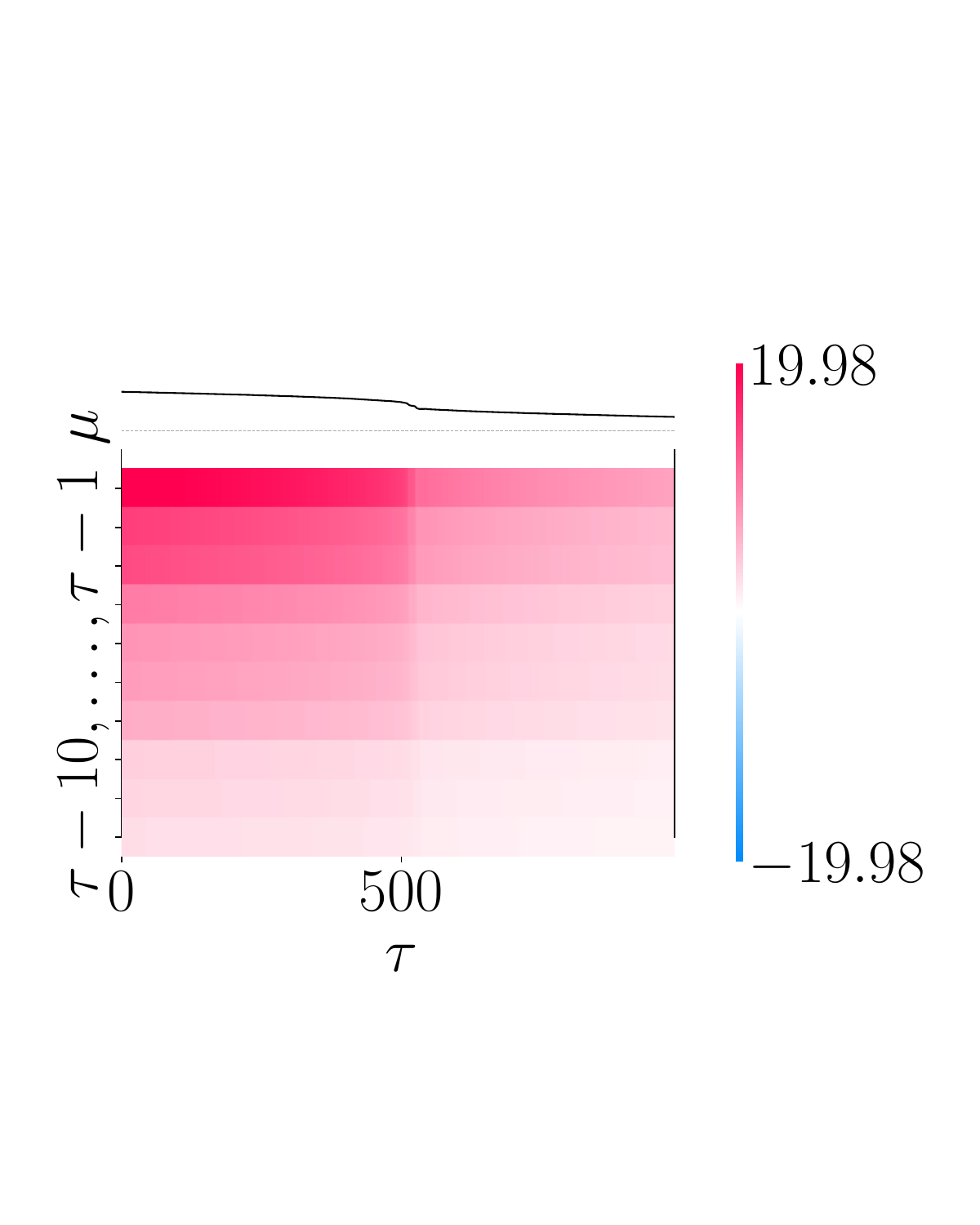} \\
	& RNN
	& ---
	& ---
	& \includegraphics[width=\linewidth,valign=m]{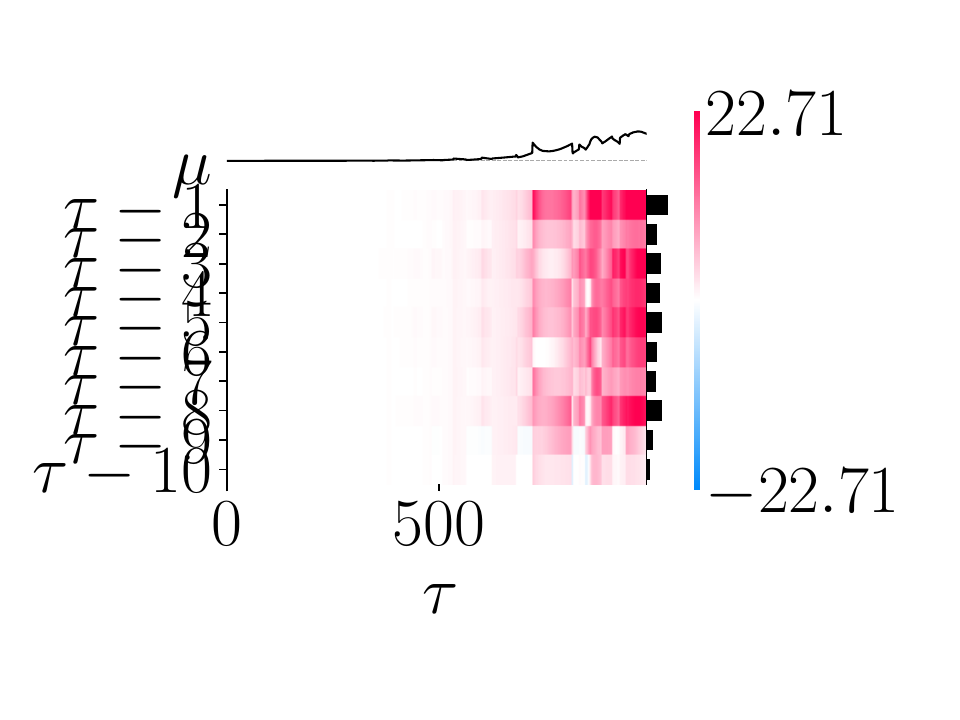}
	& \includegraphics[width=\linewidth,valign=m]{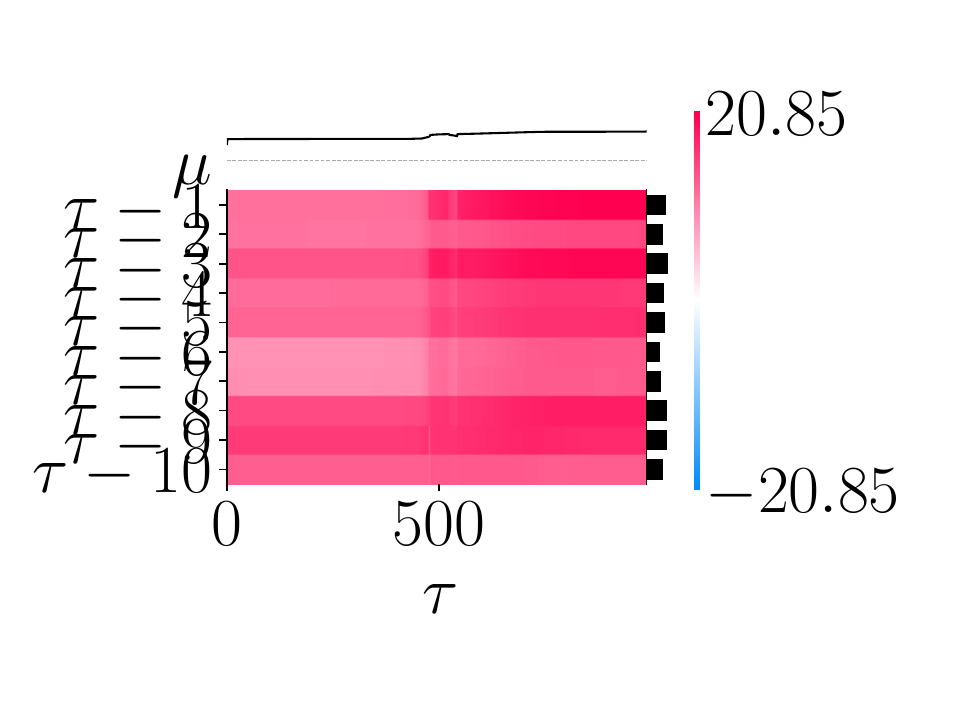} \\

	\midrule

	\multirow[c]{3}{*}{\rotatebox[origin=c]{90}{\(F_s\): Sinusoidal (stationary)}}
	& BNN
	& \includegraphics[width=\linewidth,valign=m]{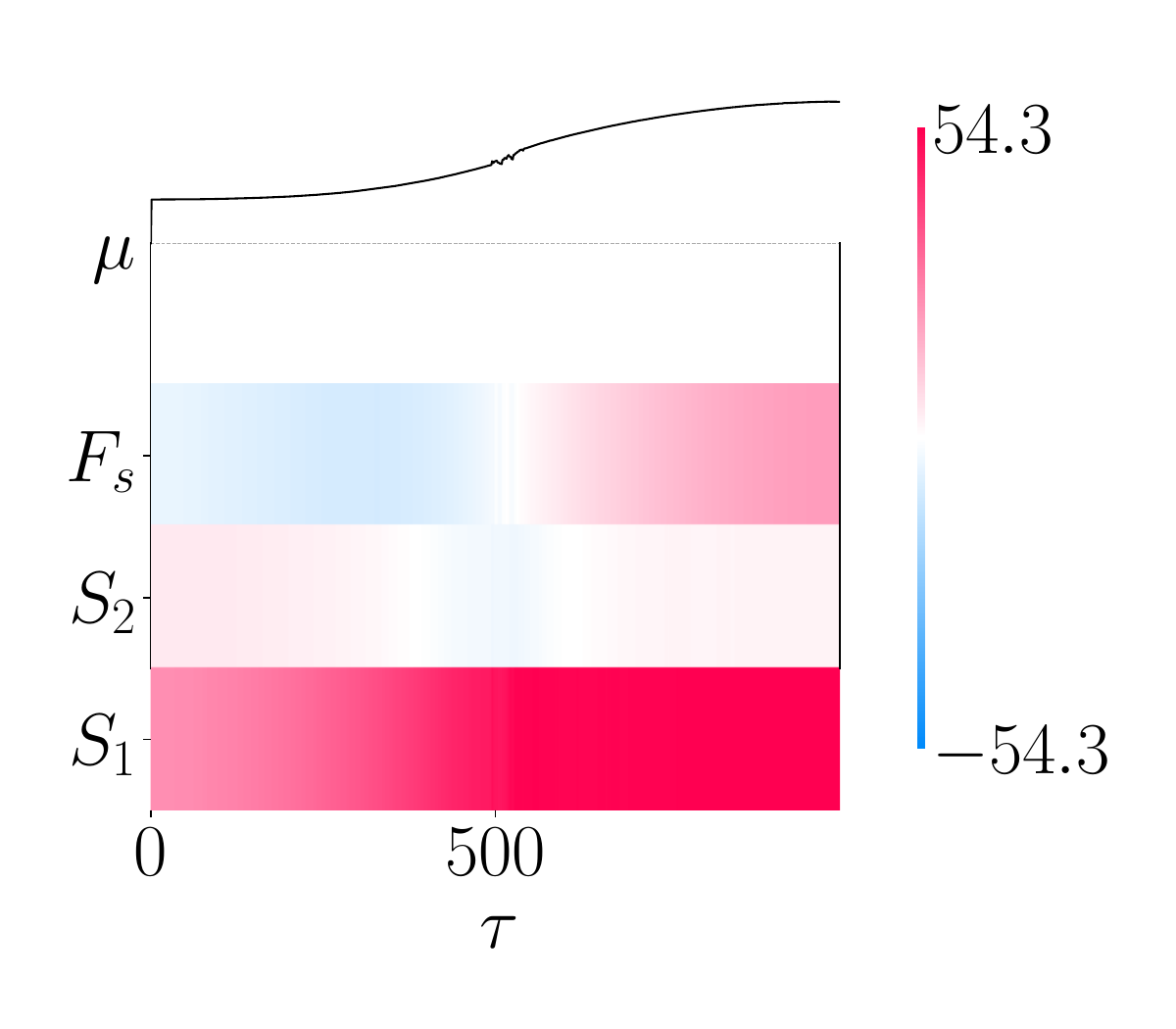}
	& \includegraphics[width=\linewidth,valign=m]{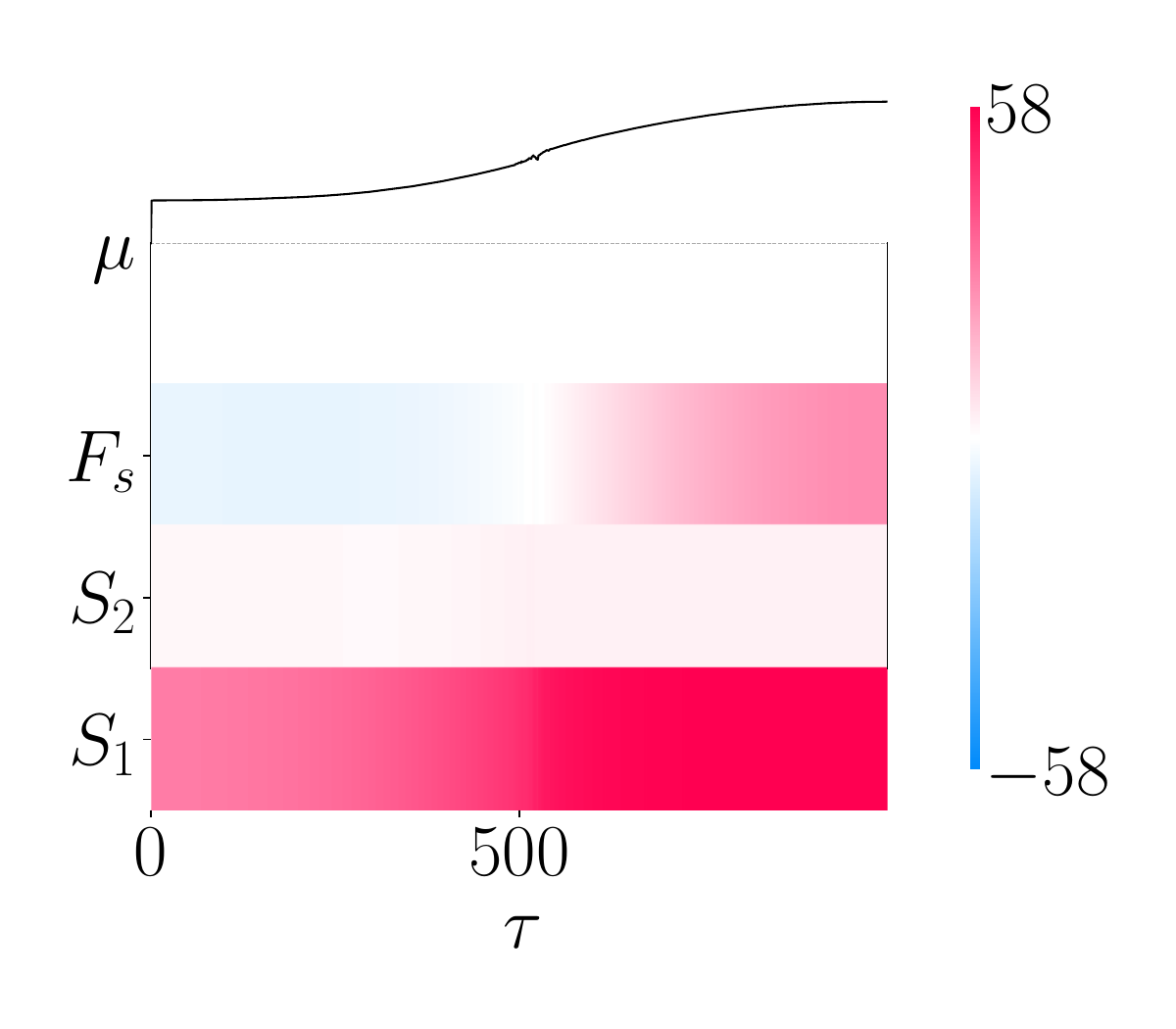}
	& \includegraphics[width=\linewidth,valign=m]{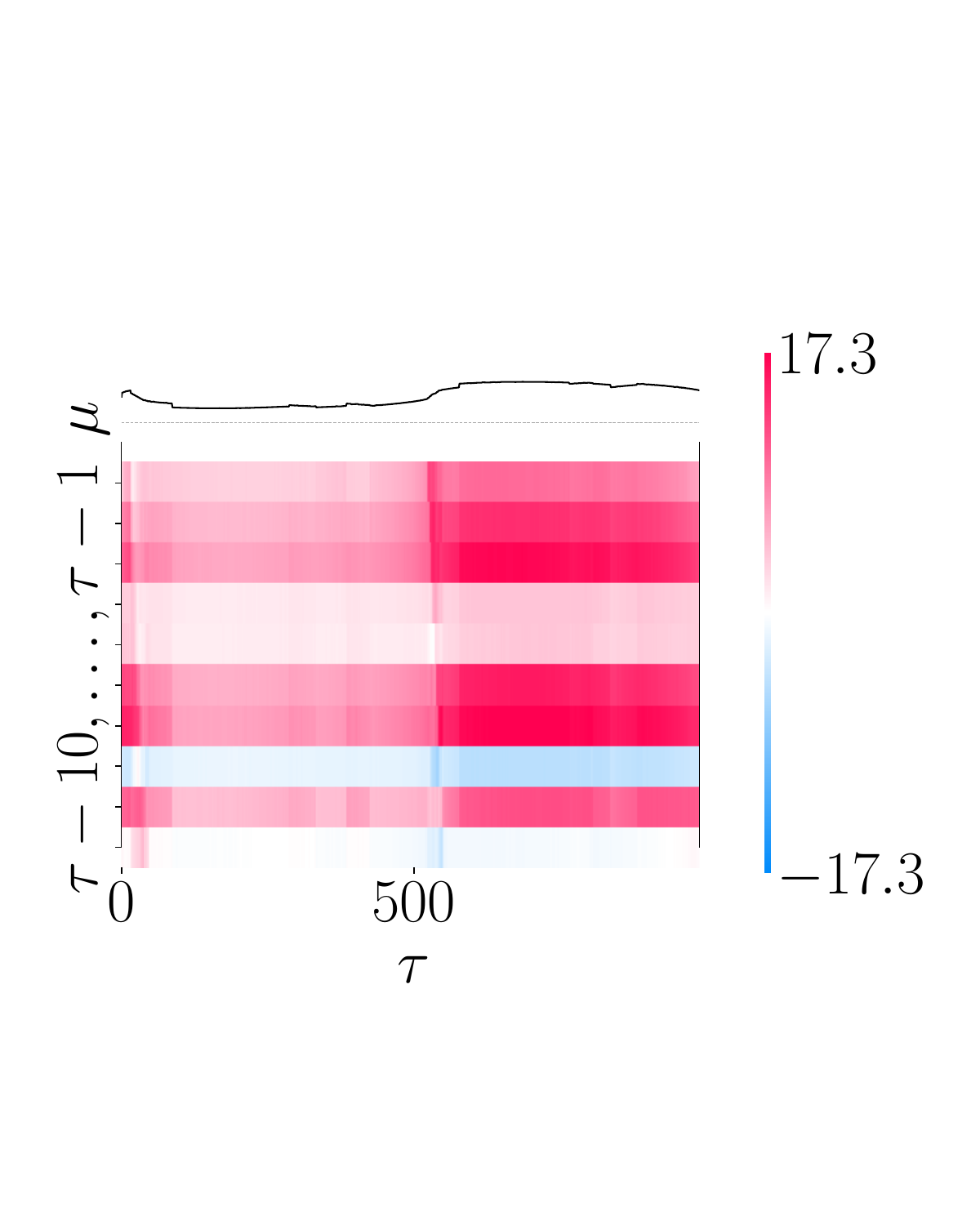}
	& \includegraphics[width=\linewidth,valign=m]{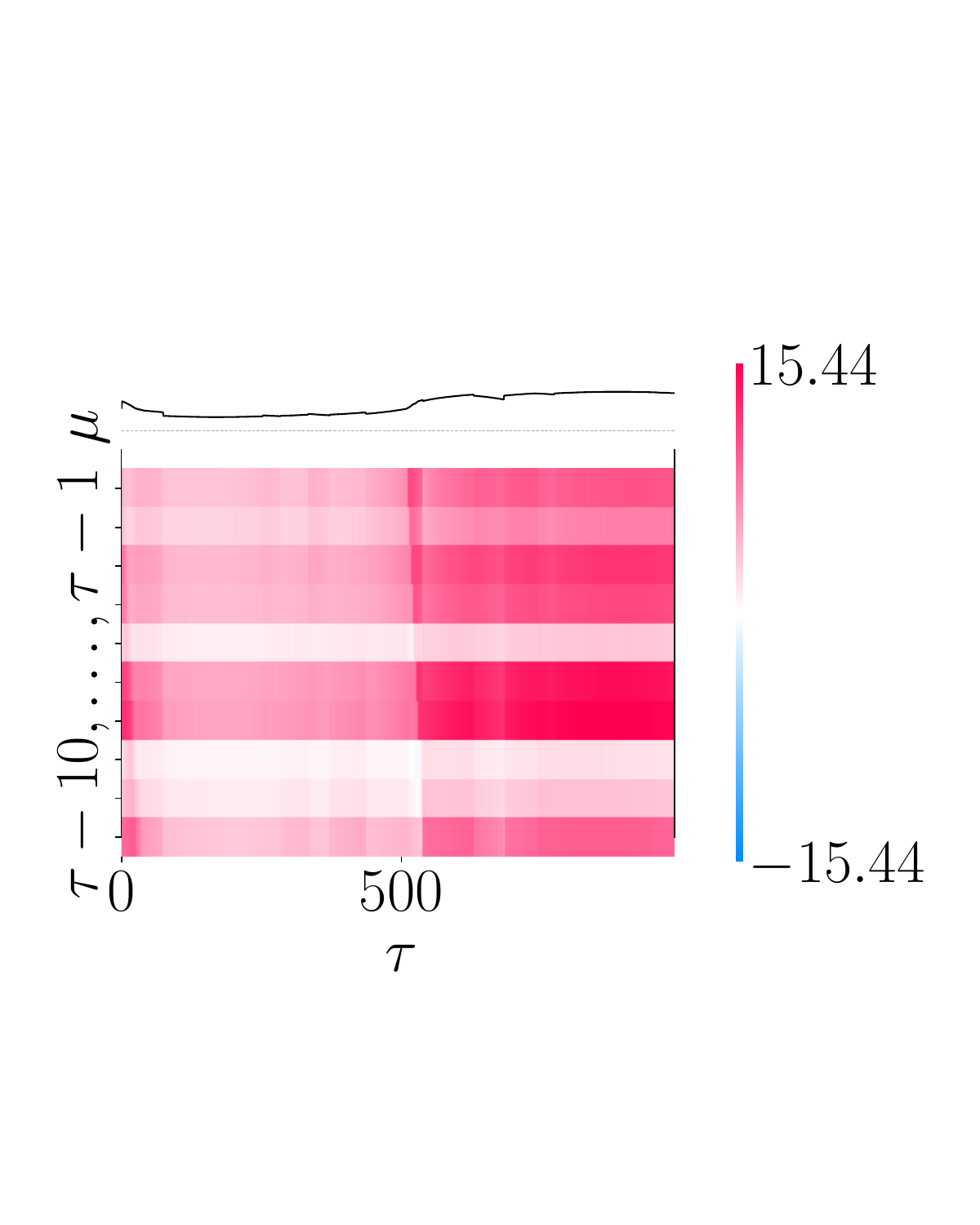} \\
	& MLP
	& \includegraphics[width=\linewidth,valign=m]{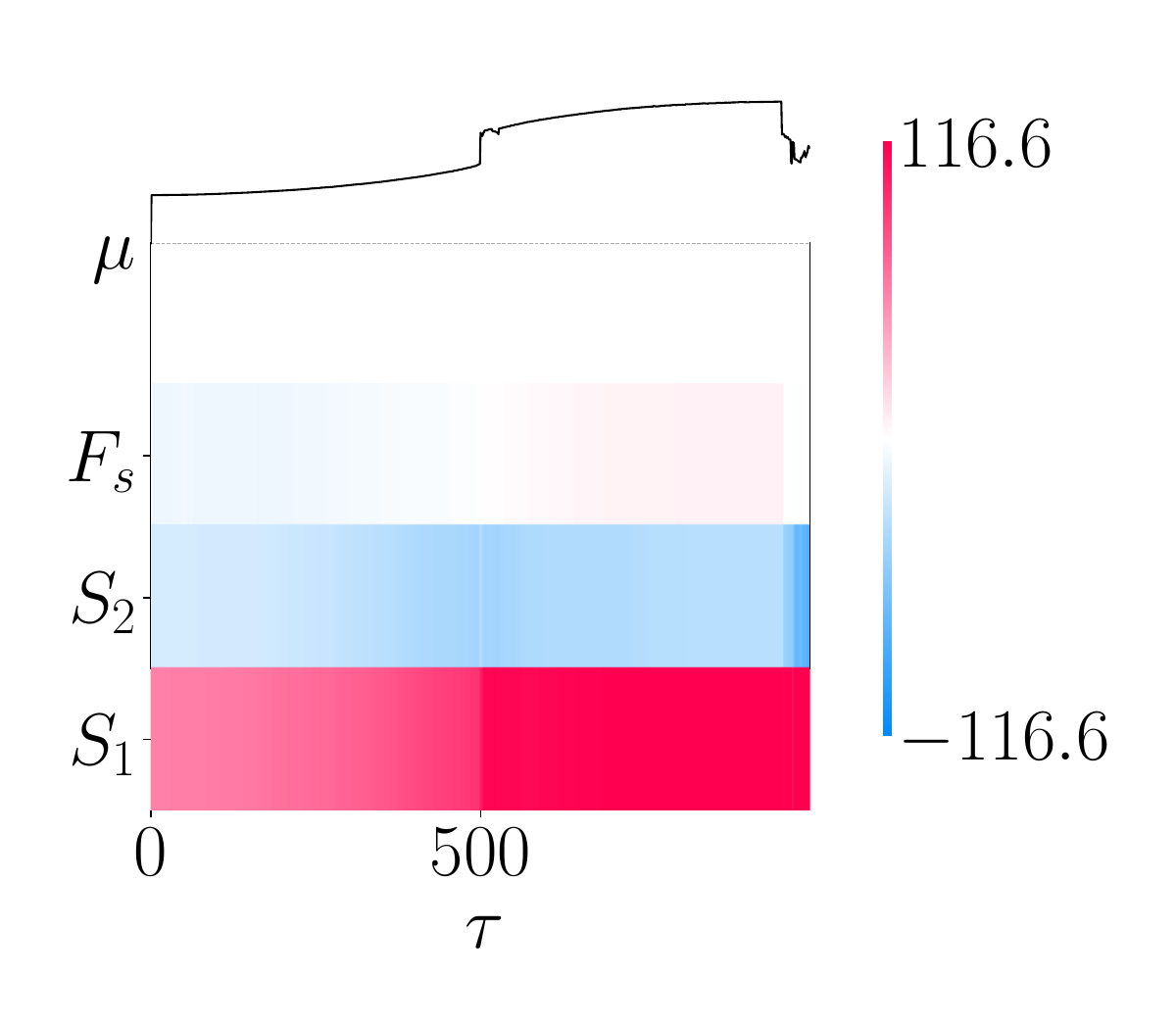}
	& \includegraphics[width=\linewidth,valign=m]{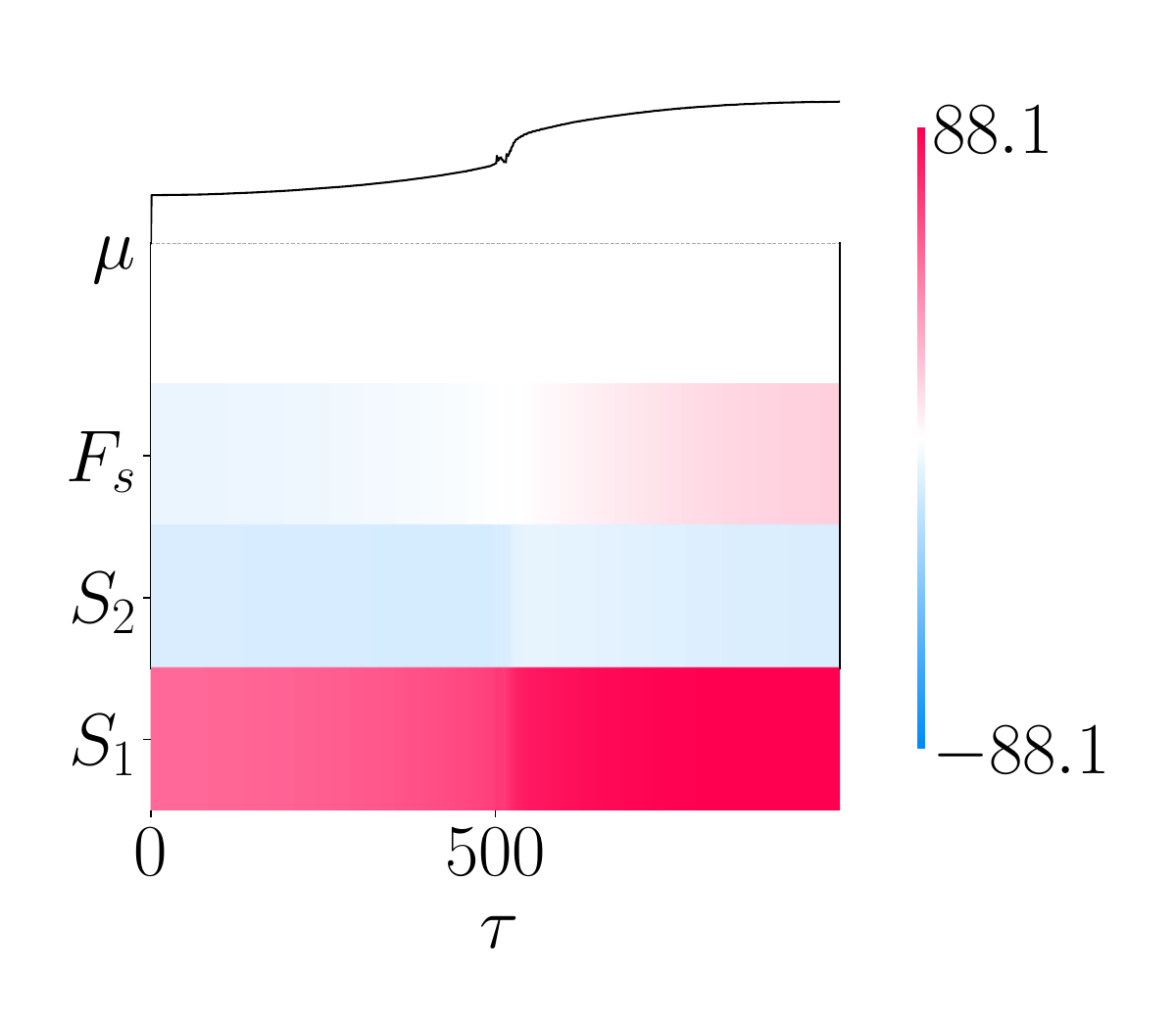}
	& \includegraphics[width=\linewidth,valign=m]{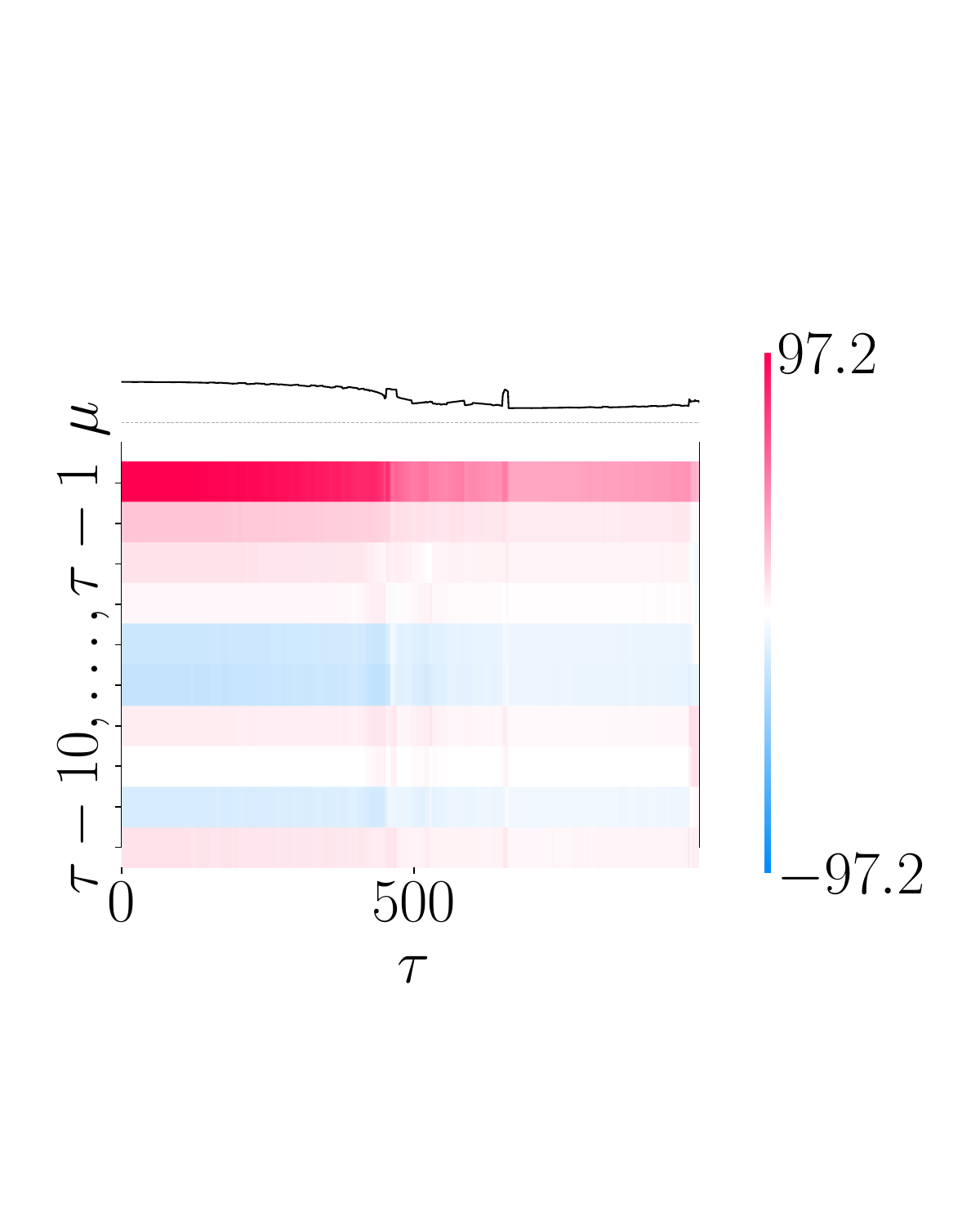}
	& \includegraphics[width=\linewidth,valign=m]{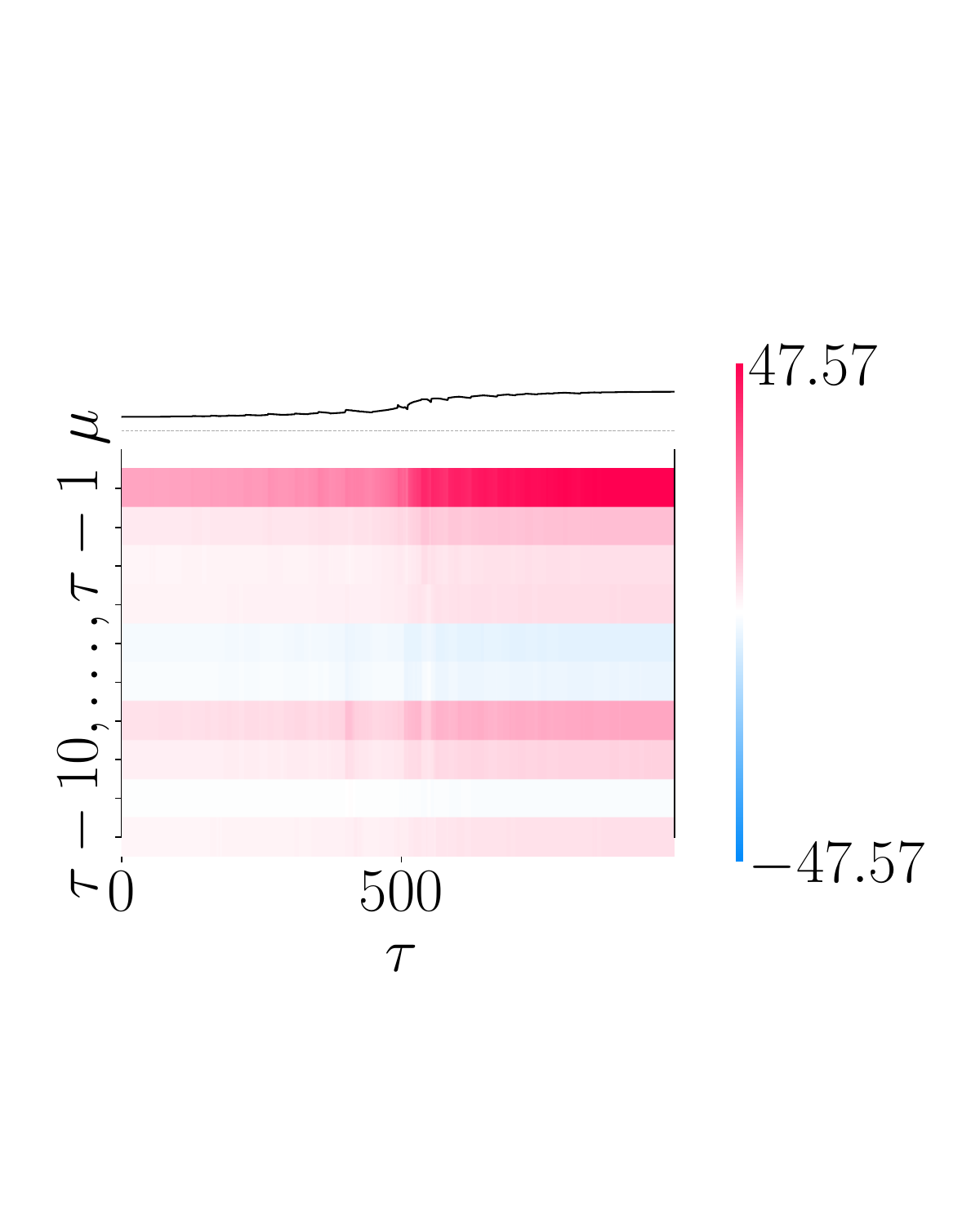} \\
	& Deep Ensemble
	& \includegraphics[width=\linewidth,valign=m]{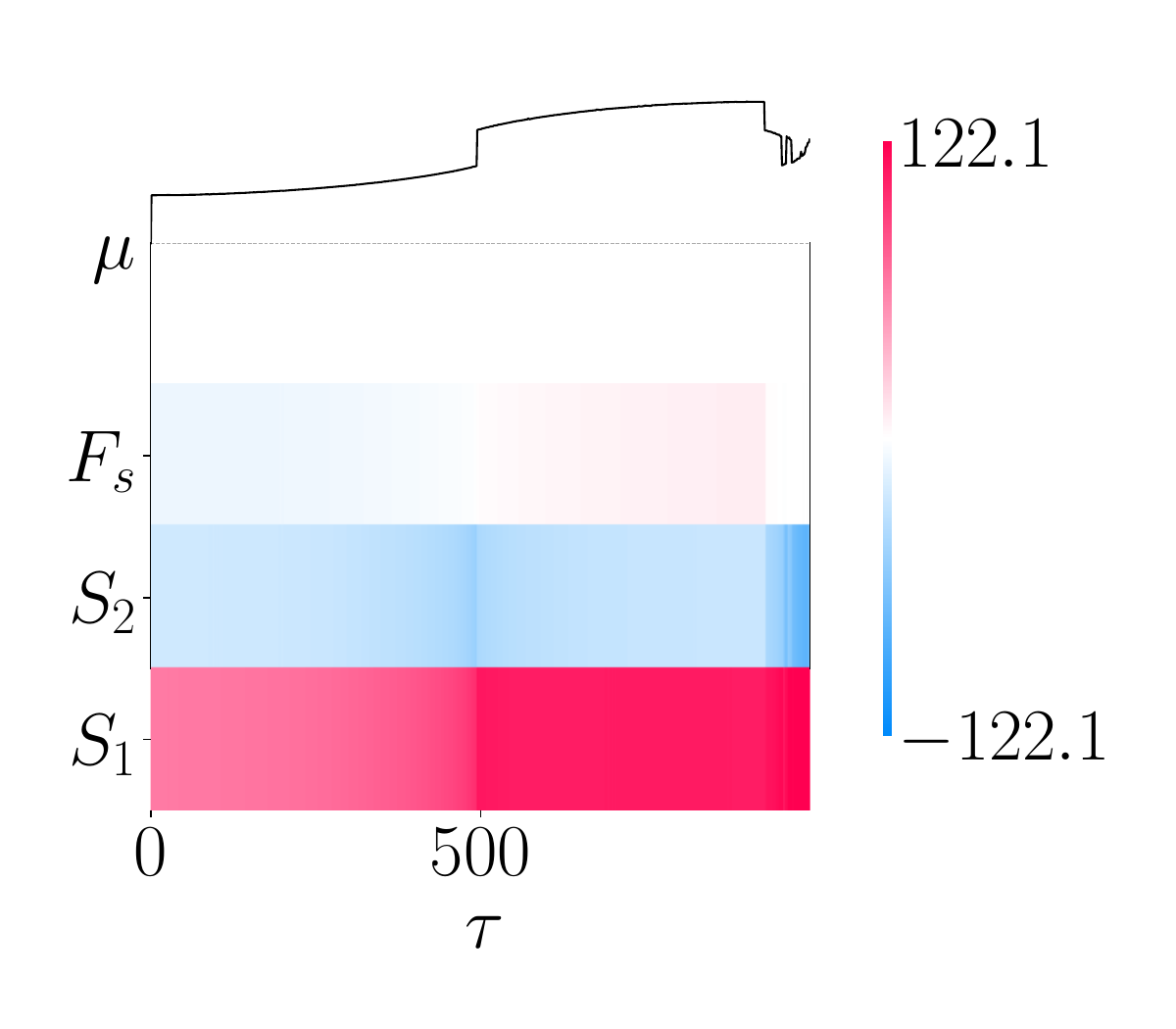}
	& \includegraphics[width=\linewidth,valign=m]{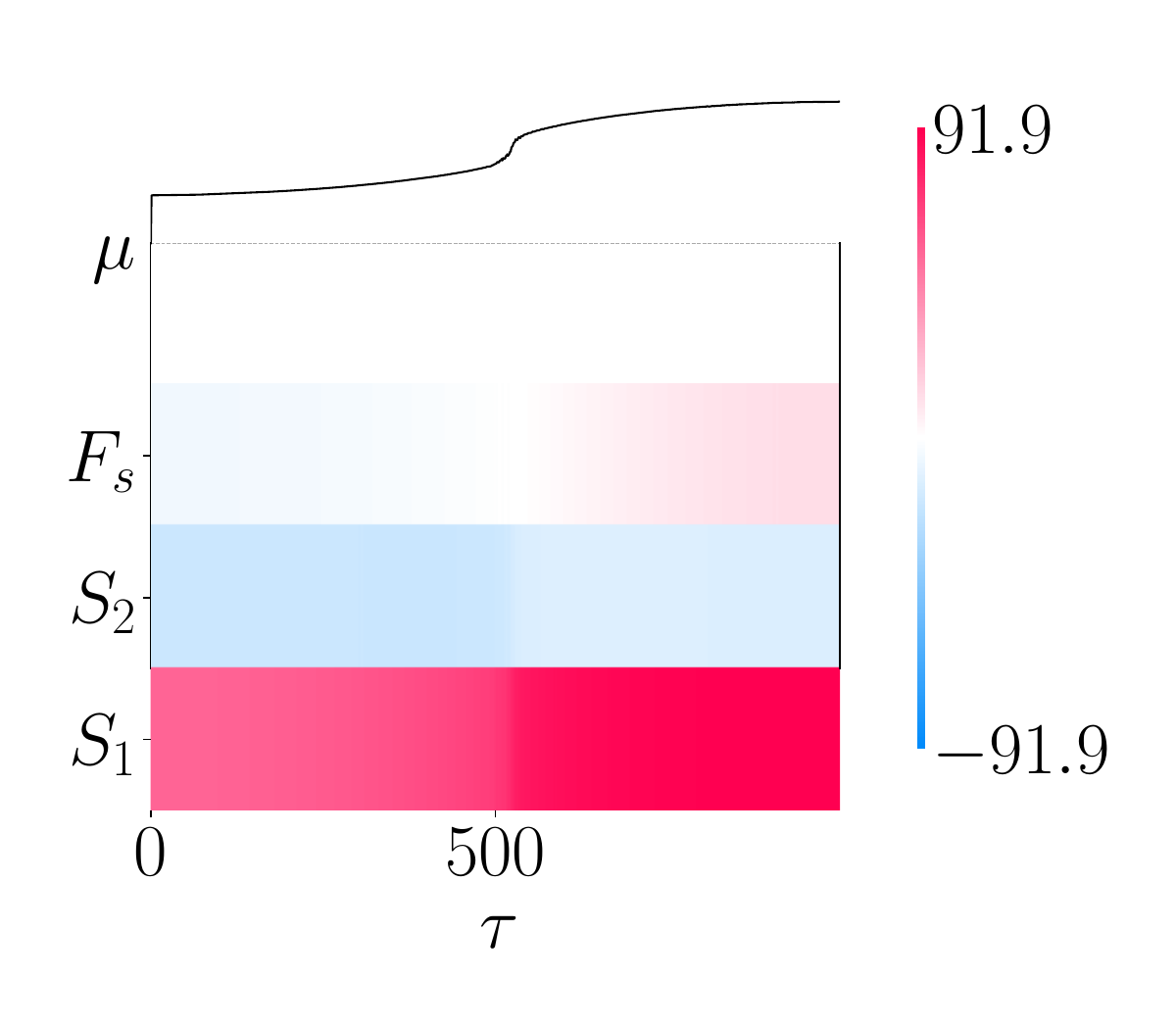}
	& \includegraphics[width=\linewidth,valign=m]{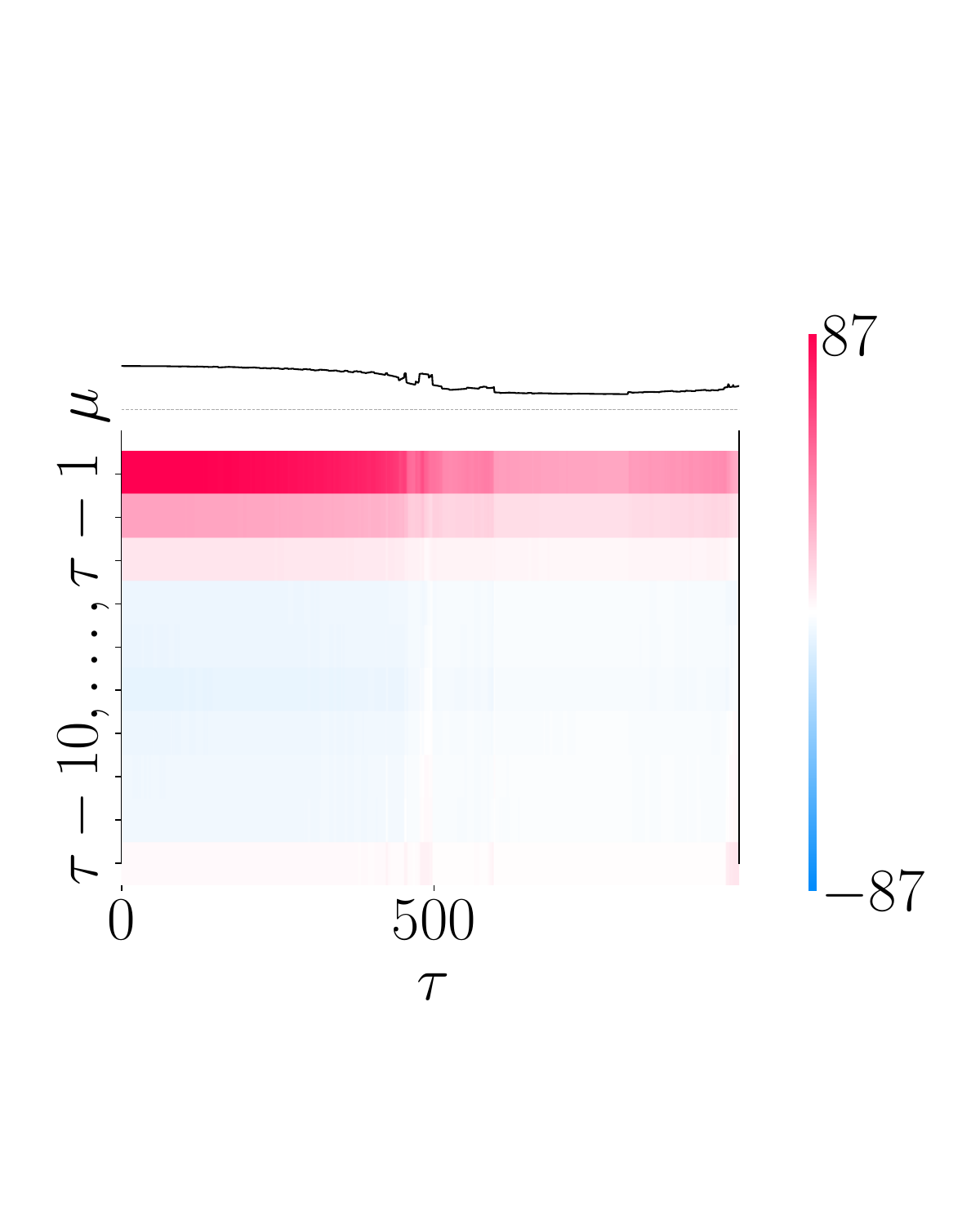}
	& \includegraphics[width=\linewidth,valign=m]{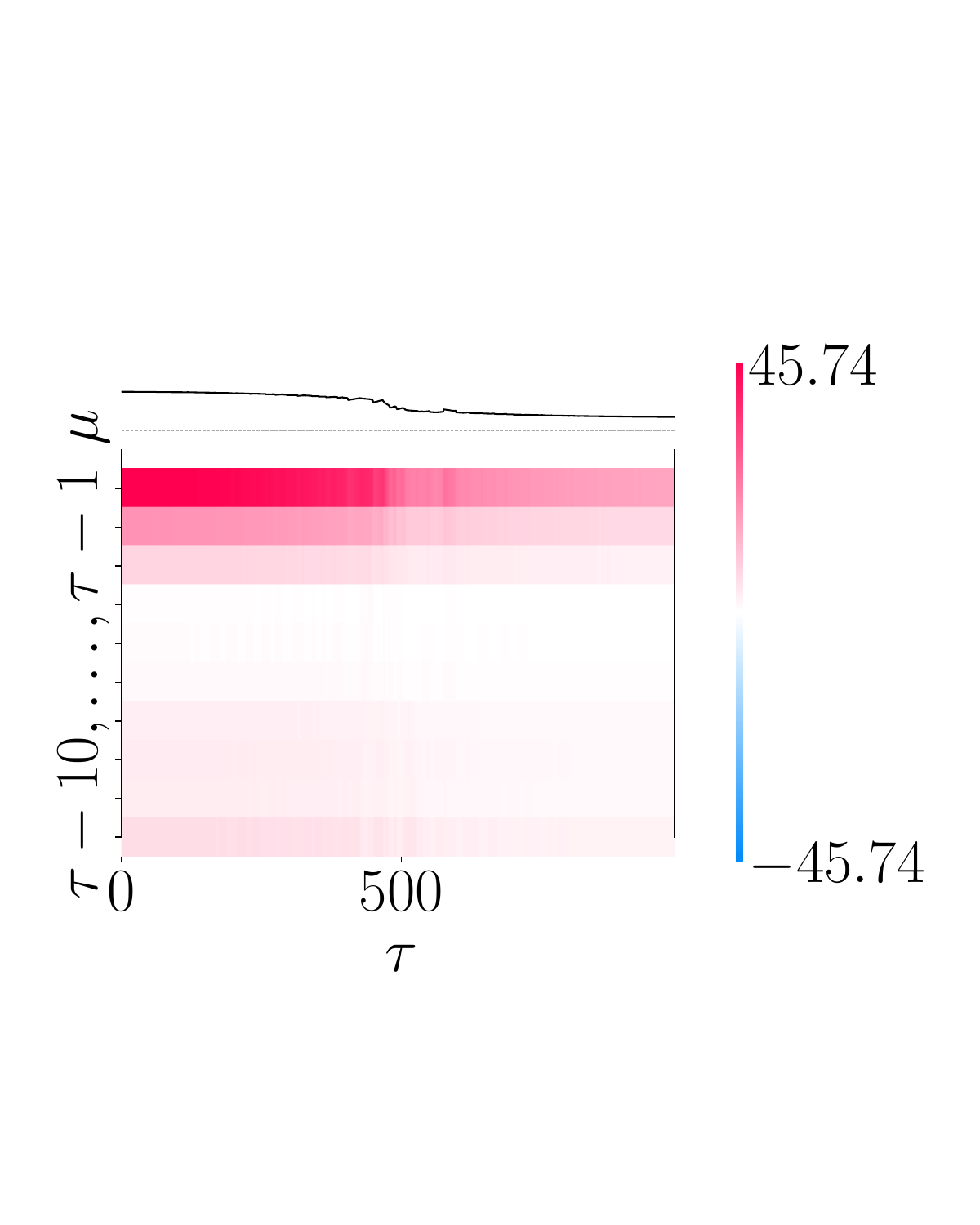} \\
	& RNN
	& ---
	& ---
	& \includegraphics[width=\linewidth,valign=m]{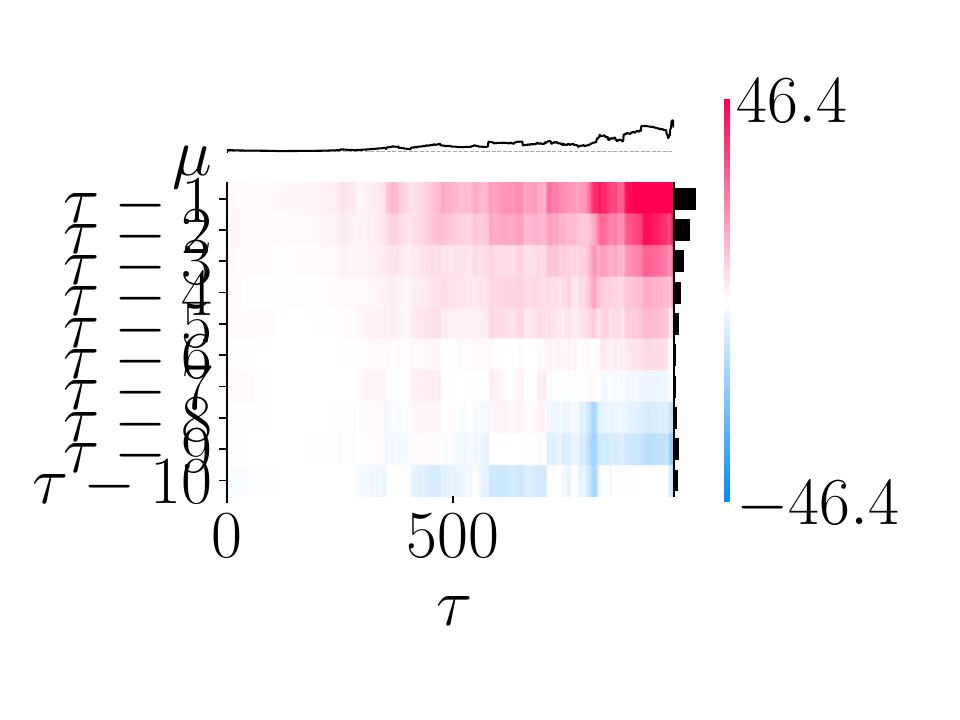}
	& \includegraphics[width=\linewidth,valign=m]{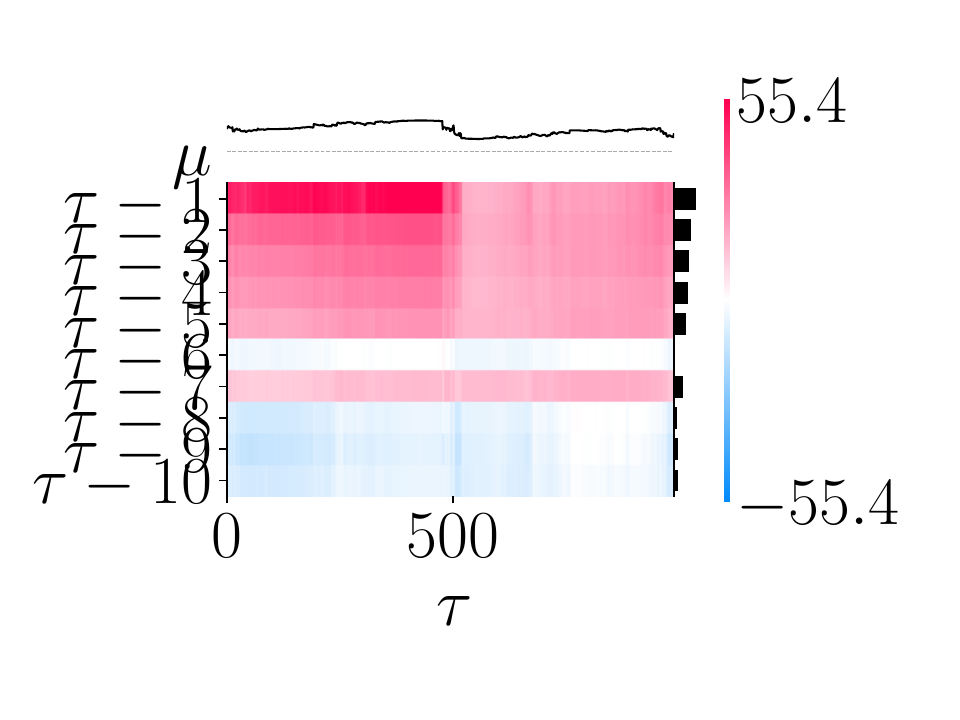} \\

	\midrule

	\multirow[c]{3}{*}{\rotatebox[origin=c]{90}{\(F_s\): Sinusoidal (nonstationary)}}
	& BNN
	& \includegraphics[width=\linewidth,valign=m]{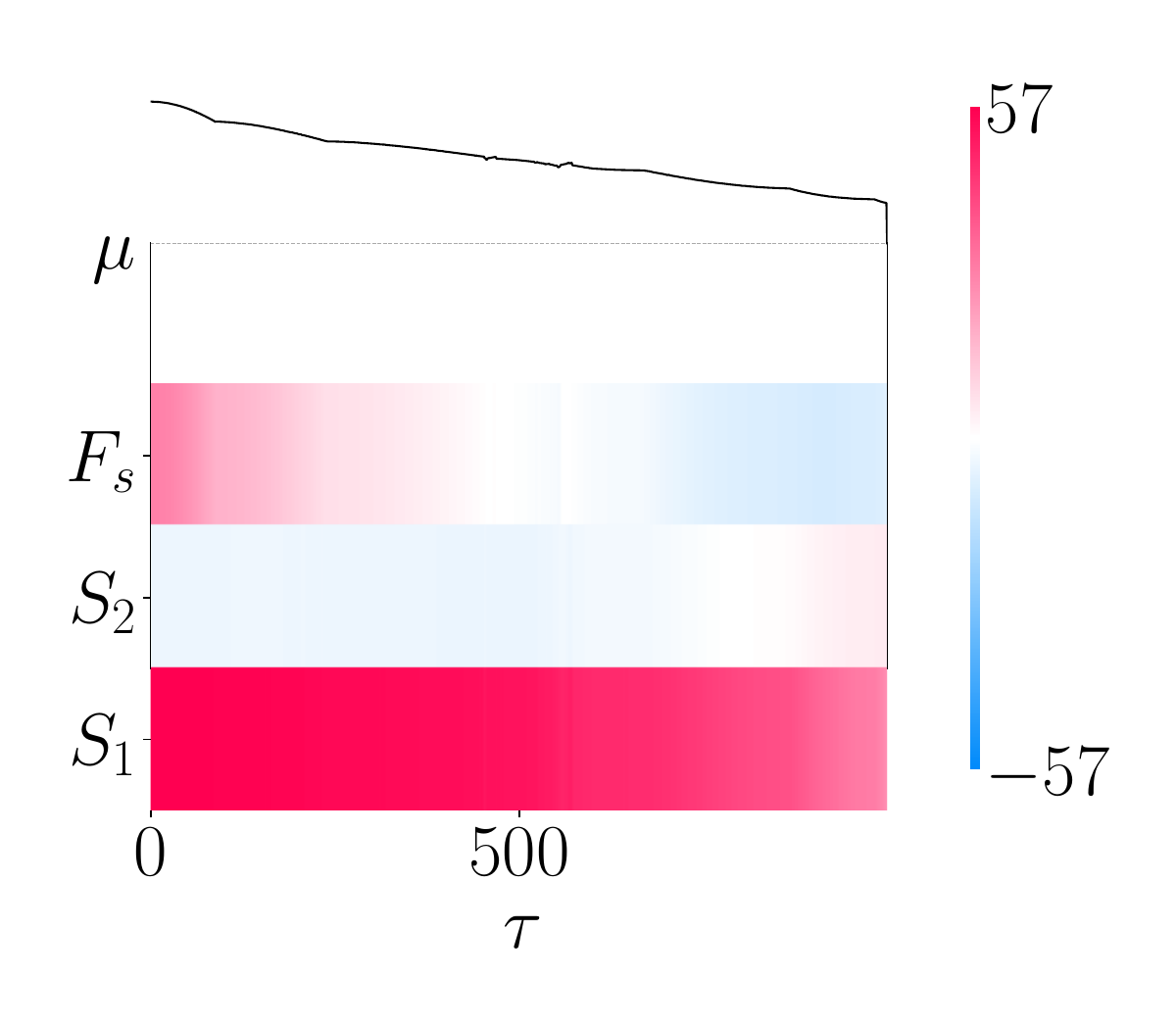}
	& \includegraphics[width=\linewidth,valign=m]{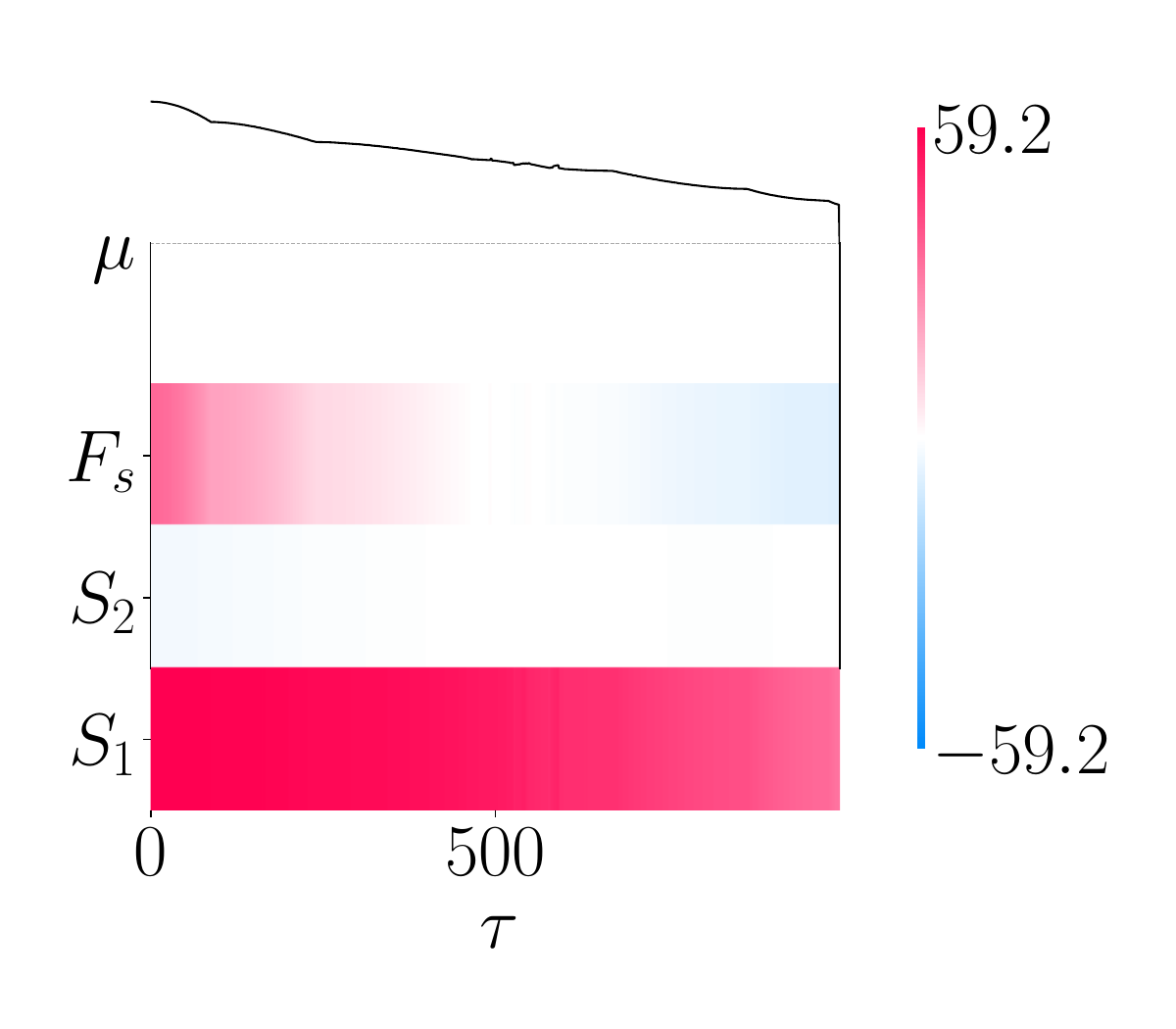}
	& \includegraphics[width=\linewidth,valign=m]{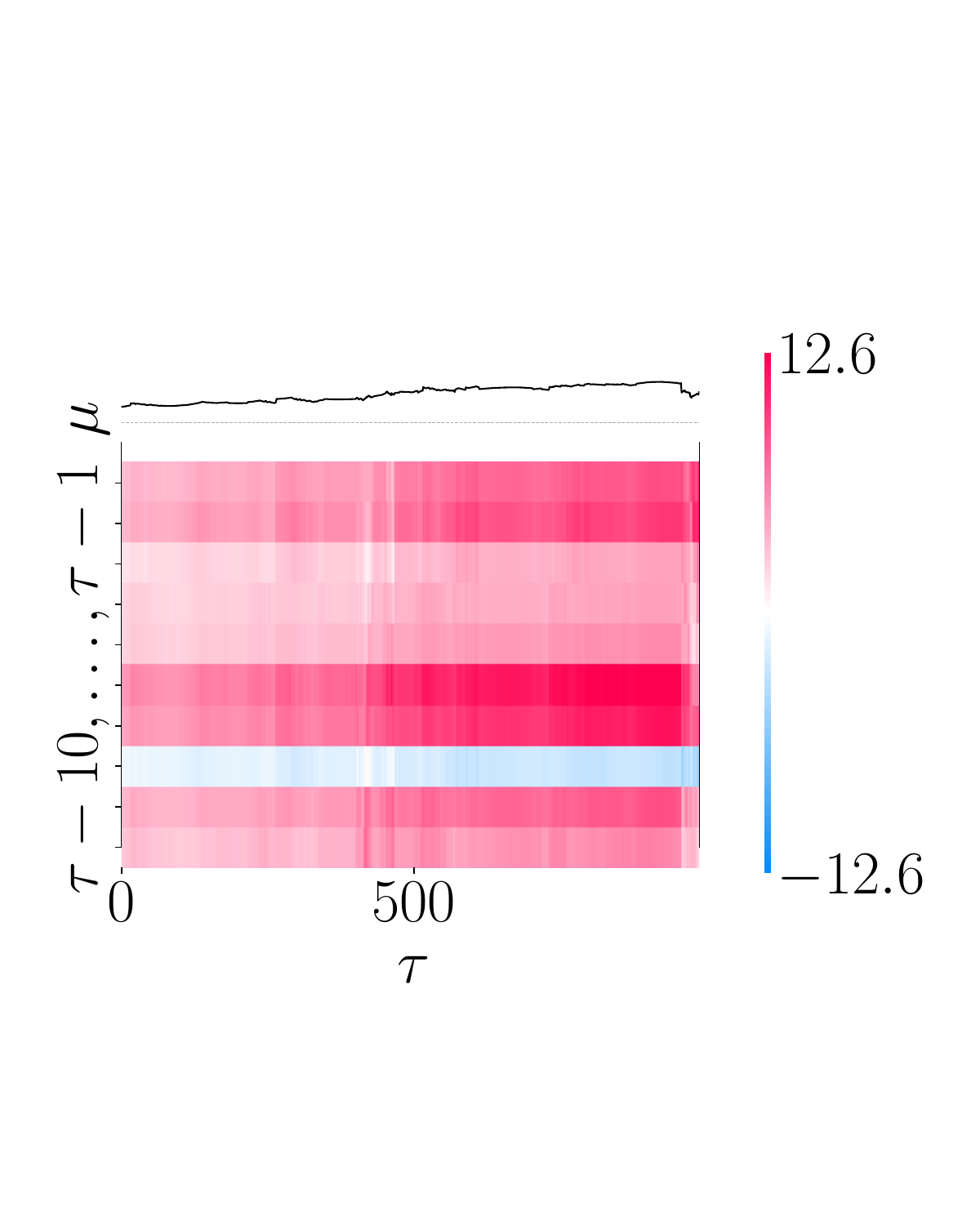}
	& \includegraphics[width=\linewidth,valign=m]{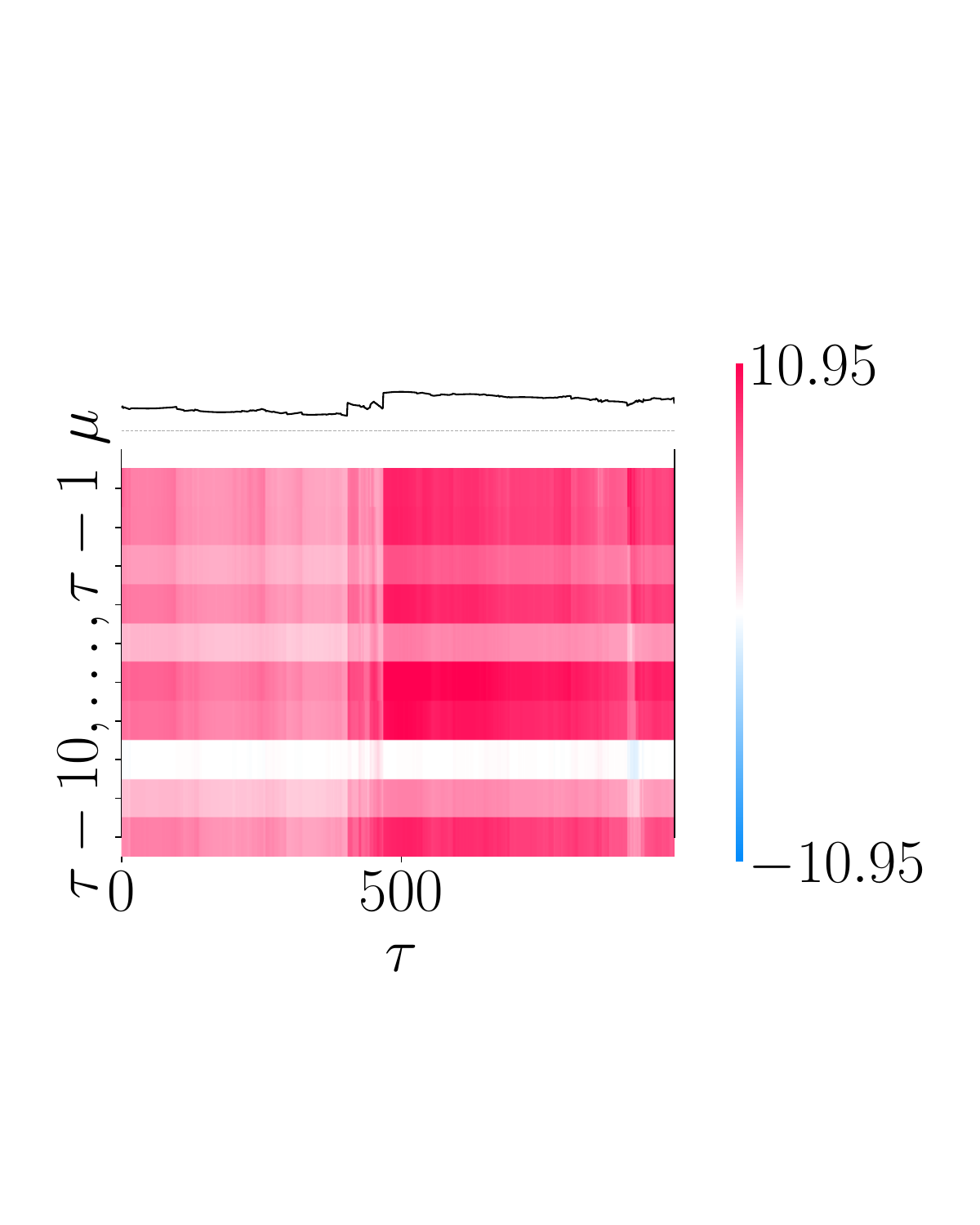} \\
	& MLP
	& \includegraphics[width=\linewidth,valign=m]{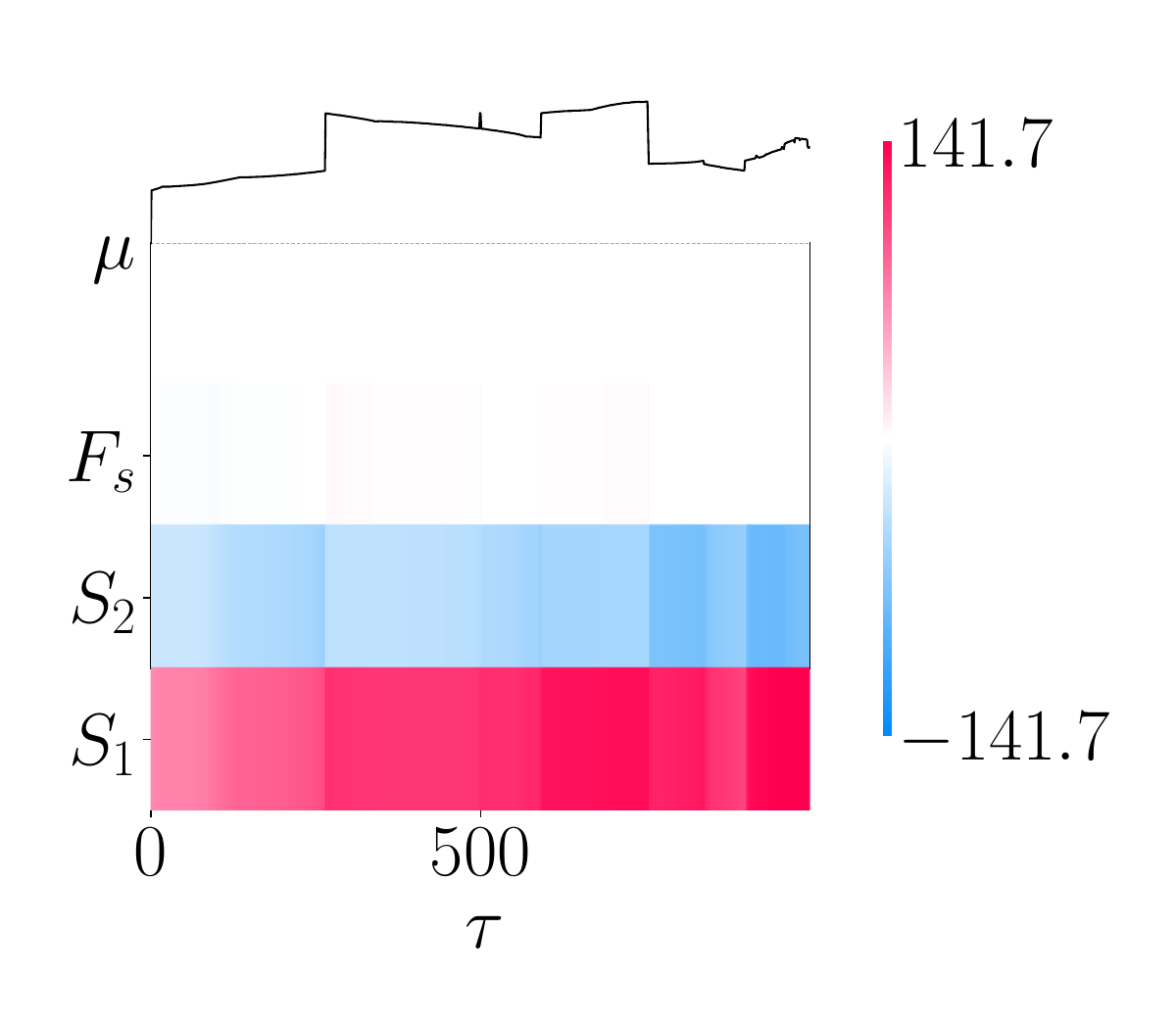}
	& \includegraphics[width=\linewidth,valign=m]{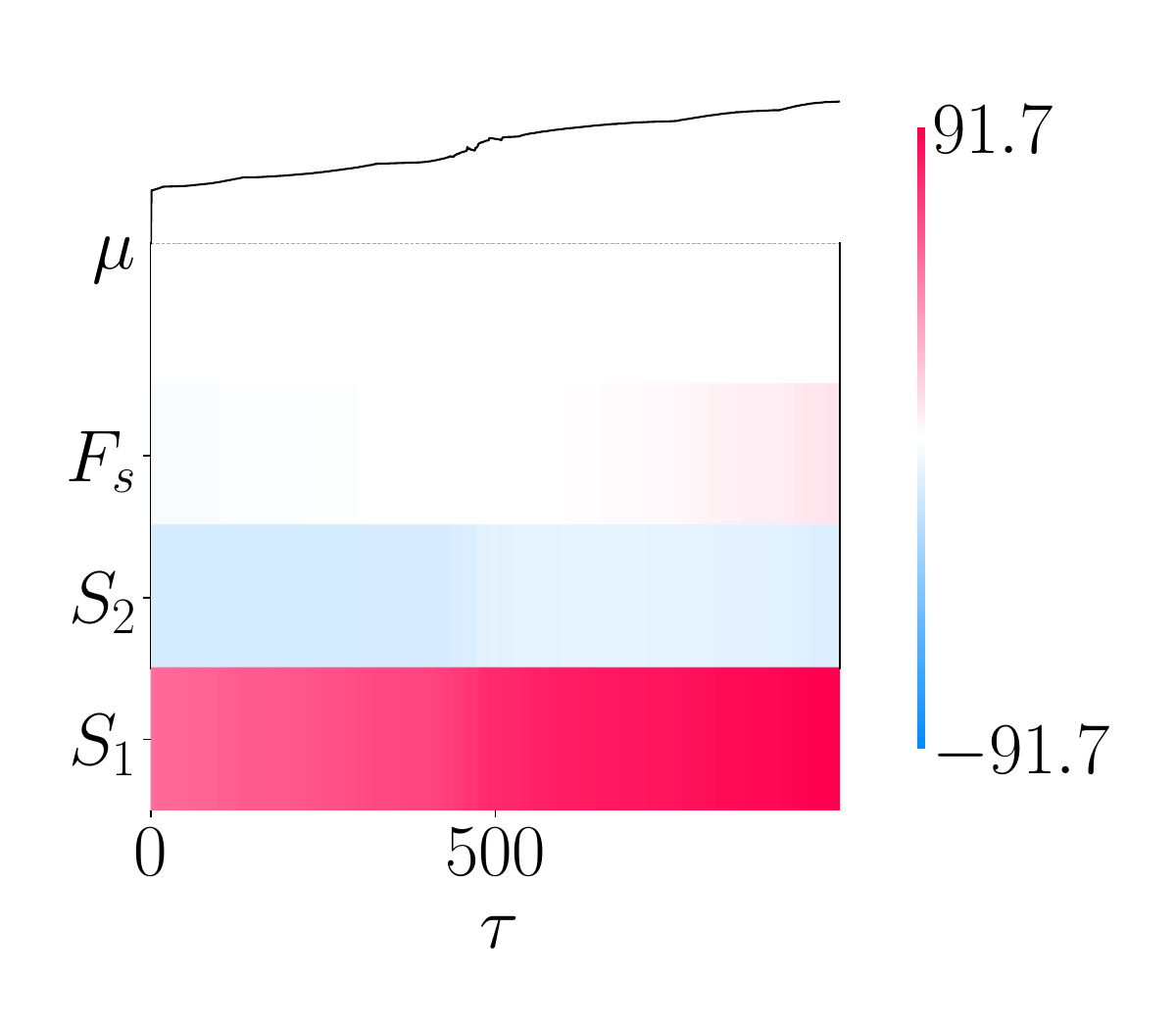}
	& \includegraphics[width=\linewidth,valign=m]{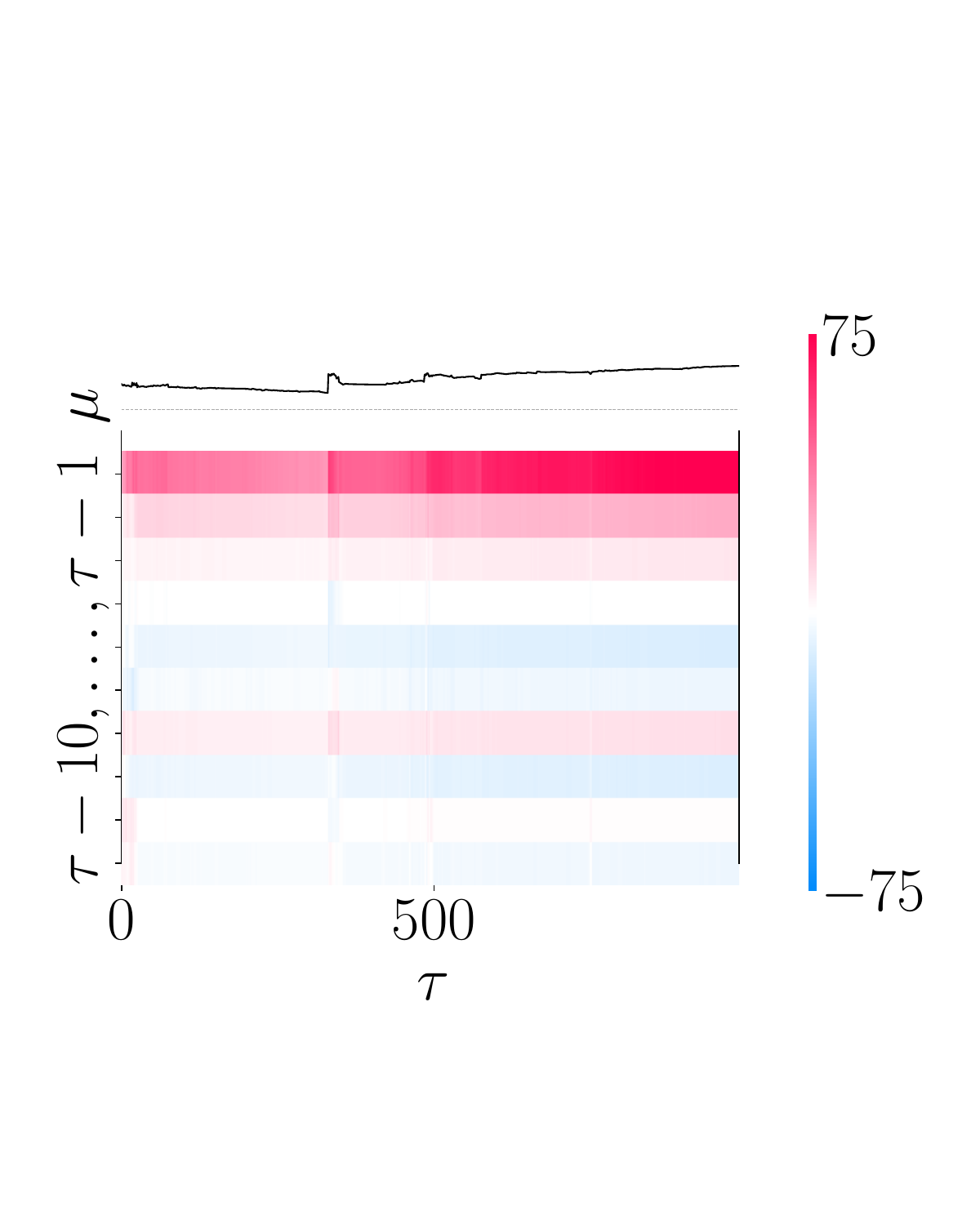}
	& \includegraphics[width=\linewidth,valign=m]{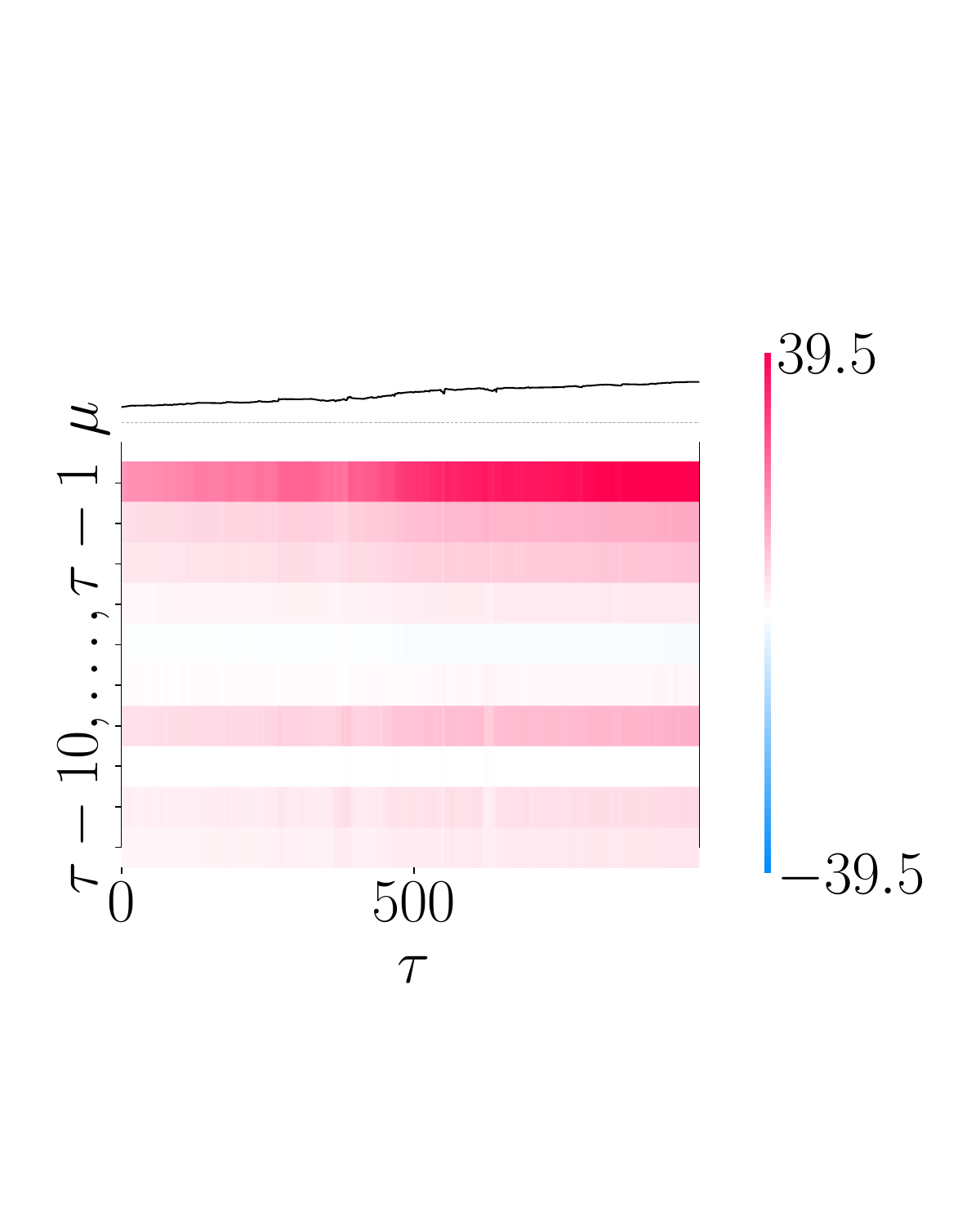} \\
	& Deep Ensemble
	& \includegraphics[width=\linewidth,valign=m]{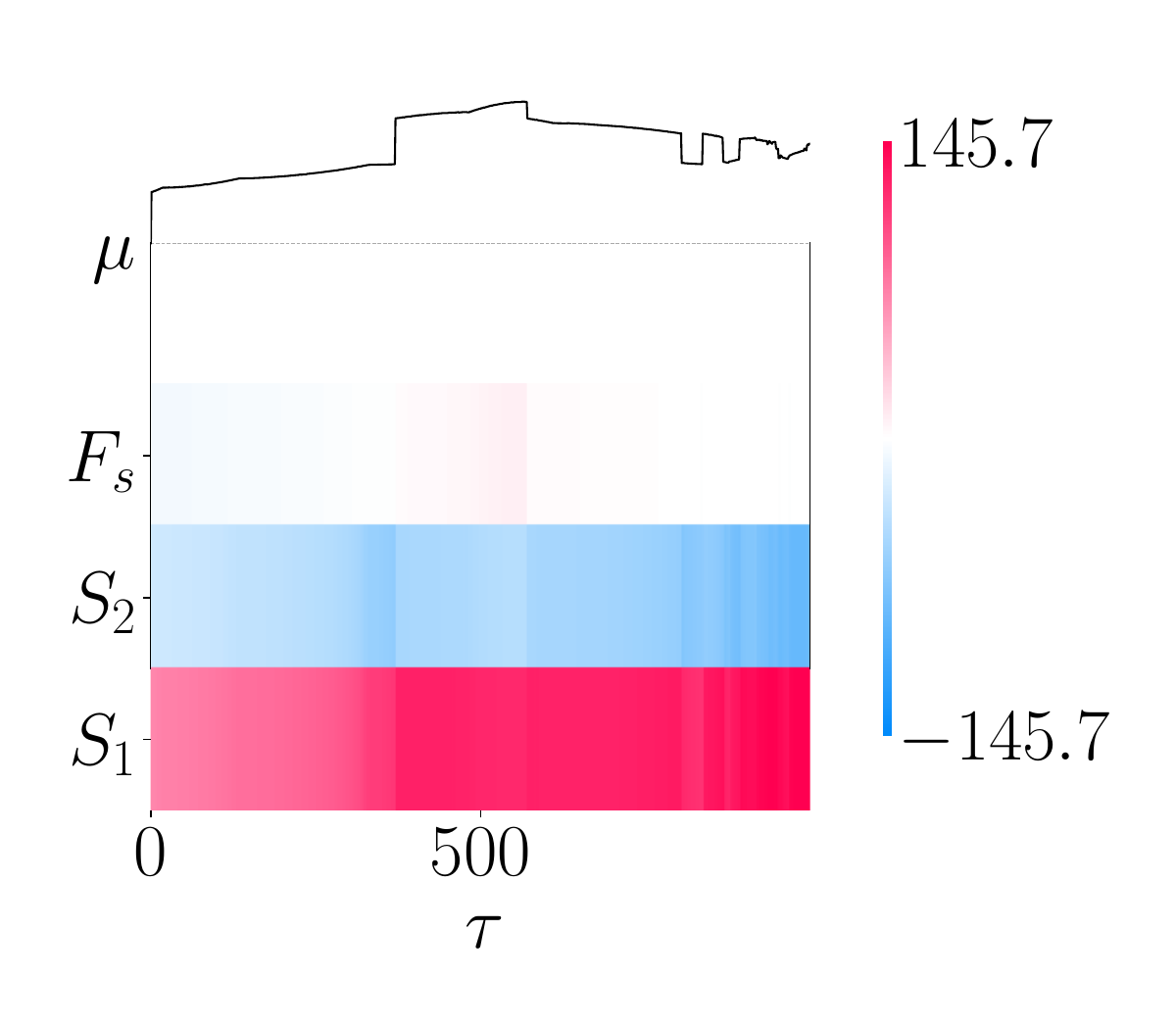}
	& \includegraphics[width=\linewidth,valign=m]{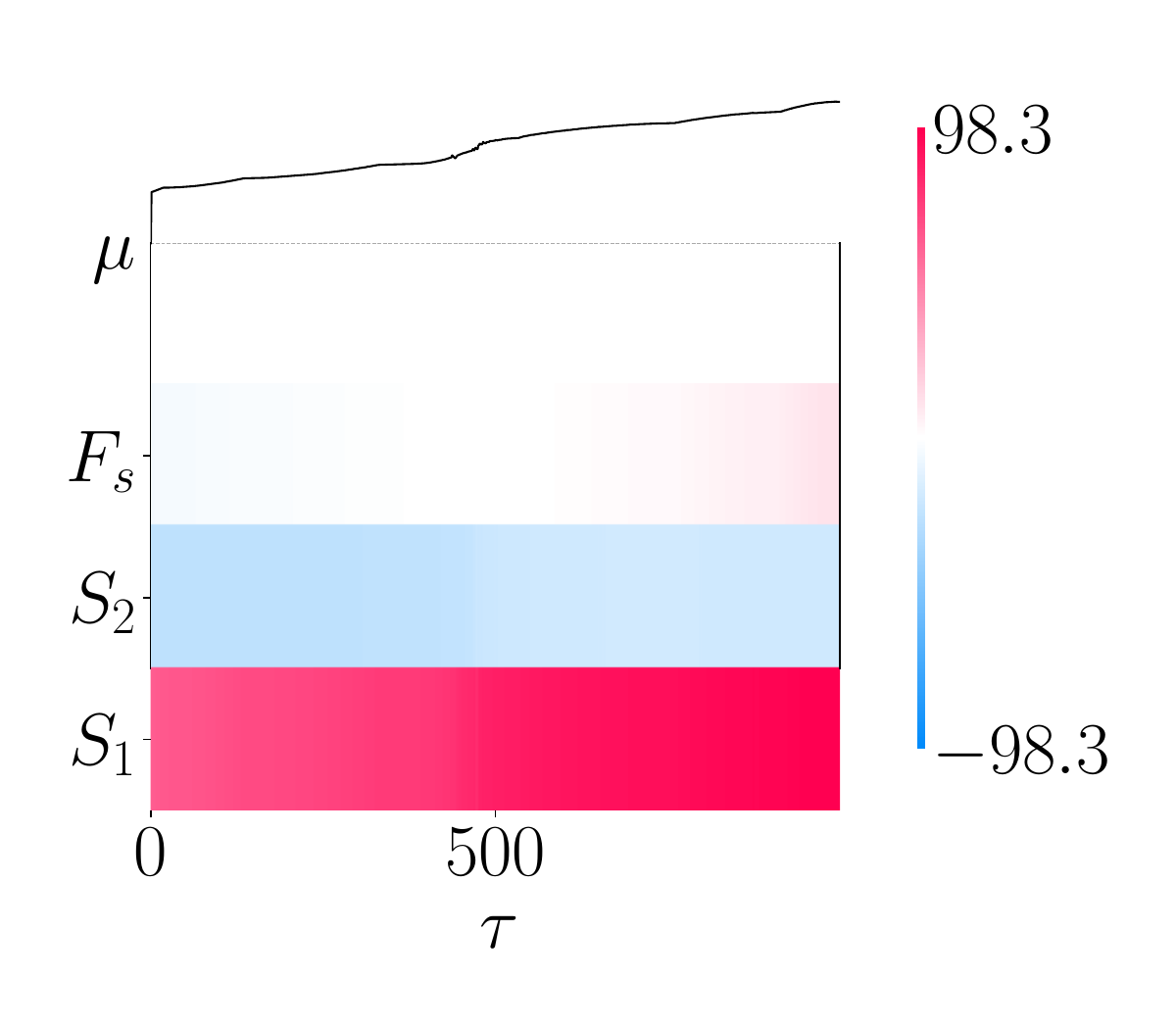}
	& \includegraphics[width=\linewidth,valign=m]{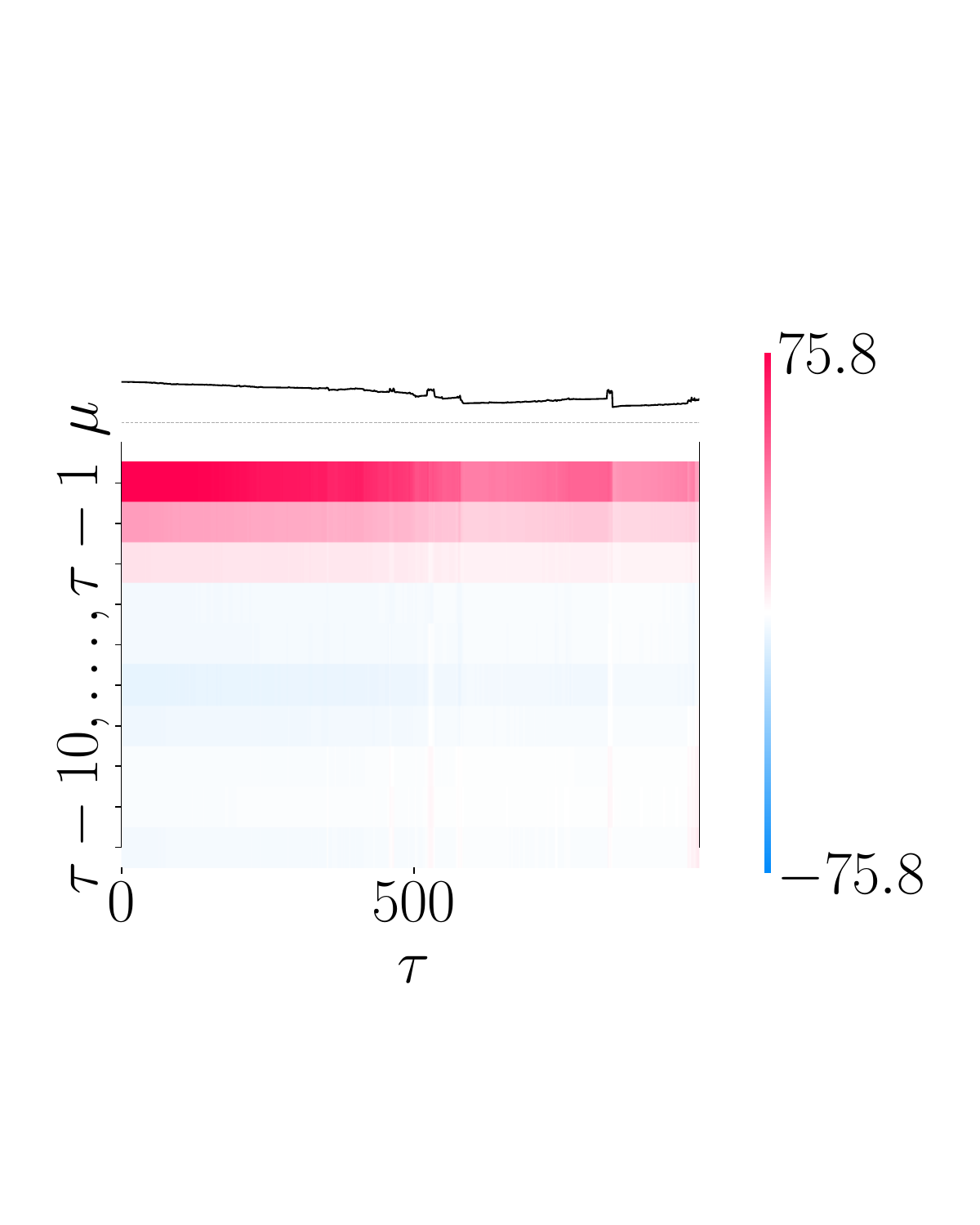}
	& \includegraphics[width=\linewidth,valign=m]{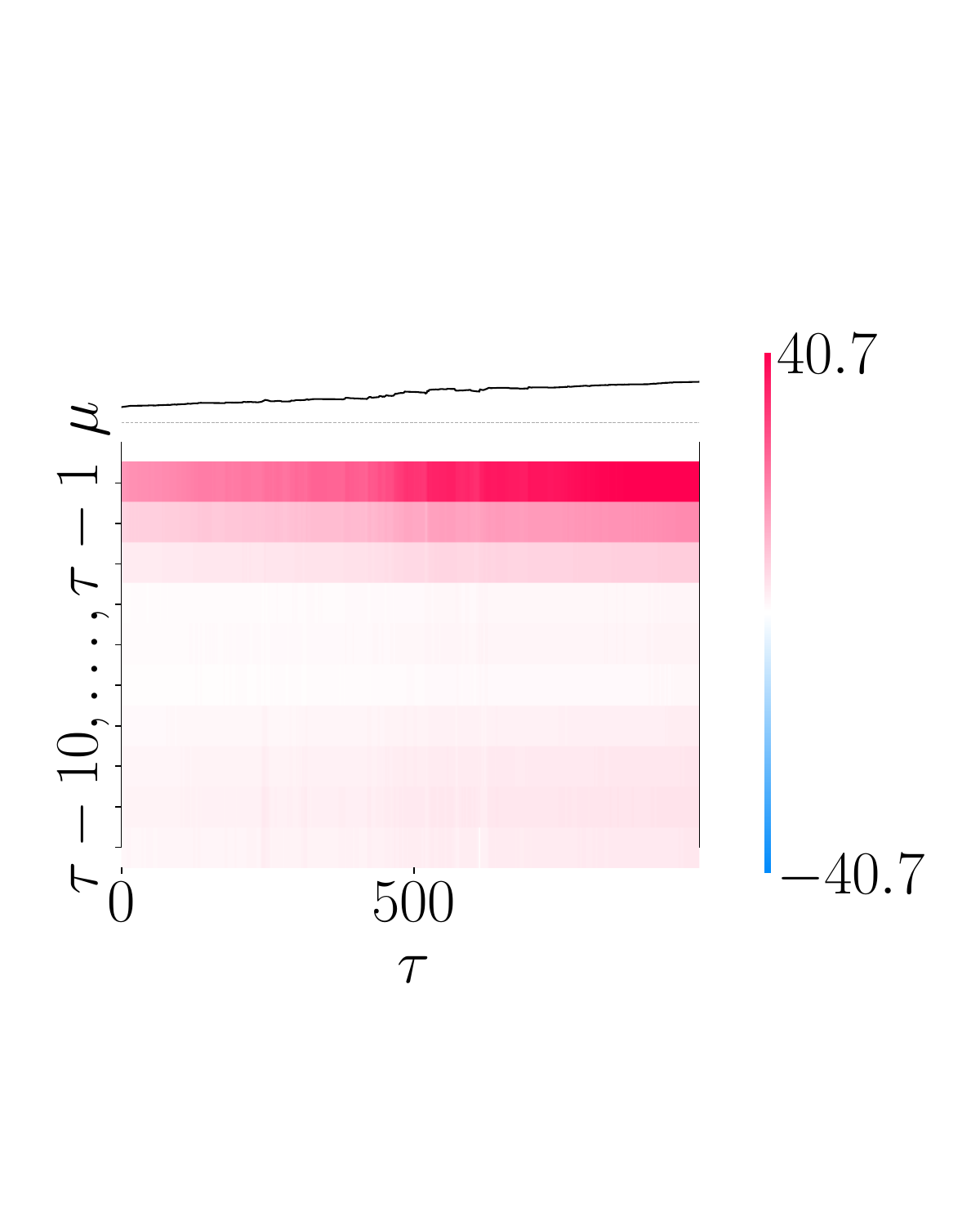} \\
	& RNN
	& ---
	& ---
	& \includegraphics[width=\linewidth,valign=m]{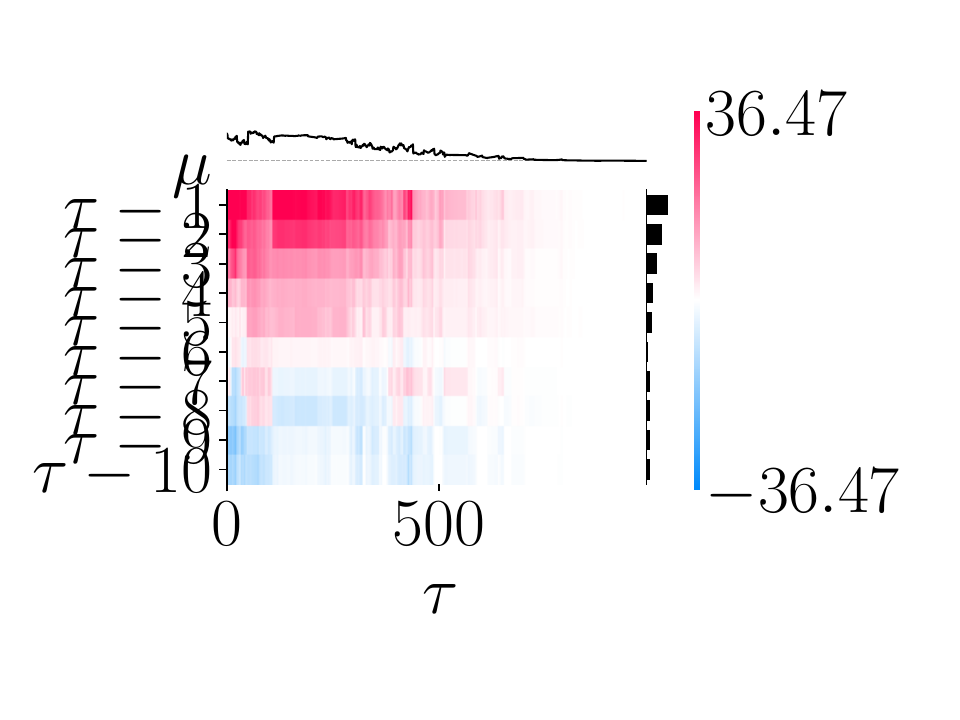}
	& \includegraphics[width=\linewidth,valign=m]{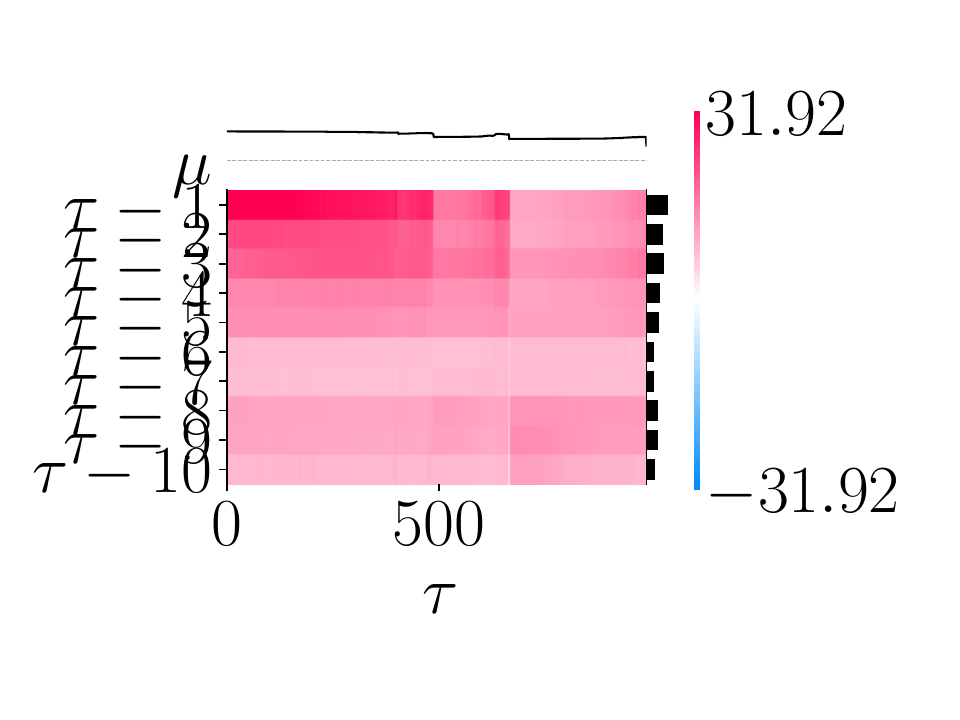} \\

	\bottomrule
	\caption{Attribution maps using \(F_s\) and \(\rho^{\text{EOS-80}}\).}%
	\label{tab:attributions-standard-nonlinear}%
\end{longtable}

\section{BNN Prior Standard Deviation}%
\label{sec:bnn-prior}

We conduct additional experiments using the BNN architecture and varying
values of the prior standard deviation \(\sigma\). We set \(\sigma = 1^{-2}, 1^{-3}, 1^{-4}, 1^{-5}, 1^{-6}\).
Note that the value in the main experiments was \(\sigma = 0.1\).


\begin{figure}%
\begin{tcbitemize}[raster equal height=rows, raster columns=3, raster halign=center, raster every box/.style=blankest]%
\tcbitem[raster multicolumn=3,boxed title style={center},halign=center]\includegraphics[width=0.33\textwidth]{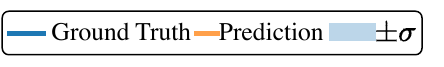}%
\tcbitem\subfloat[AR, \(\sigma = 1^{-2}\)]{\includegraphics[width=\linewidth,keepaspectratio]{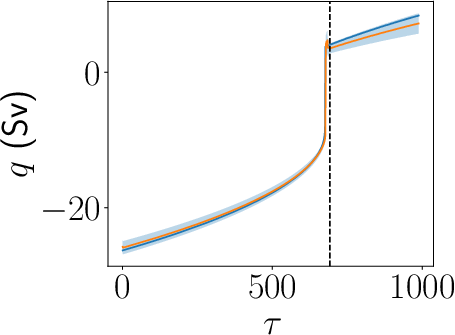}}%
\tcbitem\subfloat[AR, \(\sigma = 1^{-3}\)]{\includegraphics[width=\linewidth,keepaspectratio]{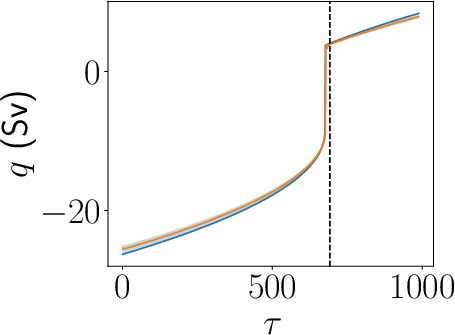}}%
\tcbitem\subfloat[AR, \(\sigma = 1^{-6}\)]{\includegraphics[width=\linewidth,keepaspectratio]{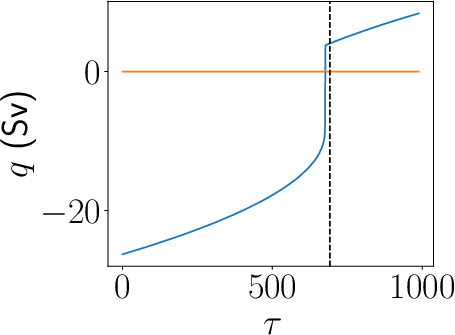}}%
\tcbitem\subfloat[AR, \(\sigma = 1^{-5}\)]{\includegraphics[width=\linewidth,keepaspectratio]{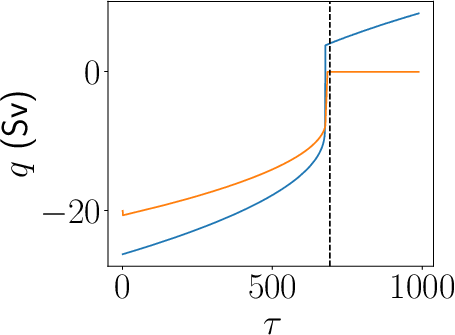}}%
\tcbitem\subfloat[AR, \(\sigma = 1^{-4}\)]{\includegraphics[width=\linewidth,keepaspectratio]{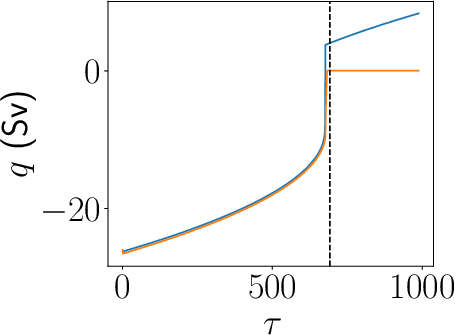}}%
\tcbitem\subfloat[PI, \(\sigma = 1^{-2}\)]{\includegraphics[width=\linewidth,keepaspectratio]{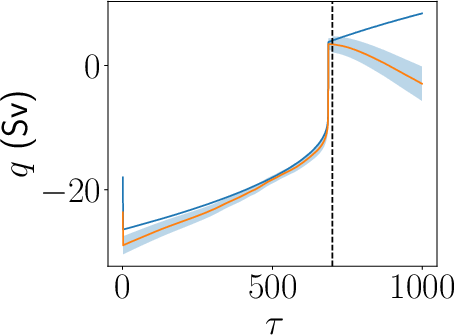}}%
\tcbitem\subfloat[PI, \(\sigma = 1^{-3}\)]{\includegraphics[width=\linewidth,keepaspectratio]{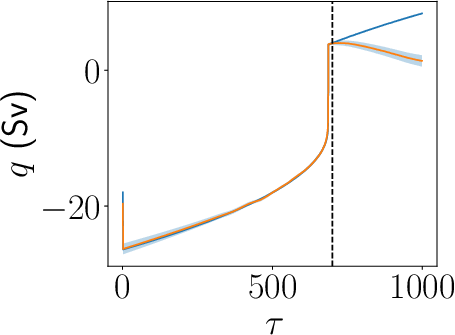}}%
\tcbitem\subfloat[PI, \(\sigma = 1^{-6}\)]{\includegraphics[width=\linewidth,keepaspectratio]{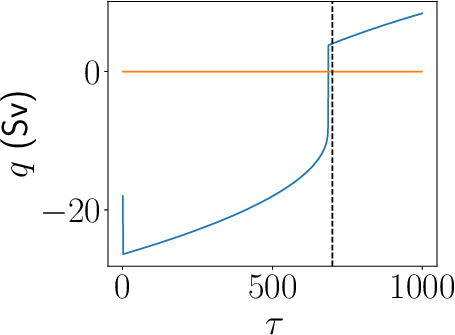}}%
\tcbitem\subfloat[PI, \(\sigma = 1^{-5}\)]{\includegraphics[width=\linewidth,keepaspectratio]{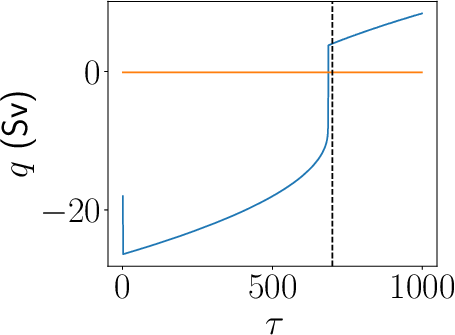}}%
\tcbitem\subfloat[PI, \(\sigma = 1^{-4}\)]{\includegraphics[width=\linewidth,keepaspectratio]{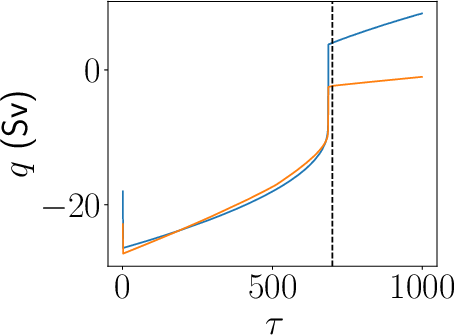}}%
\end{tcbitemize}%
\caption{BNN predictions under varying \(\sigma\) for \(\mathcal{F}_1\).}%
\label{fig:pgt-1}%
\end{figure}

%

\begin{figure}[h]%
\begin{tcbitemize}[raster equal height=rows, raster columns=4, raster halign=center, raster every box/.style=blankest]%
\tcbitem\subfloat[SHAP, AR, \(\sigma = 1^{-2}\)]{\includegraphics[width=\linewidth,keepaspectratio]{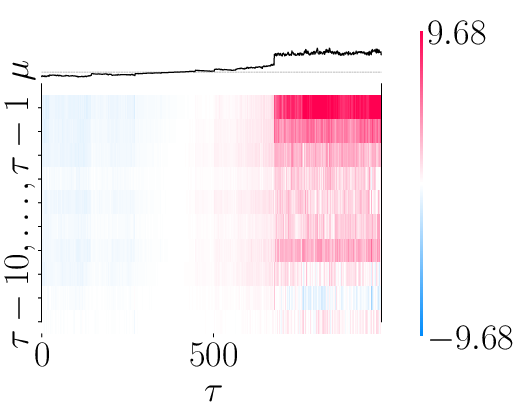}}%
\tcbitem\subfloat[DeepLIFT, AR, \(\sigma = 1^{-2}\)]{\includegraphics[width=\linewidth,keepaspectratio]{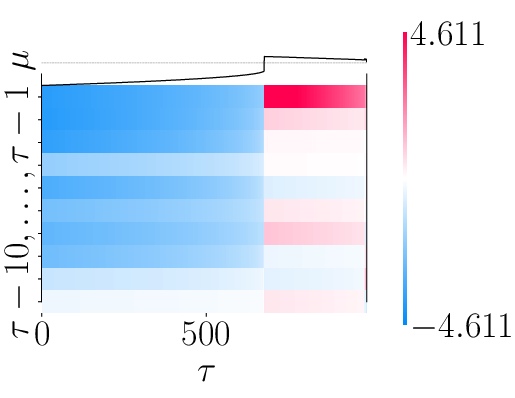}}%
\tcbitem\subfloat[SHAP, AR, \(\sigma = 1^{-3}\)]{\includegraphics[width=\linewidth,keepaspectratio]{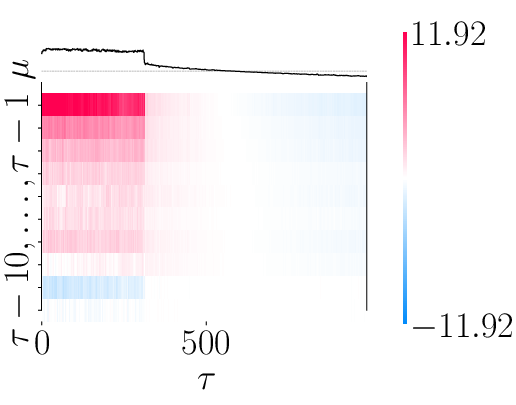}}%
\tcbitem\subfloat[DeepLIFT, AR, \(\sigma = 1^{-3}\)]{\includegraphics[width=\linewidth,keepaspectratio]{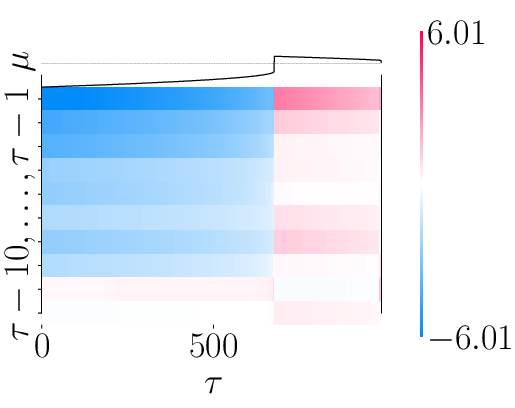}}%
\tcbitem\subfloat[SHAP, AR, \(\sigma = 1^{-6}\)]{\includegraphics[width=\linewidth,keepaspectratio]{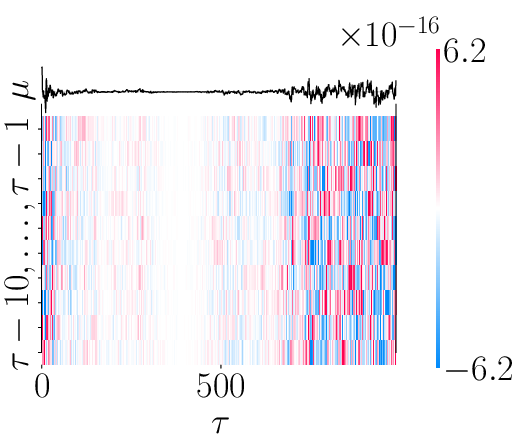}}%
\tcbitem\subfloat[DeepLIFT, AR, \(\sigma = 1^{-6}\)]{\includegraphics[width=\linewidth,keepaspectratio]{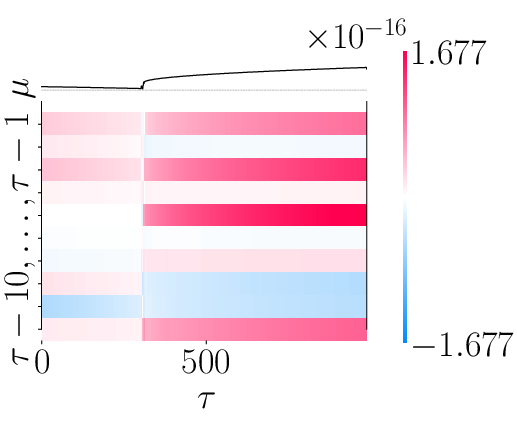}}%
\tcbitem\subfloat[SHAP, AR, \(\sigma = 1^{-5}\)]{\includegraphics[width=\linewidth,keepaspectratio]{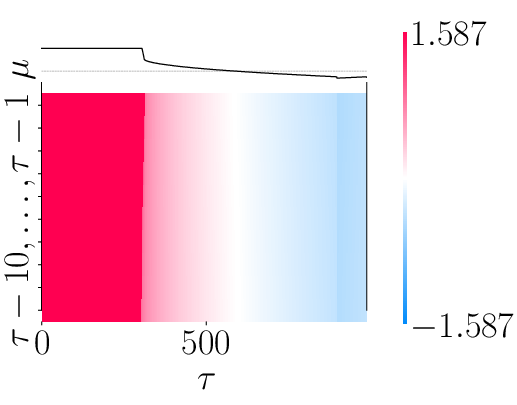}}%
\tcbitem\subfloat[DeepLIFT, AR, \(\sigma = 1^{-5}\)]{\includegraphics[width=\linewidth,keepaspectratio]{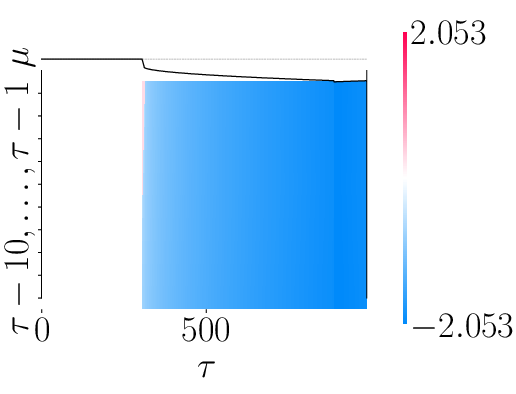}}%
\tcbitem\subfloat[SHAP, AR, \(\sigma = 1^{-4}\)]{\includegraphics[width=\linewidth,keepaspectratio]{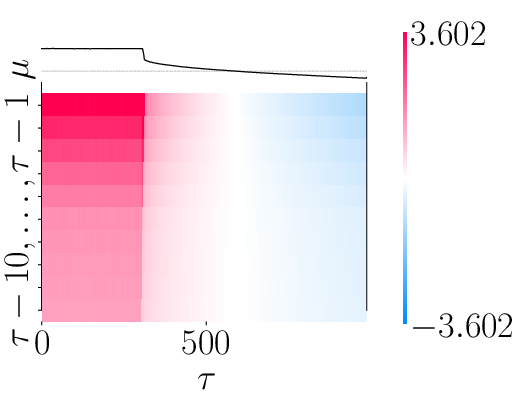}}%
\tcbitem\subfloat[DeepLIFT, AR, \(\sigma = 1^{-4}\)]{\includegraphics[width=\linewidth,keepaspectratio]{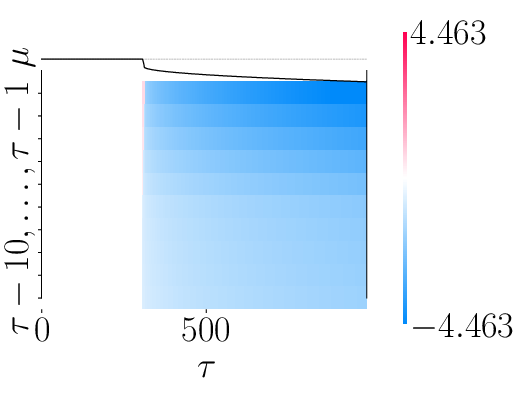}}%
\tcbitem\subfloat[SHAP, PI, \(\sigma = 1^{-2}\)]{\includegraphics[width=\linewidth,keepaspectratio]{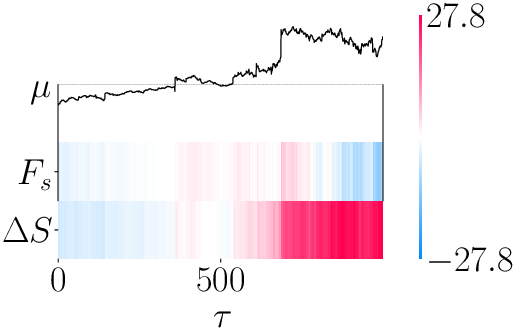}}%
\tcbitem\subfloat[DeepLIFT, PI, \(\sigma = 1^{-2}\)]{\includegraphics[width=\linewidth,keepaspectratio]{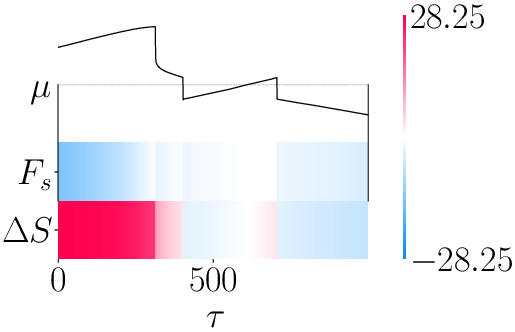}}%
\tcbitem\subfloat[SHAP, PI, \(\sigma = 1^{-3}\)]{\includegraphics[width=\linewidth,keepaspectratio]{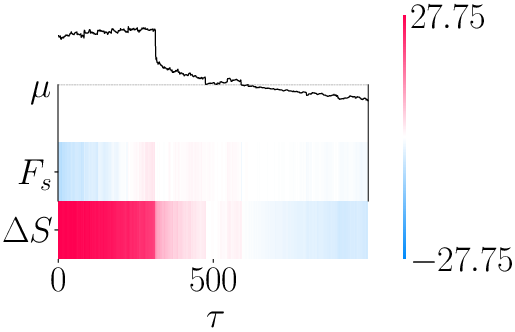}}%
\tcbitem\subfloat[DeepLIFT, PI, \(\sigma = 1^{-3}\)]{\includegraphics[width=\linewidth,keepaspectratio]{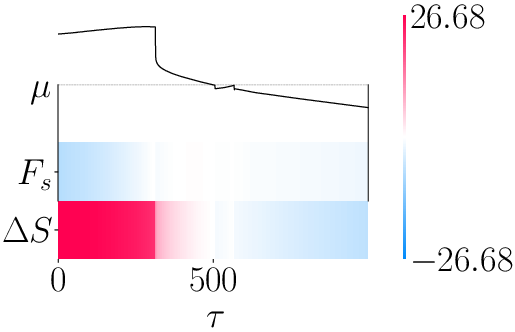}}%
\tcbitem\subfloat[SHAP, PI, \(\sigma = 1^{-6}\)]{\includegraphics[width=\linewidth,keepaspectratio]{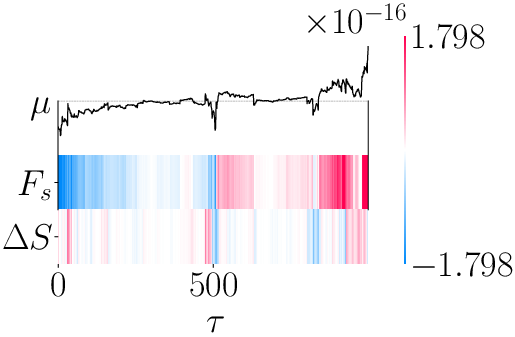}}%
\tcbitem\subfloat[DeepLIFT, PI, \(\sigma = 1^{-6}\)]{\includegraphics[width=\linewidth,keepaspectratio]{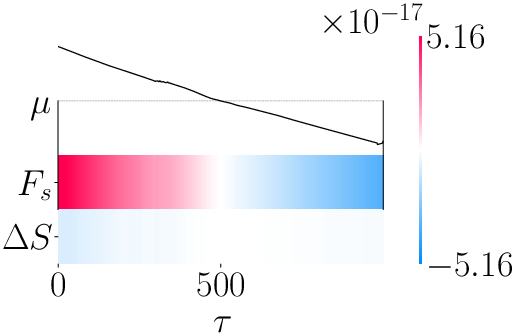}}%
\tcbitem\subfloat[SHAP, PI, \(\sigma = 1^{-5}\)]{\includegraphics[width=\linewidth,keepaspectratio]{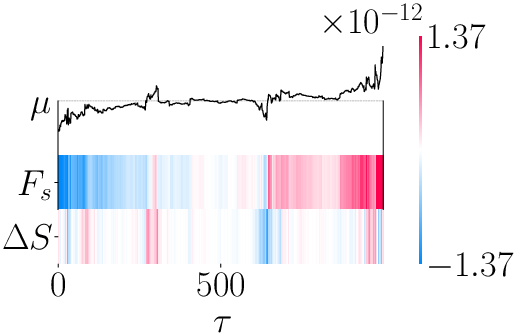}}%
\tcbitem\subfloat[DeepLIFT, PI, \(\sigma = 1^{-5}\)]{\includegraphics[width=\linewidth,keepaspectratio]{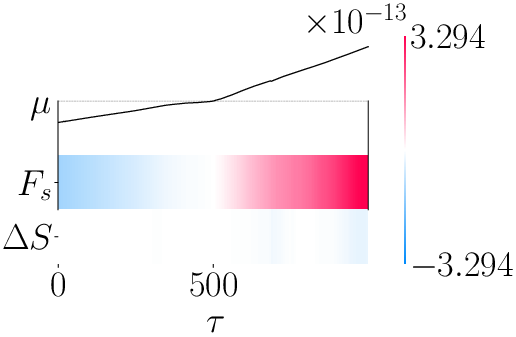}}%
\tcbitem\subfloat[SHAP, PI, \(\sigma = 1^{-4}\)]{\includegraphics[width=\linewidth,keepaspectratio]{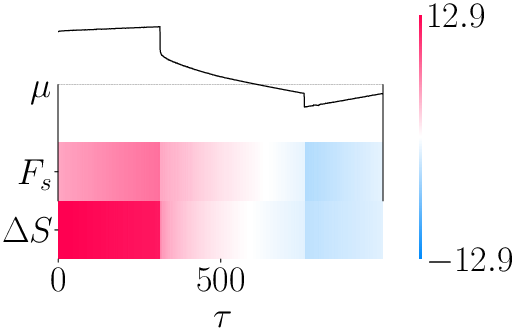}}%
\tcbitem\subfloat[DeepLIFT, PI, \(\sigma = 1^{-4}\)]{\includegraphics[width=\linewidth,keepaspectratio]{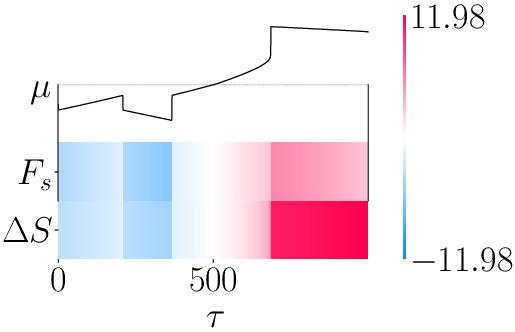}}%
\end{tcbitemize}%
\caption{BNN attributions under varying \(\sigma\) for \(\mathcal{F}_1\).}%
\label{fig:xai-1}%
\end{figure}


\begin{figure}%
\begin{tcbitemize}[raster equal height=rows, raster columns=3, raster halign=center, raster every box/.style=blankest]%
\tcbitem[raster multicolumn=3,boxed title style={center},halign=center]\includegraphics[width=0.33\textwidth]{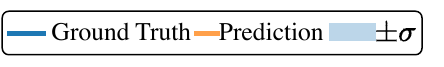}%
\tcbitem\subfloat[AR, \(\sigma = 1^{-2}\)]{\includegraphics[width=\linewidth,keepaspectratio]{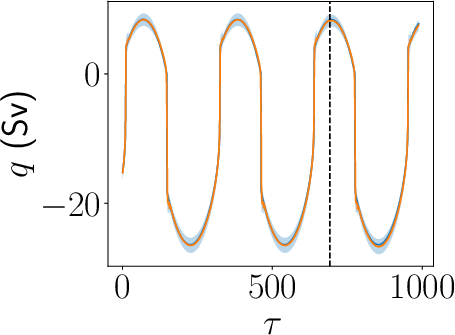}}%
\tcbitem\subfloat[AR, \(\sigma = 1^{-3}\)]{\includegraphics[width=\linewidth,keepaspectratio]{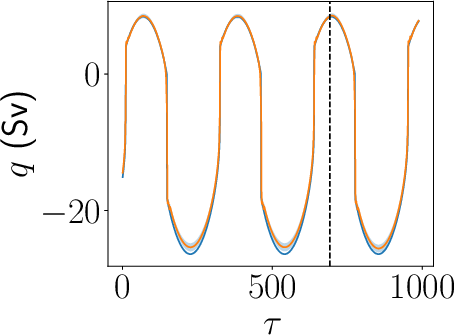}}%
\tcbitem\subfloat[AR, \(\sigma = 1^{-6}\)]{\includegraphics[width=\linewidth,keepaspectratio]{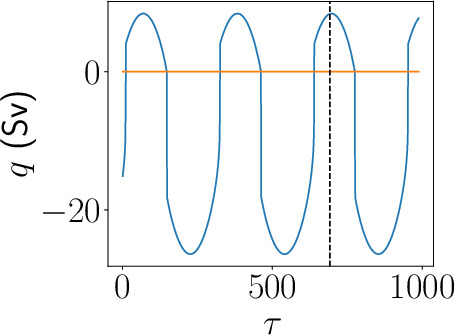}}%
\tcbitem\subfloat[AR, \(\sigma = 1^{-5}\)]{\includegraphics[width=\linewidth,keepaspectratio]{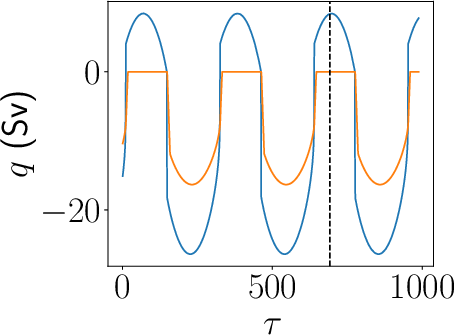}}%
\tcbitem\subfloat[AR, \(\sigma = 1^{-4}\)]{\includegraphics[width=\linewidth,keepaspectratio]{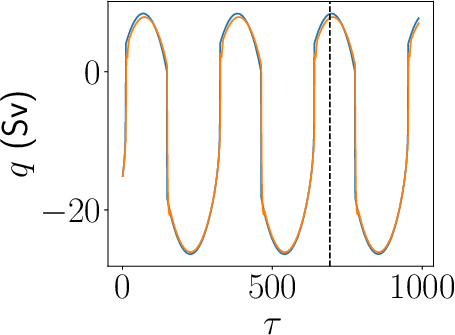}}%
\tcbitem\subfloat[PI, \(\sigma = 1^{-2}\)]{\includegraphics[width=\linewidth,keepaspectratio]{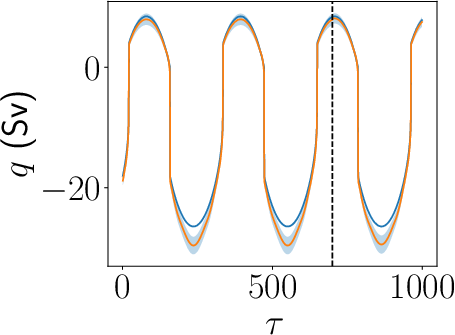}}%
\tcbitem\subfloat[PI, \(\sigma = 1^{-3}\)]{\includegraphics[width=\linewidth,keepaspectratio]{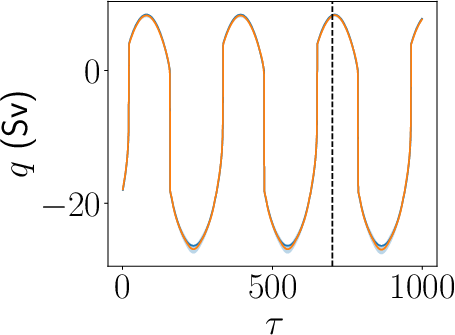}}%
\tcbitem\subfloat[PI, \(\sigma = 1^{-6}\)]{\includegraphics[width=\linewidth,keepaspectratio]{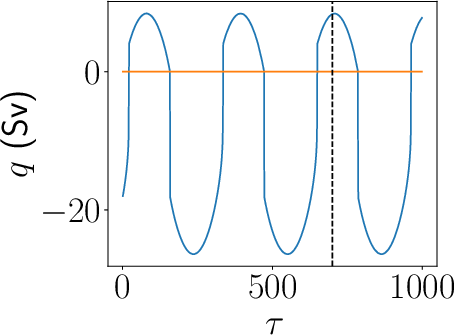}}%
\tcbitem\subfloat[PI, \(\sigma = 1^{-5}\)]{\includegraphics[width=\linewidth,keepaspectratio]{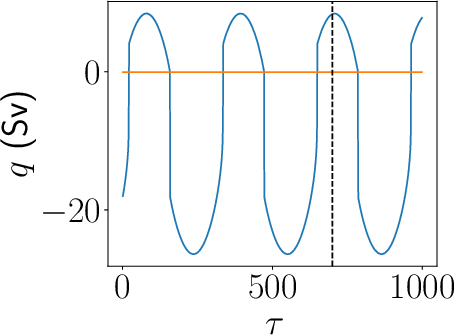}}%
\tcbitem\subfloat[PI, \(\sigma = 1^{-4}\)]{\includegraphics[width=\linewidth,keepaspectratio]{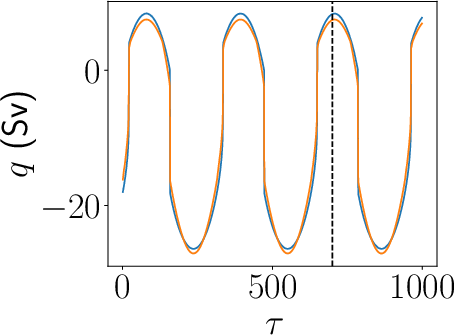}}%
\end{tcbitemize}%
\caption{BNN predictions under varying \(\sigma\) for \(\mathcal{F}_2\).}%
\label{fig:pgt-2}%
\end{figure}

%

\begin{figure}[h]%
\begin{tcbitemize}[raster equal height=rows, raster columns=4, raster halign=center, raster every box/.style=blankest]%
\tcbitem\subfloat[SHAP, AR, \(\sigma = 1^{-2}\)]{\includegraphics[width=\linewidth,keepaspectratio]{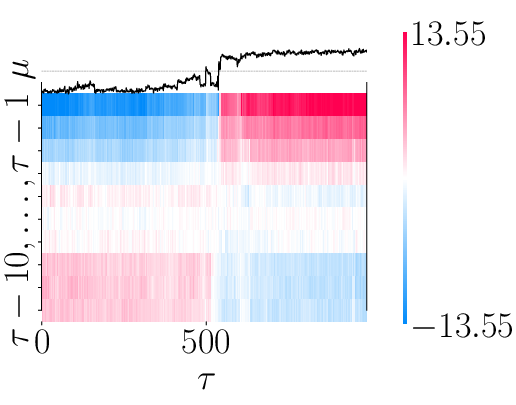}}%
\tcbitem\subfloat[DeepLIFT, AR, \(\sigma = 1^{-2}\)]{\includegraphics[width=\linewidth,keepaspectratio]{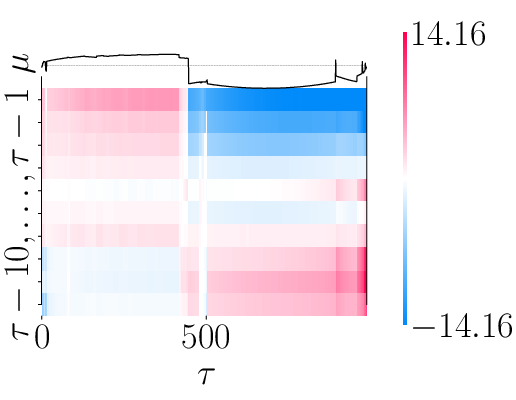}}%
\tcbitem\subfloat[SHAP, AR, \(\sigma = 1^{-3}\)]{\includegraphics[width=\linewidth,keepaspectratio]{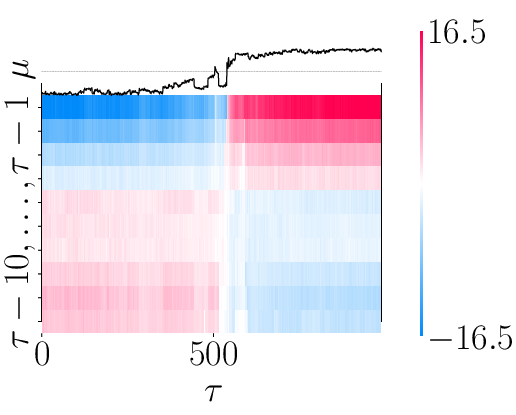}}%
\tcbitem\subfloat[DeepLIFT, AR, \(\sigma = 1^{-3}\)]{\includegraphics[width=\linewidth,keepaspectratio]{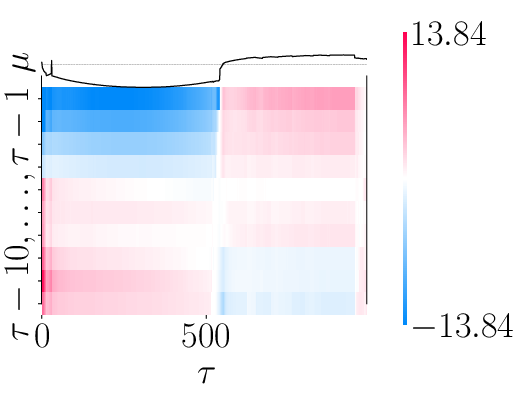}}%
\tcbitem\subfloat[SHAP, AR, \(\sigma = 1^{-6}\)]{\includegraphics[width=\linewidth,keepaspectratio]{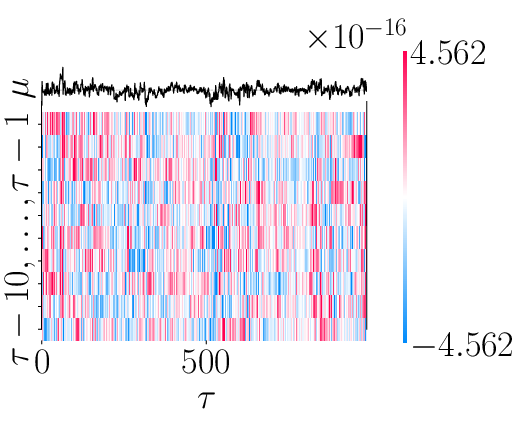}}%
\tcbitem\subfloat[DeepLIFT, AR, \(\sigma = 1^{-6}\)]{\includegraphics[width=\linewidth,keepaspectratio]{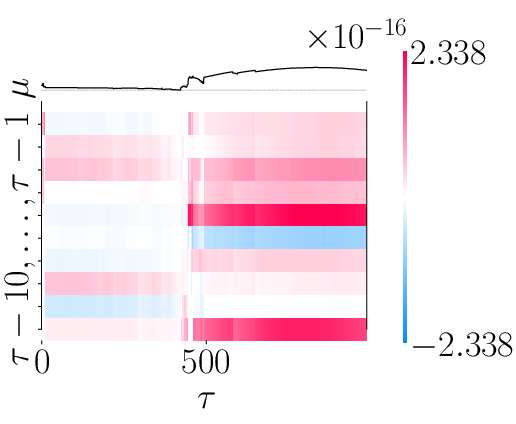}}%
\tcbitem\subfloat[SHAP, AR, \(\sigma = 1^{-5}\)]{\includegraphics[width=\linewidth,keepaspectratio]{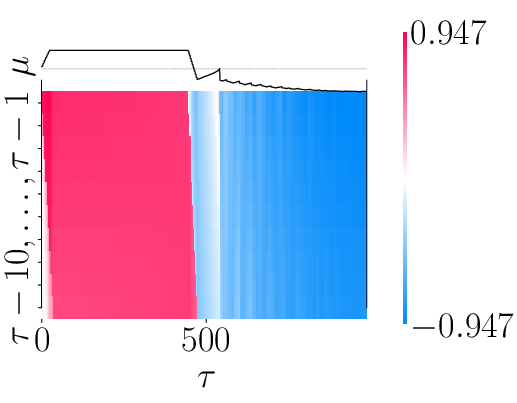}}%
\tcbitem\subfloat[DeepLIFT, AR, \(\sigma = 1^{-5}\)]{\includegraphics[width=\linewidth,keepaspectratio]{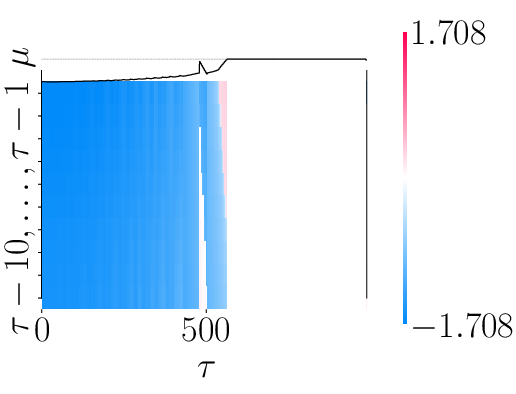}}%
\tcbitem\subfloat[SHAP, AR, \(\sigma = 1^{-4}\)]{\includegraphics[width=\linewidth,keepaspectratio]{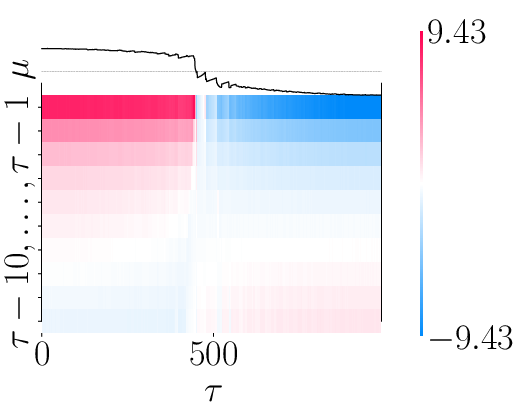}}%
\tcbitem\subfloat[DeepLIFT, AR, \(\sigma = 1^{-4}\)]{\includegraphics[width=\linewidth,keepaspectratio]{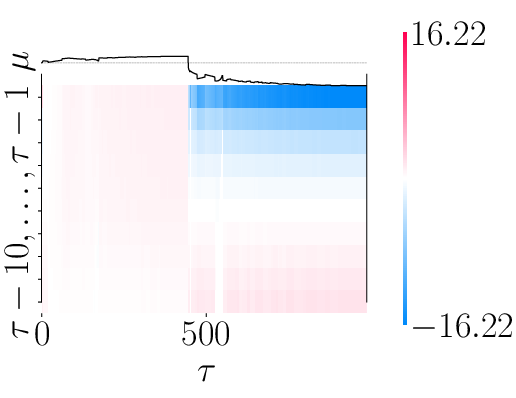}}%
\tcbitem\subfloat[SHAP, PI, \(\sigma = 1^{-2}\)]{\includegraphics[width=\linewidth,keepaspectratio]{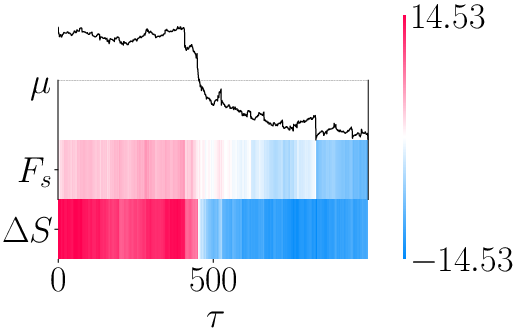}}%
\tcbitem\subfloat[DeepLIFT, PI, \(\sigma = 1^{-2}\)]{\includegraphics[width=\linewidth,keepaspectratio]{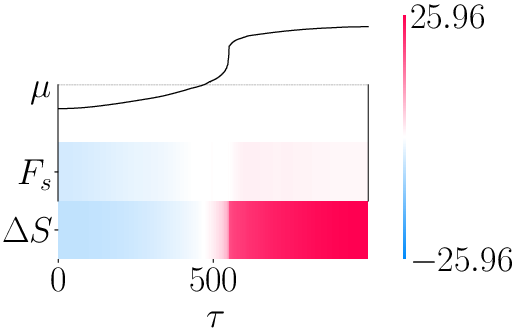}}%
\tcbitem\subfloat[SHAP, PI, \(\sigma = 1^{-3}\)]{\includegraphics[width=\linewidth,keepaspectratio]{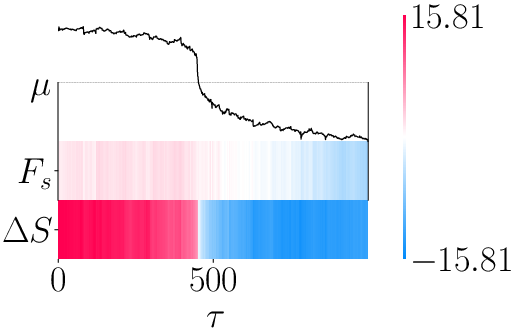}}%
\tcbitem\subfloat[DeepLIFT, PI, \(\sigma = 1^{-3}\)]{\includegraphics[width=\linewidth,keepaspectratio]{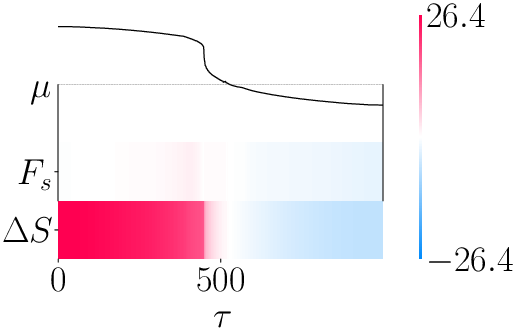}}%
\tcbitem\subfloat[SHAP, PI, \(\sigma = 1^{-6}\)]{\includegraphics[width=\linewidth,keepaspectratio]{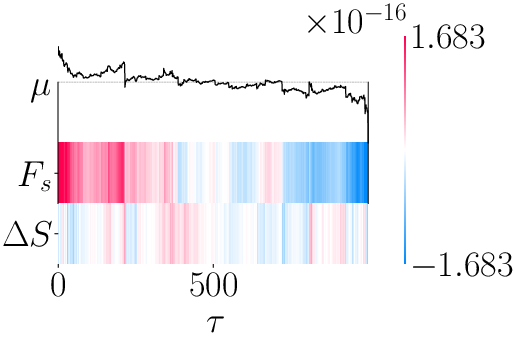}}%
\tcbitem\subfloat[DeepLIFT, PI, \(\sigma = 1^{-6}\)]{\includegraphics[width=\linewidth,keepaspectratio]{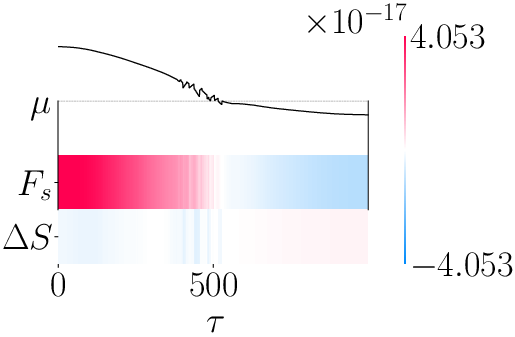}}%
\tcbitem\subfloat[SHAP, PI, \(\sigma = 1^{-5}\)]{\includegraphics[width=\linewidth,keepaspectratio]{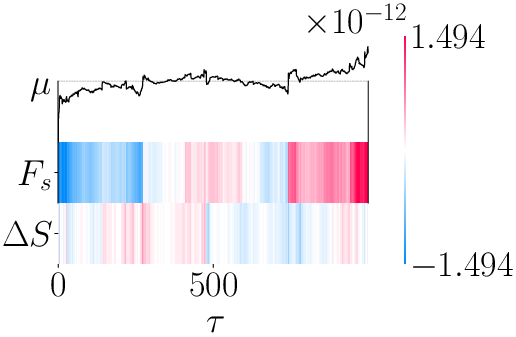}}%
\tcbitem\subfloat[DeepLIFT, PI, \(\sigma = 1^{-5}\)]{\includegraphics[width=\linewidth,keepaspectratio]{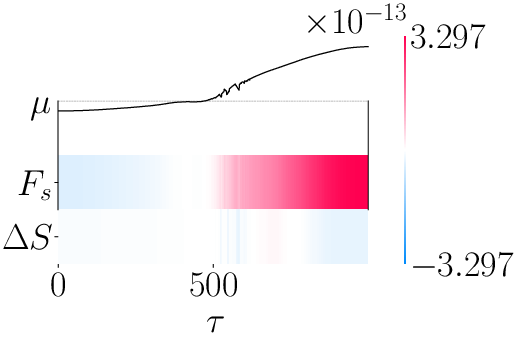}}%
\tcbitem\subfloat[SHAP, PI, \(\sigma = 1^{-4}\)]{\includegraphics[width=\linewidth,keepaspectratio]{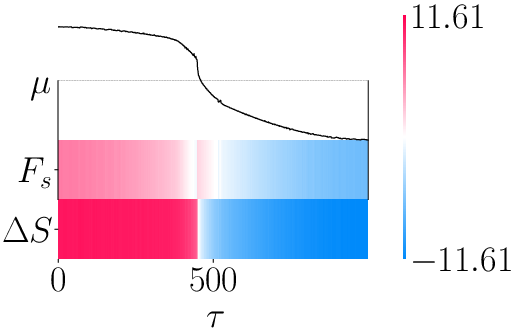}}%
\tcbitem\subfloat[DeepLIFT, PI, \(\sigma = 1^{-4}\)]{\includegraphics[width=\linewidth,keepaspectratio]{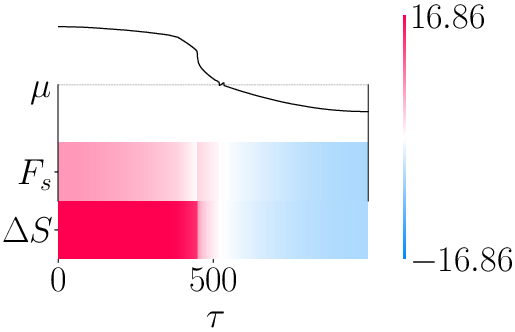}}%
\end{tcbitemize}%
\caption{BNN attributions under varying \(\sigma\) for \(\mathcal{F}_2\).}%
\label{fig:xai-2}%
\end{figure}

%

\begin{figure}%
\begin{tcbitemize}[raster equal height=rows, raster columns=3, raster halign=center, raster every box/.style=blankest]%
\tcbitem[raster multicolumn=3,boxed title style={center},halign=center]\includegraphics[width=0.33\textwidth]{../icml2024/fig/legend_pred_gt.pdf}%
\tcbitem\subfloat[AR, \(\sigma = 1^{-2}\)]{\includegraphics[width=\linewidth,keepaspectratio]{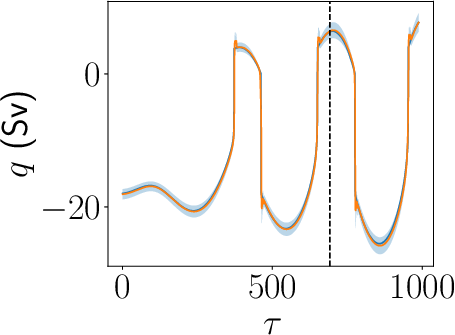}}%
\tcbitem\subfloat[AR, \(\sigma = 1^{-3}\)]{\includegraphics[width=\linewidth,keepaspectratio]{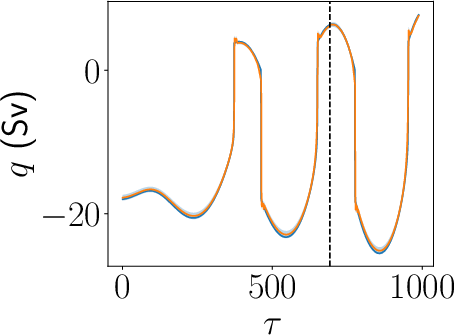}}%
\tcbitem\subfloat[AR, \(\sigma = 1^{-6}\)]{\includegraphics[width=\linewidth,keepaspectratio]{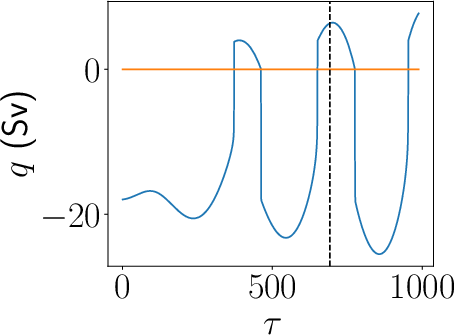}}%
\tcbitem\subfloat[AR, \(\sigma = 1^{-5}\)]{\includegraphics[width=\linewidth,keepaspectratio]{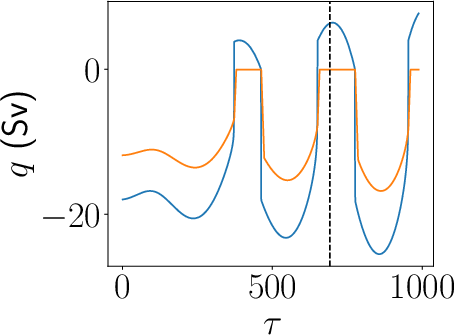}}%
\tcbitem\subfloat[AR, \(\sigma = 1^{-4}\)]{\includegraphics[width=\linewidth,keepaspectratio]{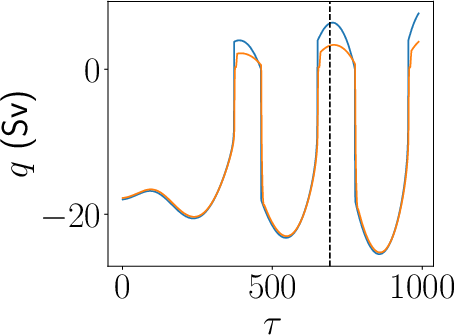}}%
\tcbitem\subfloat[PI, \(\sigma = 1^{-2}\)]{\includegraphics[width=\linewidth,keepaspectratio]{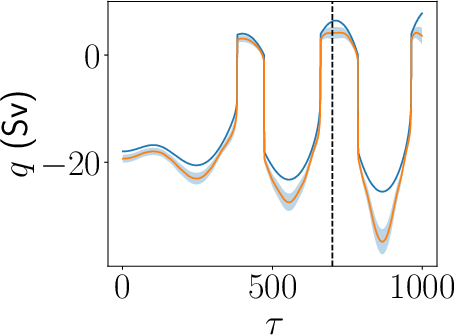}}%
\tcbitem\subfloat[PI, \(\sigma = 1^{-3}\)]{\includegraphics[width=\linewidth,keepaspectratio]{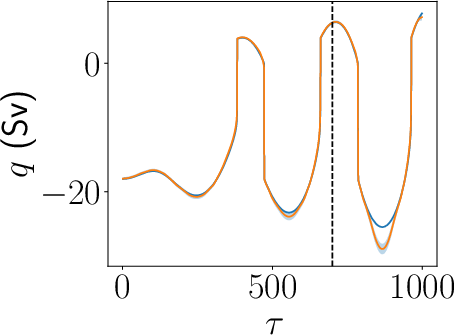}}%
\tcbitem\subfloat[PI, \(\sigma = 1^{-6}\)]{\includegraphics[width=\linewidth,keepaspectratio]{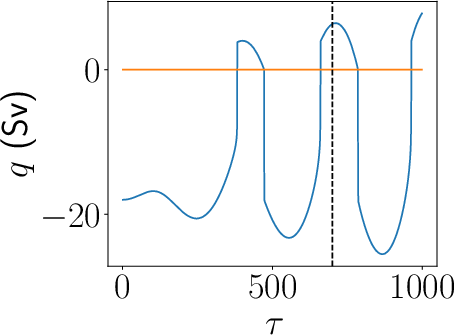}}%
\tcbitem\subfloat[PI, \(\sigma = 1^{-5}\)]{\includegraphics[width=\linewidth,keepaspectratio]{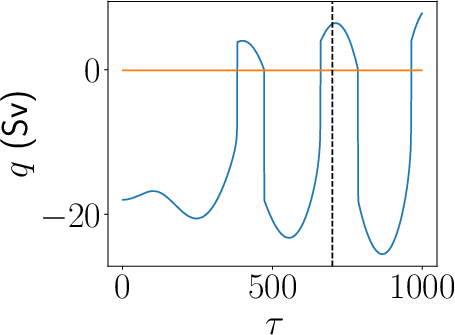}}%
\tcbitem\subfloat[PI, \(\sigma = 1^{-4}\)]{\includegraphics[width=\linewidth,keepaspectratio]{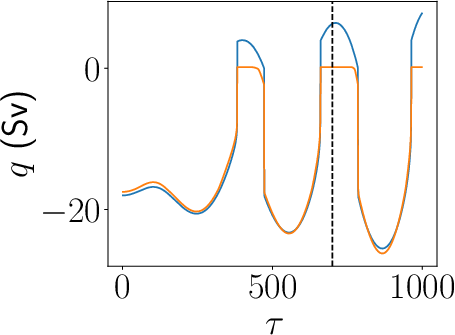}}%
\end{tcbitemize}%
\caption{BNN predictions under varying \(\sigma\) for \(\mathcal{F}_3\).}%
\label{fig:pgt-3}%
\end{figure}

%

\begin{figure}[h]%
\begin{tcbitemize}[raster equal height=rows, raster columns=4, raster halign=center, raster every box/.style=blankest]%
\tcbitem\subfloat[SHAP, AR, \(\sigma = 1^{-2}\)]{\includegraphics[width=\linewidth,keepaspectratio]{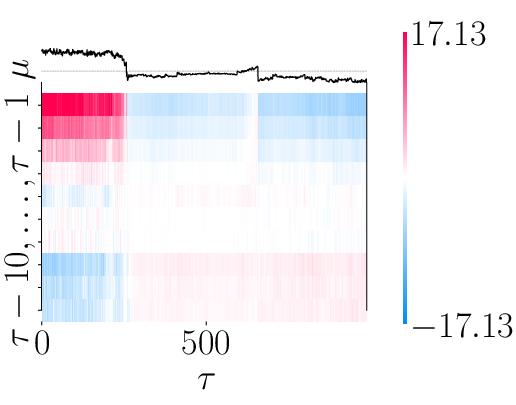}}%
\tcbitem\subfloat[DeepLIFT, AR, \(\sigma = 1^{-2}\)]{\includegraphics[width=\linewidth,keepaspectratio]{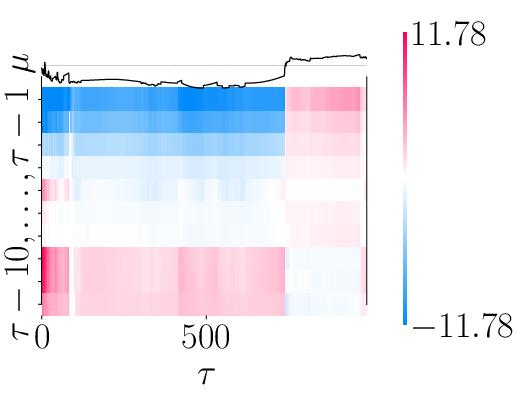}}%
\tcbitem\subfloat[SHAP, AR, \(\sigma = 1^{-3}\)]{\includegraphics[width=\linewidth,keepaspectratio]{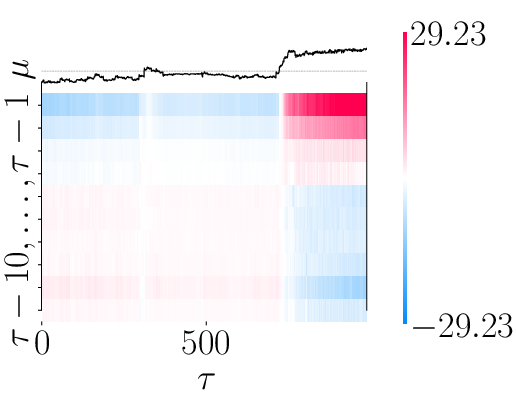}}%
\tcbitem\subfloat[DeepLIFT, AR, \(\sigma = 1^{-3}\)]{\includegraphics[width=\linewidth,keepaspectratio]{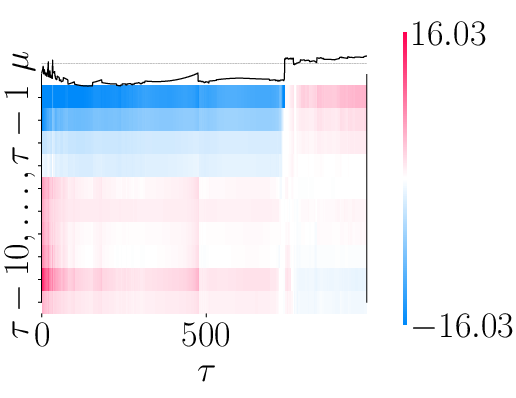}}%
\tcbitem\subfloat[SHAP, AR, \(\sigma = 1^{-6}\)]{\includegraphics[width=\linewidth,keepaspectratio]{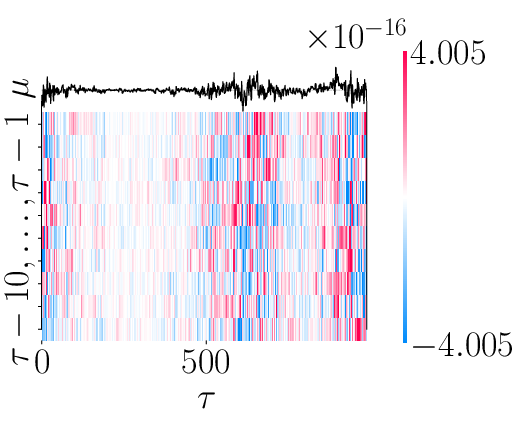}}%
\tcbitem\subfloat[DeepLIFT, AR, \(\sigma = 1^{-6}\)]{\includegraphics[width=\linewidth,keepaspectratio]{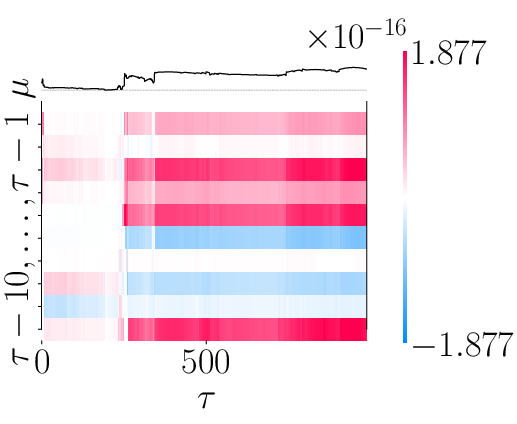}}%
\tcbitem\subfloat[SHAP, AR, \(\sigma = 1^{-5}\)]{\includegraphics[width=\linewidth,keepaspectratio]{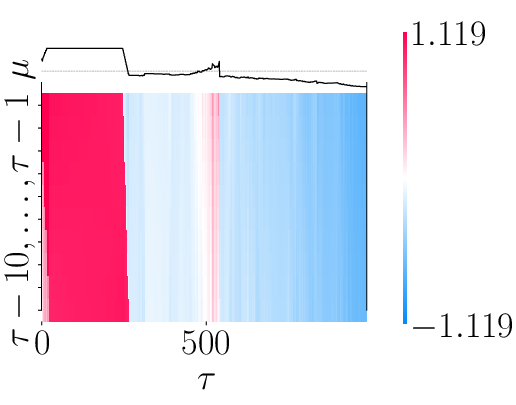}}%
\tcbitem\subfloat[DeepLIFT, AR, \(\sigma = 1^{-5}\)]{\includegraphics[width=\linewidth,keepaspectratio]{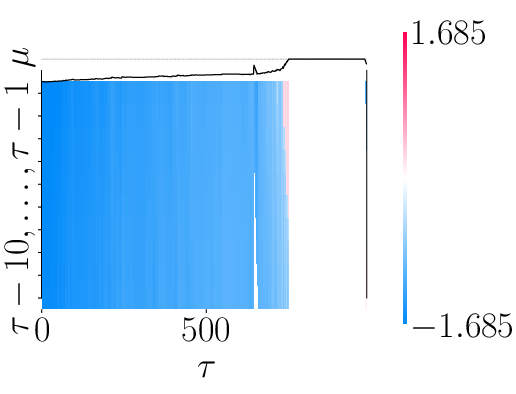}}%
\tcbitem\subfloat[SHAP, AR, \(\sigma = 1^{-4}\)]{\includegraphics[width=\linewidth,keepaspectratio]{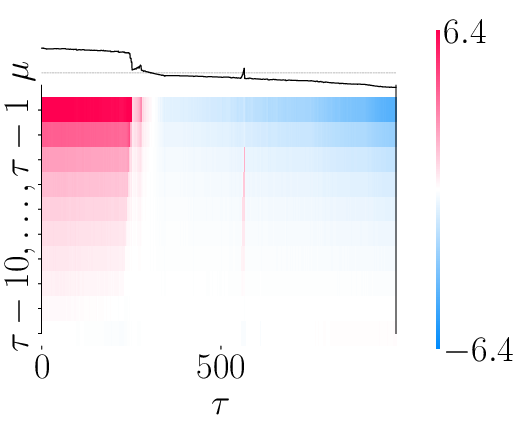}}%
\tcbitem\subfloat[DeepLIFT, AR, \(\sigma = 1^{-4}\)]{\includegraphics[width=\linewidth,keepaspectratio]{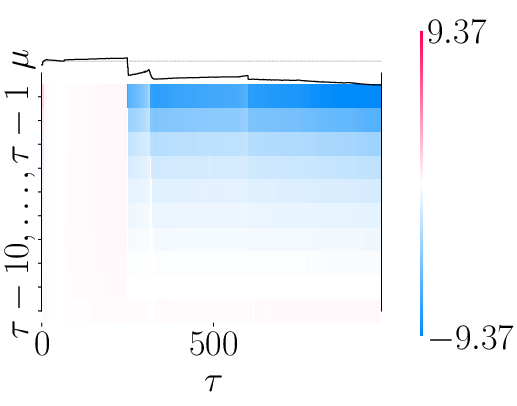}}%
\tcbitem\subfloat[SHAP, PI, \(\sigma = 1^{-2}\)]{\includegraphics[width=\linewidth,keepaspectratio]{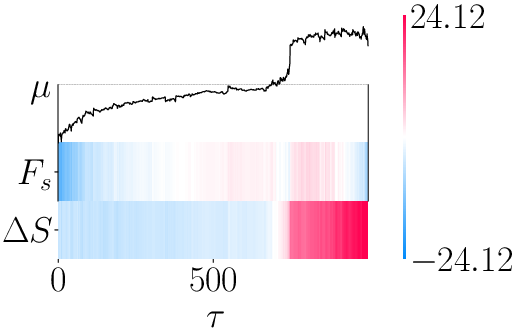}}%
\tcbitem\subfloat[DeepLIFT, PI, \(\sigma = 1^{-2}\)]{\includegraphics[width=\linewidth,keepaspectratio]{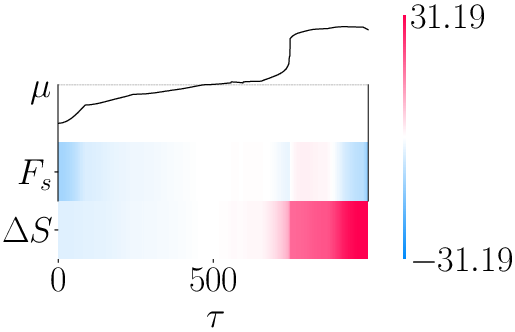}}%
\tcbitem\subfloat[SHAP, PI, \(\sigma = 1^{-3}\)]{\includegraphics[width=\linewidth,keepaspectratio]{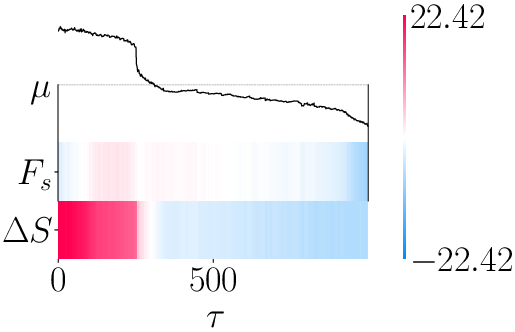}}%
\tcbitem\subfloat[DeepLIFT, PI, \(\sigma = 1^{-3}\)]{\includegraphics[width=\linewidth,keepaspectratio]{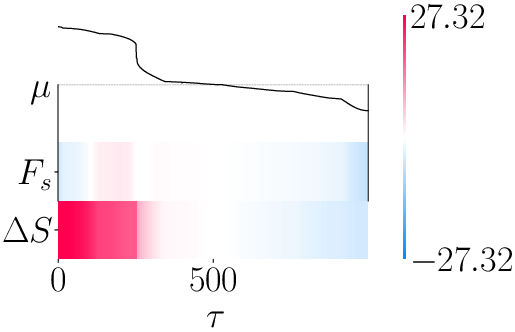}}%
\tcbitem\subfloat[SHAP, PI, \(\sigma = 1^{-6}\)]{\includegraphics[width=\linewidth,keepaspectratio]{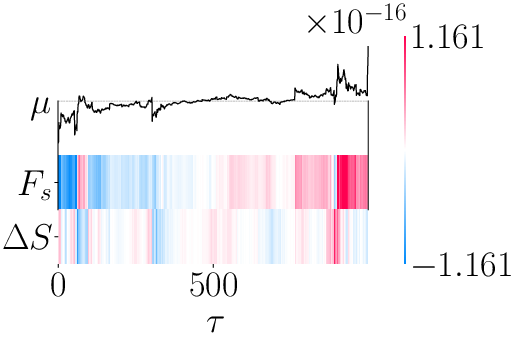}}%
\tcbitem\subfloat[DeepLIFT, PI, \(\sigma = 1^{-6}\)]{\includegraphics[width=\linewidth,keepaspectratio]{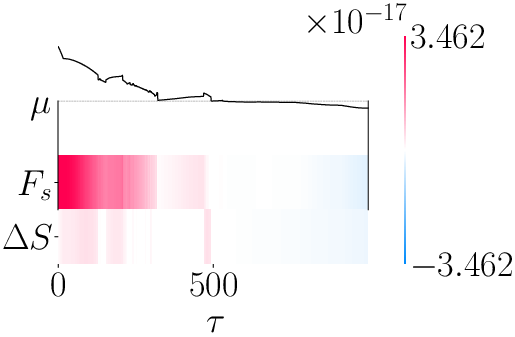}}%
\tcbitem\subfloat[SHAP, PI, \(\sigma = 1^{-5}\)]{\includegraphics[width=\linewidth,keepaspectratio]{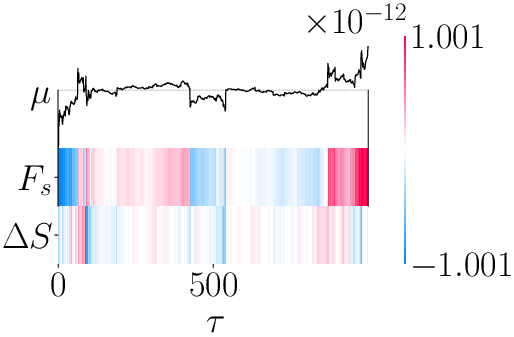}}%
\tcbitem\subfloat[DeepLIFT, PI, \(\sigma = 1^{-5}\)]{\includegraphics[width=\linewidth,keepaspectratio]{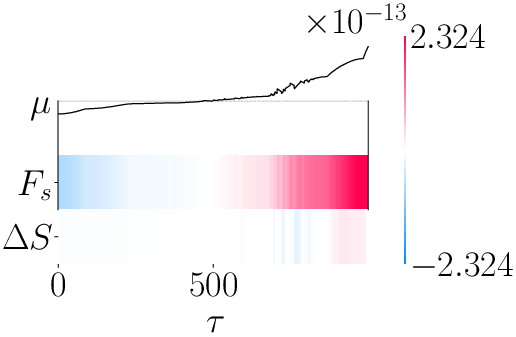}}%
\tcbitem\subfloat[SHAP, PI, \(\sigma = 1^{-4}\)]{\includegraphics[width=\linewidth,keepaspectratio]{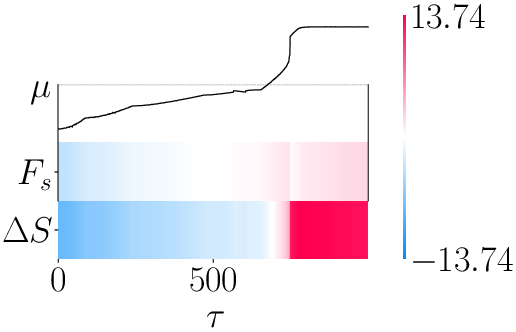}}%
\tcbitem\subfloat[DeepLIFT, PI, \(\sigma = 1^{-4}\)]{\includegraphics[width=\linewidth,keepaspectratio]{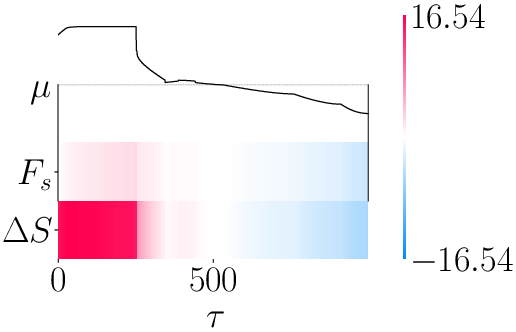}}%
\end{tcbitemize}%
\caption{BNN attributions under varying \(\sigma\) for \(\mathcal{F}_3\).}%
\label{fig:xai-3}%
\end{figure}

%

\begin{figure}%
\begin{tcbitemize}[raster equal height=rows, raster columns=3, raster halign=center, raster every box/.style=blankest]%
\tcbitem[raster multicolumn=3,boxed title style={center},halign=center]\includegraphics[width=0.33\textwidth]{../icml2024/fig/legend_pred_gt.pdf}%
\tcbitem\subfloat[PI, \(\sigma = 1^{-2}\)]{\includegraphics[width=\linewidth,keepaspectratio]{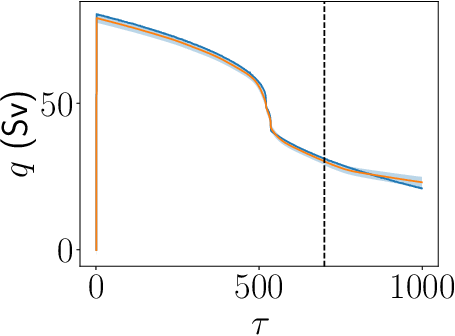}}%
\tcbitem\subfloat[PI, \(\sigma = 1^{-3}\)]{\includegraphics[width=\linewidth,keepaspectratio]{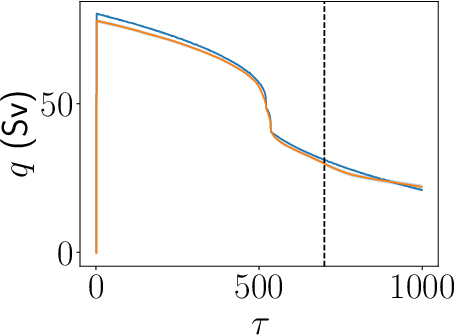}}%
\tcbitem\subfloat[PI, \(\sigma = 1^{-6}\)]{\includegraphics[width=\linewidth,keepaspectratio]{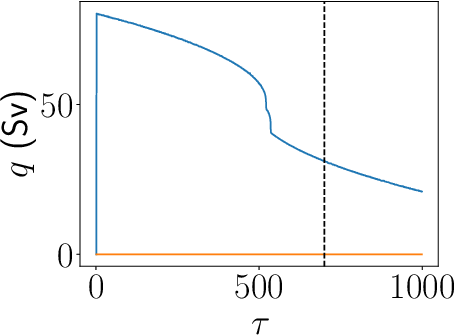}}%
\tcbitem\subfloat[PI, \(\sigma = 1^{-5}\)]{\includegraphics[width=\linewidth,keepaspectratio]{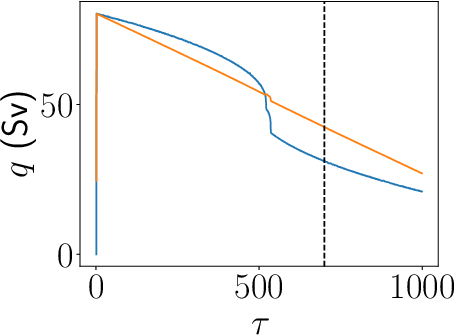}}%
\tcbitem\subfloat[PI, \(\sigma = 1^{-4}\)]{\includegraphics[width=\linewidth,keepaspectratio]{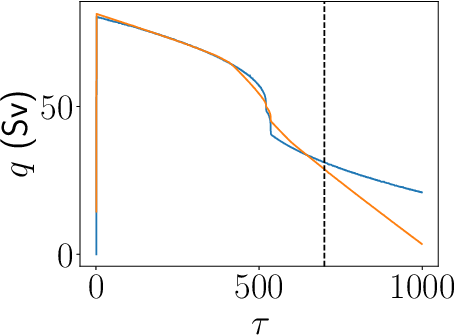}}%
\tcbitem\subfloat[AR, \(\sigma = 1^{-2}\)]{\includegraphics[width=\linewidth,keepaspectratio]{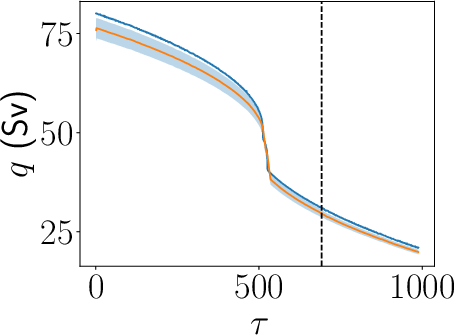}}%
\tcbitem\subfloat[AR, \(\sigma = 1^{-3}\)]{\includegraphics[width=\linewidth,keepaspectratio]{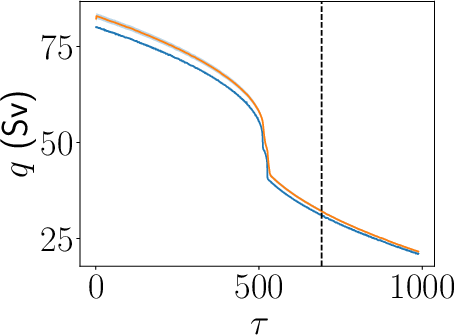}}%
\tcbitem\subfloat[AR, \(\sigma = 1^{-6}\)]{\includegraphics[width=\linewidth,keepaspectratio]{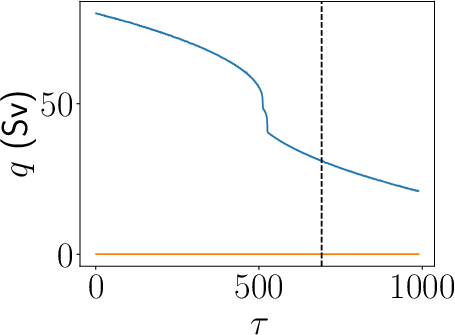}}%
\tcbitem\subfloat[AR, \(\sigma = 1^{-5}\)]{\includegraphics[width=\linewidth,keepaspectratio]{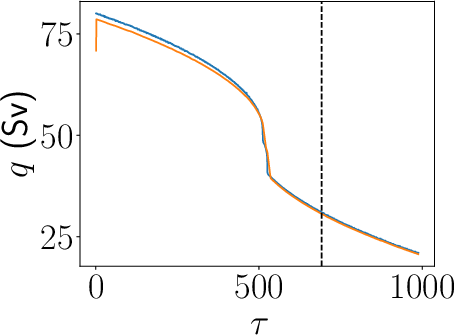}}%
\tcbitem\subfloat[AR, \(\sigma = 1^{-4}\)]{\includegraphics[width=\linewidth,keepaspectratio]{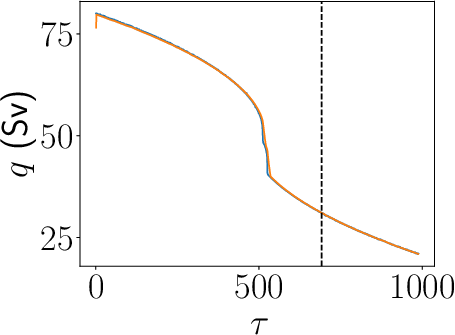}}%
\end{tcbitemize}%
\caption{BNN predictions under varying \(\sigma\) for \(\mathcal{F}_4\).}%
\label{fig:pgt-4}%
\end{figure}

%

\begin{figure}[h]%
\begin{tcbitemize}[raster equal height=rows, raster columns=4, raster halign=center, raster every box/.style=blankest]%
\tcbitem\subfloat[SHAP, PI, \(\sigma = 1^{-2}\)]{\includegraphics[width=\linewidth,keepaspectratio]{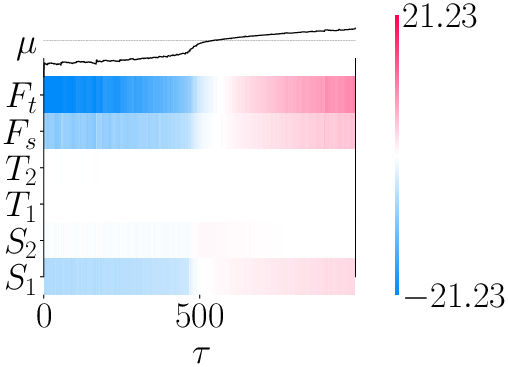}}%
\tcbitem\subfloat[DeepLIFT, PI, \(\sigma = 1^{-2}\)]{\includegraphics[width=\linewidth,keepaspectratio]{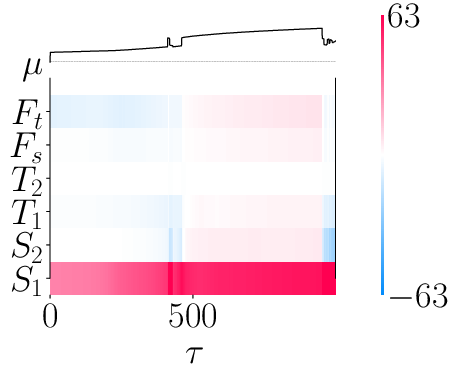}}%
\tcbitem\subfloat[SHAP, PI, \(\sigma = 1^{-3}\)]{\includegraphics[width=\linewidth,keepaspectratio]{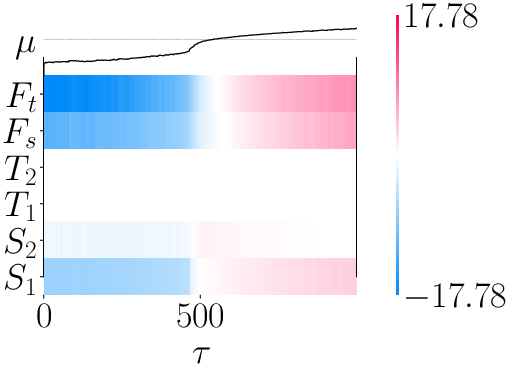}}%
\tcbitem\subfloat[DeepLIFT, PI, \(\sigma = 1^{-3}\)]{\includegraphics[width=\linewidth,keepaspectratio]{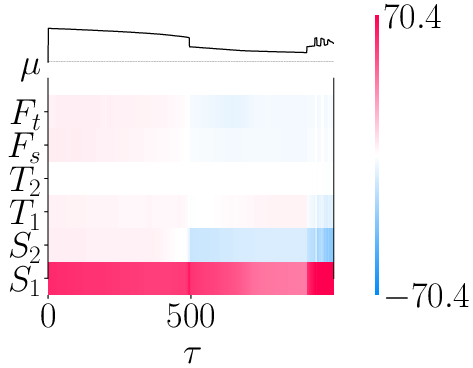}}%
\tcbitem\subfloat[SHAP, PI, \(\sigma = 1^{-6}\)]{\includegraphics[width=\linewidth,keepaspectratio]{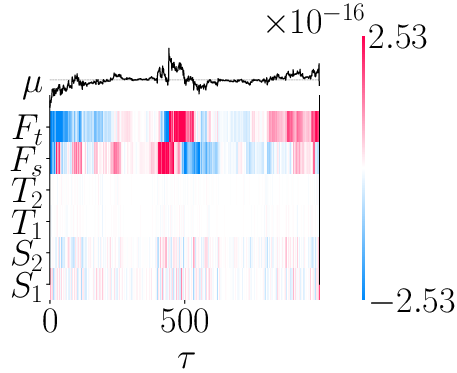}}%
\tcbitem\subfloat[DeepLIFT, PI, \(\sigma = 1^{-6}\)]{\includegraphics[width=\linewidth,keepaspectratio]{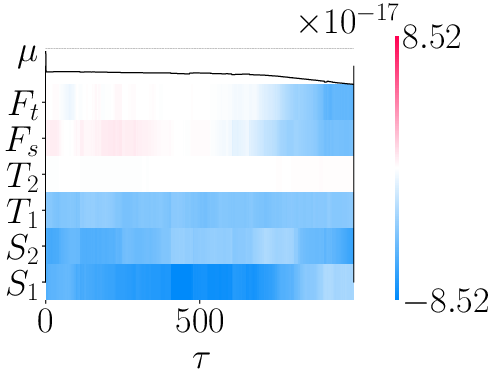}}%
\tcbitem\subfloat[SHAP, PI, \(\sigma = 1^{-5}\)]{\includegraphics[width=\linewidth,keepaspectratio]{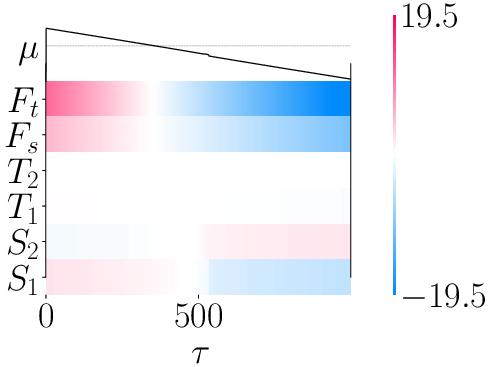}}%
\tcbitem\subfloat[DeepLIFT, PI, \(\sigma = 1^{-5}\)]{\includegraphics[width=\linewidth,keepaspectratio]{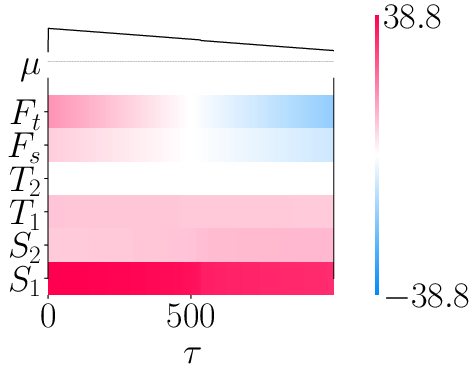}}%
\tcbitem\subfloat[SHAP, PI, \(\sigma = 1^{-4}\)]{\includegraphics[width=\linewidth,keepaspectratio]{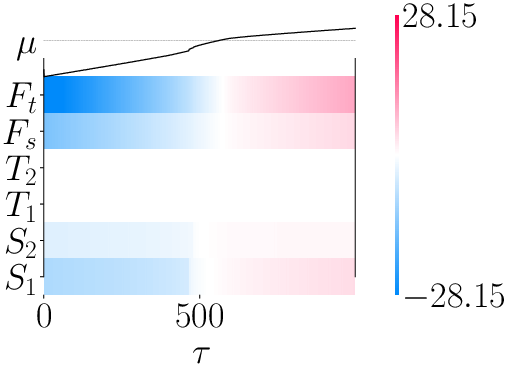}}%
\tcbitem\subfloat[DeepLIFT, PI, \(\sigma = 1^{-4}\)]{\includegraphics[width=\linewidth,keepaspectratio]{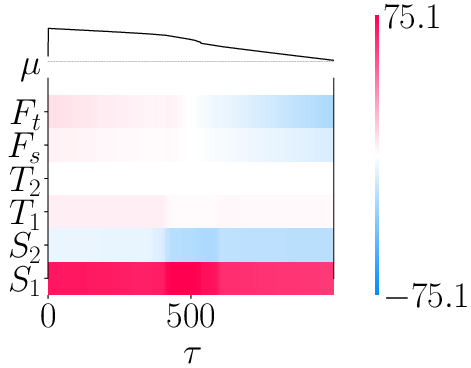}}%
\tcbitem\subfloat[SHAP, AR, \(\sigma = 1^{-2}\)]{\includegraphics[width=\linewidth,keepaspectratio]{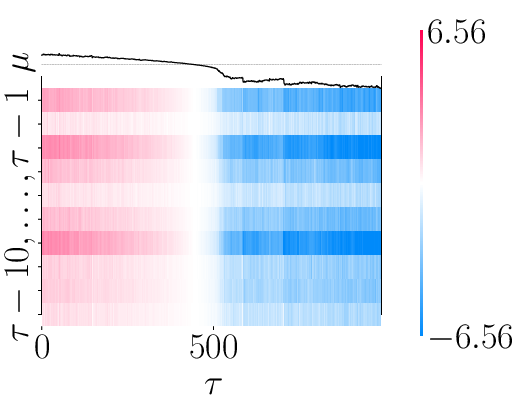}}%
\tcbitem\subfloat[DeepLIFT, AR, \(\sigma = 1^{-2}\)]{\includegraphics[width=\linewidth,keepaspectratio]{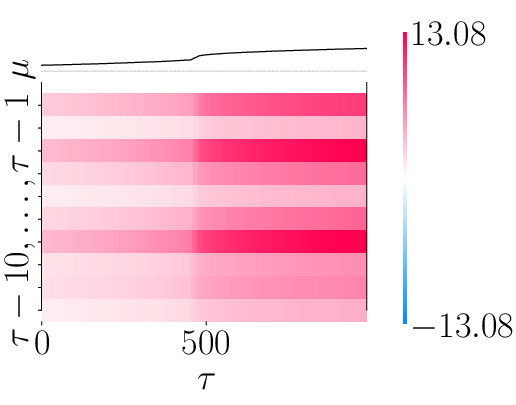}}%
\tcbitem\subfloat[SHAP, AR, \(\sigma = 1^{-3}\)]{\includegraphics[width=\linewidth,keepaspectratio]{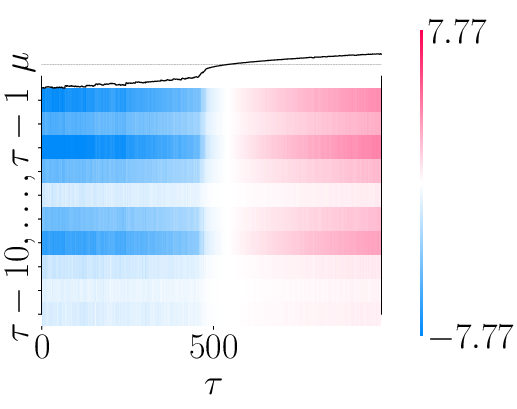}}%
\tcbitem\subfloat[DeepLIFT, AR, \(\sigma = 1^{-3}\)]{\includegraphics[width=\linewidth,keepaspectratio]{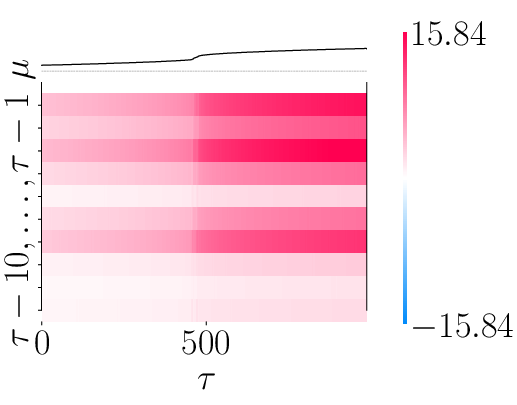}}%
\tcbitem\subfloat[SHAP, AR, \(\sigma = 1^{-6}\)]{\includegraphics[width=\linewidth,keepaspectratio]{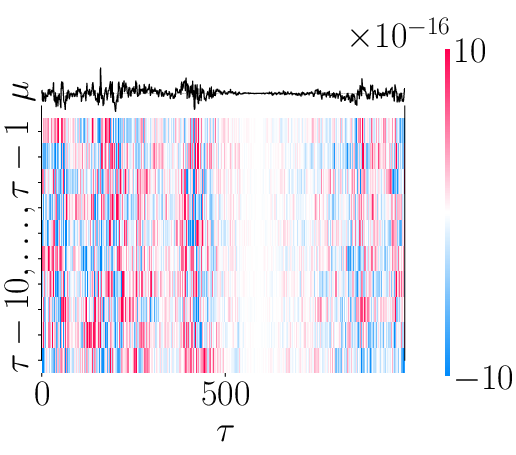}}%
\tcbitem\subfloat[DeepLIFT, AR, \(\sigma = 1^{-6}\)]{\includegraphics[width=\linewidth,keepaspectratio]{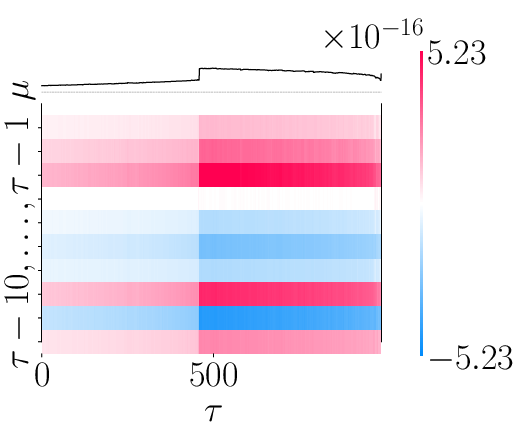}}%
\tcbitem\subfloat[SHAP, AR, \(\sigma = 1^{-5}\)]{\includegraphics[width=\linewidth,keepaspectratio]{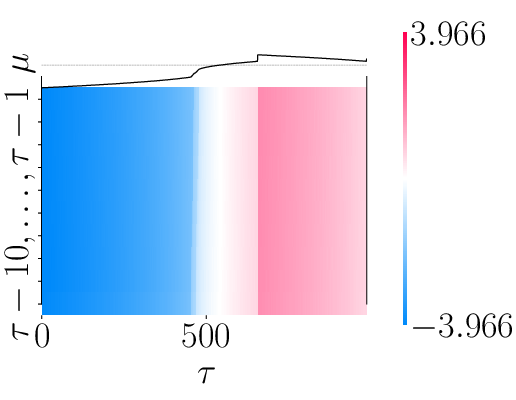}}%
\tcbitem\subfloat[DeepLIFT, AR, \(\sigma = 1^{-5}\)]{\includegraphics[width=\linewidth,keepaspectratio]{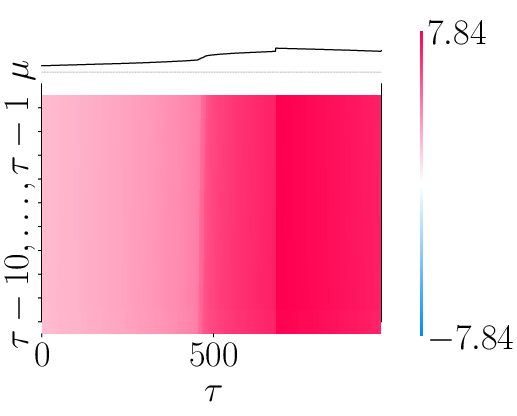}}%
\tcbitem\subfloat[SHAP, AR, \(\sigma = 1^{-4}\)]{\includegraphics[width=\linewidth,keepaspectratio]{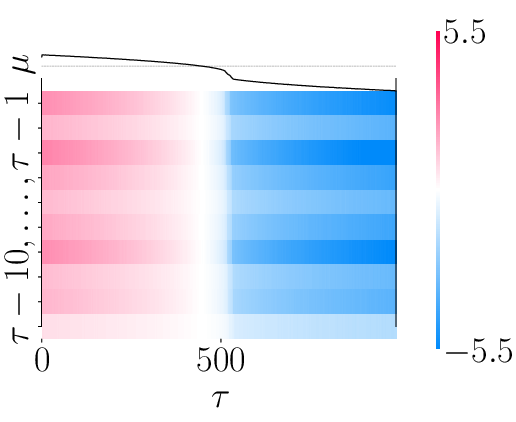}}%
\tcbitem\subfloat[DeepLIFT, AR, \(\sigma = 1^{-4}\)]{\includegraphics[width=\linewidth,keepaspectratio]{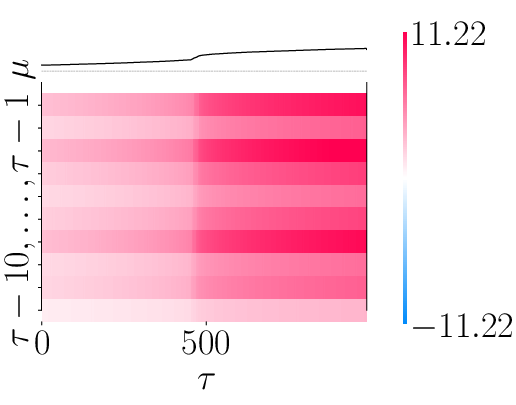}}%
\end{tcbitemize}%
\caption{BNN attributions under varying \(\sigma\) for \(\mathcal{F}_4\).}%
\label{fig:xai-4}%
\end{figure}

%

\begin{figure}%
\begin{tcbitemize}[raster equal height=rows, raster columns=3, raster halign=center, raster every box/.style=blankest]%
\tcbitem[raster multicolumn=3,boxed title style={center},halign=center]\includegraphics[width=0.33\textwidth]{../icml2024/fig/legend_pred_gt.pdf}%
\tcbitem\subfloat[PI, \(\sigma = 1^{-2}\)]{\includegraphics[width=\linewidth,keepaspectratio]{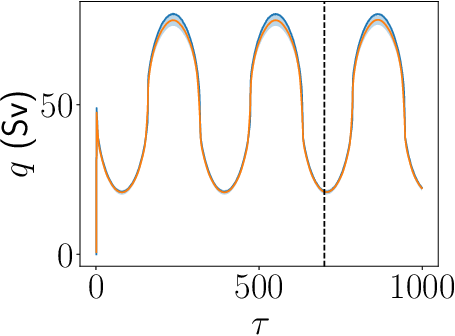}}%
\tcbitem\subfloat[PI, \(\sigma = 1^{-3}\)]{\includegraphics[width=\linewidth,keepaspectratio]{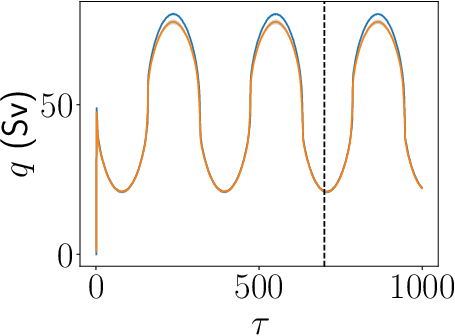}}%
\tcbitem\subfloat[PI, \(\sigma = 1^{-6}\)]{\includegraphics[width=\linewidth,keepaspectratio]{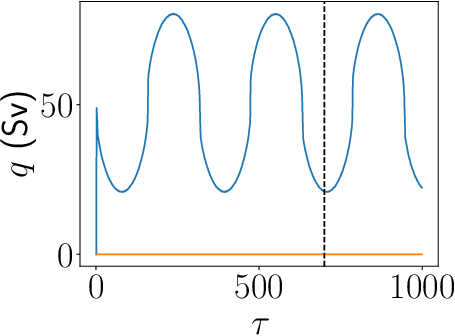}}%
\tcbitem\subfloat[PI, \(\sigma = 1^{-5}\)]{\includegraphics[width=\linewidth,keepaspectratio]{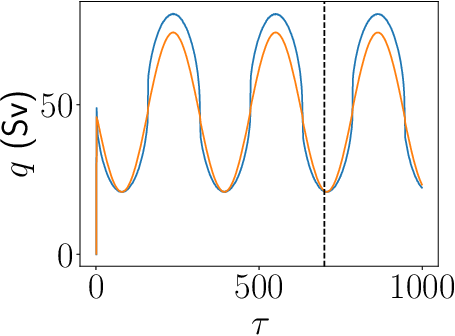}}%
\tcbitem\subfloat[PI, \(\sigma = 1^{-4}\)]{\includegraphics[width=\linewidth,keepaspectratio]{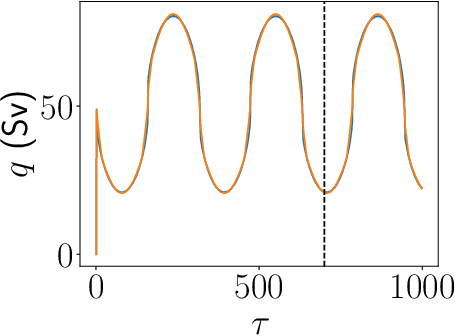}}%
\tcbitem\subfloat[AR, \(\sigma = 1^{-2}\)]{\includegraphics[width=\linewidth,keepaspectratio]{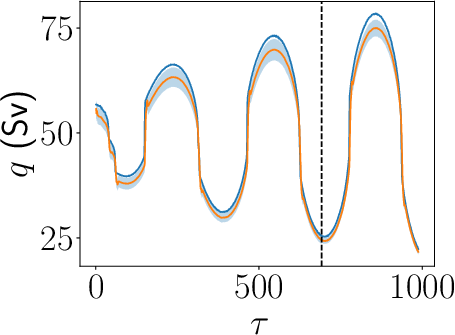}}%
\tcbitem\subfloat[AR, \(\sigma = 1^{-3}\)]{\includegraphics[width=\linewidth,keepaspectratio]{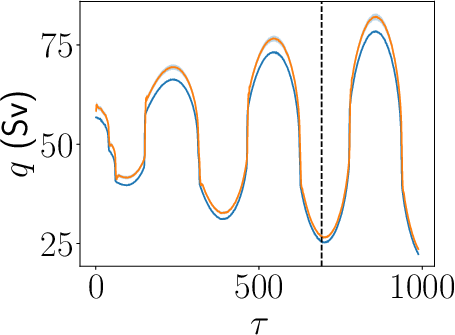}}%
\tcbitem\subfloat[AR, \(\sigma = 1^{-6}\)]{\includegraphics[width=\linewidth,keepaspectratio]{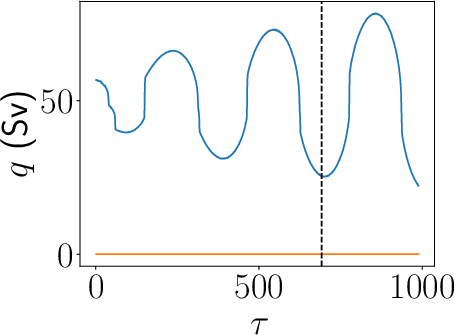}}%
\tcbitem\subfloat[AR, \(\sigma = 1^{-5}\)]{\includegraphics[width=\linewidth,keepaspectratio]{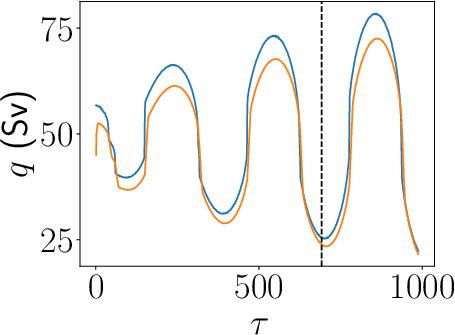}}%
\tcbitem\subfloat[AR, \(\sigma = 1^{-4}\)]{\includegraphics[width=\linewidth,keepaspectratio]{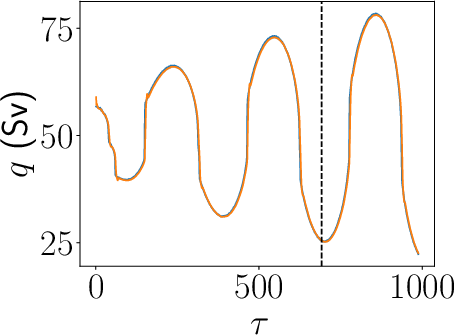}}%
\end{tcbitemize}%
\caption{BNN predictions under varying \(\sigma\) for \(\mathcal{F}_5\).}%
\label{fig:pgt-5}%
\end{figure}

%

\begin{figure}[h]%
\begin{tcbitemize}[raster equal height=rows, raster columns=4, raster halign=center, raster every box/.style=blankest]%
\tcbitem\subfloat[SHAP, PI, \(\sigma = 1^{-2}\)]{\includegraphics[width=\linewidth,keepaspectratio]{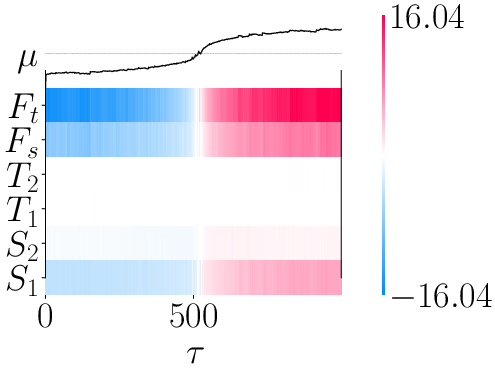}}%
\tcbitem\subfloat[DeepLIFT, PI, \(\sigma = 1^{-2}\)]{\includegraphics[width=\linewidth,keepaspectratio]{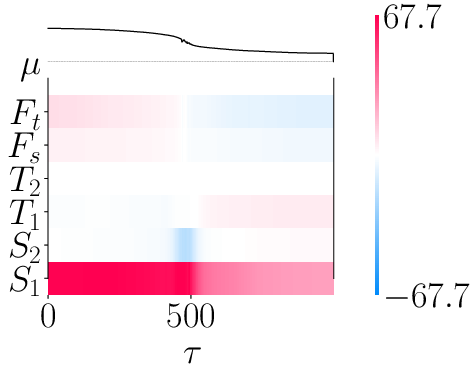}}%
\tcbitem\subfloat[SHAP, PI, \(\sigma = 1^{-3}\)]{\includegraphics[width=\linewidth,keepaspectratio]{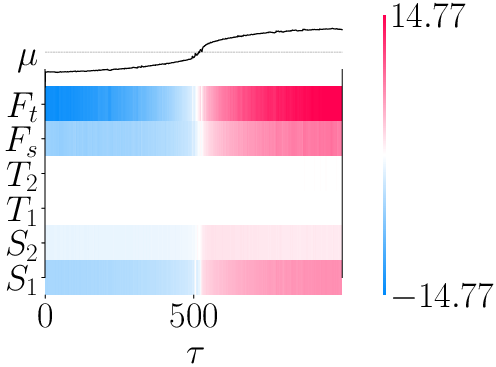}}%
\tcbitem\subfloat[DeepLIFT, PI, \(\sigma = 1^{-3}\)]{\includegraphics[width=\linewidth,keepaspectratio]{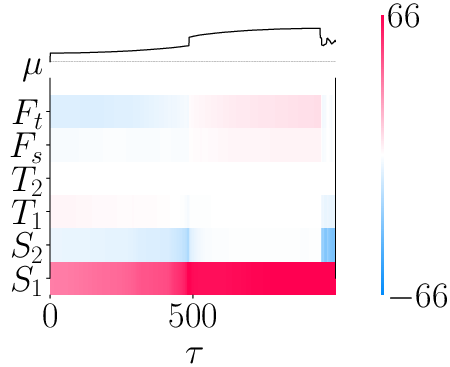}}%
\tcbitem\subfloat[SHAP, PI, \(\sigma = 1^{-6}\)]{\includegraphics[width=\linewidth,keepaspectratio]{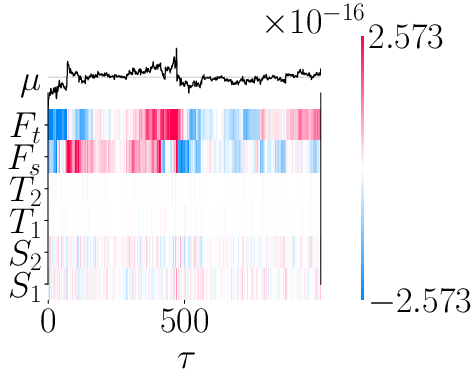}}%
\tcbitem\subfloat[DeepLIFT, PI, \(\sigma = 1^{-6}\)]{\includegraphics[width=\linewidth,keepaspectratio]{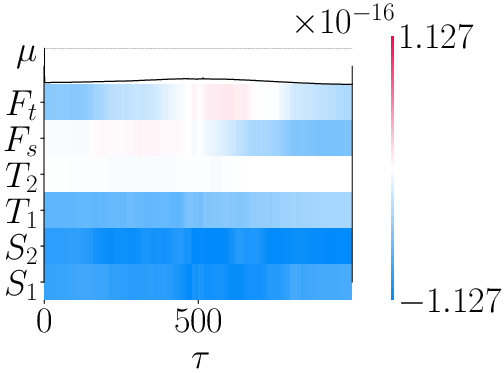}}%
\tcbitem\subfloat[SHAP, PI, \(\sigma = 1^{-5}\)]{\includegraphics[width=\linewidth,keepaspectratio]{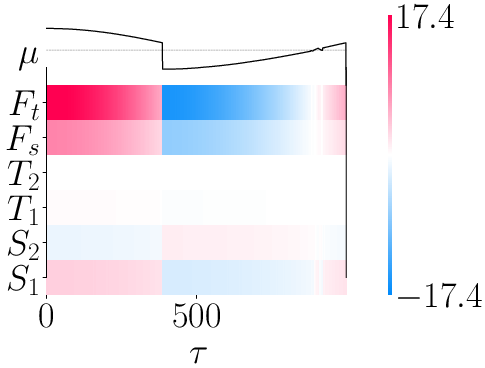}}%
\tcbitem\subfloat[DeepLIFT, PI, \(\sigma = 1^{-5}\)]{\includegraphics[width=\linewidth,keepaspectratio]{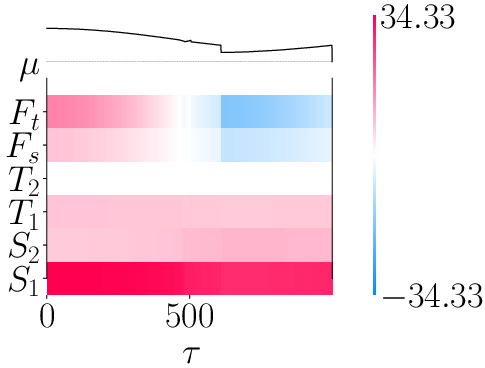}}%
\tcbitem\subfloat[SHAP, PI, \(\sigma = 1^{-4}\)]{\includegraphics[width=\linewidth,keepaspectratio]{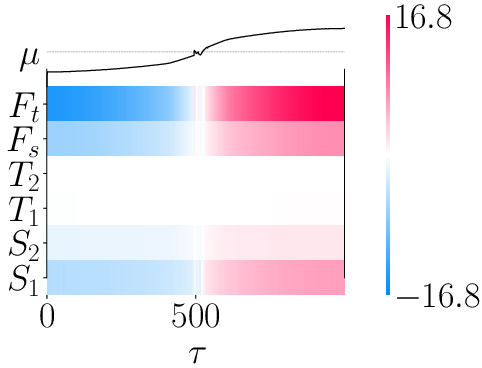}}%
\tcbitem\subfloat[DeepLIFT, PI, \(\sigma = 1^{-4}\)]{\includegraphics[width=\linewidth,keepaspectratio]{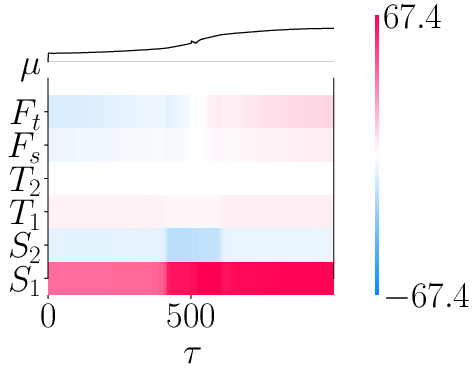}}%
\tcbitem\subfloat[SHAP, AR, \(\sigma = 1^{-2}\)]{\includegraphics[width=\linewidth,keepaspectratio]{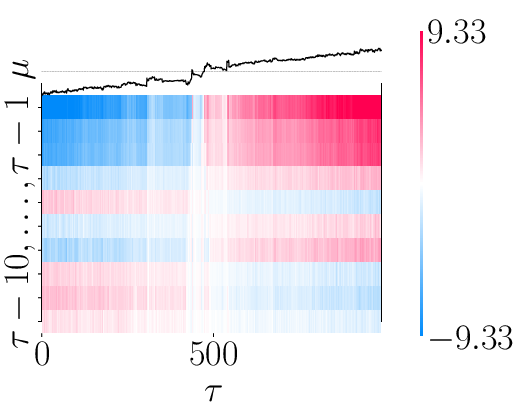}}%
\tcbitem\subfloat[DeepLIFT, AR, \(\sigma = 1^{-2}\)]{\includegraphics[width=\linewidth,keepaspectratio]{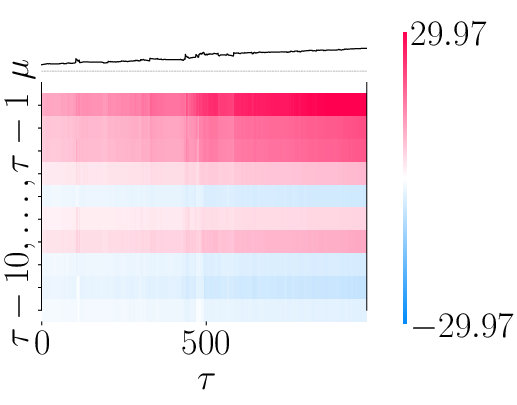}}%
\tcbitem\subfloat[SHAP, AR, \(\sigma = 1^{-3}\)]{\includegraphics[width=\linewidth,keepaspectratio]{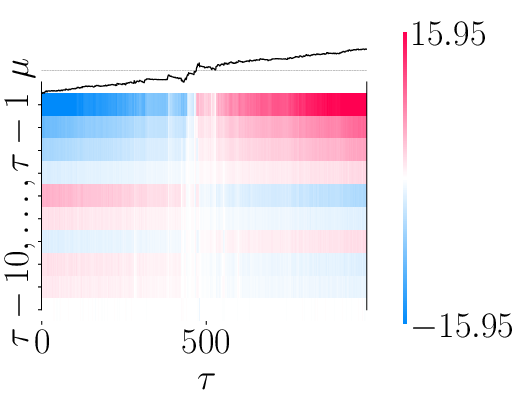}}%
\tcbitem\subfloat[DeepLIFT, AR, \(\sigma = 1^{-3}\)]{\includegraphics[width=\linewidth,keepaspectratio]{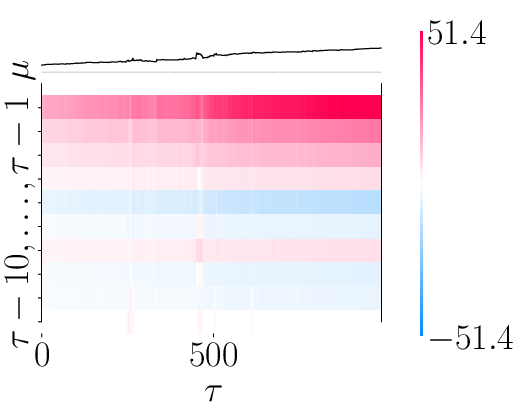}}%
\tcbitem\subfloat[SHAP, AR, \(\sigma = 1^{-6}\)]{\includegraphics[width=\linewidth,keepaspectratio]{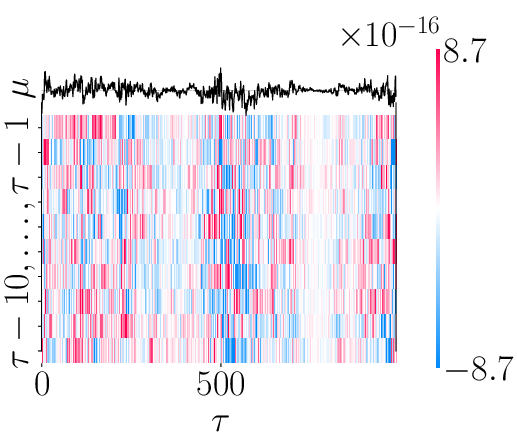}}%
\tcbitem\subfloat[DeepLIFT, AR, \(\sigma = 1^{-6}\)]{\includegraphics[width=\linewidth,keepaspectratio]{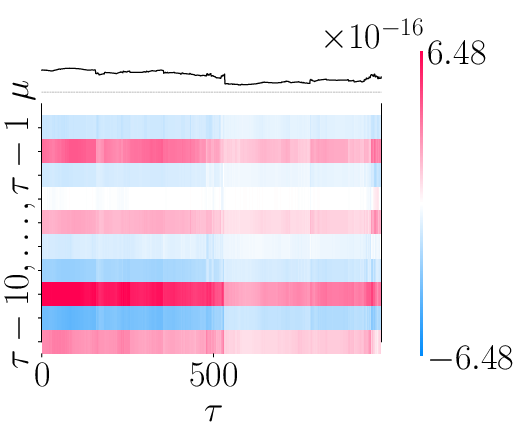}}%
\tcbitem\subfloat[SHAP, AR, \(\sigma = 1^{-5}\)]{\includegraphics[width=\linewidth,keepaspectratio]{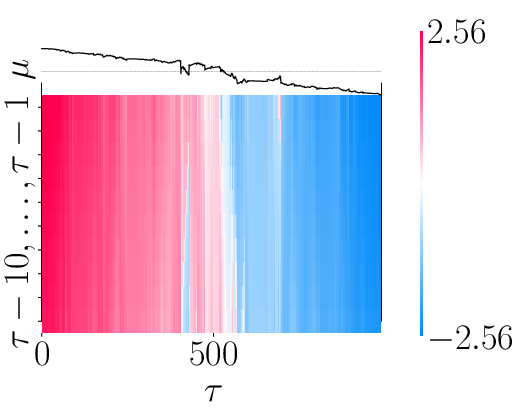}}%
\tcbitem\subfloat[DeepLIFT, AR, \(\sigma = 1^{-5}\)]{\includegraphics[width=\linewidth,keepaspectratio]{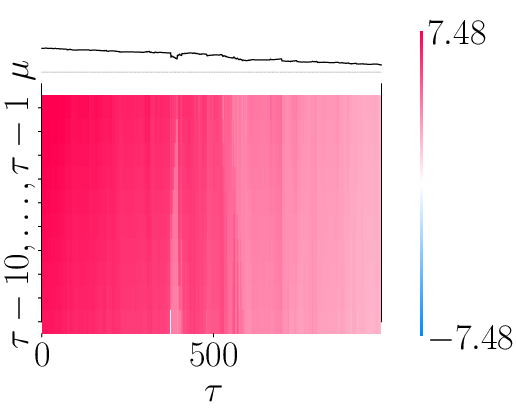}}%
\tcbitem\subfloat[SHAP, AR, \(\sigma = 1^{-4}\)]{\includegraphics[width=\linewidth,keepaspectratio]{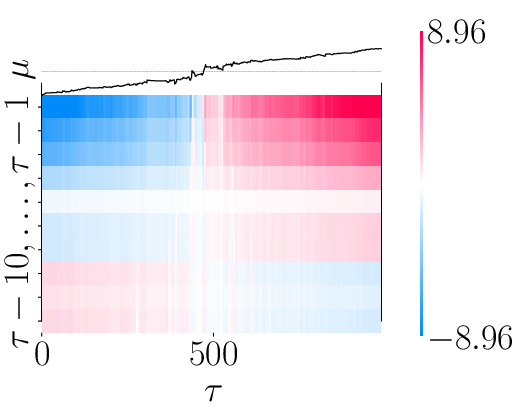}}%
\tcbitem\subfloat[DeepLIFT, AR, \(\sigma = 1^{-4}\)]{\includegraphics[width=\linewidth,keepaspectratio]{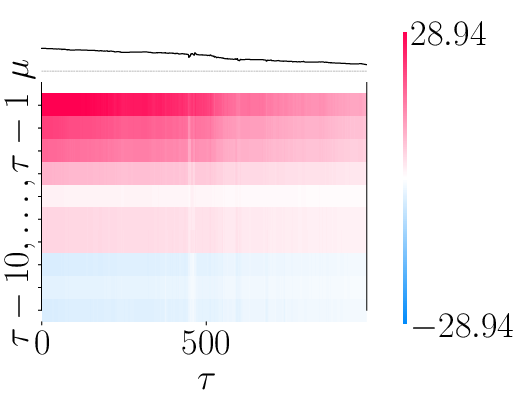}}%
\end{tcbitemize}%
\caption{BNN attributions under varying \(\sigma\) for \(\mathcal{F}_5\).}%
\label{fig:xai-5}%
\end{figure}

%

\begin{figure}%
\begin{tcbitemize}[raster equal height=rows, raster columns=3, raster halign=center, raster every box/.style=blankest]%
\tcbitem[raster multicolumn=3,boxed title style={center},halign=center]\includegraphics[width=0.33\textwidth]{../icml2024/fig/legend_pred_gt.pdf}%
\tcbitem\subfloat[PI, \(\sigma = 1^{-2}\)]{\includegraphics[width=\linewidth,keepaspectratio]{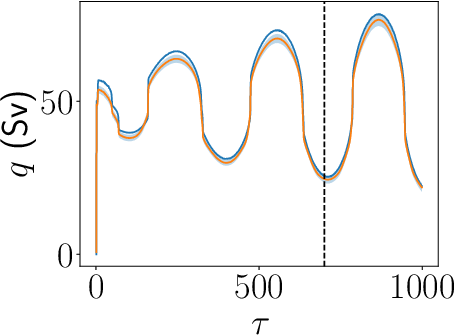}}%
\tcbitem\subfloat[PI, \(\sigma = 1^{-3}\)]{\includegraphics[width=\linewidth,keepaspectratio]{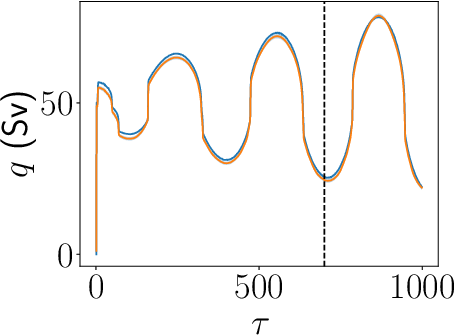}}%
\tcbitem\subfloat[PI, \(\sigma = 1^{-6}\)]{\includegraphics[width=\linewidth,keepaspectratio]{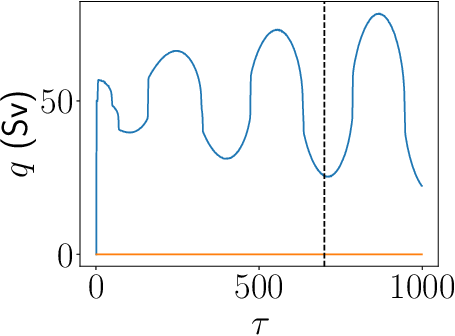}}%
\tcbitem\subfloat[PI, \(\sigma = 1^{-5}\)]{\includegraphics[width=\linewidth,keepaspectratio]{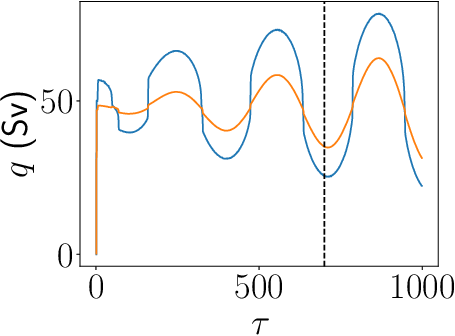}}%
\tcbitem\subfloat[PI, \(\sigma = 1^{-4}\)]{\includegraphics[width=\linewidth,keepaspectratio]{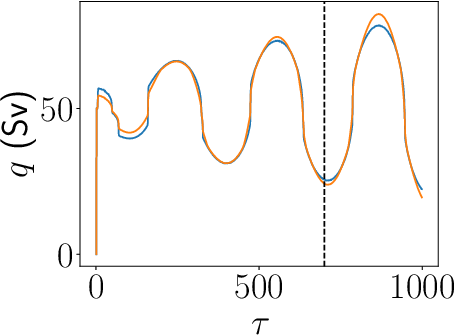}}%
\tcbitem\subfloat[AR, \(\sigma = 1^{-2}\)]{\includegraphics[width=\linewidth,keepaspectratio]{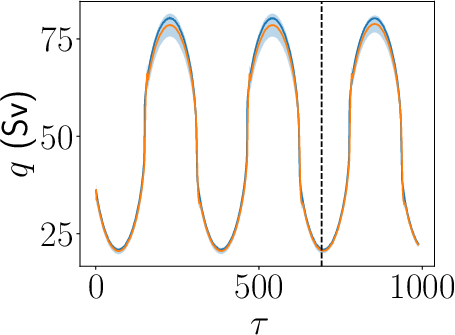}}%
\tcbitem\subfloat[AR, \(\sigma = 1^{-3}\)]{\includegraphics[width=\linewidth,keepaspectratio]{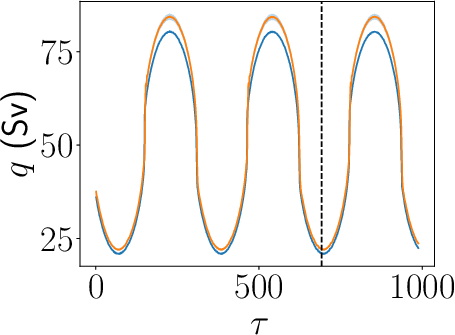}}%
\tcbitem\subfloat[AR, \(\sigma = 1^{-6}\)]{\includegraphics[width=\linewidth,keepaspectratio]{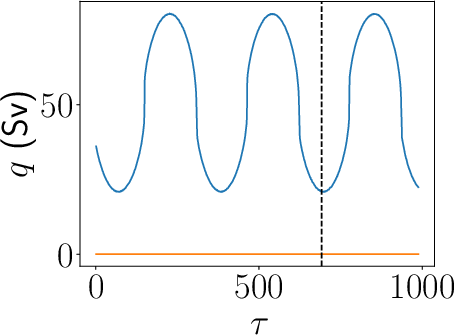}}%
\tcbitem\subfloat[AR, \(\sigma = 1^{-5}\)]{\includegraphics[width=\linewidth,keepaspectratio]{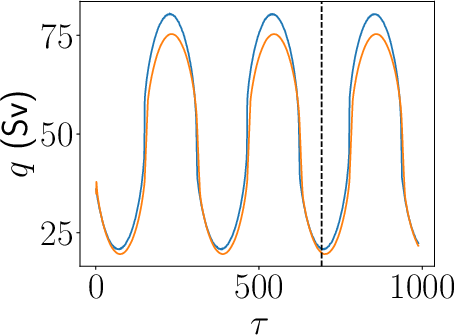}}%
\tcbitem\subfloat[AR, \(\sigma = 1^{-4}\)]{\includegraphics[width=\linewidth,keepaspectratio]{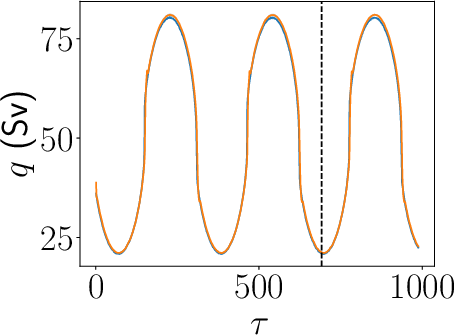}}%
\end{tcbitemize}%
\caption{BNN predictions under varying \(\sigma\) for \(\mathcal{F}_6\).}%
\label{fig:pgt-6}%
\end{figure}

%

\begin{figure}[h]%
\begin{tcbitemize}[raster equal height=rows, raster columns=4, raster halign=center, raster every box/.style=blankest]%
\tcbitem\subfloat[SHAP, PI, \(\sigma = 1^{-2}\)]{\includegraphics[width=\linewidth,keepaspectratio]{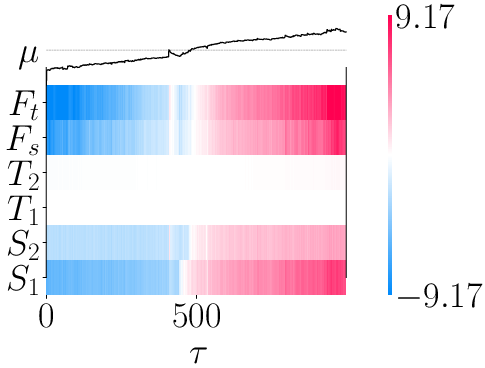}}%
\tcbitem\subfloat[DeepLIFT, PI, \(\sigma = 1^{-2}\)]{\includegraphics[width=\linewidth,keepaspectratio]{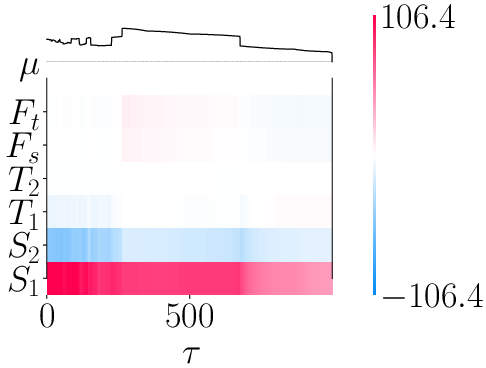}}%
\tcbitem\subfloat[SHAP, PI, \(\sigma = 1^{-3}\)]{\includegraphics[width=\linewidth,keepaspectratio]{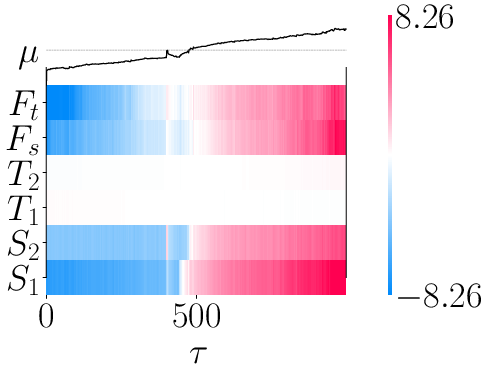}}%
\tcbitem\subfloat[DeepLIFT, PI, \(\sigma = 1^{-3}\)]{\includegraphics[width=\linewidth,keepaspectratio]{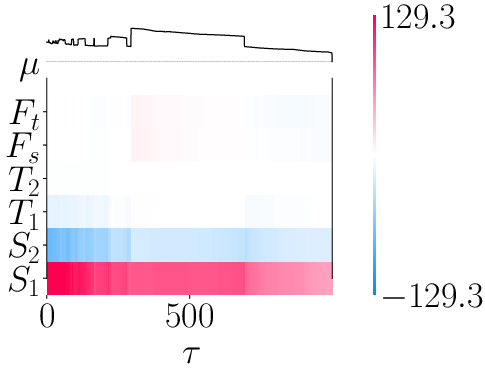}}%
\tcbitem\subfloat[SHAP, PI, \(\sigma = 1^{-6}\)]{\includegraphics[width=\linewidth,keepaspectratio]{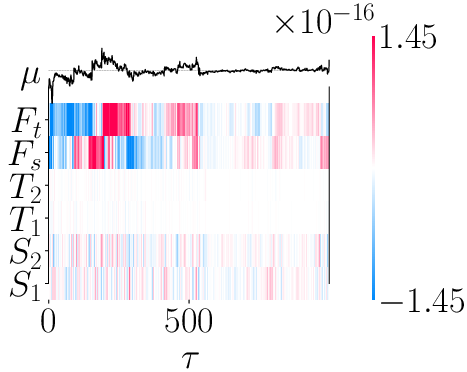}}%
\tcbitem\subfloat[DeepLIFT, PI, \(\sigma = 1^{-6}\)]{\includegraphics[width=\linewidth,keepaspectratio]{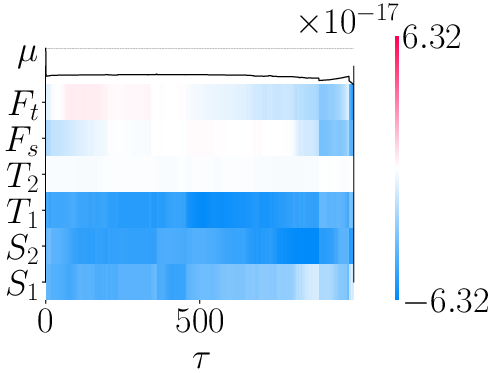}}%
\tcbitem\subfloat[SHAP, PI, \(\sigma = 1^{-5}\)]{\includegraphics[width=\linewidth,keepaspectratio]{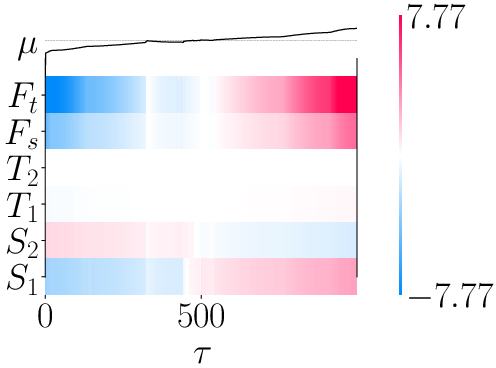}}%
\tcbitem\subfloat[DeepLIFT, PI, \(\sigma = 1^{-5}\)]{\includegraphics[width=\linewidth,keepaspectratio]{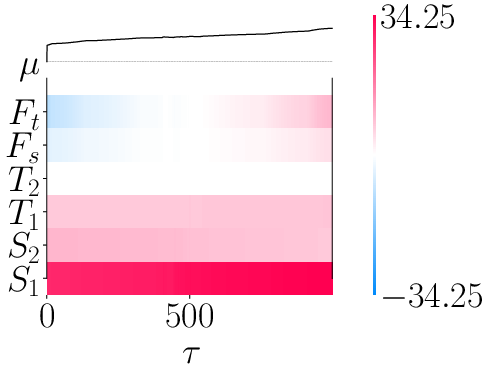}}%
\tcbitem\subfloat[SHAP, PI, \(\sigma = 1^{-4}\)]{\includegraphics[width=\linewidth,keepaspectratio]{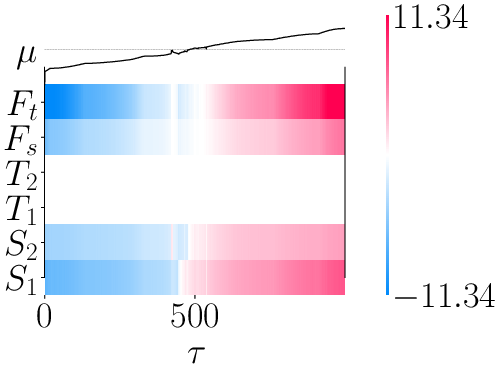}}%
\tcbitem\subfloat[DeepLIFT, PI, \(\sigma = 1^{-4}\)]{\includegraphics[width=\linewidth,keepaspectratio]{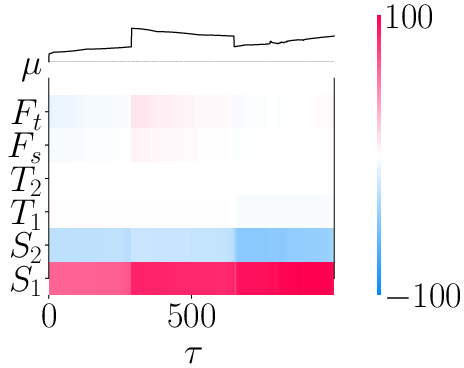}}%
\tcbitem\subfloat[SHAP, AR, \(\sigma = 1^{-2}\)]{\includegraphics[width=\linewidth,keepaspectratio]{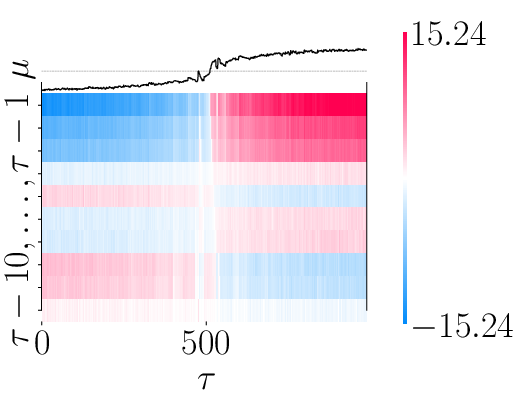}}%
\tcbitem\subfloat[DeepLIFT, AR, \(\sigma = 1^{-2}\)]{\includegraphics[width=\linewidth,keepaspectratio]{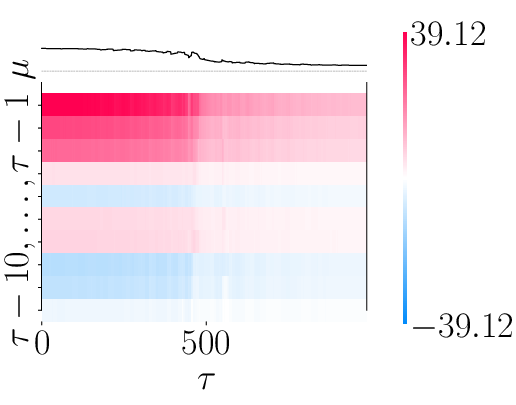}}%
\tcbitem\subfloat[SHAP, AR, \(\sigma = 1^{-3}\)]{\includegraphics[width=\linewidth,keepaspectratio]{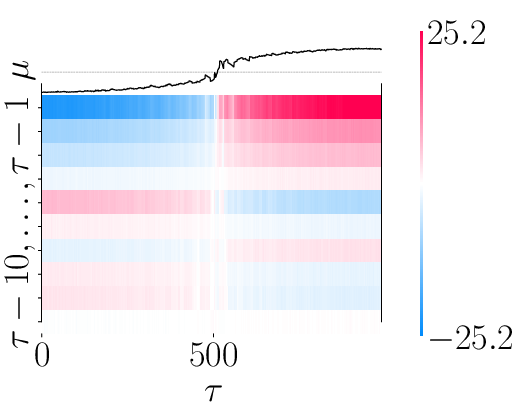}}%
\tcbitem\subfloat[DeepLIFT, AR, \(\sigma = 1^{-3}\)]{\includegraphics[width=\linewidth,keepaspectratio]{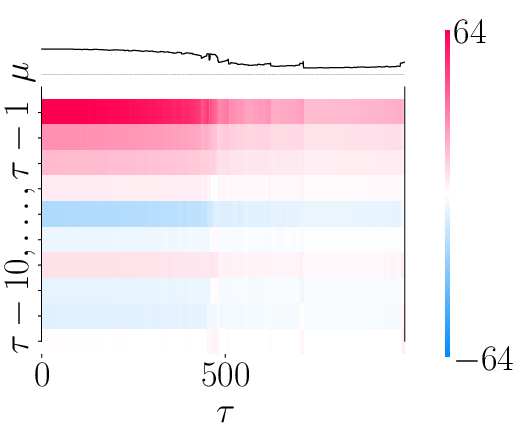}}%
\tcbitem\subfloat[SHAP, AR, \(\sigma = 1^{-6}\)]{\includegraphics[width=\linewidth,keepaspectratio]{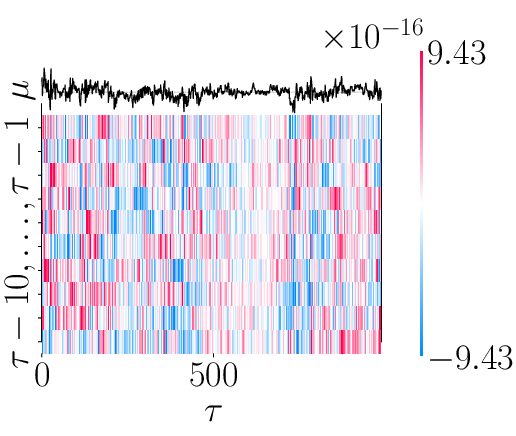}}%
\tcbitem\subfloat[DeepLIFT, AR, \(\sigma = 1^{-6}\)]{\includegraphics[width=\linewidth,keepaspectratio]{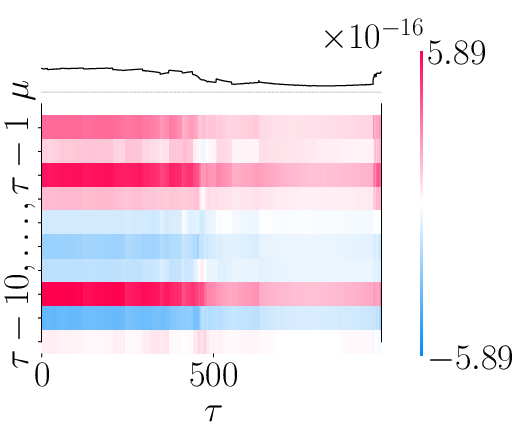}}%
\tcbitem\subfloat[SHAP, AR, \(\sigma = 1^{-5}\)]{\includegraphics[width=\linewidth,keepaspectratio]{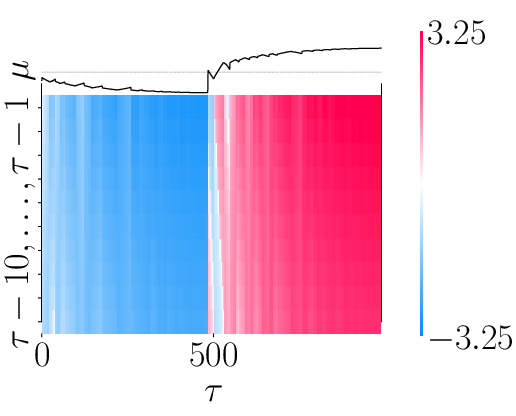}}%
\tcbitem\subfloat[DeepLIFT, AR, \(\sigma = 1^{-5}\)]{\includegraphics[width=\linewidth,keepaspectratio]{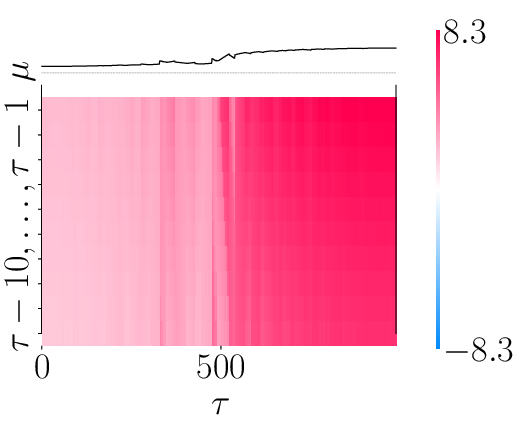}}%
\tcbitem\subfloat[SHAP, AR, \(\sigma = 1^{-4}\)]{\includegraphics[width=\linewidth,keepaspectratio]{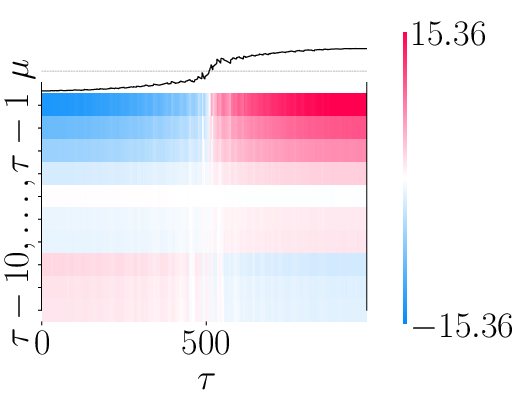}}%
\tcbitem\subfloat[DeepLIFT, AR, \(\sigma = 1^{-4}\)]{\includegraphics[width=\linewidth,keepaspectratio]{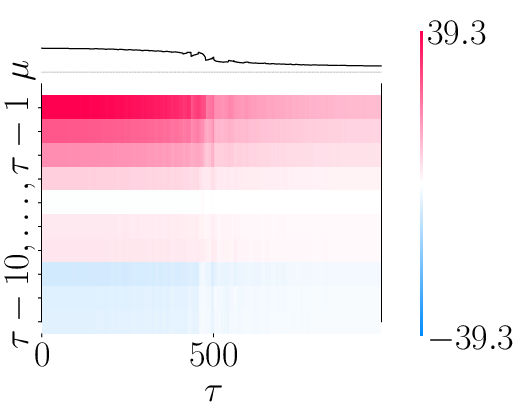}}%
\end{tcbitemize}%
\caption{BNN attributions under varying \(\sigma\) for \(\mathcal{F}_6\).}%
\label{fig:xai-6}%
\end{figure}

%

\end{document}